\documentclass{article}

\PassOptionsToPackage{numbers, compress}{natbib}
\usepackage{natbib}
\setcitestyle{numbers, square, comma, compress}

\usepackage{microtype}
\usepackage{graphicx}
\usepackage{subfigure}
\usepackage{booktabs} 

\usepackage[nohyperref,accepted]{icml2025}

\usepackage{amsmath}
\usepackage{amssymb}
\usepackage{mathtools}
\usepackage{amsthm}

\usepackage{amsfonts}       
\usepackage{nicefrac}       
\usepackage{microtype}      

\usepackage{color}
\usepackage{arydshln}
\usepackage{bm}
\usepackage{array}
\usepackage{comment}
\usepackage{subfigure}
\usepackage{caption}
\usepackage{colortbl} 
\usepackage{multirow}
\usepackage{makecell}
\usepackage{wrapfig}
\usepackage{setspace}
\usepackage{diagbox} 
\usepackage{longtable}

\usepackage{arydshln} 
\usepackage{enumitem}

\usepackage{pifont}

\usepackage{tabularray}
\usepackage{paralist}
\usepackage{makecell}
\usepackage{colortbl}
\usepackage{etoolbox}
\usepackage{cellspace}
\usepackage{wrapfig}
\usepackage{xspace}

\usepackage{tcolorbox}
\usepackage{adjustbox}
\usepackage{rotating}

\usepackage{blindtext}
\usepackage{titlesec}
\usepackage{arydshln}
\usepackage[textsize=tiny]{todonotes}

\usepackage{etoc}
\usepackage{titletoc}
\usepackage{xcolor}         
\definecolor{mydarkblue}{rgb}{0,0.08,0.45}
\usepackage[colorlinks=true,citecolor=mydarkblue]{hyperref}

\newcommand\DoToC{%
  \startcontents
  \printcontents{l}{1}{\noindent \textbf{\Large{Table of Contents}}\vskip3pt\vskip5pt}
  \vskip3pt\vskip5pt
}

\tcbuselibrary{breakable}

\usepackage[capitalize,noabbrev]{cleveref}
\theoremstyle{plain}

\theoremstyle{definition}

\theoremstyle{remark}

\newcommand{\leveliilogo}{\raisebox{-10pt}{\includegraphics[width=1.6em]{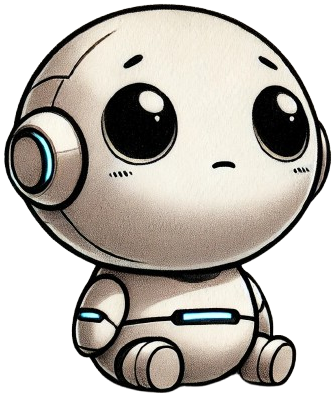}}\xspace\xspace}

\newcommand{\leveliiilogo}{\raisebox{-16pt}{\includegraphics[width=1.6em]{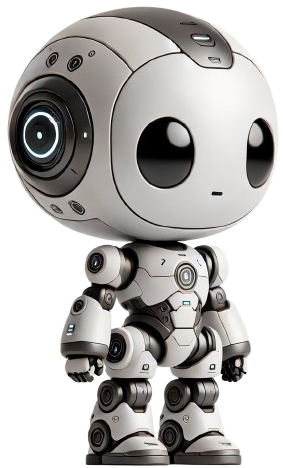}}\xspace\xspace}

\newcommand{\levelivlogo}{\raisebox{-21pt}{\includegraphics[width=2.1em]{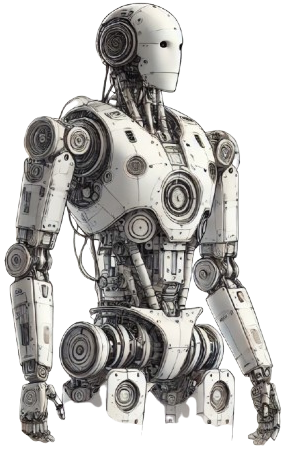}}\xspace\xspace}

\newcommand{\levelvlogo}{\raisebox{-16pt}{\includegraphics[width=2.7em]{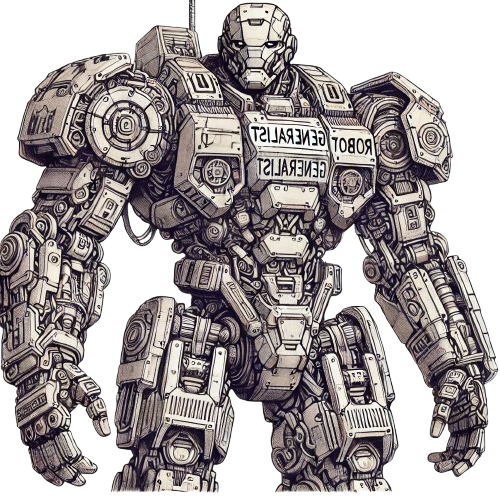}}\xspace\xspace}

\newcommand{\championlogo}{\raisebox{-0pt}{\includegraphics[width=0.9em]{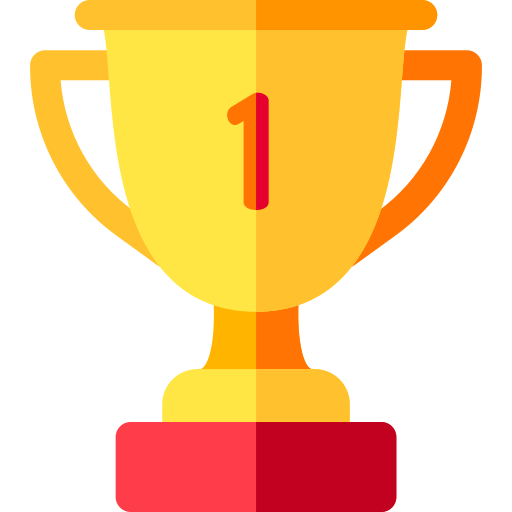}}\xspace\xspace}

\newcommand{\silverlogo}{\raisebox{-0pt}{\includegraphics[width=0.9em]{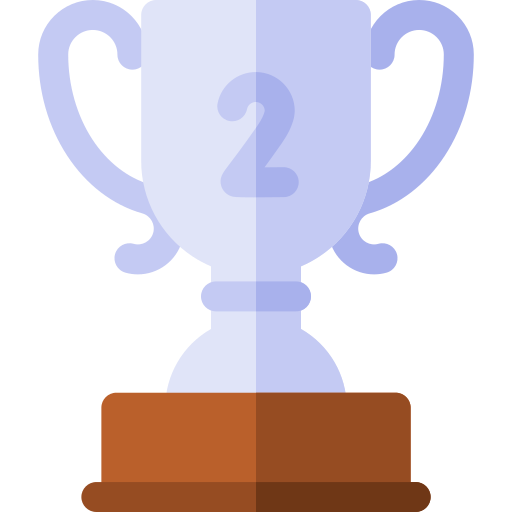}}\xspace\xspace}

\newcommand{\bronzelogo}{\raisebox{-0pt}{\includegraphics[width=0.9em]{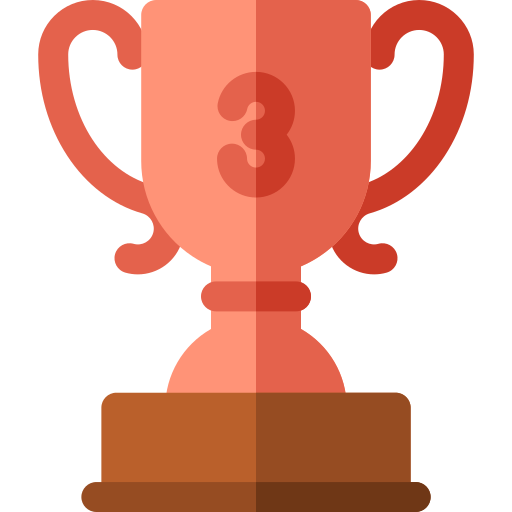}}\xspace\xspace}

\newcommand{\languagelogo}{\raisebox{-0pt}{\includegraphics[width=0.9em]{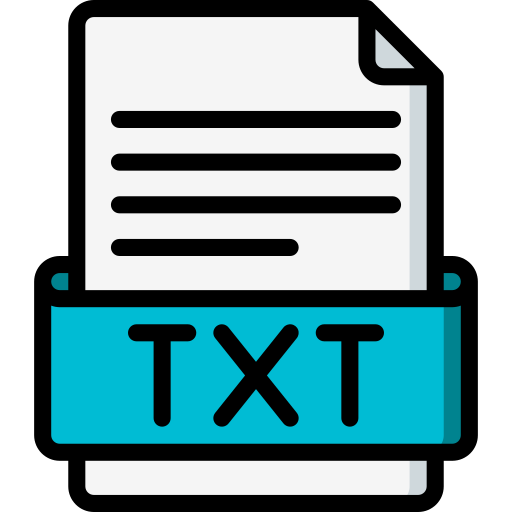}}\xspace\xspace}

\newcommand{\imagelogo}{\raisebox{-0pt}{\includegraphics[width=0.9em]{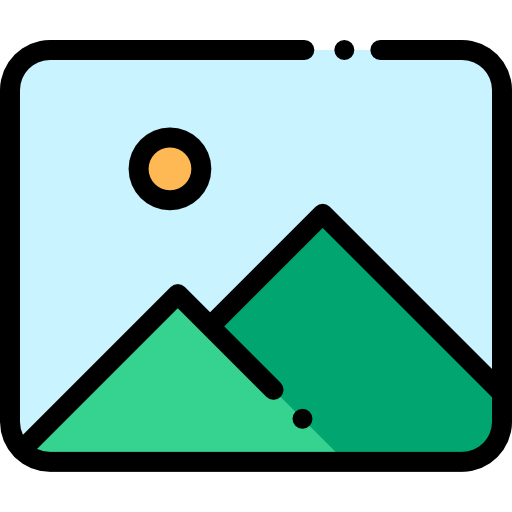}}\xspace\xspace}

\newcommand{\videologo}{\raisebox{-0pt}{\includegraphics[width=0.9em]{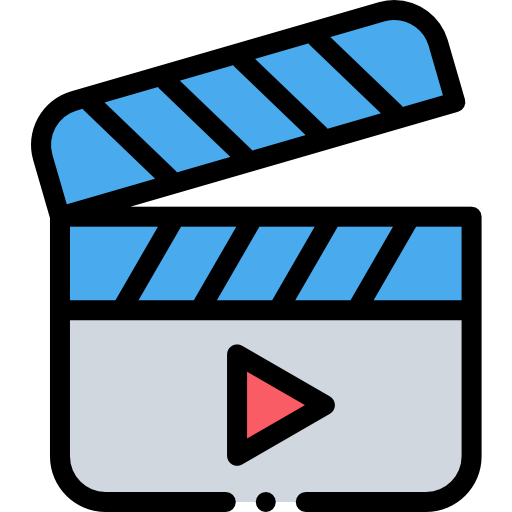}}\xspace\xspace}

\newcommand{\audiologo}{\raisebox{-0pt}{\includegraphics[width=0.9em]{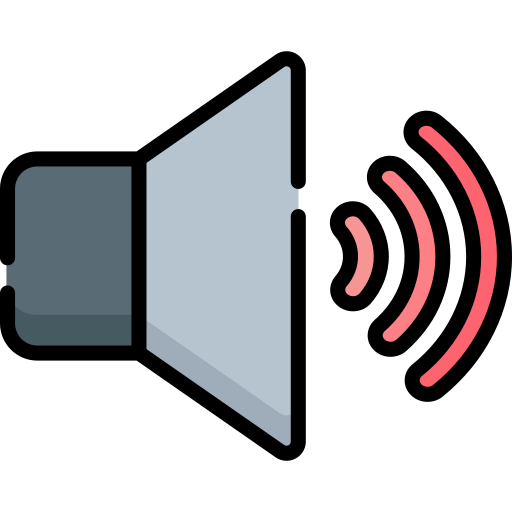}}\xspace\xspace}

\newcommand{\dlogo}{\raisebox{-0pt}{\includegraphics[width=0.9em]{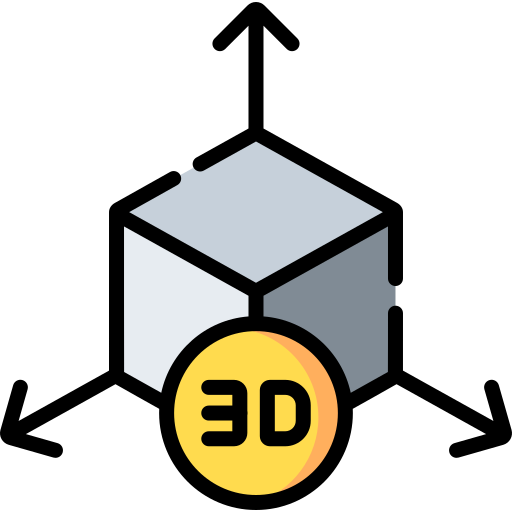}}\xspace\xspace}

\definecolor{placeholdertxt}{RGB}{255,209,227}

\definecolor{myblue}{RGB}{52,218,247}
\definecolor{myred}{RGB}{255,90,90}
\definecolor{mypink}{RGB}{239,43,159}
\definecolor{myupdate}{RGB}{254,243,222}
\definecolor{myfrozen}{RGB}{237,255,255}
\definecolor{ired}{RGB}{229,72,72}
\definecolor{igreen}{RGB}{80,219,144}
\definecolor{nmblue}{RGB}{216,234,247}

\definecolor{linkred}{RGB}{255,0,49}

\definecolor{textred}{RGB}{255,242,234}
\definecolor{imageblue}{RGB}{224,250,255}
\definecolor{videogreen}{RGB}{214,250,232}

\definecolor{mheadertgreen}{RGB}{76,175,185}
\definecolor{lightgreen}{RGB}{237,246,247}

\definecolor{lightgray}{RGB}{242,242,242}
\definecolor{lightgraygrey}{RGB}{240,240,240}
\definecolor{deepgray}{RGB}{196,196,196}

\definecolor{mydeepblue}{RGB}{188,252,216}
\definecolor{mypinky}{RGB}{255,149,211}

\definecolor{bg-tb-heavey-nlp}{RGB}{214,255,232}
\definecolor{bg-tb-light-nlp}{RGB}{237,252,247}

\definecolor{bg-tb-heavey-vision}{RGB}{255,242,227}
\definecolor{bg-tb-light-vision}{RGB}{255,250,247}

\definecolor{bg-tb-heavey-video}{RGB}{219,247,255}
\definecolor{bg-tb-light-video}{RGB}{240,252,252}

\definecolor{bg-tb-heavey-audio}{RGB}{240,219,255}
\definecolor{bg-tb-light-audio}{RGB}{250,245,252}

\definecolor{bg-tb-heavey-3D}{RGB}{255,229,229}
\definecolor{bg-tb-light-3D}{RGB}{255,245,245}

\definecolor{greenCom}{RGB}{3,214,131}
\definecolor{blueGen}{RGB}{0,160,216}
\definecolor{redNLP}{RGB}{250,59,59}

\icmltitlerunning{On Path to Multimodal Generalist: General-Level and General-Bench}

\begin{document}

\onecolumn

\icmltitle{\emph{On Path to Multimodal Generalist}: General-Level and General-Bench}

\icmlsetsymbol{equal}{*}

\vspace{-4mm}

\begin{icmlauthorlist}
\icmlauthor{Hao Fei}{equal,nus}
\icmlauthor{Yuan Zhou}{equal,ntu}
\icmlauthor{Juncheng Li}{equal,zju}
\icmlauthor{Xiangtai Li}{equal,ntu}
\icmlauthor{Qingshan Xu}{equal,ntu}
\icmlauthor{Bobo Li}{equal,nus}
\icmlauthor{Shengqiong Wu}{equal,nus}
\icmlauthor{Yaoting Wang}{kaust}
\icmlauthor{Junbao Zhou}{ntu}
\icmlauthor{Jiahao Meng}{pku}
\icmlauthor{Qingyu Shi}{pku}
\icmlauthor{Zhiyuan Zhou}{hfut}
\icmlauthor{Liangtao Shi}{hfut}
\icmlauthor{Minghe Gao}{zju}
\icmlauthor{Daoan Zhang}{ur}
\icmlauthor{Zhiqi Ge}{zju}
\icmlauthor{Weiming Wu}{nju}
\icmlauthor{Siliang Tang}{zju}
\icmlauthor{Kaihang Pan}{zju}
\icmlauthor{Yaobo Ye}{zju}
\icmlauthor{Haobo Yuan}{ntu}
\icmlauthor{Tao Zhang}{whu}
\icmlauthor{Tianjie Ju}{sjtu}
\icmlauthor{Zixiang Meng}{whu}
\icmlauthor{Shilin Xu}{pku}
\icmlauthor{Liyu Jia}{ntu}
\icmlauthor{Wentao Hu}{ntu}
\icmlauthor{Meng Luo}{nus}
\icmlauthor{Jiebo Luo}{ur}
\icmlauthor{Tat-Seng Chua}{nus}
\icmlauthor{Shuicheng Yan}{nus}
\icmlauthor{Hanwang Zhang}{ntu}
\end{icmlauthorlist}

\icmlaffiliation{nus}{NUS}
\icmlaffiliation{ntu}{NTU}
\icmlaffiliation{zju}{ZJU}
\icmlaffiliation{nju}{NJU}
\icmlaffiliation{kaust}{KAUST}
\icmlaffiliation{pku}{PKU}
\icmlaffiliation{whu}{WHU}
\icmlaffiliation{sjtu}{SJTU}
\icmlaffiliation{hfut}{HFUT}
\icmlaffiliation{ur}{UR}

\icmlProjectLeader{Hao Fei}{haofei37@nus.edu.sg}

\icmlcorrespondingauthor{Shuicheng Yan}{yansc@nus.edu.sg}
\icmlcorrespondingauthor{Hanwang Zhang}{hanwangzhang@ntu.edu.sg}

\vspace{2mm}
\begin{center}    
    \large{\color{mypink}{\textbf{Project Page:} 
    \url{https://generalist.top}}}
    
    \large{\color{mypink}{\textbf{Leaderboard:} 
    \url{https://generalist.top/leaderboard}}}
    
    \large{\color{mypink}{\textbf{Benchmark:} 
    \url{https://huggingface.co/General-Level}}}    
\end{center}
\vspace{2mm}

{%
\renewcommand\twocolumn[1][]{#1}%
\begin{center}
\includegraphics[width=0.99\linewidth]{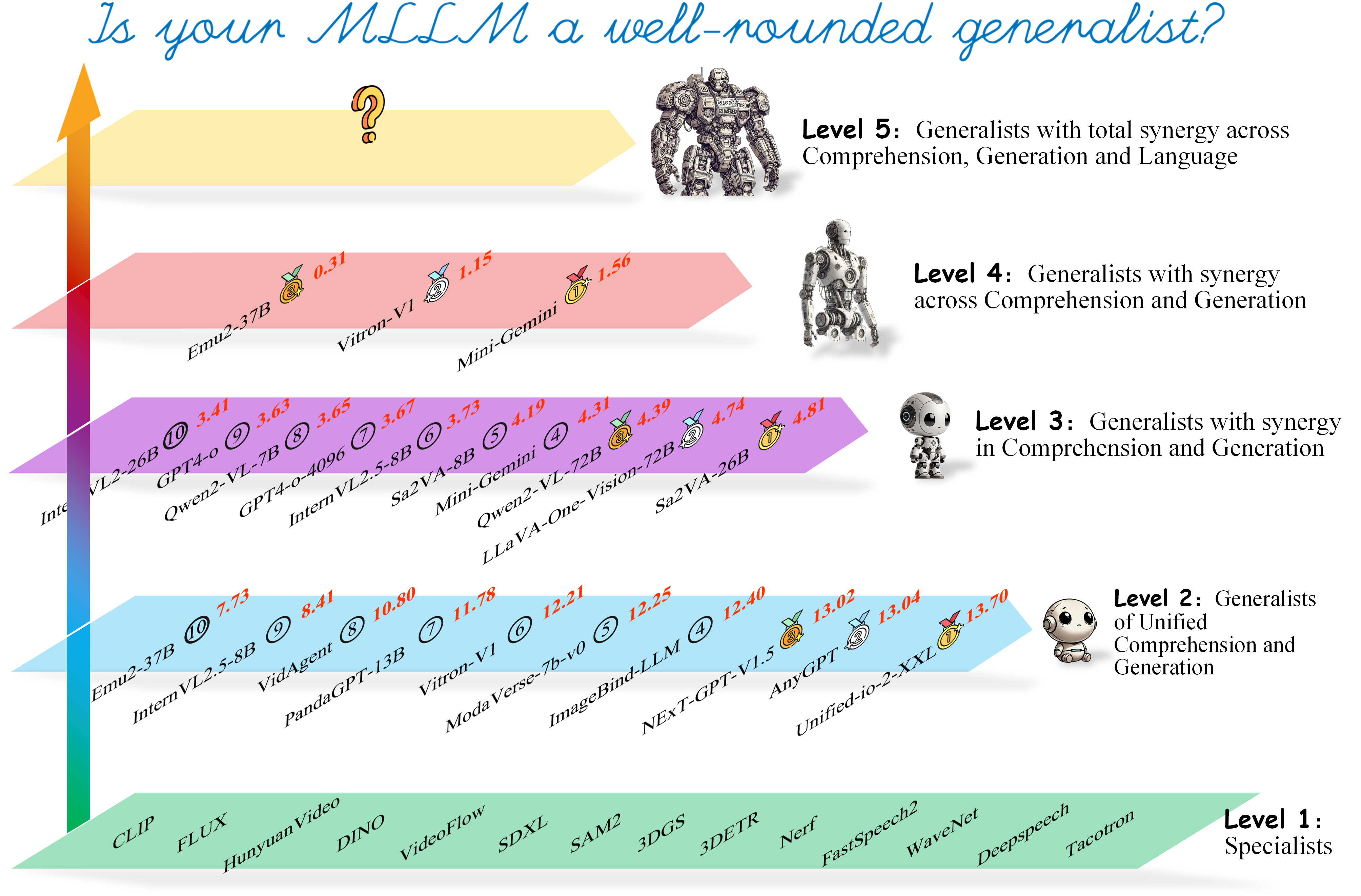}
\vspace{-1mm}
\captionof{figure}{
Leaderboard of multimodal generalists over \texttt{\textbf{General-Level}} (only top-performing ones shown here).
}
\label{fig:leaderboard-intro}
\vspace{4mm}
\end{center}%
}

\printAffiliationsAndNotice{\icmlEqualContribution} 

\begin{abstract}
The Multimodal Large Language Model (MLLM) is currently experiencing rapid growth, driven by the advanced capabilities of language-based LLMs. 
Unlike their specialist predecessors, existing MLLMs are evolving towards a Multimodal Generalist paradigm. 
Initially limited to understanding multiple modalities, these models have advanced to not only comprehend but also generate across modalities. 
Their capabilities have expanded from coarse-grained to fine-grained multimodal understanding and from supporting singular modalities to accommodating a wide array of or even arbitrary modalities. 
To assess the capabilities of various MLLMs, a diverse array of benchmark test sets has been proposed. 
This leads to a critical question: \emph{Can we simply assume that higher performance across tasks indicates a stronger MLLM capability, bringing us closer to human-level AI?}

\vspace{0.2mm}
We argue that the answer is not as straightforward as it seems. 
In this project, we introduce an evaluation framework to delineate the capabilities and behaviors of current multimodal generalists. 
This framework, named \texttt{\textbf{General-Level}}, establishes 5-scale levels of MLLM performance and generality, offering a methodology to compare MLLMs and gauge the progress of existing systems towards more robust multimodal generalists and, ultimately, towards AGI (Artificial General Intelligence). 
Central to our framework is the use of \textbf{Synergy} as the evaluative criterion, categorizing capabilities based on whether MLLMs preserve synergy across comprehension and generation, as well as across multimodal interactions.
To evaluate the comprehensive abilities of various generalists, we present a massive multimodal benchmark, \texttt{\textbf{General-Bench}}, which encompasses a broader spectrum of skills, modalities, formats, and capabilities, including over 700 tasks and 325,800 instances. 
The evaluation results that involve over 100 existing state-of-the-art MLLMs uncover the capability rankings of generalists, highlighting the challenges in reaching genuine AI. 
We expect this project to pave the way for future research on next-generation multimodal foundation models, providing a robust infrastructure to accelerate the realization of AGI.
\end{abstract}

\vspace{4mm}



{\color{red} \DoToC}

\clearpage


\section{Introduction}
\label{Introduction}

Large Language Models (LLMs, e.g., ChatGPT \cite{chatgpt} and LLaMA \cite{abs-2302-13971}) have revolutionized the NLP field by serving as generalists addressing a vast spectrum of NLP tasks. 
This breadth of capability has edged humans ever closer to the realization of Artificial General Intelligence (AGI). 
Yet, human intelligence inherently operates across multiple modalities, not solely through language. 
This observation has spurred the development of multimodal LLMs \cite{AlayracDLMBHLMM22,0008LSH23,abs-2304-08485,gpt4}, i.e., multimodal generalists, which are rapidly gaining traction and evolving towards AGI. 
The recent progress in MLLMs is marked by significant advancements.
For example, the initial multimodal agents where LLMs serve as mere task schedulers, later have evolved into joint foundation MLLMs \cite{abs-2304-10592,abs-2304-08485,abs-2305-11000,gpt4,wu2023next,chen2024ll3da,sun2024generative}.
Also, MLLMs have progressed from understanding only multimodal signals to both comprehending and generating multimodal content, even editing capabilities \cite{wang2023gpt4video,munasinghe2023pg,zhang2024omg,fei2024vitron}.
Further, these models have advanced from coarse-grained modal understanding to fine-grained multimodal comprehension, such as pixel-level visual modeling \cite{ren2023pixellm,yuan2023osprey,rasheed2023glamm}.
More significantly, MLLMs that initially support only singleton non-textual modalities have now facilitated the understanding and generation of signals across various modalities, even simultaneously accommodating any modality \cite{wu2023next,zhan2024anygpt,lu2024unified}.

Accordingly, the community has introduced various benchmarks to evaluate those MLLMs \cite{wu2023q,xia2024mmie,yue2024mmmu,meng2024mmiu,liu2025mmbench,li2024seed,ying2024mmt,li2024survey}.
The prevailing evaluation mindset might yet be largely outdated,
simplistically assuming that superior performance across tasks presents a stronger generalist capability \cite{xu2023lvlm,yu2023mm,fu2024MME,chen2024mega}, and then being closer to AGI. 
We contend this perspective overly simplifies the implication inherent in real multimodal generalization. 
Theoretically, it's effortless to assemble a ``super agent'' from all singleton state-of-the-art (SoTA) specialists to achieve the above goal, while such a simplistic integration would never suffice to realize genuine AGI. 
We argue that the key to advancing towards AGI lies in the \emph{\textbf{synergy}} effect—\textcolor{blue}{a capability that enables knowledge learned in one modality or task to generalize and enhance mastery in other modalities or tasks, fostering mutual improvement across different modalities and tasks through interconnected learning}.\footnote{Synergy, in essence, can be understood as a form of generalization ability.} 
As illustrated in Figure \ref{fig:intro}, most current MLLMs predominantly build on the language intelligence of LLMs to simulate the indirect intelligence of multimodality, which is merely extending language intelligence to aid multimodal understanding.
While LLMs (e.g., ChatGPT) have already demonstrated such synergy in NLP, reflecting language intelligence, unfortunately, the vast majority of MLLMs do not really achieve it across modalities and tasks.

In this project, we introduce a sophisticated evaluation framework, \texttt{\textbf{General-Level}}, for more accurately positioning and assessing the capabilities of current MLLM generalists, charting a path toward authentic multimodal AGI.
Drawing inspiration from the tiered classification mechanism in the automotive industry for autonomous vehicles \cite{yurtsever2020survey}, \texttt{{General-Level}} defines five principal levels of model performance and generality. 
Central to the framework is the synergy ability as the evaluative criterion, categorizing capabilities based on whether generalists preserve synergy in and across multimodal comprehension and generation, as well as cross-modal interactions.
From the lowest to the highest level, the scope of synergy ability required progressively escalates from single tasks or modalities to total synergy. 
As a generalist strives to advance to a higher level, it must demonstrate significant enhancements in its synergy capabilities, during which the difficulty of progression is also inherently increasing.

\begin{figure*}[!t]
\centering
\includegraphics[width=0.99\linewidth]{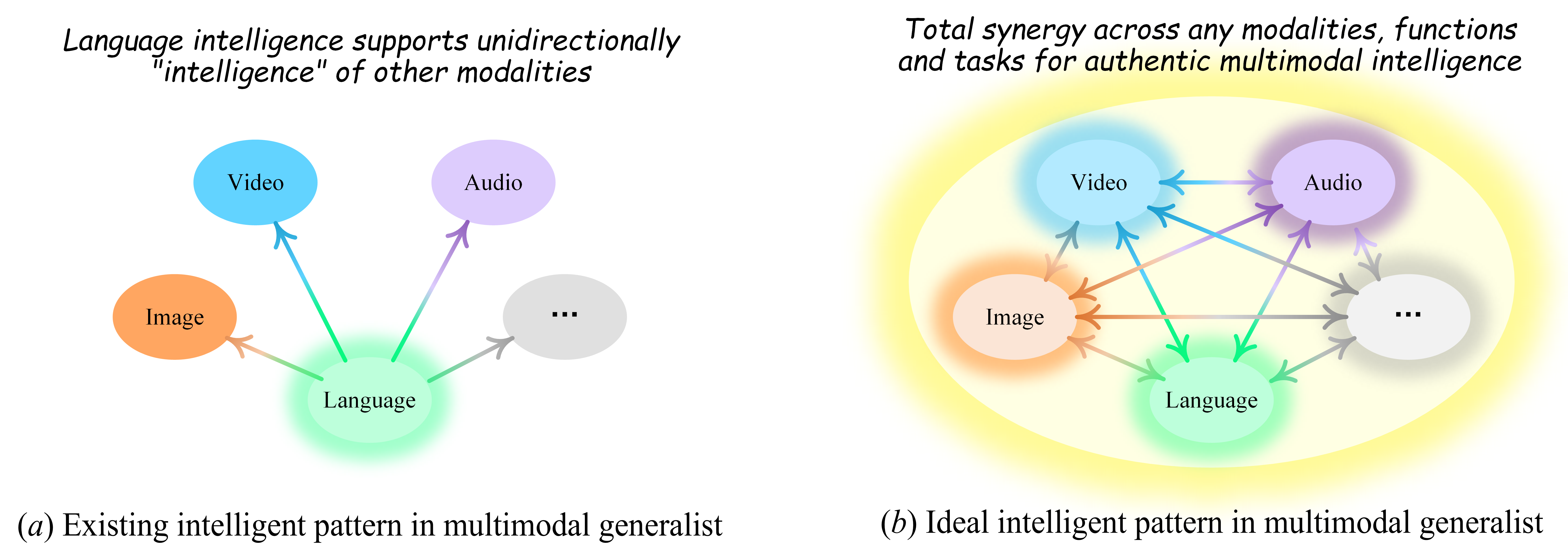}
\vspace{-2mm}
\caption{
The ``\emph{intelligence}'' in most existing multimodal generalists (i.e., MLLMs) hinges on language intelligence (i.e., from LLMs) \textbf{(a)}, whereas the ideal intelligence mode should be maintaining synergy across all modalities and tasks \textbf{(b)}.}
\label{fig:intro}
\vspace{-3mm}
\end{figure*}

To effectively evaluate within the \texttt{{General-Level}} framework, a suitable benchmark is essential. 
While there are numerous MLLM evaluation benchmarks, e.g., LVLM-eHub \cite{xu2023lvlm}, MME \cite{fu2024MME}, MMMU \cite{yue2024mmmu}, SEED-Bench \cite{li2024seed}, MMT-Bench \cite{ying2024mmt}, and MEGA-Bench \cite{chen2024mega}, they might have certain limitations that render them inadequate for our needs.
Firstly, existing benchmarks often convert all tasks into a uniform multiple-choice QA format \cite{fu2024MME,ying2024mmt}, simplifying the evaluation process but consequently restricting assessments to only the models' multimodal comprehension capabilities. 
However, a true multimodal generalist should support not only comprehension, but also possess capabilities in multimodal generation, editing, and beyond.
Second, the majority of current benchmarks \cite{wu2023q,liu2025mmbench,li2024seed} predominantly focus on the image modality and overlook other crucial modalities such as video, audio, even 3D and beyond, which are vital for a robust multimodal generalist. 
Third, these benchmarks are typically limited to coarse-grained multimodal understanding \cite{xu2023lvlm,yu2023mm,fu2024MME} and fail to adequately assess finer-grained ones, which actually lag far behind the current advancements in MLLMs, i.e., supporting pixel-level image understanding and generation \cite{fei2024vitron,zhang2024omg}.
In response to these challenges, we propose \texttt{\textbf{General-Bench}}, which is a massive multimodal evaluation benchmark, spanning from various modalities (e.g., image, video, audio, 3D, language, and beyond) in diverse native formats, covering a wide range of tasks that thoroughly assess the full capabilities of a multimodal generalist.

Our evaluation of over 100 existing top-performing LLM/MLLM systems has uncovered critical insights into their capabilities and rankings as multimodal generalists. 
The most notable finding is that most MLLMs lack the cross-task or cross-modal synergy ability required for higher-level classifications, with even advanced models like GPT-4V and GPT-4o not achieving top ranks.
This highlights a considerable gap in achieving the goals of multimodal generalists. 
Also, the majority of existing MLLMs manage only a few basic multimodal tasks and skills, which negatively affects their scoring. 
Most critically, no model has yet demonstrated the ability to enhance language intelligence through non-language modalities, underscoring the substantial challenges in the pursuit of genuine AGI.

\textbf{Contributions:}
1) We introduce a tiered classification system called \texttt{General-Level} for multimodal generalists, establishing a rigorous standard or norm that can guide future MLLM research.
2) We contribute a new evaluation benchmark (\texttt{General-Bench}) that provides the most comprehensive coverage of modalities and tasks available to date.
We hope this project will serve as an infrastructure to facilitate the development of next-generation multimodal foundation models in achieving more capable and general-purpose multimodal intelligence.

\vspace{-1mm}
\section{Background and Related Work}

\vspace{-1mm}
More and more tend to recognize that LLMs have unlocked the potential of language intelligence, bringing unprecedented hope to achieve AGI.
Essentially, an LLM serves as a generalist capable of tackling nearly all downstream NLP tasks.
LLMs have subsequently evolved in an effort to extend this intelligence across various other modalities, i.e., MLLMs \cite{bai2023qwen,zhang2023internlm,jin2023unified,li2024mini,fei2024multimodal,fei2024enhancing}.
Unlike the past `smaller' specialists \cite{van2016wavenet,radford2021learning,rombach2022high,liu2023grounding}, MLLMs represent an important advancement of unification to handle all modalities and tasks with one foundation model, i.e., multimodal generalists.
Naturally, empowering a multimodal generalist with strong multimodal intelligence capabilities is an essential pathway toward realizing AGI.

Technically, the vast majority of existing MLLMs have frameworks that are anchored by an LLM to serve as the core for reasoning and decision-making.
By integrating various well-trained modules of different modalities or tasks (typically existing specialists, e.g., CLIP \cite{radford2021learning} and Stable Diffusion \cite{rombach2022high}), MLLMs are facilitated with the comprehension and even generation of diverse modalities.
Representative MLLMs include Blip2 \cite{0008LSH23}, LLAVA \cite{abs-2304-08485}, MiniGPT-4 \cite{abs-2304-10592}, Flamingo \cite{AlayracDLMBHLMM22}, and NExT-GPT \cite{wu2023next}, among others.
However, such an architectural setup merely simulates `pseudo' multimodal intelligence, as it still fundamentally relies on the language intelligence of LLMs without genuine non-language modality intelligence.
As emphasized earlier, a capable generalist must possess synergy capabilities across all modalities and tasks, akin to how an LLM (e.g., ChatGPT) generalizes well to unseen NLP tasks, despite not being exposed to all tasks during its training.
While these current multimodal generalists can deliver strong performances on multimodal benchmarks, sometimes even on par with SoTA specialists, they do not fundamentally achieve true synergy.

Consequently, this paper positions synergy as the central criterion for evaluating multimodal generalists on their journey toward AGI.
Current evaluation methods \cite{li2024survey} for MLLMs still adhere to the traditional approach used for specialists, simply comparing the MLLM performance on multimodal tasks, assuming that higher scores indicate greater strength and closer proximity to AGI.
Going beyond that, we propose a new evaluation framework—not only do we compare whether models support various modalities and tasks and their performance, but we also rank them based on the synergy capabilities of multimodal generalists.
Meanwhile, we significantly expand the scope of current MLLM benchmark datasets in terms of modality and task coverages, as well as task formats, contributing to the most comprehensive benchmark dataset to date in the community.

\vspace{-1mm}
\section{\emph{General-Level}: A 5-Level Taxonomy of Multimodal Generalists}

\subsection{Preliminary}

\subsubsection{Observations and Principles}

\paragraph{Observation-1: Multimodal Comprehension vs. Simultaneous Multimodal Comprehension and Generation.}

Initially, MLLMs are capable only of interpreting multimodal signals, meaning their responses are limited to textual outputs based on user-provided multimodal inputs. 
However, an MLLM that only offers multimodal comprehension operates at the most basic and rudimentary level. 
More advanced MLLMs have since emerged, equipped with not only multimodal comprehension but also the ability to generate and even edit content across various modalities. 
It is widely believed that the more advanced a multimodal generalist is, the more it should encompass advanced functionalities, encompassing both comprehension and generation.

\vspace{-1mm}
\paragraph{Observation-2: Covering Broader Modalities.}

Being a multimodal generalist requires the ability to extensively support and handle a wide range of modal data, including, but not limited to, text, images, videos, audio, and even 3D. 
The extent of modal support is indicative of the breadth of an AI system’s capabilities. 
Initially, MLLMs could manage only a singleton non-linguistic modality, e.g., images, videos, or audio signals. 
To date, these models have evolved to simultaneously support multiple non-linguistic modalities—such as combining images with videos, videos with audio, and even any modality in the current most advanced cases.

\vspace{-1mm}
\paragraph{Observation-3: Supporting Various Tasks and Paradigms.}

To qualify as a true multimodal generalist, it must be capable of handling a broad range of tasks with different definitions and requirements. 
The greater the variety of tasks supported, the stronger the generalist's overall versatility. 
For example, early visual MLLMs could only manage coarse-grained image understanding, but recent advancements have enabled them to achieve fine-grained, pixel-level multimodal comprehension, such as pixel-level image/video grounding and editing. 
This advancement necessitates that the model's decoding components should be versatile enough to generate outputs in various task formats, not merely restricted to text. 
These functional heads must handle different task types such as object localization, pixel-level modifications, and multimodal content creation.

\vspace{-1mm}
\paragraph{Observation-4: Multimodal Agent vs. Multimodal Foundation Model.}

Initially, researchers approach multimodal tasks by using LLMs as task schedulers, where an LLM orchestrates the execution of tasks by invoking external tools and modules (often specialists) to handle specific multimodal tasks. 
This setup is referred to as a multimodal agent. 
Subsequently, attention shifted towards building joint MLLMs, where the LLM is tightly integrated with other modules, such as multimodal understanding components (front-end) and multimodal generation components (back-end), through a shared embedding space. 
This setup allows for joint training, where the entire system, including all parameters, can be updated end-to-end. 
While it's theoretically possible to create a `super agent' by combining all singleton SoTA specialists to handle various modalities and tasks, such a straightforward aggregation does not lead to true AGI. 
The complexity of AGI requires deeper integration and generalization across tasks and modalities.

\begin{figure*}[!t]
\centering
\includegraphics[width=0.99\linewidth]{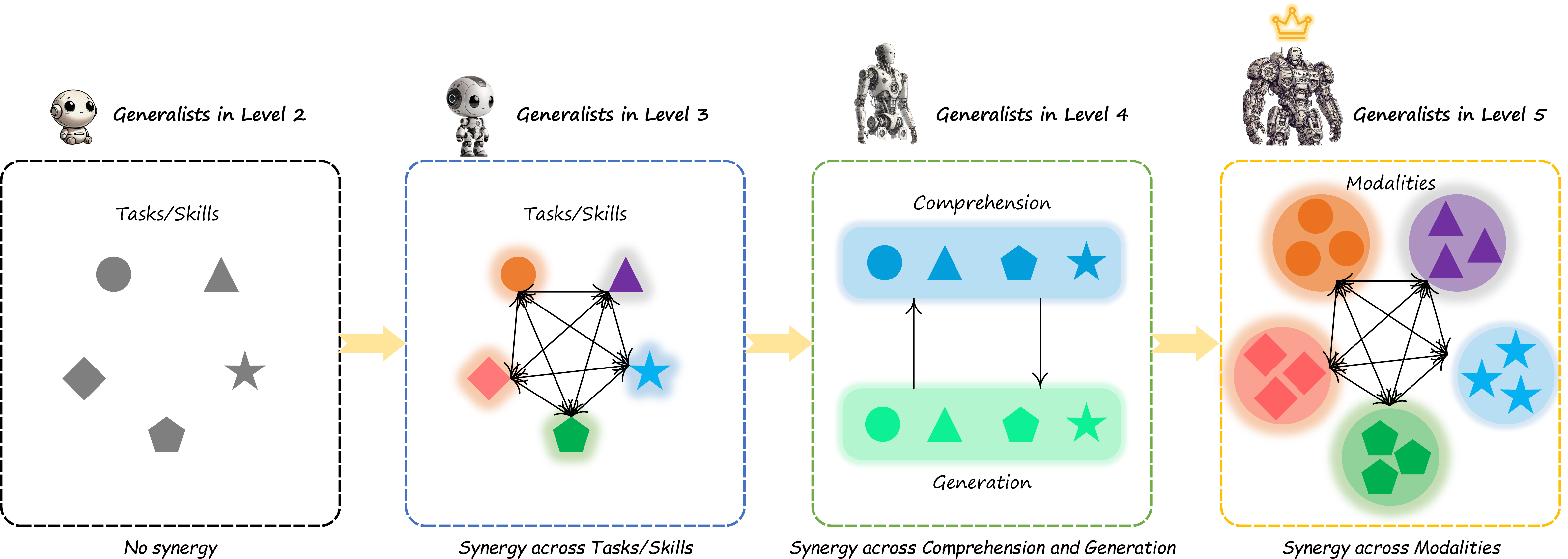}
\vspace{-2mm}
\caption{
A specific illustration on \textbf{synergy} effect.
}
\label{fig:synergy-idea}
\end{figure*}

\subsubsection{Synergy as Core to Multimodal Generalists}

We argue that determining whether a multimodal generalist is stronger cannot be simplistically equated with achieving higher scores on a benchmark or/and supporting as many multimodal tasks as possible compared to other models—a common practice in current MLLM benchmarking and evaluation. 
A simple counterexample can illustrate this point: it could be comparatively easier to construct a `super agent' by integrating all SoTA specialists for various multimodal tasks into a single system. 
Such an agent could achieve top-level performance across all tasks (on par with the strongest individual specialist models) while supporting a wide range of multimodal functionalities.  
However, such agents can be far from the multimodal generalist we expect as a pathway to AGI. 
Such a type of agent lacks inherent multimodal intelligence and capabilities, as it relies on an ensemble of specialized systems rather than embodying true, native multimodal generalization.

Instead, the ideal multimodal generalist (and ultimately AGI) we envision should be a multimodal counterpart of an all-capable OpenAI ChatGPT series. 
Such a model would not only surpass SoTA specialists in task-wise performance across various tasks and modalities but also exhibit exceptional \emph{cross-task}, \emph{cross-comprehension-generation}, and \emph{cross-modality} generalization capabilities. 
In other words, the knowledge learned from certain tasks, skills, and modalities should be transferable to other tasks, skills, and modalities—extrapolating the understanding to effectively engage with other tasks and modalities, and vise versa, creating a synergistic effect where the combined result exceeds the sum of individual contributions, achieving a 1+1$>$2 effect.
ChatGPT on the language side can be a good example: it outperforms SoTA specialists in unseen tasks without having undergone specific training for those tasks.
This generalizability is what we claim as the \textbf{synergy effect}.

\begin{figure*}[!t]
\centering
\includegraphics[width=0.9\linewidth]{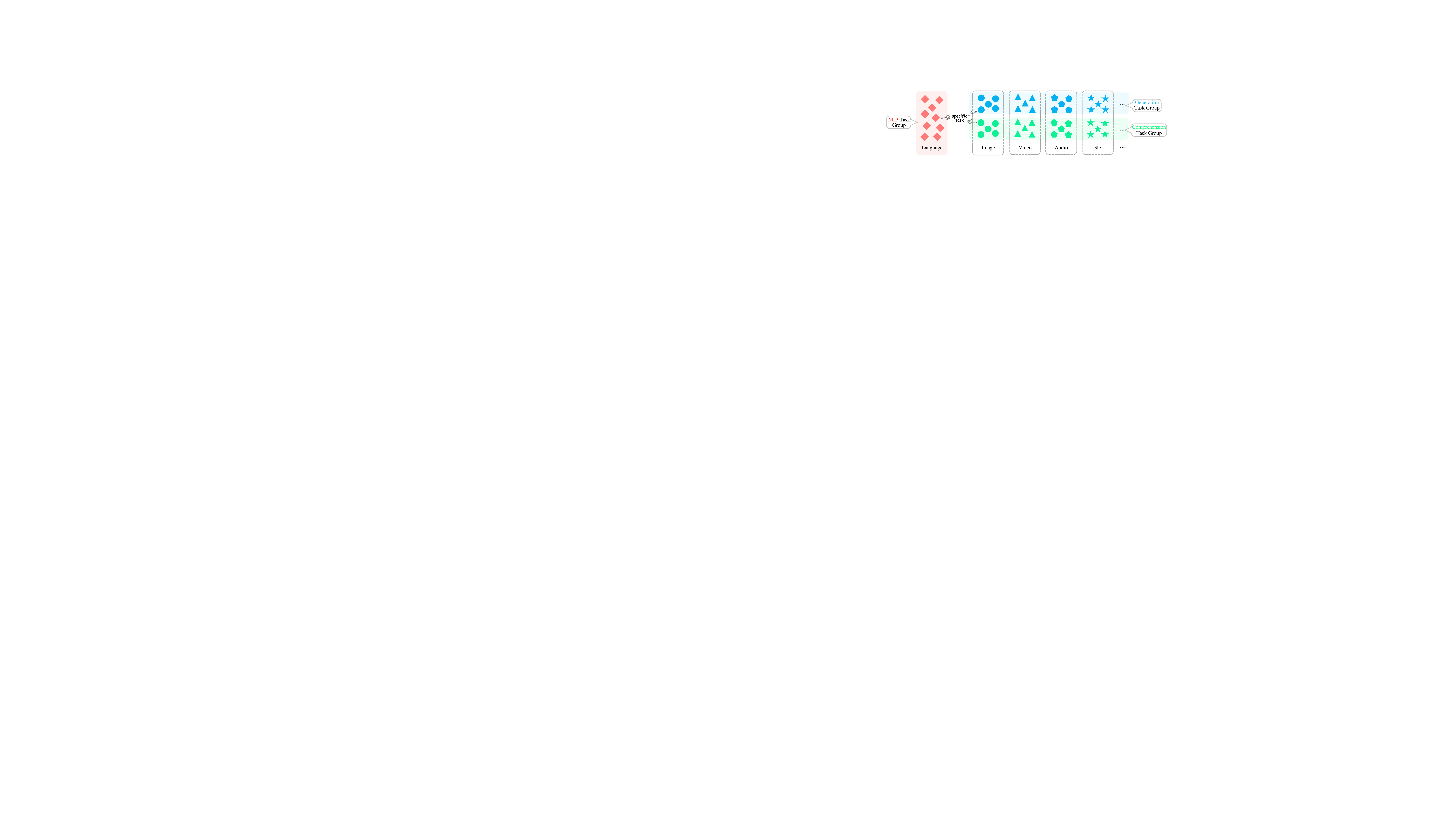}
\vspace{-2mm}
\caption{We categorize tasks of various modalities into \textcolor{greenCom}{Comprehension} group, \textcolor{blueGen}{Generation} group and \textcolor{redNLP}{NLP} group. 
Each colored stylish symbol represents a specific task of a certain modality.
}
\label{fig:level-task}
\vspace{-4mm}
\end{figure*}

\vspace{-2mm}
\subsection{Defining Levels Centered on Synergy}
\label{Definition of Levels}

\vspace{-1mm}
Based on the above principles, we introduce a 5-level taxonomy of multimodal generalists, \texttt{General-Level}.
General-Level framework evaluates generalists based on the levels and strengths of the synergy they preserve. 
Specifically, we define three levels and scopes of synergy, ranked from low to high: `task-task', `comprehension-generation', and `modality-modality', as illustrated in Figure~\ref{fig:synergy-idea}. 
Achieving these levels of synergy becomes progressively more challenging, corresponding to higher degrees of general intelligence.
Assume we have a benchmark of various modalities and tasks, where we can categorize tasks under these modalities into the Comprehension group and the Generation group, as well as the language (i.e., NLP) group, as illustrated in Figure \ref{fig:level-task}.
Now, we can define the scoring specification of \texttt{General-Level} as in Table \ref{tab:levels}.

{
\begin{table*}[!h]
\centering
\fontsize{8.5}{8}\selectfont
\setlength{\tabcolsep}{1.5mm}
\caption{
\texttt{\textbf{General-Level}} framework toward classifying multimodal generalists into \textbf{FIVE} levels based on the synergy abilities models preserve.
We denote the number of tasks within the \textcolor{greenCom}{Comprehension} group by $M$; the number within the \textcolor{blueGen}{Generation} group by $N$; and the number of \textcolor{redNLP}{NLP} tasks by $T$.
}
\label{tab:levels}
\vspace{-2mm}
\begin{tabular}{p{0.17\textwidth}p{0.31\textwidth}p{0.33\textwidth}p{0.12\textwidth}}

\hline
\rowcolor{mheadertgreen}
\vspace{0.3pt} \bf Level \vspace{-\baselineskip}	& \vspace{0.3pt} \bf Definition \vspace{-\baselineskip}	& \vspace{0.3pt} \bf Scoring \vspace{-\baselineskip}	& \vspace{0.3pt} \bf Example \vspace{-\baselineskip} \\[10pt]
\hline

\addlinespace[2pt]
\textbf{Level-1}: \newline  Specialists 
&  Various current models, each fine-tuned on a specific task or dataset of specific modalities, are task-specific players (i.e., SoTA specialists). This includes various learning tasks, such as linguistic/visual recognition, classification, generation, segmentation, grounding, inpainting, and more.		&  For each task in the benchmark ($i$-th task), the current SoTA specialist's score is recorded as:
\begin{equation*}
    \sigma^{sota}_{i}
\end{equation*}
& 
CLIP \cite{0001LXH22}, 
FLUX \cite{flux2023}, 
FastSpeech2 \cite{FastSpeech-0006H0QZZL21}, 
$\cdots$
\\
\hline

\addlinespace[2pt]
\rowcolor{lightgreen} 
\multicolumn{4}{l}{\em $\downarrow$ \textbf{Upgrading Condition}: Supporting as many tasks and functionalities as possible \vspace{-\baselineskip}}\\[2pt]

\hline
\addlinespace[2pt]
\leveliilogo \textbf{Level-2}: \newline  Generalists of Unified \textcolor{greenCom}{Comprehension} and/or \textcolor{blueGen}{Generation}	& Models are task-unified players, e.g., MLLMs, capable of supporting different modalities and tasks. Such MLLMs can integrate various models through existing encoding and decoding technologies to achieve aggregation and unification of various modalities and tasks (such as comprehension and generation tasks).		& The average score between \textcolor{greenCom}{Comprehension} and \textcolor{blueGen}{Generation} tasks (i.e., across all tasks) represents the score at this level. A model that can score non-zero on the data is considered capable of supporting that task. The more supported tasks and the higher the scores, the higher its overall score:
\setlength{\abovedisplayskip}{0pt}
\setlength{\belowdisplayskip}{0pt}
\setlength{\abovedisplayshortskip}{0pt}
\setlength{\belowdisplayshortskip}{0pt}
\begin{equation*}
    S_2 = \frac{1}{2} \left( \frac{1}{M} \sum^{M}_{i=1} \sigma_{i}^{C} + \frac{1}{N} \sum^{N}_{j=1} \sigma_{i}^{G}   \right)
\end{equation*}
& 
Unified-io-2 \cite{lu2024unified}, 
AnyGPT \cite{zhan2024anygpt}, 
NExT-GPT \cite{wu2023next}, 
SEED-LLaMA \cite{ge2023making}, 
GPT-4V \cite{gpt4}, 
$\cdots$
\\
\hline

\addlinespace[2pt]
\rowcolor{lightgreen} \multicolumn{4}{l}{\em$\downarrow$ \textbf{Upgrading Condition}: Generalists achieving as stronger synergy and cross as many tasks as possible \vspace{-\baselineskip}}\\[2pt]

\hline

\addlinespace[2pt]
\leveliiilogo \textbf{Level-3}: \newline  Generalists with \textbf{synergy} in \textcolor{greenCom}{Comprehension} and/or \textcolor{blueGen}{Generation}	& Models are task-unified players, and synergy is in \textcolor{greenCom}{Comprehension} and/or \textcolor{blueGen}{Generation}. MLLMs enhance several tasks' performance beyond corresponding SoTA scores through joint learning across multiple tasks due to the synergy effect.	& Assign a mask weight of 0 or 1 to each task; mask=1 only if the corresponding score ($\sigma_i^{C}$ or $\sigma_i^{G}$) exceeds the SoTA specialist's score, otherwise mask=0. Then, calculate the average score between $S_C$ and $S_G$. The more tasks to surpass the SoTA specialist, the higher the $S_3$:
\setlength{\abovedisplayskip}{0pt}
\setlength{\belowdisplayskip}{0pt}
\setlength{\abovedisplayshortskip}{0pt}
\setlength{\belowdisplayshortskip}{0pt}
\begin{equation*}
\begin{aligned}
    S_3 &= \frac{1}{2} \left( S_G + S_C  \right) \,, \text{where} \\
    S_C &= \frac{1}{M} \sum_{i=1}^M 
        \begin{cases} 
        \sigma^{C}_i & \text{if } \sigma_i^{C} \geq \sigma^{C}_{{sota}} \\
        0 & \text{otherwise}
        \end{cases} \\
    S_G &= \frac{1}{N} \sum_{j=1}^N 
        \begin{cases} 
        \sigma^{G}_j & \text{if } \sigma^{G}_j \geq \sigma^{G}_{{sota}} \\
        0 & \text{otherwise}
        \end{cases}
\end{aligned}
\end{equation*}
& 
GPT-4o \cite{gpt4}, 
Gemini-1.5 \cite{team2024gemini}, 
Claude-3.5 \cite{TheClaude3}, 
DeepSeek-VL \cite{lu2024deepseek}, 
LLaVA-One-Vision \cite{li2024llavaonevision}, 
Qwen2-VL \cite{wang2024qwen2}, 
InternVL2.5 \cite{chen2024internvl}, 
Phi-3.5-Vision \cite{abdin2024phi},  
$\cdots$
\\
\hline

\addlinespace[2pt]
\rowcolor{lightgreen} \multicolumn{4}{l}{\em$\downarrow$ \textbf{Upgrading Condition}: Generalists in unified comprehension and generation capability with synergy in between \vspace{-\baselineskip}}\\[2pt]

\hline

\addlinespace[2pt]
\levelivlogo \textbf{Level-4}: \newline Generalists with \textbf{synergy} across \textcolor{greenCom}{Comprehension} and \textcolor{blueGen}{Generation}	& Models are task-unified players, and synergy is across \textcolor{greenCom}{Comprehension} and \textcolor{blueGen}{Generation}.		& Calculate the harmonic mean between \textcolor{greenCom}{Comprehension} and \textcolor{blueGen}{Generation} scores.
The stronger synergy a model has between \textcolor{greenCom}{Comprehension} and \textcolor{blueGen}{Generation} tasks, the higher the score:
\setlength{\abovedisplayskip}{2pt}
\setlength{\belowdisplayskip}{0pt}
\setlength{\abovedisplayshortskip}{0pt}
\setlength{\belowdisplayshortskip}{0pt}
\begin{equation*}
    S_4 = \frac{2 S_C S_G}{S_C + S_G} 
\end{equation*}
& Mini-Gemini \cite{li2024mini}, Vitron-V1 \cite{fei2024vitron}, Emu2-37B \cite{sun2024generative}, 
$\cdots$
\\
\hline

\addlinespace[2pt]
\rowcolor{lightgreen} \multicolumn{4}{l}{\em$\downarrow$ \textbf{Upgrading Condition}: Generalists achieving cross-modal synergy with abductive reasoning ability \vspace{-\baselineskip}}\\[2pt]

\hline

\addlinespace[2pt]
\levelvlogo \textbf{Level-5}: \newline Generalists with \textbf{total synergy} across \textcolor{greenCom}{Comprehension}, \textcolor{blueGen}{Generation} and \textcolor{redNLP}{Language}	& Models are task-unified players, preserving the synergy effect across \textcolor{greenCom}{Comprehension}, \textcolor{blueGen}{Generation}, and \textcolor{redNLP}{Language}. In other words, the model not only achieves cross-modality synergy between \textcolor{greenCom}{Comprehension} and \textcolor{blueGen}{Generation} groups but also further realizes synergy with language. The \textcolor{redNLP}{Language} intelligence can enhance multimodal intelligence and vice versa; understanding multimodal information can also aid in understanding language.		& Calculate the model's average score exceeding SoTA NLP specialists on NLP benchmark data; normalize it to a [0,1] weight, and multiply it by the score from level-4 as the level-5 score:
\setlength{\abovedisplayskip}{2pt}
\setlength{\belowdisplayskip}{0pt}
\setlength{\abovedisplayshortskip}{0pt}
\setlength{\belowdisplayshortskip}{0pt}
\begin{equation*}
\begin{aligned}
   S_{5} &= S_{4} \times w_{L} \,, \text{where} \\
   w_L &= \frac{S_L}{S_{\text{total}}} \,, \text{where} \\
   S_L &= \frac{1}{T} \sum_{k=1}^T 
    \begin{cases} 
    \sigma_k & \text{if } \sigma_k \geq \sigma_{\text{sota}} \\
    0 & \text{otherwise}
    \end{cases}
\end{aligned}
\end{equation*}
& \color{mypink}{\emph{None found yet (Let's wait for multimodal ChatGPT moment!)}} \\
\hline

\end{tabular}
\vspace{-5mm}
\end{table*}
}

\vspace{-1mm}
\subsubsection{Scoring Specification}

When calculating scores using the corresponding formula, we normalize all task metrics to a 100-point scale. 
While most task evaluation scores typically range from 0-100, such as \emph{F1} and \emph{Accuracy}, certain metrics, e.g., \emph{FID}, \emph{MAE}, and \emph{PSNR}, yet yield scores outside this usual range. 
Thus, we design some mapping functions to standardize performance scores.
Our framework also incorporates the principle of diminishing scores: an MLLM (i.e., multimodal generalist) can achieve scores at multiple levels, but it is classified at its highest level, where it achieves a non-zero score.

We assume that current MLLMs have already demonstrated synergy mode from language to non-language modalities. 
Then the remaining mission is to confirm the existence of synergy in the reverse direction, from non-language to language modalities. 
Therefore, for level 5—measuring total synergy—we do not measure the generality across all modalities and tasks. 
Instead, we assess whether a model can improve NLP task performance to exceed that of NLP SoTA specialists.

Also, except for Level-1 and Level-5, when calculating $S_2$, $S_3$, and $S_4$, we consider a reasonable approach when handling different modalities. 
First, we calculate the specific score component $S_k^{i}$ of a generalist in the $i$-th modality (assuming there are $N$ modalities in total) for the score $S_k$.
This modality-specific component can accurately reflect the model's Level-$k$ capability in the $i$-th modality.
Next, by decomposing each score into its components across different modalities, we sum the components of each modality with equal weights to obtain the overall score for each level.
\begin{equation*}
S_k = \sum_i^N \frac{1}{N} S_k^{i} 
\end{equation*}
The advantage of this method is that it reduces the bias introduced by the number of tasks in different modalities. 
For example, in our benchmark, image-related tasks (especially comprehension-type tasks) are overwhelmingly more numerous compared to other modalities, such as audio tasks. 
Therefore, two generalists with similar capability levels, say one for image tasks and the other for audio tasks, would have a higher $S_k$ score for the image-generalist over the audio-generalist, due to the larger number of image tasks. 
This discrepancy is unrealistic and contrary to our core idea for evaluating multimodal generalists. 
To eliminate the bias caused by the number of tasks within each modality, we propose the above calculation method, which treats the capabilities of different modalities equally.
Meanwhile, this method also prioritizes generalists that can support more modalities. 
For instance, a model that supports more modalities will certainly have a higher overall score compared to a generalist that supports only one modality.

This scoring method ensures that as an MLLM climbs to higher levels, its scores progressively decrease, which should indicate the increasing difficulty of advancing levels. 
Climbing from level $n$ to level $n+1$ requires specific capabilities, i.e., demonstrating sufficient synergy capability associated with that level, which we highlight as critical factors in Table \ref{tab:levels}.
Within the same level, to achieve a higher score, a model must: 1) support as many tasks and modalities as possible, and simultaneously 2) achieve the highest possible performance on individual tasks.

\vspace{-2mm}
\subsubsection{Scoring Relaxation}

A central aspect of our \texttt{General-Level} framework lies in how synergy effects are computed. According to the standard understanding of the `synergy' concept, e.g., \emph{the performance of a generalist model on joint modeling of tasks A and B (e.g., $P_\theta(y|A,B)$) should exceed its performance when modeling task A alone (e.g., $P_\theta(y|A)$) or task B alone (e.g., $P_\theta(y|B)$)}. However, adopting this approach poses a significant challenge that hinders the measurement of synergy: there is no feasible way to establish two independent distributions, $P_\theta(y|A)$ and $P_\theta(y|B)$, and a joint distribution $P_\theta(y|A,B)$. 
This limitation arises because a given generalist model has already undergone extensive pre-training and fine-tuning, where tasks A and B have likely been jointly modeled. It is impractical to retrain such a generalist to isolate the learning and modeling of tasks A or B independently in order to derive these distributions. 
Otherwise, such an approach would result in excessive redundant computation and inference on the benchmark data.

To simplify and relax the evaluation of synergy, we introduce a key assumption in the scoring algorithm:  
\vspace{-2mm}
\begin{tcolorbox}[fontupper=\linespread{0.9}\selectfont,breakable]
{
\vspace{-1mm}
\emph{
Theoretically, we posit that the stronger a model's synergy capability, the more likely it is to surpass the task performance of SoTA specialists when synergy is effectively employed. 
Then, we can simplify the synergy measurement as: if a generalist outperforms a SoTA specialist in a specific task, we consider it as evidence of a synergy effect, i.e., leveraging the knowledge learned from other tasks or modalities to enhance its performance in the targeted task.
\vspace{-1mm}
}
}
\end{tcolorbox}
By making this assumption, we avoid the need for direct pairwise measurements between `task-task', `comprehension-generation', or `modality-modality', which would otherwise require complex and computationally intensive algorithms.

\vspace{-2mm}
\subsubsection{Properties of General-Level}

The \texttt{General-Level} framework possesses several important attributes that play a critical role in supporting the hierarchical classification and ranking of MLLMs. 
These properties are also well-grounded in mathematical theory.

\paragraph{Property-1: Independence from Peer Generalists}

In our scoring framework, the scores of any generalist depend solely on the dataset and the reference scores of SoTA specialists, without relying on the scores of other tested generalists. These two components are entirely independent. The dataset defines the specific tasks, while the specialists provide baseline reference scores used for the calculation of the experimental generalists' scores.
This property ensures that the evaluation of generalists is free from interdependence, maintaining objectivity and fairness among all systems participating in the ranking.

\paragraph{Property-2: Monotonicity Across Levels}

Generally, if a generalist is rated at the highest level-$k$, it is expected to achieve scores at all levels from 2 to $k$. 
We further expect that as the level increases, the corresponding scores for the generalist will decrease, i.e., $S_{k-1} > S_k$.  
This is a reasonable and realistic requirement, as higher levels impose stricter demands on the generalist’s capabilities, naturally leading to lower scores for the same model.  
Below, we provide proof that the scoring algorithm of \texttt{General-Level} framework mathematically guarantees the strictly monotonic score decline across levels.

\begin{tcolorbox}[
    fontupper=\linespread{0.9}\selectfont,
    breakable,
    title={$\blacktriangleright$ \textbf{The proof for $S_3\leq S_2$}}
]
{
\small
\begin{equation*}
\begin{aligned}
S_3 =& \frac{1}{2} \left( S_G + S_C  \right)  \\
    =& \frac{1}{2} \left(
        \frac{1}{M} \sum_{i=1}^M 
        \begin{cases} 
        \sigma^{C}_i & \text{if } \sigma_i^{C} \geq \sigma^{C}_{{sota}} \\
        0 & \text{otherwise}
        \end{cases} 
        + 
        \frac{1}{N} \sum_{j=1}^N 
        \begin{cases} 
        \sigma^{G}_j & \text{if } \sigma^{G}_j \geq \sigma^{G}_{{sota}} \\
        0 & \text{otherwise}
        \end{cases}
         \right) \\
    \leq & \frac{1}{2} \left( \frac{1}{M} \sum^{M}_{i=1} \sigma_{i}^{C} + \frac{1}{N} \sum^{N}_{j=1} \sigma_{i}^{G}  \right) \\
    = & S_2
\end{aligned}  
\end{equation*} 
}
\end{tcolorbox}

\begin{tcolorbox}[
    fontupper=\linespread{0.9}\selectfont,
    breakable,
    title={$\blacktriangleright$ \textbf{The proof for $S_4 \leq S_3$}}
]
{
\small
\par\textbf{Suppose:}\\
\begin{equation*}
\begin{aligned}
    S_G &= \frac{1}{M} \sum_{i=1}^M 
        \begin{cases} 
        \sigma_i & \text{if } \sigma_i \geq \sigma_{{sota}} \\
        0 & \text{otherwise}
        \end{cases} \\
    S_C &= \frac{1}{N} \sum_{j=1}^N 
        \begin{cases} 
        \sigma_j & \text{if } \sigma_j \geq \sigma_{{sota}} \\
        0 & \text{otherwise}
        \end{cases} \\
\end{aligned}
\end{equation*}

\par\textbf{According to \emph{Cauchy-Schwarz Inequality}, let's represent}\\
\begin{equation*}
   \left( \frac{S_C + S_G}{2} \right)^{2} \geq \left( \frac{2 S_C S_G}{S_C + S_G} \right) 
\end{equation*}

\par\textbf{Expanding this,}\\
\begin{equation*}
    \frac{(S_C + S_G)^2}{4} \geq \frac{2 S_C S_G}{S_C + S_G}
\end{equation*}

\par\textbf{Multiplying both sides by $4(S_C + S_G)$,}\\
\begin{equation*}
    (S_C + S_G)^3 \geq 8 S_C S_G(S_C + S_G)
\end{equation*}

\par\textbf{Simplifying further}\\
\begin{equation*}
    S_C^3 + S_G^3 \geq 2 S_C S_G(S_C + S_G)
\end{equation*}

\par\textbf{This factorizes to}\\
\begin{equation*}
    (S_C - S_G)^2(S_C + S_G) \geq 0 \,.
\end{equation*}

\par\textbf{Finally, we have}\\
\begin{equation*}
\begin{aligned}
    S_4 &= \frac{2 S_C S_G}{S_C + S_G}  \\
        &\leq  \frac{1}{2} \left( S_C + S_G \right)  \\
        &=  S_3 \,.
\end{aligned}
\end{equation*}
}
\end{tcolorbox}

\begin{tcolorbox}[
    fontupper=\linespread{0.9}\selectfont,
    breakable,
    title={$\blacktriangleright$ \textbf{The proof for $S_5\leq S_4$}}
]
{
\small

\par\textbf{We have} \\
\begin{equation*}
\begin{aligned}
   w_L &= \frac{S_L}{S_{\text{total}}} \,, \text{where} \\
   S_L &= \frac{1}{T} \sum_{k=1}^T 
    \begin{cases} 
    \sigma_k & \text{if } \sigma_k \geq \sigma_{\text{sota}} \\
    0 & \text{otherwise}
    \end{cases}
\end{aligned}
\end{equation*}

\par\textbf{which means,} \\
\begin{equation*}
    w_L \leq 1 \,.
\end{equation*}

\par\textbf{Then} \\
\begin{equation*}
\begin{aligned}
    S_5 - S_4 &= S_4 * w_L - S_4 \\
              &= S_4 * (w_L - 1) \\
              &\leq 0
\end{aligned}
\end{equation*}

\par\textbf{Thus,} \\
\begin{equation*}
\begin{aligned}
    S_5 - S_4 &\leq 0
\end{aligned}
\end{equation*}
}
\end{tcolorbox}

\vspace{2mm}
\noindent\textbf{Property-3: Encouraging Rich and Balanced Multimodal Task Support.}

\vspace{0.5em}

\indent\textbf{$\blacktriangleright$ More Task, The Better.}
A good multimodal evaluation system should not only reward models for achieving higher scores on individual tasks and surpassing SoTA specialists but also incentivize a trend where multimodal generalists support as many diverse multimodal tasks as possible. This is a reasonable expectation, as an ideal multimodal generalist should inherently support a broader range of modalities and tasks.
The scoring algorithm of our \texttt{General-Level} framework aligns with this objective. For instance, in the case of level-2 scoring:
\begin{equation*}
    S_2 = \frac{1}{M+N} \sum^{M+N}_{i=1} \sigma_{i} \,,
\end{equation*}
a model that achieves nonzero scores across a greater number of modalities and tasks will naturally obtain a higher average score, thereby ranking higher within the same level.

\textbf{$\blacktriangleright$ More Balance, The Better.}
Moreover, our scoring algorithm also promotes models that achieve more balanced performance across tasks. For example, in the case of level-4 scoring,
consider the following scenarios:  
\begin{compactitem}
    \item[1)] Model A achieves SoTA specialist performance on X tasks in the comprehension category but only Y tasks (where $X \gg Y$) in the generation category.  
    \item[2)] Model B achieves SoTA specialist performance on X tasks in both the comprehension and generation categories.  
\end{compactitem}
According to the properties of the harmonic mean inequality, $S_4^A < S_4^B$.

\begin{tcolorbox}[
    fontupper=\linespread{0.9}\selectfont,
    breakable,
    title={$\blacktriangleright$ \textbf{The proof for $S_4^A < S_4^B$} when $X \gg Y$ in level-4}
]
{
\small

\par\textbf{Extreme Assumptions:}\\
- For Model A, the \( X \) tasks in the comprehension group have scores of \(\sigma_C^A = 1\), and the \( Y \) tasks in the generation group have scores of \(\sigma_G^A = 1\), while all other scores are 0.\\
- For Model B, both comprehension and generation groups have \( X \) tasks with scores of \(\sigma_C^B = 1\) and \(\sigma_G^B = 1\), while all other scores are 0.

\par\textbf{Model-A Scores:}\\
For Model A, the comprehension and generation scores are:
\[
S_C^A = \frac{X}{M}, \quad S_G^A = \frac{Y}{N}.
\]
The overall score for Model A is:
\[
S_4^A = \frac{2 \cdot S_C^A \cdot S_G^A}{S_C^A + S_G^A} = \frac{2 \cdot \frac{X}{M} \cdot \frac{Y}{N}}{\frac{X}{M} + \frac{Y}{N}} = \frac{2XY}{XN + YM}.
\]

\par\textbf{Model-B Scores:}\\
For Model B, both comprehension and generation groups have \( X \) tasks with scores of 1, so:
\[
S_C^B = \frac{X}{M}, \quad S_G^B = \frac{X}{N}.
\]
The overall score for Model B is:
\[
S_4^B = \frac{2 \cdot S_C^B \cdot S_G^B}{S_C^B + S_G^B} = \frac{2 \cdot \frac{X}{M} \cdot \frac{X}{N}}{\frac{X}{M} + \frac{X}{N}} = \frac{X^2}{XN + XM}.
\]

\par\textbf{Comparison:}\\
We need to compare:
\[
\frac{2XY}{XN + YM} \quad \text{and} \quad \frac{X^2}{XN + XM}.
\]
Given \( X \gg Y \), it follows that:
\[
\frac{2XY}{XN + YM} < \frac{X^2}{XN + XM}.
\]
Thus, \( S_4^A < S_4^B \).

}
\end{tcolorbox}
Through the above mathematical analysis, we have proven that under the same task distribution, the uneven generation score distribution of Model A results in its level-4 score being lower than that of Model B.
This ensures that models with more balanced performance across comprehension and generation are ranked higher.

\paragraph{Property-4: Dynamic Update on Benchmarking and Specialists}

Finally, we observe an important point: the more tasks included in the benchmark used to evaluate models, the more accurate and objective the resulting evaluations and conclusions. 
This requirement for the evaluation benchmark to have dynamic properties aligns well with real-world needs. 
In practice, new tasks, data, and even new modalities are constantly being introduced, and a generalist should be capable of covering these newly added tasks and functionalities.  
Accordingly, in our evaluation system, we allow the benchmark to evolve dynamically, such as by adding new tasks under various modalities and categories. 
Once new tasks are added, we update the scores and rankings of all tested generalists to reflect the expanded benchmark.

On the other hand, we also allow updates to the SoTA specialist models timely for each task, as scoring at higher levels is anchored to the performance of the SoTA models. 
This is a reasonable act, as specialists are continually being developed and improved. 
Once a baseline specialist advances, generalists must also improve to remain competitive, or risk being surpassed.  
Thus, in \texttt{General-Level} framework, the scores corresponding to SoTA specialists are subject to periodic updates. 
Also, we dynamically and regularly update the scoring and ranking of all generalists to ensure the evaluation remains accurate and reflective of the current state of the field.

\subsection{Receipt to Leveling Upper in General-Level}

Here we provide a guideline to help better understand how to achieve higher levels in \texttt{General-Level} framework.

\vspace{-2mm}
\paragraph{Level-1$\to$Level-2: Supporting as many tasks and functionalities as possible.}

Transitioning from specialists to generalists requires making the system compatible with various task modeling paradigms, i.e., supporting diverse modality types and input formats, as well as handling a wide range of model types and output formats (whether for comprehension and/or generation).  
Currently, the most popular and widely adopted practice is to use an LLM as the backbone/intelligence medium, integrating various specialists to build generalists. 
There are two primary implementation strategies.

First, agent-based generalists \cite{abs-2303-04671,abs-2303-17580}.
In this approach, the LLM acts as a task scheduler and dispatcher, facilitating message passing through hard integration (explicit text). This is essentially a pipeline architecture. However, since gradient propagation across the entire system is not feasible, this method is prone to error propagation.  
The performance upper bound of generalists built with this approach is equivalent to the SoTA specialists for all supported tasks, primarily due to the lack of features, information sharing, and limited task collaboration.

Second, end-to-end generalists \cite{liu2023llava,0008LSH23,abs-2304-10592}.
In this type, the entire system is constructed as a continuous joint model, allowing for full-stack updates via gradient propagation.  
The most common architecture in this category uses an LLM as the backbone, achieving soft integration of various encoders and decoders through input tokenization and feature embedding, combined with overall fine-tuning.

\vspace{-2mm}
\paragraph{Level-2 $\to$ Level-3: Generalists achieving as stronger synergy and cross as many tasks as possible.}

To advance from a vanilla generalist to Level-3, the system must demonstrate cross-task synergy capabilities, enabling at least two tasks (regardless of whether both involve comprehension, generation, or one involves comprehension while the other involves generation) to share features and achieve mutual performance improvements.  
The most direct method to realize cross-task synergy is through multi-task joint training.  
Specifically, during joint learning, the system must ensure it can maintain task-shared/persistent common features while preserving each task's specific features without degradation, e.g., Vitron \cite{fei2024vitron}.  
Moreover, the model must support synergy across as many tasks as possible and ensure that the synergy effect is significant enough to achieve higher evaluations at Level-3.

\vspace{-2mm}
\paragraph{Level-3$\to$Level-4: Generalists in unified comprehension and generation capability with synergy in between.}

To advance to Level-4, generalists must first achieve unified comprehension and generation capabilities, regardless of whether they support a single modality (non-NLP) or multiple modalities.  
At the same time, the system must meet the requirement that its capabilities in comprehension and generation synergize and enhance one another.  
Generally speaking, compared to acquiring comprehension capabilities, obtaining generation capabilities at the technical level is relatively more challenging.  
For instance, the visual comprehension abilities of most visual LLMs tend to be significantly stronger than their visual generation capabilities.  
If a generalist can score at Level-4, it indicates that the system not only possesses strong comprehension capabilities but also maintains these capabilities while further learning and training its generation abilities.  
To achieve this, Morph-Token \cite{pan2024auto} introduces a disentangling visual reconstruction loss for generation learning to avoid interference with the comprehension learning loss.

\vspace{-2mm}
\paragraph{Level-4$\to$Level-5: Generalists achieving cross-modal synergy with abductive reasoning ability.}
Achieving Level-5 represents the ultimate goal for generalists, where features, knowledge, and even intelligence learned from tasks in certain modalities can (to varying degrees) transfer to tasks in other supported modalities.  
Currently, most multimodal generalists are limited by architectural developments, primarily enabling language intelligence to support intelligence in other modalities (as illustrated in Figure \ref{fig:intro}).  
However, to truly achieve Level-5, synergy must exist across all modalities.  
For instance, in the current MLLM community, this would require MLLMs to enhance performance on NLP tasks as well, while most of the MLLMs perform unsatisfactorily in NLP tasks.  
From a technical perspective, generalists must be capable of abductive reasoning, i.e., the ability to infer and generalize across everything.  
Also, they need to ensure modality-agnostic context consistency during reasoning.

\vspace{-2mm}
\section{\emph{General-Bench}: A Holistic Benchmark for Multimodal Generalists}

We introduce \texttt{General-Bench}, a new benchmark to meet the outlined criteria and serve as the standard dataset for our evaluation framework.

\vspace{-1mm}
\subsection{Data Construction}

\subsubsection{Design Criterion}

As previously noted, the current benchmarks that rank MLLMs based solely on their performance have significant limitations, which hinder the encouragement of MLLMs to evolve toward becoming more capable multimodal generalists. 
Primarily, nearly all existing benchmarks focus on evaluating MLLMs' capabilities in visual modalities, particularly images, while significantly neglecting tasks in other modalities such as video, audio, 3D, etc. 
Moreover, they often assume that MLLMs already possess satisfied NLP capabilities, thus omitting evaluations in language.

\begin{figure*}[!t]
\centering
\includegraphics[width=0.99\linewidth]{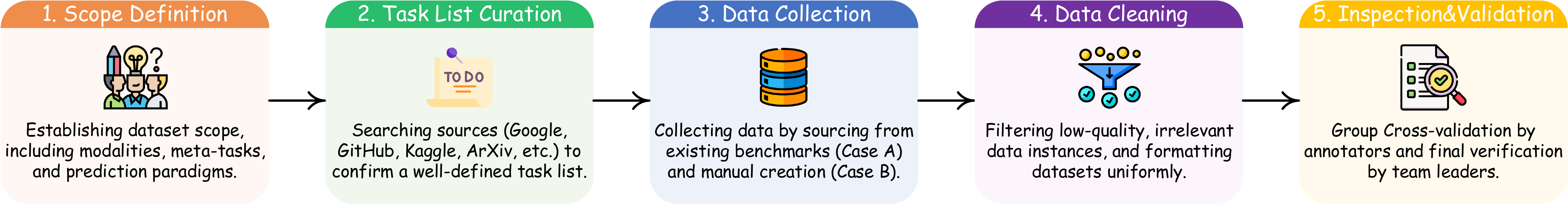}
\vspace{-2mm}
\caption{An illustration of the data construction pipeline of \texttt{General-Bench}.}
\label{fig:data-construction}
\vspace{-4mm}
\end{figure*}

Secondly, these benchmarks tend to simply convert free-form predictions into fixed QA format of pre-defined choices—essentially a compromise that reflects the current limitations of MLLM capabilities—allowing many tasks that MLLMs cannot produce in specific formats to still be executed. 
We believe that a genuine multimodal generalist should support tasks in their original formats. 
Furthermore, most benchmarks only assess MLLMs' understanding of visual information; however, a multimodal generalist should inherently possess a wide range of capabilities beyond mere comprehension, such as generation, editing, etc.
Therefore, we expect to construct a benchmark that possesses these characteristics:
\vspace{-2mm}
\begin{compactitem}
    \item Covering as broad a range of tasks, skills and modalities as possible.
    
    \item Encompassing both comprehension and generation of tasks.

    \item Including a rich diversity of tasks across various scenarios and domains. 
    
    \item Preserving the original task-prediction formats.
    
    \item Timely maintaining and expanding the dataset dynamically.

\end{compactitem}

\subsubsection{Construction Process}

The construction of our \texttt{General-Bench} dataset follows a structured 5-step process to ensure both comprehensiveness and quality.
Figure \ref{fig:data-construction} presents the data construction pipeline.

\vspace{-2mm}
\paragraph{Step-1: Defining Scope and Range.}  
We begin by conducting a series of panel discussions to establish the scope of the dataset. 
This involves determining the modalities to include, identifying the core general skills (meta-tasks), and specifying the prediction paradigms to address. 
These discussions help outline a comprehensive framework for the dataset, ensuring that it accommodates diverse tasks and capabilities required for evaluating multimodal generalists.

\vspace{-2mm}
\paragraph{Step-2: Curating Task List.}  
Based on the defined scope, we curate a comprehensive task list by systematically searching various sources, including Google, GitHub, Kaggle, ArXiv, and PaperWithCode, etc. 
For each task, we specify its input-output targets, select appropriate evaluation metrics, and also identify SoTA specialists as reference points. 
This step ensures that each task is well-defined and aligned with existing SoTA practices.

\vspace{-2mm}
\paragraph{Step-3: Collecting Data.}  
Next, we start collecting the data instances.
The data collection process is divided into two cases for handling two different scenarios:

\begin{compactitem}
    \item \textbf{Case A:} If the data could be sourced from existing benchmark datasets (only from their test sets), modifications are made to enhance diversity. We will show all the data sources of our benchmark in the following subsections. 
    For textual data, rephrasing is done using ChatGPT. For non-textual modalities such as images, videos, and audio, semantically equivalent replacements are identified through retrieval or direct recording from relevant databases or websites.
    
    \item \textbf{Case B:} For tasks without available datasets or insufficient enough numbers of samples, we manually create instances. This involves crafting input-output pairs according to the task definition, running existing models to generate predictions, and performing manual verification and correction of the results.
\end{compactitem}

We ensure that each task includes (at least) 500 data samples.
Also, we ensure that all tasks faithfully retain their original input-output prediction structure or format, i.e., not reformatted into QA-based multiple-choice questions.

\vspace{-2mm}
\paragraph{Step-4: Data Filtering and Cleaning.}  
After collecting datasets for all modalities and tasks, we proceed with data filtering and cleaning.  
First, we filter out low-quality instances, including those that do not align well with the task's evaluation purpose, lack target modality information, or fail to meet the defined prediction paradigms. 
For tasks where the number of instances is insufficient, we restart the data annotation process to supplement the required quantity.  
Afterward, we organize all data into a unified storage format according to the designed specifications. 
For example, textual data is standardized into JSON files with consistent naming conventions applied to all files.

\vspace{-2mm}
\paragraph{Step-5: Data Inspection and Validation.}  
Finally, we conduct a rigorous inspection and validation process to guarantee data quality and consistency. 
Annotators work in groups of three, independently reviewing the same instance. 
An instance is accepted only if all three annotators reach consensus. 
Finally, team leaders or supervisors conduct an additional round of verification to ensure the dataset meets the highest standards of consistency and accuracy.

\begin{figure*}[!t]
\centering
\includegraphics[width=0.99\linewidth]{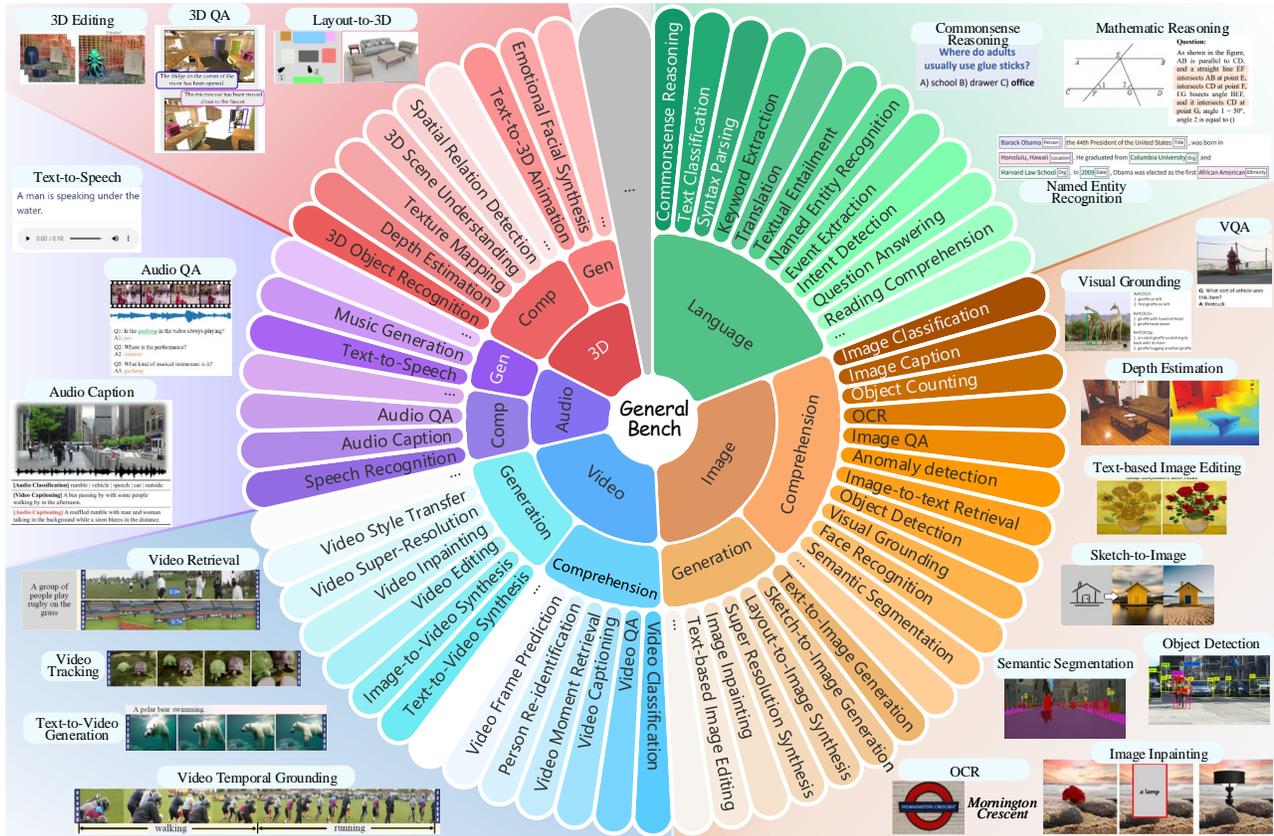}
\vspace{-2mm}
\caption{
Overview of \texttt{\textbf{General-Bench}}, which covers 145 skills for more than 700 tasks with over 325,800 samples under comprehension and generation categories in various modalities.
Appendix $\S$ \ref{Overall Data Taxonomy and Hierarchy} gives holistic hierarchical taxonomies.
}
\label{fig:dataset-sum}
\vspace{-4mm}
\end{figure*}

\subsection{Evaluation and Splitting}

As each task follows the original format, our evaluation metrics vary in rich task types. 
For instance, we evaluate $X$-to-text generation tasks using BLEU/ROUGE/CIDEr scores, image segmentation tasks with mIoU for generating masks, and image generation tasks using FID, etc.
Also, we design some mapping functions to standardize performance scores.
In Appendix \S\ref{Evaluation Metrics} we present the evaluation metrics as well as the mapping tricks in detail.

For most of the tasks, we maintain around 500 testing instances each.
Considering that not all practitioners in the community may be interested in participating in the leaderboard—for example, some may simply wish to use our dataset for their research or publications—we propose dividing the test set for each task into a closed set and an open set.
The closed set is reserved for leaderboard evaluations: only the input data is released, and users are required to submit their model’s predicted outputs for centralized assessment.
In contrast, the open set provides full access to both inputs and corresponding outputs, enabling practitioners to explore and utilize the data more freely.
Each task's test set is split into closed and open subsets with a ratio of 2:3.

\begin{figure*}[!t]
\centering
\includegraphics[width=0.93\linewidth]{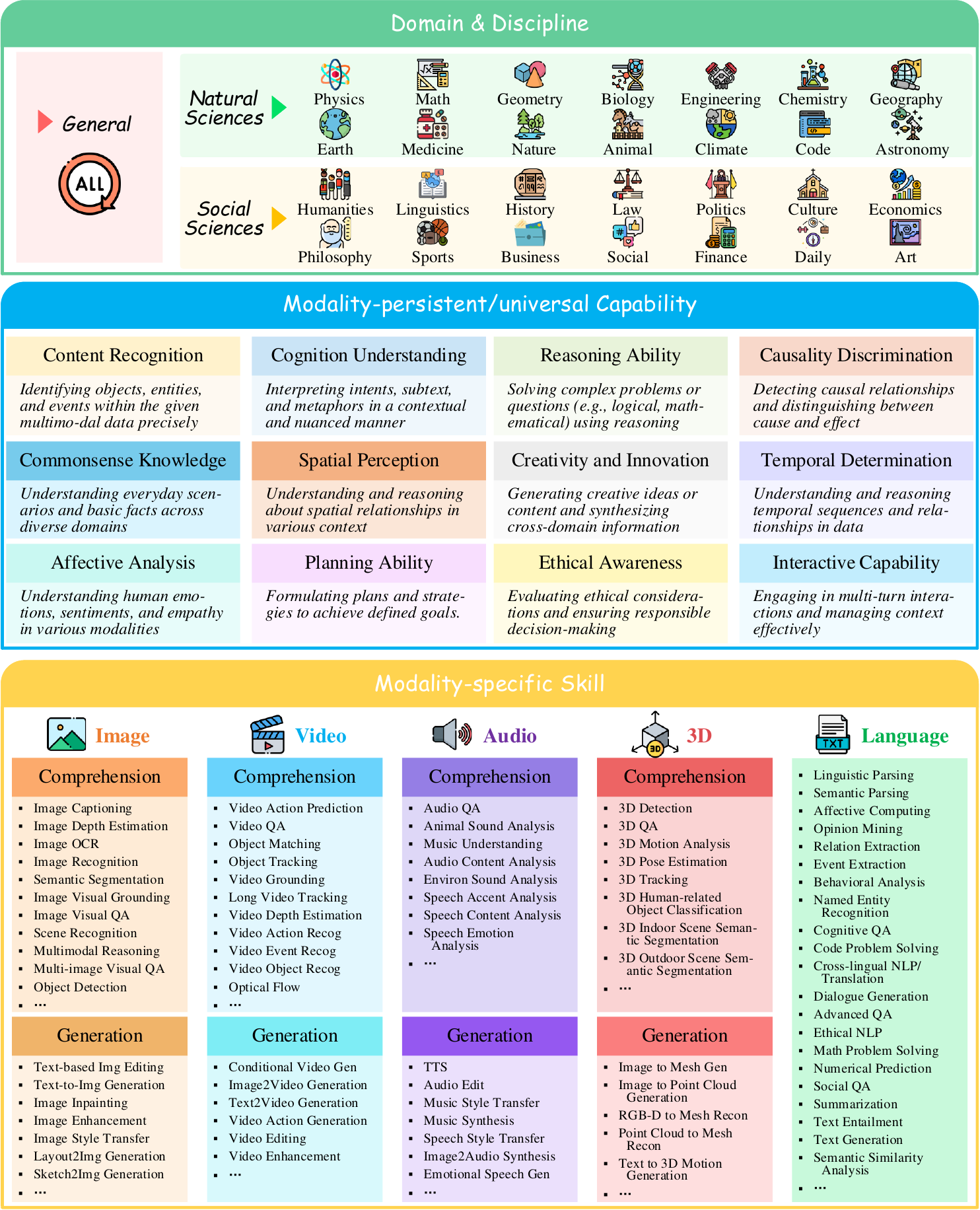}
\vspace{-3mm}
\caption{
\texttt{\textbf{General-Bench}} covers over 29 domains, evaluating more than 12 modality-persistent capabilities of generalists, as well as 145 modality-specific skills.
In Appendix $\S$\ref{Data Specification and Distributions} we showcase all tasks and data specification in detail.
}
\label{fig:dataset-insight}
\vspace{-5mm}
\end{figure*}

\begin{table}[!t]
\centering
\fontsize{8.5}{11}\selectfont
\setlength{\tabcolsep}{2.5mm}
\caption{Summary of numbers of skills, tasks and data instances across modalities.}
\label{tab:skills-tasks-stat}
\begin{tabular}{cccccccccccc}
\hline
& & \multicolumn{2}{c}{\bf Image} & \multicolumn{2}{c}{\bf Video} & \multicolumn{2}{c}{\bf Audio} & \multicolumn{2}{c}{\bf 3D} & \multicolumn{1}{c}{\multirow{2}{*}{\bf Language}} & \multicolumn{1}{c}{\multirow{2}{*}{\emph{\textbf{TOTAL}}}} \\
\cmidrule(r){3-4} \cmidrule(r){5-6} \cmidrule(r){7-8} \cmidrule(r){9-10} 
& & Comp & Gen & Comp & Gen & Comp & Gen & Comp & Gen & & \\
\hline

\multirow{2}{*}{\#Skill} & Single & 40 & 15 & 20 & 6 & 9 & 11 & 13 & 9 & \multirow{2}{*}{22} & \multirow{2}{*}{145} \\
\cdashline{2-10}
& Sum & \multicolumn{2}{c}{55} & \multicolumn{2}{c}{26} & \multicolumn{2}{c}{20} & \multicolumn{2}{c}{22} &  & \\

\hline

\multirow{2}{*}{\#Task} & Single & 271 & 45 & 126 & 46 & 24 & 20 & 30 & 22 & \multirow{2}{*}{118} & \multirow{2}{*}{702} \\
\cdashline{2-10}
& Sum & \multicolumn{2}{c}{316} & \multicolumn{2}{c}{170} & \multicolumn{2}{c}{44} & \multicolumn{2}{c}{52} & & \\

\hline

\multirow{2}{*}{\#Instance} & Single & 124,880 & 26,610 & 44,442  & 16,430  &  11,247 & 9,516  & 23,705  &  10,614 &  \multirow{2}{*}{58,432} & \multirow{2}{*}{325,876} \\
\cdashline{2-10}
& Sum & \multicolumn{2}{c}{151,490} & \multicolumn{2}{c}{60,872} & \multicolumn{2}{c}{20,763} & \multicolumn{2}{c}{34,319} & & \\
\hline
\end{tabular}
\vspace{-2mm}
\end{table}

{
\begin{table*}[!t]
\centering
\caption{
Comparison of \texttt{\textbf{General-Bench}} with existing representative MLLM benchmarks. 
`Comp.': Comprehension; `Gen.': Generation.
Appendix~$\S$\ref{Full Data Statistics} presents a complete view for more comparisons of exiting benchmarks.
 }
\label{tab:data-statistics}
\vspace{-2mm}
\fontsize{8}{10}\selectfont
\setlength{\tabcolsep}{1.0mm}
\begin{tabular}{lcccccccc}
\toprule
 \textbf{Benchmark} & SEED-Bench & MMBench & MMMU   & LVLM-eHub  & MMIU  &MMT-Bench &MEGA-Bench  & \bf General-Bench \\

\hline

\rowcolor{lightgreen}\textbf{\bf  Modality} & Txt,Img,Vid &  Txt,Img  &  Txt,Img & Txt,Img  & \makecell{Txt,Img,Vid,\\Point-Cloud,Depth}  & \makecell{Txt,Img,Vid,\\Point-Cloud} & Txt,Img,Vid & \makecell{Txt,Img,Vid,Aud,\\Time,Depth,3D-RGB,\\Point-Cloud,Infrared,\\Spectrogram,Radar,\\Code,Doc,Graph,$\cdots$} \\

\textbf{\bf Task Scheme}  & Comp. &  Comp.  & Comp.  & Comp.  &  Comp. &  Comp. &  Comp. & Comp.+Gen. \\

\bottomrule

\rowcolor{lightgreen} \textbf{\bf  \# Domain} & 1 & 1   & 6  & 1  &  1 & 4 & 5 & \color{black}{\bf 29} \\

\textbf{\bf  \# Skill} & 12 & 2 & 6  &  6 &  7 & 32 & 10 & \color{black}{\bf 145} \\

\rowcolor{lightgreen} \textbf{\bf  \# Task} & 12 & 20  &  30  &  47 & 52  & 162 & 505 &\bf 702 \\

\textbf{\bf  \# Sample} & 19K &  3K  &  11.5K  & 2.1K & 11.7K  &  31K & 8K & \color{black}{\bf 325.8K} \\

\bottomrule

\textbf{\bf  Answer Form}  & MC-QA &  MC-QA  & MC-QA  & MC-QA  &  MC-QA &  MC-QA &  Free-Form &  Free-Form \\

\rowcolor{lightgreen} \textbf{\bf  \# Metric}  & Acc. &  Acc.  & Acc.  & Acc.  &  Acc. &  Acc. &  Origin (45) & Origin ({\bf 58}) \\

\textbf{\bf  Annotation} & Manual &   Repurposed &  Manual & Repurposed  &  Repurposed & Repurposed  & Manual & Manual \\
\bottomrule

\rowcolor{lightgreen} \textbf{\bf \# Tested Models}  & 12 & 21  & 24  & 8  &  22 & 30 & 22 & \color{black}{\bf 172+102} \\

\bottomrule
\end{tabular}
\vspace{-5mm}
\end{table*}
}

\vspace{-2mm}
\subsection{Data Insights}

First, Table \ref{tab:skills-tasks-stat} summarizes the statistics of task and skill numbers in \texttt{General-Bench}.
The data compiled for \texttt{General-Bench} is visualized in 
Figure \ref{fig:dataset-sum} visualizes the \texttt{General-Bench} highlights of task/modality support.
Overall, the current version of the dataset includes those most common modalities (inner ring), and except for NLP tasks, all modalities distinguish between comprehension and generation tasks (middle ring). 
\texttt{General-Bench} particularly places a strong emphasis on the diversity of its evaluation data, covering a wide range of fields and scenarios to assess different aspects of model capabilities, as depicted in Figure \ref{fig:dataset-insight}.
First, the dataset spans a variety of domains and disciplines, incorporating 28 major areas within both the physical sciences (e.g., Physics, Math, Geometry, Biology) and the social sciences (e.g., Humanities, Linguistics, History, Social). 
The evaluation of a generalist's skills and capabilities is categorized into universal modality-invariant abilities and modality-specific skills. 
The modality-invariant abilities comprehensively include 12 categories, such as content recognition, commonsense knowledge, reasoning ability, causality discrimination, affective analysis, creativity, and innovation, etc.
For modality-specific skills, we explicitly detail the main capabilities under both comprehension and generation for each modality, which correspond to the meta-tasks (skills) of our dataset.

In Table \ref{tab:data-statistics}, we further present a comparison with several existing popular benchmarks. 
It also covers the broadest range of disciplines and supports the widest array of modalities. 
\texttt{General-Bench} comprises 130 multimodal skills, containing 702 tasks with over 325,800 annotations across various formats and domains. 
The volume of tasks and data in \texttt{General-Bench} significantly exceeds that of current benchmarks. 
Moreover, our dataset facilitates original free-form task prediction, allowing for a more diverse array of task types.

\subsection{Leaderboard Re-Scoping}

Given the large scale of our dataset, it would be highly costly for practitioners to run the entire dataset under our proposed General-Level evaluation protocol.  
Moreover, it's realized that most existing multimodal generalists (e.g., MLLMs) have not yet reached the level of capability required to cover a wide range of modalities and tasks, as envisioned in our framework.  
As a result, many current models may find it difficult to fully demonstrate their potential on our leaderboard.  
To improve usability and encourage broader participation, we further propose a graded structure for the leaderboard by dividing its scope into four levels of increasing difficulty:

\begin{compactitem}
    \item  \textbf{Scope-A}: Full-spectrum leaderboard covering all modalities and tasks, designed for highly capable, general-purpose multimodal models. This scope has one leaderboard encompassing all levels in General-Level, making it the most challenging track.
    We further derive a full version and a quick version leaderboard for easier participation.

    \item  \textbf{Scope-B}: Modality-specific leaderboards, each focusing on a single modality or partially joint modality, and designed for modality-wise generalists. This scope maintains 4 separate leaderboards, one per modality (except for language).

    \item  \textbf{Scope-C}: Leaderboards focused on either comprehension or generation within a single modality. This scope includes 8 leaderboards: $2 \times 4$ for comprehension/generation across multimodal tasks, with a lower entry barrier for participation.

    \item  \textbf{Scope-D}: Finer-grained, skill-level (task-cluster-specific) leaderboards within each modality, tailored for partial generalists. This scope includes a large number of specific leaderboards, offering the lowest difficulty for participation.
\end{compactitem}

\begin{figure*}[!t]
\centering
\includegraphics[width=0.96\linewidth]{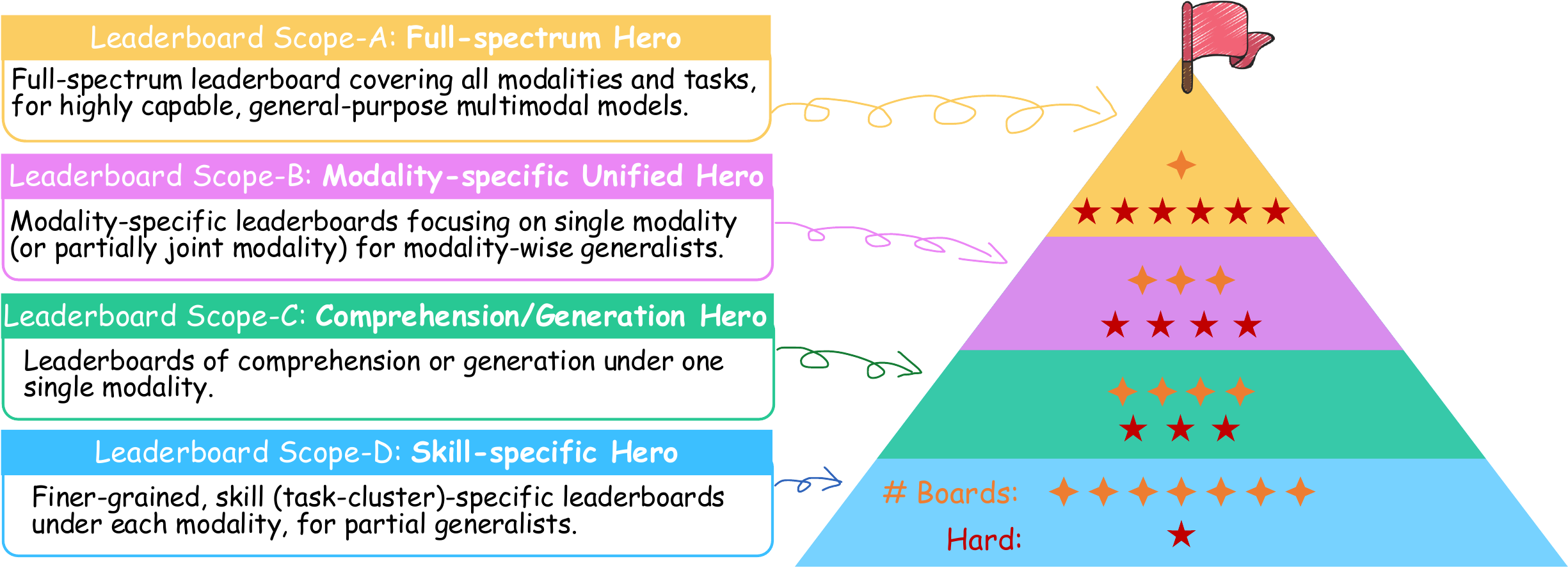}
\vspace{-2mm}
\caption{
We reorganize \textbf{General-Bench} into 4 scopes, categorized by the level of participation difficulty for practitioners.
}
\label{fig:leaderboard-scope}
\vspace{-3mm}
\end{figure*}

Figure~\ref{fig:leaderboard-scope} illustrates this design.  
Each leaderboard scope reflects a different level of difficulty, allowing practitioners to flexibly choose which leaderboard to participate in based on the capabilities of their models and the amount of resources they are willing to invest.

\vspace{-2mm}
\section{Experiments}

\vspace{-1mm}
In this section, we conduct a comprehensive evaluation on \texttt{General-Bench}, from which we gain observations and jump to some conclusions.
Note that our experiments are based on the full-spectrum leaderboard (Scope-A).

\vspace{-2mm}
\subsection{Multimodal Specialist and Generalist Systems}

\vspace{-2mm}
\paragraph{SoTA Specialist.}
For each specific task under a specific modality, we select a SoTA specialist to generate benchmark results. The selection of specialists is determined based on two criteria:  
1) their performance on each task using public benchmarks and leaderboards, i.e., they must demonstrate top performance; and  
2) whether they are widely recognized and utilized by the community.
Meanwhile, we exclude models that lack reliable open-source code or parameters (as we are unable to run our own data through them), even if such models claim to be SoTA in their own papers. 
It is important to note that the specialists we use must have undergone large-scale supervised pretraining on the corresponding tasks, enabling them to achieve SoTA performances.
In our implementation, we directly load their released parameters and perform inference on the \texttt{General-Bench} test sets. 
Table \ref{tab:task-specification-img-comp} to Table \ref{tab:task-specification-nlp} in Appendix \S\ref{Tasks-and-Skills} lists all the specialists used along with their corresponding tasks.
In total, we have 172 specialists.

\vspace{-2mm}
\paragraph{Multimodal Generalists.}
We consider a diverse set of existing popular MLLMs that are capable of handling specific or various modalities and tasks. 
This includes both open-source systems and closed-source ones (such as the OpenAI GPT series). 
For open-source models, we implement them by loading their released parameters and directly performing inference on the \texttt{General-Bench} test sets. 
For closed-source models, we utilize their APIs to access the services.
We note that, despite the release of a vast number of MLLMs in the community, due to resource constraints, we only consider a subset of MLLMs that demonstrate strong and stable capabilities and are widely recognized and utilized. 
However, our evaluation system remains open, and we encourage more MLLMs interested in our benchmarking system to participate by running their own evaluations and submitting their scores.
Table~\ref{tab:MLLM-list} summarizes all the multimodal generalists employed, including their corresponding modality support, characterized skills, parameter sizes, and backbone LLM architectures.

{
\fontsize{8.5}{9}\selectfont 
\setlength{\tabcolsep}{1.2mm}
\captionof{table}{
A complete list of (multimodal) generalists evaluated on General-Bench.
}
\vspace{-2mm}
\label{tab:MLLM-list}
\begin{longtable}{>{\centering\arraybackslash}p{0.5cm}
    p{3.2cm}
    p{3cm}
    >{\centering\arraybackslash}p{0.8cm}
    >{\centering\arraybackslash}p{5cm}
    >{\centering\arraybackslash}p{3.3cm}
    }

\toprule
\textbf{\#} & \textbf{Model} & \textbf{Backbone} & \textbf{Size} & \textbf{Modality Support} & \multicolumn{1}{c}{\textbf{Paradigm}} \\\addlinespace[3pt]
\hline
\endfirsthead
\hline
\textbf{\#} & \textbf{Model} & \textbf{Backbone} & \textbf{Size} & \textbf{Modality Support} & \multicolumn{1}{c}{\textbf{Paradigm}} \\\addlinespace[3pt]
\hline
\endhead
\hline
\endfoot

\rowcolor{deepgray} \multicolumn{6}{l}{$\bullet$ \textbf{Language-oriented (Closed/Open-sourced) Models}} \\\addlinespace[3pt]

\addlinespace[3pt]
1 & Meta-Llama-3.1-8B-Instruct \cite{abs-2302-13971} & Llama & 8B & Language & \multicolumn{1}{c}{/} \\\addlinespace[3pt]

\rowcolor{lightgray}
2 & Gemma-2-9b-it \cite{team2024gemma} & Gemma & 9B & Language & \multicolumn{1}{c}{/}\\\addlinespace[3pt]

3 & GPT-J \cite{gpt-j} & GPT-J & 6B & Language & \multicolumn{1}{c}{/} \\\addlinespace[3pt]

\rowcolor{lightgray}
4 & ChatGLM-6B \cite{glm2024chatglm} & ChatGLM & 6B & Language & \multicolumn{1}{c}{/} \\\addlinespace[3pt]

5 & Qwen2.5-7B-Instruct \cite{yang2024qwen2} & Qwen2.5 & 7B & Language & \multicolumn{1}{c}{/} \\\addlinespace[3pt]

\rowcolor{lightgray}
6 & InternLM2-Chat-7B \cite{cai2024internlm2} & InternLM2 & 7B & Language & \multicolumn{1}{c}{/} \\\addlinespace[3pt]

7 & Baichuan2-7B-Chat \cite{yang2023baichuan} & Baichuan2 & 7B & Language & \multicolumn{1}{c}{/} \\\addlinespace[3pt]

\rowcolor{lightgray}
8 & Vicuna-7b-V1.5 \cite{chiang2023vicuna} & Vicuna & 7B & Language & \multicolumn{1}{c}{/} \\\addlinespace[3pt]

9 & Falcon3-7B-Instruct \cite{falcon40b} & Falcon3 & 7B & Language & \multicolumn{1}{c}{/} \\\addlinespace[3pt]

\rowcolor{lightgray}
10 & Ministral-8B-Instruct-2410 \cite{jiang2024mixtral} & Ministral & 8B & Language & \multicolumn{1}{c}{/} \\\addlinespace[3pt]
		
11 & Yi-lightning \cite{young2024yi} & Llama & 6B & Language & \multicolumn{1}{c}{/} \\\addlinespace[3pt]

\rowcolor{lightgray}
12 & GPT-3.5-turbo \cite{chatgpt} & GPT3.5 & / & Language & \multicolumn{1}{c}{/} \\\addlinespace[3pt]

\hline
\rowcolor{deepgray} \multicolumn{6}{l}{$\bullet$ \textbf{Multimodal Close-sourced  Models}} \\\addlinespace[3pt]

1 & GPT4-V \cite{gpt4} & GPT4 & / & Language, Image & Comprehension \\\addlinespace[3pt]

\rowcolor{lightgray}
2 & GPT4-o-mini \cite{gpt4} & GPT4 & / & Language, Image & Comprehension \\\addlinespace[3pt]

3 & GPT4-o \cite{gpt4} & GPT4 & / & Language, Image & Comprehension \\\addlinespace[3pt]

\rowcolor{lightgray}
4 & GPT4-o-4096 \cite{gpt4} & GPT4 & / & Language, Image & Comprehension \\\addlinespace[3pt]

5 & ChatGPT-o-latest \cite{gpt4} & GPT4 & / & Language, Image & Comprehension \\\addlinespace[3pt]

\rowcolor{lightgray}
6 & Claude-3.5-Sonnet \cite{TheClaude3} & Claude-3.5-Sonnet & / & Language, Image & Comprehension \\\addlinespace[3pt]

7 & Claude-3.5-Opus \cite{TheClaude3} & Claude-3.5-Opus & / & Language, Image & Comprehension \\\addlinespace[3pt]

\rowcolor{lightgray}
8 & Gemini-1.5-Pro \cite{team2024gemini} & Gemini & / & Language, Image & Comprehension \\\addlinespace[3pt]

9 & Gemini-1.5-Flash \cite{team2024gemini} & Gemini & / & Language, Image & Comprehension \\\addlinespace[3pt]

\hline
\rowcolor{deepgray} \multicolumn{6}{l}{$\bullet$ \textbf{Multimodal Open-sourced Models}} \\\addlinespace[3pt]

1 & Yi-vision-v2 \cite{young2024yi} & LLaVa & 6B & Language, Image & Comprehension \\\addlinespace[3pt]

\rowcolor{lightgray}
2 & Emu2-37B \cite{sun2024generative} & LLaMA-33B & 37B & Language, Image & Comprehension+Generation \\\addlinespace[3pt]

3 & InternVL2.5-2B \cite{chen2024internvl} & internlm2\_5-1\_8b-chat & 2B & Language, Image & Comprehension \\\addlinespace[3pt]

\rowcolor{lightgray}
4 & InternVL2.5-4B \cite{chen2024internvl} & Qwen2.5-3B-Instruct & 4B & Language, Image & Comprehension \\\addlinespace[3pt]

5 & InternVL2.5-8B \cite{chen2024internvl} & internlm2\_5-7b-chat & 8B & Language, Image & Comprehension \\\addlinespace[3pt]

\rowcolor{lightgray}
6 & Mini‑InternVL-Chat-\newline2B-V1-5 \cite{gao2024mini} & InternLM2-Chat-1.8B & 2B & Language, Image & Comprehension \\\addlinespace[3pt]

7 & Mini‑InternVL‑Chat‑\newline4B‑V1‑5 \cite{gao2024mini} & Phi-3-mini-128k-instruct & 4B & Language, Image & Comprehension \\\addlinespace[3pt]

\rowcolor{lightgray}
8 & InternLM-XComposer2-VL-1.8B \cite{dong2024internlm} & InternLM2-Chat-1.8B & 1.8B & Language, Image & Comprehension \\\addlinespace[3pt]

9 & MoE-LLAVA-Phi2-2.7B-4e-384 \cite{lin2024moe} & Phi2 & 2.7B & Language, Image & Comprehension \\\addlinespace[3pt]

\rowcolor{lightgray}
10 & Monkey-10B-chat \cite{li2024monkey} & Qwev-7B & 10B & Language, Image & Comprehension \\\addlinespace[3pt]

11 & mPLUG-Owl2-LLaMA2-7b \cite{ye2024mplug} & LLaMA2-7b & 7B & Language, Image & Comprehension \\\addlinespace[3pt]

\rowcolor{lightgray}
12 & Phi-3.5-Vision-Instruct \cite{abdin2024phi} & Phi-3 Mini & 4.2B & Language, Image & Comprehension \\\addlinespace[3pt]

13 & Cambrian-1-8B \cite{tong2024cambrian} & LLaMA3-8B-Instruct & 8B & Language, Image & Comprehension \\\addlinespace[3pt]

\rowcolor{lightgray}
14 & DetGPT \cite{pi2023detgpt} & Vicuna-7B & 7B & Language, Image & Comprehension \\\addlinespace[3pt]

15 & Otter \cite{li2023otter} & LLaMA-7B & 7B & Language, image & Comprehension \\\addlinespace[3pt]

\rowcolor{lightgray}
16 & NExT-Chat \cite{zhang2023next} & LLaVA & 7B & Language, Image & Comprehension \\\addlinespace[3pt]

17 & GPT4RoI-7B \cite{zhang2023gpt4roi} & LLaMA-7B & 7B & Language, Image & Comprehension \\\addlinespace[3pt]

\rowcolor{lightgray}
18 & GLaMM \cite{rasheed2024glamm} & Vicuna-7B & 7B & Language, Image & Comprehension \\\addlinespace[3pt]

19 & Pixtral-12B \cite{agrawal2024pixtral} & Mistral-Nemo-12B & 12B & Language, Image & Comprehension \\\addlinespace[3pt]

\rowcolor{lightgray}
20 & BLIP-2 \cite{0008LSH23} & Flan T5-xl & 3B & Language, Image & Comprehension \\\addlinespace[3pt]

21 & BLIP-3 (XGen-MM) \cite{xue2024xgen} & Phi3-mini & 4B & Language, Image & Comprehension \\\addlinespace[3pt]

\rowcolor{lightgray}
22 & miniMonkey \cite{li2024monkey} & Qwev-7B & 7B & Language, Image & Comprehension \\\addlinespace[3pt]

23 & MiniGPT4-LLaMA2-7B \cite{abs-2304-10592} & LLaMA2-7B-instruct & 7B & Language, Image & Comprehension \\\addlinespace[3pt]

\rowcolor{lightgray}
24 & Show-o \cite{xie2024showo} &	Show-o & 1.3B &	Language, Image &	Comprehension+Generation \\\addlinespace[3pt]

25 & DeepSeek-VL-7B-Base \cite{lu2024deepseek} & DeepSeek-LLM-7b-base & 7B & Language, Image & Comprehension \\\addlinespace[3pt]

\rowcolor{lightgray}
26 & DeepSeek-VL-7B-Chat \cite{lu2024deepseek} & DeepSeek & 7B & Language, Image & Comprehension \\\addlinespace[3pt]

27 & LISA \cite{lai2024lisa} & LLaMA-7B & 7B & Language, Image & Comprehension \\\addlinespace[3pt]

\rowcolor{lightgray}
28 & CogVLM-Chat \cite{wang2023cogvlm} & Vicuna-v1.5-7B & 17B & Language, Image & Comprehension \\\addlinespace[3pt]

29 & ShareGPT4V-7B \cite{chen2025sharegpt4v} & Vicuna-v1.5-7B & 7B & Language, Image & Comprehension \\\addlinespace[3pt]

\rowcolor{lightgray}
30 & ShareGPT4V-13B \cite{chen2025sharegpt4v} & Vicuna-v1.5-13B & 13B & Language, Image & Comprehension \\\addlinespace[3pt]

31 & GLM-VL-Chat \cite{du2021glm} & GLM-4V & 9B & Language, Image & Comprehension \\\addlinespace[3pt]

\rowcolor{lightgray}
32 & OMG-LLaVA-InternLM20B \cite{zhang2024omg} & internlm2-7b & 7B & Language, Image & Comprehension \\\addlinespace[3pt]

33 & Idefics3-8B-Llama3 \cite{laurenccon2024building} & Llama-3.1-8B & 8B & Language, Image & Comprehension \\\addlinespace[3pt]

\rowcolor{lightgray}
34 & MiniCPM3-4B \cite{hu2024minicpm} & MiniCPM3-4B & 4B & Language, Image & Comprehension \\\addlinespace[3pt]

35 & SEED-LLaMA-13B \cite{ge2023making} & Llama2-chat-13B & 14B & Language, Image & Comprehension+Generation \\\addlinespace[3pt]

\rowcolor{lightgray}
36 & LaVIT-V2 (7B) \cite{jin2309unified} & LLaMA-7B & 7B & Language, Image & Comprehension+Generation \\\addlinespace[3pt]

37 & LM4LV \cite{zheng2024lm4lv} & LLaMA2-7B instruct & 7B & Language, Video & Generation \\\addlinespace[3pt]

\rowcolor{lightgray}
38 & CoLVA-2B \cite{zhou2025CoLVA} & Qwen2-2B & 2B & Language, Image, Video & Comprehension \\\addlinespace[3pt]

39 & CoLVA-4B \cite{zhou2025CoLVA} & Phi3-3.8B & 4.1B & Language, Image, Video & Comprehension \\\addlinespace[3pt]

\rowcolor{lightgray}
40 & Long-LLaVA-9B \cite{wang2024longllava} & Jamba-9B-Instruct & 9B & Language, Video & Comprehension \\\addlinespace[3pt]

41 & DeepSeek-VL-2-small \cite{lu2024deepseek} & DeepSeekMoE-16B & 2.8B & Language, Image & Comprehension \\\addlinespace[3pt]

\rowcolor{lightgray}
42 & DeepSeek-VL-2 \cite{lu2024deepseek} & DeepSeekMoE-27B & 4.5B & Language, Image & Comprehension \\\addlinespace[3pt]

43 & Qwen-VL-Chat \cite{bai2023qwen} & Qwen-7B & 7B & Language, Image, Video & Comprehension \\\addlinespace[3pt]

\rowcolor{lightgray}
44 & Qwen-Audio-Chat \cite{chu2023qwen} & Qwen-7B & 7B & Language, Audio & Comprehension \\\addlinespace[3pt]

45 & Qwen2-VL-7B \cite{wang2024qwen2} & Qwen2-7B & 7B & Language, Image, Video & Comprehension \\\addlinespace[3pt]

\rowcolor{lightgray}
46 & Qwen2-Audio-Instruct \cite{chu2024qwen2} & Qwen-7B & 7B & Language, Audio & Comprehension \\\addlinespace[3pt]

47 & Qwen2-VL-72B \cite{wang2024qwen2} & Qwen2-72B & 72B & Language, Image, Video & Comprehension \\\addlinespace[3pt]

\rowcolor{lightgray}
48 & LLaVA-NeXT-13B \cite{liu2024llava} & Vicuna-13B & 13B & Language, Image & Comprehension \\\addlinespace[3pt]

49 & LLaVA-NeXT-34B \cite{liu2024llava} & Nous-Hermes-2-Yi-34B & 34B & Language, Image & Comprehension \\\addlinespace[3pt]

\rowcolor{lightgray}
50 & LLaVA-One-Vision-7B \cite{li2024llavaonevision} & Qwen2-7B & 7B & Language, Image, Video & Comprehension \\\addlinespace[3pt]

51 & LLaVA-One-Vision-72B \cite{li2024llavaonevision} & Qwen2-72B & 72B & Language, Image, Video & Comprehension \\\addlinespace[3pt]

\rowcolor{lightgray}
52 & Sa2VA-8B \cite{sa2va} & InternLM2-7B & 8B & Language, Image, Video & Comprehension \\\addlinespace[3pt]

53 & Sa2VA-26B \cite{sa2va} & InternLM2-20B & 26B & Language, Image, Video & Comprehension \\\addlinespace[3pt]

\rowcolor{lightgray}
54 & InternVL-2-8B \cite{chen2024internvl} & InternLM2-7B & 8B & Language, Image, Video & Comprehension \\\addlinespace[3pt]

55 & InternVL-2.5-8B \cite{chen2024internvl} & internlm2\_5-7b-chat & 8B & Language, Image, Video & Comprehension \\\addlinespace[3pt]

\rowcolor{lightgray}
56 & InternVL-2-26B \cite{chen2024internvl} & InternLM2-20B & 26B & Language, Image, Video & Comprehension \\\addlinespace[3pt]

57 & InternVL-2.5-26B \cite{chen2024internvl} & internlm2\_5-20b-chat & 26B & Language, Image, Video & Comprehension \\\addlinespace[3pt]

\rowcolor{lightgray}
58 & Vitron-V1 \cite{fei2024vitron} & vicuna-7b-v0 & 7B & Language, Image, Video & Comprehension+Generation \\\addlinespace[3pt]

59 & Mini-Gemini \cite{li2024mini} & Nous-Hermes-2-Yi-34B & 34B & Language, Image &Comprehension+Generation \\\addlinespace[3pt]

\rowcolor{lightgray}
60 & 3D-LLM-2.1B \cite{hong20233d} & BLIP2 & 2.1B & Language, 3D & Comprehension \\\addlinespace[3pt]

61 & PointLLM-7B \cite{xu2025pointllm} & LLaMA & 7B & Language, 3D & Comrehension \\\addlinespace[3pt]

\rowcolor{lightgray}
62 & PointLLM-13B \cite{xu2025pointllm} & LLaMA & 13B & Language, 3D & Comprehension \\\addlinespace[3pt]

63 & 3D-VisTA \cite{zhu20233d} & BERT & 1.3B & Language, 3D & Comprehension \\\addlinespace[3pt]

\rowcolor{lightgray}
64 & AvatarGPT \cite{zhou2024avatargpt} & T5-large & 770M & Language, 3D & Comprehension \\\addlinespace[3pt]

65 & MotionGPT-T5 \cite{jiang2024motiongpt} & T5 & 220M & Language, 3D & Generation \\\addlinespace[3pt]

\rowcolor{lightgray}
66 & MotionGPT-LLaMA \cite{zhang2023motiongpt} & LLaMA & 13B & Language, 3D & Generation \\\addlinespace[3pt]
		
67 & LLaMA-mesh \cite{zhang2023motiongpt} & LLaMA & 7B & Language, 3D & Generation \\\addlinespace[3pt]

\rowcolor{lightgray}
68 & GAMA \cite{ghosh-etal-2024-gama} & Llama-2-7b-chat & 7B & Language, Audio & Comprehension \\\addlinespace[3pt]

69 & Pengi \cite{deshmukh2023pengi} & GPT2-base & 124M & Language, Audio & Comprehension \\\addlinespace[3pt]

\rowcolor{lightgray}
70 & WavLLM \cite{hu2024wavllm} & LLaMA-2-7B-chat & 7B & Language, Audio & Comprehension \\\addlinespace[3pt]

71 & SALMONN-7B \cite{tang2023salmonn} & Vicuna-7B & 7B & Language, Audio (Speech) & Comprehension \\\addlinespace[3pt]

\rowcolor{lightgray}
72 & SALMONN-13B \cite{tang2023salmonn} & Vicuna-13B & 13B & Language, Audio (Speech) & Comprehension \\\addlinespace[3pt]

73 & SpeechGPT-7B-com \cite{abs-2305-11000} & LLaMA-2 & 7B & Language, Audio (Speech) & Generation \\\addlinespace[3pt]

\rowcolor{lightgray}
74 & AudioGPT-GPT4 \cite{abs-2304-12995} & GPT-4 & / & Language, Audio (Speech, Sound) & Generation \\\addlinespace[3pt]

75 & AnyGPT \cite{zhan2024anygpt} & LLaMA-2-7B & 8B & Language, Image, Audio (Speech, Music) & Comprehension+Generation \\\addlinespace[3pt]

\rowcolor{lightgray}
76 & PandaGPT-13B \cite{abs-2305-16355} & Vicuna-13B-v0 & 13B & Language, Image, Video, Audio & Comprehension \\\addlinespace[3pt]

77 & ImageBind-LLM \cite{han2023imagebind} & LLama-1-7B & 7B & Language, Image, Video, Audio & Comprehension \\\addlinespace[3pt]

\rowcolor{lightgray}
78 & ModaVerse-7b-v0 \cite{wang2024modaverse} & Vicuna-7b-V0 & 7B & Language, Image, Video, Audio & Comprehension+Generation \\\addlinespace[3pt]

79 & Unified-io-2-XXL \cite{lu2024unified} & UIO-2-XXL & 6.8B & Language, Image, Video, Audio & Comprehension+Generation \\\addlinespace[3pt]

\rowcolor{lightgray}
80 & NExT-GPT-V1.5 \cite{wu2023next} & vicuna-7b-v1.5 & 7B & Language, Image, Video, Audio & Comprehension+Generation \\\addlinespace[3pt]

81 & VidAgent$^{\dag}$ \cite{abs-2303-17580} & vicuna-7b-v0 & 7B & Language, Image, Video & Comprehension+Generation \\\addlinespace[3pt]

\hline
\end{longtable}
\vspace{-4mm}
}

Note that, for VidAgent$^{\dag}$, we implement HuggingGPT as the prototype agent, and integrate InternVL-2.5-8B \cite{chen2024internvl} as video comprehension module, and integrate CogVideo \cite{abs-2205-15868} as video generation module.

\subsection{Experimental Settings}

For different models, we consistently follow the settings provided in their respective GitHub repositories, including model parameters and hyperparameters. 
We do not perform additional pre-training or fine-tuning. 
Each task and dataset comes with a predefined instruction prompt text. 
During evaluation, we use the same default prompt across all MLLMs to ensure fairness.
The inference time varies across models. Smaller models complete evaluations within a few minutes, while larger models require significantly more time. 
On pure text-based NLP tasks, model inference is highly efficient; however, on video tasks, models demand more memory and have slower inference speeds. 
Our open-source codebase supports multi-GPU distributed inference, effectively accelerating the evaluation process. 
Also, we organize personnel into multiple groups to run models in parallel, further optimizing efficiency. 
For each task, we provide predefined evaluation scripts. Once the model generates outputs, the scripts are used to evaluate performance systematically.

{
\begin{table*}[t!]
\centering
  \fontsize{7.5}{7.5}\selectfont 
  \setlength{\tabcolsep}{1.2mm}
\caption{
Performance of multimodal generalists on various image comprehension skills.
Skill full names and specific tasks are listed in Appendix $\S$ \ref{Tasks-and-Skills}.
The full performance records of more generalists are shown in Appendix $\S$ \ref{Extension on Experimental Results}.
}
\vspace{-2mm}
\label{tab:main-overall-results-image-comp}
\begin{tabular}{p{2cm} cccccccccc cc ccc}
\toprule
\rowcolor{bg-tb-heavey-vision} & \multicolumn{10}{c}{\bf \textcolor{greenCom}{Image Comprehension Skill (Avg within each \#I-C Group)}} & \multicolumn{2}{c}{\bf Task Completion} & \multicolumn{3}{c}{\bf Level Score on Image} \\ 
\rowcolor{bg-tb-heavey-vision} & \bf\#1 &\bf \#2 &\bf \#3 &\bf \#4 & \bf\#5 & \bf\#6 & \bf\#7 & \bf\#8 &\bf \#9 & \bf\#10 & 
 &  &  &  &  \\
\rowcolor{bg-tb-heavey-vision}  & \bf\#11 & \bf\#12 & \bf\#13 & \bf\#14 &\bf \#15 &\bf \#16 & \bf\#17 & \bf\#18 & \bf\#19 & \bf\#20 & 
 &  &  &  &  \\
\rowcolor{bg-tb-heavey-vision} &\bf \#21 & \bf\#22 & \bf\#23 &\bf \#24 &\bf \#25 & \bf\#26 & \bf\#27 & \bf\#28 & \bf\#29 & \bf\#30 & 
 &  &  &  &  \\
\rowcolor{bg-tb-heavey-vision} \multicolumn{1}{c}{\multirow{-5}{*}{\bf  Model}}  & \bf \#31 & \bf\#32 & \bf\#33 & \bf\#34 & \bf\#35 & \bf\#36 & \bf\#37 & \bf\#38 & \bf\#39 & \bf\#40 & 
\multicolumn{1}{c}{\multirow{-4}{*}{\parbox{1.2cm}{\#Supported Task}}} & \multicolumn{1}{c}{\multirow{-4}{*}{\parbox{1.2cm}{\#Win-over-Specialist}}} & \multicolumn{1}{c}{\multirow{-4}{*}{Level-2}}  & \multicolumn{1}{c}{\multirow{-4}{*}{Level-3}}  & \multicolumn{1}{c}{\multirow{-4}{*}{Level-4}}  \\
  \midrule

\addlinespace[4pt]

\rowcolor{bg-tb-light-vision} & 51.27 & 53.32 & 42.04 & 22.30 & 39.02 & 22.42 & 46.02 & 15.67 & 51.20 & 28.01 &  &  &  &  &  \\
\rowcolor{bg-tb-light-vision} & 36.40 & 65.15 & 43.78 & 58.90 & 63.73 & 87.84 & 58.66 & 72.25 & 34.51 & 95.70 &  &  &  &  &  \\
\rowcolor{bg-tb-light-vision} & 70.00 & 50.40 & 65.97 & 16.60 & 78.00 & 50.48 & 19.90 & 53.55 & 64.10 & 35.90 &  &  &  &  &  \\
\bottomrule
\end{tabular}%
\vspace{-4mm}
\end{table*}
}

{
\begin{table*}[t!]
\centering
  \fontsize{7.5}{8}\selectfont 
  \setlength{\tabcolsep}{1.5mm}
\caption{
Performance of part of multimodal generalists on image generation skills.
}
\vspace{-2mm}
\label{tab:main-overall-results-image-gen}
\begin{tabular}{p{2cm} cccccccc cc ccc}
\toprule
\rowcolor{bg-tb-heavey-vision} & \multicolumn{8}{c}{\bf \textcolor{blueGen}{Image Generation Skill (Avg within each \#I-G Group)}} & \multicolumn{2}{c}{\bf Task Completion} & \multicolumn{3}{c}{\bf Level Score on Image} \\ 
\rowcolor{bg-tb-heavey-vision} & \bf\#1 & \bf\#2 & \bf\#3 & \bf\#4 & \bf\#5 & \bf\#6 & \bf\#7 & \bf\#8 &  
 &  &  &  &  \\
\rowcolor{bg-tb-heavey-vision} \multicolumn{1}{c}{\multirow{-3}{*}{\bf  Model}}  & \bf \#9 & \bf\#10 & \bf\#11 & \bf\#12 & \bf\#13 & \bf\#14 & \bf\#15 &  & 
\multicolumn{1}{c}{\multirow{-2}{*}{\parbox{1.2cm}{\#Supported Task}}} & \multicolumn{1}{c}{\multirow{-2}{*}{\parbox{1.2cm}{\#Winning-Specialist}}} & \multicolumn{1}{c}{\multirow{-2}{*}{Level-2}}  & \multicolumn{1}{c}{\multirow{-2}{*}{Level-3}}  & \multicolumn{1}{c}{\multirow{-2}{*}{Level-4}}  \\

  \midrule

\rowcolor{bg-tb-light-vision} & 18.70 & 45.40 & 33.77 & 16.30 & 4.86 & 24.00 & 99.29 & 15.06 &  &  &  &  &  \\
\rowcolor{bg-tb-light-vision} \multirow{-2}{*}{\bf SoTA Specialist} & 53.16 & 16.47 & 25.33 & 43.93 & 20.35 & 67.44 & 36.11 &  & \multirow{-2}{*}{/} & \multirow{-2}{*}{/} & \multirow{-2}{*}{/} & \multirow{-2}{*}{/} & \multirow{-2}{*}{/} \\

\midrule\midrule

\multirow{2}{*}{SEED-LLaMA-14B} & 127.10 & 0.00 & 37.10 & 7.51 & 127.42 & 98.33 & 0.00 & 0.00 & \multirow{2}{*}{35 (77.8\%)} & \multirow{2}{*}{0 (0.0\%)} & \multirow{2}{*}{26.81} & \multirow{2}{*}{3.49} & \multirow{2}{*}{0.00} \\
&30.18 & 87.90 & 14.58 & 175.33 & 0.00 & 51.82 & 62.60 &  &  &  &  & & \\
\addlinespace[4pt]

\rowcolor{bg-tb-light-vision} & 93.52 & 0.00 & 34.85 & 8.53 & 101.80 & 81.95 & 0.00 & 0.00 &  &  &  &  &  \\
\bottomrule
\end{tabular}%
\end{table*}
}

\begin{table*}[t!]
\centering
\caption{
Performance of multimodal generalists on video comprehension and generation skills.
}
\label{tab:main-overall-results-video}
\fontsize{7.5}{8}\selectfont 
\setlength{\tabcolsep}{1mm}
\vspace{-2mm}
\begin{tabular}{p{2cm} cccccccccc cc ccc}
\toprule
\rowcolor{bg-tb-heavey-video} & \multicolumn{10}{c}{\bf \textcolor{greenCom}{Video Comprehension Skill (Avg within each \#V-C Group)}} & \multicolumn{2}{c}{\bf Task Completion} & \multicolumn{3}{c}{\bf Level Score on Video} \\ 
\rowcolor{bg-tb-heavey-video} & \bf\#1 & \bf\#2 & \bf\#3 & \bf\#4 & \bf\#5 & \bf\#6 & \bf\#7 & \bf\#8 & \bf\#9 & \bf\#10 & 
 &  &  &  &  \\
\rowcolor{bg-tb-heavey-video} \multicolumn{1}{c}{\multirow{-3}{*}{\bf  Model}}  & \bf \#11 & \bf\#12 & \bf\#13 & \bf\#14 & \bf\#15 & \bf\#16 & \bf\#17 & \bf\#18 & \bf\#19 & \bf\#20 & 
\multicolumn{1}{c}{\multirow{-2}{*}{\parbox{1.2cm}{\#Supported Task}}} & \multicolumn{1}{c}{\multirow{-2}{*}{\parbox{1.2cm}{\#Win-over-Specialist}}} & \multicolumn{1}{c}{\multirow{-2}{*}{Level-2}}  & \multicolumn{1}{c}{\multirow{-2}{*}{Level-3}}  & \multicolumn{1}{c}{\multirow{-2}{*}{Level-4}}  \\

\midrule

\rowcolor{bg-tb-light-video} & 37.43 & 49.64 & 21.31 & 23.06 & 81.85 & 85.43 & 54.53 & 64.83 & 40.65 & 30.80 &  &  &  &  &  \\
\rowcolor{bg-tb-light-video} \multirow{-2}{*}{\bf SoTA Specialist} & 45.84 & 13.92 & 0.14 & 48.06 & 68.96 & 63.62 & 77.02 & 75.08 & 37.20 & 44.00 & \multirow{-2}{*}{/} & \multirow{-2}{*}{/} & \multirow{-2}{*}{/} & \multirow{-2}{*}{/} & \multirow{-2}{*}{/} \\

\midrule\midrule

\multirow{2}{*}{ InternVL-2.5-8B} & 33.15 & 27.54 & 14.51 & 18.83 & 0.00 & 0.00 & 0.00 & 0.00 & 0.00 &0.00 &  &  &  &  &  \\
& 0.00 & 0.00 & 0.00 & 0.00 & 0.00 & 0.00 & 0.00 & 0.00 & 0.00 & 4.85 & \multirow{-2}{*}{55 (43.7\%)} & \multirow{-2}{*}{5 (4.0\%)} & \multirow{-2}{*}{5.76} & \multirow{-2}{*}{1.24} & \multirow{-2}{*}{0.00} \\

\addlinespace[4pt]

\rowcolor{bg-tb-light-video} & 37.03 & 32.01 & 18.71 & 21.57 & 0.00 & 0.00 & 0.00 & 0.00 & 0.00 &0.00 &  &  &  &  &  \\
\end{tabular}%
{
\fontsize{7.5}{8}\selectfont 
\setlength{\tabcolsep}{2.2mm}
\begin{tabular}{p{2cm} cccccccccc cc ccc}

\addlinespace[4pt]
\rowcolor{bg-tb-heavey-video} & \multicolumn{6}{c}{\bf \textcolor{blueGen}{Video Generation Skill (Avg within each \#V-G Group)}} & \multicolumn{2}{c}{\bf Task Completion} & \multicolumn{3}{c}{\bf Level Score on Video} \\ 

\rowcolor{bg-tb-heavey-video} \multicolumn{1}{c}{\multirow{-2}{*}{\bf  Model}}  & \bf\#1 & \bf\#2 & \bf\#3 & \bf\#4 & \bf\#5 & \bf\#6  & 
\multirow{1}{*}{\parbox{1.6cm}{\#Task-Supprt}} & \multirow{1}{*}{\parbox{1.6cm}{\#Win-Spclst}} & \multirow{1}{*}{Level-2}  & \multirow{1}{*}{Level-3}  & \multirow{1}{*}{Level-4}  \\

\midrule

\rowcolor{bg-tb-light-video} \multirow{1}{*}{\bf SoTA Specialist} & 69.09 & 55.79 & 88.94 & 62.90 & 37.79 & 51.46 & \multirow{1}{*}{/} & \multirow{1}{*}{/} & \multirow{1}{*}{/} & \multirow{1}{*}{/} & \multirow{1}{*}{/} \\

\midrule\midrule

\multirow{1}{*}{ VidAgent} & 52.42 & 47.73 & 88.84 & 63.61 & 0.00 & 0.00 & 30 (65.2\%) & 0 (0.0\%) & 25.00 & 0.00 & 0.00 \\
\addlinespace[4pt]

\rowcolor{bg-tb-light-video} \multirow{1}{*}{ LM4LV} & 0.00 & 0.00 & 0.00 & 0.00 & 25.90 & 5.93 & 8 (17.4\%) & 0 (0.0\%) & 6.74 & 0.00 & 0.00 \\
\addlinespace[4pt]

\multirow{1}{*}{ NExT-GPT-V1.5} & 26.78 & 6.72 & 130.22 & 16.03 & 0.08 & 0.06 & 40 (87.0\%) & 0 (0.0\%) & 8.34 & 0.71 & 0.00 \\
\addlinespace[4pt]

\rowcolor{bg-tb-light-video} \multirow{1}{*}{ Vitron-V1} & 36.74 & 19.32 & 116.31 & 25.09 & 0.08 & 0.06 & 40 (87.0\%) & 0 (0.0\%) & 18.72 & 3.04 & 0.00 \\

\bottomrule
\end{tabular}%
}
\vspace{-3mm}
\end{table*}

{
\begin{table*}[t!]
\centering
  \fontsize{7.5}{8}\selectfont 
  \setlength{\tabcolsep}{1mm}
\caption{
Performance of multimodal generalists on audio comprehension and generation skills.
}
\vspace{-2mm}
\label{tab:main-overall-results-audio}
\begin{tabular}{p{2.2cm} ccccccccc cc ccc}
\toprule
\rowcolor{bg-tb-heavey-audio} & \multicolumn{9}{c}{\bf \textcolor{greenCom}{Audio Comprehension Skill (Avg within each \#A-C Group)}} & \multicolumn{2}{c}{\bf Task Completion} & \multicolumn{3}{c}{\bf Level Score on Audio} \\ 

\rowcolor{bg-tb-heavey-audio} \multicolumn{1}{c}{\multirow{-2}{*}{\bf  Model}}  &  \bf\#1 & \bf\#2 & \bf\#3 & \bf\#4 & \bf\#5 & \bf\#6 & \bf\#7 & \bf\#8 & \bf\#9  & 
\multirow{1}{*}{\parbox{1.6cm}{\#Task-Supprt}} & \multirow{1}{*}{\parbox{1.6cm}{\#Win-Spclst}} & \multirow{-1}{*}{Level-2}  & \multirow{-1}{*}{Level-3}  & \multirow{-1}{*}{Level-4}  \\
  \midrule

\rowcolor{bg-tb-light-audio} \multirow{-1}{*}{\bf SoTA Specialist} & 87.27 & 79.08 & 70.62 & 79.00 & 71.87 & 62.90 & 58.70 & 77.90 & 78.07 & / & / & / & / & / \\

\midrule\midrule

\multirow{1}{*}{ Qwen-Audio-Chat} & 56.93 & 68.77 & 76.80 & 37.70 & 47.71 & 19.79 & 56.44 & 85.15 & 78.50 & 30 (100.0\%) & 6 (25.0\%) & 28.39 & 10.57 & 0.00 \\
\addlinespace[4pt]

\rowcolor{bg-tb-light-audio} \multirow{1}{*}{ Qwen2-Audio-Instruct} & 72.65 & 74.80 & 61.40 & 36.80 & 45.82 & 13.45 & 61.68 & 78.95 & 67.99 & 24 (100.0\%) & 6 (25.0\%) & 28.61 & 8.53 & 0.00 \\
\addlinespace[4pt]

\multirow{1}{*}{ GAMA} & 57.00 & 64.20 & 68.00 & 53.20 & 18.43 & 26.95 & 48.85 & 85.55 & 61.80 & 23 (95.8\%) & 4 (16.7\%) & 26.35 & 7.15 & 0.00 \\
\addlinespace[4pt]

\rowcolor{bg-tb-light-audio} \multirow{1}{*}{ Pengi} & 52.88 & 60.07 & 56.70 & 36.78 & 19.77 & 19.55 & 42.95 & 77.40 & 61.17 & 23 (95.8\%) & 1 (4.2\%) & 23.29 & 1.74 & 0.00 \\
\addlinespace[4pt]

\multirow{1}{*}{ SALMONN-13B} & 67.89 & 56.33 & 67.80 & 29.45 & 24.67 & 19.36 & 43.95 & 76.55 & 56.67 & 23 (95.8\%) & 2 (8.3\%) & 23.95 & 3.61 & 0.00 \\
\end{tabular}%
{
\fontsize{7.5}{8}\selectfont 
\setlength{\tabcolsep}{0.7mm}
\begin{tabular}{p{2cm} ccccccccccc cc ccc}

\addlinespace[2pt]

 \rowcolor{bg-tb-heavey-audio} & \multicolumn{11}{c}{\bf \textcolor{blueGen}{Audio Generation Skill (Avg within each \#A-G Group)}} & \multicolumn{2}{c}{\bf Task Completion} & \multicolumn{3}{c}{\bf Level Score on Audio} \\ 

\rowcolor{bg-tb-heavey-audio} \multicolumn{1}{c}{\multirow{-2}{*}{\bf  Model}}  &  \bf\#1 & \bf\#2 & \bf\#3 & \bf\#4 & \bf\#5 & \bf\#6 & \bf\#7 & \bf\#8 & \bf\#9 & \bf\#10 & \bf\#11 & 
\multirow{1}{*}{\parbox{1.6cm}{\#Task-Supprt}} & \multirow{1}{*}{\parbox{1.6cm}{\#Win-Spclst}} & \multirow{1}{*}{Level-2}  & \multirow{1}{*}{Level-3}  & \multirow{1}{*}{Level-4}  \\
  \midrule

\rowcolor{bg-tb-light-audio} \multirow{1}{*}{\bf SoTA Specialist} & 31.50 & 3.82 & 3.64 & 4.68 & 41.54 & 51.40 & 11.52 & 6.80 & 8.33 & 22.88 & 20.33 & \multirow{1}{*}{/} & \multirow{1}{*}{/} & \multirow{1}{*}{/} & \multirow{1}{*}{/} & \multirow{1}{*}{/} \\

\midrule\midrule

\rowcolor{bg-tb-light-audio} \multirow{1}{*}{Unified-io-2-XXL} & 18.36 & 2.03 & 5.11 & 40.52 & 16.41 & 24.31 & 16.97 & 86.23 & 94.52 & 0.25 & 2.24 & 17 (85.0\%) & 0 (0.0\%) & 25.63 & 1.01 & 0.00 \\
\addlinespace[4pt]
 
\multirow{1}{*}{Any-GPT} & 23.50 & 3.24 & 4.57 & 33.58 & 13.38 & 14.05 & 27.49 & 45.36 & 83.89 & 0.25 & 2.47 & 17 (85.0\%) & 1 (5.0\%) & 29.06 & 3.29 & 0.00 \\
\addlinespace[4pt]

\rowcolor{bg-tb-light-audio} \multirow{1}{*}{NExT-GPT-V1.5} & 13.60 & 1.15 & 4.07 & 50.51 & 34.51 & 1.35 & 12.36 & 96.70 & 99.23 & 0.25 & 7.77 & 17 (85.0\%) & 1 (5.0\%) & 25.05 & 1.34 & 0.00 \\
\addlinespace[4pt]

\multirow{1}{*}{AudioGPT} & 0.50 & 1.32 & 4.61 & 23.10 & 29.48 & 0.00 & 0.00 & 46.30 & 79.98 & 0.25 & 0.00 & 13 (65.0\%) & 1 (5.0\%) & 8.80 & 3.02 & 0.00 \\
\addlinespace[4pt]

\rowcolor{bg-tb-light-audio} \multirow{1}{*}{SpeechGPT} & 0.10 & 2.79 & 4.44 & 32.35 & 0.00 & 0.00 & 0.00 & 30.24 & 85.54 & 0.25 & 0.00 & 11 (55.0\%) & 0 (0.0\%) & 7.22 & 0.00 & 0.00 \\
\bottomrule
\end{tabular}%
}
\end{table*}
}

{
\begin{table*}[t!]
\centering
  \fontsize{7.5}{8}\selectfont 
  \setlength{\tabcolsep}{0.4mm}
\caption{
Performance of multimodal generalists on 3D comprehension and generation skills.
}
\vspace{-2mm}
\label{tab:main-overall-results-3D}
\begin{tabular}{p{2cm} ccccccccccccc cc ccc}
\toprule
\rowcolor{bg-tb-heavey-3D} & \multicolumn{13}{c}{\bf \textcolor{greenCom}{3D Comprehension Skill (Avg within each \#D-C Group)}} & \multicolumn{2}{c}{\bf Task Completion} & \multicolumn{3}{c}{\bf Level Score on 3D} \\ 

\rowcolor{bg-tb-heavey-3D} \multicolumn{1}{c}{\multirow{-2}{*}{\bf  Model}}  & \bf\#1 & \bf\#2 & \bf\#3 & \bf\#4 & \bf\#5 & \bf\#6 & \bf\#7 & \bf\#8 & \bf\#9 & \bf\#10 & \bf\#11 & \bf\#12 & \bf\#13 & 
\multirow{1}{*}{\parbox{1.6cm}{\#Task-Supprt}} & \multirow{1}{*}{\parbox{1.6cm}{\#Win-Spclst}} & \multirow{-1}{*}{Level-2}  & \multirow{-1}{*}{Level-3}  & \multirow{-1}{*}{Level-4}  \\
  \midrule

\rowcolor{bg-tb-light-3D} \multirow{1}{*}{\bf SoTA Specialist} & 96.24 & 98.35 & 97.78 & 78.50 & 70.02 & 81.20 & 55.00 & 88.28 & 75.20 & 9.96 & 68.52 & 47.14 & 22.30 & \multirow{1}{*}{/} & \multirow{1}{*}{/} & \multirow{1}{*}{/} & \multirow{1}{*}{/} & \multirow{1}{*}{/} \\

\midrule\midrule

\multirow{1}{*}{3D-VisTA} & 0.00 & 0.00 & 0.00 & 0.00 & 0.00 & 0.00 & 0.00 & 0.00 & 0.00 & 0.00 & 0.00 & 46.37 & 0.00 & 7 (23.3\%) & 2 (6.7\%) & 5.41 & 1.07 & 0.00 \\
\addlinespace[2pt]

\rowcolor{bg-tb-light-3D} \multirow{1}{*}{PointLLM-7B} & 46.16 & 7.50 & 72.86 & 0.00 & 0.00 & 0.00 & 0.00 & 0.00 &0.00 &  0.00 & 0.00 & 0.00 & 0.00 & 8 (26.7\%) & 0 (0.0\%) & 6.53 & 0.00 & 0.00 \\
\addlinespace[2pt]

\multirow{1}{*}{PointLLM-13B} & 48.79 & 10.00 & 78.14 & 0.00 & 0.00 & 0.00 & 0.00 & 0.00 & 0.00 & 0.00 & 0.00 & 0.00 &0.00 &  9 (30.0\%) & 0 (0.0\%) & 7.00 & 0.00 & 0.00 \\
\addlinespace[2pt]

\rowcolor{bg-tb-light-3D} \multirow{1}{*}{3D-LLM} & 0.00 & 0.00 & 0.00 & 0.00 & 0.00 & 0.00 & 0.00 & 0.00 & 0.00 & 0.00 & 0.00 & 46.34 & 0.00 & 7 (23.3\%) & 1 (3.3\%) & 5.41 & 1.38 & 0.00 \\
\addlinespace[2pt]

\multirow{1}{*}{AvatarGPT} & 0.00 &  0.00 & 0.00 & 0.00 & 0.00 & 0.00 & 0.00 & 0.00 & 0.00 & 0.00 & 0.00 & 0.00 & 12.70 & 1 (3.3\%) & 0 (0.0\%) & 0.21 & 0.21 & 0.00 \\
\addlinespace[2pt]

\Xhline{1.5pt}
\end{tabular}%
{
\fontsize{7.5}{8}\selectfont 
\setlength{\tabcolsep}{0.9mm}
\begin{tabular}{p{2.2cm} ccccccccc cc ccc}

\addlinespace[2pt]

 \rowcolor{bg-tb-heavey-3D} & \multicolumn{9}{c}{\bf \textcolor{blueGen}{3D Generation Skill (Avg within each \#D-G Group)}} & \multicolumn{2}{c}{\bf Task Completion} & \multicolumn{3}{c}{\bf Level Score on 3D} \\ 

\rowcolor{bg-tb-heavey-3D} \multicolumn{1}{c}{\multirow{-2}{*}{\bf  Model}}  & \bf\#1 & \bf\#2 & \bf\#3 & \bf\#4 & \bf\#5 & \bf\#6 & \bf\#7 & \bf\#8 & \bf\#9  & 
\multirow{1}{*}{\parbox{1.6cm}{\#Task-Supprt}} & \multirow{1}{*}{\parbox{1.6cm}{\#Win-Spclst}} & \multirow{1}{*}{Level-2}  & \multirow{1}{*}{Level-3}  & \multirow{1}{*}{Level-4}  \\
  \midrule

\rowcolor{bg-tb-light-3D} \multirow{1}{*}{\bf SoTA Specialist} & 0.22 & 7.12E-5 & 24.42 & 25.69 & 78.06 & 83.64 & 6540.02 & 6540.02 & 0.23 & \multirow{1}{*}{/} & \multirow{1}{*}{/} & \multirow{1}{*}{/} & \multirow{1}{*}{/} & \multirow{1}{*}{/} \\

\midrule\midrule

\multirow{1}{*}{MotionGPT-T5} & 0.00 & 0.00 & 0.00 & 0.00 & 0.00 & 0.00 & 0.00 & 0.00 & 0.51 & 1 (4.5\%) & 0 (0.0\%) & 0.00 & 0.00 & 0.00 \\
\addlinespace[2pt]

\rowcolor{bg-tb-light-3D} \multirow{1}{*}{MotionGPT-LLaMA} & 0.00 & 0.00 & 0.00 & 0.00 & 0.00 & 0.00 & 0.00 & 0.00 & 0.60 & 1 (4.5\%) & 0 (0.0\%) & 0.00 & 0.00 & 0.00 \\
\addlinespace[2pt]

\multirow{1}{*}{LLaMA-Mesh} & 0.00 & 0.00 & 0.00 & 17.55 & 0.00 & 0.00 & 0.00 & 0.00 & 0.00 & 1 (4.5\%) & 0 (0.0\%) & 1.60 & 0.00 & 0.00 \\
\addlinespace[2pt]

\bottomrule
\end{tabular}%
}
\vspace{-3mm}
\end{table*}
}

{
\begin{table*}[t!]
\centering
  \fontsize{7.5}{8}\selectfont 
  \setlength{\tabcolsep}{1.4mm}
\caption{
Performance of generalists on language-related (NLP) skills.
}
\vspace{-2mm}
\label{tab:main-overall-results-nlp}
\begin{tabular}{p{2.2cm} ccccccccccc cc c}
\toprule
\rowcolor{bg-tb-heavey-nlp} & \multicolumn{11}{c}{\bf \textcolor{redNLP}{Language Skill (Avg within each \#L Group)}} & \multicolumn{2}{c}{\bf Task Completion} & \multicolumn{1}{c}{\bf Level Score} \\ 
\rowcolor{bg-tb-heavey-nlp} & \bf  \#1 & \bf\#2 & \bf\#3 & \bf\#4 & \bf\#5 & \bf\#6 & \bf\#7 & \bf\#8 & \bf\#9 & \bf\#10 & \bf\#11 & 
 &  &  \\
\rowcolor{bg-tb-heavey-nlp} \multicolumn{1}{c}{\multirow{-3}{*}{\bf  Model}}  & \bf\#12 & \bf\#13 & \bf\#14 & \bf\#15 & \bf\#16 & \bf\#17 & \bf\#18 & \bf\#19 & \bf\#20 & \bf\#21  & \bf  \#22 & 
\multicolumn{1}{c}{\multirow{-2}{*}{\parbox{1.2cm}{\#Supported Task}}} & \multicolumn{1}{c}{\multirow{-2}{*}{\parbox{1.2cm}{\#Win-over-Specialist}}} & \multirow{-2}{*}{Level-5}  \\

  \midrule

\rowcolor{bg-tb-light-nlp} & 62.62 	&	86.23 	&	76.78 	&	71.00 	&	58.02 	&	62.80 	&	75.11 	&	77.84 	&	79.70 	&	71.91 	&	28.27  & \multirow{2}{*}{/} & \multirow{2}{*}{/} & \multirow{2}{*}{/}  \\
\rowcolor{bg-tb-light-nlp} \multirow{-2}{*}{\bf SoTA Specialist} & 86.95 	&	0.31 	&	94.40 	&	91.41 	&	86.05 	&	86.03 	&	84.72 	&	83.67 	&	58.61 	&	77.73 	&	92.38  &  &  &  \\

\midrule\midrule

\multirow{2}{*}{\parbox{2cm}{Meta-Llama-3.1-8B-Instruct}} & 39.75 	&	56.76 	&	54.21 	&	60.52 	&	20.01 	&	37.17 	&	36.23 	&	29.12 	&	53.23 	&	44.49 	&	14.80  & \multirow{2}{*}{113 (98.3\%)} & \multirow{2}{*}{0 (0.0\%)} & \multirow{2}{*}{0.00} \\
 &  45.34 	&	7.95 	&	76.40 	&	51.80 	&	65.90 	&	41.10 	&	24.49 	&	30.70 	&	8.08 	&	32.40 	&	54.35  &  &  &  \\
\addlinespace[4pt]

\rowcolor{bg-tb-light-nlp} & 28.97 	&	33.24 	&	37.24 	&	46.10 	&	19.39 	&	27.84 	&	18.85 	&	35.88 	&	27.85 	&	38.51 	&	13.93   &   &  &  \\
\bottomrule
\end{tabular}%
\vspace{-3mm}
\end{table*}
}

\vspace{-2mm}
\subsection{Overall Evaluation Results}

\vspace{-2mm}
We note that all the generalists run the evaluation on our \texttt{General-Bench} data set under a zero-shot setting.
The overall results of part of the models on image comprehension and generation are presented in Table~\ref{tab:main-overall-results-image-comp} and Table~\ref{tab:main-overall-results-image-gen}, respectively;
video results are shown in Table~\ref{tab:main-overall-results-video};
audio results are shown in Table~\ref{tab:main-overall-results-audio};
3D results are shown in Table~\ref{tab:main-overall-results-3D};
The results of all generalists on NLP tasks are shown in Table~\ref{tab:main-overall-results-nlp}.
The complete performing scores of all MLLMs across all tasks and datasets are presented in Appendix $\S$\ref{Extension on Experimental Results}.
Overall, we have the following observations.

\textbf{Observation-1: Lack of task support.}
From these results, the first observation is that the vast majority of MLLMs exhibit a lack of support for a wide range of tasks in our benchmarks.
Even models like OpenAI's GPT-4V and GPT-4o, which achieve top rankings on many existing MLLM benchmarks and leaderboards \cite{SEED-Bench-abs-2307-16125,MMBench-LiuDZLZZYWHLCL24}, fail to demonstrate satisfactory task support on our benchmark. Specifically, GPT-4V and GPT-4o support only 177 out of 271 image comprehension tasks (65.1\%).
Among open-source models, InternVL2.5-8B achieves a task support rate of 71\% for image comprehension tasks, outperforming GPT-4V and GPT-4o. 
For other modalities—such as video, audio, and 3D—the task-supporting rates are much less.
Only Vitron-V1 supports over 90\% of image tasks, and Sa2VA-8B achieves 72.2\% supporting rate in the video comprehension group.
This highlights a pervasive issue: current MLLMs require significant improvements in their architectural design to support as many tasks as possible.

\textbf{Observation-2: Few generalists surpass the SoTA specialist.}
Also, we can notice that there are few models capable of surpassing the SoTA generalist.
Overall, the tasks and skills that various MLLMs can surpass the SoTA specialists are quite few.
As seen, closed-sourced models (e.g., GPT-4V, GPT-4o, Gemini-1.5, and Claude-3.5) have the highest winning rate, with over 30\% 
The best open-sourced Qwen2-VL-72B achieves a rate of 36.4\% image comprehension by surpassing SoTA specialists. 
In other modalities such as video, audio, 3D, and language, the chances to surpass SoTA specialists are much lower.
If an MLLM cannot outperform the SoTA specialist, it implies that the foundational conditions of cross-task/ability synergy for these MLLMs to become multimodal generalists are not met.

\textbf{Observation-3: Focus more on content comprehension than supporting generation.}
For instance, GPT-4V and GPT-4o achieve better results than the SoTA specialist in certain skills within image comprehension tasks, and this improvement is significantly more pronounced than that of other models.
However, GPT-4V and GPT-4o are limited to image comprehension tasks and provide zero support for image generation tasks. 
It is thus evident that GPT-4V and GPT-4o are not well-rounded multimodal generalists.\footnote{
It would thus be more rational to claim the current OpenAI GPT-4V/4o series as partial generalists, or visual generalists.
}
This trend becomes even more evident in other modalities. 
A significantly higher number of MLLMs support multimodal understanding compared to those supporting multimodal generation. 
Furthermore, the rate at which MLLMs surpass SoTA specialists in multimodal understanding benchmarks is much higher than in multimodal generation benchmarks.
We emphasize that this imbalance reflects a critical limitation in the capability building of current multimodal generalists.

\textbf{Observation-4: Insufficient support for all modalities.}  
We also found that many MLLMs are unable to support all modalities simultaneously. Moreover, the vast majority of existing MLLMs are predominantly focused on understanding or generating image-based modalities. 
In contrast, much less attention has been devoted to video, audio, and 3D modalities (attention: image $>$ video $>$ 3D $>$ audio), with relatively few multimodal generalists addressing these areas.
Most MLLMs, including the strongest ones, primarily handle image and language tasks, offering little to no support for other modalities. 
The completeness of support across various modalities and functionalities is insufficient for existing MLLMs to qualify as true multimodal generalists.
We emphasize that to be considered a multimodal generalist, a model must be capable of understanding and generating signals from as many modalities as possible simultaneously.

\textbf{Observation-5: Multimodality does NOT really enhance language.} 
The ideal multimodal generalists should enable mutual enhancement across modalities. 
Unfortunately, our experimental results (as shown in Table~\ref{tab:main-overall-results-nlp}) reveal that none of the current MLLMs provide any improvements in NLP tasks.
Although various MLLMs achieve certain scores on NLP tasks, none of them surpass the performance of SoTA specialists in NLP. 
Furthermore, the performance gap between MLLMs and SoTA specialists in NLP tasks is larger than the gap observed in other modalities.
While certain relevant research suggests that models, such as Vicuna, Qwen2, and LLaMA, trained with multimodal data (e.g., images) can also improve NLP tasks, such improvement has not yet enabled models to outperform SoTA NLP specialists on core language tasks.
Our large-scale evaluation shows they still fall short of outperforming fine-tuned language specialists. 
We hypothesize that existing MLLMs, despite utilizing language-centered LLMs as their core, have significantly weakened their language capabilities due to an excessive focus on training and fine-tuning on non-language modalities. 
This trade-off not only undermines their language understanding but also fails to leverage multimodal information to enhance language-related tasks.

{
\fontsize{8.5}{11}\selectfont 
\setlength{\tabcolsep}{1.5mm}
\captionof{table}{
Leaderboard of multimodal generalists (MLLMs) at level-2.
}
\vspace{-2mm}
\label{tab:ranking-level-2}
\begin{longtable}{p{4.2cm} cccccccc}
\hline
\multirow{2}{*}{\textbf{Model}} & \multirow{2}{*}{\textbf{Modality}} &\multirow{2}{*}{\textbf{Paradigm}} & \multicolumn{5}{c}{\textbf{Level 2 Score}} & \multirow{2}{*}{\textbf{Ranking}} \\
\cmidrule{4-8}
& & & \textbf{of Image}	& \textbf{of Video}	& 	\textbf{of Audio}	& 	\textbf{of 3D}	& \textbf{of Overall} & \\
\hline
\endfirsthead
\hline
\multirow{2}{*}{\textbf{Model}} & \multirow{2}{*}{\textbf{Modality}} &\multirow{2}{*}{\textbf{Paradigm}} & \multicolumn{5}{c}{\textbf{Level 2 Score}} & \multirow{2}{*}{\textbf{Ranking}} \\
\cmidrule{4-8}
& & & \textbf{of Image}	& \textbf{of Video}	& 	\textbf{of Audio}	& 	\textbf{of 3D}	& \textbf{of Overall} & \\
\hline
\endhead
\hline
\endfoot
\hline
\endlastfoot

\rowcolor{bg-tb-light-video} Unified-io-2-XXL & \languagelogo \imagelogo \videologo \audiologo & \textcolor{greenCom}{C}+\textcolor{blueGen}{G} & 20.62  & 8.56  & 25.63  & 0.00  & 13.70  & 1\championlogo \\
AnyGPT & \languagelogo \imagelogo \audiologo & \textcolor{greenCom}{C}+\textcolor{blueGen}{G} & 23.10  & 0.00  & 29.06  & 0.00  & 13.04  & 2\silverlogo \\
\rowcolor{bg-tb-light-video} NExT-GPT-V1.5 & \languagelogo \imagelogo \videologo \audiologo & \textcolor{greenCom}{C}+\textcolor{blueGen}{G} & 18.69  & 8.34  & 25.05  & 0.00  & 13.02  & 3\bronzelogo \\
ImageBind-LLM & \languagelogo \imagelogo \videologo \audiologo & \textcolor{greenCom}{C} & 19.54  & 12.54  & 17.52  & 0.00  & 12.40  & 4 \\
\rowcolor{bg-tb-light-video} ModaVerse-7b-v0 & \languagelogo \imagelogo \videologo \audiologo & \textcolor{greenCom}{C}+\textcolor{blueGen}{G} & 15.56  & 7.32  & 26.10  & 0.00  & 12.25  & 5 \\
Vitron-V1 & \languagelogo \imagelogo \videologo & \textcolor{greenCom}{C}+\textcolor{blueGen}{G} & 30.13  & 18.72  & 0.00  & 0.00  & 12.21  & 6 \\
\rowcolor{bg-tb-light-video} PandaGPT-13B & \languagelogo \imagelogo \videologo \audiologo & \textcolor{greenCom}{C} & 20.78  & 9.34  & 16.98  & 0.00  & 11.78  & 7 \\
VidAgent & \languagelogo \imagelogo \videologo & \textcolor{greenCom}{C}+\textcolor{blueGen}{G} & 18.21  & 25.00  & 0.00  & 0.00  & 10.80  & 8 \\
\rowcolor{bg-tb-light-video} InternVL2\_5-8B & \languagelogo \imagelogo & \textcolor{greenCom}{C} & 25.20  & 8.44  & 0.00  & 0.00  & 8.41  & 9 \\
Emu2-37B & \languagelogo \imagelogo & \textcolor{greenCom}{C}+\textcolor{blueGen}{G} & 30.90  & 0.00  & 0.00  & 0.00  & 7.73  & 10 \\
\cdashline{1-9}
\rowcolor{bg-tb-light-video} Sa2VA-26B & \languagelogo \imagelogo \videologo & \textcolor{greenCom}{C} & 21.88  & 8.81  & 0.00  & 0.00  & 7.67  & 11 \\
LaVIT-V2 (7B) & \languagelogo \imagelogo & \textcolor{greenCom}{C}+\textcolor{blueGen}{G} & 29.50  & 0.00  & 0.00  & 0.00  & 7.38  & 12 \\
\rowcolor{bg-tb-light-video} LLaVA-One-Vision-72B & \languagelogo \imagelogo \videologo & \textcolor{greenCom}{C} & 23.12  & 5.83  & 0.00  & 0.00  & 7.24  & 13 \\
Qwen2-Audio-Instruct & \languagelogo \audiologo & \textcolor{greenCom}{C} & 0.00  & 0.00  & 28.61  & 0.00  & 7.15  & 14 \\
\rowcolor{bg-tb-light-video} Qwen-Audio-Chat & \languagelogo \audiologo & \textcolor{greenCom}{C} & 0.00  & 0.00  & 28.39  & 0.00  & 7.10  & 15 \\
Mini-Gemini & \languagelogo \imagelogo & \textcolor{greenCom}{C}+\textcolor{blueGen}{G} & 27.90 & 0.00 & 0.00 & 0.00 & 6.975 & 16 \\
\rowcolor{bg-tb-light-video} SEED-LLaMA-13B & \languagelogo \imagelogo & \textcolor{greenCom}{C}+\textcolor{blueGen}{G} & 26.81  & 0.00  & 0.00  & 0.00  & 6.70  & 17 \\
GAMA & \languagelogo \audiologo & \textcolor{greenCom}{C} & 0.00  & 0.00  & 26.35  & 0.00  & 6.59  & 18 \\
\rowcolor{bg-tb-light-video} Qwen2-VL-72B & \languagelogo \imagelogo \videologo & \textcolor{greenCom}{C} & 19.41  & 6.89  & 0.00  & 0.00  & 6.58  & 19 \\
Sa2VA-8B & \languagelogo \imagelogo \videologo & \textcolor{greenCom}{C} & 17.33  & 8.31  & 0.00  & 0.00  & 6.41  & 20 \\
\rowcolor{bg-tb-light-video} InternVL-2.5-26B & \languagelogo \imagelogo \videologo & \textcolor{greenCom}{C} & 18.73  & 6.70  & 0.00  & 0.00  & 6.36  & 21 \\
Qwen2-VL-7B & \languagelogo \imagelogo \videologo & \textcolor{greenCom}{C} & 18.42  & 6.00  & 0.00  & 0.00  & 6.11  & 22 \\
\rowcolor{bg-tb-light-video} InternVL2\_5-4B & \languagelogo \imagelogo & \textcolor{greenCom}{C} & 24.41  & 0.00  & 0.00  & 0.00  & 6.10  & 23 \\
SALMONN-13B & \languagelogo \audiologo & \textcolor{greenCom}{C} & 0.00  & 0.00  & 23.95  & 0.00  & 5.99  & 24 \\
\rowcolor{bg-tb-light-video} InternVL-2-26B & \languagelogo \imagelogo \videologo & \textcolor{greenCom}{C} & 17.55  & 6.36  & 0.00  & 0.00  & 5.98  & 25 \\
WavLLM & \languagelogo \audiologo & \textcolor{greenCom}{C} & 0.00  & 0.00  & 23.49  & 0.00  & 5.87  & 26 \\
\rowcolor{bg-tb-light-video} Monkey-10B-chat & \languagelogo \imagelogo & \textcolor{greenCom}{C} & 23.51  & 0.00  & 0.00  & 0.00  & 5.87  & 27 \\
InternVL2\_5-2B & \languagelogo \imagelogo & \textcolor{greenCom}{C} & 23.32  & 0.00  & 0.00  & 0.00  & 5.83  & 28 \\
\rowcolor{bg-tb-light-video} Pengi & \languagelogo \audiologo & \textcolor{greenCom}{C} & 0.00  & 0.00  & 23.29  & 0.00  & 5.82  & 29 \\
LLaVA-One-Vision-7B & \languagelogo \imagelogo \videologo & \textcolor{greenCom}{C} & 18.32  & 4.34  & 0.00  & 0.00  & 5.67  & 30 \\
\rowcolor{bg-tb-light-video} SALMONN-7B & \languagelogo \audiologo & \textcolor{greenCom}{C} & 0.00  & 0.00  & 21.09  & 0.00  & 5.27  & 31 \\
InternVL-2.5-8B & \languagelogo \imagelogo \videologo & \textcolor{greenCom}{C} & 14.70  & 5.76  & 0.00  & 0.00  & 5.12  & 32 \\
\rowcolor{bg-tb-light-video} DeepSeek-VL-7B-Chat & \languagelogo \imagelogo & \textcolor{greenCom}{C} & 19.89  & 0.00  & 0.00  & 0.00  & 4.97  & 33 \\
InternVL-2-8B & \languagelogo \imagelogo \videologo & \textcolor{greenCom}{C} & 14.06  & 5.64  & 0.00  & 0.00  & 4.93  & 34 \\
\rowcolor{bg-tb-light-video} GPT4-o & \languagelogo \imagelogo & \textcolor{greenCom}{C} & 19.67  & 0.00  & 0.00  & 0.00  & 4.92  & 35 \\
GPT4-o-4096 & \languagelogo \imagelogo & \textcolor{greenCom}{C} & 19.68  & 0.00  & 0.00  & 0.00  & 4.92  & 36 \\
\rowcolor{bg-tb-light-video} Gemini-1.5-Pro & \languagelogo \imagelogo & \textcolor{greenCom}{C} & 19.67  & 0.00  & 0.00  & 0.00  & 4.92  & 37 \\
Claude-3.5-Sonnet & \languagelogo \imagelogo & \textcolor{greenCom}{C} & 19.38  & 0.00  & 0.00  & 0.00  & 4.85  & 38 \\
\rowcolor{bg-tb-light-video} Claude-3.5-Opus & \languagelogo \imagelogo & \textcolor{greenCom}{C} & 19.00  & 0.00  & 0.00  & 0.00  & 4.75  & 39 \\
chatgpt4-o-latest & \languagelogo \imagelogo & \textcolor{greenCom}{C} & 18.98  & 0.00  & 0.00  & 0.00  & 4.74  & 40 \\
\rowcolor{bg-tb-light-video} Gemini-1.5-Flash & \languagelogo \imagelogo & \textcolor{greenCom}{C} & 18.54  & 0.00  & 0.00  & 0.00  & 4.64  & 41 \\
CoLVA-4B & \languagelogo \imagelogo \videologo & \textcolor{greenCom}{C} & 13.59  & 4.78  & 0.00  & 0.00  & 4.59  & 42 \\
\rowcolor{bg-tb-light-video} GPT4-V & \languagelogo \imagelogo & \textcolor{greenCom}{C} & 18.16  & 0.00  & 0.00  & 0.00  & 4.54  & 43 \\
GPT4-o-mini & \languagelogo \imagelogo & \textcolor{greenCom}{C} & 17.79  & 0.00  & 0.00  & 0.00  & 4.45  & 44 \\
\rowcolor{bg-tb-light-video} GLM-VL-Chat & \languagelogo \imagelogo & \textcolor{greenCom}{C} & 17.00  & 0.00  & 0.00  & 0.00  & 4.25  & 45 \\
Idefics3-8B-Llama3 & \languagelogo \imagelogo & \textcolor{greenCom}{C} & 16.71  & 0.00  & 0.00  & 0.00  & 4.18  & 46 \\
\rowcolor{bg-tb-light-video} LLaVA-NeXT-34B & \languagelogo \imagelogo & \textcolor{greenCom}{C} & 16.58  & 0.00  & 0.00  & 0.00  & 4.15  & 47 \\
Phi-3.5-Vision-Instruct & \languagelogo \imagelogo & \textcolor{greenCom}{C} & 16.46  & 0.00  & 0.00  & 0.00  & 4.12  & 48 \\
\rowcolor{bg-tb-light-video} MiniCPM3-4B & \languagelogo \imagelogo & \textcolor{greenCom}{C} & 16.46  & 0.00  & 0.00  & 0.00  & 4.12  & 49 \\
CogVLM-Chat & \languagelogo \imagelogo & \textcolor{greenCom}{C} & 16.31  & 0.00  & 0.00  & 0.00  & 4.08  & 50 \\
\rowcolor{bg-tb-light-video} CoLVA-2B & \languagelogo \imagelogo \videologo & \textcolor{greenCom}{C} & 11.73  & 4.47  & 0.00  & 0.00  & 4.05  & 51 \\
InternVL-Chat-V1-5 & \languagelogo \imagelogo & \textcolor{greenCom}{C} & 16.16  & 0.00  & 0.00  & 0.00  & 4.04  & 52 \\
\rowcolor{bg-tb-light-video} DetGPT & \languagelogo \imagelogo & \textcolor{greenCom}{C} & 16.05  & 0.00  & 0.00  & 0.00  & 4.01  & 53 \\
BLIP-3 (XGen-MM) & \languagelogo \imagelogo & \textcolor{greenCom}{C} & 15.40  & 0.00  & 0.00  & 0.00  & 3.85  & 54 \\
\rowcolor{bg-tb-light-video} LLaVA-NeXT-13B & \languagelogo \imagelogo & \textcolor{greenCom}{C} & 15.11  & 0.00  & 0.00  & 0.00  & 3.78  & 55 \\
Pixtral-12B & \languagelogo \imagelogo & \textcolor{greenCom}{C} & 14.74  & 0.00  & 0.00  & 0.00  & 3.69  & 56 \\
\rowcolor{bg-tb-light-video} ShareGPT4V-13B & \languagelogo \imagelogo & \textcolor{greenCom}{C} & 14.72  & 0.00  & 0.00  & 0.00  & 3.68  & 57 \\
Yi-vision-v2 & \languagelogo \imagelogo & \textcolor{greenCom}{C} & 14.61  & 0.00  & 0.00  & 0.00  & 3.65  & 58 \\
\rowcolor{bg-tb-light-video} Qwen-VL-Chat & \languagelogo \imagelogo \videologo & \textcolor{greenCom}{C} & 13.91  & 5.34  & 0.00  & 0.00  & 3.48  & 59 \\
ShareGPT4V-7B & \languagelogo \imagelogo & \textcolor{greenCom}{C} & 13.78  & 0.00  & 0.00  & 0.00  & 3.45  & 60 \\
\rowcolor{bg-tb-light-video} Mini‑InternVL‑Chat‑4B‑V1‑5 & \languagelogo \imagelogo & \textcolor{greenCom}{C} & 13.53  & 0.00  & 0.00  & 0.00  & 3.38  & 61 \\
InternLM-XComposer2-VL-1.8B & \languagelogo \imagelogo & \textcolor{greenCom}{C} & 13.31  & 0.00  & 0.00  & 0.00  & 3.33  & 62 \\
\rowcolor{bg-tb-light-video} DeepSeek-VL-7B-Base & \languagelogo \imagelogo & \textcolor{greenCom}{C} & 13.13  & 0.00  & 0.00  & 0.00  & 3.28  & 63 \\
MiniGPT4-LLaMA2-7B & \languagelogo \imagelogo & \textcolor{greenCom}{C} & 12.89  & 0.00  & 0.00  & 0.00  & 3.22  & 64 \\
\rowcolor{bg-tb-light-video} MoE-LLAVA-Phi2-2.7B-4e-384 & \languagelogo \imagelogo & \textcolor{greenCom}{C} & 12.55  & 0.00  & 0.00  & 0.00  & 3.14  & 65 \\
mPLUG-Owl2-LLaMA2-7b & \languagelogo \imagelogo & \textcolor{greenCom}{C} & 12.21  & 0.00  & 0.00  & 0.00  & 3.05  & 66 \\
\rowcolor{bg-tb-light-video} Cambrian-1-8B & \languagelogo \imagelogo & \textcolor{greenCom}{C} & 11.76  & 0.00  & 0.00  & 0.00  & 2.94  & 67 \\
BLIP2 & \languagelogo \imagelogo & \textcolor{greenCom}{C} & 11.65  & 0.00  & 0.00  & 0.00  & 2.91  & 68 \\
\rowcolor{bg-tb-light-video} miniMonkey & \languagelogo \imagelogo & \textcolor{greenCom}{C} & 11.31  & 0.00  & 0.00  & 0.00  & 2.83  & 69 \\
NExT-Chat & \languagelogo \imagelogo & \textcolor{greenCom}{C} & 10.65  & 0.00  & 0.00  & 0.00  & 2.66  & 70 \\
\rowcolor{bg-tb-light-video} Audio-GPT4 & \languagelogo \audiologo & \textcolor{blueGen}{G} & 0.00  & 0.00  & 8.80  & 0.00  & 2.20  & 71 \\
GPT4RoI-7B & \languagelogo \imagelogo & \textcolor{greenCom}{C} & 8.49  & 0.00  & 0.00  & 0.00  & 2.12  & 72 \\
\rowcolor{bg-tb-light-video} Show-o & \languagelogo \imagelogo & \textcolor{greenCom}{C}+\textcolor{blueGen}{G} & 7.78 & 0.00 & 0.00 & 0.00 & 1.95 & 73 \\
SpeechGPT-7B-com & \languagelogo \audiologo & \textcolor{blueGen}{G} & 0.00  & 0.00  & 7.22  & 0.00  & 1.81  & 74 \\
\rowcolor{bg-tb-light-video} PointLLM-13B & \languagelogo \dlogo & \textcolor{greenCom}{C} & 0.00  & 0.00  & 0.00  & 7.00  & 1.75  & 75 \\
LM4LV & \languagelogo \videologo & \textcolor{blueGen}{G} & 0.00  & 6.74  & 0.00  & 0.00  & 1.69  & 76 \\
\rowcolor{bg-tb-light-video} PointLLM-7B & \languagelogo \dlogo & \textcolor{greenCom}{C} & 0.00  & 0.00  & 0.00  & 6.53  & 1.63  & 77 \\
Long-LLaVA-9B & \languagelogo \videologo & \textcolor{greenCom}{C} & 10.23  & 5.84  & 0.00  & 0.00  & 1.46  & 78 \\
\rowcolor{bg-tb-light-video} 3D-VisTA & \languagelogo \dlogo & \textcolor{greenCom}{C} & 0.00  & 0.00  & 0.00  & 5.41  & 1.35  & 79 \\
3D-LLM-2.1B & \languagelogo \dlogo & \textcolor{greenCom}{C} & 0.00  & 0.00  & 0.00  & 5.41  & 1.35  & 80 \\
\rowcolor{bg-tb-light-video} OMG-LLaVA-InternLM20B & \languagelogo \imagelogo & \textcolor{greenCom}{C} & 4.56  & 0.00  & 0.00  & 0.00  & 1.14  & 81 \\
DeepSeek-VL-2 & \languagelogo \imagelogo & \textcolor{greenCom}{C} & 19.21  & 3.98  & 0.00  & 0.00  & 1.00  & 82 \\
\rowcolor{bg-tb-light-video} DeepSeek-VL-2-small & \languagelogo \imagelogo & \textcolor{greenCom}{C} & 17.40  & 3.64  & 0.00  & 0.00  & 0.91  & 83 \\
Otter & \languagelogo \imagelogo & \textcolor{greenCom}{C} & 3.15  & 0.00  & 0.00  & 0.00  & 0.79  & 84 \\
\rowcolor{bg-tb-light-video} LLaMA-mesh & \languagelogo \dlogo & \textcolor{blueGen}{G} & 0.00  & 0.00  & 0.00  & 1.60  & 0.40  & 85 \\
LISA & \languagelogo \imagelogo & \textcolor{greenCom}{C} & 1.27  & 0.00  & 0.00  & 0.00  & 0.32  & 86 \\
\rowcolor{bg-tb-light-video} GLaMM & \languagelogo \imagelogo & \textcolor{greenCom}{C} & 0.94  & 0.00  & 0.00  & 0.00  & 0.24  & 87 \\
AvatarGPT & \languagelogo \dlogo & \textcolor{greenCom}{C} & 0.00  & 0.00  & 0.00  & 0.21  & 0.05  & 88 \\
\rowcolor{bg-tb-light-video} MotionGPT-T5 & \languagelogo \dlogo & \textcolor{blueGen}{G} & 0.00  & 0.00  & 0.00  & 0.00  & 0.00  & / \\
MotionGPT-LLaMA & \languagelogo \dlogo & \textcolor{blueGen}{G} & 0.00  & 0.00  & 0.00  & 0.00  & 0.00  & / \\

\hline

\rowcolor{bg-tb-light-video} Meta-Llama-3.1-8B-Instruct & \languagelogo & / & 0.00  & 0.00  & 0.00  & 0.00  & 0.00  & / \\
Gemma-2-9b-it & \languagelogo & / & 0.00  & 0.00  & 0.00  & 0.00  & 0.00  & / \\
\rowcolor{bg-tb-light-video} GPT-J & \languagelogo & / & 0.00  & 0.00  & 0.00  & 0.00  & 0.00  & / \\
ChatGLM-6B & \languagelogo & / & 0.00  & 0.00  & 0.00  & 0.00  & 0.00  & / \\
\rowcolor{bg-tb-light-video} Qwen2.5-7B-Instruct & \languagelogo & / & 0.00  & 0.00  & 0.00  & 0.00  & 0.00  & / \\
InternLM2-Chat-7B & \languagelogo & / & 0.00  & 0.00  & 0.00  & 0.00  & 0.00  & / \\
\rowcolor{bg-tb-light-video} Baichuan2-7B-Base & \languagelogo & / & 0.00  & 0.00  & 0.00  & 0.00  & 0.00  & / \\
Vicuna-7b-V1.5 & \languagelogo & / & 0.00  & 0.00  & 0.00  & 0.00  & 0.00  & / \\
\rowcolor{bg-tb-light-video} Falcon3-7B-Instruct & \languagelogo & / & 0.00  & 0.00  & 0.00  & 0.00  & 0.00  & / \\
Ministral-8B-Instruct-2410 & \languagelogo & / & 0.00  & 0.00  & 0.00  & 0.00  & 0.00  & / \\
\rowcolor{bg-tb-light-video} Yi-lightning & \languagelogo & / & 0.00  & 0.00  & 0.00  & 0.00  & 0.00  & / \\
GPT-3.5-turbo & \languagelogo & / & 0.00  & 0.00  & 0.00  & 0.00  & 0.00  & / \\
\hline
\end{longtable}
}


{
\fontsize{8.5}{11}\selectfont 
\setlength{\tabcolsep}{1.5mm}
\captionof{table}{
Leaderboard of multimodal generalists (MLLMs) at level-3 where \textcolor{greenCom}{\underline{C}omprehension} and \textcolor{blueGen}{\underline{G}eneration}.
}
\vspace{-2mm}
\label{tab:ranking-level-3}
\begin{longtable}{p{4.2cm} cccccccc}
\hline
\multirow{2}{*}{\textbf{Model}} & \multirow{2}{*}{\textbf{Modality}} &\multirow{2}{*}{\textbf{Paradigm}} & \multicolumn{5}{c}{\textbf{Level 3 Score}} & \multirow{2}{*}{\textbf{Ranking}} \\
\cmidrule{4-8}
& & & \textbf{of Image}	& \textbf{of Video}	& 	\textbf{of Audio}	& 	\textbf{of 3D}	& \textbf{of Overall} & \\
\hline
\endfirsthead
\hline
\multirow{2}{*}{\textbf{Model}} & \multirow{2}{*}{\textbf{Modality}} &\multirow{2}{*}{\textbf{Paradigm}} & \multicolumn{5}{c}{\textbf{Level 3 Score}} & \multirow{2}{*}{\textbf{Ranking}} \\
\cmidrule{4-8}
& & & \textbf{of Image}	& \textbf{of Video}	& 	\textbf{of Audio}	& 	\textbf{of 3D}	& \textbf{of Overall} & \\
\hline
\endhead
\hline
\endfoot
\hline
\endlastfoot

\rowcolor{bg-tb-light-video} Sa2VA-26B & \languagelogo \imagelogo \videologo & \textcolor{greenCom}{C} & 14.65  & 4.58  & 0.00  & 0.00  & 4.81  & 1\championlogo \\
LLaVA-One-Vision-72B & \languagelogo \imagelogo \videologo & \textcolor{greenCom}{C} & 15.21  & 3.75  & 0.00  & 0.00  & 4.74  & 2\silverlogo \\
\rowcolor{bg-tb-light-video} Qwen2-VL-72B & \languagelogo \imagelogo \videologo & \textcolor{greenCom}{C} & 12.34  & 5.22  & 0.00  & 0.00  & 4.39  & 3\bronzelogo \\
Mini-Gemini & \languagelogo \imagelogo & \textcolor{greenCom}{C}+\textcolor{blueGen}{G} & 17.23 & 0.00 & 0.00 & 0.00 & 4.31  & 4 \\
\rowcolor{bg-tb-light-video} Sa2VA-8B & \languagelogo \imagelogo \videologo & \textcolor{greenCom}{C} & 12.39  & 4.38  & 0.00  & 0.00  & 4.19  & 5 \\
InternVL2\_5-8B & \languagelogo \imagelogo & \textcolor{greenCom}{C} & 13.09  & 1.82  & 0.00  & 0.00  & 3.73  & 6 \\
\rowcolor{bg-tb-light-video} GPT4-o-4096 & \languagelogo \imagelogo & \textcolor{greenCom}{C} & 14.68  & 0.00  & 0.00  & 0.00  & 3.67  & 7 \\
Qwen2-VL-7B & \languagelogo \imagelogo \videologo & \textcolor{greenCom}{C} & 12.13  & 2.47  & 0.00  & 0.00  & 3.65  & 8 \\
\rowcolor{bg-tb-light-video} GPT4-o & \languagelogo \imagelogo & \textcolor{greenCom}{C} & 14.51  & 0.00  & 0.00  & 0.00  & 3.63  & 9 \\
InternVL-2-26B & \languagelogo \imagelogo \videologo & \textcolor{greenCom}{C} & 8.81  & 4.81  & 0.00  & 0.00  & 3.41  & 10 \\
\cdashline{1-9}
\rowcolor{bg-tb-light-video} InternVL-2.5-26B & \languagelogo \imagelogo \videologo & \textcolor{greenCom}{C} & 9.51  & 3.76  & 0.00  & 0.00  & 3.32  & 11 \\
ChatGPT-o-latest & \languagelogo \imagelogo & \textcolor{greenCom}{C} & 13.02  & 0.00  & 0.00  & 0.00  & 3.26  & 12 \\
\rowcolor{bg-tb-light-video} GPT4-V & \languagelogo \imagelogo & \textcolor{greenCom}{C} & 12.85  & 0.00  & 0.00  & 0.00  & 3.21  & 13 \\
Gemini-1.5-Pro & \languagelogo \imagelogo & \textcolor{greenCom}{C} & 12.66  & 0.00  & 0.00  & 0.00  & 3.17  & 14 \\
\rowcolor{bg-tb-light-video} Claude-3.5-Sonnet & \languagelogo \imagelogo & \textcolor{greenCom}{C} & 11.98  & 0.00  & 0.00  & 0.00  & 3.00  & 15 \\
GPT4-o-mini & \languagelogo \imagelogo & \textcolor{greenCom}{C} & 11.94  & 0.00  & 0.00  & 0.00  & 2.99  & 16 \\
\rowcolor{bg-tb-light-video} LLaVA-One-Vision-7B & \languagelogo \imagelogo \videologo & \textcolor{greenCom}{C} & 10.21  & 1.54  & 0.00  & 0.00  & 2.94  & 17 \\
InternVL2\_5-4B & \languagelogo \imagelogo & \textcolor{greenCom}{C} & 11.59  & 0.00  & 0.00  & 0.00  & 2.90  & 18 \\
\rowcolor{bg-tb-light-video} Monkey-10B-chat & \languagelogo \imagelogo & \textcolor{greenCom}{C} & 11.59  & 0.00  & 0.00  & 0.00  & 2.90  & 19 \\
InternVL2\_5-2B & \languagelogo \imagelogo & \textcolor{greenCom}{C} & 11.45  & 0.00  & 0.00  & 0.00  & 2.86  & 20 \\
\rowcolor{bg-tb-light-video} Claude-3.5-Opus & \languagelogo \imagelogo & \textcolor{greenCom}{C} & 11.08  & 0.00  & 0.00  & 0.00  & 2.77  & 21 \\
Gemini-1.5-Flash & \languagelogo \imagelogo & \textcolor{greenCom}{C} & 10.85  & 0.00  & 0.00  & 0.00  & 2.71  & 22 \\
\rowcolor{bg-tb-light-video} Vitron-V1 & \languagelogo \imagelogo \videologo & \textcolor{greenCom}{C}+\textcolor{blueGen}{G} & 7.65  & 3.04  & 0.00  & 0.00  & 2.67  & 23 \\
CoLVA-4B & \languagelogo \imagelogo \videologo & \textcolor{greenCom}{C} & 9.45  & 1.24  & 0.00  & 0.00  & 2.67  & 24 \\
\rowcolor{bg-tb-light-video} Qwen-Audio-Chat & \languagelogo \audiologo & \textcolor{greenCom}{C} & 0.00  & 0.00  & 10.57  & 0.00  & 2.64  & 25 \\
InternVL-Chat-V1-5 & \languagelogo \imagelogo & \textcolor{greenCom}{C} & 9.42  & 0.00  & 0.00  & 0.00  & 2.36  & 26 \\
\rowcolor{bg-tb-light-video} Phi-3.5-Vision-Instruct & \languagelogo \imagelogo & \textcolor{greenCom}{C} & 9.39  & 0.00  & 0.00  & 0.00  & 2.35  & 27 \\
DeepSeek-VL-2 & \languagelogo \imagelogo & \textcolor{greenCom}{C} & 8.32  & 0.64  & 0.00  & 0.00  & 2.24  & 28 \\
\rowcolor{bg-tb-light-video} InternVL-2.5-8B & \languagelogo \imagelogo \videologo & \textcolor{greenCom}{C} & 7.63  & 1.24  & 0.00  & 0.00  & 2.22  & 29 \\
GLM-VL-Chat & \languagelogo \imagelogo & \textcolor{greenCom}{C} & 8.67  & 0.00  & 0.00  & 0.00  & 2.17  & 30 \\
\rowcolor{bg-tb-light-video} Qwen2-Audio-Instruct & \languagelogo \audiologo & \textcolor{greenCom}{C} & 0.00  & 0.00  & 8.53  & 0.00  & 2.13  & 31 \\
LLaVA-NeXT-34B & \languagelogo \imagelogo & \textcolor{greenCom}{C} & 8.24  & 0.00  & 0.00  & 0.00  & 2.06  & 32 \\
\rowcolor{bg-tb-light-video} DeepSeek-VL-7B-Chat & \languagelogo \imagelogo & \textcolor{greenCom}{C} & 8.19  & 0.00  & 0.00  & 0.00  & 2.05  & 33 \\
MiniCPM3-4B & \languagelogo \imagelogo & \textcolor{greenCom}{C} & 8.11  & 0.00  & 0.00  & 0.00  & 2.03  & 34 \\
\rowcolor{bg-tb-light-video} Long-LLaVA-9B & \languagelogo \videologo & \textcolor{greenCom}{C} & 4.21  & 3.81  & 0.00  & 0.00  & 2.01  & 35 \\
Yi-vision-v2 & \languagelogo \imagelogo & \textcolor{greenCom}{C} & 7.85  & 0.00  & 0.00  & 0.00  & 1.96  & 36 \\
\rowcolor{bg-tb-light-video} CogVLM-Chat & \languagelogo \imagelogo & \textcolor{greenCom}{C} & 7.77  & 0.00  & 0.00  & 0.00  & 1.94  & 37 \\
InternVL-2-8B & \languagelogo \imagelogo \videologo & \textcolor{greenCom}{C} & 7.28  & 0.46  & 0.00  & 0.00  & 1.94  & 38 \\
\rowcolor{bg-tb-light-video} Idefics3-8B-Llama3 & \languagelogo \imagelogo & \textcolor{greenCom}{C} & 7.70  & 0.00  & 0.00  & 0.00  & 1.93  & 39 \\
CoLVA-2B & \languagelogo \imagelogo \videologo & \textcolor{greenCom}{C} & 6.60  & 1.04  & 0.00  & 0.00  & 1.91  & 40 \\
\rowcolor{bg-tb-light-video} GAMA & \languagelogo \audiologo & \textcolor{greenCom}{C} & 0.00  & 0.00  & 7.15  & 0.00  & 1.79  & 41 \\
LLaVA-NeXT-13B & \languagelogo \imagelogo & \textcolor{greenCom}{C} & 6.87  & 0.00  & 0.00  & 0.00  & 1.72  & 42 \\
\rowcolor{bg-tb-light-video} BLIP-3 (XGen-MM) & \languagelogo \imagelogo & \textcolor{greenCom}{C} & 6.42  & 0.00  & 0.00  & 0.00  & 1.61  & 43 \\
ShareGPT4V-13B & \languagelogo \imagelogo & \textcolor{greenCom}{C} & 5.97  & 0.00  & 0.00  & 0.00  & 1.49  & 44 \\
\rowcolor{bg-tb-light-video} Qwen-VL-Chat & \languagelogo \imagelogo \videologo & \textcolor{greenCom}{C} & 5.88  & 0.00  & 0.00  & 0.00  & 1.47  & 45 \\
DeepSeek-VL-7B-Base & \languagelogo \imagelogo & \textcolor{greenCom}{C} & 5.75  & 0.00  & 0.00  & 0.00  & 1.44  & 46 \\
\rowcolor{bg-tb-light-video} Pixtral-12B & \languagelogo \imagelogo & \textcolor{greenCom}{C} & 5.72  & 0.00  & 0.00  & 0.00  & 1.43  & 47 \\
DeepSeek-VL-2-small & \languagelogo \imagelogo & \textcolor{greenCom}{C} & 5.12  & 0.52  & 0.00  & 0.00  & 1.41  & 48 \\
\rowcolor{bg-tb-light-video} MoE-LLAVA-Phi2-2.7B-4e-384 & \languagelogo \imagelogo & \textcolor{greenCom}{C} & 5.47  & 0.00  & 0.00  & 0.00  & 1.37  & 49 \\
NExT-GPT-V1.5 & \languagelogo \imagelogo \videologo \audiologo & \textcolor{greenCom}{C}+\textcolor{blueGen}{G} & 3.24  & 0.71  & 1.34  & 0.00  & 1.32  & 50 \\
\rowcolor{bg-tb-light-video} Mini‑InternVL‑Chat‑4B‑V1‑5 & \languagelogo \imagelogo & \textcolor{greenCom}{C} & 5.21  & 0.00  & 0.00  & 0.00  & 1.30  & 51 \\
Emu2-37B & \languagelogo \imagelogo & \textcolor{greenCom}{C}+\textcolor{blueGen}{G} & 5.18  & 0.00  & 0.00  & 0.00  & 1.30  & 52 \\
\rowcolor{bg-tb-light-video} InternLM-XComposer2-VL-1.8B & \languagelogo \imagelogo & \textcolor{greenCom}{C} & 4.78  & 0.00  & 0.00  & 0.00  & 1.20  & 53 \\
ShareGPT4V-7B & \languagelogo \imagelogo & \textcolor{greenCom}{C} & 4.78  & 0.00  & 0.00  & 0.00  & 1.20  & 54 \\
\rowcolor{bg-tb-light-video} MiniGPT4-LLaMA2-7B & \languagelogo \imagelogo & \textcolor{greenCom}{C} & 4.68  & 0.00  & 0.00  & 0.00  & 1.17  & 55 \\
mPLUG-Owl2-LLaMA2-7b & \languagelogo \imagelogo & \textcolor{greenCom}{C} & 4.60  & 0.00  & 0.00  & 0.00  & 1.15  & 56 \\
\rowcolor{bg-tb-light-video} AnyGPT & \languagelogo \imagelogo \audiologo & \textcolor{greenCom}{C}+\textcolor{blueGen}{G} & 1.29  & 0.00  & 3.29  & 0.00  & 1.15  & 57 \\
miniMonkey & \languagelogo \imagelogo & \textcolor{greenCom}{C} & 4.51  & 0.00  & 0.00  & 0.00  & 1.13  & 58 \\
\rowcolor{bg-tb-light-video} Cambrian-1-8B & \languagelogo \imagelogo & \textcolor{greenCom}{C} & 3.84  & 0.00  & 0.00  & 0.00  & 0.96  & 59 \\
DetGPT & \languagelogo \imagelogo & \textcolor{greenCom}{C} & 3.77  & 0.00  & 0.00  & 0.00  & 0.94  & 60 \\
\rowcolor{bg-tb-light-video} LaVIT-V2 (7B) & \languagelogo \imagelogo & \textcolor{greenCom}{C}+\textcolor{blueGen}{G} & 3.71  & 0.00  & 0.00  & 0.00  & 0.93  & 61 \\
SALMONN-13B & \languagelogo \audiologo & \textcolor{greenCom}{C} & 0.00  & 0.00  & 3.61  & 0.00  & 0.90  & 62 \\
\rowcolor{bg-tb-light-video} ImageBind-LLM & \languagelogo \imagelogo \videologo \audiologo & \textcolor{greenCom}{C} & 1.56  & 0.72  & 1.26  & 0.00  & 0.89  & 63 \\
NExT-Chat & \languagelogo \imagelogo & \textcolor{greenCom}{C} & 3.51  & 0.00  & 0.00  & 0.00  & 0.88  & 64 \\
\rowcolor{bg-tb-light-video} SEED-LLaMA-13B & \languagelogo \imagelogo & \textcolor{greenCom}{C}+\textcolor{blueGen}{G} & 3.49  & 0.00  & 0.00  & 0.00  & 0.87  & 65 \\
WavLLM & \languagelogo \audiologo & \textcolor{greenCom}{C} & 0.00  & 0.00  & 3.28  & 0.00  & 0.82  & 66 \\
\rowcolor{bg-tb-light-video} Unified-io-2-XXL & \languagelogo \imagelogo \videologo \audiologo & \textcolor{greenCom}{C}+\textcolor{blueGen}{G} & 2.11  & 0.14  & 1.01  & 0.00  & 0.82  & 67 \\
ModaVerse-7b-v0 & \languagelogo \imagelogo \videologo \audiologo & \textcolor{greenCom}{C}+\textcolor{blueGen}{G} & 0.98  & 0.23  & 1.14  & 0.78  & 0.78  & 68 \\
\rowcolor{bg-tb-light-video} PandaGPT-13B & \languagelogo \imagelogo \videologo \audiologo & \textcolor{greenCom}{C} & 2.35  & 0.05  & 0.65  & 0.00  & 0.76  & 69 \\
Audio-GPT4 & \languagelogo \audiologo & \textcolor{blueGen}{G} & 0.00  & 0.00  & 3.02  & 0.00  & 0.76  & 70 \\
\rowcolor{bg-tb-light-video} BLIP2 & \languagelogo \imagelogo & \textcolor{greenCom}{C} & 2.79  & 0.00  & 0.00  & 0.00  & 0.70  & 71 \\
GPT4RoI-7B & \languagelogo \imagelogo & \textcolor{greenCom}{C} & 2.36  & 0.00  & 0.00  & 0.00  & 0.59  & 72 \\
\rowcolor{bg-tb-light-video} Pengi & \languagelogo \audiologo & \textcolor{greenCom}{C} & 0.00  & 0.00  & 1.74  & 0.00  & 0.44  & 73 \\
3D-LLM-2.1B & \languagelogo \dlogo & \textcolor{greenCom}{C} & 0.00  & 0.00  & 0.00  & 1.38  & 0.35  & 74 \\
\rowcolor{bg-tb-light-video} 3D-VisTA & \languagelogo \dlogo & \textcolor{greenCom}{C} & 0.00  & 0.00  & 0.00  & 1.07  & 0.27  & 75 \\
Show-o & \languagelogo \imagelogo & \textcolor{greenCom}{C}+\textcolor{blueGen}{G} & 0.84 & 0.00 & 0.00 & 0.00 & 0.21  & 76 \\
\rowcolor{bg-tb-light-video} LISA & \languagelogo \imagelogo & \textcolor{greenCom}{C} & 0.82  & 0.00  & 0.00  & 0.00  & 0.21  & 77 \\
Otter & \languagelogo \imagelogo & \textcolor{greenCom}{C} & 0.68  & 0.00  & 0.00  & 0.00  & 0.17  & 78 \\
\rowcolor{bg-tb-light-video} OMG-LLaVA-InternLM20B & \languagelogo \imagelogo & \textcolor{greenCom}{C} & 0.44  & 0.00  & 0.00  & 0.00  & 0.11  & 79 \\
GLaMM & \languagelogo \imagelogo & \textcolor{greenCom}{C} & 0.41  & 0.00  & 0.00  & 0.00  & 0.10  & 80 \\
\rowcolor{bg-tb-light-video} AvatarGPT & \languagelogo \dlogo & \textcolor{greenCom}{C} & 0.00  & 0.00  & 0.00  & 0.21  & 0.05  & 81 \\

PointLLM-7B & \languagelogo \dlogo & \textcolor{greenCom}{C} & 0.00  & 0.00  & 0.00  & 0.00  & 0.00  & / \\
\rowcolor{bg-tb-light-video} PointLLM-13B & \languagelogo \dlogo & \textcolor{greenCom}{C} & 0.00  & 0.00  & 0.00  & 0.00  & 0.00  & / \\
MotionGPT-T5 & \languagelogo \dlogo & \textcolor{blueGen}{G} & 0.00  & 0.00  & 0.00  & 0.00  & 0.00  & / \\
\rowcolor{bg-tb-light-video} MotionGPT-LLaMA & \languagelogo \dlogo & \textcolor{blueGen}{G} & 0.00  & 0.00  & 0.00  & 0.00  & 0.00  & / \\
LLaMA-mesh & \languagelogo \dlogo & \textcolor{blueGen}{G} & 0.00  & 0.00  & 0.00  & 0.00  & 0.00  & / \\
\rowcolor{bg-tb-light-video} SALMONN-7B & \languagelogo \audiologo & \textcolor{greenCom}{C} & 0.00  & 0.00  & 0.00  & 0.00  & 0.00  & / \\
SpeechGPT-7B-com & \languagelogo \audiologo & \textcolor{blueGen}{G} & 0.00  & 0.00  & 0.00  & 0.00  & 0.00  & / \\

LM4LV & \languagelogo \videologo & \textcolor{blueGen}{G} & 0.00  & 0.00  & 0.00  & 0.00  & 0.00  & / \\
\rowcolor{bg-tb-light-video} VidAgent & \languagelogo \imagelogo \videologo & \textcolor{greenCom}{C}+\textcolor{blueGen}{G} & 0.00  & 0.00  & 0.00  & 0.00  & 0.00  & / \\

\hline

Meta-Llama-3.1-8B-Instruct & \languagelogo & / & 0.00  & 0.00  & 0.00  & 0.00  & 0.00  & / \\
\rowcolor{bg-tb-light-video} Gemma-2-9b-it & \languagelogo & / & 0.00  & 0.00  & 0.00  & 0.00  & 0.00  & / \\
GPT-J & \languagelogo & / & 0.00  & 0.00  & 0.00  & 0.00  & 0.00  & / \\
\rowcolor{bg-tb-light-video} ChatGLM-6B & \languagelogo & / & 0.00  & 0.00  & 0.00  & 0.00  & 0.00  & / \\
Qwen2.5-7B-Instruct & \languagelogo & / & 0.00  & 0.00  & 0.00  & 0.00  & 0.00  & / \\
\rowcolor{bg-tb-light-video} InternLM2-Chat-7B & \languagelogo & / & 0.00  & 0.00  & 0.00  & 0.00  & 0.00  & / \\
Baichuan2-7B-Base & \languagelogo & / & 0.00  & 0.00  & 0.00  & 0.00  & 0.00  & / \\
\rowcolor{bg-tb-light-video} Vicuna-7b-V1.5 & \languagelogo & / & 0.00  & 0.00  & 0.00  & 0.00  & 0.00  & / \\
Falcon3-7B-Instruct & \languagelogo & / & 0.00  & 0.00  & 0.00  & 0.00  & 0.00  & / \\
\rowcolor{bg-tb-light-video} Ministral-8B-Instruct-2410 & \languagelogo & / & 0.00  & 0.00  & 0.00  & 0.00  & 0.00  & / \\
Yi-lightning & \languagelogo & / & 0.00  & 0.00  & 0.00  & 0.00  & 0.00  & / \\
\rowcolor{bg-tb-light-video} GPT-3.5-turbo & \languagelogo & / & 0.00  & 0.00  & 0.00  & 0.00  & 0.00  & / \\

\hline
\end{longtable}
}

\begin{table}[!h]
\centering
\fontsize{8.5}{12}\selectfont 
\setlength{\tabcolsep}{1.2mm}
\caption{Leaderboard of multimodal generalists (MLLMs) at level-4, where \textcolor{greenCom}{\underline{C}omprehension} and \textcolor{blueGen}{\underline{G}eneration}.
}
\label{tab:ranking-level-4}
\begin{tabular}{p{4.2cm} cccccccc}
\hline
\multirow{2}{*}{\textbf{Model}} & \multirow{2}{*}{\textbf{Modality}} &\multirow{2}{*}{\textbf{Paradigm}} & \multicolumn{5}{c}{\textbf{Level 4 Score}} & \multirow{2}{*}{\textbf{Ranking}} \\
\cmidrule{4-8}
& & & \textbf{of Image}	& \textbf{of Video}	& 	\textbf{of Audio}	& 	\textbf{of 3D}	& \textbf{of Overall} & \\
\hline

\rowcolor{bg-tb-light-video} Mini-Gemini & \languagelogo \imagelogo & \textcolor{greenCom}{C}+\textcolor{blueGen}{G} & 6.23 & 0.00 & 0.00 & 0.00 & 1.56  & 1\championlogo\\
Vitron-V1 & \languagelogo \imagelogo \videologo & \textcolor{greenCom}{C}+\textcolor{blueGen}{G} & 4.59  & 0.00  & 0.00  & 0.00  & 1.15  & 2\silverlogo \\
\rowcolor{bg-tb-light-video} Emu2-37B & \languagelogo \imagelogo & \textcolor{greenCom}{C}+\textcolor{blueGen}{G} & 1.25  & 0.00  & 0.00  & 0.00  & 0.31  & 3\bronzelogo \\

\hline
\end{tabular}
\end{table}


\begin{figure*}[!t]
\centering
\includegraphics[width=0.99\textwidth]{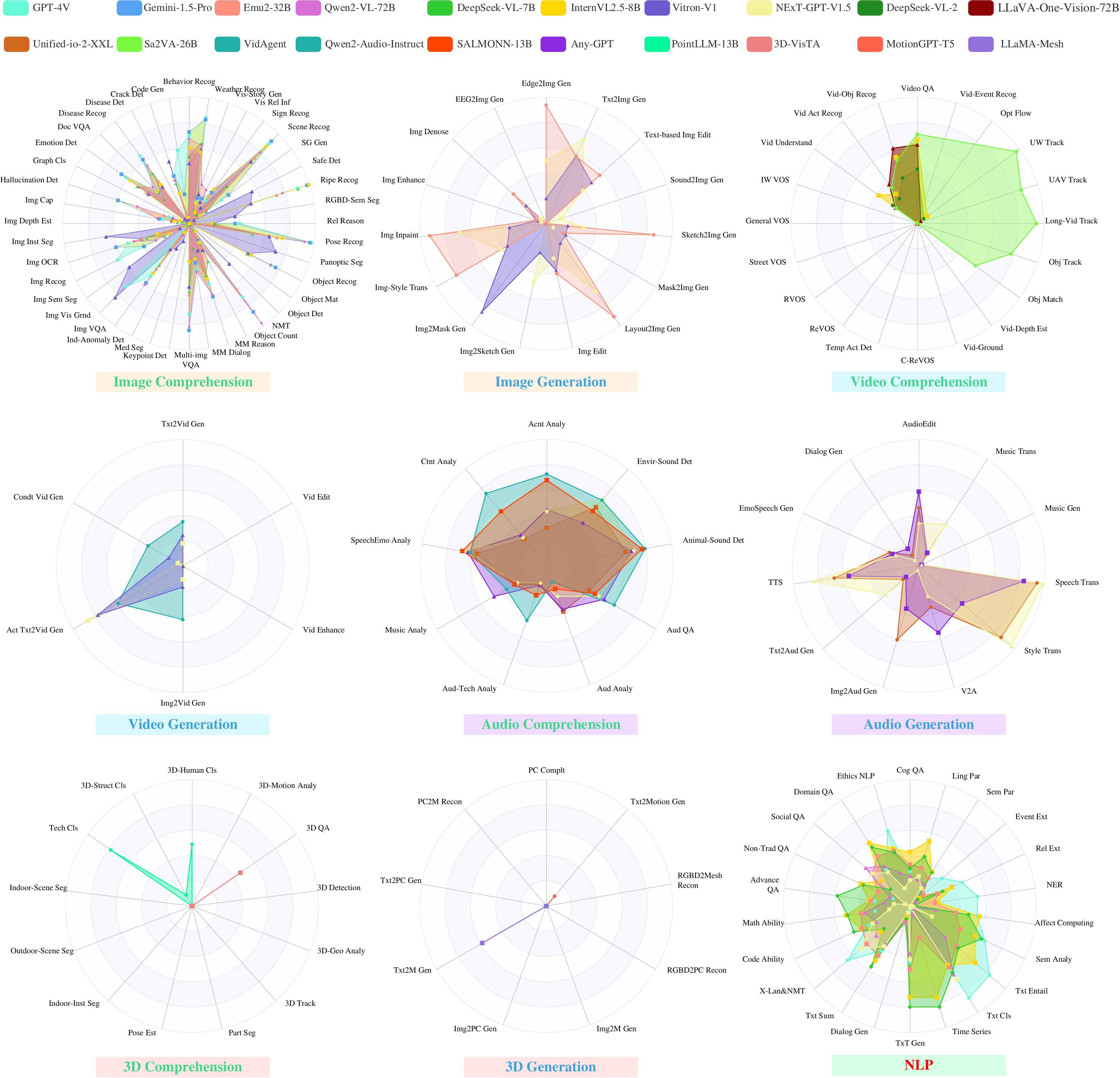}
\caption{
Visualization of skill support in various multimodal generalists.
}
\label{fig:task-supporting}
\vspace{-3mm}
\end{figure*}

\vspace{-2mm}
\subsection{Level and Leaderboard of Multimodal Generalists}

\vspace{-1mm}
Based on the overall performance of each model across the various modalities and tasks, we rank all the compared models according to the \texttt{General-Level} scoring defined in $\S$ \ref{Definition of Levels}. 
Tables~\ref{tab:ranking-level-2}, \ref{tab:ranking-level-3} and \ref{tab:ranking-level-4} present the specific scores and rankings of multimodal generalists at different General-Levels.
Note that no generalists score non-zero at Level-5, and thus we do not show a rank at Level-5.
Figure \ref{fig:leaderboard-intro} visualizes these leaderboards.

As shown, for all the current MLLMs at level 2, Unified-io-2-XXL \cite{lu2024unified} ranks the best, followed by AnyGPT \cite{zhan2024anygpt}. 
Surprisingly, GPT-4V and GPT-4o did not achieve the expected rankings at level 2. 
While the GPT series excels in the individual tasks it supports, as generalists, they fall short in skill coverage compared to some open-source MLLMs. 
This is because, to rank higher at level 2, models must not only perform well on different tasks but also support as many modalities and tasks as possible.

Next, MLLMs that can manage to reach level 3 become different.
Sa2VA-26B \cite{sa2va} ranks at the top, while LLaVA-One-Vision-72B \cite{li2024llavaonevision} and Qwen2-VL-72B \cite{wang2024qwen2} achieve the second and third places, respectively.
Some high-ranking level-2 models lost their places at level 3.
This lies in the fact that most MLLMs are limited to multimodal content comprehension and lack support for generation tasks.
GPT-4V and GPT-4o win top-10 positions here.

Finally, only 3 MLLMs that reach level 4 can be seen, i.e., Mini-Gemini, Emu2-37B, and Vitron-V1. 
At this level, these models exhibit synergy across both comprehension and generation.
Besides these three models, no other systems exhibit such capability.

Most critically, no model has yet demonstrated the ability to enhance language intelligence through non-language modalities, underscoring the significant challenges in the pursuit of true AGI.
And this is definitely our goal to reach the most capable multimodal generalists.

\vspace{-2mm}

\subsection{Capability BreakDown}

We now take a closer look, as multimodal generalists, at how well different MLLMs support tasks and modalities.

\vspace{-3mm}
\paragraph{Task Supporting.}
In Figure~\ref{fig:task-supporting}, we present all skills (meta-tasks) supported by different MLLMs across various modalities and within the scopes of comprehension and generation.
Overall, MLLMs show relatively lower task support for 3D tasks and skills, compared with the status for other modalities.
Also, the coverage of comprehension-related skills by MLLMs should be generally higher than that of generation-related skills. 
This trend is consistent with the results observed in previous experiments.
A significant trend we identified is that MLLMs tend to support skills within only one (or a few) task paradigms. 
This results in differentiated skill support across different MLLMs, with few models capable of supporting a wide range of skills across diverse tasks.

\begin{figure}[!t]  
    \centering
    \begin{minipage}{0.43\textwidth}  
        \centering
        \includegraphics[width=\linewidth]{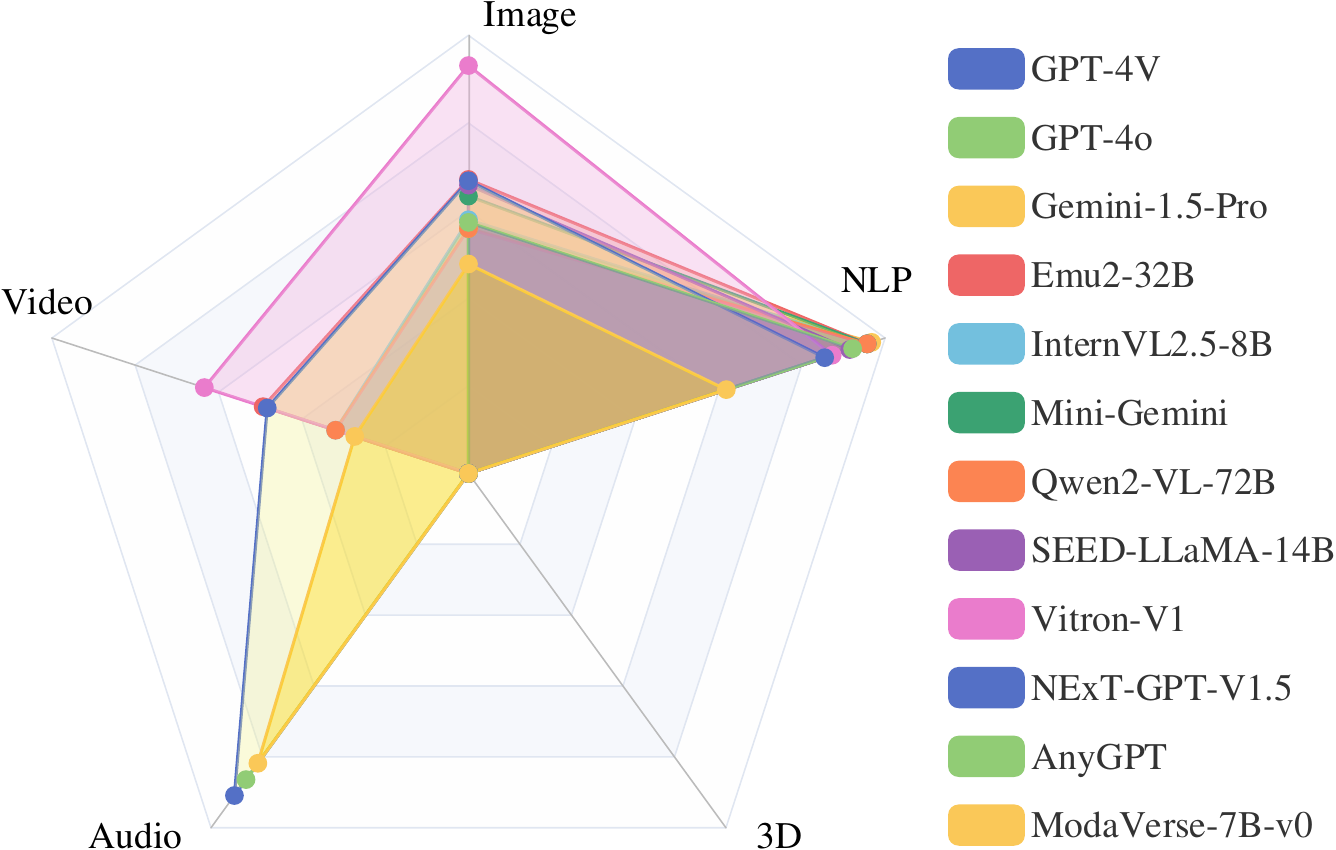}
        \caption{Supporting modality.}
        \label{fig:modality-supporting}
    \end{minipage}
    \hspace{3mm}
    \begin{minipage}{0.45\textwidth}  
        \centering
        \includegraphics[width=\linewidth]{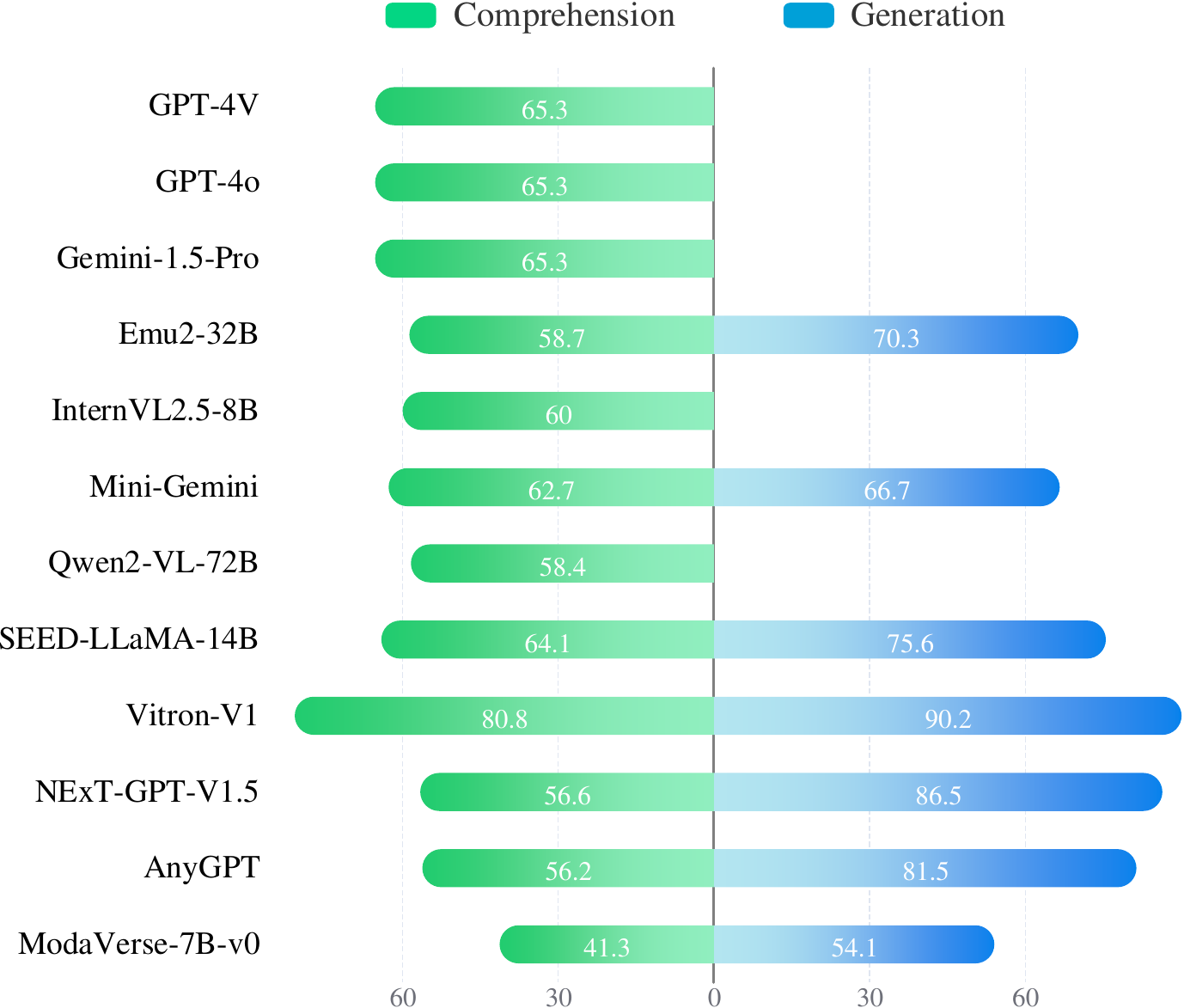}
        \caption{Supporting generation and comprehension.}
        \label{fig:g-c-supporting}
    \end{minipage}
    \vspace{-3mm}
\end{figure}

Moreover, most MLLMs are inclined to focus on basic skills or tasks with simpler and more straightforward definitions, while tasks requiring complex content output and advanced skills are supported by far fewer models.
For instance, compared to coarse-grained visual understanding tasks (e.g., captioning and classification), tasks with more complex definitions—such as pixel-level object detection, image/3D segmentation, video tracking, and image generation—are supported by far fewer existing MLLMs.
However, we observe that a few MLLMs stand out for their broader support of cross-modal skills, such as Vitron-V1 \cite{fei2024vitron}.
Thanks to their architectural designs, these models demonstrate a wider range of task support compared to others.
We emphasize that supporting as many task paradigms as possible is a critical requirement for developing more capable multimodal generalists.

\vspace{-2mm}
\paragraph{Modality Supporting.}
The broader the range of supported modalities, the more general and versatile the model's capabilities are. 
Our benchmark emphasizes the evaluation of MLLMs' all-modality capabilities.
As shown in the experimental results above, most MLLMs support only a single modality (excluding the language modality, which is inherently supported by LLMs). 
To further illustrate this, Figure \ref{fig:modality-supporting} compares the multimodal support capabilities of several top-performing MLLMs.
In general, there are very few MLLMs capable of supporting multiple modalities simultaneously. 
In most cases, MLLMs support one non-language modality, e.g., GPT-4V \cite{gpt4}, Emu2-32B \cite{sun2024generative}, Mini-Gemini \cite{li2024mini}, InternVL2.5-8B \cite{chen2024internvl}. 
Only a few MLLMs stand out with broader cross-modal or even all-modality support capabilities, encompassing language, image, video, and audio modalities. 
Examples include NExT-GPT-V1.5 \cite{wu2023next}, Unified-io-2-XXL \cite{lu2024unified}, and AnyGPT \cite{zhan2024anygpt}, etc. 
Thanks to their architectural designs, these systems demonstrate a wider range of modality and task support compared to others.

\begin{figure*}[!t]
\centering
\includegraphics[width=1\textwidth]{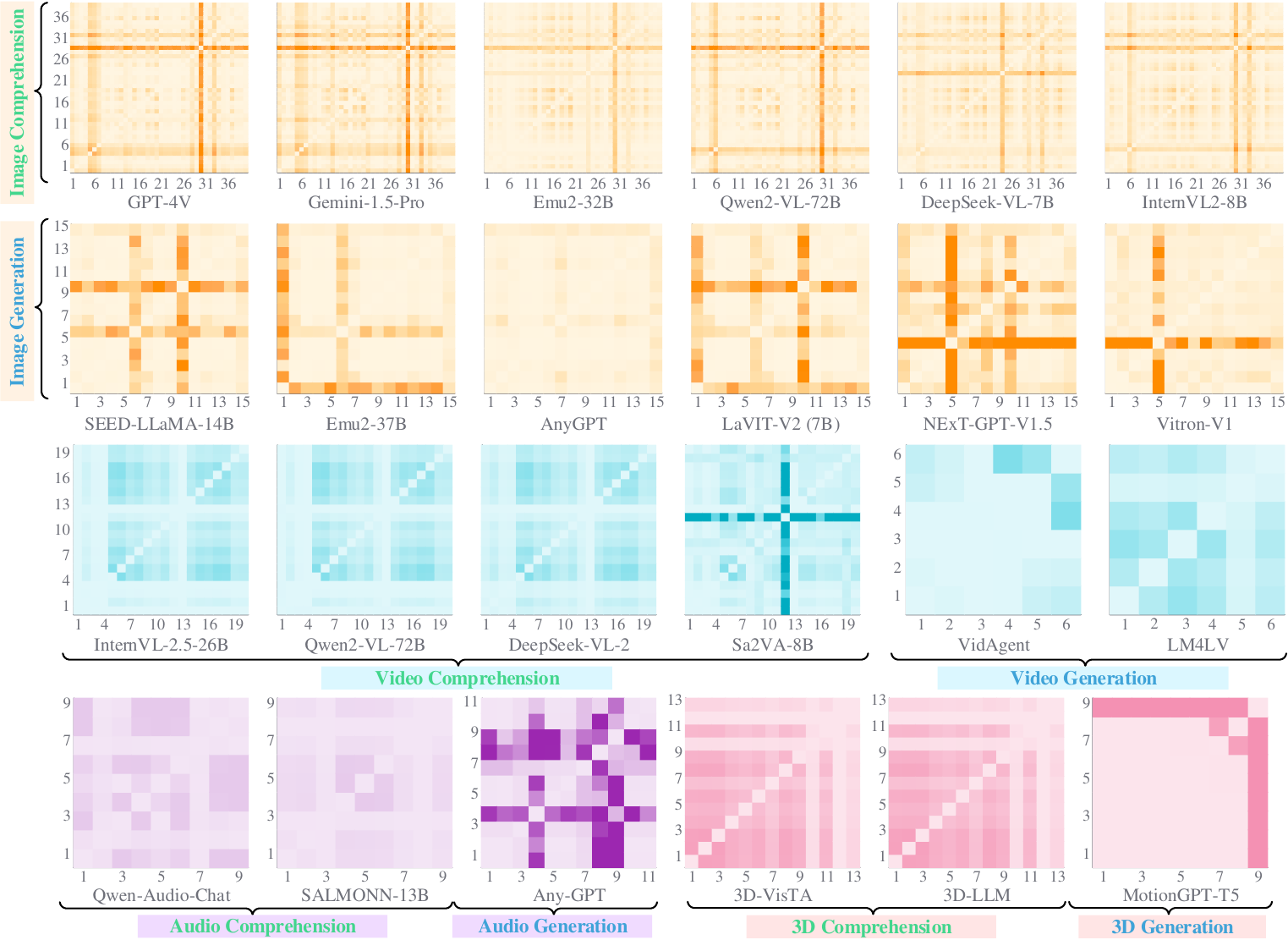}
\vspace{-5mm}
\caption{
Visualizations of synergy effects between all different skills of various MLLMs.
}
\label{fig:synergy-heatmap-skill}
\vspace{-5mm}
\end{figure*}

\paragraph{Capabilities on Comprehension vs. Generation.}
Within a single modality, tasks and skills can be categorized into comprehension and generation types. Here, we explore the capabilities of different MLLMs in supporting these two paradigms.
We select several representative MLLMs that can or cannot support both comprehension and generation within various modalities for comparison. 
Figure~\ref{fig:g-c-supporting} directly presents the statistics on these models' capabilities in these two aspects.
Overall, support for content comprehension significantly outweighs support for content generation. 
This phenomenon aligns with practical realities, as modeling tasks for comprehension typically involve expressing the understood content in language, which is relatively straightforward. 
In contrast, generating multimodal content requires additional efforts in model decoding and extra training, making it a more challenging capability to achieve.
It is also evident that different models exhibit varying balances between comprehension and generation capabilities. 
Among them, Vitron-V1 \cite{fei2024vitron} demonstrates the most comprehensive and well-rounded capabilities in both comprehension and generation to date, i.e., supporting the largest number of both paradigms.

\vspace{-2mm}
\subsection{Analysis and Discussion on Synergy}

Next, we conduct a finer-grained analysis of the synergy performance of various MLLMs.

\vspace{-3mm}
\paragraph{Synergy Across Skills.}
First, we examine the different synergy effects exhibited by models across various skills. 
Skills are categorized based on different modalities and further divided into comprehension and generation categories. 
We explore the synergy effects displayed by top-performing MLLMs within these skill groups.
Technically, we calculate the synergy score for tasks (skills) based on the level-3 score algorithm from \texttt{General-Level} ($\S$ \ref{Definition of Levels}). 
Then we count the number of synergy tasks for each model and use scores exceeding SoTA specialists as weights. 
Figure~\ref{fig:synergy-heatmap-skill} visualizes the results with heatmaps. 
It is shown that different models exhibit varying levels of cross-task synergy capabilities. 
Overall, models that achieve higher level-3 scores tend to display denser clusters of highlighted cells in the heatmap.
Also, we observe that synergy effects are more likely to occur among closely related skills, as knowledge and information are more easily transferable between similar tasks. 
This trend is consistent across different modalities. 
Furthermore, generation tasks seem to exhibit stronger synergy effects compared to tasks within the comprehension category.

\begin{figure*}[!t]
\centering
\includegraphics[width=0.99\textwidth]{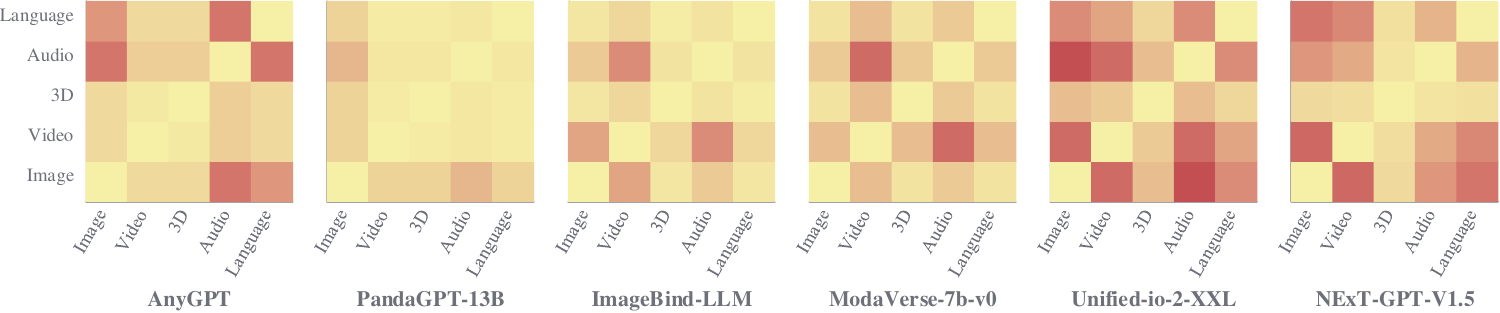}
\vspace{-2mm}
\caption{
Visualizations (symmetrised) of synergy effects between modalities of various MLLMs.
}
\label{fig:synergy-heatmap-modality}
\vspace{-2mm}
\end{figure*}

\vspace{-3mm}
\paragraph{Synergy Across Modalities.}
Finally, we analyze whether different models have learned synergy effects across modalities. 
The approach is similar to the previous methods, where we calculate instances where performance across modalities exceeds that of SoTA specialists.
Figure~\ref{fig:synergy-heatmap-modality} visualizes the performance of 6 representative strong-performing MLLMs that cover as many modalities as possible. 
Several notable trends emerge.
First, we observe that there is significant synergy occurs between the image and video modalities. 
This is reasonable, as static image information and dynamic video content are both fundamentally visual in nature, allowing for substantial information sharing that enhances performance across tasks.

Most strikingly, although language appears to exhibit a synergy effect with various other modalities, this effect is in fact unidirectional, specifically from language $\to$ other modalities. 
Based on our previous results on NLP tasks, no synergy has been observed from any other modality to the language modality.
While in theory, the audio modality should be closely related to language, and we would expect to observe synergy between audio and language tasks, this is not reflected in current results. 
This limitation is likely due to the reliance of audio-based LLM architectures on the language intelligence of LLMs, which has not yet translated into performance improvements that exceed those of NLP SoTA specialists.
We thus strongly urge the MLLM community to prioritize enhancing cross-modality synergy capabilities to advance the development of truly comprehensive multimodal generalists.

\begin{wrapfigure}[12]{r}{6.2cm}
\vspace{-3mm}
\centering
\includegraphics[width=0.98\linewidth]{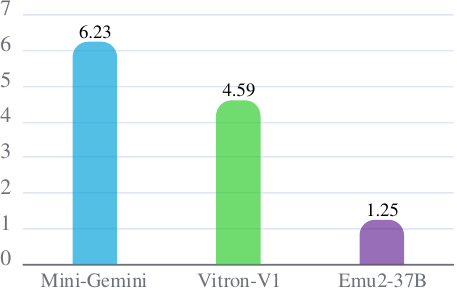}
\vspace{-2mm}
\caption{
Synergy strength between comprehension and generation.
}
\label{fig:synergy-comp-gen}
\end{wrapfigure}

\vspace{-3mm}
\paragraph{Synergy Across Comprehension and Generation.}
We further investigate the synergy across comprehension and generation, which represents a broader type of synergy than at the task level. 
Using a similar approach, we calculate the synergy score based on the level-4 score algorithm from \texttt{General-Level}. 
Specifically, we count the frequency of synergy occurrences for each model and use the performance increments exceeding SoTA specialists as weights. 
The total sum (after normalization) constitutes the synergy score across comprehension and generation.
Figure~\ref{fig:synergy-comp-gen} presents the comparisons for only 3 models. 
As shown, Mini-Gemini \cite{li2024mini} demonstrates the best synergy effects at this level, securing the top rank on our \texttt{General-Level} leaderboard.
Also, we observe that the cross-comprehension-generation synergy is only pronounced in the image modality, as these 3 models all support only image modalities.

\vspace{-2mm}
\section{Discussions and Future Investigation}

We propose a leveled evaluation of MLLMs, with a hierarchical framework called \texttt{General-Level}, and a large-scale benchmark dataset \texttt{General-Bench}. 
Yet we believe several aspects of this work can be further improved.
In addition, as the next step to achieve more capable multimodal generalists toward AGI, we believe some points are worth further investigation.

\vspace{-2mm}

\paragraph{Further refinement of the \texttt{General-Level} framework.}  
Although this framework is a concrete starting point for building true multimodal generalists, there are still areas for improvement, especially in terms of the algorithms. For example, the coordination average used to compute Level-3 in the \texttt{General-Level} framework assumes a balance between the number of comprehension and generation tasks, which is an unrealistic assumption. Additionally, we have relaxed the measurement of synergy by assuming that a model’s synergy capability is reflected in its ability to surpass SoTA specialists in task performance, avoiding a direct measurement of the synergy effect. Future work will consider optimizing this aspect to provide a more robust definition.

\vspace{-2mm}

\paragraph{Expanding the \texttt{General-Bench} dataset to include more comprehensive tasks and modalities.}  
To ensure the evaluation of multimodal generalists is complete and unbiased, the \texttt{General-Bench} dataset should be further expanded. 
Currently, the dataset is somewhat imbalanced across different modalities and tasks; for example, there is more data for image-related tasks than for audio and 3D modalities, and there are more comprehension tasks than generation tasks. 
Moreover, multimodality should be more broadly defined to include not just visible information such as language, vision, and sound, but also other signals and types of information. 
For instance, LLMs’ reasoning abilities have been shown to be significantly enhanced through learning code. 
As multimodal models, it is necessary to support some coding capabilities, which, in turn, could theoretically improve the model's understanding and reasoning in other modalities, such as vision.
Finally, our current benchmark primarily considers tasks in which individual modality is operated in isolation. 
However, in reality, a good multimodal generalist should be capable of modality-switching and modality-interleaved reasoning (in both comprehension and generation) under multi-turn user-machine interactions. 
In the future, we plan to incorporate tasks and datasets that assess interleaved modality capabilities.

\vspace{-2mm}

\paragraph{Rethinking Evaluation Paradigm for Model Capabilities.}  
Many current task evaluation methodologies still follow conventional paradigms. In most cases, automatic and scalable evaluation strategies are preferred (we provide a detailed overview in Appendix $\S$\ref{Evaluation Metrics}). While such approaches may suffice for relatively straightforward tasks—such as multiple-choice or classification—they often fall short when applied to format-free tasks, particularly those involving multimodal generation.
For instance, in video or 3D generation, traditional metrics like FID or FVD are increasingly considered inadequate, as they fail to reliably capture the quality and fidelity of the generated content. 
Consequently, there is a growing reliance on human evaluations. 
To improve scalability, many recent works have begun employing LLMs to simulate human-level judgment \cite{zheng2023judging}. 
However, this ``LLM-as-a-judge'' approach introduces challenges in terms of evaluation stability and reproducibility, which remain open research problems.
In addition, our current General-Level evaluation framework adopts a single primary metric per task, which may inherently introduce bias. 
We argue that future evaluations should incorporate multiple complementary metrics to provide a more comprehensive assessment.
Lastly, as multimodal generalist models continue to evolve with stronger reasoning capabilities, corresponding benchmarks should also be upgraded to evaluate the interpretability and traceability of their intermediate reasoning processes.

\vspace{-2mm}
\paragraph{Optimizing model architecture to support more functionalities and modalities with stronger performance.}  
Our above experiments reveal that very few MLLMs currently achieve unified capabilities. Most models support only 1-2 modalities or abilities, which severely limits their qualifications as generalists. 
Some models, while supporting multiple tasks and modalities, still exhibit limited performance in individual tasks. 
Future research could focus on optimizing model architectures to support as many functions and modalities as possible while also delivering stronger performance. 
We note that simply integrating multiple models or modules using an agent-based approach can increase the number of supported functions and modalities but does not necessarily improve performance (i.e., it still cannot surpass individual specialists). 
A more promising approach may be to leverage the Mixture of Experts (MoE) strategy to construct more unified MLLMs.
Recently, more advanced understanding and generation capabilities have also been achieved through some of the latest architectures, such as those that combine autoregressive and diffusion frameworks.

\vspace{-2mm}
\paragraph{Strengthening synergy capabilities is the key focus.}  
As emphasized in this work, achieving synergy is the fundamental requirement, and it is crucial for ensuring the model has more powerful capabilities (compared to specialists). 
To accomplish this, several aspects need to be considered. 
First, at the architectural level, it is necessary to design essential modules or mechanisms that allow the model to flexibly and effectively transfer features learned from different tasks and modalities \cite{fei2024vitron,pan2025transfer}. 
This is key to enabling the ``learning by analogy'' capability. 
Second, at the learning level, to achieve stronger capabilities across multiple tasks and modalities, the model must be trained in a way that prevents it from forgetting previously learned knowledge when learning new tasks.
Also, recently, by incorporating training techniques such as Reinforcement Learning from Human Feedback (RLHF), models have achieved more powerful reasoning capabilities and improved generalization.

\vspace{-2mm}
\section{Conclusion}

\vspace{-1mm}
Inspired by the concept of capability levels defined in the autonomous driving industry, we propose \texttt{General-Level}, a framework that evaluates and categorizes the capabilities of existing MLLMs through a 5-level hierarchical rating mechanism based on their ability to maintain synergy across comprehension, generation, and multimodal interactions.
\texttt{General-Level} provides a structured methodology to assess MLLMs across diverse tasks, modalities, and synergies, particularly in comprehension and generation.
To support this evaluation, we further present \texttt{General-Bench}, a large-scale multimodal benchmark that spans a wide spectrum of tasks (702), modalities (language, image, video, audio, 3D, etc.), domains (29), and original formats with 325,800 instances. 
By benchmarking over 100 popular LLMs/MLLMs, we uncover the current capability limitations and provide a clear ranking of generalist performance.
We hope the \texttt{General-Level} and \texttt{General-Bench} in this study will propel the community to develop next-generation multimodal foundation models to achieve more sophisticated, general-purpose multimodal intelligence.



\newpage

{
\bibliographystyle{unsrtnat}
\bibliography{ref}

\begin{thebibliography}{679}
\providecommand{\natexlab}[1]{#1}
\providecommand{\url}[1]{\texttt{#1}}
\expandafter\ifx\csname urlstyle\endcsname\relax
  \providecommand{\doi}[1]{doi: #1}\else
  \providecommand{\doi}{doi: \begingroup \urlstyle{rm}\Url}\fi

\bibitem[OpenAI(2022{\natexlab{a}})]{chatgpt}
OpenAI.
\newblock Introducing chatgpt.
\newblock 2022{\natexlab{a}}.

\bibitem[Touvron et~al.(2023)Touvron, Lavril, Izacard, Martinet, Lachaux, Lacroix, Rozi{\`{e}}re, Goyal, Hambro, Azhar, Rodriguez, Joulin, Grave, and Lample]{abs-2302-13971}
Hugo Touvron, Thibaut Lavril, Gautier Izacard, Xavier Martinet, Marie{-}Anne Lachaux, Timoth{\'{e}}e Lacroix, Baptiste Rozi{\`{e}}re, Naman Goyal, Eric Hambro, Faisal Azhar, Aur{\'{e}}lien Rodriguez, Armand Joulin, Edouard Grave, and Guillaume Lample.
\newblock Llama: Open and efficient foundation language models.
\newblock \emph{CoRR}, abs/2302.13971, 2023.

\bibitem[Alayrac et~al.(2022)Alayrac, Donahue, Luc, Miech, Barr, Hasson, Lenc, Mensch, Millican, Reynolds, Ring, Rutherford, Cabi, Han, Gong, Samangooei, Monteiro, Menick, Borgeaud, Brock, Nematzadeh, Sharifzadeh, Binkowski, Barreira, Vinyals, Zisserman, and Simonyan]{AlayracDLMBHLMM22}
Jean{-}Baptiste Alayrac, Jeff Donahue, Pauline Luc, Antoine Miech, Iain Barr, Yana Hasson, Karel Lenc, Arthur Mensch, Katherine Millican, Malcolm Reynolds, Roman Ring, Eliza Rutherford, Serkan Cabi, Tengda Han, Zhitao Gong, Sina Samangooei, Marianne Monteiro, Jacob~L. Menick, Sebastian Borgeaud, Andy Brock, Aida Nematzadeh, Sahand Sharifzadeh, Mikolaj Binkowski, Ricardo Barreira, Oriol Vinyals, Andrew Zisserman, and Kar{\'{e}}n Simonyan.
\newblock Flamingo: a visual language model for few-shot learning.
\newblock In \emph{Proceedings of the NeurIPS}, 2022.

\bibitem[Li et~al.(2023{\natexlab{a}})Li, Li, Savarese, and Hoi]{0008LSH23}
Junnan Li, Dongxu Li, Silvio Savarese, and Steven C.~H. Hoi.
\newblock {BLIP-2:} bootstrapping language-image pre-training with frozen image encoders and large language models.
\newblock In \emph{Proceedings of the ICML}, pages 19730--19742, 2023{\natexlab{a}}.

\bibitem[Liu et~al.(2023{\natexlab{a}})Liu, Li, Wu, and Lee]{abs-2304-08485}
Haotian Liu, Chunyuan Li, Qingyang Wu, and Yong~Jae Lee.
\newblock Visual instruction tuning.
\newblock \emph{CoRR}, abs/2304.08485, 2023{\natexlab{a}}.

\bibitem[OpenAI(2022{\natexlab{b}})]{gpt4}
OpenAI.
\newblock Gpt-4 technical report.
\newblock 2022{\natexlab{b}}.

\bibitem[Zhu et~al.(2023{\natexlab{a}})Zhu, Chen, Shen, Li, and Elhoseiny]{abs-2304-10592}
Deyao Zhu, Jun Chen, Xiaoqian Shen, Xiang Li, and Mohamed Elhoseiny.
\newblock Minigpt-4: Enhancing vision-language understanding with advanced large language models.
\newblock \emph{CoRR}, abs/2304.10592, 2023{\natexlab{a}}.

\bibitem[Zhang et~al.(2023{\natexlab{a}})Zhang, Li, Zhang, Zhan, Wang, Zhou, and Qiu]{abs-2305-11000}
Dong Zhang, Shimin Li, Xin Zhang, Jun Zhan, Pengyu Wang, Yaqian Zhou, and Xipeng Qiu.
\newblock Speechgpt: Empowering large language models with intrinsic cross-modal conversational abilities.
\newblock \emph{CoRR}, abs/2305.11000, 2023{\natexlab{a}}.

\bibitem[Wu et~al.(2024{\natexlab{a}})Wu, Fei, Qu, Ji, and Chua]{wu2023next}
Shengqiong Wu, Hao Fei, Leigang Qu, Wei Ji, and Tat-Seng Chua.
\newblock Next-gpt: Any-to-any multimodal llm.
\newblock In \emph{Proceedings of the International Conference on Machine Learning}, 2024{\natexlab{a}}.

\bibitem[Chen et~al.(2024{\natexlab{a}})Chen, Chen, Zhang, Li, Yu, Fei, Zhu, Fan, and Chen]{chen2024ll3da}
Sijin Chen, Xin Chen, Chi Zhang, Mingsheng Li, Gang Yu, Hao Fei, Hongyuan Zhu, Jiayuan Fan, and Tao Chen.
\newblock Ll3da: Visual interactive instruction tuning for omni-3d understanding reasoning and planning.
\newblock In \emph{Proceedings of the IEEE/CVF Conference on Computer Vision and Pattern Recognition}, pages 26428--26438, 2024{\natexlab{a}}.

\bibitem[Sun et~al.(2024)Sun, Cui, Zhang, Zhang, Yu, Wang, Rao, Liu, Huang, and Wang]{sun2024generative}
Quan Sun, Yufeng Cui, Xiaosong Zhang, Fan Zhang, Qiying Yu, Yueze Wang, Yongming Rao, Jingjing Liu, Tiejun Huang, and Xinlong Wang.
\newblock Generative multimodal models are in-context learners.
\newblock In \emph{Proceedings of the IEEE/CVF Conference on Computer Vision and Pattern Recognition}, pages 14398--14409, 2024.

\bibitem[Wang et~al.(2023{\natexlab{a}})Wang, Wang, Zhao, Wu, Lyu, Li, Cai, Zhou, Shi, and Tu]{wang2023gpt4video}
Zhanyu Wang, Longyue Wang, Zhen Zhao, Minghao Wu, Chenyang Lyu, Huayang Li, Deng Cai, Luping Zhou, Shuming Shi, and Zhaopeng Tu.
\newblock Gpt4video: A unified multimodal large language model for lnstruction-followed understanding and safety-aware generation.
\newblock \emph{arXiv preprint arXiv:2311.16511}, 2023{\natexlab{a}}.

\bibitem[Munasinghe et~al.(2023)Munasinghe, Thushara, Maaz, Rasheed, Khan, Shah, and Khan]{munasinghe2023pg}
Shehan Munasinghe, Rusiru Thushara, Muhammad Maaz, Hanoona~Abdul Rasheed, Salman Khan, Mubarak Shah, and Fahad Khan.
\newblock Pg-video-llava: Pixel grounding large video-language models.
\newblock \emph{arXiv preprint arXiv:2311.13435}, 2023.

\bibitem[Zhang et~al.(2024{\natexlab{a}})Zhang, Li, Fei, Yuan, Wu, Ji, Loy, and Yan]{zhang2024omg}
Tao Zhang, Xiangtai Li, Hao Fei, Haobo Yuan, Shengqiong Wu, Shunping Ji, Chen~Change Loy, and Shuicheng Yan.
\newblock Omg-llava: Bridging image-level, object-level, pixel-level reasoning and understanding.
\newblock \emph{arXiv preprint arXiv:2406.19389}, 2024{\natexlab{a}}.

\bibitem[Fei et~al.(2024{\natexlab{a}})Fei, Wu, Zhang, Chua, and Yan]{fei2024vitron}
Hao Fei, Shengqiong Wu, Hanwang Zhang, Tat-Seng Chua, and Shuicheng Yan.
\newblock Vitron: A unified pixel-level vision llm for understanding, generating, segmenting, editing.
\newblock 2024{\natexlab{a}}.

\bibitem[Ren et~al.(2023)Ren, Huang, Wei, Zhao, Fu, Feng, and Jin]{ren2023pixellm}
Zhongwei Ren, Zhicheng Huang, Yunchao Wei, Yao Zhao, Dongmei Fu, Jiashi Feng, and Xiaojie Jin.
\newblock Pixellm: Pixel reasoning with large multimodal model.
\newblock \emph{arXiv preprint arXiv:2312.02228}, 2023.

\bibitem[Yuan et~al.(2023{\natexlab{a}})Yuan, Li, Liu, Tang, Luo, Qin, Zhang, and Zhu]{yuan2023osprey}
Yuqian Yuan, Wentong Li, Jian Liu, Dongqi Tang, Xinjie Luo, Chi Qin, Lei Zhang, and Jianke Zhu.
\newblock Osprey: Pixel understanding with visual instruction tuning.
\newblock \emph{arXiv preprint arXiv:2312.10032}, 2023{\natexlab{a}}.

\bibitem[Rasheed et~al.(2023)Rasheed, Maaz, Shaji, Shaker, Khan, Cholakkal, Anwer, Xing, Yang, and Khan]{rasheed2023glamm}
Hanoona Rasheed, Muhammad Maaz, Sahal Shaji, Abdelrahman Shaker, Salman Khan, Hisham Cholakkal, Rao~M Anwer, Erix Xing, Ming-Hsuan Yang, and Fahad~S Khan.
\newblock Glamm: Pixel grounding large multimodal model.
\newblock \emph{arXiv preprint arXiv:2311.03356}, 2023.

\bibitem[Zhan et~al.(2024)Zhan, Dai, Ye, Zhou, Zhang, Liu, Zhang, Yuan, Zhang, Li, et~al.]{zhan2024anygpt}
Jun Zhan, Junqi Dai, Jiasheng Ye, Yunhua Zhou, Dong Zhang, Zhigeng Liu, Xin Zhang, Ruibin Yuan, Ge~Zhang, Linyang Li, et~al.
\newblock Anygpt: Unified multimodal llm with discrete sequence modeling.
\newblock \emph{arXiv preprint arXiv:2402.12226}, 2024.

\bibitem[Lu et~al.(2024{\natexlab{a}})Lu, Clark, Lee, Zhang, Khosla, Marten, Hoiem, and Kembhavi]{lu2024unified}
Jiasen Lu, Christopher Clark, Sangho Lee, Zichen Zhang, Savya Khosla, Ryan Marten, Derek Hoiem, and Aniruddha Kembhavi.
\newblock Unified-io 2: Scaling autoregressive multimodal models with vision language audio and action.
\newblock In \emph{Proceedings of the IEEE/CVF Conference on Computer Vision and Pattern Recognition}, pages 26439--26455, 2024{\natexlab{a}}.

\bibitem[Wu et~al.(2023{\natexlab{a}})Wu, Zhang, Zhang, Chen, Liao, Wang, Li, Sun, Yan, Zhai, et~al.]{wu2023q}
Haoning Wu, Zicheng Zhang, Erli Zhang, Chaofeng Chen, Liang Liao, Annan Wang, Chunyi Li, Wenxiu Sun, Qiong Yan, Guangtao Zhai, et~al.
\newblock Q-bench: A benchmark for general-purpose foundation models on low-level vision.
\newblock \emph{arXiv preprint arXiv:2309.14181}, 2023{\natexlab{a}}.

\bibitem[Xia et~al.(2024{\natexlab{a}})Xia, Han, Qiu, Zhou, Wang, Zheng, Chen, Cui, Ding, Li, et~al.]{xia2024mmie}
Peng Xia, Siwei Han, Shi Qiu, Yiyang Zhou, Zhaoyang Wang, Wenhao Zheng, Zhaorun Chen, Chenhang Cui, Mingyu Ding, Linjie Li, et~al.
\newblock Mmie: Massive multimodal interleaved comprehension benchmark for large vision-language models.
\newblock \emph{arXiv preprint arXiv:2410.10139}, 2024{\natexlab{a}}.

\bibitem[Yue et~al.(2024{\natexlab{a}})Yue, Ni, Zhang, Zheng, Liu, Zhang, Stevens, Jiang, Ren, Sun, et~al.]{yue2024mmmu}
Xiang Yue, Yuansheng Ni, Kai Zhang, Tianyu Zheng, Ruoqi Liu, Ge~Zhang, Samuel Stevens, Dongfu Jiang, Weiming Ren, Yuxuan Sun, et~al.
\newblock Mmmu: A massive multi-discipline multimodal understanding and reasoning benchmark for expert agi.
\newblock In \emph{Proceedings of the IEEE/CVF Conference on Computer Vision and Pattern Recognition}, pages 9556--9567, 2024{\natexlab{a}}.

\bibitem[Meng et~al.(2024{\natexlab{a}})Meng, Wang, Li, Lu, Tian, Liao, Zhu, Dai, Qiao, Luo, et~al.]{meng2024mmiu}
Fanqing Meng, Jin Wang, Chuanhao Li, Quanfeng Lu, Hao Tian, Jiaqi Liao, Xizhou Zhu, Jifeng Dai, Yu~Qiao, Ping Luo, et~al.
\newblock Mmiu: Multimodal multi-image understanding for evaluating large vision-language models.
\newblock \emph{arXiv preprint arXiv:2408.02718}, 2024{\natexlab{a}}.

\bibitem[Liu et~al.(2025)Liu, Duan, Zhang, Li, Zhang, Zhao, Yuan, Wang, He, Liu, et~al.]{liu2025mmbench}
Yuan Liu, Haodong Duan, Yuanhan Zhang, Bo~Li, Songyang Zhang, Wangbo Zhao, Yike Yuan, Jiaqi Wang, Conghui He, Ziwei Liu, et~al.
\newblock Mmbench: Is your multi-modal model an all-around player?
\newblock In \emph{European Conference on Computer Vision}, pages 216--233. Springer, 2025.

\bibitem[Li et~al.(2024{\natexlab{a}})Li, Ge, Ge, Wang, Wang, Zhang, and Shan]{li2024seed}
Bohao Li, Yuying Ge, Yixiao Ge, Guangzhi Wang, Rui Wang, Ruimao Zhang, and Ying Shan.
\newblock Seed-bench: Benchmarking multimodal large language models.
\newblock In \emph{Proceedings of the IEEE/CVF Conference on Computer Vision and Pattern Recognition}, pages 13299--13308, 2024{\natexlab{a}}.

\bibitem[Ying et~al.(2024{\natexlab{a}})Ying, Meng, Wang, Li, Lin, Yang, Zhang, Zhang, Lin, Liu, et~al.]{ying2024mmt}
Kaining Ying, Fanqing Meng, Jin Wang, Zhiqian Li, Han Lin, Yue Yang, Hao Zhang, Wenbo Zhang, Yuqi Lin, Shuo Liu, et~al.
\newblock Mmt-bench: A comprehensive multimodal benchmark for evaluating large vision-language models towards multitask agi.
\newblock \emph{arXiv preprint arXiv:2404.16006}, 2024{\natexlab{a}}.

\bibitem[Li et~al.(2024{\natexlab{b}})Li, Lu, Fei, Luo, Dai, Xia, Jin, Gan, Qi, Fu, Tai, Yang, Wang, and Wang]{li2024survey}
Jian Li, Weiheng Lu, Hao Fei, Meng Luo, Ming Dai, Min Xia, Yizhang Jin, Zhenye Gan, Ding Qi, Chaoyou Fu, Ying Tai, Wankou Yang, Yabiao Wang, and Chengjie Wang.
\newblock A survey on benchmarks of multimodal large language models.
\newblock \emph{arXiv preprint arXiv:2408.08632}, 2024{\natexlab{b}}.

\bibitem[Xu et~al.(2023{\natexlab{a}})Xu, Shao, Zhang, Gao, Liu, Lei, Meng, Huang, Qiao, and Luo]{xu2023lvlm}
Peng Xu, Wenqi Shao, Kaipeng Zhang, Peng Gao, Shuo Liu, Meng Lei, Fanqing Meng, Siyuan Huang, Yu~Qiao, and Ping Luo.
\newblock Lvlm-ehub: A comprehensive evaluation benchmark for large vision-language models.
\newblock \emph{arXiv preprint arXiv:2306.09265}, 2023{\natexlab{a}}.

\bibitem[Yu et~al.(2023)Yu, Yang, Li, Wang, Lin, Liu, Wang, and Wang]{yu2023mm}
Weihao Yu, Zhengyuan Yang, Linjie Li, Jianfeng Wang, Kevin Lin, Zicheng Liu, Xinchao Wang, and Lijuan Wang.
\newblock Mm-vet: Evaluating large multimodal models for integrated capabilities.
\newblock \emph{arXiv preprint arXiv:2308.02490}, 2023.

\bibitem[Fu et~al.(2024{\natexlab{a}})Fu, Chen, Shen, Qin, Zhang, Lin, Yang, Zheng, Li, Sun, Wu, and Ji]{fu2024MME}
Chaoyou Fu, Peixian Chen, Yunhang Shen, Yulei Qin, Mengdan Zhang, Xu~Lin, Jinrui Yang, Xiawu Zheng, Ke~Li, Xing Sun, Yunsheng Wu, and Rongrong Ji.
\newblock Mme: A comprehensive evaluation benchmark for multimodal large language models.
\newblock \emph{CoRR}, abs/2306.13394, 2024{\natexlab{a}}.

\bibitem[Chen et~al.(2024{\natexlab{b}})Chen, Liang, Siu, Wang, Wang, Wang, Ni, Zhu, Jiang, Lyu, et~al.]{chen2024mega}
Jiacheng Chen, Tianhao Liang, Sherman Siu, Zhengqing Wang, Kai Wang, Yubo Wang, Yuansheng Ni, Wang Zhu, Ziyan Jiang, Bohan Lyu, et~al.
\newblock Mega-bench: Scaling multimodal evaluation to over 500 real-world tasks.
\newblock \emph{arXiv preprint arXiv:2410.10563}, 2024{\natexlab{b}}.

\bibitem[Yurtsever et~al.(2020)Yurtsever, Lambert, Carballo, and Takeda]{yurtsever2020survey}
Ekim Yurtsever, Jacob Lambert, Alexander Carballo, and Kazuya Takeda.
\newblock A survey of autonomous driving: Common practices and emerging technologies.
\newblock \emph{IEEE access}, 8:\penalty0 58443--58469, 2020.

\bibitem[Bai et~al.(2023)Bai, Bai, Yang, Wang, Tan, Wang, Lin, Zhou, and Zhou]{bai2023qwen}
Jinze Bai, Shuai Bai, Shusheng Yang, Shijie Wang, Sinan Tan, Peng Wang, Junyang Lin, Chang Zhou, and Jingren Zhou.
\newblock Qwen-vl: A frontier large vision-language model with versatile abilities.
\newblock \emph{arXiv preprint arXiv:2308.12966}, 2023.

\bibitem[Zhang et~al.(2023{\natexlab{b}})Zhang, Dong, Wang, Cao, Xu, Ouyang, Zhao, Duan, Zhang, Ding, et~al.]{zhang2023internlm}
Pan Zhang, Xiaoyi Dong, Bin Wang, Yuhang Cao, Chao Xu, Linke Ouyang, Zhiyuan Zhao, Haodong Duan, Songyang Zhang, Shuangrui Ding, et~al.
\newblock Internlm-xcomposer: A vision-language large model for advanced text-image comprehension and composition.
\newblock \emph{arXiv preprint arXiv:2309.15112}, 2023{\natexlab{b}}.

\bibitem[Jin et~al.(2023)Jin, Xu, Chen, Liao, Tan, Chen, Lei, Liu, Song, Lei, et~al.]{jin2023unified}
Yang Jin, Kun Xu, Liwei Chen, Chao Liao, Jianchao Tan, Bin Chen, Chenyi Lei, An~Liu, Chengru Song, Xiaoqiang Lei, et~al.
\newblock Unified language-vision pretraining with dynamic discrete visual tokenization.
\newblock \emph{arXiv preprint arXiv:2309.04669}, 2023.

\bibitem[Li et~al.(2024{\natexlab{c}})Li, Zhang, Wang, Zhong, Chen, Chu, Liu, and Jia]{li2024mini}
Yanwei Li, Yuechen Zhang, Chengyao Wang, Zhisheng Zhong, Yixin Chen, Ruihang Chu, Shaoteng Liu, and Jiaya Jia.
\newblock Mini-gemini: Mining the potential of multi-modality vision language models.
\newblock \emph{arXiv preprint arXiv:2403.18814}, 2024{\natexlab{c}}.

\bibitem[Fei et~al.(2024{\natexlab{b}})Fei, Yao, Zhang, Liu, Zhang, and Chua]{fei2024multimodal}
Hao Fei, Yuan Yao, Zhuosheng Zhang, Fuxiao Liu, Ao~Zhang, and Tat-Seng Chua.
\newblock From multimodal llm to human-level ai: Modality, instruction, reasoning, efficiency and beyond.
\newblock In \emph{Proceedings of the 2024 Joint International Conference on Computational Linguistics, Language Resources and Evaluation (LREC-COLING 2024): Tutorial Summaries}, pages 1--8, 2024{\natexlab{b}}.

\bibitem[Fei et~al.(2024{\natexlab{c}})Fei, Wu, Zhang, Zhang, Chua, and Yan]{fei2024enhancing}
Hao Fei, Shengqiong Wu, Meishan Zhang, Min Zhang, Tat-Seng Chua, and Shuicheng Yan.
\newblock Enhancing video-language representations with structural spatio-temporal alignment.
\newblock \emph{IEEE Transactions on Pattern Analysis and Machine Intelligence}, 2024{\natexlab{c}}.

\bibitem[Van Den~Oord et~al.(2016)Van Den~Oord, Dieleman, Zen, Simonyan, Vinyals, Graves, Kalchbrenner, Senior, Kavukcuoglu, et~al.]{van2016wavenet}
Aaron Van Den~Oord, Sander Dieleman, Heiga Zen, Karen Simonyan, Oriol Vinyals, Alex Graves, Nal Kalchbrenner, Andrew Senior, Koray Kavukcuoglu, et~al.
\newblock Wavenet: A generative model for raw audio.
\newblock \emph{arXiv preprint arXiv:1609.03499}, 12, 2016.

\bibitem[Radford et~al.(2021)Radford, Kim, Hallacy, Ramesh, Goh, Agarwal, Sastry, Askell, Mishkin, Clark, et~al.]{radford2021learning}
Alec Radford, Jong~Wook Kim, Chris Hallacy, Aditya Ramesh, Gabriel Goh, Sandhini Agarwal, Girish Sastry, Amanda Askell, Pamela Mishkin, Jack Clark, et~al.
\newblock Learning transferable visual models from natural language supervision.
\newblock In \emph{International conference on machine learning}, pages 8748--8763. PMLR, 2021.

\bibitem[Rombach et~al.(2022)Rombach, Blattmann, Lorenz, Esser, and Ommer]{rombach2022high}
Robin Rombach, Andreas Blattmann, Dominik Lorenz, Patrick Esser, and Bj{\"o}rn Ommer.
\newblock High-resolution image synthesis with latent diffusion models.
\newblock In \emph{Proceedings of the IEEE/CVF conference on computer vision and pattern recognition}, pages 10684--10695, 2022.

\bibitem[Liu et~al.(2023{\natexlab{b}})Liu, Zeng, Ren, Li, Zhang, Yang, Li, Yang, Su, Zhu, et~al.]{liu2023grounding}
Shilong Liu, Zhaoyang Zeng, Tianhe Ren, Feng Li, Hao Zhang, Jie Yang, Chunyuan Li, Jianwei Yang, Hang Su, Jun Zhu, et~al.
\newblock Grounding dino: Marrying dino with grounded pre-training for open-set object detection.
\newblock \emph{arXiv preprint arXiv:2303.05499}, 2023{\natexlab{b}}.

\bibitem[Li et~al.(2022)Li, Li, Xiong, and Hoi]{0001LXH22}
Junnan Li, Dongxu Li, Caiming Xiong, and Steven C.~H. Hoi.
\newblock {BLIP:} bootstrapping language-image pre-training for unified vision-language understanding and generation.
\newblock In \emph{Proceedings of the ICML}, pages 12888--12900, 2022.

\bibitem[Labs(2023)]{flux2023}
Black~Forest Labs.
\newblock Flux.
\newblock \url{https://github.com/black-forest-labs/flux}, 2023.

\bibitem[Ren et~al.(2021)Ren, Hu, Tan, Qin, Zhao, Zhao, and Liu]{FastSpeech-0006H0QZZL21}
Yi~Ren, Chenxu Hu, Xu~Tan, Tao Qin, Sheng Zhao, Zhou Zhao, and Tie{-}Yan Liu.
\newblock Fastspeech 2: Fast and high-quality end-to-end text to speech.
\newblock In \emph{9th International Conference on Learning Representations, {ICLR} 2021, Virtual Event, Austria, May 3-7, 2021}, 2021.

\bibitem[Ge et~al.(2023)Ge, Zhao, Zeng, Ge, Li, Wang, and Shan]{ge2023making}
Yuying Ge, Sijie Zhao, Ziyun Zeng, Yixiao Ge, Chen Li, Xintao Wang, and Ying Shan.
\newblock Making llama see and draw with seed tokenizer.
\newblock \emph{arXiv preprint arXiv:2310.01218}, 2023.

\bibitem[Team et~al.(2024{\natexlab{a}})Team, Georgiev, Lei, Burnell, Bai, Gulati, Tanzer, Vincent, Pan, Wang, et~al.]{team2024gemini}
Gemini Team, Petko Georgiev, Ving~Ian Lei, Ryan Burnell, Libin Bai, Anmol Gulati, Garrett Tanzer, Damien Vincent, Zhufeng Pan, Shibo Wang, et~al.
\newblock Gemini 1.5: Unlocking multimodal understanding across millions of tokens of context.
\newblock \emph{arXiv preprint arXiv:2403.05530}, 2024{\natexlab{a}}.

\bibitem[Team(2024)]{TheClaude3}
Anthropic Team.
\newblock The claude 3 model family: Opus, sonnet, haiku.
\newblock \emph{preprint}, 2024.
\newblock URL \url{https://api.semanticscholar.org/CorpusID:268232499}.

\bibitem[Lu et~al.(2024{\natexlab{b}})Lu, Liu, Zhang, Wang, Dong, Liu, Sun, Ren, Li, Yang, et~al.]{lu2024deepseek}
Haoyu Lu, Wen Liu, Bo~Zhang, Bingxuan Wang, Kai Dong, Bo~Liu, Jingxiang Sun, Tongzheng Ren, Zhuoshu Li, Hao Yang, et~al.
\newblock Deepseek-vl: towards real-world vision-language understanding.
\newblock \emph{arXiv preprint arXiv:2403.05525}, 2024{\natexlab{b}}.

\bibitem[Li et~al.(2024{\natexlab{d}})Li, Zhang, Guo, Zhang, Li, Zhang, Zhang, Zhang, Li, Liu, et~al.]{li2024llavaonevision}
Bo~Li, Yuanhan Zhang, Dong Guo, Renrui Zhang, Feng Li, Hao Zhang, Kaichen Zhang, Peiyuan Zhang, Yanwei Li, Ziwei Liu, et~al.
\newblock Llava-onevision: Easy visual task transfer.
\newblock \emph{arXiv preprint arXiv:2408.03326}, 2024{\natexlab{d}}.

\bibitem[Wang et~al.(2024{\natexlab{a}})Wang, Bai, Tan, Wang, Fan, Bai, Chen, Liu, Wang, Ge, et~al.]{wang2024qwen2}
Peng Wang, Shuai Bai, Sinan Tan, Shijie Wang, Zhihao Fan, Jinze Bai, Keqin Chen, Xuejing Liu, Jialin Wang, Wenbin Ge, et~al.
\newblock Qwen2-vl: Enhancing vision-language model's perception of the world at any resolution.
\newblock \emph{arXiv preprint arXiv:2409.12191}, 2024{\natexlab{a}}.

\bibitem[Chen et~al.(2024{\natexlab{c}})Chen, Wu, Wang, Su, Chen, Xing, Zhong, Zhang, Zhu, Lu, et~al.]{chen2024internvl}
Zhe Chen, Jiannan Wu, Wenhai Wang, Weijie Su, Guo Chen, Sen Xing, Muyan Zhong, Qinglong Zhang, Xizhou Zhu, Lewei Lu, et~al.
\newblock Internvl: Scaling up vision foundation models and aligning for generic visual-linguistic tasks.
\newblock In \emph{Proceedings of the IEEE/CVF Conference on Computer Vision and Pattern Recognition}, pages 24185--24198, 2024{\natexlab{c}}.

\bibitem[Abdin et~al.(2024)Abdin, Aneja, Awadalla, Awadallah, Awan, Bach, Bahree, Bakhtiari, Bao, Behl, et~al.]{abdin2024phi}
Marah Abdin, Jyoti Aneja, Hany Awadalla, Ahmed Awadallah, Ammar~Ahmad Awan, Nguyen Bach, Amit Bahree, Arash Bakhtiari, Jianmin Bao, Harkirat Behl, et~al.
\newblock Phi-3 technical report: A highly capable language model locally on your phone.
\newblock \emph{arXiv preprint arXiv:2404.14219}, 2024.

\bibitem[Wu et~al.(2023{\natexlab{b}})Wu, Yin, Qi, Wang, Tang, and Duan]{abs-2303-04671}
Chenfei Wu, Shengming Yin, Weizhen Qi, Xiaodong Wang, Zecheng Tang, and Nan Duan.
\newblock Visual chatgpt: Talking, drawing and editing with visual foundation models.
\newblock \emph{CoRR}, abs/2303.04671, 2023{\natexlab{b}}.

\bibitem[Shen et~al.(2023)Shen, Song, Tan, Li, Lu, and Zhuang]{abs-2303-17580}
Yongliang Shen, Kaitao Song, Xu~Tan, Dongsheng Li, Weiming Lu, and Yueting Zhuang.
\newblock Hugginggpt: Solving {AI} tasks with chatgpt and its friends in huggingface.
\newblock \emph{CoRR}, abs/2303.17580, 2023.

\bibitem[Liu et~al.(2023{\natexlab{c}})Liu, Cheng, Liu, Zhang, Li, Ren, Zou, Yang, Su, Zhu, et~al.]{liu2023llava}
Shilong Liu, Hao Cheng, Haotian Liu, Hao Zhang, Feng Li, Tianhe Ren, Xueyan Zou, Jianwei Yang, Hang Su, Jun Zhu, et~al.
\newblock Llava-plus: Learning to use tools for creating multimodal agents.
\newblock \emph{arXiv preprint arXiv:2311.05437}, 2023{\natexlab{c}}.

\bibitem[Pan et~al.(2024)Pan, Tang, Li, Fan, Chow, Yan, Chua, Zhuang, and Zhang]{pan2024auto}
Kaihang Pan, Siliang Tang, Juncheng Li, Zhaoyu Fan, Wei Chow, Shuicheng Yan, Tat-Seng Chua, Yueting Zhuang, and Hanwang Zhang.
\newblock Auto-encoding morph-tokens for multimodal llm.
\newblock \emph{arXiv preprint arXiv:2405.01926}, 2024.

\bibitem[Team et~al.(2024{\natexlab{b}})Team, Riviere, Pathak, Sessa, Hardin, Bhupatiraju, Hussenot, Mesnard, Shahriari, Ram{\'e}, et~al.]{team2024gemma}
Gemma Team, Morgane Riviere, Shreya Pathak, Pier~Giuseppe Sessa, Cassidy Hardin, Surya Bhupatiraju, L{\'e}onard Hussenot, Thomas Mesnard, Bobak Shahriari, Alexandre Ram{\'e}, et~al.
\newblock Gemma 2: Improving open language models at a practical size.
\newblock \emph{arXiv preprint arXiv:2408.00118}, 2024{\natexlab{b}}.

\bibitem[Wang and Komatsuzaki(2021)]{gpt-j}
Ben Wang and Aran Komatsuzaki.
\newblock {GPT-J-6B: A 6 Billion Parameter Autoregressive Language Model}.
\newblock \url{https://github.com/kingoflolz/mesh-transformer-jax}, May 2021.

\bibitem[GLM et~al.(2024)GLM, Zeng, Xu, Wang, Zhang, Yin, Zhang, Rojas, Feng, Zhao, et~al.]{glm2024chatglm}
Team GLM, Aohan Zeng, Bin Xu, Bowen Wang, Chenhui Zhang, Da~Yin, Dan Zhang, Diego Rojas, Guanyu Feng, Hanlin Zhao, et~al.
\newblock Chatglm: A family of large language models from glm-130b to glm-4 all tools.
\newblock \emph{arXiv preprint arXiv:2406.12793}, 2024.

\bibitem[Yang et~al.(2024{\natexlab{a}})Yang, Yang, Zhang, Hui, Zheng, Yu, Li, Liu, Huang, Wei, et~al.]{yang2024qwen2}
An~Yang, Baosong Yang, Beichen Zhang, Binyuan Hui, Bo~Zheng, Bowen Yu, Chengyuan Li, Dayiheng Liu, Fei Huang, Haoran Wei, et~al.
\newblock Qwen2. 5 technical report.
\newblock \emph{arXiv preprint arXiv:2412.15115}, 2024{\natexlab{a}}.

\bibitem[Cai et~al.(2024)Cai, Cao, Chen, Chen, Chen, Chen, Chen, Chen, Chen, Chu, et~al.]{cai2024internlm2}
Zheng Cai, Maosong Cao, Haojiong Chen, Kai Chen, Keyu Chen, Xin Chen, Xun Chen, Zehui Chen, Zhi Chen, Pei Chu, et~al.
\newblock Internlm2 technical report.
\newblock \emph{arXiv preprint arXiv:2403.17297}, 2024.

\bibitem[Yang et~al.(2023)Yang, Xiao, Wang, Zhang, Bian, Yin, Lv, Pan, Wang, Yan, et~al.]{yang2023baichuan}
Aiyuan Yang, Bin Xiao, Bingning Wang, Borong Zhang, Ce~Bian, Chao Yin, Chenxu Lv, Da~Pan, Dian Wang, Dong Yan, et~al.
\newblock Baichuan 2: Open large-scale language models.
\newblock \emph{arXiv preprint arXiv:2309.10305}, 2023.

\bibitem[Chiang et~al.(2023)Chiang, Li, Lin, Sheng, Wu, Zhang, Zheng, Zhuang, Zhuang, Gonzalez, et~al.]{chiang2023vicuna}
Wei-Lin Chiang, Zhuohan Li, Zi~Lin, Ying Sheng, Zhanghao Wu, Hao Zhang, Lianmin Zheng, Siyuan Zhuang, Yonghao Zhuang, Joseph~E Gonzalez, et~al.
\newblock Vicuna: An open-source chatbot impressing gpt-4 with 90\%* chatgpt quality.
\newblock \emph{See https://vicuna. lmsys. org (accessed 14 April 2023)}, 2\penalty0 (3):\penalty0 6, 2023.

\bibitem[Almazrouei et~al.(2023)Almazrouei, Alobeidli, Alshamsi, Cappelli, Cojocaru, Debbah, Goffinet, Heslow, Launay, Malartic, Noune, Pannier, and Penedo]{falcon40b}
Ebtesam Almazrouei, Hamza Alobeidli, Abdulaziz Alshamsi, Alessandro Cappelli, Ruxandra Cojocaru, Merouane Debbah, Etienne Goffinet, Daniel Heslow, Julien Launay, Quentin Malartic, Badreddine Noune, Baptiste Pannier, and Guilherme Penedo.
\newblock {Falcon-40B}: an open large language model with state-of-the-art performance.
\newblock 2023.

\bibitem[Jiang et~al.(2024{\natexlab{a}})Jiang, Sablayrolles, Roux, Mensch, Savary, Bamford, Chaplot, Casas, Hanna, Bressand, et~al.]{jiang2024mixtral}
Albert~Q Jiang, Alexandre Sablayrolles, Antoine Roux, Arthur Mensch, Blanche Savary, Chris Bamford, Devendra~Singh Chaplot, Diego de~las Casas, Emma~Bou Hanna, Florian Bressand, et~al.
\newblock Mixtral of experts.
\newblock \emph{arXiv preprint arXiv:2401.04088}, 2024{\natexlab{a}}.

\bibitem[Young et~al.(2024)Young, Chen, Li, Huang, Zhang, Zhang, Li, Zhu, Chen, Chang, et~al.]{young2024yi}
Alex Young, Bei Chen, Chao Li, Chengen Huang, Ge~Zhang, Guanwei Zhang, Heng Li, Jiangcheng Zhu, Jianqun Chen, Jing Chang, et~al.
\newblock Yi: Open foundation models by 01. ai.
\newblock \emph{arXiv preprint arXiv:2403.04652}, 2024.

\bibitem[Gao et~al.(2024)Gao, Chen, Cui, Ren, Wang, Zhu, Tian, Ye, He, Zhu, et~al.]{gao2024mini}
Zhangwei Gao, Zhe Chen, Erfei Cui, Yiming Ren, Weiyun Wang, Jinguo Zhu, Hao Tian, Shenglong Ye, Junjun He, Xizhou Zhu, et~al.
\newblock Mini-internvl: a flexible-transfer pocket multi-modal model with 5\% parameters and 90\% performance.
\newblock \emph{Visual Intelligence}, 2\penalty0 (1):\penalty0 1--17, 2024.

\bibitem[Dong et~al.(2024)Dong, Zhang, Zang, Cao, Wang, Ouyang, Wei, Zhang, Duan, Cao, et~al.]{dong2024internlm}
Xiaoyi Dong, Pan Zhang, Yuhang Zang, Yuhang Cao, Bin Wang, Linke Ouyang, Xilin Wei, Songyang Zhang, Haodong Duan, Maosong Cao, et~al.
\newblock Internlm-xcomposer2: Mastering free-form text-image composition and comprehension in vision-language large model.
\newblock \emph{arXiv preprint arXiv:2401.16420}, 2024.

\bibitem[Lin et~al.(2024{\natexlab{a}})Lin, Tang, Ye, Cui, Zhu, Jin, Zhang, Ning, and Yuan]{lin2024moe}
Bin Lin, Zhenyu Tang, Yang Ye, Jiaxi Cui, Bin Zhu, Peng Jin, Junwu Zhang, Munan Ning, and Li~Yuan.
\newblock Moe-llava: Mixture of experts for large vision-language models.
\newblock \emph{arXiv preprint arXiv:2401.15947}, 2024{\natexlab{a}}.

\bibitem[Li et~al.(2024{\natexlab{e}})Li, Yang, Liu, Ma, Zhang, Yang, Sun, Liu, and Bai]{li2024monkey}
Zhang Li, Biao Yang, Qiang Liu, Zhiyin Ma, Shuo Zhang, Jingxu Yang, Yabo Sun, Yuliang Liu, and Xiang Bai.
\newblock Monkey: Image resolution and text label are important things for large multi-modal models.
\newblock In \emph{Proceedings of the IEEE/CVF Conference on Computer Vision and Pattern Recognition}, pages 26763--26773, 2024{\natexlab{e}}.

\bibitem[Ye et~al.(2024)Ye, Xu, Ye, Yan, Hu, Liu, Qian, Zhang, and Huang]{ye2024mplug}
Qinghao Ye, Haiyang Xu, Jiabo Ye, Ming Yan, Anwen Hu, Haowei Liu, Qi~Qian, Ji~Zhang, and Fei Huang.
\newblock mplug-owl2: Revolutionizing multi-modal large language model with modality collaboration.
\newblock In \emph{Proceedings of the IEEE/CVF Conference on Computer Vision and Pattern Recognition}, pages 13040--13051, 2024.

\bibitem[Tong et~al.(2024{\natexlab{a}})Tong, Brown, Wu, Woo, Middepogu, Akula, Yang, Yang, Iyer, Pan, et~al.]{tong2024cambrian}
Shengbang Tong, Ellis Brown, Penghao Wu, Sanghyun Woo, Manoj Middepogu, Sai~Charitha Akula, Jihan Yang, Shusheng Yang, Adithya Iyer, Xichen Pan, et~al.
\newblock Cambrian-1: A fully open, vision-centric exploration of multimodal llms.
\newblock \emph{arXiv preprint arXiv:2406.16860}, 2024{\natexlab{a}}.

\bibitem[Pi et~al.(2023)Pi, Gao, Diao, Pan, Dong, Zhang, Yao, Han, Xu, Kong, et~al.]{pi2023detgpt}
Renjie Pi, Jiahui Gao, Shizhe Diao, Rui Pan, Hanze Dong, Jipeng Zhang, Lewei Yao, Jianhua Han, Hang Xu, Lingpeng Kong, et~al.
\newblock Detgpt: Detect what you need via reasoning.
\newblock \emph{arXiv preprint arXiv:2305.14167}, 2023.

\bibitem[Li et~al.(2023{\natexlab{b}})Li, Zhang, Chen, Wang, Yang, and Liu]{li2023otter}
Bo~Li, Yuanhan Zhang, Liangyu Chen, Jinghao Wang, Jingkang Yang, and Ziwei Liu.
\newblock Otter: A multi-modal model with in-context instruction tuning.
\newblock \emph{arXiv preprint arXiv:2305.03726}, 2023{\natexlab{b}}.

\bibitem[Zhang et~al.(2023{\natexlab{c}})Zhang, Zhao, Xie, Zheng, Ji, and Chua]{zhang2023next}
Ao~Zhang, Liming Zhao, Chen-Wei Xie, Yun Zheng, Wei Ji, and Tat-Seng Chua.
\newblock Next-chat: An lmm for chat, detection and segmentation.
\newblock \emph{arXiv preprint arXiv:2311.04498}, 2023{\natexlab{c}}.

\bibitem[Zhang et~al.(2023{\natexlab{d}})Zhang, Sun, Chen, Xiao, Shao, Zhang, Chen, and Luo]{zhang2023gpt4roi}
Shilong Zhang, Peize Sun, Shoufa Chen, Min Xiao, Wenqi Shao, Wenwei Zhang, Kai Chen, and Ping Luo.
\newblock Gpt4roi: Instruction tuning large language model on region-of-interest.
\newblock \emph{arXiv preprint arXiv:2307.03601}, 2023{\natexlab{d}}.

\bibitem[Rasheed et~al.(2024)Rasheed, Maaz, Shaji, Shaker, Khan, Cholakkal, Anwer, Xing, Yang, and Khan]{rasheed2024glamm}
Hanoona Rasheed, Muhammad Maaz, Sahal Shaji, Abdelrahman Shaker, Salman Khan, Hisham Cholakkal, Rao~M Anwer, Eric Xing, Ming-Hsuan Yang, and Fahad~S Khan.
\newblock Glamm: Pixel grounding large multimodal model.
\newblock In \emph{Proceedings of the IEEE/CVF Conference on Computer Vision and Pattern Recognition}, pages 13009--13018, 2024.

\bibitem[Agrawal et~al.(2024)Agrawal, Antoniak, Hanna, Bout, Chaplot, Chudnovsky, Costa, De~Monicault, Garg, Gervet, et~al.]{agrawal2024pixtral}
Pravesh Agrawal, Szymon Antoniak, Emma~Bou Hanna, Baptiste Bout, Devendra Chaplot, Jessica Chudnovsky, Diogo Costa, Baudouin De~Monicault, Saurabh Garg, Theophile Gervet, et~al.
\newblock Pixtral 12b.
\newblock \emph{arXiv preprint arXiv:2410.07073}, 2024.

\bibitem[Xue et~al.(2024)Xue, Shu, Awadalla, Wang, Yan, Purushwalkam, Zhou, Prabhu, Dai, Ryoo, et~al.]{xue2024xgen}
Le~Xue, Manli Shu, Anas Awadalla, Jun Wang, An~Yan, Senthil Purushwalkam, Honglu Zhou, Viraj Prabhu, Yutong Dai, Michael~S Ryoo, et~al.
\newblock xgen-mm (blip-3): A family of open large multimodal models.
\newblock \emph{arXiv preprint arXiv:2408.08872}, 2024.

\bibitem[Xie et~al.(2024)Xie, Mao, Bai, Zhang, Wang, Lin, Gu, Chen, Yang, and Shou]{xie2024showo}
Jinheng Xie, Weijia Mao, Zechen Bai, David~Junhao Zhang, Weihao Wang, Kevin~Qinghong Lin, Yuchao Gu, Zhijie Chen, Zhenheng Yang, and Mike~Zheng Shou.
\newblock Show-o: One single transformer to unify multimodal understanding and generation.
\newblock \emph{arXiv preprint arXiv:2408.12528}, 2024.

\bibitem[Lai et~al.(2024)Lai, Tian, Chen, Li, Yuan, Liu, and Jia]{lai2024lisa}
Xin Lai, Zhuotao Tian, Yukang Chen, Yanwei Li, Yuhui Yuan, Shu Liu, and Jiaya Jia.
\newblock Lisa: Reasoning segmentation via large language model.
\newblock In \emph{Proceedings of the IEEE/CVF Conference on Computer Vision and Pattern Recognition}, pages 9579--9589, 2024.

\bibitem[Wang et~al.(2023{\natexlab{b}})Wang, Lv, Yu, Hong, Qi, Wang, Ji, Yang, Zhao, Song, et~al.]{wang2023cogvlm}
Weihan Wang, Qingsong Lv, Wenmeng Yu, Wenyi Hong, Ji~Qi, Yan Wang, Junhui Ji, Zhuoyi Yang, Lei Zhao, Xixuan Song, et~al.
\newblock Cogvlm: Visual expert for pretrained language models.
\newblock \emph{arXiv preprint arXiv:2311.03079}, 2023{\natexlab{b}}.

\bibitem[Chen et~al.(2025)Chen, Li, Dong, Zhang, He, Wang, Zhao, and Lin]{chen2025sharegpt4v}
Lin Chen, Jinsong Li, Xiaoyi Dong, Pan Zhang, Conghui He, Jiaqi Wang, Feng Zhao, and Dahua Lin.
\newblock Sharegpt4v: Improving large multi-modal models with better captions.
\newblock In \emph{European Conference on Computer Vision}, pages 370--387. Springer, 2025.

\bibitem[Du et~al.(2021)Du, Qian, Liu, Ding, Qiu, Yang, and Tang]{du2021glm}
Zhengxiao Du, Yujie Qian, Xiao Liu, Ming Ding, Jiezhong Qiu, Zhilin Yang, and Jie Tang.
\newblock Glm: General language model pretraining with autoregressive blank infilling.
\newblock \emph{arXiv preprint arXiv:2103.10360}, 2021.

\bibitem[Lauren{\c{c}}on et~al.(2024)Lauren{\c{c}}on, Marafioti, Sanh, and Tronchon]{laurenccon2024building}
Hugo Lauren{\c{c}}on, Andr{\'e}s Marafioti, Victor Sanh, and L{\'e}o Tronchon.
\newblock Building and better understanding vision-language models: insights and future directions.
\newblock In \emph{Workshop on Responsibly Building the Next Generation of Multimodal Foundational Models}, 2024.

\bibitem[Hu et~al.(2024{\natexlab{a}})Hu, Tu, Han, He, Cui, Long, Zheng, Fang, Huang, Zhao, et~al.]{hu2024minicpm}
Shengding Hu, Yuge Tu, Xu~Han, Chaoqun He, Ganqu Cui, Xiang Long, Zhi Zheng, Yewei Fang, Yuxiang Huang, Weilin Zhao, et~al.
\newblock Minicpm: Unveiling the potential of small language models with scalable training strategies.
\newblock \emph{arXiv preprint arXiv:2404.06395}, 2024{\natexlab{a}}.

\bibitem[Jin et~al.()Jin, Xu, Xu, Chen, Liao, Tan, Huang, Chen, Lei, Liu, et~al.]{jin2309unified}
Yang Jin, Kun Xu, Kun Xu, Liwei Chen, Chao Liao, Jianchao Tan, Quzhe Huang, Bin Chen, Chenyi Lei, An~Liu, et~al.
\newblock Unified language-vision pretraining in llm with dynamic discrete visual tokenization. arxiv 2024.
\newblock \emph{arXiv preprint arXiv:2309.04669}.

\bibitem[Zheng et~al.(2024)Zheng, Gu, Li, and Dong]{zheng2024lm4lv}
Boyang Zheng, Jinjin Gu, Shijun Li, and Chao Dong.
\newblock Lm4lv: A frozen large language model for low-level vision tasks.
\newblock \emph{arXiv preprint arXiv:2405.15734}, 2024.

\bibitem[Zhou et~al.(2025)Zhou, Zhang, Xu, Chen, Zhou, Tong, Ji, Zhang, Li, and Qi]{zhou2025CoLVA}
Yikang Zhou, Tao Zhang, Shilin Xu, Shihao Chen, Qianyu Zhou, Yunhai Tong, Shunping Ji, Jiangning Zhang, Xiangtai Li, and Lu~Qi.
\newblock Are they the same? exploring visual correspondence shortcomings of multimodal llms, 2025.

\bibitem[Wang et~al.(2024{\natexlab{b}})Wang, Song, Chen, Zhang, and Wang]{wang2024longllava}
Xidong Wang, Dingjie Song, Shunian Chen, Chen Zhang, and Benyou Wang.
\newblock Longllava: Scaling multi-modal llms to 1000 images efficiently via a hybrid architecture.
\newblock \emph{arXiv preprint arXiv:2409.02889}, 2024{\natexlab{b}}.

\bibitem[Chu et~al.(2023)Chu, Xu, Zhou, Yang, Zhang, Yan, Zhou, and Zhou]{chu2023qwen}
Yunfei Chu, Jin Xu, Xiaohuan Zhou, Qian Yang, Shiliang Zhang, Zhijie Yan, Chang Zhou, and Jingren Zhou.
\newblock Qwen-audio: Advancing universal audio understanding via unified large-scale audio-language models.
\newblock \emph{arXiv preprint arXiv:2311.07919}, 2023.

\bibitem[Chu et~al.(2024)Chu, Xu, Yang, Wei, Wei, Guo, Leng, Lv, He, Lin, et~al.]{chu2024qwen2}
Yunfei Chu, Jin Xu, Qian Yang, Haojie Wei, Xipin Wei, Zhifang Guo, Yichong Leng, Yuanjun Lv, Jinzheng He, Junyang Lin, et~al.
\newblock Qwen2-audio technical report.
\newblock \emph{arXiv preprint arXiv:2407.10759}, 2024.

\bibitem[Liu et~al.(2024{\natexlab{a}})Liu, Li, Li, Li, Zhang, Shen, and Lee]{liu2024llava}
Haotian Liu, Chunyuan Li, Yuheng Li, Bo~Li, Yuanhan Zhang, Sheng Shen, and Yong~Jae Lee.
\newblock Llava-next: Improved reasoning, ocr, and world knowledge, 2024{\natexlab{a}}.

\bibitem[Yuan et~al.(2025)Yuan, Li, Zhang, Huang, Xu, Ji, Tong, Qi, Feng, and Yang]{sa2va}
Haobo Yuan, Xiangtai Li, Tao Zhang, Zilong Huang, Shilin Xu, Shunping Ji, Yunhai Tong, Lu~Qi, Jiashi Feng, and Ming-Hsuan Yang.
\newblock Sa2va: Marrying sam2 with llava for dense grounded understanding of images and videos.
\newblock \emph{arXiv}, 2025.

\bibitem[Hong et~al.(2023)Hong, Zhen, Chen, Zheng, Du, Chen, and Gan]{hong20233d}
Yining Hong, Haoyu Zhen, Peihao Chen, Shuhong Zheng, Yilun Du, Zhenfang Chen, and Chuang Gan.
\newblock 3d-llm: Injecting the 3d world into large language models.
\newblock \emph{Advances in Neural Information Processing Systems}, 36:\penalty0 20482--20494, 2023.

\bibitem[Xu et~al.(2025)Xu, Wang, Wang, Chen, Pang, and Lin]{xu2025pointllm}
Runsen Xu, Xiaolong Wang, Tai Wang, Yilun Chen, Jiangmiao Pang, and Dahua Lin.
\newblock Pointllm: Empowering large language models to understand point clouds.
\newblock In \emph{European Conference on Computer Vision}, pages 131--147. Springer, 2025.

\bibitem[Zhu et~al.(2023{\natexlab{b}})Zhu, Ma, Chen, Deng, Huang, and Li]{zhu20233d}
Ziyu Zhu, Xiaojian Ma, Yixin Chen, Zhidong Deng, Siyuan Huang, and Qing Li.
\newblock 3d-vista: Pre-trained transformer for 3d vision and text alignment.
\newblock In \emph{Proceedings of the IEEE/CVF International Conference on Computer Vision}, pages 2911--2921, 2023{\natexlab{b}}.

\bibitem[Zhou et~al.(2024{\natexlab{a}})Zhou, Wan, and Wang]{zhou2024avatargpt}
Zixiang Zhou, Yu~Wan, and Baoyuan Wang.
\newblock Avatargpt: All-in-one framework for motion understanding planning generation and beyond.
\newblock In \emph{Proceedings of the IEEE/CVF Conference on Computer Vision and Pattern Recognition}, pages 1357--1366, 2024{\natexlab{a}}.

\bibitem[Jiang et~al.(2024{\natexlab{b}})Jiang, Chen, Liu, Yu, Yu, and Chen]{jiang2024motiongpt}
Biao Jiang, Xin Chen, Wen Liu, Jingyi Yu, Gang Yu, and Tao Chen.
\newblock Motiongpt: Human motion as a foreign language.
\newblock \emph{Advances in Neural Information Processing Systems}, 36, 2024{\natexlab{b}}.

\bibitem[Zhang et~al.(2023{\natexlab{e}})Zhang, Huang, Liu, Tang, Lu, Chen, Bai, Chu, Yu, and Ouyang]{zhang2023motiongpt}
Yaqi Zhang, Di~Huang, Bin Liu, Shixiang Tang, Yan Lu, Lu~Chen, Lei Bai, Qi~Chu, Nenghai Yu, and Wanli Ouyang.
\newblock Motiongpt: Finetuned llms are general-purpose motion generators.
\newblock \emph{arXiv preprint arXiv:2306.10900}, 2023{\natexlab{e}}.

\bibitem[Ghosh et~al.(2024)Ghosh, Kumar, Seth, Evuru, Tyagi, Sakshi, Nieto, Duraiswami, and Manocha]{ghosh-etal-2024-gama}
Sreyan Ghosh, Sonal Kumar, Ashish Seth, Chandra Kiran~Reddy Evuru, Utkarsh Tyagi, S~Sakshi, Oriol Nieto, Ramani Duraiswami, and Dinesh Manocha.
\newblock {GAMA}: A large audio-language model with advanced audio understanding and complex reasoning abilities.
\newblock In \emph{Proceedings of the 2024 Conference on Empirical Methods in Natural Language Processing}, pages 6288--6313, 2024.

\bibitem[Deshmukh et~al.(2023)Deshmukh, Elizalde, Singh, and Wang]{deshmukh2023pengi}
Soham Deshmukh, Benjamin Elizalde, Rita Singh, and Huaming Wang.
\newblock Pengi: An audio language model for audio tasks.
\newblock \emph{Advances in Neural Information Processing Systems}, 36:\penalty0 18090--18108, 2023.

\bibitem[Hu et~al.(2024{\natexlab{b}})Hu, Zhou, Liu, Chen, Meng, Hao, Pan, Liu, Li, Sivasankaran, et~al.]{hu2024wavllm}
Shujie Hu, Long Zhou, Shujie Liu, Sanyuan Chen, Lingwei Meng, Hongkun Hao, Jing Pan, Xunying Liu, Jinyu Li, Sunit Sivasankaran, et~al.
\newblock Wavllm: Towards robust and adaptive speech large language model.
\newblock \emph{arXiv preprint arXiv:2404.00656}, 2024{\natexlab{b}}.

\bibitem[Tang et~al.(2023)Tang, Yu, Sun, Chen, Tan, Li, Lu, Ma, and Zhang]{tang2023salmonn}
Changli Tang, Wenyi Yu, Guangzhi Sun, Xianzhao Chen, Tian Tan, Wei Li, Lu~Lu, Zejun Ma, and Chao Zhang.
\newblock Salmonn: Towards generic hearing abilities for large language models.
\newblock \emph{arXiv preprint arXiv:2310.13289}, 2023.

\bibitem[Huang et~al.(2023{\natexlab{a}})Huang, Li, Yang, Shi, Chang, Ye, Wu, Hong, Huang, Liu, Ren, Zhao, and Watanabe]{abs-2304-12995}
Rongjie Huang, Mingze Li, Dongchao Yang, Jiatong Shi, Xuankai Chang, Zhenhui Ye, Yuning Wu, Zhiqing Hong, Jiawei Huang, Jinglin Liu, Yi~Ren, Zhou Zhao, and Shinji Watanabe.
\newblock Audiogpt: Understanding and generating speech, music, sound, and talking head.
\newblock \emph{CoRR}, abs/2304.12995, 2023{\natexlab{a}}.

\bibitem[Su et~al.(2023)Su, Lan, Li, Xu, Wang, and Cai]{abs-2305-16355}
Yixuan Su, Tian Lan, Huayang Li, Jialu Xu, Yan Wang, and Deng Cai.
\newblock Pandagpt: One model to instruction-follow them all.
\newblock \emph{CoRR}, abs/2305.16355, 2023.

\bibitem[Han et~al.(2023)Han, Zhang, Shao, Gao, Xu, Xiao, Zhang, Liu, Wen, Guo, et~al.]{han2023imagebind}
Jiaming Han, Renrui Zhang, Wenqi Shao, Peng Gao, Peng Xu, Han Xiao, Kaipeng Zhang, Chris Liu, Song Wen, Ziyu Guo, et~al.
\newblock Imagebind-llm: Multi-modality instruction tuning.
\newblock \emph{arXiv preprint arXiv:2309.03905}, 2023.

\bibitem[Wang et~al.(2024{\natexlab{c}})Wang, Zhuang, and Wu]{wang2024modaverse}
Xinyu Wang, Bohan Zhuang, and Qi~Wu.
\newblock Modaverse: Efficiently transforming modalities with llms.
\newblock In \emph{Proceedings of the IEEE/CVF Conference on Computer Vision and Pattern Recognition}, pages 26606--26616, 2024{\natexlab{c}}.

\bibitem[Hong et~al.(2022)Hong, Ding, Zheng, Liu, and Tang]{abs-2205-15868}
Wenyi Hong, Ming Ding, Wendi Zheng, Xinghan Liu, and Jie Tang.
\newblock Cogvideo: Large-scale pretraining for text-to-video generation via transformers.
\newblock \emph{CoRR}, abs/2205.15868, 2022.

\bibitem[Li et~al.(2023{\natexlab{c}})Li, Wang, Wang, Ge, Ge, and Shan]{SEED-Bench-abs-2307-16125}
Bohao Li, Rui Wang, Guangzhi Wang, Yuying Ge, Yixiao Ge, and Ying Shan.
\newblock Seed-bench: Benchmarking multimodal llms with generative comprehension.
\newblock \emph{CoRR}, abs/2307.16125, 2023{\natexlab{c}}.

\bibitem[Liu et~al.(2024{\natexlab{b}})Liu, Duan, Zhang, Li, Zhang, Zhao, Yuan, Wang, He, Liu, Chen, and Lin]{MMBench-LiuDZLZZYWHLCL24}
Yuan Liu, Haodong Duan, Yuanhan Zhang, Bo~Li, Songyang Zhang, Wangbo Zhao, Yike Yuan, Jiaqi Wang, Conghui He, Ziwei Liu, Kai Chen, and Dahua Lin.
\newblock Mmbench: Is your multi-modal model an all-around player?
\newblock In \emph{Computer Vision - {ECCV} 2024 - 18th European Conference, Milan, Italy, September 29-October 4, 2024, Proceedings, Part {VI}}, pages 216--233, 2024{\natexlab{b}}.

\bibitem[Zheng et~al.(2023{\natexlab{a}})Zheng, Chiang, Sheng, Zhuang, Wu, Zhuang, Lin, Li, Li, Xing, et~al.]{zheng2023judging}
Lianmin Zheng, Wei-Lin Chiang, Ying Sheng, Siyuan Zhuang, Zhanghao Wu, Yonghao Zhuang, Zi~Lin, Zhuohan Li, Dacheng Li, Eric Xing, et~al.
\newblock Judging llm-as-a-judge with mt-bench and chatbot arena.
\newblock \emph{Advances in Neural Information Processing Systems}, 36:\penalty0 46595--46623, 2023{\natexlab{a}}.

\bibitem[Pan et~al.(2025)Pan, Shukla, Singh, Zhao, Mishra, Wang, Xu, Chen, Li, Juefei-Xu, et~al.]{pan2025transfer}
Xichen Pan, Satya~Narayan Shukla, Aashu Singh, Zhuokai Zhao, Shlok~Kumar Mishra, Jialiang Wang, Zhiyang Xu, Jiuhai Chen, Kunpeng Li, Felix Juefei-Xu, et~al.
\newblock Transfer between modalities with metaqueries.
\newblock \emph{arXiv preprint arXiv:2504.06256}, 2025.

\bibitem[Lu et~al.(2022{\natexlab{a}})Lu, Mishra, Xia, Qiu, Chang, Zhu, Tafjord, Clark, and Kalyan]{lu2022learn}
Pan Lu, Swaroop Mishra, Tanglin Xia, Liang Qiu, Kai-Wei Chang, Song-Chun Zhu, Oyvind Tafjord, Peter Clark, and Ashwin Kalyan.
\newblock Learn to explain: Multimodal reasoning via thought chains for science question answering.
\newblock \emph{Advances in Neural Information Processing Systems}, 35:\penalty0 2507--2521, 2022{\natexlab{a}}.

\bibitem[Ma et~al.(2024)Ma, Zhan, Wong, Li, Sun, Chan, and Chao]{ma2024visaidmath}
Jingkun Ma, Runzhe Zhan, Derek~F Wong, Yang Li, Di~Sun, Hou~Pong Chan, and Lidia~S Chao.
\newblock Visaidmath: Benchmarking visual-aided mathematical reasoning.
\newblock \emph{arXiv preprint arXiv:2410.22995}, 2024.

\bibitem[Li et~al.(2024{\natexlab{f}})Li, Lin, Peng, Nyandwi, Jiang, Ma, Khanuja, Krishna, Neubig, and Ramanan]{li2024naturalbench}
Baiqi Li, Zhiqiu Lin, Wenxuan Peng, Jean de~Dieu Nyandwi, Daniel Jiang, Zixian Ma, Simran Khanuja, Ranjay Krishna, Graham Neubig, and Deva Ramanan.
\newblock Naturalbench: Evaluating vision-language models on natural adversarial samples.
\newblock \emph{arXiv preprint arXiv:2410.14669}, 2024{\natexlab{f}}.

\bibitem[Wang et~al.(2024{\natexlab{d}})Wang, Zou, Lin, Sun, Liu, Zhang, Liu, Aw, and Chen]{AudioBench-abs-2406-16020}
Bin Wang, Xunlong Zou, Geyu Lin, Shuo Sun, Zhuohan Liu, Wenyu Zhang, Zhengyuan Liu, AiTi Aw, and Nancy~F. Chen.
\newblock Audiobench: {A} universal benchmark for audio large language models.
\newblock \emph{CoRR}, abs/2406.16020, 2024{\natexlab{d}}.

\bibitem[Yuan et~al.(2023{\natexlab{b}})Yuan, Ma, Li, Zhang, Chen, Yin, Zhuo, Liu, Huang, Tian, Deng, Wang, Lin, Benetos, Ragni, Gyenge, Dannenberg, Chen, Xia, Xue, Liu, Wang, Liu, Guo, and Fu]{MARBLE-YuanMLZCYZLHTDW23}
Ruibin Yuan, Yinghao Ma, Yizhi Li, Ge~Zhang, Xingran Chen, Hanzhi Yin, Le~Zhuo, Yiqi Liu, Jiawen Huang, Zeyue Tian, Binyue Deng, Ningzhi Wang, Chenghua Lin, Emmanouil Benetos, Anton Ragni, Norbert Gyenge, Roger~B. Dannenberg, Wenhu Chen, Gus Xia, Wei Xue, Si~Liu, Shi Wang, Ruibo Liu, Yike Guo, and Jie Fu.
\newblock {MARBLE:} music audio representation benchmark for universal evaluation.
\newblock In \emph{Advances in Neural Information Processing Systems 36: Annual Conference on Neural Information Processing Systems 2023, NeurIPS 2023, New Orleans, LA, USA, December 10 - 16, 2023}, 2023{\natexlab{b}}.

\bibitem[Sakshi et~al.(2024)Sakshi, Tyagi, Kumar, Seth, Selvakumar, Nieto, Duraiswami, Ghosh, and Manocha]{MMAU-abs-2410-19168}
S.~Sakshi, Utkarsh Tyagi, Sonal Kumar, Ashish Seth, Ramaneswaran Selvakumar, Oriol Nieto, Ramani Duraiswami, Sreyan Ghosh, and Dinesh Manocha.
\newblock {MMAU:} {A} massive multi-task audio understanding and reasoning benchmark.
\newblock \emph{CoRR}, abs/2410.19168, 2024.

\bibitem[He et~al.(2023)He, Bai, Lin, Zhao, Hu, Sheng, Yi, Li, and Liu]{T3Bench-abs-2310-02977}
Yuze He, Yushi Bai, Matthieu Lin, Wang Zhao, Yubin Hu, Jenny Sheng, Ran Yi, Juanzi Li, and Yong{-}Jin Liu.
\newblock T\({}^{\mbox{3}}\)bench: Benchmarking current progress in text-to-3d generation.
\newblock \emph{CoRR}, abs/2310.02977, 2023.

\bibitem[Zhang et~al.(2024{\natexlab{b}})Zhang, Hu, Huang, Gong, and Zeng]{3DBench-0002HHG024}
Junjie Zhang, Tianci Hu, Xiaoshui Huang, Yongshun Gong, and Dan Zeng.
\newblock 3dbench: {A} scalable 3d benchmark and instruction-tuning dataset.
\newblock In \emph{Proceedings of the Thirty-Third International Joint Conference on Artificial Intelligence, {IJCAI} 2024, Jeju, South Korea, August 3-9, 2024}, pages 1706--1714, 2024{\natexlab{b}}.

\bibitem[Ning et~al.(2023)Ning, Zhu, Xie, Lin, Cui, Yuan, Chen, and Yuan]{Video-Bench-abs-2311-16103}
Munan Ning, Bin Zhu, Yujia Xie, Bin Lin, Jiaxi Cui, Lu~Yuan, Dongdong Chen, and Li~Yuan.
\newblock Video-bench: {A} comprehensive benchmark and toolkit for evaluating video-based large language models.
\newblock \emph{CoRR}, abs/2311.16103, 2023.

\bibitem[Fu et~al.(2024{\natexlab{b}})Fu, Dai, Luo, Li, Ren, Zhang, Wang, Zhou, Shen, Zhang, Chen, Li, Lin, Zhao, Li, Xu, Zheng, Chen, Ji, and Sun]{Video-MME}
Chaoyou Fu, Yuhan Dai, Yondong Luo, Lei Li, Shuhuai Ren, Renrui Zhang, Zihan Wang, Chenyu Zhou, Yunhang Shen, Mengdan Zhang, Peixian Chen, Yanwei Li, Shaohui Lin, Sirui Zhao, Ke~Li, Tong Xu, Xiawu Zheng, Enhong Chen, Rongrong Ji, and Xing Sun.
\newblock Video-mme: The first-ever comprehensive evaluation benchmark of multi-modal llms in video analysis.
\newblock \emph{CoRR}, abs/2405.21075, 2024{\natexlab{b}}.

\bibitem[Zhou et~al.(2024{\natexlab{b}})Zhou, Shu, Zhao, Wu, Xiao, Yang, Xiong, Zhang, Huang, and Liu]{MLVU-video}
Junjie Zhou, Yan Shu, Bo~Zhao, Boya Wu, Shitao Xiao, Xi~Yang, Yongping Xiong, Bo~Zhang, Tiejun Huang, and Zheng Liu.
\newblock {MLVU:} {A} comprehensive benchmark for multi-task long video understanding.
\newblock \emph{CoRR}, abs/2406.04264, 2024{\natexlab{b}}.

\bibitem[Li et~al.(2024{\natexlab{g}})Li, Wang, He, Li, Wang, Liu, Wang, Xu, Chen, Lou, Wang, and Qiao]{MVBench-video}
Kunchang Li, Yali Wang, Yinan He, Yizhuo Li, Yi~Wang, Yi~Liu, Zun Wang, Jilan Xu, Guo Chen, Ping Lou, Limin Wang, and Yu~Qiao.
\newblock Mvbench: {A} comprehensive multi-modal video understanding benchmark.
\newblock In \emph{{IEEE/CVF} Conference on Computer Vision and Pattern Recognition, {CVPR} 2024, Seattle, WA, USA, June 16-22, 2024}, pages 22195--22206, 2024{\natexlab{g}}.

\bibitem[Chen et~al.(2024{\natexlab{d}})Chen, Lin, Zhang, and Huang]{AutoEval-Video-ChenLZH24}
Xiuyuan Chen, Yuan Lin, Yuchen Zhang, and Weiran Huang.
\newblock Autoeval-video: An automatic benchmark for assessing large vision language models in open-ended video question answering.
\newblock In \emph{Computer Vision - {ECCV} 2024 - 18th European Conference, Milan, Italy, September 29-October 4, 2024, Proceedings, Part {LXVIII}}, pages 179--195, 2024{\natexlab{d}}.

\bibitem[Huang et~al.(2024)Huang, He, Yu, Zhang, Si, Jiang, Zhang, Wu, Jin, Chanpaisit, Wang, Chen, Wang, Lin, Qiao, and Liu]{VBench-video}
Ziqi Huang, Yinan He, Jiashuo Yu, Fan Zhang, Chenyang Si, Yuming Jiang, Yuanhan Zhang, Tianxing Wu, Qingyang Jin, Nattapol Chanpaisit, Yaohui Wang, Xinyuan Chen, Limin Wang, Dahua Lin, Yu~Qiao, and Ziwei Liu.
\newblock Vbench: Comprehensive benchmark suite for video generative models.
\newblock In \emph{{IEEE/CVF} Conference on Computer Vision and Pattern Recognition, {CVPR} 2024, Seattle, WA, USA, June 16-22, 2024}, pages 21807--21818, 2024.

\bibitem[Fu et~al.(2023)Fu, Chen, Shen, Qin, Zhang, Lin, Qiu, Lin, Yang, Zheng, Li, Sun, and Ji]{MME-abs-2306-13394}
Chaoyou Fu, Peixian Chen, Yunhang Shen, Yulei Qin, Mengdan Zhang, Xu~Lin, Zhenyu Qiu, Wei Lin, Jinrui Yang, Xiawu Zheng, Ke~Li, Xing Sun, and Rongrong Ji.
\newblock {MME:} {A} comprehensive evaluation benchmark for multimodal large language models.
\newblock \emph{CoRR}, abs/2306.13394, 2023.

\bibitem[Xu et~al.(2023{\natexlab{b}})Xu, Shao, Zhang, Gao, Liu, Lei, Meng, Huang, Qiao, and Luo]{LVLM-eHub-abs-2306-09265}
Peng Xu, Wenqi Shao, Kaipeng Zhang, Peng Gao, Shuo Liu, Meng Lei, Fanqing Meng, Siyuan Huang, Yu~Qiao, and Ping Luo.
\newblock Lvlm-ehub: {A} comprehensive evaluation benchmark for large vision-language models.
\newblock \emph{CoRR}, abs/2306.09265, 2023{\natexlab{b}}.

\bibitem[Yu et~al.(2024)Yu, Yang, Li, Wang, Lin, Liu, Wang, and Wang]{MM-Vet-YuYLWL0WW24}
Weihao Yu, Zhengyuan Yang, Linjie Li, Jianfeng Wang, Kevin Lin, Zicheng Liu, Xinchao Wang, and Lijuan Wang.
\newblock Mm-vet: Evaluating large multimodal models for integrated capabilities.
\newblock In \emph{Forty-first International Conference on Machine Learning, {ICML} 2024, Vienna, Austria, July 21-27, 2024}, 2024.

\bibitem[Wu and Xie(2024)]{V*-bench-WuX24a}
Penghao Wu and Saining Xie.
\newblock V*: Guided visual search as a core mechanism in multimodal llms.
\newblock In \emph{{IEEE/CVF} Conference on Computer Vision and Pattern Recognition, {CVPR} 2024, Seattle, WA, USA, June 16-22, 2024}, pages 13084--13094, 2024.

\bibitem[Xia et~al.(2024{\natexlab{b}})Xia, Han, Qiu, Zhou, Wang, Zheng, Chen, Cui, Ding, Li, Wang, and Yao]{MMIE-abs-2410-10139}
Peng Xia, Siwei Han, Shi Qiu, Yiyang Zhou, Zhaoyang Wang, Wenhao Zheng, Zhaorun Chen, Chenhang Cui, Mingyu Ding, Linjie Li, Lijuan Wang, and Huaxiu Yao.
\newblock {MMIE:} massive multimodal interleaved comprehension benchmark for large vision-language models.
\newblock \emph{CoRR}, abs/2410.10139, 2024{\natexlab{b}}.

\bibitem[Qian et~al.(2024)Qian, Ye, Fauconnier, Grasch, Yang, and Gan]{MIA-Bench-abs-2407-01509}
Yusu Qian, Hanrong Ye, Jean{-}Philippe Fauconnier, Peter Grasch, Yinfei Yang, and Zhe Gan.
\newblock Mia-bench: Towards better instruction following evaluation of multimodal llms.
\newblock \emph{CoRR}, abs/2407.01509, 2024.

\bibitem[Zhang et~al.(2024{\natexlab{c}})Zhang, Zhang, Tian, Fu, Zhang, Wu, Li, Wang, Wen, Zhang, Wang, Jin, and Tan]{MME-RealWorld-abs-2408-13257}
Yifan Zhang, Huanyu Zhang, Haochen Tian, Chaoyou Fu, Shuangqing Zhang, Junfei Wu, Feng Li, Kun Wang, Qingsong Wen, Zhang Zhang, Liang Wang, Rong Jin, and Tieniu Tan.
\newblock Mme-realworld: Could your multimodal {LLM} challenge high-resolution real-world scenarios that are difficult for humans?
\newblock \emph{CoRR}, abs/2408.13257, 2024{\natexlab{c}}.

\bibitem[Ge et~al.(2024{\natexlab{a}})Ge, Chen, Chen, Chen, Chen, Chen, Xie, Yan, Zhu, Lin, Dingjie, Wang, Gao, Zhiyi, Li, Wan, and Wang]{MLLM-Bench}
Wentao Ge, Shunian Chen, Guiming~Hardy Chen, Junying Chen, Zhihong Chen, Nuo Chen, Wenya Xie, Shuo Yan, Chenghao Zhu, Ziyue Lin, Song Dingjie, Xidong Wang, Anningzhe Gao, Zhang Zhiyi, Jianquan Li, Xiang Wan, and Benyou Wang.
\newblock Mllm-bench: Evaluating multimodal llms with per-sample criteria, 2024{\natexlab{a}}.

\bibitem[Wu et~al.(2024{\natexlab{b}})Wu, Zhang, Zhang, Chen, Liao, Wang, Li, Sun, Yan, Zhai, and Lin]{Q-Bench-0001Z0CLWLSYZL24}
Haoning Wu, Zicheng Zhang, Erli Zhang, Chaofeng Chen, Liang Liao, Annan Wang, Chunyi Li, Wenxiu Sun, Qiong Yan, Guangtao Zhai, and Weisi Lin.
\newblock Q-bench: {A} benchmark for general-purpose foundation models on low-level vision.
\newblock In \emph{The Twelfth International Conference on Learning Representations, {ICLR} 2024, Vienna, Austria, May 7-11, 2024}, 2024{\natexlab{b}}.

\bibitem[Wang et~al.(2024{\natexlab{e}})Wang, Fu, Huang, Li, Liu, Liu, Ma, Xu, Zhou, Zhang, Yan, Mo, Liu, Lu, Li, Xiao, Chang, Roth, Zhang, Poon, and Chen]{MuirBench-abs-2406-09411}
Fei Wang, Xingyu Fu, James~Y. Huang, Zekun Li, Qin Liu, Xiaogeng Liu, Mingyu~Derek Ma, Nan Xu, Wenxuan Zhou, Kai Zhang, Tianyi~Lorena Yan, Wenjie~Jacky Mo, Hsiang{-}Hui Liu, Pan Lu, Chunyuan Li, Chaowei Xiao, Kai{-}Wei Chang, Dan Roth, Sheng Zhang, Hoifung Poon, and Muhao Chen.
\newblock Muirbench: {A} comprehensive benchmark for robust multi-image understanding.
\newblock \emph{CoRR}, abs/2406.09411, 2024{\natexlab{e}}.

\bibitem[Song et~al.(2024)Song, Chen, Chen, Yu, Wan, and Wang]{MileBench-abs-2404-18532}
Dingjie Song, Shunian Chen, Guiming~Hardy Chen, Fei Yu, Xiang Wan, and Benyou Wang.
\newblock Milebench: Benchmarking mllms in long context.
\newblock \emph{CoRR}, abs/2404.18532, 2024.

\bibitem[Li et~al.(2023{\natexlab{d}})Li, Ge, Ge, Wang, Wang, Zhang, and Shan]{SEED-Bench-2-abs-2311-17092}
Bohao Li, Yuying Ge, Yixiao Ge, Guangzhi Wang, Rui Wang, Ruimao Zhang, and Ying Shan.
\newblock Seed-bench-2: Benchmarking multimodal large language models.
\newblock \emph{CoRR}, abs/2311.17092, 2023{\natexlab{d}}.

\bibitem[Tong et~al.(2024{\natexlab{b}})Tong, Brown, Wu, Woo, Middepogu, Akula, Yang, Yang, Iyer, Pan, Wang, Fergus, LeCun, and Xie]{Cambrian-1-abs-2406-16860}
Shengbang Tong, Ellis Brown, Penghao Wu, Sanghyun Woo, Manoj Middepogu, Sai~Charitha Akula, Jihan Yang, Shusheng Yang, Adithya Iyer, Xichen Pan, Austin Wang, Rob Fergus, Yann LeCun, and Saining Xie.
\newblock Cambrian-1: {A} fully open, vision-centric exploration of multimodal llms.
\newblock \emph{CoRR}, abs/2406.16860, 2024{\natexlab{b}}.

\bibitem[Meng et~al.(2024{\natexlab{b}})Meng, Wang, Li, Lu, Tian, Liao, Zhu, Dai, Qiao, Luo, Zhang, and Shao]{MMIU-abs-2408-02718}
Fanqing Meng, Jin Wang, Chuanhao Li, Quanfeng Lu, Hao Tian, Jiaqi Liao, Xizhou Zhu, Jifeng Dai, Yu~Qiao, Ping Luo, Kaipeng Zhang, and Wenqi Shao.
\newblock {MMIU:} multimodal multi-image understanding for evaluating large vision-language models.
\newblock \emph{CoRR}, abs/2408.02718, 2024{\natexlab{b}}.

\bibitem[Ying et~al.(2024{\natexlab{b}})Ying, Meng, Wang, Li, Lin, Yang, Zhang, Zhang, Lin, Liu, Lei, Lu, Chen, Xu, Zhang, Zhang, Gao, Wang, Qiao, Luo, Zhang, and Shao]{MMT-Bench-YingMWLLYZZLLLL24}
Kaining Ying, Fanqing Meng, Jin Wang, Zhiqian Li, Han Lin, Yue Yang, Hao Zhang, Wenbo Zhang, Yuqi Lin, Shuo Liu, Jiayi Lei, Quanfeng Lu, Runjian Chen, Peng Xu, Renrui Zhang, Haozhe Zhang, Peng Gao, Yali Wang, Yu~Qiao, Ping Luo, Kaipeng Zhang, and Wenqi Shao.
\newblock Mmt-bench: {A} comprehensive multimodal benchmark for evaluating large vision-language models towards multitask {AGI}.
\newblock In \emph{Forty-first International Conference on Machine Learning, {ICML} 2024, Vienna, Austria, July 21-27, 2024}, 2024{\natexlab{b}}.

\bibitem[Chen et~al.(2024{\natexlab{e}})Chen, Liang, Siu, Wang, Wang, Wang, Ni, Zhu, Jiang, Lyu, Jiang, He, Liu, Hu, Yue, and Chen]{MEGA-Bench-abs-2410-10563}
Jiacheng Chen, Tianhao Liang, Sherman Siu, Zhengqing Wang, Kai Wang, Yubo Wang, Yuansheng Ni, Wang Zhu, Ziyan Jiang, Bohan Lyu, Dongfu Jiang, Xuan He, Yuan Liu, Hexiang Hu, Xiang Yue, and Wenhu Chen.
\newblock Mega-bench: Scaling multimodal evaluation to over 500 real-world tasks.
\newblock \emph{CoRR}, abs/2410.10563, 2024{\natexlab{e}}.

\bibitem[fur(2023)]{furtado2023atmpositions}
Anomaly detection for atm booths, 2023.
\newblock URL \url{https://www.kaggle.com/datasets/ashlinfurtado/atm-positions}.

\bibitem[Hum({\natexlab{a}})]{Human-Action-Recognition}
Human action recognition (har) dataset, {\natexlab{a}}.
\newblock URL \url{https://www.kaggle.com/datasets/meetnagadia/human-action-recognition-har-dataset}.

\bibitem[Spo()]{Sports-Image-Classification}
100 sports image classification.
\newblock URL \url{https://www.kaggle.com/datasets/gpiosenka/sports-classification}.

\bibitem[Conti et~al.(2023)Conti, Fini, Mancini, Rota, Wang, and Ricci]{CaSED-conti2023vocabularyfree}
Alessandro Conti, Enrico Fini, Massimiliano Mancini, Paolo Rota, Yiming Wang, and Elisa Ricci.
\newblock Vocabulary-free image classification, 2023.

\bibitem[Ske()]{Sketch2Code}
Sketch2code.
\newblock URL \url{https://www.kaggle.com/datasets/vshantam/sketch2code}.

\bibitem[Kumar()]{sketchcode}
Ashwin Kumar.
\newblock Generating html code from a hand-drawn wireframe.
\newblock URL \url{https://github.com/ashnkumar/sketch-code}.

\bibitem[Siegel(2021)]{DVN/Z3ZYLI_2021}
Joshua Siegel.
\newblock {Oxidized and non-oxidized tire sidewall and tread images}.
\newblock 2021.
\newblock \doi{10.7910/DVN/Z3ZYLI}.
\newblock URL \url{https://doi.org/10.7910/DVN/Z3ZYLI}.

\bibitem[Shi et~al.(2016)Shi, Cui, Qi, Meng, and Chen]{shi2016automatic}
Yong Shi, Limeng Cui, Zhiquan Qi, Fan Meng, and Zhensong Chen.
\newblock Automatic road crack detection using random structured forests.
\newblock \emph{IEEE Transactions on Intelligent Transportation Systems}, 17\penalty0 (12):\penalty0 3434--3445, 2016.

\bibitem[Minderer et~al.(2022{\natexlab{a}})Minderer, Gritsenko, Stone, Neumann, Weissenborn, Dosovitskiy, Mahendran, Arnab, Dehghani, Shen, Wang, Zhai, Kipf, and Houlsby]{OWLVIT-abs-2205-06230}
Matthias Minderer, Alexey~A. Gritsenko, Austin Stone, Maxim Neumann, Dirk Weissenborn, Alexey Dosovitskiy, Aravindh Mahendran, Anurag Arnab, Mostafa Dehghani, Zhuoran Shen, Xiao Wang, Xiaohua Zhai, Thomas Kipf, and Neil Houlsby.
\newblock Simple open-vocabulary object detection with vision transformers.
\newblock \emph{CoRR}, abs/2205.06230, 2022{\natexlab{a}}.

\bibitem[Beans, Strawberry and Tomato diseases()]{Plant-Disease-Detection}
Beans, Strawberry and Tomato diseases.
\newblock URL \url{https://universe.roboflow.com/artificial-intelligence-82oex/detecting-diseases}.

\bibitem[Li* et~al.(2022)Li*, Zhang*, Zhang*, Yang, Li, Zhong, Wang, Yuan, Zhang, Hwang, Chang, and Gao]{GLIP-li2021grounded}
Liunian~Harold Li*, Pengchuan Zhang*, Haotian Zhang*, Jianwei Yang, Chunyuan Li, Yiwu Zhong, Lijuan Wang, Lu~Yuan, Lei Zhang, Jenq-Neng Hwang, Kai-Wei Chang, and Jianfeng Gao.
\newblock Grounded language-image pre-training.
\newblock In \emph{CVPR}, 2022.

\bibitem[Gua()]{Guava-Fruit-Disease}
Guava fruit disease dataset.
\newblock URL \url{https://www.kaggle.com/datasets/asadullahgalib/guava-disease-dataset}.

\bibitem[ECG()]{ECG-Dataset}
National heart foundation 2023 ecg dataset.
\newblock URL \url{https://www.kaggle.com/datasets/drkhaledmohsin/national-heart-foundation-2023-ecg-dataset}.

\bibitem[Bra()]{Brain-Tumor-MRI-Dataset}
Brain tumor mri dataset.
\newblock URL \url{https://www.kaggle.com/datasets/masoudnickparvar/brain-tumor-mri-dataset}.

\bibitem[Pum()]{Pumpkin-Leaf-Diseases-Dataset}
Pumpkin leaf diseases dataset.
\newblock URL \url{https://www.kaggle.com/datasets/tahmidmir/pumpkin-leaf-diseases-dataset-from-bangladesh}.

\bibitem[S. et~al.()S., Kant, P., A., and R.]{Leukemia-Classification}
Mourya S., S.~Kant, Kumar P., Gupta A., and \&~Gupta R.
\newblock All challenge dataset of isbi 2019 (c-nmc 2019) (version 1).
\newblock URL \url{https://www.cancerimagingarchive.net/collection/c-nmc-2019/}.

\bibitem[Tito et~al.(2022)Tito, Karatzas, and Valveny]{MP-DocVQA-abs-2212-05935}
Rub{\`{e}}n Tito, Dimosthenis Karatzas, and Ernest Valveny.
\newblock Hierarchical multimodal transformers for multi-page docvqa.
\newblock \emph{CoRR}, abs/2212.05935, 2022.

\bibitem[Ye et~al.(2023{\natexlab{a}})Ye, Hu, Xu, Ye, Yan, Dan, Zhao, Xu, Li, Tian, Qi, Zhang, and Huang]{DocOwl-abs-2307-02499}
Jiabo Ye, Anwen Hu, Haiyang Xu, Qinghao Ye, Ming Yan, Yuhao Dan, Chenlin Zhao, Guohai Xu, Chenliang Li, Junfeng Tian, Qian Qi, Ji~Zhang, and Fei Huang.
\newblock mplug-docowl: Modularized multimodal large language model for document understanding.
\newblock \emph{CoRR}, abs/2307.02499, 2023{\natexlab{a}}.

\bibitem[Kim et~al.(2022)Kim, Hong, Yim, Nam, Park, Yim, Hwang, Yun, Han, and Park]{Donut-kim2022donut}
Geewook Kim, Teakgyu Hong, Moonbin Yim, JeongYeon Nam, Jinyoung Park, Jinyeong Yim, Wonseok Hwang, Sangdoo Yun, Dongyoon Han, and Seunghyun Park.
\newblock Ocr-free document understanding transformer.
\newblock In \emph{European Conference on Computer Vision (ECCV)}, 2022.

\bibitem[doc()]{docquery}
Docquery: Document query engine powered by large language models.
\newblock URL \url{https://github.com/impira/docquery}.

\bibitem[Emo()]{Emotion-Detection}
Emotion detection.
\newblock URL \url{https://www.kaggle.com/datasets/ananthu017/emotion-detection-fer}.

\bibitem[Serengil and Ozpinar(2020)]{deepface-serengil2020lightface}
Sefik~Ilkin Serengil and Alper Ozpinar.
\newblock Lightface: A hybrid deep face recognition framework.
\newblock In \emph{2020 Innovations in Intelligent Systems and Applications Conference (ASYU)}, pages 23--27. IEEE, 2020.
\newblock \doi{10.1109/ASYU50717.2020.9259802}.
\newblock URL \url{https://ieeexplore.ieee.org/document/9259802}.

\bibitem[Pet()]{Pet-Facial-Expression-Image-Dataset}
Pet's facial expression image dataset.
\newblock URL \url{https://www.kaggle.com/datasets/anshtanwar/pets-facial-expression-dataset}.

\bibitem[Fac({\natexlab{a}})]{Facial-Emotion-Recognition}
Facial emotion recognition image dataset, {\natexlab{a}}.
\newblock URL \url{https://www.kaggle.com/datasets/sujaykapadnis/emotion-recognition-dataset}.

\bibitem[Zhang et~al.(2023{\natexlab{f}})Zhang, Zhang, Zhang, Wang, and Song]{DDAMFN-electronics12173595}
Saining Zhang, Yuhang Zhang, Ye~Zhang, Yufei Wang, and Zhigang Song.
\newblock A dual-direction attention mixed feature network for facial expression recognition.
\newblock \emph{Electronics}, 12, 2023{\natexlab{f}}.

\bibitem[Eme()]{Emergency-Vehicle-Siren-Sounds}
Emergency vehicle siren sounds.
\newblock URL \url{https://www.kaggle.com/datasets/vishnu0399/emergency-vehicle-siren-sounds}.

\bibitem[Jiang et~al.(2024{\natexlab{c}})Jiang, He, Zeng, Wei, Ku, Liu, and Chen]{MANTIS-Jiang2024MANTISIM}
Dongfu Jiang, Xuan He, Huaye Zeng, Cong Wei, Max~W.F. Ku, Qian Liu, and Wenhu Chen.
\newblock Mantis: Interleaved multi-image instruction tuning.
\newblock \emph{Transactions on Machine Learning Research}, 2024, 2024{\natexlab{c}}.
\newblock URL \url{https://openreview.net/forum?id=skLtdUVaJa}.

\bibitem[Liu et~al.(2024{\natexlab{c}})Liu, Fu, Xie, Xie, Sun, Lian, Kang, and Li]{PhD-liu2024phd}
Jiazhen Liu, Yuhan Fu, Ruobing Xie, Runquan Xie, Xingwu Sun, Fengzong Lian, Zhanhui Kang, and Xirong Li.
\newblock Phd: A prompted visual hallucination evaluation dataset, 2024{\natexlab{c}}.

\bibitem[Zhang et~al.(2023{\natexlab{g}})Zhang, Li, Das, Malin, and Kumar]{zhang2023sac}
Jiaxin Zhang, Zhuohang Li, Kamalika Das, Bradley~A Malin, and Sricharan Kumar.
\newblock Sac3: Reliable hallucination detection in black-box language models via semantic-aware cross-check consistency.
\newblock \emph{arXiv preprint arXiv:2311.01740}, 2023{\natexlab{g}}.

\bibitem[Ins()]{Instagram-Images-with-Captions}
Instagram images with captions.
\newblock URL \url{https://www.kaggle.com/datasets/prithvijaunjale/instagram-images-with-captions}.

\bibitem[Nguyen et~al.(2022)Nguyen, Suganuma, and Okatani]{Grit-nguyen2022}
Van-Quang Nguyen, Masanori Suganuma, and Takayuki Okatani.
\newblock Grit: Faster and better image captioning transformer using dual visual features.
\newblock In \emph{Computer Vision--ECCV 2022: 17th European Conference, Tel Aviv, Israel, October 23--27, 2022, Proceedings, Part XXXVI}, pages 167--184, 2022.

\bibitem[Lin et~al.(2014)Lin, Maire, Belongie, Hays, Perona, Ramanan, Doll{\'{a}}r, and Zitnick]{COCO-LinMBHPRDZ14}
Tsung{-}Yi Lin, Michael Maire, Serge~J. Belongie, James Hays, Pietro Perona, Deva Ramanan, Piotr Doll{\'{a}}r, and C.~Lawrence Zitnick.
\newblock Microsoft {COCO:} common objects in context.
\newblock In \emph{Computer Vision - {ECCV} 2014 - 13th European Conference, Zurich, Switzerland, September 6-12, 2014, Proceedings, Part {V}}, pages 740--755, 2014.

\bibitem[Lu et~al.(2018)Lu, Wang, Zheng, and Li]{Satellite-Image-Caption-Generation-LuWZL18}
Xiaoqiang Lu, Binqiang Wang, Xiangtao Zheng, and Xuelong Li.
\newblock Exploring models and data for remote sensing image caption generation.
\newblock \emph{{IEEE} Trans. Geosci. Remote. Sens.}, 56\penalty0 (4):\penalty0 2183--2195, 2018.

\bibitem[Hum({\natexlab{b}})]{Human-Related-Image-Captioning}
Human related image captioning, {\natexlab{b}}.
\newblock URL \url{https://freerangestock.com/}.

\bibitem[Chi({\natexlab{a}})]{Chinese-Image-Captioning}
Chinese image captioning, {\natexlab{a}}.
\newblock URL \url{https://www.kaggle.com/datasets/allanyiinai/chineseimagecaptioning/data}.

\bibitem[Alam et~al.(2024)Alam, Imam, Guizani, and Karray]{SpaceNet-abs-2405-13267}
Mohammed~Talha Alam, Raza Imam, Mohsen Guizani, and Fakhri Karray.
\newblock {FLARE} up your data: Diffusion-based augmentation method in astronomical imaging.
\newblock \emph{CoRR}, abs/2405.13267, 2024.

\bibitem[Liu et~al.(2024{\natexlab{d}})Liu, Li, Li, and Lee]{LLava-15-LiuLLL24}
Haotian Liu, Chunyuan Li, Yuheng Li, and Yong~Jae Lee.
\newblock Improved baselines with visual instruction tuning.
\newblock In \emph{{IEEE/CVF} Conference on Computer Vision and Pattern Recognition, {CVPR} 2024, Seattle, WA, USA, June 16-22, 2024}, pages 26286--26296, 2024{\natexlab{d}}.

\bibitem[Silberman et~al.(2012)Silberman, Hoiem, Kohli, and Fergus]{Nyudv2-SilbermanHKF12}
Nathan Silberman, Derek Hoiem, Pushmeet Kohli, and Rob Fergus.
\newblock Indoor segmentation and support inference from {RGBD} images.
\newblock In \emph{Computer Vision - {ECCV} 2012 - 12th European Conference on Computer Vision, Florence, Italy, October 7-13, 2012, Proceedings, Part {V}}, pages 746--760, 2012.

\bibitem[Bhat et~al.(2021)Bhat, Alhashim, and Wonka]{AdaBins-BhatAW21}
Shariq~Farooq Bhat, Ibraheem Alhashim, and Peter Wonka.
\newblock Adabins: Depth estimation using adaptive bins.
\newblock In \emph{{IEEE} Conference on Computer Vision and Pattern Recognition, {CVPR} 2021, virtual, June 19-25, 2021}, pages 4009--4018, 2021.

\bibitem[Cordts et~al.(2016)Cordts, Omran, Ramos, Rehfeld, Enzweiler, Benenson, Franke, Roth, and Schiele]{Cityscapes-CordtsORREBFRS16}
Marius Cordts, Mohamed Omran, Sebastian Ramos, Timo Rehfeld, Markus Enzweiler, Rodrigo Benenson, Uwe Franke, Stefan Roth, and Bernt Schiele.
\newblock The cityscapes dataset for semantic urban scene understanding.
\newblock In \emph{2016 {IEEE} Conference on Computer Vision and Pattern Recognition, {CVPR} 2016, Las Vegas, NV, USA, June 27-30, 2016}, pages 3213--3223, 2016.

\bibitem[Tang et~al.(2021)Tang, Chen, Li, Li, Zhang, and Hu]{BPR-TangCLLZH21}
Chufeng Tang, Hang Chen, Xiao Li, Jianmin Li, Zhaoxiang Zhang, and Xiaolin Hu.
\newblock Look closer to segment better: Boundary patch refinement for instance segmentation.
\newblock In \emph{{IEEE} Conference on Computer Vision and Pattern Recognition, {CVPR} 2021, virtual, June 19-25, 2021}, pages 13926--13935, 2021.

\bibitem[Han()]{Handwriting-Recognition(OCR)}
Handwriting recognition (ocr).
\newblock URL \url{https://www.kaggle.com/datasets/ssarkar445/handwriting-recognitionocr}.

\bibitem[tes()]{tesseract}
Tesseract open source ocr engine (main repository).
\newblock URL \url{https://github.com/tesseract-ocr/tesseract}.

\bibitem[sta({\natexlab{a}})]{standard-OCR-dataset}
Standard ocr dataset, {\natexlab{a}}.
\newblock URL \url{https://www.kaggle.com/datasets/preatcher/standard-ocr-dataset}.

\bibitem[Car({\natexlab{a}})]{Car-License-Plate-Detection}
Car license plate detection, {\natexlab{a}}.
\newblock URL \url{https://www.kaggle.com/datasets/andrewmvd/car-plate-detection}.

\bibitem[Rad({\natexlab{a}})]{Radical-Level-Oracle-Bone-Character-Dataset}
Radical-level oracle bone character dataset, {\natexlab{a}}.
\newblock URL \url{https://www.kaggle.com/datasets/ycfanglab/radical-level-oracle-bone-character-dataset}.

\bibitem[Acn()]{Acne-Recognition-Dataset}
Acne recognition dataset.
\newblock URL \url{https://www.kaggle.com/datasets/imtkaggleteam/acne-computer-vision}.

\bibitem[LaT()]{LaTeX-OCR}
Latex ocr.
\newblock URL \url{https://github.com/lukas-blecher/LaTeX-OCR}.

\bibitem[Blecher()]{pix2tex}
Lukas Blecher.
\newblock Using a vit to convert images of equations into latex code.
\newblock URL \url{https://github.com/lukas-blecher/LaTeX-OCR}.

\bibitem[Seg()]{Segmented-Bob-Ross-Images}
Segmented bob ross images.
\newblock URL \url{https://www.kaggle.com/datasets/residentmario/segmented-bob-ross-images}.

\bibitem[L\"uddecke and Ecker(2022)]{CLIPSeg-lueddecke22}
Timo L\"uddecke and Alexander Ecker.
\newblock Image segmentation using text and image prompts.
\newblock In \emph{Proceedings of the IEEE/CVF Conference on Computer Vision and Pattern Recognition (CVPR)}, pages 7086--7096, 2022.

\bibitem[Chen et~al.(2022{\natexlab{a}})Chen, Liu, Zhao, Zhou, and Zhang]{Cerberus-ChenL0ZZ22}
Xiaoxue Chen, Tianyu Liu, Hao Zhao, Guyue Zhou, and Ya{-}Qin Zhang.
\newblock Cerberus transformer: Joint semantic, affordance and attribute parsing.
\newblock In \emph{{IEEE/CVF} Conference on Computer Vision and Pattern Recognition, {CVPR} 2022, New Orleans, LA, USA, June 18-24, 2022}, pages 19617--19626, 2022{\natexlab{a}}.

\bibitem[Chen et~al.(2018)Chen, Zhu, Papandreou, Schroff, and Adam]{deeplabv3plus2018}
Liang-Chieh Chen, Yukun Zhu, George Papandreou, Florian Schroff, and Hartwig Adam.
\newblock Encoder-decoder with atrous separable convolution for semantic image segmentation.
\newblock In \emph{ECCV}, 2018.

\bibitem[Yang et~al.(2013)Yang, Luo, and Lin]{yang2014clothing}
Wei Yang, Ping Luo, and Liang Lin.
\newblock Clothing co-parsing by joint image segmentation and labeling.
\newblock In \emph{Computer Vision and Pattern Recognition (CVPR), 2014 IEEE Conference on}, 2013.

\bibitem[Flo()]{Flood-Area-Segmentation}
Flood area segmentation.
\newblock URL \url{https://www.kaggle.com/datasets/faizalkarim/flood-area-segmentation}.

\bibitem[Demir et~al.(2018)Demir, Koperski, Lindenbaum, Pang, Huang, Basu, Hughes, Tuia, and Raskar]{DeepGlobe18}
Ilke Demir, Krzysztof Koperski, David Lindenbaum, Guan Pang, Jing Huang, Saikat Basu, Forest Hughes, Devis Tuia, and Ramesh Raskar.
\newblock Deepglobe 2018: A challenge to parse the earth through satellite images.
\newblock In \emph{The IEEE Conference on Computer Vision and Pattern Recognition (CVPR) Workshops}, 2018.

\bibitem[Yu et~al.(2016)Yu, Poirson, Yang, Berg, and Berg]{Refcocog-YuPYBB16}
Licheng Yu, Patrick Poirson, Shan Yang, Alexander~C. Berg, and Tamara~L. Berg.
\newblock Modeling context in referring expressions.
\newblock In \emph{Computer Vision - {ECCV} 2016 - 14th European Conference, Amsterdam, The Netherlands, October 11-14, 2016, Proceedings, Part {II}}, pages 69--85, 2016.

\bibitem[Liu et~al.(2023{\natexlab{d}})Liu, Ding, Cai, Zhang, Satzoda, Mahadevan, and Manmatha]{PolyFormer-liu2023}
Jiang Liu, Hui Ding, Zhaowei Cai, Yuting Zhang, Ravi~Kumar Satzoda, Vijay Mahadevan, and R~Manmatha.
\newblock Polyformer: Referring image segmentation as sequential polygon generation.
\newblock In \emph{CVPR}, 2023{\natexlab{d}}.

\bibitem[Kazemzadeh et~al.(2014)Kazemzadeh, Ordonez, Matten, and Berg]{Refcoco-KazemzadehOMB14}
Sahar Kazemzadeh, Vicente Ordonez, Mark Matten, and Tamara~L. Berg.
\newblock Referitgame: Referring to objects in photographs of natural scenes.
\newblock In \emph{Proceedings of the 2014 Conference on Empirical Methods in Natural Language Processing, {EMNLP} 2014, October 25-29, 2014, Doha, Qatar, {A} meeting of SIGDAT, a Special Interest Group of the {ACL}}, pages 787--798, 2014.

\bibitem[Gurari et~al.(2018)Gurari, Li, Stangl, Guo, Lin, Grauman, Luo, and Bigham]{VizWiz-Gurari0SGLGLB18}
Danna Gurari, Qing Li, Abigale~J. Stangl, Anhong Guo, Chi Lin, Kristen Grauman, Jiebo Luo, and Jeffrey~P. Bigham.
\newblock Vizwiz grand challenge: Answering visual questions from blind people.
\newblock In \emph{2018 {IEEE} Conference on Computer Vision and Pattern Recognition, {CVPR} 2018, Salt Lake City, UT, USA, June 18-22, 2018}, pages 3608--3617, 2018.

\bibitem[Wang et~al.(2022{\natexlab{a}})Wang, Yang, Hu, Li, Lin, Gan, Liu, Liu, and Wang]{GIT-WangYHLLGLLW22}
Jianfeng Wang, Zhengyuan Yang, Xiaowei Hu, Linjie Li, Kevin Lin, Zhe Gan, Zicheng Liu, Ce~Liu, and Lijuan Wang.
\newblock {GIT:} {A} generative image-to-text transformer for vision and language.
\newblock \emph{Trans. Mach. Learn. Res.}, 2022{\natexlab{a}}.

\bibitem[Marino et~al.(2019)Marino, Rastegari, Farhadi, and Mottaghi]{okvqa}
Kenneth Marino, Mohammad Rastegari, Ali Farhadi, and Roozbeh Mottaghi.
\newblock Ok-vqa: A visual question answering benchmark requiring external knowledge.
\newblock In \emph{Conference on Computer Vision and Pattern Recognition (CVPR)}, 2019.

\bibitem[med()]{medical-visual-question-answering}
Medical visual question answering.
\newblock URL \url{https://www.kaggle.com/datasets/mitanshuchakrawarty/medical-visual-question-answering}.

\bibitem[Zhang et~al.(2023{\natexlab{h}})Zhang, Wu, Zhao, Lin, Zhang, Wang, and Xie]{PMC-VQA-zhang2023}
Xiaoman Zhang, Chaoyi Wu, Ziheng Zhao, Weixiong Lin, Ya~Zhang, Yanfeng Wang, and Weidi Xie.
\newblock Pmc-vqa: Visual instruction tuning for medical visual question answering.
\newblock \emph{arXiv preprint arXiv:2305.10415}, 2023{\natexlab{h}}.

\bibitem[Hudson and Manning(2019)]{GQA-HudsonM19}
Drew~A. Hudson and Christopher~D. Manning.
\newblock {GQA:} {A} new dataset for real-world visual reasoning and compositional question answering.
\newblock In \emph{{IEEE} Conference on Computer Vision and Pattern Recognition, {CVPR} 2019, Long Beach, CA, USA, June 16-20, 2019}, pages 6700--6709, 2019.

\bibitem[Lu et~al.(2022{\natexlab{b}})Lu, Mishra, Xia, Qiu, Chang, Zhu, Tafjord, Clark, and Kalyan]{ScienceQA-LuMX0CZTCK22}
Pan Lu, Swaroop Mishra, Tanglin Xia, Liang Qiu, Kai{-}Wei Chang, Song{-}Chun Zhu, Oyvind Tafjord, Peter Clark, and Ashwin Kalyan.
\newblock Learn to explain: Multimodal reasoning via thought chains for science question answering.
\newblock In \emph{Advances in Neural Information Processing Systems 35: Annual Conference on Neural Information Processing Systems 2022, NeurIPS 2022, New Orleans, LA, USA, November 28 - December 9, 2022}, 2022{\natexlab{b}}.

\bibitem[Zhang et~al.(2024{\natexlab{d}})Zhang, Zhang, Li, Zhao, Karypis, and Smola]{MMCoT-0001Z00KS24}
Zhuosheng Zhang, Aston Zhang, Mu~Li, Hai Zhao, George Karypis, and Alex Smola.
\newblock Multimodal chain-of-thought reasoning in language models.
\newblock \emph{Trans. Mach. Learn. Res.}, 2024{\natexlab{d}}.

\bibitem[Blo()]{Blood-Cell-Images}
Blood cell images.
\newblock URL \url{https://www.kaggle.com/datasets/paultimothymooney/blood-cells}.

\bibitem[cat()]{cataract-dataset}
Cataract dataset.
\newblock URL \url{https://www.kaggle.com/datasets/jr2ngb/cataractdataset}.

\bibitem[Lu et~al.(2021{\natexlab{a}})Lu, Gong, Jiang, Qiu, Huang, Liang, and Zhu]{Inter-GPS-LuGJQHLZ20}
Pan Lu, Ran Gong, Shibiao Jiang, Liang Qiu, Siyuan Huang, Xiaodan Liang, and Song{-}Chun Zhu.
\newblock Inter-gps: Interpretable geometry problem solving with formal language and symbolic reasoning.
\newblock In \emph{Proceedings of the 59th Annual Meeting of the Association for Computational Linguistics and the 11th International Joint Conference on Natural Language Processing, {ACL/IJCNLP} 2021, (Volume 1: Long Papers), Virtual Event, August 1-6, 2021}, pages 6774--6786, 2021{\natexlab{a}}.

\bibitem[Lun({\natexlab{a}})]{Lung-cance-CAM-attempting}
Lung cancer cam attempting, {\natexlab{a}}.
\newblock URL \url{https://www.kaggle.com/code/zaibunnisaa/lung-cancer-cam-attempting-92-accuracy-ver5/input}.

\bibitem[Lu et~al.(2024{\natexlab{c}})Lu, Bansal, Xia, Liu, Li, Hajishirzi, Cheng, Chang, Galley, and Gao]{MathVista-lu2024}
Pan Lu, Hritik Bansal, Tony Xia, Jiacheng Liu, Chunyuan Li, Hannaneh Hajishirzi, Hao Cheng, Kai-Wei Chang, Michel Galley, and Jianfeng Gao.
\newblock Mathvista: Evaluating mathematical reasoning of foundation models in visual contexts.
\newblock In \emph{International Conference on Learning Representations (ICLR)}, 2024{\natexlab{c}}.

\bibitem[Peng et~al.(2024)Peng, Li, Zhou, Xia, Zhang, Bai, Mao, Wang, He, Zhou, Shi, Chen, Zhang, and Yue]{Chimera-peng2024}
Tianshuo Peng, Mingsheng Li, Hongbin Zhou, Renqiu Xia, Renrui Zhang, Lei Bai, Song Mao, Bin Wang, Conghui He, Aojun Zhou, Botian Shi, Tao Chen, Bo~Zhang, and Xiangyu Yue.
\newblock Chimera: Improving generalist model with domain-specific experts, 2024.

\bibitem[Wang et~al.(2020{\natexlab{a}})Wang, Lin, and Wong]{COVID-Net-Wang2020}
Linda Wang, Zhong~Qiu Lin, and Alexander Wong.
\newblock Covid-net: a tailored deep convolutional neural network design for detection of covid-19 cases from chest x-ray images.
\newblock \emph{Scientific Reports}, 10\penalty0 (1):\penalty0 19549, Nov 2020{\natexlab{a}}.

\bibitem[Wang et~al.(2024{\natexlab{f}})Wang, Zheng, Chen, Ma, and Zhong]{EarthVQA-WangZCMZ24}
Junjue Wang, Zhuo Zheng, Zihang Chen, Ailong Ma, and Yanfei Zhong.
\newblock Earthvqa: Towards queryable earth via relational reasoning-based remote sensing visual question answering.
\newblock In \emph{Thirty-Eighth {AAAI} Conference on Artificial Intelligence, {AAAI} 2024, Thirty-Sixth Conference on Innovative Applications of Artificial Intelligence, {IAAI} 2024, Fourteenth Symposium on Educational Advances in Artificial Intelligence, {EAAI} 2014, February 20-27, 2024, Vancouver, Canada}, pages 5481--5489, 2024{\natexlab{f}}.

\bibitem[Lobry et~al.(2020)Lobry, Marcos, Murray, and Tuia]{RSVQA-LobryMMT20}
Sylvain Lobry, Diego Marcos, Jesse Murray, and Devis Tuia.
\newblock {RSVQA:} visual question answering for remote sensing data.
\newblock \emph{{IEEE} Trans. Geosci. Remote. Sens.}, 58\penalty0 (12):\penalty0 8555--8566, 2020.

\bibitem[Ski()]{Skin-Cancer-Dateset}
Skin cancer dateset.
\newblock URL \url{https://www.kaggle.com/datasets/fanconic/skin-cancer-malignant-vs-benign}.

\bibitem[Chang et~al.(2022)Chang, Cao, Narang, Gao, Suzuki, and Bisk]{WebQA-ChangCNGSB22}
Yingshan Chang, Guihong Cao, Mridu Narang, Jianfeng Gao, Hisami Suzuki, and Yonatan Bisk.
\newblock Webqa: Multihop and multimodal {QA}.
\newblock In \emph{{IEEE/CVF} Conference on Computer Vision and Pattern Recognition, {CVPR} 2022, New Orleans, LA, USA, June 18-24, 2022}, pages 16474--16483, 2022.

\bibitem[Wei et~al.(2024)Wei, Su, Xu, Zeng, Liu, and Wang]{MLoEM-WeiSXZLW24}
Jingyu Wei, Yi~Su, Kele Xu, Lingbin Zeng, Bo~Liu, and Huaimin Wang.
\newblock Demonstrative instruction following in multimodal llms via integrating low-rank adaptation with ensemble learning.
\newblock In \emph{Proceedings of the 32nd {ACM} International Conference on Multimedia, {MM} 2024, Melbourne, VIC, Australia, 28 October 2024 - 1 November 2024}, pages 11435--11441, 2024.

\bibitem[Li et~al.(2022{\natexlab{a}})Li, Li, and Nie]{MMCoQA-LiLN22}
Yongqi Li, Wenjie Li, and Liqiang Nie.
\newblock Mmcoqa: Conversational question answering over text, tables, and images.
\newblock In \emph{Proceedings of the 60th Annual Meeting of the Association for Computational Linguistics (Volume 1: Long Papers), {ACL} 2022, Dublin, Ireland, May 22-27, 2022}, pages 4220--4231, 2022{\natexlab{a}}.

\bibitem[Li et~al.(2024{\natexlab{h}})Li, Zhang, Zhang, Zhang, Li, Li, Ma, and Li]{li2024llava}
Feng Li, Renrui Zhang, Hao Zhang, Yuanhan Zhang, Bo~Li, Wei Li, Zejun Ma, and Chunyuan Li.
\newblock Llava-next-interleave: Tackling multi-image, video, and 3d in large multimodal models.
\newblock \emph{arXiv preprint arXiv:2407.07895}, 2024{\natexlab{h}}.

\bibitem[Li et~al.(2023{\natexlab{e}})Li, Pan, Ge, Gao, Ji, Zhang, Chua, Tang, Zhang, and Zhuang]{VPG-C-li2023fine}
Juncheng Li, Kaihang Pan, Zhiqi Ge, Minghe Gao, Wei Ji, Wenqiao Zhang, Tat-Seng Chua, Siliang Tang, Hanwang Zhang, and Yueting Zhuang.
\newblock Fine-tuning multimodal llms to follow zero-shot demonstrative instructions.
\newblock In \emph{The Twelfth International Conference on Learning Representations}, 2023{\natexlab{e}}.

\bibitem[Tanaka et~al.(2023)Tanaka, Nishida, Nishida, Hasegawa, Saito, and Saito]{SlideVQA-TanakaNNHSS23}
Ryota Tanaka, Kyosuke Nishida, Kosuke Nishida, Taku Hasegawa, Itsumi Saito, and Kuniko Saito.
\newblock Slidevqa: {A} dataset for document visual question answering on multiple images.
\newblock In \emph{Thirty-Seventh {AAAI} Conference on Artificial Intelligence, {AAAI} 2023, Thirty-Fifth Conference on Innovative Applications of Artificial Intelligence, {IAAI} 2023, Thirteenth Symposium on Educational Advances in Artificial Intelligence, {EAAI} 2023, Washington, DC, USA, February 7-14, 2023}, pages 13636--13645, 2023.

\bibitem[Mathew et~al.(2021)Mathew, Karatzas, and Jawahar]{DocVQA-MathewKJ21}
Minesh Mathew, Dimosthenis Karatzas, and C.~V. Jawahar.
\newblock Docvqa: {A} dataset for {VQA} on document images.
\newblock In \emph{{IEEE} Winter Conference on Applications of Computer Vision, {WACV} 2021, Waikoloa, HI, USA, January 3-8, 2021}, pages 2199--2208, 2021.

\bibitem[Kembhavi et~al.(2017)Kembhavi, Seo, Schwenk, Choi, Farhadi, and Hajishirzi]{TQA-KembhaviSSCFH17}
Aniruddha Kembhavi, Min~Joon Seo, Dustin Schwenk, Jonghyun Choi, Ali Farhadi, and Hannaneh Hajishirzi.
\newblock Are you smarter than a sixth grader? textbook question answering for multimodal machine comprehension.
\newblock In \emph{2017 {IEEE} Conference on Computer Vision and Pattern Recognition, {CVPR} 2017, Honolulu, HI, USA, July 21-26, 2017}, pages 5376--5384, 2017.

\bibitem[Mishra et~al.(2019)Mishra, Shekhar, Singh, and Chakraborty]{OCR-VQA-0001SSC19}
Anand Mishra, Shashank Shekhar, Ajeet~Kumar Singh, and Anirban Chakraborty.
\newblock {OCR-VQA:} visual question answering by reading text in images.
\newblock In \emph{2019 International Conference on Document Analysis and Recognition, {ICDAR} 2019, Sydney, Australia, September 20-25, 2019}, pages 947--952, 2019.

\bibitem[Jiaqing()]{Handwritten-Formula-VQA}
Li~Jiaqing.
\newblock Handwritten formula vqa.
\newblock URL \url{https://huggingface.co/datasets/Kitajiang/test2_CROHME2016}.

\bibitem[AI()]{Face-Comparison-VQA}
BitMind AI.
\newblock Face comparison visual question answering.
\newblock URL \url{https://huggingface.co/datasets/bitmind/lfw}.

\bibitem[Hu et~al.(2024{\natexlab{c}})Hu, Xu, Zhang, Ye, Yan, Zhang, Jin, Huang, and Zhou]{mPLUG-DocOwl2-abs-2409-03420}
Anwen Hu, Haiyang Xu, Liang Zhang, Jiabo Ye, Ming Yan, Ji~Zhang, Qin Jin, Fei Huang, and Jingren Zhou.
\newblock mplug-docowl2: High-resolution compressing for ocr-free multi-page document understanding.
\newblock \emph{CoRR}, abs/2409.03420, 2024{\natexlab{c}}.

\bibitem[Rosenfeld et~al.(2018)Rosenfeld, Solbach, and Tsotsos]{ttl-RosenfeldST18}
Amir Rosenfeld, Markus~D. Solbach, and John~K. Tsotsos.
\newblock Totally looks like - how humans compare, compared to machines.
\newblock In \emph{Computer Vision - {ACCV} 2018 - 14th Asian Conference on Computer Vision, Perth, Australia, December 2-6, 2018, Revised Selected Papers, Part {I}}, pages 282--297, 2018.

\bibitem[Garcia et~al.(2020)Garcia, Ye, Liu, Hu, Otani, Chu, Nakashima, and Mitamura]{garcia2020AQUA}
Noa Garcia, Chentao Ye, Zihua Liu, Qingtao Hu, Mayu Otani, Chenhui Chu, Yuta Nakashima, and Teruko Mitamura.
\newblock A dataset and baselines for visual question answering on art.
\newblock In \emph{Proceedings of the European Conference in Computer Vision Workshops}, 2020.

\bibitem[Li et~al.(2022{\natexlab{b}})Li, Xu, Tian, Wang, Yan, Bi, Ye, Chen, Xu, Cao, Zhang, Huang, Huang, Zhou, and Si]{mPLUG-LiXTWYBYCXCZHHZ22}
Chenliang Li, Haiyang Xu, Junfeng Tian, Wei Wang, Ming Yan, Bin Bi, Jiabo Ye, He~Chen, Guohai Xu, Zheng Cao, Ji~Zhang, Songfang Huang, Fei Huang, Jingren Zhou, and Luo Si.
\newblock mplug: Effective and efficient vision-language learning by cross-modal skip-connections.
\newblock In \emph{Proceedings of the 2022 Conference on Empirical Methods in Natural Language Processing, {EMNLP} 2022, Abu Dhabi, United Arab Emirates, December 7-11, 2022}, pages 7241--7259, 2022{\natexlab{b}}.

\bibitem[Zhang et~al.(2024{\natexlab{e}})Zhang, Feng, Bai, Du, Hou, Deng, Han, Li, Wang, Liu, Qu, Zhang, Zhao, Liang, Liu, Fang, Yang, Huang, Lin, Zhang, and Ni]{CII-Bench-abs-2410-13854}
Chenhao Zhang, Xi~Feng, Yuelin Bai, Xinrun Du, Jinchang Hou, Kaixin Deng, Guangzeng Han, Qinrui Li, Bingli Wang, Jiaheng Liu, Xingwei Qu, Yifei Zhang, Qixuan Zhao, Yiming Liang, Ziqiang Liu, Feiteng Fang, Min Yang, Wenhao Huang, Chenghua Lin, Ge~Zhang, and Shiwen Ni.
\newblock Can mllms understand the deep implication behind chinese images?
\newblock \emph{CoRR}, abs/2410.13854, 2024{\natexlab{e}}.

\bibitem[Han et~al.(2024)Han, Mao, Jiang, Pan, and Zhang]{StyleBooth-abs-2404-12154}
Zhen Han, Chaojie Mao, Zeyinzi Jiang, Yulin Pan, and Jingfeng Zhang.
\newblock Stylebooth: Image style editing with multimodal instruction.
\newblock \emph{CoRR}, abs/2404.12154, 2024.

\bibitem[Bergmann et~al.(2021)Bergmann, Batzner, Fauser, Sattlegger, and Steger]{MVTec-BergmannBFSS21}
Paul Bergmann, Kilian Batzner, Michael Fauser, David Sattlegger, and Carsten Steger.
\newblock The mvtec anomaly detection dataset: {A} comprehensive real-world dataset for unsupervised anomaly detection.
\newblock \emph{Int. J. Comput. Vis.}, 129\penalty0 (4):\penalty0 1038--1059, 2021.

\bibitem[Zhang et~al.(2023{\natexlab{i}})Zhang, Li, Li, Huang, Shan, and Chen]{DeSTSeg-ZhangL0HSC23}
Xuan Zhang, Shiyu Li, Xi~Li, Ping Huang, Jiulong Shan, and Ting Chen.
\newblock Destseg: Segmentation guided denoising student-teacher for anomaly detection.
\newblock In \emph{{IEEE/CVF} Conference on Computer Vision and Pattern Recognition, {CVPR} 2023, Vancouver, BC, Canada, June 17-24, 2023}, pages 3914--3923, 2023{\natexlab{i}}.

\bibitem[Horak et~al.(2022)Horak, Bilik, Bostik, Kratochvila, and Zemcik]{Industry-Biscuit}
Karel Horak, Simon Bilik, Ondrej Bostik, Lukas Kratochvila, and Tomas Zemcik.
\newblock Industry biscuit (cookie) dataset, 2022.
\newblock URL \url{https://www.kaggle.com/dsv/4311115}.

\bibitem[Zou et~al.(2022)Zou, Jeong, Pemula, Zhang, and Dabeer]{VisA-ZouJPZD22}
Yang Zou, Jongheon Jeong, Latha Pemula, Dongqing Zhang, and Onkar Dabeer.
\newblock Spot-the-difference self-supervised pre-training for anomaly detection and segmentation.
\newblock In \emph{Computer Vision - {ECCV} 2022 - 17th European Conference, Tel Aviv, Israel, October 23-27, 2022, Proceedings, Part {XXX}}, pages 392--408, 2022.

\bibitem[Mar()]{Marble-Surface-Anomaly-Detection}
Marble surface anomaly detection.
\newblock URL \url{https://www.kaggle.com/datasets/wardaddy24/marble-surface-anomaly-detection}.

\bibitem[COV({\natexlab{a}})]{COVID-19-CT-scan-lesion-segmentation}
Covid-19 ct scan lesion segmentation dataset, {\natexlab{a}}.
\newblock URL \url{https://www.kaggle.com/datasets/maedemaftouni/covid19-ct-scan-lesion-segmentation-dataset?select=masks}.

\bibitem[COV({\natexlab{b}})]{COVID-19-CT-scan-lesion-segmentation-Unet}
Pc5 3 - mia class ortsu, {\natexlab{b}}.
\newblock URL \url{https://www.kaggle.com/code/aryan6043/pc5-3-mia-class-ortsu}.

\bibitem[Hub()]{Hubmap-Organ-Segmentation}
Hubmap - organ segmentation competition.
\newblock URL \url{https://www.kaggle.com/datasets/manojprabhaakr/hubmap-organ-512512}.

\bibitem[Lun({\natexlab{b}})]{Lung-Mask-Image-Dataset}
Lung mask image dataset, {\natexlab{b}}.
\newblock URL \url{https://www.kaggle.com/datasets/newra008/lung-mask-image-dataset}.

\bibitem[Bac()]{Bacteria-detection}
Bacteria detection with darkfield microscopy dataset for spirochaeta segmentation with image and manually annotated masks.
\newblock URL \url{https://tianchi.aliyun.com/dataset/94411?spm=a2c22.28136470.J_3941670930.9.2fe04a0aRphTJa&from=search}.

\bibitem[Buda et~al.(2019{\natexlab{a}})Buda, Saha, and Mazurowski]{BudaSM19}
Mateusz Buda, Ashirbani Saha, and Maciej~A. Mazurowski.
\newblock Association of genomic subtypes of lower-grade gliomas with shape features automatically extracted by a deep learning algorithm.
\newblock \emph{Comput. Biol. Medicine}, 109:\penalty0 218--225, 2019{\natexlab{a}}.

\bibitem[Buda et~al.(2019{\natexlab{b}})Buda, Saha, and Mazurowski]{BrainUNet-buda2019association}
Mateusz Buda, Ashirbani Saha, and Maciej~A Mazurowski.
\newblock Association of genomic subtypes of lower-grade gliomas with shape features automatically extracted by a deep learning algorithm.
\newblock \emph{Computers in Biology and Medicine}, 109, 2019{\natexlab{b}}.

\bibitem[Che()]{Chest-X-ray-Segmentation}
Chest x-ray dataset for tuberculosis segmentation.
\newblock URL \url{https://www.kaggle.com/datasets/iamtapendu/chest-x-ray-lungs-segmentation}.

\bibitem[Zhang et~al.(2019)Zhang, Li, Dong, Rosin, Cai, Han, Yang, Huang, and Hu]{OCHuman-ZhangLDRCHYHH19}
Song{-}Hai Zhang, Ruilong Li, Xin Dong, Paul~L. Rosin, Zixi Cai, Xi~Han, Dingcheng Yang, Haozhi Huang, and Shi{-}Min Hu.
\newblock Pose2seg: Detection free human instance segmentation.
\newblock In \emph{{IEEE} Conference on Computer Vision and Pattern Recognition, {CVPR} 2019, Long Beach, CA, USA, June 16-20, 2019}, pages 889--898, 2019.

\bibitem[Xu et~al.(2022)Xu, Zhang, Zhang, and Tao]{ViTPose-XuZZT22}
Yufei Xu, Jing Zhang, Qiming Zhang, and Dacheng Tao.
\newblock Vitpose: Simple vision transformer baselines for human pose estimation.
\newblock In \emph{Advances in Neural Information Processing Systems 35: Annual Conference on Neural Information Processing Systems 2022, NeurIPS 2022, New Orleans, LA, USA, November 28 - December 9, 2022}, 2022.

\bibitem[Suhr et~al.(2019)Suhr, Zhou, Zhang, Zhang, Bai, and Artzi]{nlvr2-SuhrZZZBA19}
Alane Suhr, Stephanie Zhou, Ally Zhang, Iris Zhang, Huajun Bai, and Yoav Artzi.
\newblock A corpus for reasoning about natural language grounded in photographs.
\newblock In \emph{Proceedings of the 57th Conference of the Association for Computational Linguistics, {ACL} 2019, Florence, Italy, July 28- August 2, 2019, Volume 1: Long Papers}, pages 6418--6428, 2019.

\bibitem[fgv()]{fgvc-aircraft}
Aircraft model matching dataset.
\newblock URL \url{https://huggingface.co/datasets/Multimodal-Fatima/FGVC_Aircraft_test}.

\bibitem[sta({\natexlab{b}})]{stanford-car}
Car model matching dataset, {\natexlab{b}}.
\newblock URL \url{https://huggingface.co/datasets/Multimodal-Fatima/StanfordCars_test}.

\bibitem[cuh()]{cuhk03}
Action consistency verification.
\newblock URL \url{https://drive.google.com/file/d/0B7TOZKXmIjU3OUhfd3BPaVRHZVE/view?resourcekey=0-hU4gyE6hFsBgizIh9DFqtA}.

\bibitem[LeCun()]{mnist}
Yann LeCun.
\newblock Digital consistency comparison.
\newblock URL \url{https://huggingface.co/datasets/ylecun/mnist}.

\bibitem[Shridhar et~al.(2020)Shridhar, Thomason, Gordon, Bisk, Han, Mottaghi, Zettlemoyer, and Fox]{ALFRED20}
Mohit Shridhar, Jesse Thomason, Daniel Gordon, Yonatan Bisk, Winson Han, Roozbeh Mottaghi, Luke Zettlemoyer, and Dieter Fox.
\newblock {ALFRED: A Benchmark for Interpreting Grounded Instructions for Everyday Tasks}.
\newblock In \emph{The IEEE Conference on Computer Vision and Pattern Recognition (CVPR)}, 2020.

\bibitem[Iyyer et~al.(2017)Iyyer, Manjunatha, Guha, Vyas, Boyd-Graber, {Daum\'{e} III}, and Davis]{IyyerComics2016}
Mohit Iyyer, Varun Manjunatha, Anupam Guha, Yogarshi Vyas, Jordan Boyd-Graber, Hal {Daum\'{e} III}, and Larry Davis.
\newblock The amazing mysteries of the gutter: Drawing inferences between panels in comic book narratives.
\newblock In \emph{IEEE Conference on Computer Vision and Pattern Recognition}, 2017.

\bibitem[Han et~al.(2017)Han, Wu, Huang, Zhang, Zhu, Li, Zhao, and Davis]{Fashion200K-HanWHZZLZD17}
Xintong Han, Zuxuan Wu, Phoenix~X. Huang, Xiao Zhang, Menglong Zhu, Yuan Li, Yang Zhao, and Larry~S. Davis.
\newblock Automatic spatially-aware fashion concept discovery.
\newblock In \emph{{IEEE} International Conference on Computer Vision, {ICCV} 2017, Venice, Italy, October 22-29, 2017}, pages 1472--1480, 2017.

\bibitem[Yagcioglu et~al.(2018)Yagcioglu, Erdem, Erdem, and Ikizler{-}Cinbis]{RecipeQA-YagciogluEEI18}
Semih Yagcioglu, Aykut Erdem, Erkut Erdem, and Nazli Ikizler{-}Cinbis.
\newblock Recipeqa: {A} challenge dataset for multimodal comprehension of cooking recipes.
\newblock In \emph{Proceedings of the 2018 Conference on Empirical Methods in Natural Language Processing, Brussels, Belgium, October 31 - November 4, 2018}, pages 1358--1368, 2018.

\bibitem[Shullani et~al.(2017)Shullani, Fontani, Iuliani, Shaya, and Piva]{VISION-ShullaniFISP17}
Dasara Shullani, Marco Fontani, Massimo Iuliani, Omar~Al Shaya, and Alessandro Piva.
\newblock {VISION:} a video and image dataset for source identification.
\newblock \emph{{EURASIP} J. Inf. Secur.}, 2017:\penalty0 15, 2017.

\bibitem[Isola et~al.(2015)Isola, Lim, and Adelson]{MIT-States-IsolaLA15}
Phillip Isola, Joseph~J. Lim, and Edward~H. Adelson.
\newblock Discovering states and transformations in image collections.
\newblock In \emph{{IEEE} Conference on Computer Vision and Pattern Recognition, {CVPR} 2015, Boston, MA, USA, June 7-12, 2015}, pages 1383--1391, 2015.

\bibitem[Caesar et~al.(2020)Caesar, Bankiti, Lang, Vora, Liong, Xu, Krishnan, Pan, Baldan, and Beijbom]{nuScenes-CaesarBLVLXKPBB20}
Holger Caesar, Varun Bankiti, Alex~H. Lang, Sourabh Vora, Venice~Erin Liong, Qiang Xu, Anush Krishnan, Yu~Pan, Giancarlo Baldan, and Oscar Beijbom.
\newblock nuscenes: {A} multimodal dataset for autonomous driving.
\newblock In \emph{2020 {IEEE/CVF} Conference on Computer Vision and Pattern Recognition, {CVPR} 2020, Seattle, WA, USA, June 13-19, 2020}, pages 11618--11628, 2020.

\bibitem[Loy et~al.(2013{\natexlab{a}})Loy, Gong, and Xiang]{Crowd-Counting-LoyGX13}
Chen~Change Loy, Shaogang Gong, and Tao Xiang.
\newblock From semi-supervised to transfer counting of crowds.
\newblock In \emph{{IEEE} International Conference on Computer Vision, {ICCV} 2013, Sydney, Australia, December 1-8, 2013}, pages 2256--2263, 2013{\natexlab{a}}.

\bibitem[Jiang et~al.(2023)Jiang, Liu, and Chen]{CLIP-Count-JiangLC23}
Ruixiang Jiang, Lingbo Liu, and Changwen Chen.
\newblock Clip-count: Towards text-guided zero-shot object counting.
\newblock In \emph{Proceedings of the 31st {ACM} International Conference on Multimedia, {MM} 2023, Ottawa, ON, Canada, 29 October 2023- 3 November 2023}, pages 4535--4545, 2023.

\bibitem[Cou({\natexlab{a}})]{Count-number-Faces}
Count the number of faces present in an image, {\natexlab{a}}.
\newblock URL \url{https://www.kaggle.com/datasets/vin1234/count-the-number-of-faces-present-in-an-image}.

\bibitem[FIT()]{FITS-Images}
Fits images for object counting and detection.
\newblock URL \url{https://www.kaggle.com/datasets/santurini/fits-images-for-object-counting-and-detection}.

\bibitem[See()]{Seeds-Counting}
Seeds counting.
\newblock URL \url{https://www.kaggle.com/datasets/raj123verma/seeds-counting}.

\bibitem[Hsieh et~al.(2017)Hsieh, Lin, and Hsu]{CARPK-HsiehLH17}
Meng{-}Ru Hsieh, Yen{-}Liang Lin, and Winston~H. Hsu.
\newblock Drone-based object counting by spatially regularized regional proposal network.
\newblock In \emph{{IEEE} International Conference on Computer Vision, {ICCV} 2017, Venice, Italy, October 22-29, 2017}, pages 4165--4173, 2017.

\bibitem[Loy et~al.(2013{\natexlab{b}})Loy, Gong, and Xiang]{Mall-Dataset-LoyGX13}
Chen~Change Loy, Shaogang Gong, and Tao Xiang.
\newblock From semi-supervised to transfer counting of crowds.
\newblock In \emph{{IEEE} International Conference on Computer Vision, {ICCV} 2013, Sydney, Australia, December 1-8, 2013}, pages 2256--2263, 2013{\natexlab{b}}.

\bibitem[Cou({\natexlab{b}})]{Count-Paperclips}
Count the paperclips, {\natexlab{b}}.
\newblock URL \url{https://www.kaggle.com/datasets/jeffheaton/count-the-paperclips}.

\bibitem[David et~al.(2020)David, Madec, Sadeghi-Tehran, Aasen, Zheng, Liu, Kirchgessner, Ishikawa, Nagasawa, Badhon, et~al.]{david2020global}
Etienne David, Simon Madec, Pouria Sadeghi-Tehran, Helge Aasen, Bangyou Zheng, Shouyang Liu, Norbert Kirchgessner, Goro Ishikawa, Koichi Nagasawa, Minhajul~A Badhon, et~al.
\newblock Global wheat head detection (gwhd) dataset: a large and diverse dataset of high-resolution rgb-labelled images to develop and benchmark wheat head detection methods.
\newblock \emph{Plant Phenomics}, 2020, 2020.

\bibitem[Cou({\natexlab{c}})]{Counting-Pipe}
Object counting using pipe dataset, {\natexlab{c}}.
\newblock URL \url{https://www.kaggle.com/datasets/satya12389/object-counting-using-pipes-dataset}.

\bibitem[Chumpu()]{Multimodal-Neural-Translation}
Romrawin Chumpu.
\newblock Multimodal neural translation.
\newblock URL \url{https://huggingface.co/datasets/romrawinjp/multi30k/viewer/default/test}.

\bibitem[Minderer et~al.(2022{\natexlab{b}})Minderer, Gritsenko, Stone, Neumann, Weissenborn, Dosovitskiy, Mahendran, Arnab, Dehghani, Shen, Wang, Zhai, Kipf, and Houlsby]{OWL-VIT-abs-2205-06230}
Matthias Minderer, Alexey~A. Gritsenko, Austin Stone, Maxim Neumann, Dirk Weissenborn, Alexey Dosovitskiy, Aravindh Mahendran, Anurag Arnab, Mostafa Dehghani, Zhuoran Shen, Xiao Wang, Xiaohua Zhai, Thomas Kipf, and Neil Houlsby.
\newblock Simple open-vocabulary object detection with vision transformers.
\newblock \emph{CoRR}, abs/2205.06230, 2022{\natexlab{b}}.

\bibitem[mak()]{make-ml}
Mask dataset.
\newblock URL \url{https://makeml.app/datasets/mask}.

\bibitem[Cheng et~al.(2024)Cheng, Song, Ge, Liu, Wang, and Shan]{YOLO-World-ChengSG0WS24}
Tianheng Cheng, Lin Song, Yixiao Ge, Wenyu Liu, Xinggang Wang, and Ying Shan.
\newblock Yolo-world: Real-time open-vocabulary object detection.
\newblock In \emph{{IEEE/CVF} Conference on Computer Vision and Pattern Recognition, {CVPR} 2024, Seattle, WA, USA, June 16-22, 2024}, pages 16901--16911. {IEEE}, 2024.
\newblock \doi{10.1109/CVPR52733.2024.01599}.
\newblock URL \url{https://doi.org/10.1109/CVPR52733.2024.01599}.

\bibitem[Far()]{Far-Vehicle-Det}
Small vehicle detection| far vehicle detection.
\newblock URL \url{https://www.kaggle.com/datasets/dataclusterlabs/small-vehicle-images-far-vehicle-detection}.

\bibitem[sma()]{small-object-detection}
Grand dataset for small object detection.
\newblock URL \url{https://www.kaggle.com/datasets/waelyahyayaseen/grand-dataset-for-small-object-detection}.

\bibitem[Tra()]{Traffic-Signs-Detection}
Traffic signs detection.
\newblock URL \url{https://www.kaggle.com/datasets/pkdarabi/cardetection}.

\bibitem[Bon()]{Bone-Fracture-Detection}
Bone fracture detection: Computer vision project.
\newblock URL \url{https://www.kaggle.com/datasets/pkdarabi/bone-fracture-detection-computer-vision-project}.

\bibitem[AhmedMohsen(2022)]{drone-detection-new-peksv_dataset}
AhmedMohsen.
\newblock drone-detection-new dataset, 2022.
\newblock URL \url{https://universe.roboflow.com/ahmedmohsen/drone-detection-new-peksv}.

\bibitem[Hospital(2024)]{hospital-i5qoz_dataset}
Hospital.
\newblock Hospital dataset, 2024.
\newblock URL \url{https://universe.roboflow.com/hospital-peeox/hospital-i5qoz}.

\bibitem[Project(2024)]{wound-care-oufqn_dataset}
Final~Year Project.
\newblock Wound care dataset, 2024.
\newblock URL \url{https://universe.roboflow.com/final-year-project-ybo2r/wound-care-oufqn}.

\bibitem[Wannakrairot(2023)]{gauge-detection-cohbz_dataset}
Anupong Wannakrairot.
\newblock Gauge detection dataset, 2023.
\newblock URL \url{https://universe.roboflow.com/anupong-wannakrairot-ud9ty/gauge-detection-cohbz}.

\bibitem[Workspace(2024)]{car-parking-occupation_dataset}
Minervas Workspace.
\newblock Car parking occupation dataset, 2024.
\newblock URL \url{https://universe.roboflow.com/minervas-workspace-7comk/car-parking-occupation}.

\bibitem[patrick Cai(2023)]{tre-o6dcm_dataset}
patrick Cai.
\newblock tre dataset, 2023.
\newblock URL \url{https://universe.roboflow.com/patrick-cai-gfgl7/tre-o6dcm}.

\bibitem[artz(2024)]{traffic_light-28zwo_dataset}
artz.
\newblock traffic\_light dataset, 2024.
\newblock URL \url{https://universe.roboflow.com/artz/traffic\_light-28zwo}.

\bibitem[Roboflow(2025)]{rock-paper-scissors-sxsw_dataset}
Roboflow.
\newblock rock-paper-scissors dataset, 2025.
\newblock URL \url{https://universe.roboflow.com/roboflow-58fyf/rock-paper-scissors-sxsw}.

\bibitem[alkud12(2024)]{pricetag-4d8lm_dataset}
alkud12.
\newblock Pricetag dataset, 2024.
\newblock URL \url{https://universe.roboflow.com/alkud12/pricetag-4d8lm}.

\bibitem[Polyu(2022)]{retail-object-detection-bqtzf_dataset}
Polyu.
\newblock Retail object detection dataset, 2022.
\newblock URL \url{https://universe.roboflow.com/polyu-9ziss/retail-object-detection-bqtzf}.

\bibitem[KIT(2023)]{exit-sign-extended-version_dataset}
KIT.
\newblock exit-sign-extended version dataset, 2023.
\newblock URL \url{https://universe.roboflow.com/kit-g1foh/exit-sign-extended-version}.

\bibitem[Hoang(2024)]{handwritten-text-recognition-wb2o4_dataset}
Hoang.
\newblock Handwritten-text-recognition dataset, 2024.
\newblock URL \url{https://universe.roboflow.com/hoang-r9hhb/handwritten-text-recognition-wb2o4}.

\bibitem[language(2024)]{signlanguage-5kmtf_dataset}
Signs language.
\newblock Signlanguage dataset, 2024.
\newblock URL \url{https://universe.roboflow.com/signs-language/signlanguage-5kmtf}.

\bibitem[FootballVideoTrackingApp(2024)]{football-video-tracking-project_dataset}
FootballVideoTrackingApp.
\newblock football-video-tracking-project dataset, 2024.
\newblock URL \url{https://universe.roboflow.com/footballvideotrackingapp/football-video-tracking-project}.

\bibitem[intern(2024)]{pest-detection-m0inx_dataset}
intern.
\newblock pest detection dataset, 2024.
\newblock URL \url{https://universe.roboflow.com/intern-tvkth/pest-detection-m0inx}.

\bibitem[faiza rehman(2024)]{underwater-garbage-detection-d4dnn_dataset}
faiza rehman.
\newblock Underwater garbage detection dataset, 2024.
\newblock URL \url{https://universe.roboflow.com/faiza-rehman/underwater-garbage-detection-d4dnn}.

\bibitem[Myworkspace(2024)]{ingredient-detection-d6nwz_dataset}
Myworkspace.
\newblock Ingredient detection dataset, 2024.
\newblock URL \url{https://universe.roboflow.com/myworkspace-awwul/ingredient-detection-d6nwz}.

\bibitem[100(2024)]{circuit-elements_dataset}
Roboflow 100.
\newblock circuit elements dataset, 2024.
\newblock URL \url{https://universe.roboflow.com/roboflow-100/circuit-elements}.

\bibitem[100(2023{\natexlab{a}})]{tabular-data-wf9uh_dataset}
Roboflow 100.
\newblock tabular data dataset, 2023{\natexlab{a}}.
\newblock URL \url{https://universe.roboflow.com/roboflow-100/tabular-data-wf9uh}.

\bibitem[100(2023{\natexlab{b}})]{radio-signal_dataset}
Roboflow 100.
\newblock radio signal dataset.
\newblock \url{ https://universe.roboflow.com/roboflow-100/radio-signal }, 2023{\natexlab{b}}.
\newblock URL \url{https://universe.roboflow.com/roboflow-100/radio-signal}.

\bibitem[week3day3(2024)]{empty-shelf-detector-wt4im_dataset}
week3day3.
\newblock Empty shelf detector dataset, 2024.
\newblock URL \url{https://universe.roboflow.com/week3day3/empty-shelf-detector-wt4im}.

\bibitem[Tom()]{Tomato-Dataset}
Tomato dataset.
\newblock URL \url{https://makeml.app/datasets/tomato}.

\bibitem[Liu et~al.(2024{\natexlab{e}})Liu, Zeng, Ren, Li, Zhang, Yang, Jiang, Li, Yang, Su, Zhu, and Zhang]{Grounding-DINO-LiuZRLZYJLYSZZ24}
Shilong Liu, Zhaoyang Zeng, Tianhe Ren, Feng Li, Hao Zhang, Jie Yang, Qing Jiang, Chunyuan Li, Jianwei Yang, Hang Su, Jun Zhu, and Lei Zhang.
\newblock Grounding {DINO:} marrying {DINO} with grounded pre-training for open-set object detection.
\newblock In \emph{Computer Vision - {ECCV} 2024 - 18th European Conference, Milan, Italy, September 29-October 4, 2024, Proceedings, Part {XLVII}}, pages 38--55, 2024{\natexlab{e}}.

\bibitem[Dro()]{Drone-Detection}
Drone detection.
\newblock URL \url{https://www.kaggle.com/datasets/cybersimar08/drone-detection}.

\bibitem[Bee()]{Bee-Detection-Dataset}
Bee detection dataset.
\newblock URL \url{https://www.kaggle.com/datasets/lara311/bee-detection-dataset}.

\bibitem[Cat()]{Cat-Faces-Detection}
Cat faces detection.
\newblock URL \url{https://www.kaggle.com/datasets/aleksandrdremov/cat-faces-detection}.

\bibitem[CUB()]{CUB-200-Bird-Species}
Cub 200 bird species xml detection dataset.
\newblock URL \url{https://www.kaggle.com/datasets/sovitrath/cub-200-bird-species-xml-detection-dataset}.

\bibitem[Dee()]{Deep-Fish-Object-Detection}
Deep fish object detection.
\newblock URL \url{https://www.kaggle.com/datasets/vencerlanz09/deep-fish-object-detection}.

\bibitem[Tran()]{Trash-Panda-Detection}
Dat Tran.
\newblock Trash panda detection.
\newblock URL \url{https://www.kaggle.com/datasets/andrewmvd/racoon-detection}.

\bibitem[Shi()]{Ships-Data}
Ships data.

\bibitem[Wee()]{Weed-Detection}
Weed detection.
\newblock URL \url{https://www.kaggle.com/datasets/jaidalmotra/weed-detection}.

\bibitem[SAR()]{SARscope}
Sarscope: Synthetic aperture radar maritime images.
\newblock URL \url{https://www.kaggle.com/datasets/kailaspsudheer/sarscope-unveiling-the-maritime-landscape/data}.

\bibitem[Li et~al.(2022{\natexlab{c}})Li, Zhang, Maybank, and Tao]{li2022bridging}
Jizhizi Li, Jing Zhang, Stephen~J Maybank, and Dacheng Tao.
\newblock Bridging composite and real: towards end-to-end deep image matting.
\newblock \emph{International Journal of Computer Vision}, 130\penalty0 (2):\penalty0 246--266, 2022{\natexlab{c}}.

\bibitem[Li et~al.(2023{\natexlab{f}})Li, Jain, and Shi]{MAM-li2023matting}
Jiachen Li, Jitesh Jain, and Humphrey Shi.
\newblock Matting anything.
\newblock \emph{arXiv: 2306.05399}, 2023{\natexlab{f}}.

\bibitem[Li et~al.(2021)Li, Ma, Zhang, and Tao]{PM-10K}
Jizhizi Li, Sihan Ma, Jing Zhang, and Dacheng Tao.
\newblock Privacy-preserving portrait matting.
\newblock In \emph{Proceedings of the 29th ACM International Conference on Multimedia}, page 3501–3509, 2021.

\bibitem[Tex()]{Text-Manipulation-Detection}
Icdar 2023 dtt in images 2: Text manipulation detection.
\newblock URL \url{https://tianchi.aliyun.com/competition/entrance/532052/information}.

\bibitem[Xiao et~al.(2021)Xiao, Yu, Han, Zheng, and Fu]{SketchHairSalon-xiao2021}
Chufeng Xiao, Deng Yu, Xiaoguang Han, Youyi Zheng, and Hongbo Fu.
\newblock Sketchhairsalon: Deep sketch-based hair image synthesis.
\newblock \emph{ACM Transactions on Graphics (Proceedings of ACM SIGGRAPH Asia 2021)}, pages 1--16, 2021.

\bibitem[Ani({\natexlab{a}})]{Animals-10}
Animals-10, {\natexlab{a}}.
\newblock URL \url{https://www.kaggle.com/datasets/alessiocorrado99/animals10}.

\bibitem[Big()]{Big-Cats-Image}
Big cats image classification dataset.
\newblock URL \url{https://www.kaggle.com/datasets/patriciabrezeanu/big-cats-image-classification-dataset}.

\bibitem[Fru()]{Fruit-Recognition}
Fruit recognition.
\newblock URL \url{https://www.kaggle.com/datasets/chrisfilo/fruit-recognition}.

\bibitem[Ind()]{Indonesian-Wayang-Types}
Indonesian wayang types.
\newblock URL \url{https://www.kaggle.com/datasets/gibranfadilla/indonesian-wayang-types}.

\bibitem[Veg()]{Vegetable-Image-Dataset}
Vegetable image dataset.
\newblock URL \url{https://www.kaggle.com/datasets/misrakahmed/vegetable-image-dataset}.

\bibitem[Car({\natexlab{b}})]{Car_Logo_Dataset}
Car logo dataset, {\natexlab{b}}.
\newblock URL \url{https://www.kaggle.com/datasets/mahaarajab/car-logo-dataset}.

\bibitem[Egg()]{Egg-Image-Dataset}
Egg image dataset.
\newblock URL \url{https://www.kaggle.com/datasets/abdullahkhanuet22/eggs-images-classification-damaged-or-not}.

\bibitem[Dog()]{Dog-Breeds}
Dog breeds.
\newblock URL \url{https://www.kaggle.com/datasets/mohamedchahed/dog-breeds}.

\bibitem[Gar()]{Garbage-Classification}
Garbage classification.
\newblock URL \url{https://www.kaggle.com/datasets/quangtheng/garbage-classification-6-classes-775class}.

\bibitem[Mam()]{Mammals-classification}
Mammals classification.
\newblock URL \url{https://www.kaggle.com/datasets/anirudhg15/mammals-classification}.

\bibitem[Ute()]{Utensil-Image-Recognition}
Utensil image recognition.
\newblock URL \url{https://www.kaggle.com/datasets/jehanbhathena/utensil-image-recognition}.

\bibitem[Boa()]{Boat-Types-Recognition}
Boat types recognition.
\newblock URL \url{https://www.kaggle.com/datasets/clorichel/boat-types-recognition}.

\bibitem[Wang et~al.(2009)Wang, Markert, and Everingham]{Wang09}
Josiah Wang, Katja Markert, and Mark Everingham.
\newblock Learning models for object recognition from natural language descriptions.
\newblock In \emph{Proceedings of the British Machine Vision Conference}, 2009.

\bibitem[Fis()]{Fish-Recognition}
Fish recognition ground-truth dataset.
\newblock URL \url{https://www.kaggle.com/datasets/madhushreesannigrahi/fish-recognition-ground-truth-data}.

\bibitem[Bal()]{Ball_Classification}
Ball classification.
\newblock URL \url{https://huggingface.co/datasets/Asseh/Ball_Classification}.

\bibitem[flo()]{flower_classification}
Flower classification.
\newblock URL \url{https://huggingface.co/datasets/tian9/flower_classification}.

\bibitem[dat()]{data-food-category-classification}
Food category classification.
\newblock URL \url{https://huggingface.co/datasets/Kaludi/data-food-category-classification}.

\bibitem[Mat()]{Material_Classification}
Material classification.
\newblock URL \url{https://huggingface.co/datasets/Factral/Material_Classification}.

\bibitem[pla()]{plant_classification_dataset}
Plant classification.
\newblock URL \url{https://huggingface.co/datasets/sxdave/plant_classification_dataset}.

\bibitem[tra()]{trash_classification}
Trash classification.
\newblock URL \url{https://huggingface.co/datasets/ethanwan/trash_classification}.

\bibitem[veh()]{vehicle-classification}
Vehicle classification.
\newblock URL \url{https://huggingface.co/datasets/aryadytm/vehicle-classification}.

\bibitem[Bir()]{Bird-Species-Classification}
Bird species classification.
\newblock URL \url{https://www.kaggle.com/datasets/akash2907/bird-species-classification}.

\bibitem[Chi({\natexlab{b}})]{Chinese-Fine-Art}
Chinese fine art, {\natexlab{b}}.
\newblock URL \url{https://www.kaggle.com/datasets/rickyjli/chinese-fine-art}.

\bibitem[Fam()]{Famous-Iconic-Women}
Famous iconic women.
\newblock URL \url{https://www.kaggle.com/datasets/fatiimaezzahra/famous-iconic-women}.

\bibitem[Fac({\natexlab{b}})]{Face-Dataset}
Gender detection and classification - face dataset, {\natexlab{b}}.
\newblock URL \url{https://www.kaggle.com/datasets/trainingdatapro/gender-detection-and-classification-image-dataset}.

\bibitem[Rad({\natexlab{b}})]{Radar-Threat-Object}
Radar threat object classification, {\natexlab{b}}.
\newblock URL \url{https://www.kaggle.com/datasets/rauldbcet/radar-threat-object-classification}.

\bibitem[Rea({\natexlab{a}})]{RealWaste-Image-Classification}
Realwaste image classification, {\natexlab{a}}.
\newblock URL \url{https://www.kaggle.com/datasets/joebeachcapital/realwaste}.

\bibitem[Mic()]{Micro-Organism-Image-Classification}
Micro-organism image classification.
\newblock URL \url{https://www.kaggle.com/datasets/mdwaquarazam/microorganism-image-classification}.

\bibitem[Yu et~al.(2020{\natexlab{a}})Yu, Jiang, Yang, and Xia]{Twitter-17-YuJYX20}
Jianfei Yu, Jing Jiang, Li~Yang, and Rui Xia.
\newblock Improving multimodal named entity recognition via entity span detection with unified multimodal transformer.
\newblock In \emph{Proceedings of the 58th Annual Meeting of the Association for Computational Linguistics, {ACL} 2020, Online, July 5-10, 2020}, pages 3342--3352, 2020{\natexlab{a}}.

\bibitem[Yu et~al.(2020{\natexlab{b}})Yu, Jiang, Yang, and Xia]{UMT-YuJYX20}
Jianfei Yu, Jing Jiang, Li~Yang, and Rui Xia.
\newblock Improving multimodal named entity recognition via entity span detection with unified multimodal transformer.
\newblock In \emph{Proceedings of the 58th Annual Meeting of the Association for Computational Linguistics, {ACL} 2020, Online, July 5-10, 2020}, pages 3342--3352, 2020{\natexlab{b}}.

\bibitem[Yan et~al.(2024{\natexlab{a}})Yan, Yao, Chen, Zhao, Fu, Zhu, Luo, Yuan, Wang, Ding, et~al.]{DF40-yan2024}
Zhiyuan Yan, Taiping Yao, Shen Chen, Yandan Zhao, Xinghe Fu, Junwei Zhu, Donghao Luo, Li~Yuan, Chengjie Wang, Shouhong Ding, et~al.
\newblock Df40: Toward next-generation deepfake detection.
\newblock \emph{arXiv preprint arXiv:2406.13495}, 2024{\natexlab{a}}.

\bibitem[Tan et~al.(2024)Tan, Zhao, Wei, Gu, Liu, and Wei]{tan2024rethinking}
Chuangchuang Tan, Yao Zhao, Shikui Wei, Guanghua Gu, Ping Liu, and Yunchao Wei.
\newblock Rethinking the up-sampling operations in cnn-based generative network for generalizable deepfake detection.
\newblock In \emph{Proceedings of the IEEE/CVF Conference on Computer Vision and Pattern Recognition}, pages 28130--28139, 2024.

\bibitem[Cheng et~al.(2020)Cheng, Collins, Zhu, Liu, Huang, Adam, and Chen]{Panoptic-DeepLab-ChengCZ0HAC20}
Bowen Cheng, Maxwell~D. Collins, Yukun Zhu, Ting Liu, Thomas~S. Huang, Hartwig Adam, and Liang{-}Chieh Chen.
\newblock Panoptic-deeplab: {A} simple, strong, and fast baseline for bottom-up panoptic segmentation.
\newblock In \emph{2020 {IEEE/CVF} Conference on Computer Vision and Pattern Recognition, {CVPR} 2020, Seattle, WA, USA, June 13-19, 2020}, pages 12472--12482, 2020.

\bibitem[Yog()]{Yoga-Pose-Classification}
Yoga pose classification.
\newblock URL \url{https://www.kaggle.com/datasets/ujjwalchowdhury/yoga-pose-classification}.

\bibitem[Zhao et~al.(2022)Zhao, Wei, Lin, Sun, Zhang, and Zhang]{VSD-ZhaoWLSZ022}
Yu~Zhao, Jianguo Wei, Zhichao Lin, Yueheng Sun, Meishan Zhang, and Min Zhang.
\newblock Visual spatial description: Controlled spatial-oriented image-to-text generation.
\newblock In \emph{Proceedings of the 2022 Conference on Empirical Methods in Natural Language Processing, {EMNLP} 2022, Abu Dhabi, United Arab Emirates, December 7-11, 2022}, pages 1437--1449, 2022.

\bibitem[Yang et~al.(2024{\natexlab{b}})Yang, Wang, Ji, and Wu]{yang2024multi}
Shuo Yang, Yongqi Wang, Xiaofeng Ji, and Xinxiao Wu.
\newblock Multi-modal prompting for open-vocabulary video visual relationship detection.
\newblock In \emph{Proceedings of the AAAI Conference on Artificial Intelligence}, volume~38, pages 6513--6521, 2024{\natexlab{b}}.

\bibitem[Yin et~al.(2024)Yin, Zhang, Li, Liu, Cheng, and Hou]{DFormer-YinZLLCH24}
Bowen Yin, Xuying Zhang, Zhong{-}Yu Li, Li~Liu, Ming{-}Ming Cheng, and Qibin Hou.
\newblock Dformer: Rethinking {RGBD} representation learning for semantic segmentation.
\newblock In \emph{The Twelfth International Conference on Learning Representations, {ICLR} 2024, Vienna, Austria, May 7-11, 2024}, 2024.

\bibitem[Str()]{Strawberry-Classification}
Strawberry classification.
\newblock URL \url{https://www.kaggle.com/datasets/abdulbasit31/strawberry-dataset}.

\bibitem[RFES(2024)]{fire-smoke-detection-eozii-nfuzs-vgo4c_dataset}
RFES.
\newblock Fire and smoke detection dataset, 2024.
\newblock URL \url{https://universe.roboflow.com/rfes/fire-smoke-detection-eozii-nfuzs-vgo4c}.

\bibitem[Jiraphat(2023)]{gun-with-webcam-views_dataset}
Jiraphat.
\newblock Gun with webcam views dataset, 2023.
\newblock URL \url{https://universe.roboflow.com/jiraphat-stgwm/gun-with-webcam-views}.

\bibitem[100(2023{\natexlab{c}})]{construction-safety-gsnvb_dataset}
Roboflow 100.
\newblock Construction safety dataset, 2023{\natexlab{c}}.
\newblock URL \url{https://universe.roboflow.com/roboflow-100/construction-safety-gsnvb}.

\bibitem[Projects(2024)]{safety-vests_dataset}
Roboflow~Universe Projects.
\newblock Safety vests dataset, 2024.
\newblock URL \url{https://universe.roboflow.com/roboflow-universe-projects/safety-vests}.

\bibitem[custom yolov5(2024)]{fall_person_dataset}
custom yolov5.
\newblock Fall person dataset, 2024.
\newblock URL \url{https://universe.roboflow.com/custom-yolov5-wm0zy/fall_person}.

\bibitem[heejin(2024)]{smoke-detection1-owusa_dataset}
heejin.
\newblock Smoke detection1 dataset, 2024.
\newblock URL \url{https://universe.roboflow.com/heejin/smoke-detection1-owusa}.

\bibitem[Bik()]{Bikes-Helmets-Dataset}
Bikes helmets dataset.
\newblock URL \url{https://www.kaggle.com/datasets/andrewmvd/helmet-detection?select=images}.

\bibitem[Krishna et~al.(2017)Krishna, Zhu, Groth, Johnson, Hata, Kravitz, Chen, Kalantidis, Li, Shamma, Bernstein, and Fei{-}Fei]{Visual-Genome-KrishnaZGJHKCKL17}
Ranjay Krishna, Yuke Zhu, Oliver Groth, Justin Johnson, Kenji Hata, Joshua Kravitz, Stephanie Chen, Yannis Kalantidis, Li{-}Jia Li, David~A. Shamma, Michael~S. Bernstein, and Li~Fei{-}Fei.
\newblock Visual genome: Connecting language and vision using crowdsourced dense image annotations.
\newblock \emph{Int. J. Comput. Vis.}, 123\penalty0 (1):\penalty0 32--73, 2017.

\bibitem[Yang et~al.(2022{\natexlab{a}})Yang, Liu, Wu, Wang, Qian, He, and Cai]{Graph-RCNN-YangLWWQHC22}
Honghui Yang, Zili Liu, Xiaopei Wu, Wenxiao Wang, Wei Qian, Xiaofei He, and Deng Cai.
\newblock Graph {R-CNN:} towards accurate 3d object detection with semantic-decorated local graph.
\newblock In \emph{Computer Vision - {ECCV} 2022 - 17th European Conference, Tel Aviv, Israel, October 23-27, 2022, Proceedings, Part {VIII}}, pages 662--679, 2022{\natexlab{a}}.

\bibitem[Ter()]{Terrain-Recognition}
Terrain recognition.
\newblock URL \url{https://www.kaggle.com/datasets/krishuppal/terrain-recognition}.

\bibitem[Lan()]{Landscape-Recognition}
Landscape recognition.
\newblock URL \url{https://www.kaggle.com/datasets/utkarshsaxenadn/landscape-recognition-image-dataset-12k-images}.

\bibitem[Ame()]{American-Sign-Language-Dataset}
American sign language dataset.
\newblock URL \url{https://www.kaggle.com/datasets/ayuraj/asl-dataset}.

\bibitem[Jocher(2020)]{yolov5}
Glenn Jocher.
\newblock Ultralytics yolov5, 2020.
\newblock URL \url{https://github.com/ultralytics/yolov5}.

\bibitem[Bodur et~al.(2024)Bodur, Gundogdu, Bhattarai, Kim, Donoser, and Bazzani]{iEdit-BodurGB0DB22}
Rumeysa Bodur, Erhan Gundogdu, Binod Bhattarai, Tae{-}Kyun Kim, Michael Donoser, and Loris Bazzani.
\newblock iedit: Localised text-guided image editing with weak supervision.
\newblock In \emph{{IEEE/CVF} Conference on Computer Vision and Pattern Recognition, {CVPR} 2024 - Workshops, Seattle, WA, USA, June 17-18, 2024}, pages 7426--7435, 2024.

\bibitem[Jhamtani and Berg-Kirkpatrick(2018)]{jhamtani2018learning}
Harsh Jhamtani and Taylor Berg-Kirkpatrick.
\newblock Learning to describe differences between pairs of similar images.
\newblock In \emph{Proceedings of the 2018 Conference on Empirical Methods in Natural Language Processing (EMNLP)}, 2018.

\bibitem[Johnson et~al.(2017)Johnson, Hariharan, van~der Maaten, Fei{-}Fei, Zitnick, and Girshick]{CLEVR-JohnsonHMFZG17}
Justin Johnson, Bharath Hariharan, Laurens van~der Maaten, Li~Fei{-}Fei, C.~Lawrence Zitnick, and Ross~B. Girshick.
\newblock {CLEVR:} {A} diagnostic dataset for compositional language and elementary visual reasoning.
\newblock In \emph{2017 {IEEE} Conference on Computer Vision and Pattern Recognition, {CVPR} 2017, Honolulu, HI, USA, July 21-26, 2017}, pages 1988--1997, 2017.

\bibitem[Forbes et~al.(2019)Forbes, Kaeser{-}Chen, Sharma, and Belongie]{Birds-to-Words-ForbesKSB19}
Maxwell Forbes, Christine Kaeser{-}Chen, Piyush Sharma, and Serge~J. Belongie.
\newblock Neural naturalist: Generating fine-grained image comparisons.
\newblock In \emph{Proceedings of the 2019 Conference on Empirical Methods in Natural Language Processing and the 9th International Joint Conference on Natural Language Processing, {EMNLP-IJCNLP} 2019, Hong Kong, China, November 3-7, 2019}, pages 708--717, 2019.

\bibitem[Zheng et~al.(2021)Zheng, Wu, Feng, Fu, and Cai]{MNRE-ZhengWFF021}
Changmeng Zheng, Zhiwei Wu, Junhao Feng, Ze~Fu, and Yi~Cai.
\newblock {MNRE:} {A} challenge multimodal dataset for neural relation extraction with visual evidence in social media posts.
\newblock In \emph{2021 {IEEE} International Conference on Multimedia and Expo, {ICME} 2021, Shenzhen, China, July 5-9, 2021}, pages 1--6, 2021.

\bibitem[Soares et~al.(2019)Soares, FitzGerald, Ling, and Kwiatkowski]{BertNRE-SoaresFLK19}
Livio~Baldini Soares, Nicholas FitzGerald, Jeffrey Ling, and Tom Kwiatkowski.
\newblock Matching the blanks: Distributional similarity for relation learning.
\newblock In \emph{Proceedings of the 57th Conference of the Association for Computational Linguistics, {ACL} 2019, Florence, Italy, July 28- August 2, 2019, Volume 1: Long Papers}, pages 2895--2905, 2019.

\bibitem[Ravi et~al.(2021)Ravi, Kafle, Cohen, Brandt, and Kapadia]{Ravi_2018_CVPR}
Hareesh Ravi, Kushal Kafle, Scott Cohen, Jonathan Brandt, and Mubbasir Kapadia.
\newblock Aesop: Abstract encoding of stories, objects and pictures.
\newblock In \emph{Proceedings of the IEEE/CVF International Conference on Computer Vision (ICCV)}, pages 2052--2063, October 2021.

\bibitem[Li et~al.(2019{\natexlab{a}})Li, Gan, Shen, Liu, Cheng, Wu, Carin, Carlson, and Gao]{StoryGAN-LiGSLCWCCG19}
Yitong Li, Zhe Gan, Yelong Shen, Jingjing Liu, Yu~Cheng, Yuexin Wu, Lawrence Carin, David~E. Carlson, and Jianfeng Gao.
\newblock Storygan: {A} sequential conditional {GAN} for story visualization.
\newblock In \emph{{IEEE} Conference on Computer Vision and Pattern Recognition, {CVPR} 2019, Long Beach, CA, USA, June 16-20, 2019}, pages 6329--6338, 2019{\natexlab{a}}.

\bibitem[Gupta et~al.(2018)Gupta, Schwenk, Farhadi, Hoiem, and Kembhavi]{FlintstonesSV-GuptaSFHK18}
Tanmay Gupta, Dustin Schwenk, Ali Farhadi, Derek Hoiem, and Aniruddha Kembhavi.
\newblock Imagine this! scripts to compositions to videos.
\newblock In \emph{Computer Vision - {ECCV} 2018 - 15th European Conference, Munich, Germany, September 8-14, 2018, Proceedings, Part {VIII}}, pages 610--626, 2018.

\bibitem[Huang et~al.(2016)Huang, Ferraro, Mostafazadeh, Misra, Agrawal, Devlin, Girshick, He, Kohli, Batra, Zitnick, Parikh, Vanderwende, Galley, and Mitchell]{VIST-HuangFMMADGHKBZ16}
Ting{-}Hao~(Kenneth) Huang, Francis Ferraro, Nasrin Mostafazadeh, Ishan Misra, Aishwarya Agrawal, Jacob Devlin, Ross~B. Girshick, Xiaodong He, Pushmeet Kohli, Dhruv Batra, C.~Lawrence Zitnick, Devi Parikh, Lucy Vanderwende, Michel Galley, and Margaret Mitchell.
\newblock Visual storytelling.
\newblock In \emph{{NAACL} {HLT} 2016, The 2016 Conference of the North American Chapter of the Association for Computational Linguistics: Human Language Technologies, San Diego California, USA, June 12-17, 2016}, pages 1233--1239, 2016.

\bibitem[Maharana et~al.(2022)Maharana, Hannan, and Bansal]{DiDeMoSV-MaharanaHB22}
Adyasha Maharana, Darryl Hannan, and Mohit Bansal.
\newblock Storydall-e: Adapting pretrained text-to-image transformers for story continuation.
\newblock In \emph{Computer Vision - {ECCV} 2022 - 17th European Conference, Tel Aviv, Israel, October 23-27, 2022, Proceedings, Part {XXXVII}}, pages 70--87, 2022.

\bibitem[Xiao(2021)]{DVN/M8JQCR_2021}
Haixia Xiao.
\newblock Weather phenomenon database (weapd), 2021.
\newblock URL \url{https://doi.org/10.7910/DVN/M8JQCR}.

\bibitem[Gao et~al.(2020)Gao, Liu, Xu, Wang, Liu, and Zou]{SketchyCOCO-GaoLXWLZ20}
Chengying Gao, Qi~Liu, Qi~Xu, Limin Wang, Jianzhuang Liu, and Changqing Zou.
\newblock Sketchycoco: Image generation from freehand scene sketches.
\newblock In \emph{2020 {IEEE/CVF} Conference on Computer Vision and Pattern Recognition, {CVPR} 2020, Seattle, WA, USA, June 13-19, 2020}, pages 5173--5182, 2020.

\bibitem[Wang et~al.(2022{\natexlab{b}})Wang, Zhang, Zhang, Ouyang, Chen, Chen, and Wen]{PITI-wang2022pretraining}
Tengfei Wang, Ting Zhang, Bo~Zhang, Hao Ouyang, Dong Chen, Qifeng Chen, and Fang Wen.
\newblock Pretraining is all you need for image-to-image translation.
\newblock \emph{arXiv:2205.12952}, 2022{\natexlab{b}}.

\bibitem[Bai et~al.(2024)Bai, Wang, Cao, Ge, Yuan, and Shan]{DreamDiffusion-BaiWCGYS24}
Yunpeng Bai, Xintao Wang, Yan{-}Pei Cao, Yixiao Ge, Chun Yuan, and Ying Shan.
\newblock Dreamdiffusion: High-quality eeg-to-image generation with temporal masked signal modeling and {CLIP} alignment.
\newblock In \emph{Computer Vision - {ECCV} 2024 - 18th European Conference, Milan, Italy, September 29-October 4, 2024, Proceedings, Part {XXXI}}, pages 472--488, 2024.

\bibitem[Wenhan~Yang and Yan(2017)]{Rain100H-Yang2017RainRemoval}
Jiashi Feng Jiaying Liu Zongming~Guo Wenhan~Yang, Robby T.~Tan and Shuicheng Yan.
\newblock Joint rain detection and removal from a single image.
\newblock \emph{IEEE Conference on Computer Vision and Pattern Recognition}, 2017.

\bibitem[Yue et~al.(2024{\natexlab{b}})Yue, Peng, Ma, Du, Wei, and Zhang]{GOUB-YuePMDWZ24}
Conghan Yue, Zhengwei Peng, Junlong Ma, Shiyan Du, Pengxu Wei, and Dongyu Zhang.
\newblock Image restoration through generalized ornstein-uhlenbeck bridge.
\newblock In \emph{Forty-first International Conference on Machine Learning, {ICML} 2024, Vienna, Austria, July 21-27, 2024}, 2024{\natexlab{b}}.

\bibitem[Qian et~al.(2018)Qian, Tan, Yang, Su, and Liu]{Raindrop-Qian-2018}
Rui Qian, Robby~T. Tan, Wenhan Yang, Jiajun Su, and Jiaying Liu.
\newblock Attentive generative adversarial network for raindrop removal from a single image.
\newblock In \emph{The IEEE Conference on Computer Vision and Pattern Recognition (CVPR)}, 2018.

\bibitem[Valanarasu et~al.(2021)Valanarasu, Yasarla, and Patel]{TransWeather-valanarasu2021}
Jeya Maria~Jose Valanarasu, Rajeev Yasarla, and Vishal~M. Patel.
\newblock Transweather: Transformer-based restoration of images degraded by adverse weather conditions, 2021.

\bibitem[Li et~al.(2019{\natexlab{b}})Li, Ren, Fu, Tao, Feng, Zeng, and Wang]{RESIDE-6K-LiRFTFZW19}
Boyi Li, Wenqi Ren, Dengpan Fu, Dacheng Tao, Dan Feng, Wenjun Zeng, and Zhangyang Wang.
\newblock Benchmarking single-image dehazing and beyond.
\newblock \emph{{IEEE} Trans. Image Process.}, 28\penalty0 (1):\penalty0 492--505, 2019{\natexlab{b}}.

\bibitem[Song et~al.(2023)Song, He, Qian, and Du]{DehazeFormer-SongHQD23}
Yuda Song, Zhuqing He, Hui Qian, and Xin Du.
\newblock Vision transformers for single image dehazing.
\newblock \emph{{IEEE} Trans. Image Process.}, 32:\penalty0 1927--1941, 2023.

\bibitem[Liu et~al.(2018)Liu, Jaw, Huang, and Hwang]{Snow100K-LiuJHH18}
Yun{-}Fu Liu, Da{-}Wei Jaw, Shih{-}Chia Huang, and Jenq{-}Neng Hwang.
\newblock Desnownet: Context-aware deep network for snow removal.
\newblock \emph{{IEEE} Trans. Image Process.}, 27\penalty0 (6):\penalty0 3064--3073, 2018.

\bibitem[Cui et~al.(2024)Cui, Ren, Cao, and Knoll]{ConvIR-CuiRCK24a}
Yuning Cui, Wenqi Ren, Xiaochun Cao, and Alois Knoll.
\newblock Revitalizing convolutional network for image restoration.
\newblock \emph{{IEEE} Trans. Pattern Anal. Mach. Intell.}, 46\penalty0 (12):\penalty0 9423--9438, 2024.

\bibitem[Nah et~al.(2017)Nah, Kim, and Lee]{GoPro-Nah-2017R}
Seungjun Nah, Tae~Hyun Kim, and Kyoung~Mu Lee.
\newblock Deep multi-scale convolutional neural network for dynamic scene deblurring.
\newblock In \emph{CVPR}, July 2017.

\bibitem[Xintian~Mao and Wang(2024)]{AdaRevD-xintm2024}
Qingli~Li Xintian~Mao and Yan Wang.
\newblock Adarevd: Adaptive patch exiting reversible decoder pushes the limit of image deblurring.
\newblock In \emph{Proc. CVPR}, 2024.

\bibitem[Agustsson and Timofte(2017)]{Agustsson_2017_CVPR_Workshops}
Eirikur Agustsson and Radu Timofte.
\newblock Ntire 2017 challenge on single image super-resolution: Dataset and study.
\newblock In \emph{The IEEE Conference on Computer Vision and Pattern Recognition (CVPR) Workshops}, 2017.

\bibitem[Luo et~al.(2024)Luo, Gustafsson, Zhao, Sj{\"{o}}lund, and Sch{\"{o}}n]{DA-CLIP-LuoG0SS24}
Ziwei Luo, Fredrik~K. Gustafsson, Zheng Zhao, Jens Sj{\"{o}}lund, and Thomas~B. Sch{\"{o}}n.
\newblock Controlling vision-language models for multi-task image restoration.
\newblock In \emph{The Twelfth International Conference on Learning Representations, {ICLR} 2024, Vienna, Austria, May 7-11, 2024}. OpenReview.net, 2024.

\bibitem[Huang et~al.(2020)Huang, Li, Yang, Sun, and Song]{OD-GAN-HuangLYSS20}
Binghui Huang, Zhi Li, Chao Yang, Fuchun Sun, and Yixu Song.
\newblock Single satellite optical imagery dehazing using {SAR} image prior based on conditional generative adversarial networks.
\newblock In \emph{{IEEE} Winter Conference on Applications of Computer Vision, {WACV} 2020, Snowmass Village, CO, USA, March 1-5, 2020}, pages 1795--1802, 2020.

\bibitem[Guo et~al.(2024{\natexlab{a}})Guo, Wang, Wang, Huang, Yang, Kot, and Wen]{SRD-guo2024single}
Laniqng Guo, Chong Wang, Yufei Wang, Siyu Huang, Wenhan Yang, Alex~C Kot, and Bihan Wen.
\newblock Single-image shadow removal using deep learning: A comprehensive survey.
\newblock \emph{arXiv preprint arXiv:2407.08865}, 2024{\natexlab{a}}.

\bibitem[Hu et~al.(2020)Hu, Fu, Zhu, Qin, and Heng]{DSC-HuFZQH20}
Xiaowei Hu, Chi{-}Wing Fu, Lei Zhu, Jing Qin, and Pheng{-}Ann Heng.
\newblock Direction-aware spatial context features for shadow detection and removal.
\newblock \emph{{IEEE} Trans. Pattern Anal. Mach. Intell.}, 42\penalty0 (11):\penalty0 2795--2808, 2020.

\bibitem[Dai et~al.(2022{\natexlab{a}})Dai, Li, Zhou, Feng, and Loy]{Flare7K-DaiLZFL22}
Yuekun Dai, Chongyi Li, Shangchen Zhou, Ruicheng Feng, and Chen~Change Loy.
\newblock Flare7k: {A} phenomenological nighttime flare removal dataset.
\newblock In \emph{Advances in Neural Information Processing Systems 35: Annual Conference on Neural Information Processing Systems 2022, NeurIPS 2022, New Orleans, LA, USA, November 28 - December 9, 2022}, 2022{\natexlab{a}}.

\bibitem[Zhang et~al.(2023{\natexlab{j}})Zhang, Ouyang, Liu, Wang, Kong, and Jin]{FF-Former-ZhangOLWKJ23}
Dafeng Zhang, Jia Ouyang, Guanqun Liu, Xiaobing Wang, Xiangyu Kong, and Zhezhu Jin.
\newblock Ff-former: Swin fourier transformer for nighttime flare removal.
\newblock In \emph{{IEEE/CVF} Conference on Computer Vision and Pattern Recognition, {CVPR} 2023 - Workshops, Vancouver, BC, Canada, June 17-24, 2023}, pages 2824--2832, 2023{\natexlab{j}}.

\bibitem[Peng et~al.(2023)Peng, Zhu, and Bian]{LSUI-10129222}
Lintao Peng, Chunli Zhu, and Liheng Bian.
\newblock U-shape transformer for underwater image enhancement.
\newblock \emph{IEEE Transactions on Image Processing}, 32:\penalty0 3066--3079, 2023.
\newblock \doi{10.1109/TIP.2023.3276332}.

\bibitem[Pucci and Martinel(2024)]{CE-VAE-pucci2024}
Rita Pucci and Niki Martinel.
\newblock Ce-vae: Capsule enhanced variational autoencoder for underwater image reconstruction.
\newblock Feb 2024.

\bibitem[Ma et~al.(2018)Ma, Shu, Bai, Wang, and Samaras]{DocUNet-MaSBWS18}
Ke~Ma, Zhixin Shu, Xue Bai, Jue Wang, and Dimitris Samaras.
\newblock Docunet: Document image unwarping via a stacked u-net.
\newblock In \emph{2018 {IEEE} Conference on Computer Vision and Pattern Recognition, {CVPR} 2018, Salt Lake City, UT, USA, June 18-22, 2018}, pages 4700--4709, 2018.

\bibitem[Zhu et~al.(2021)Zhu, Xu, Ke, and Lau]{FDRNet-Zhu0KL21}
Lei Zhu, Ke~Xu, Zhanghan Ke, and Rynson W.~H. Lau.
\newblock Mitigating intensity bias in shadow detection via feature decomposition and reweighting.
\newblock In \emph{2021 {IEEE/CVF} International Conference on Computer Vision, {ICCV} 2021, Montreal, QC, Canada, October 10-17, 2021}, pages 4682--4691, 2021.

\bibitem[Islam et~al.(2020)Islam, Xia, and Sattar]{FUnIE-GAN-IslamXS20}
Md~Jahidul Islam, Youya Xia, and Junaed Sattar.
\newblock Fast underwater image enhancement for improved visual perception.
\newblock \emph{{IEEE} Robotics Autom. Lett.}, 5\penalty0 (2):\penalty0 3227--3234, 2020.

\bibitem[Ma et~al.(2022)Ma, Ma, Liu, Fan, and Luo]{SCI-ma2022toward}
Long Ma, Tengyu Ma, Risheng Liu, Xin Fan, and Zhongxuan Luo.
\newblock Toward fast, flexible, and robust low-light image enhancement.
\newblock In \emph{Proceedings of the IEEE/CVF Conference on Computer Vision and Pattern Recognition}, pages 5637--5646, 2022.

\bibitem[Karras et~al.(2018)Karras, Aila, Laine, and Lehtinen]{CelebaHQ-256-KarrasALL18}
Tero Karras, Timo Aila, Samuli Laine, and Jaakko Lehtinen.
\newblock Progressive growing of gans for improved quality, stability, and variation.
\newblock In \emph{6th International Conference on Learning Representations, {ICLR} 2018, Vancouver, BC, Canada, April 30 - May 3, 2018, Conference Track Proceedings}, 2018.

\bibitem[Li et~al.(2022{\natexlab{d}})Li, Lin, Zhou, Qi, Wang, and Jia]{MAT-li2022}
Wenbo Li, Zhe Lin, Kun Zhou, Lu~Qi, Yi~Wang, and Jiaya Jia.
\newblock Mat: Mask-aware transformer for large hole image inpainting.
\newblock In \emph{Proceedings of the IEEE/CVF Conference on Computer Vision and Pattern Recognition}, 2022{\natexlab{d}}.

\bibitem[Degerli et~al.(2022)Degerli, Kiranyaz, Chowdhury, and Gabbouj]{QaTa-COV19-DegerliKCG22}
Aysen Degerli, Serkan Kiranyaz, Muhammad Enamul~Hoque Chowdhury, and Moncef Gabbouj.
\newblock Osegnet: Operational segmentation network for covid-19 detection using chest x-ray images.
\newblock In \emph{2022 {IEEE} International Conference on Image Processing, {ICIP} 2022, Bordeaux, France, 16-19 October 2022}, pages 2306--2310, 2022.

\bibitem[Degerli et~al.(2021)Degerli, Ahishali, Yamac, Kiranyaz, Chowdhury, Hameed, Hamid, Mazhar, and Gabbouj]{DDLA-DegerliAYKCHHMG21}
Aysen Degerli, Mete Ahishali, Mehmet Yamac, Serkan Kiranyaz, Muhammad Enamul~Hoque Chowdhury, Khalid Hameed, Tahir Hamid, Rashid Mazhar, and Moncef Gabbouj.
\newblock {COVID-19} infection map generation and detection from chest x-ray images.
\newblock \emph{Health Inf. Sci. Syst.}, 9\penalty0 (1):\penalty0 15, 2021.

\bibitem[Deng-Ping et~al.(2022)Deng-Ping, Ziling, Peng, Hong, Xuebin, and Luc]{FS2K-Fan2022}
Fan Deng-Ping, Huang Ziling, Zheng Peng, Liu Hong, Qin Xuebin, and Van~Gool Luc.
\newblock Facial-sketch synthesis: A new challenge.
\newblock \emph{Machine Intelligence Research}, 2022.

\bibitem[Gao et~al.(2023)Gao, Zhu, Jiang, and Wang]{HIDA-gao2023human}
Fei Gao, Yifan Zhu, Chang Jiang, and Nannan Wang.
\newblock Human-inspired facial sketch synthesis with dynamic adaptation.
\newblock In \emph{Proceedings of the IEEE/CVF International Conference on Computer Vision}, pages 7237--7247, 2023.

\bibitem[Brooks et~al.(2023)Brooks, Holynski, and Efros]{InstructPix2Pix-BrooksHE23}
Tim Brooks, Aleksander Holynski, and Alexei~A. Efros.
\newblock Instructpix2pix: Learning to follow image editing instructions.
\newblock In \emph{{IEEE/CVF} Conference on Computer Vision and Pattern Recognition, {CVPR} 2023, Vancouver, BC, Canada, June 17-24, 2023}, pages 18392--18402, 2023.

\bibitem[Pretty Face()]{Pretty-Face}
Pretty Face.
\newblock URL \url{https://www.kaggle.com/datasets/yewtsing/pretty-face}.

\bibitem[Zheng et~al.(2023{\natexlab{b}})Zheng, Zhou, Li, Qi, Shan, and Li]{LayoutDiffusion-ZhengZLQSL23}
Guangcong Zheng, Xianpan Zhou, Xuewei Li, Zhongang Qi, Ying Shan, and Xi~Li.
\newblock Layoutdiffusion: Controllable diffusion model for layout-to-image generation.
\newblock In \emph{{IEEE/CVF} Conference on Computer Vision and Pattern Recognition, {CVPR} 2023, Vancouver, BC, Canada, June 17-24, 2023}, pages 22490--22499, 2023{\natexlab{b}}.

\bibitem[Qin et~al.(2023{\natexlab{a}})Qin, Zhang, Yu, Feng, Yang, Zhou, Wang, Niebles, Xiong, Savarese, Ermon, Fu, and Xu]{UniControl-QinZYFYZWNXSE0X23}
Can Qin, Shu Zhang, Ning Yu, Yihao Feng, Xinyi Yang, Yingbo Zhou, Huan Wang, Juan~Carlos Niebles, Caiming Xiong, Silvio Savarese, Stefano Ermon, Yun Fu, and Ran Xu.
\newblock Unicontrol: {A} unified diffusion model for controllable visual generation in the wild.
\newblock In \emph{Advances in Neural Information Processing Systems 36: Annual Conference on Neural Information Processing Systems 2023, NeurIPS 2023, New Orleans, LA, USA, December 10 - 16, 2023}, 2023{\natexlab{a}}.

\bibitem[Wei et~al.(2022)Wei, Chen, Zhou, Liao, Tan, Yuan, Zhang, and Yu]{HairCLIP-Wei0Z0TY0Y22}
Tianyi Wei, Dongdong Chen, Wenbo Zhou, Jing Liao, Zhentao Tan, Lu~Yuan, Weiming Zhang, and Nenghai Yu.
\newblock Hairclip: Design your hair by text and reference image.
\newblock In \emph{{IEEE/CVF} Conference on Computer Vision and Pattern Recognition, {CVPR} 2022, New Orleans, LA, USA, June 18-24, 2022}, pages 18051--18060, 2022.

\bibitem[Qin et~al.(2023{\natexlab{b}})Qin, Yu, Xing, Zhang, Chen, Ermon, Fu, Xiong, and Xu]{GlueGen-QinYXZCE0XX23}
Can Qin, Ning Yu, Chen Xing, Shu Zhang, Zeyuan Chen, Stefano Ermon, Yun Fu, Caiming Xiong, and Ran Xu.
\newblock Gluegen: Plug and play multi-modal encoders for x-to-image generation.
\newblock In \emph{{IEEE/CVF} International Conference on Computer Vision, {ICCV} 2023, Paris, France, October 1-6, 2023}, pages 23028--23039, 2023{\natexlab{b}}.

\bibitem[Ima()]{Image-Colorization}
Image colorization dataset.
\newblock URL \url{https://www.kaggle.com/datasets/aayush9753/image-colorization-dataset}.

\bibitem[Wang et~al.(2023{\natexlab{c}})Wang, Yu, and Zhang]{DDNM-wang2022zero}
Yinhuai Wang, Jiwen Yu, and Jian Zhang.
\newblock Zero-shot image restoration using denoising diffusion null-space model.
\newblock \emph{The Eleventh International Conference on Learning Representations}, 2023{\natexlab{c}}.

\bibitem[Ge et~al.(2024{\natexlab{b}})Ge, Zhao, Li, Ge, and Shan]{SEED-Data-Edit-ge2024seed}
Yuying Ge, Sijie Zhao, Chen Li, Yixiao Ge, and Ying Shan.
\newblock Seed-data-edit technical report: A hybrid dataset for instructional image editing.
\newblock \emph{arXiv preprint arXiv:2405.04007}, 2024{\natexlab{b}}.

\bibitem[Ge et~al.(2024{\natexlab{c}})Ge, Zhao, Zhu, Ge, Yi, Song, Li, Ding, and Shan]{SEED-X-ge2024seed}
Yuying Ge, Sijie Zhao, Jinguo Zhu, Yixiao Ge, Kun Yi, Lin Song, Chen Li, Xiaohan Ding, and Ying Shan.
\newblock Seed-x: Multimodal models with unified multi-granularity comprehension and generation.
\newblock \emph{arXiv preprint arXiv:2404.14396}, 2024{\natexlab{c}}.

\bibitem[Huang et~al.(2023{\natexlab{b}})Huang, Chan, Jiang, and Liu]{huang2023collaborative}
Ziqi Huang, Kelvin~CK Chan, Yuming Jiang, and Ziwei Liu.
\newblock Collaborative diffusion for multi-modal face generation and editing.
\newblock In \emph{Proceedings of the IEEE/CVF Conference on Computer Vision and Pattern Recognition}, pages 6080--6090, 2023{\natexlab{b}}.

\bibitem[Wu et~al.(2023{\natexlab{c}})Wu, Li, He, Shou, Shen, Cheng, Li, Gao, Zhang, and Wang]{ParaDiffusion-abs-2311-14284}
Weijia Wu, Zhuang Li, Yefei He, Mike~Zheng Shou, Chunhua Shen, Lele Cheng, Yan Li, Tingting Gao, Di~Zhang, and Zhongyuan Wang.
\newblock Paragraph-to-image generation with information-enriched diffusion model.
\newblock \emph{CoRR}, abs/2311.14284, 2023{\natexlab{c}}.

\bibitem[Castro et~al.(2022)Castro, Deng, Huang, Burzo, and Mihalcea]{WildQA-2022-in-the-wild}
Santiago Castro, Naihao Deng, Pingxuan Huang, Mihai~G. Burzo, and Rada Mihalcea.
\newblock In-the-wild video question answering.
\newblock In \emph{COLING}, pages 5613--5635, October 2022.

\bibitem[Zhang et~al.(2024{\natexlab{f}})Zhang, Wu, Li, Li, Ma, Liu, and Li]{LLaVA-Video-72B-Qwen2}
Yuanhan Zhang, Jinming Wu, Wei Li, Bo~Li, Zejun Ma, Ziwei Liu, and Chunyuan Li.
\newblock Video instruction tuning with synthetic data.
\newblock \emph{CoRR}, abs/2410.02713, 2024{\natexlab{f}}.

\bibitem[Maaz et~al.(2024)Maaz, Rasheed, Khan, and Khan]{Maaz2024VideoGPT}
Muhammad Maaz, Hanoona Rasheed, Salman Khan, and Fahad~Shahbaz Khan.
\newblock Videogpt+: Integrating image and video encoders for enhanced video understanding.
\newblock \emph{arxiv}, 2024.
\newblock URL \url{https://arxiv.org/abs/2406.09418}.

\bibitem[Li et~al.(2024{\natexlab{i}})Li, Liu, Wu, Wang, Shen, Qu, Niu, Wang, Chen, and Li]{aria-dongxu}
Dongxu Li, Yudong Liu, Haoning Wu, Yue Wang, Zhiqi Shen, Bowen Qu, Xinyao Niu, Guoyin Wang, Bei Chen, and Junnan Li.
\newblock Aria: An open multimodal native mixture-of-experts model.
\newblock \emph{arXiv preprint arXiv:2410.05993}, 2024{\natexlab{i}}.

\bibitem[Li et~al.(2024{\natexlab{j}})Li, Deng, Ke, Liu, Rahmani, Guo, Schiele, and Chen]{Sports-QA-abs-2401-01505}
Haopeng Li, Andong Deng, Qiuhong Ke, Jun Liu, Hossein Rahmani, Yulan Guo, Bernt Schiele, and Chen Chen.
\newblock Sports-qa: {A} large-scale video question answering benchmark for complex and professional sports.
\newblock \emph{CoRR}, abs/2401.01505, 2024{\natexlab{j}}.

\bibitem[Yu et~al.(2019)Yu, Xu, Yu, Yu, Zhao, Zhuang, and Tao]{ActivityNet-QA-YuXYYZZT19}
Zhou Yu, Dejing Xu, Jun Yu, Ting Yu, Zhou Zhao, Yueting Zhuang, and Dacheng Tao.
\newblock Activitynet-qa: {A} dataset for understanding complex web videos via question answering.
\newblock In \emph{The Thirty-Third {AAAI} Conference on Artificial Intelligence, {AAAI} 2019, The Thirty-First Innovative Applications of Artificial Intelligence Conference, {IAAI} 2019, The Ninth {AAAI} Symposium on Educational Advances in Artificial Intelligence, {EAAI} 2019, Honolulu, Hawaii, USA, January 27 - February 1, 2019}, pages 9127--9134, 2019.

\bibitem[Wang et~al.(2024{\natexlab{g}})Wang, Bai, Tan, Wang, Fan, Bai, Chen, Liu, Wang, Ge, Fan, Dang, Du, Ren, Men, Liu, Zhou, Zhou, and Lin]{Qwen2-VL-abs-2409-12191}
Peng Wang, Shuai Bai, Sinan Tan, Shijie Wang, Zhihao Fan, Jinze Bai, Keqin Chen, Xuejing Liu, Jialin Wang, Wenbin Ge, Yang Fan, Kai Dang, Mengfei Du, Xuancheng Ren, Rui Men, Dayiheng Liu, Chang Zhou, Jingren Zhou, and Junyang Lin.
\newblock Qwen2-vl: Enhancing vision-language model's perception of the world at any resolution.
\newblock \emph{CoRR}, abs/2409.12191, 2024{\natexlab{g}}.

\bibitem[Li et~al.(2024{\natexlab{k}})Li, Wang, He, Li, Wang, Liu, Wang, Xu, Chen, Lou, Wang, and Qiao]{MVBench-0002WH00LWX0L0024}
Kunchang Li, Yali Wang, Yinan He, Yizhuo Li, Yi~Wang, Yi~Liu, Zun Wang, Jilan Xu, Guo Chen, Ping Lou, Limin Wang, and Yu~Qiao.
\newblock Mvbench: {A} comprehensive multi-modal video understanding benchmark.
\newblock In \emph{{IEEE/CVF} Conference on Computer Vision and Pattern Recognition, {CVPR} 2024, Seattle, WA, USA, June 16-22, 2024}, pages 22195--22206, 2024{\natexlab{k}}.

\bibitem[Fu et~al.(2024{\natexlab{c}})Fu, Dai, Luo, Li, Ren, Zhang, Wang, Zhou, Shen, Zhang, Chen, Li, Lin, Zhao, Li, Xu, Zheng, Chen, Ji, and Sun]{Video-MME-abs-2405-21075}
Chaoyou Fu, Yuhan Dai, Yondong Luo, Lei Li, Shuhuai Ren, Renrui Zhang, Zihan Wang, Chenyu Zhou, Yunhang Shen, Mengdan Zhang, Peixian Chen, Yanwei Li, Shaohui Lin, Sirui Zhao, Ke~Li, Tong Xu, Xiawu Zheng, Enhong Chen, Rongrong Ji, and Xing Sun.
\newblock Video-mme: The first-ever comprehensive evaluation benchmark of multi-modal llms in video analysis.
\newblock \emph{CoRR}, abs/2405.21075, 2024{\natexlab{c}}.

\bibitem[Liu et~al.(2024{\natexlab{f}})Liu, Dong, Liu, Hu, Lu, and Rao]{Oryx-abs-2409-12961}
Zuyan Liu, Yuhao Dong, Ziwei Liu, Winston Hu, Jiwen Lu, and Yongming Rao.
\newblock Oryx {MLLM:} on-demand spatial-temporal understanding at arbitrary resolution.
\newblock \emph{CoRR}, abs/2409.12961, 2024{\natexlab{f}}.

\bibitem[Fang et~al.(2024)Fang, Mao, Duan, Zhao, Li, Lin, and Chen]{MMBench-Video-abs-2406-14515}
Xinyu Fang, Kangrui Mao, Haodong Duan, Xiangyu Zhao, Yining Li, Dahua Lin, and Kai Chen.
\newblock Mmbench-video: {A} long-form multi-shot benchmark for holistic video understanding.
\newblock \emph{CoRR}, abs/2406.14515, 2024.

\bibitem[Li et~al.(2013)Li, Kim, Humayun, Tsai, and Rehg]{LiKHTR13}
Fuxin Li, Taeyoung Kim, Ahmad Humayun, David Tsai, and James~M. Rehg.
\newblock Video segmentation by tracking many figure-ground segments.
\newblock In \emph{{IEEE} International Conference on Computer Vision, {ICCV} 2013, Sydney, Australia, December 1-8, 2013}, pages 2192--2199, 2013.

\bibitem[Ochs et~al.(2014)Ochs, Malik, and Brox]{OchsMB14}
Peter Ochs, Jitendra Malik, and Thomas Brox.
\newblock Segmentation of moving objects by long term video analysis.
\newblock \emph{{IEEE} Trans. Pattern Anal. Mach. Intell.}, 36\penalty0 (6):\penalty0 1187--1200, 2014.

\bibitem[Lin et~al.(2024{\natexlab{b}})Lin, Zhu, Shen, Fu, Zhang, and Wang]{ViDSOD-100-LinZSFZW24}
Junhao Lin, Lei Zhu, Jiaxing Shen, Huazhu Fu, Qing Zhang, and Liansheng Wang.
\newblock Vidsod-100: {A} new dataset and a baseline model for {RGB-D} video salient object detection.
\newblock \emph{Int. J. Comput. Vis.}, 132\penalty0 (11):\penalty0 5173--5191, 2024{\natexlab{b}}.

\bibitem[Cho et~al.(2024)Cho, Lee, Lee, and Lee]{cho2024transforming}
Suhwan Cho, Minhyeok Lee, Jungho Lee, and Sangyoun Lee.
\newblock Transforming static images using generative models for video salient object detection.
\newblock \emph{arXiv preprint arXiv:2411.13975}, 2024.

\bibitem[Lamdouar et~al.(2020)Lamdouar, Yang, Xie, and Zisserman]{LamdouarYXZ20}
Hala Lamdouar, Charig Yang, Weidi Xie, and Andrew Zisserman.
\newblock Betrayed by motion: Camouflaged object discovery via motion segmentation.
\newblock In \emph{Computer Vision - {ACCV} 2020 - 15th Asian Conference on Computer Vision, Kyoto, Japan, November 30 - December 4, 2020, Revised Selected Papers, Part {II}}, pages 488--503, 2020.

\bibitem[Meeran et~al.(2024)Meeran, T, and Mantha]{SAM-PM-MeeranTM22}
Muhammad~Nawfal Meeran, Gokul~Adethya T, and Bhanu~Pratyush Mantha.
\newblock {SAM-PM:} enhancing video camouflaged object detection using spatio-temporal attention.
\newblock In \emph{{IEEE/CVF} Conference on Computer Vision and Pattern Recognition, {CVPR} 2024 - Workshops, Seattle, WA, USA, June 17-18, 2024}, pages 1857--1866, 2024.

\bibitem[Soomro et~al.(2012{\natexlab{a}})Soomro, Zamir, and Shah]{UCF101-abs-1212-0402}
Khurram Soomro, Amir~Roshan Zamir, and Mubarak Shah.
\newblock {UCF101:} {A} dataset of 101 human actions classes from videos in the wild.
\newblock \emph{CoRR}, abs/1212.0402, 2012{\natexlab{a}}.

\bibitem[Kuehne et~al.(2011)Kuehne, Jhuang, Garrote, Poggio, and Serre]{HMDB-KuehneJGPS11}
Hildegard Kuehne, Hueihan Jhuang, Est{\'{\i}}baliz Garrote, Tomaso~A. Poggio, and Thomas Serre.
\newblock {HMDB:} {A} large video database for human motion recognition.
\newblock In \emph{{IEEE} International Conference on Computer Vision, {ICCV} 2011, Barcelona, Spain, November 6-13, 2011}, pages 2556--2563, 2011.

\bibitem[Carreira et~al.(2018)Carreira, Noland, Banki{-}Horvath, Hillier, and Zisserman]{Kinetics-600-abs-1808-01340}
Jo{\~{a}}o Carreira, Eric Noland, Andras Banki{-}Horvath, Chloe Hillier, and Andrew Zisserman.
\newblock A short note about kinetics-600.
\newblock \emph{CoRR}, abs/1808.01340, 2018.

\bibitem[KunChang~Li and Qiao(2023)]{li2023videochat}
Yi~Wang Yizhuo Li Wenhai Wang Ping Luo Yali Wang Limin~Wang KunChang~Li, Yinan~He and Yu~Qiao.
\newblock Videochat: Chat-centric video understanding.
\newblock \emph{arXiv preprint arXiv:2305.06355}, 2023.

\bibitem[Li et~al.(2020)Li, Rodriguez, Yu, and Li]{li2020word}
Dongxu Li, Cristian Rodriguez, Xin Yu, and Hongdong Li.
\newblock Word-level deep sign language recognition from video: A new large-scale dataset and methods comparison.
\newblock In \emph{The IEEE Winter Conference on Applications of Computer Vision}, pages 1459--1469, 2020.

\bibitem[Zuo et~al.(2023)Zuo, Wei, and Mak]{NLA-SLR-ZuoWM23}
Ronglai Zuo, Fangyun Wei, and Brian Mak.
\newblock Natural language-assisted sign language recognition.
\newblock In \emph{{IEEE/CVF} Conference on Computer Vision and Pattern Recognition, {CVPR} 2023, Vancouver, BC, Canada, June 17-24, 2023}, pages 14890--14900. {IEEE}, 2023.

\bibitem[Shu et~al.(2024)Shu, Zhang, Liu, Qin, Zhou, Huang, and Zhao]{Video-XL-abs-2409-14485}
Yan Shu, Peitian Zhang, Zheng Liu, Minghao Qin, Junjie Zhou, Tie{-}Jun Huang, and Bo~Zhao.
\newblock Video-xl: Extra-long vision language model for hour-scale video understanding.
\newblock \emph{CoRR}, abs/2409.14485, 2024.

\bibitem[Zhou et~al.(2024{\natexlab{c}})Zhou, Shu, Zhao, Wu, Xiao, Yang, Xiong, Zhang, Huang, and Liu]{MLVU-zhou}
Junjie Zhou, Yan Shu, Bo~Zhao, Boya Wu, Shitao Xiao, Xi~Yang, Yongping Xiong, Bo~Zhang, Tiejun Huang, and Zheng Liu.
\newblock Mlvu: A comprehensive benchmark for multi-task long video understanding.
\newblock \emph{arXiv preprint arXiv:2406.04264}, 2024{\natexlab{c}}.

\bibitem[Miao et~al.(2022)Miao, Wang, Wu, Li, Zhang, Wei, and Yang]{MiaoWWLZWY22}
Jiaxu Miao, Xiaohan Wang, Yu~Wu, Wei Li, Xu~Zhang, Yunchao Wei, and Yi~Yang.
\newblock Large-scale video panoptic segmentation in the wild: {A} benchmark.
\newblock In \emph{{IEEE/CVF} Conference on Computer Vision and Pattern Recognition, {CVPR} 2022, New Orleans, LA, USA, June 18-24, 2022}, pages 21001--21011, 2022.

\bibitem[Ravi et~al.(2024)Ravi, Gabeur, Hu, Hu, Ryali, Ma, Khedr, R{\"{a}}dle, Rolland, Gustafson, Mintun, Pan, Alwala, Carion, Wu, Girshick, Doll{\'{a}}r, and Feichtenhofer]{SAM2-abs-2408-00714}
Nikhila Ravi, Valentin Gabeur, Yuan{-}Ting Hu, Ronghang Hu, Chaitanya Ryali, Tengyu Ma, Haitham Khedr, Roman R{\"{a}}dle, Chlo{\'{e}} Rolland, Laura Gustafson, Eric Mintun, Junting Pan, Kalyan~Vasudev Alwala, Nicolas Carion, Chao{-}Yuan Wu, Ross~B. Girshick, Piotr Doll{\'{a}}r, and Christoph Feichtenhofer.
\newblock {SAM} 2: Segment anything in images and videos.
\newblock \emph{CoRR}, abs/2408.00714, 2024.

\bibitem[Xu et~al.(2018)Xu, Yang, Fan, Yue, Liang, Yang, and Huang]{YouTube-VOS-abs-1809-03327}
Ning Xu, Linjie Yang, Yuchen Fan, Dingcheng Yue, Yuchen Liang, Jianchao Yang, and Thomas~S. Huang.
\newblock Youtube-vos: {A} large-scale video object segmentation benchmark.
\newblock \emph{CoRR}, abs/1809.03327, 2018.

\bibitem[Seo et~al.(2020)Seo, Lee, and Han]{URVOS-SeoLH20}
Seonguk Seo, Joon{-}Young Lee, and Bohyung Han.
\newblock {URVOS:} unified referring video object segmentation network with a large-scale benchmark.
\newblock In \emph{Computer Vision - {ECCV} 2020 - 16th European Conference, Glasgow, UK, August 23-28, 2020, Proceedings, Part {XV}}, pages 208--223, 2020.

\bibitem[Wu et~al.(2023{\natexlab{d}})Wu, Jiang, Yan, Lu, Yuan, and Luo]{UniRef-abs-2312-15715}
Jiannan Wu, Yi~Jiang, Bin Yan, Huchuan Lu, Zehuan Yuan, and Ping Luo.
\newblock Uniref++: Segment every reference object in spatial and temporal spaces.
\newblock \emph{CoRR}, abs/2312.15715, 2023{\natexlab{d}}.

\bibitem[Yan et~al.(2024{\natexlab{b}})Yan, Wang, Yan, Jiang, Hu, Kang, Xie, and Gavves]{yan2024visa}
Cilin Yan, Haochen Wang, Shilin Yan, Xiaolong Jiang, Yao Hu, Guoliang Kang, Weidi Xie, and Efstratios Gavves.
\newblock Visa: Reasoning video object segmentation via large language models.
\newblock \emph{arXiv preprint arXiv:2407.11325}, 2024{\natexlab{b}}.

\bibitem[Tang et~al.(2022)Tang, Liao, Liu, Li, Jin, Jiang, Yu, and Xu]{HC-STVG2-TangLLLJJYX22}
Zongheng Tang, Yue Liao, Si~Liu, Guanbin Li, Xiaojie Jin, Hongxu Jiang, Qian Yu, and Dong Xu.
\newblock Human-centric spatio-temporal video grounding with visual transformers.
\newblock \emph{{IEEE} Trans. Circuits Syst. Video Technol.}, 32\penalty0 (12):\penalty0 8238--8249, 2022.

\bibitem[Yang et~al.(2022{\natexlab{b}})Yang, Miech, Sivic, Laptev, and Schmid]{TubeDETR-YangMSLS22}
Antoine Yang, Antoine Miech, Josef Sivic, Ivan Laptev, and Cordelia Schmid.
\newblock Tubedetr: Spatio-temporal video grounding with transformers.
\newblock In \emph{{IEEE/CVF} Conference on Computer Vision and Pattern Recognition, {CVPR} 2022, New Orleans, LA, USA, June 18-24, 2022}, pages 16421--16432, 2022{\natexlab{b}}.

\bibitem[Zhang et~al.(2020)Zhang, Zhao, Zhao, Wang, Liu, and Gao]{VidSTG-ZhangZZWLG20}
Zhu Zhang, Zhou Zhao, Yang Zhao, Qi~Wang, Huasheng Liu, and Lianli Gao.
\newblock Where does it exist: Spatio-temporal video grounding for multi-form sentences.
\newblock In \emph{2020 {IEEE/CVF} Conference on Computer Vision and Pattern Recognition, {CVPR} 2020, Seattle, WA, USA, June 13-19, 2020}, pages 10665--10674, 2020.

\bibitem[Butler et~al.(2012)Butler, Wulff, Stanley, and Black]{ButlerWSB12}
Daniel~J. Butler, Jonas Wulff, Garrett~B. Stanley, and Michael~J. Black.
\newblock A naturalistic open source movie for optical flow evaluation.
\newblock In \emph{Computer Vision - {ECCV} 2012 - 12th European Conference on Computer Vision, Florence, Italy, October 7-13, 2012, Proceedings, Part {VI}}, pages 611--625, 2012.

\bibitem[Hu et~al.(2024{\natexlab{d}})Hu, Gao, Li, Zhao, Cun, Zhang, Quan, and Shan]{DepthCrafter-abs-2409-02095}
Wenbo Hu, Xiangjun Gao, Xiaoyu Li, Sijie Zhao, Xiaodong Cun, Yong Zhang, Long Quan, and Ying Shan.
\newblock Depthcrafter: Generating consistent long depth sequences for open-world videos.
\newblock \emph{CoRR}, abs/2409.02095, 2024{\natexlab{d}}.

\bibitem[Dai et~al.(2017)Dai, Chang, Savva, Halber, Funkhouser, and Nie{\ss}ner]{DaiCSHFN17}
Angela Dai, Angel~X. Chang, Manolis Savva, Maciej Halber, Thomas~A. Funkhouser, and Matthias Nie{\ss}ner.
\newblock Scannet: Richly-annotated 3d reconstructions of indoor scenes.
\newblock In \emph{2017 {IEEE} Conference on Computer Vision and Pattern Recognition, {CVPR} 2017, Honolulu, HI, USA, July 21-26, 2017}, pages 2432--2443, 2017.

\bibitem[Palazzolo et~al.(2019)Palazzolo, Behley, Lottes, Gigu{\`{e}}re, and Stachniss]{PalazzoloBLGS19}
Emanuele Palazzolo, Jens Behley, Philipp Lottes, Philippe Gigu{\`{e}}re, and Cyrill Stachniss.
\newblock Refusion: 3d reconstruction in dynamic environments for {RGB-D} cameras exploiting residuals.
\newblock In \emph{2019 {IEEE/RSJ} International Conference on Intelligent Robots and Systems, {IROS} 2019, Macau, SAR, China, November 3-8, 2019}, pages 7855--7862, 2019.

\bibitem[Geiger et~al.(2012)Geiger, Lenz, and Urtasun]{GeigerLU12}
Andreas Geiger, Philip Lenz, and Raquel Urtasun.
\newblock Are we ready for autonomous driving? the {KITTI} vision benchmark suite.
\newblock In \emph{2012 {IEEE} Conference on Computer Vision and Pattern Recognition, Providence, RI, USA, June 16-21, 2012}, pages 3354--3361, 2012.

\bibitem[MMV()]{MMVMBench}
Mmvmbench.
\newblock URL \url{https://huggingface.co/zhouyik/MMVMBench}.

\bibitem[Kristan et~al.(2018)Kristan, Leonardis, Matas, Felsberg, Pflugfelder, Zajc, Voj{\'{\i}}r, Bhat, Lukezic, Eldesokey, Fern{\'{a}}ndez, Garc{\'{\i}}a{-}Mart{\'{\i}}n, Iglesias{-}Arias, Alatan, Gonz{\'{a}}lez{-}Garc{\'{\i}}a, Petrosino, Memarmoghadam, Vedaldi, Muhic, He, Smeulders, Perera, Li, Chen, Kim, Xu, Xiong, Tian, Luo, Sun, Hao, Kim, Mishra, Chen, Wang, Wee, Gavves, Gundogdu, Velasco{-}Salido, Khan, Yang, Zhao, Li, Battistone, Ath, Subrahmanyam, Bastos, Ling, Galoogahi, Lee, Li, Zhao, Fan, Zhang, Possegger, Li, Lu, Zhi, Li, Lee, Chang, Drummond, Valmadre, Martin, Chahl, Choi, Li, Wang, Qi, Sung, Johnander, Henriques, Choi, van~de Weijer, Herranz, Mart{\'{\i}}nez, Kittler, Zhuang, Gao, Grm, Zhang, Wang, Yang, Rout, Si, Bertinetto, Chu, Che, Maresca, Danelljan, Yang, Abdelpakey, Shehata, Kang, Lee, Wang, Miksik, Moallem, Vicente{-}Mo{\~{n}}ivar, Senna, Li, Torr, Raju, Qian, Wang, Zhou, Guo, Nieto, Gorthi, Tao, Bowden, Everson, Wang, Yun, Choi, Vivas, Bai, Huang, Wu, Hadfield, Wang, Golodetz, Tang,
  Xu, Zhang, Fischer, Santopietro, Struc, Wang, Zuo, Feng, Wu, Zou, Hu, Zhou, Zeng, Zhang, Wu, Wu, Tian, Li, Lu, Law, Wu, Demiris, Yang, Jiao, Li, Zhang, Sun, Zhang, Zhu, Feng, Wang, and He]{VOT2018-KristanLMFPZVBL18}
Matej Kristan, Ales Leonardis, Jiri Matas, Michael Felsberg, Roman~P. Pflugfelder, Luka~Cehovin Zajc, Tom{\'{a}}s Voj{\'{\i}}r, Goutam Bhat, Alan Lukezic, Abdelrahman Eldesokey, Gustavo Fern{\'{a}}ndez, {\'{A}}lvaro Garc{\'{\i}}a{-}Mart{\'{\i}}n, {\'{A}}lvaro Iglesias{-}Arias, A.~Aydin Alatan, Abel Gonz{\'{a}}lez{-}Garc{\'{\i}}a, Alfredo Petrosino, Alireza Memarmoghadam, Andrea Vedaldi, Andrej Muhic, Anfeng He, Arnold W.~M. Smeulders, Asanka~G. Perera, Bo~Li, Boyu Chen, Changick Kim, Changsheng Xu, Changzhen Xiong, Cheng Tian, Chong Luo, Chong Sun, Cong Hao, Daijin Kim, Deepak Mishra, Deming Chen, Dong Wang, Dongyoon Wee, Efstratios Gavves, Erhan Gundogdu, Erik Velasco{-}Salido, Fahad~Shahbaz Khan, Fan Yang, Fei Zhao, Feng Li, Francesco Battistone, George~De Ath, Gorthi R. K.~Sai Subrahmanyam, Guilherme~Sousa Bastos, Haibin Ling, Hamed~Kiani Galoogahi, Hankyeol Lee, Haojie Li, Haojie Zhao, Heng Fan, Honggang Zhang, Horst Possegger, Houqiang Li, Huchuan Lu, Hui Zhi, Huiyun Li, Hyemin Lee, Hyung~Jin Chang,
  Isabela Drummond, Jack Valmadre, Jaime~Spencer Martin, Javaan~Singh Chahl, Jin~Young Choi, Jing Li, Jinqiao Wang, Jinqing Qi, Jinyoung Sung, Joakim Johnander, Jo{\~{a}}o~F. Henriques, Jongwon Choi, Joost van~de Weijer, Jorge~Rodr{\'{\i}}guez Herranz, Jos{\'{e}}~M. Mart{\'{\i}}nez, Josef Kittler, Junfei Zhuang, Junyu Gao, Klemen Grm, Lichao Zhang, Lijun Wang, Lingxiao Yang, Litu Rout, Liu Si, Luca Bertinetto, Lutao Chu, Manqiang Che, Mario~Edoardo Maresca, Martin Danelljan, Ming{-}Hsuan Yang, Mohamed~H. Abdelpakey, Mohamed~S. Shehata, Myunggu Kang, Namhoon Lee, Ning Wang, Ondrej Miksik, Payman Moallem, Pablo Vicente{-}Mo{\~{n}}ivar, Pedro Senna, Peixia Li, Philip H.~S. Torr, Priya~Mariam Raju, Ruihe Qian, Qiang Wang, Qin Zhou, Qing Guo, Rafael~Martin Nieto, Rama Krishna Sai~Subrahmanyam Gorthi, Ran Tao, Richard Bowden, Richard~M. Everson, Runling Wang, Sangdoo Yun, Seokeon Choi, Sergio Vivas, Shuai Bai, Shuangping Huang, Sihang Wu, Simon Hadfield, Siwen Wang, Stuart Golodetz, Ming Tang, Tianyang Xu, Tianzhu
  Zhang, Tobias Fischer, Vincenzo Santopietro, Vitomir Struc, Wei Wang, Wangmeng Zuo, Wei Feng, Wei Wu, Wei Zou, Weiming Hu, Wengang Zhou, Wenjun Zeng, Xiaofan Zhang, Xiaohe Wu, Xiao{-}Jun Wu, Xinmei Tian, Yan Li, Yan Lu, Yee~Wei Law, Yi~Wu, Yiannis Demiris, Yicai Yang, Yifan Jiao, Yuhong Li, Yunhua Zhang, Yuxuan Sun, Zheng Zhang, Zheng Zhu, Zhenhua Feng, Zhihui Wang, and Zhiqun He.
\newblock The sixth visual object tracking {VOT2018} challenge results.
\newblock In \emph{Computer Vision - {ECCV} 2018 Workshops - Munich, Germany, September 8-14, 2018, Proceedings, Part {I}}, pages 3--53, 2018.

\bibitem[Yan et~al.(2023)Yan, Jiang, Wu, Wang, Luo, Yuan, and Lu]{UniNext-0004JWWLYL23}
Bin Yan, Yi~Jiang, Jiannan Wu, Dong Wang, Ping Luo, Zehuan Yuan, and Huchuan Lu.
\newblock Universal instance perception as object discovery and retrieval.
\newblock In \emph{{IEEE/CVF} Conference on Computer Vision and Pattern Recognition, {CVPR} 2023, Vancouver, BC, Canada, June 17-24, 2023}, pages 15325--15336, 2023.

\bibitem[Fan et~al.(2019{\natexlab{a}})Fan, Lin, Yang, Chu, Deng, Yu, Bai, Xu, Liao, and Ling]{LaSOT-FanLYCDYBXLL19}
Heng Fan, Liting Lin, Fan Yang, Peng Chu, Ge~Deng, Sijia Yu, Hexin Bai, Yong Xu, Chunyuan Liao, and Haibin Ling.
\newblock Lasot: {A} high-quality benchmark for large-scale single object tracking.
\newblock In \emph{{IEEE} Conference on Computer Vision and Pattern Recognition, {CVPR} 2019, Long Beach, CA, USA, June 16-20, 2019}, pages 5374--5383, 2019{\natexlab{a}}.

\bibitem[Sun et~al.(2022)Sun, Cao, Jiang, Yuan, Bai, Kitani, and Luo]{DanceTrack-SunCJYBKL22}
Peize Sun, Jinkun Cao, Yi~Jiang, Zehuan Yuan, Song Bai, Kris Kitani, and Ping Luo.
\newblock Dancetrack: Multi-object tracking in uniform appearance and diverse motion.
\newblock In \emph{{IEEE/CVF} Conference on Computer Vision and Pattern Recognition, {CVPR} 2022, New Orleans, LA, USA, June 18-24, 2022}, pages 20961--20970, 2022.

\bibitem[Mueller et~al.(2016)Mueller, Smith, and Ghanem]{UAV123-MuellerSG16}
Matthias Mueller, Neil Smith, and Bernard Ghanem.
\newblock A benchmark and simulator for {UAV} tracking.
\newblock In \emph{Computer Vision - {ECCV} 2016 - 14th European Conference, Amsterdam, The Netherlands, October 11-14, 2016, Proceedings, Part {I}}, volume 9905 of \emph{Lecture Notes in Computer Science}, pages 445--461, 2016.

\bibitem[Alawode et~al.(2022)Alawode, Guo, Ummar, Werghi, Dias, Mian, and Javed]{UTB180-alawode2022utb180}
Basit Alawode, Yuhang Guo, Mehnaz Ummar, Naoufel Werghi, Jorge Dias, Ajmal Mian, and Sajid Javed.
\newblock Utb180: A high-quality benchmark for underwater tracking.
\newblock In \emph{{ACCV}}, 2022.

\bibitem[Dosovitskiy et~al.(2015)Dosovitskiy, Fischer, Ilg, H{\"{a}}usser, Hazirbas, Golkov, van~der Smagt, Cremers, and Brox]{DosovitskiyFIHH15}
Alexey Dosovitskiy, Philipp Fischer, Eddy Ilg, Philip H{\"{a}}usser, Caner Hazirbas, Vladimir Golkov, Patrick van~der Smagt, Daniel Cremers, and Thomas Brox.
\newblock Flownet: Learning optical flow with convolutional networks.
\newblock In \emph{2015 {IEEE} International Conference on Computer Vision, {ICCV} 2015, Santiago, Chile, December 7-13, 2015}, pages 2758--2766, 2015.

\bibitem[de~Armas(2019)]{deArmas2019videoflow}
Jadiel de~Armas.
\newblock Videoflow.
\newblock \url{https://github.com/videoflow/videoflow}, 2019.

\bibitem[Ilg et~al.(2017)Ilg, Mayer, Saikia, Keuper, Dosovitskiy, and Brox]{IMKDB17}
E.~Ilg, N.~Mayer, T.~Saikia, M.~Keuper, A.~Dosovitskiy, and T.~Brox.
\newblock Flownet 2.0: Evolution of optical flow estimation with deep networks.
\newblock In \emph{IEEE Conference on Computer Vision and Pattern Recognition (CVPR)}, Jul 2017.

\bibitem[Seidel et~al.(2021)Seidel, Apitzsch, and Hirtz]{seidel2021omniflow}
Roman Seidel, Andr{\'e} Apitzsch, and Gangolf Hirtz.
\newblock Omniflow: Human omnidirectional optical flow.
\newblock In \emph{Proceedings of the IEEE/CVF Conference on Computer Vision and Pattern Recognition}, pages 3678--3681, 2021.

\bibitem[Shi et~al.(2022{\natexlab{a}})Shi, Zhou, Yang, Ye, Yin, Yin, Meng, and Wang]{shi2022panoflow}
Hao Shi, Yifan Zhou, Kailun Yang, Yaozu Ye, Xiaoting Yin, Zhe Yin, Shi Meng, and Kaiwei Wang.
\newblock Panoflow: Learning optical flow for panoramic images.
\newblock \emph{arXiv preprint arXiv:2202.13388}, 2022{\natexlab{a}}.

\bibitem[Castro et~al.(2019)Castro, Hazarika, P{\'{e}}rez{-}Rosas, Zimmermann, Mihalcea, and Poria]{CastroHPZMP19}
Santiago Castro, Devamanyu Hazarika, Ver{\'{o}}nica P{\'{e}}rez{-}Rosas, Roger Zimmermann, Rada Mihalcea, and Soujanya Poria.
\newblock Towards multimodal sarcasm detection (an {\_}obviously{\_} perfect paper).
\newblock In \emph{Proceedings of the 57th Conference of the Association for Computational Linguistics, {ACL} 2019, Florence, Italy, July 28- August 2, 2019, Volume 1: Long Papers}, pages 4619--4629, 2019.

\bibitem[Sadhu et~al.(2021)Sadhu, Gupta, Yatskar, Nevatia, and Kembhavi]{Sadhu_2021_CVPR}
Arka Sadhu, Tanmay Gupta, Mark Yatskar, Ram Nevatia, and Aniruddha Kembhavi.
\newblock Visual semantic role labeling for video understanding.
\newblock In \emph{The IEEE Conference on Computer Vision and Pattern Recognition (CVPR)}, June 2021.

\bibitem[vid()]{video-diff-zeroscope}
zeroscope.
\newblock URL \url{https://huggingface.co/cerspense/zeroscope_v2_576w}.

\bibitem[Wang et~al.(2023{\natexlab{d}})Wang, Chen, Ma, Zhou, Huang, Wang, Yang, He, Yu, Yang, Guo, Wu, Si, Jiang, Chen, Loy, Dai, Lin, Qiao, and Liu]{LAVIE-abs-2309-15103}
Yaohui Wang, Xinyuan Chen, Xin Ma, Shangchen Zhou, Ziqi Huang, Yi~Wang, Ceyuan Yang, Yinan He, Jiashuo Yu, Peiqing Yang, Yuwei Guo, Tianxing Wu, Chenyang Si, Yuming Jiang, Cunjian Chen, Chen~Change Loy, Bo~Dai, Dahua Lin, Yu~Qiao, and Ziwei Liu.
\newblock {LAVIE:} high-quality video generation with cascaded latent diffusion models.
\newblock \emph{CoRR}, abs/2309.15103, 2023{\natexlab{d}}.

\bibitem[Yang et~al.(2024{\natexlab{c}})Yang, Teng, Zheng, Ding, Huang, Xu, Yang, Hong, Zhang, Feng, Yin, Gu, Zhang, Wang, Cheng, Liu, Xu, Dong, and Tang]{CogVideoX-abs-2408-06072}
Zhuoyi Yang, Jiayan Teng, Wendi Zheng, Ming Ding, Shiyu Huang, Jiazheng Xu, Yuanming Yang, Wenyi Hong, Xiaohan Zhang, Guanyu Feng, Da~Yin, Xiaotao Gu, Yuxuan Zhang, Weihan Wang, Yean Cheng, Ting Liu, Bin Xu, Yuxiao Dong, and Jie Tang.
\newblock Cogvideox: Text-to-video diffusion models with an expert transformer.
\newblock \emph{CoRR}, abs/2408.06072, 2024{\natexlab{c}}.

\bibitem[Lai()]{Laion-aesthetics}
Laion-aesthetics.
\newblock URL \url{https://laion.ai/blog/laion-aesthetics/}.

\bibitem[Liu et~al.(2023{\natexlab{e}})Liu, Xia, Zhang, Chen, Xing, Wang, Yang, and Shan]{StyleCrafter-abs-2312-00330}
Gongye Liu, Menghan Xia, Yong Zhang, Haoxin Chen, Jinbo Xing, Xintao Wang, Yujiu Yang, and Ying Shan.
\newblock Stylecrafter: Enhancing stylized text-to-video generation with style adapter.
\newblock \emph{CoRR}, abs/2312.00330, 2023{\natexlab{e}}.

\bibitem[Bain et~al.(2021)Bain, Nagrani, Varol, and Zisserman]{Bain21}
Max Bain, Arsha Nagrani, G{\"u}l Varol, and Andrew Zisserman.
\newblock Frozen in time: A joint video and image encoder for end-to-end retrieval.
\newblock In \emph{IEEE International Conference on Computer Vision}, 2021.

\bibitem[Chen et~al.(2024{\natexlab{f}})Chen, Zhang, Cun, Xia, Wang, Weng, and Shan]{chen2024videocrafter2}
Haoxin Chen, Yong Zhang, Xiaodong Cun, Menghan Xia, Xintao Wang, Chao Weng, and Ying Shan.
\newblock Videocrafter2: Overcoming data limitations for high-quality video diffusion models, 2024{\natexlab{f}}.

\bibitem[Smaira et~al.(2020)Smaira, Carreira, Noland, Clancy, Wu, and Zisserman]{Kinetics-700-abs-2010-10864}
Lucas Smaira, Jo{\~{a}}o Carreira, Eric Noland, Ellen Clancy, Amy Wu, and Andrew Zisserman.
\newblock A short note on the kinetics-700-2020 human action dataset.
\newblock \emph{CoRR}, abs/2010.10864, 2020.

\bibitem[Soomro et~al.(2012{\natexlab{b}})Soomro, Zamir, and Shah]{soomro2012ucf101}
Khurram Soomro, Amir~Roshan Zamir, and Mubarak Shah.
\newblock Ucf101: A dataset of 101 human actions classes from videos in the wild.
\newblock \emph{arXiv preprint arXiv:1212.0402}, 2012{\natexlab{b}}.

\bibitem[Dai et~al.(2022{\natexlab{b}})Dai, Yu, Ma, Zhang, Li, Li, Shen, and Qi]{dai2022video}
Peng Dai, Xin Yu, Lan Ma, Baoheng Zhang, Jia Li, Wenbo Li, Jiajun Shen, and Xiaojuan Qi.
\newblock Video demoireing with relation-based temporal consistency.
\newblock In \emph{Proceedings of the IEEE/CVF Conference on Computer Vision and Pattern Recognition}, 2022{\natexlab{b}}.

\bibitem[Yue et~al.(2023)Yue, Cheng, Liu, and Yang]{yue2023recaptured}
Huanjing Yue, Yijia Cheng, Xin Liu, and Jingyu Yang.
\newblock Recaptured raw screen image and video demoir$\backslash$'eing via channel and spatial modulations.
\newblock \emph{arXiv preprint arXiv:2310.20332}, 2023.

\bibitem[Xu et~al.(2024)Xu, Song, Chen, and Zhou]{XuSCZ24}
Shuning Xu, Binbin Song, Xiangyu Chen, and Jiantao Zhou.
\newblock Direction-aware video demoir{\'{e}}ing with temporal-guided bilateral learning.
\newblock In \emph{Thirty-Eighth {AAAI} Conference on Artificial Intelligence, {AAAI} 2024, Thirty-Sixth Conference on Innovative Applications of Artificial Intelligence, {IAAI} 2024, Fourteenth Symposium on Educational Advances in Artificial Intelligence, {EAAI} 2014, February 20-27, 2024, Vancouver, Canada}, pages 6360--6368, 2024.

\bibitem[Yue et~al.(2020)Yue, Cao, Liao, Chu, and Yang]{yue2020supervised}
Huanjing Yue, Cong Cao, Lei Liao, Ronghe Chu, and Jingyu Yang.
\newblock Supervised raw video denoising with a benchmark dataset on dynamic scenes.
\newblock In \emph{IEEE Conference on Computer Vision and Pattern Recognition}, 2020.

\bibitem[Pont-Tuset et~al.(2017)Pont-Tuset, Perazzi, Caelles, Arbel\'aez, Sorkine-Hornung, and {Van Gool}]{Pont-Tuset_arXiv_2017}
Jordi Pont-Tuset, Federico Perazzi, Sergi Caelles, Pablo Arbel\'aez, Alexander Sorkine-Hornung, and Luc {Van Gool}.
\newblock The 2017 davis challenge on video object segmentation.
\newblock \emph{arXiv:1704.00675}, 2017.

\bibitem[Qi et~al.(2022)Qi, Chen, Yang, and Chen]{qi2022BSVD}
Chenyang Qi, Junming Chen, Xin Yang, and Qifeng Chen.
\newblock Real-time streaming video denoising with bidirectional buffers.
\newblock In \emph{ACM MM}, 2022.

\bibitem[Xue et~al.(2019)Xue, Chen, Wu, Wei, and Freeman]{XueCWWF19}
Tianfan Xue, Baian Chen, Jiajun Wu, Donglai Wei, and William~T. Freeman.
\newblock Video enhancement with task-oriented flow.
\newblock \emph{Int. J. Comput. Vis.}, 127\penalty0 (8):\penalty0 1106--1125, 2019.

\bibitem[Zhang et~al.(2023{\natexlab{k}})Zhang, Zhu, Wang, Chen, Wu, and Wang]{zhang2023extracting}
Guozhen Zhang, Yuhan Zhu, Haonan Wang, Youxin Chen, Gangshan Wu, and Limin Wang.
\newblock Extracting motion and appearance via inter-frame attention for efficient video frame interpolation.
\newblock In \emph{Proceedings of the IEEE/CVF Conference on Computer Vision and Pattern Recognition}, pages 5682--5692, 2023{\natexlab{k}}.

\bibitem[MSU()]{MSU-Video-Colorization-Benchmark}
Msu video colorization benchmark.
\newblock URL \url{https://videoprocessing.ai/benchmarks/video-colorization.html}.

\bibitem[Hofinger et~al.(2022)Hofinger, Kobler, Effland, and Pock]{HofingerKobler22}
Markus Hofinger, Erich Kobler, Alexander Effland, and Thomas Pock.
\newblock Learned variational video color propagation.
\newblock In \emph{Computer Vision -- {ECCV} 2022}, Lecture Notes in Computer Science, 2022.

\bibitem[Rea({\natexlab{b}})]{Real-Haze-Video-Database}
Real haze video database, {\natexlab{b}}.
\newblock URL \url{https://qualinet.github.io/databases/video/real-haze-video-database/}.

\bibitem[Xu et~al.(2023{\natexlab{c}})Xu, Hu, Zhu, Dou, Dai, Qiao, and Heng]{xu2023map}
Jiaqi Xu, Xiaowei Hu, Lei Zhu, Qi~Dou, Jifeng Dai, Yu~Qiao, and Pheng-Ann Heng.
\newblock Video dehazing via a multi-range temporal alignment network with physical prior.
\newblock In \emph{Proceedings of the IEEE/CVF Conference on Computer Vision and Pattern Recognition (CVPR)}, 2023{\natexlab{c}}.

\bibitem[Wu et~al.(2024{\natexlab{c}})Wu, Yang, Avil{\'{e}}s{-}Rivero, Ren, Chen, Chen, and Zhu]{WuYARCCZ24}
Hongtao Wu, Yijun Yang, Angelica~I. Avil{\'{e}}s{-}Rivero, Jingjing Ren, Sixiang Chen, Haoyu Chen, and Lei Zhu.
\newblock Semi-supervised video desnowing network via temporal decoupling experts and distribution-driven contrastive regularization.
\newblock In \emph{Computer Vision - {ECCV} 2024 - 18th European Conference, Milan, Italy, September 29-October 4, 2024, Proceedings, Part {X}}, pages 70--89, 2024{\natexlab{c}}.

\bibitem[Chen et~al.(2023)Chen, Ren, Gu, Wu, Lu, Cai, and Zhu]{ChenRGWLCZ23}
Haoyu Chen, Jingjing Ren, Jinjin Gu, Hongtao Wu, Xuequan Lu, Haoming Cai, and Lei Zhu.
\newblock Snow removal in video: {A} new dataset and {A} novel method.
\newblock In \emph{{IEEE/CVF} International Conference on Computer Vision, {ICCV} 2023, Paris, France, October 1-6, 2023}, pages 13165--13176, 2023.

\bibitem[Ghasemabadi et~al.()Ghasemabadi, Janjua, Salameh, and Niu]{ghasemabadilearning}
Amirhosein Ghasemabadi, Muhammad~Kamran Janjua, Mohammad Salameh, and Di~Niu.
\newblock Learning truncated causal history model for video restoration.
\newblock In \emph{The Thirty-eighth Annual Conference on Neural Information Processing Systems}.

\bibitem[Wen et~al.(2023)Wen, Wu, and Chen]{wen2023video}
Qiang Wen, Yue Wu, and Qifeng Chen.
\newblock Video waterdrop removal via spatio-temporal fusion in driving scenes.
\newblock In \emph{International Conference on Robotics and Automation (ICRA)}. IEEE, 2023.

\bibitem[Wu et~al.(2023{\natexlab{e}})Wu, Yang, Chen, Ren, and Zhu]{wu2023mask}
Hongtao Wu, Yijun Yang, Haoyu Chen, Jingjing Ren, and Lei Zhu.
\newblock Mask-guided progressive network for joint raindrop and rain streak removal in videos.
\newblock In \emph{Proceedings of the 31st ACM International Conference on Multimedia}, pages 7216--7225, 2023{\natexlab{e}}.

\bibitem[Wu et~al.(2024{\natexlab{d}})Wu, Yang, Xu, Wang, Zhou, and Zhu]{wu2024rainmamba}
Hongtao Wu, Yijun Yang, Huihui Xu, Weiming Wang, Jinni Zhou, and Lei Zhu.
\newblock Rainmamba: Enhanced locality learning with state space models for video deraining.
\newblock \emph{arXiv preprint arXiv:2407.21773}, 2024{\natexlab{d}}.

\bibitem[Chan et~al.(2022)Chan, Zhou, Xu, and Loy]{chan2022investigating}
Kelvin~C.K. Chan, Shangchen Zhou, Xiangyu Xu, and Chen~Change Loy.
\newblock Investigating tradeoffs in real-world video super-resolution.
\newblock In \emph{IEEE Conference on Computer Vision and Pattern Recognition}, 2022.

\bibitem[Zhou et~al.(2024{\natexlab{d}})Zhou, Yang, Wang, Luo, and Loy]{zhou2024upscaleavideo}
Shangchen Zhou, Peiqing Yang, Jianyi Wang, Yihang Luo, and Chen~Change Loy.
\newblock {Upscale-A-Video}: Temporal-consistent diffusion model for real-world video super-resolution.
\newblock In \emph{CVPR}, 2024{\natexlab{d}}.

\bibitem[Zhou et~al.(2023)Zhou, Li, Chan, and Loy]{zhou2023propainter}
Shangchen Zhou, Chongyi Li, Kelvin~C.K Chan, and Chen~Change Loy.
\newblock {ProPainter}: Improving propagation and transformer for video inpainting.
\newblock In \emph{Proceedings of IEEE International Conference on Computer Vision (ICCV)}, 2023.

\bibitem[Nah et~al.(2019)Nah, Baik, Hong, Moon, Son, Timofte, and Lee]{Nah_2019_CVPR_Workshops_REDS}
Seungjun Nah, Sungyong Baik, Seokil Hong, Gyeongsik Moon, Sanghyun Son, Radu Timofte, and Kyoung~Mu Lee.
\newblock Ntire 2019 challenge on video deblurring and super-resolution: Dataset and study.
\newblock In \emph{CVPR Workshops}, June 2019.

\bibitem[Liang et~al.(2022)Liang, Cao, Fan, Zhang, Ranjan, Li, Timofte, and Van~Gool]{liang2022vrt}
Jingyun Liang, Jiezhang Cao, Yuchen Fan, Kai Zhang, Rakesh Ranjan, Yawei Li, Radu Timofte, and Luc Van~Gool.
\newblock Vrt: A video restoration transformer.
\newblock \emph{arXiv preprint arXiv:2201.12288}, 2022.

\bibitem[Yang et~al.(2024{\natexlab{d}})Yang, Zhou, Liu, , and Loy]{yang2024fresco}
Shuai Yang, Yifan Zhou, Ziwei Liu, , and Chen~Change Loy.
\newblock Fresco: Spatial-temporal correspondence for zero-shot video translation.
\newblock In \emph{CVPR}, 2024{\natexlab{d}}.

\bibitem[Yang et~al.(2022{\natexlab{c}})Yang, Jiang, Liu, and Loy]{yang2022Vtoonify}
Shuai Yang, Liming Jiang, Ziwei Liu, and Chen~Change Loy.
\newblock Vtoonify: Controllable high-resolution portrait video style transfer.
\newblock \emph{ACM Transactions on Graphics (TOG)}, 41\penalty0 (6):\penalty0 1--15, 2022{\natexlab{c}}.

\bibitem[Weinberger(2015)]{Speech-Accent-Archive}
Steven Weinberger.
\newblock Speech accent archive, 2015.
\newblock URL \url{http://accent.gmu.edu}.

\bibitem[Elizalde et~al.(2023)Elizalde, Deshmukh, Ismail, and Wang]{CLAP-ElizaldeDIW23}
Benjamin Elizalde, Soham Deshmukh, Mahmoud~Al Ismail, and Huaming Wang.
\newblock {CLAP} learning audio concepts from natural language supervision.
\newblock In \emph{{IEEE} International Conference on Acoustics, Speech and Signal Processing {ICASSP} 2023, Rhodes Island, Greece, June 4-10, 2023}, pages 1--5, 2023.

\bibitem[Nagrani et~al.(2017)Nagrani, Chung, and Zisserman]{VoxCeleb-NagraniCZ17}
Arsha Nagrani, Joon~Son Chung, and Andrew Zisserman.
\newblock Voxceleb: {A} large-scale speaker identification dataset.
\newblock In \emph{18th Annual Conference of the International Speech Communication Association, Interspeech 2017, Stockholm, Sweden, August 20-24, 2017}, pages 2616--2620, 2017.

\bibitem[Hsu et~al.(2021)Hsu, Bolte, Tsai, Lakhotia, Salakhutdinov, and Mohamed]{HuBERT-HsuBTLSM21}
Wei{-}Ning Hsu, Benjamin Bolte, Yao{-}Hung~Hubert Tsai, Kushal Lakhotia, Ruslan Salakhutdinov, and Abdelrahman Mohamed.
\newblock Hubert: Self-supervised speech representation learning by masked prediction of hidden units.
\newblock \emph{{IEEE} {ACM} Trans. Audio Speech Lang. Process.}, 29:\penalty0 3451--3460, 2021.

\bibitem[Gong et~al.(2022)Gong, Yu, and Glass]{Vocalsound-GongYG22}
Yuan Gong, Jin Yu, and James~R. Glass.
\newblock Vocalsound: {A} dataset for improving human vocal sounds recognition.
\newblock In \emph{{IEEE} International Conference on Acoustics, Speech and Signal Processing, {ICASSP} 2022, Virtual and Singapore, 23-27 May 2022}, pages 151--155, 2022.

\bibitem[Lugosch et~al.(2019)Lugosch, Ravanelli, Ignoto, Tomar, and Bengio]{Flue-speech-command-LugoschRITB19}
Loren Lugosch, Mirco Ravanelli, Patrick Ignoto, Vikrant~Singh Tomar, and Yoshua Bengio.
\newblock Speech model pre-training for end-to-end spoken language understanding.
\newblock In \emph{20th Annual Conference of the International Speech Communication Association, Interspeech 2019, Graz, Austria, September 15-19, 2019}, pages 814--818, 2019.

\bibitem[Warden(2018)]{Speech-Commands-abs-1804-03209}
Pete Warden.
\newblock Speech commands: {A} dataset for limited-vocabulary speech recognition.
\newblock \emph{CoRR}, abs/1804.03209, 2018.

\bibitem[Wang et~al.(2024{\natexlab{h}})Wang, Zhang, Fei, Zhao, Li, Wu, Ji, and Zhang]{SpeechEE-WangZ0ZLW0Z24}
Bin Wang, Meishan Zhang, Hao Fei, Yu~Zhao, Bobo Li, Shengqiong Wu, Wei Ji, and Min Zhang.
\newblock Speechee: {A} novel benchmark for speech event extraction.
\newblock In \emph{Proceedings of the 32nd {ACM} International Conference on Multimedia, {MM} 2024, Melbourne, VIC, Australia, 28 October 2024 - 1 November 2024}, pages 10449--10458, 2024{\natexlab{h}}.

\bibitem[Busso et~al.(2008)Busso, Bulut, Lee, Kazemzadeh, Mower, Kim, Chang, Lee, and Narayanan]{IEMOCAP-BussoBLKMKCLN08}
Carlos Busso, Murtaza Bulut, Chi{-}Chun Lee, Abe Kazemzadeh, Emily Mower, Samuel Kim, Jeannette~N. Chang, Sungbok Lee, and Shrikanth~S. Narayanan.
\newblock {IEMOCAP:} interactive emotional dyadic motion capture database.
\newblock \emph{Lang. Resour. Evaluation}, 42\penalty0 (4):\penalty0 335--359, 2008.

\bibitem[Chen et~al.(2022{\natexlab{b}})Chen, Wang, Chen, Wu, Liu, Chen, Li, Kanda, Yoshioka, Xiao, Wu, Zhou, Ren, Qian, Qian, Wu, Zeng, Yu, and Wei]{WavLM-ChenWCWLCLKYXWZ22}
Sanyuan Chen, Chengyi Wang, Zhengyang Chen, Yu~Wu, Shujie Liu, Zhuo Chen, Jinyu Li, Naoyuki Kanda, Takuya Yoshioka, Xiong Xiao, Jian Wu, Long Zhou, Shuo Ren, Yanmin Qian, Yao Qian, Jian Wu, Michael Zeng, Xiangzhan Yu, and Furu Wei.
\newblock Wavlm: Large-scale self-supervised pre-training for full stack speech processing.
\newblock \emph{{IEEE} J. Sel. Top. Signal Process.}, 16\penalty0 (6):\penalty0 1505--1518, 2022{\natexlab{b}}.

\bibitem[Mus()]{Music-Genre-GTZAN}
Gtzan dataset - music genre classification.
\newblock URL \url{https://www.kaggle.com/datasets/andradaolteanu/gtzan-dataset-music-genre-classification}.

\bibitem[McCallum et~al.(2022)McCallum, Korzeniowski, Oramas, Gouyon, and Ehmann]{Musicset-Sup-McCallumKOGE22}
Matthew~C. McCallum, Filip Korzeniowski, Sergio Oramas, Fabien Gouyon, and Andreas~F. Ehmann.
\newblock Supervised and unsupervised learning of audio representations for music understanding.
\newblock In \emph{Proceedings of the 23rd International Society for Music Information Retrieval Conference, {ISMIR} 2022, Bengaluru, India, December 4-8, 2022}, pages 256--263, 2022.

\bibitem[Engel et~al.(2017)Engel, Resnick, Roberts, Dieleman, Eck, Simonyan, and Norouzi]{NSynth-nsynth2017}
Jesse Engel, Cinjon Resnick, Adam Roberts, Sander Dieleman, Douglas Eck, Karen Simonyan, and Mohammad Norouzi.
\newblock Neural audio synthesis of musical notes with wavenet autoencoders, 2017.

\bibitem[Wilkins et~al.(2018)Wilkins, Seetharaman, Wahl, and Pardo]{VocalSet-WilkinsSWP18}
Julia Wilkins, Prem Seetharaman, Alison Wahl, and Bryan Pardo.
\newblock Vocalset: {A} singing voice dataset.
\newblock In \emph{Proceedings of the 19th International Society for Music Information Retrieval Conference, {ISMIR} 2018, Paris, France, September 23-27, 2018}, pages 468--474, 2018.

\bibitem[Drossos et~al.(2020)Drossos, Lipping, and Virtanen]{Clotho-cap-DrossosLV20}
Konstantinos Drossos, Samuel Lipping, and Tuomas Virtanen.
\newblock Clotho: an audio captioning dataset.
\newblock In \emph{2020 {IEEE} International Conference on Acoustics, Speech and Signal Processing, {ICASSP} 2020, Barcelona, Spain, May 4-8, 2020}, pages 736--740, 2020.

\bibitem[Kim et~al.(2023{\natexlab{a}})Kim, Sung{-}Bin, and Oh]{PTAAC-KimSO23}
Minkyu Kim, Kim Sung{-}Bin, and Tae{-}Hyun Oh.
\newblock Prefix tuning for automated audio captioning.
\newblock In \emph{{IEEE} International Conference on Acoustics, Speech and Signal Processing {ICASSP} 2023, Rhodes Island, Greece, June 4-10, 2023}, pages 1--5, 2023{\natexlab{a}}.

\bibitem[Kim et~al.(2019)Kim, Kim, Lee, and Kim]{AudioCaps-KimKLK19}
Chris~Dongjoo Kim, Byeongchang Kim, Hyunmin Lee, and Gunhee Kim.
\newblock Audiocaps: Generating captions for audios in the wild.
\newblock In \emph{Proceedings of the 2019 Conference of the North American Chapter of the Association for Computational Linguistics: Human Language Technologies, {NAACL-HLT} 2019, Minneapolis, MN, USA, June 2-7, 2019, Volume 1 (Long and Short Papers)}, pages 119--132, 2019.

\bibitem[Gong et~al.(2024)Gong, Luo, Liu, Karlinsky, and Glass]{LTU-0001LLKG24}
Yuan Gong, Hongyin Luo, Alexander~H. Liu, Leonid Karlinsky, and James~R. Glass.
\newblock Listen, think, and understand.
\newblock In \emph{The Twelfth International Conference on Learning Representations, {ICLR} 2024, Vienna, Austria, May 7-11, 2024}, 2024.

\bibitem[Lipping et~al.(2022)Lipping, Sudarsanam, Drossos, and Virtanen]{Clotho-AQA-LippingSDV22}
Samuel Lipping, Parthasaarathy Sudarsanam, Konstantinos Drossos, and Tuomas Virtanen.
\newblock Clotho-aqa: {A} crowdsourced dataset for audio question answering.
\newblock In \emph{30th European Signal Processing Conference, {EUSIPCO} 2022, Belgrade, Serbia, August 29 - Sept. 2, 2022}, pages 1140--1144, 2022.

\bibitem[Li et~al.(2023{\natexlab{g}})Li, Xu, and Hu]{MWAFM-LiX023}
Guangyao Li, Yixin Xu, and Di~Hu.
\newblock Multi-scale attention for audio question answering.
\newblock In \emph{24th Annual Conference of the International Speech Communication Association, Interspeech 2023, Dublin, Ireland, August 20-24, 2023}, pages 3442--3446, 2023{\natexlab{g}}.

\bibitem[Stowell et~al.(2018)Stowell, Stylianou, Wood, Pamula, and Glotin]{Bird-Sound-Detection-abs-1807-05812}
Dan Stowell, Yannis Stylianou, Mike Wood, Hanna Pamula, and Herv{\'{e}} Glotin.
\newblock Automatic acoustic detection of birds through deep learning: the first bird audio detection challenge.
\newblock \emph{CoRR}, abs/1807.05812, 2018.

\bibitem[Ani({\natexlab{b}})]{Animal-Sound-Detection}
Animal sound dataset, {\natexlab{b}}.
\newblock URL \url{https://github.com/YashNita/Animal-Sound-Dataset}.

\bibitem[Mesaros et~al.(2016)Mesaros, Heittola, and Virtanen]{TUT17-MesarosHV16}
Annamaria Mesaros, Toni Heittola, and Tuomas Virtanen.
\newblock {TUT} database for acoustic scene classification and sound event detection.
\newblock In \emph{24th European Signal Processing Conference, {EUSIPCO} 2016, Budapest, Hungary, August 29 - September 2, 2016}, pages 1128--1132, 2016.

\bibitem[Piczak(2015)]{ESC50-Piczak15}
Karol~J. Piczak.
\newblock {ESC:} dataset for environmental sound classification.
\newblock In \emph{Proceedings of the 23rd Annual {ACM} Conference on Multimedia Conference, {MM} '15, Brisbane, Australia, October 26 - 30, 2015}, pages 1015--1018, 2015.

\bibitem[Manco et~al.(2023)Manco, Weck, Doh, Won, Zhang, Bogdanov, Wu, Chen, Tovstogan, Benetos, Quinton, Fazekas, and Nam]{Song-Describer-Dataset-manco2023thesong}
Ilaria Manco, Benno Weck, Seungheon Doh, Minz Won, Yixiao Zhang, Dmitry Bogdanov, Yusong Wu, Ke~Chen, Philip Tovstogan, Emmanouil Benetos, Elio Quinton, György Fazekas, and Juhan Nam.
\newblock The song describer dataset: a corpus of audio captions for music-and-language evaluation.
\newblock In \emph{Machine Learning for Audio Workshop at NeurIPS 2023}, 2023.

\bibitem[Hou et~al.(2024)Hou, Liu, Yuan, Xue, Shan, Zhao, and Zhang]{Control-diff-abs-2410-05151}
Siyuan Hou, Shansong Liu, Ruibin Yuan, Wei Xue, Ying Shan, Mangsuo Zhao, and Chao Zhang.
\newblock Editing music with melody and text: Using controlnet for diffusion transformer.
\newblock \emph{CoRR}, abs/2410.05151, 2024.

\bibitem[Lee et~al.(2023)Lee, Park, and Kim]{DailyTalk-LeePK23}
Keon Lee, Kyumin Park, and Daeyoung Kim.
\newblock Dailytalk: Spoken dialogue dataset for conversational text-to-speech.
\newblock In \emph{{IEEE} International Conference on Acoustics, Speech and Signal Processing {ICASSP} 2023, Rhodes Island, Greece, June 4-10, 2023}, pages 1--5, 2023.

\bibitem[Zhou et~al.(2021)Zhou, Sisman, Liu, and Li]{emotion-speech-zhou2021seen}
Kun Zhou, Berrak Sisman, Rui Liu, and Haizhou Li.
\newblock Seen and unseen emotional style transfer for voice conversion with a new emotional speech dataset.
\newblock In \emph{ICASSP 2021-2021 IEEE International Conference on Acoustics, Speech and Signal Processing (ICASSP)}, pages 920--924. IEEE, 2021.

\bibitem[Panayotov et~al.(2015)Panayotov, Chen, Povey, and Khudanpur]{Librispeech-PanayotovCPK15}
Vassil Panayotov, Guoguo Chen, Daniel Povey, and Sanjeev Khudanpur.
\newblock Librispeech: An {ASR} corpus based on public domain audio books.
\newblock In \emph{2015 {IEEE} International Conference on Acoustics, Speech and Signal Processing, {ICASSP} 2015, South Brisbane, Queensland, Australia, April 19-24, 2015}, pages 5206--5210, 2015.

\bibitem[Zhang et~al.(2023{\natexlab{l}})Zhang, Zhang, Li, Zhou, and Qiu]{SpeechTokenizer-zhang2023}
Xin Zhang, Dong Zhang, Shimin Li, Yaqian Zhou, and Xipeng Qiu.
\newblock Speechtokenizer: Unified speech tokenizer for speech language models, 2023{\natexlab{l}}.

\bibitem[Guan et~al.(2024)Guan, Li, Li, Huang, Wang, Lin, Huang, Li, and Hong]{MM-TTS-GuanLLHWLHLH24}
Wenhao Guan, Yishuang Li, Tao Li, Hukai Huang, Feng Wang, Jiayan Lin, Lingyan Huang, Lin Li, and Qingyang Hong.
\newblock {MM-TTS:} multi-modal prompt based style transfer for expressive text-to-speech synthesis.
\newblock In \emph{Thirty-Eighth {AAAI} Conference on Artificial Intelligence, {AAAI} 2024, Thirty-Sixth Conference on Innovative Applications of Artificial Intelligence, {IAAI} 2024, Fourteenth Symposium on Educational Advances in Artificial Intelligence, {EAAI} 2014, February 20-27, 2024, Vancouver, Canada}, pages 18117--18125, 2024.

\bibitem[Liu et~al.(2024{\natexlab{g}})Liu, Yuan, Liu, Mei, Kong, Tian, Wang, Wang, Wang, and Plumbley]{AudioLDM2-LiuYLMKTWWWP24}
Haohe Liu, Yi~Yuan, Xubo Liu, Xinhao Mei, Qiuqiang Kong, Qiao Tian, Yuping Wang, Wenwu Wang, Yuxuan Wang, and Mark~D. Plumbley.
\newblock Audioldm 2: Learning holistic audio generation with self-supervised pretraining.
\newblock \emph{{IEEE} {ACM} Trans. Audio Speech Lang. Process.}, 32:\penalty0 2871--2883, 2024{\natexlab{g}}.

\bibitem[Harwath and Glass(2015)]{Flicker8k}
David Harwath and James Glass.
\newblock Deep multimodal semantic embeddings for speech and images, 2015.

\bibitem[Kim et~al.(2023{\natexlab{b}})Kim, Choi, Maiti, Yeo, Watanabe, and Ro]{Img2Sp}
Minsu Kim, Jeongsoo Choi, Soumi Maiti, Jeong~Hun Yeo, Shinji Watanabe, and Yong~Man Ro.
\newblock Towards practical and efficient image-to-speech captioning with vision-language pre-training and multi-modal tokens, 2023{\natexlab{b}}.

\bibitem[Zhang et~al.(2024{\natexlab{g}})Zhang, Mo, Zhang, and Morgado]{AVSync15}
Lin Zhang, Shentong Mo, Yijing Zhang, and Pedro Morgado.
\newblock Audio-synchronized visual animation, 2024{\natexlab{g}}.

\bibitem[Luo et~al.(2023)Luo, Yan, Hu, and Zhao]{Diff-Foley-luo2023}
Simian Luo, Chuanhao Yan, Chenxu Hu, and Hang Zhao.
\newblock Diff-foley: Synchronized video-to-audio synthesis with latent diffusion models, 2023.

\bibitem[Junichi et~al.(2019)Junichi, Christophe, and Kirsten]{VCTK-Corpus}
Yamagishi Junichi, Veaux Christophe, and MacDonald Kirsten.
\newblock Cstr vctk corpus: English multi-speaker corpus for cstr voice cloning toolkit, 2019.

\bibitem[Ye et~al.(2023{\natexlab{b}})Ye, Zhao, Ko, Meng, Wang, Wang, and Cao]{GigaST-ye2023}
Rong Ye, Chengqi Zhao, Tom Ko, Chutong Meng, Tao Wang, Mingxuan Wang, and Jun Cao.
\newblock Gigast: A 10,000-hour pseudo speech translation corpus, 2023{\natexlab{b}}.

\bibitem[Wang et~al.(2020{\natexlab{b}})Wang, Pino, Wu, and Gu]{CoVoST-wang-etal-2020-covost}
Changhan Wang, Juan Pino, Anne Wu, and Jiatao Gu.
\newblock {C}o{V}o{ST}: A diverse multilingual speech-to-text translation corpus.
\newblock In \emph{Proceedings of the Twelfth Language Resources and Evaluation Conference}, pages 4197--4203, May 2020{\natexlab{b}}.

\bibitem[Crestel and Esling(2017)]{crestel2017liveorchestralpianorealtime}
Léopold Crestel and Philippe Esling.
\newblock Live orchestral piano, a system for real-time orchestral music generation, 2017.

\bibitem[Wang* et~al.(2020)Wang*, Chen*, Jiang, Zhang, Xu, Dai, Bin, and Xia]{pop909-ismir2020}
Ziyu Wang*, Ke~Chen*, Junyan Jiang, Yiyi Zhang, Maoran Xu, Shuqi Dai, Guxian Bin, and Gus Xia.
\newblock Pop909: A pop-song dataset for music arrangement generation.
\newblock In \emph{Proceedings of 21st International Conference on Music Information Retrieval, {ISMIR}}, 2020.

\bibitem[Shih et~al.(2023)Shih, Wu, Zalkow, M{\"{u}}ller, and Yang]{Theme-Transformer-ShihWZMY23}
Yi{-}Jen Shih, Shih{-}Lun Wu, Frank Zalkow, Meinard M{\"{u}}ller, and Yi{-}Hsuan Yang.
\newblock Theme transformer: Symbolic music generation with theme-conditioned transformer.
\newblock \emph{{IEEE} Trans. Multim.}, 25:\penalty0 3495--3508, 2023.

\bibitem[Zhang et~al.(2022)Zhang, Li, Wang, Deng, Liu, Ren, He, Huang, Zhu, Chen, and Zhao]{msinger-zhang2022}
Lichao Zhang, Ruiqi Li, Shoutong Wang, Liqun Deng, Jinglin Liu, Yi~Ren, Jinzheng He, Rongjie Huang, Jieming Zhu, Xiao Chen, and Zhou Zhao.
\newblock M4singer: A multi-style, multi-singer and musical score provided mandarin singing corpus.
\newblock In \emph{Thirty-sixth Conference on Neural Information Processing Systems Datasets and Benchmarks Track}, 2022.

\bibitem[Liu et~al.(2022)Liu, Li, Ren, Chen, and Zhao]{DiffSinger-Liu00CZ22}
Jinglin Liu, Chengxi Li, Yi~Ren, Feiyang Chen, and Zhou Zhao.
\newblock Diffsinger: Singing voice synthesis via shallow diffusion mechanism.
\newblock In \emph{Thirty-Sixth {AAAI} Conference on Artificial Intelligence, {AAAI} 2022, Thirty-Fourth Conference on Innovative Applications of Artificial Intelligence, {IAAI} 2022, The Twelveth Symposium on Educational Advances in Artificial Intelligence, {EAAI} 2022 Virtual Event, February 22 - March 1, 2022}, pages 11020--11028, 2022.

\bibitem[Forsgren and Martiros(2022)]{Riffusion-Forsgren_Martiros_2022}
Seth* Forsgren and Hayk* Martiros.
\newblock {Riffusion - Stable diffusion for real-time music generation}.
\newblock 2022.
\newblock URL \url{https://riffusion.com/about}.

\bibitem[Li et~al.(2024{\natexlab{l}})Li, Zhang, Tang, Ma, Dong, and Xu]{MusicIT-LiZTMDX24}
Sifei Li, Yuxin Zhang, Fan Tang, Chongyang Ma, Weiming Dong, and Changsheng Xu.
\newblock Music style transfer with time-varying inversion of diffusion models.
\newblock In \emph{Thirty-Eighth {AAAI} Conference on Artificial Intelligence, {AAAI} 2024, Thirty-Sixth Conference on Innovative Applications of Artificial Intelligence, {IAAI} 2024, Fourteenth Symposium on Educational Advances in Artificial Intelligence, {EAAI} 2014, February 20-27, 2024, Vancouver, Canada}, pages 547--555, 2024{\natexlab{l}}.

\bibitem[C{\'{\i}}fka et~al.(2020)C{\'{\i}}fka, Simsekli, and Richard]{Groove2Groove-CifkaSR20}
Ondrej C{\'{\i}}fka, Umut Simsekli, and Ga{\"{e}}l Richard.
\newblock Groove2groove: One-shot music style transfer with supervision from synthetic data.
\newblock \emph{{IEEE} {ACM} Trans. Audio Speech Lang. Process.}, 28:\penalty0 2638--2650, 2020.

\bibitem[Wu et~al.(2015)Wu, Song, Khosla, Yu, Zhang, Tang, and Xiao]{WuSKYZTX15}
Zhirong Wu, Shuran Song, Aditya Khosla, Fisher Yu, Linguang Zhang, Xiaoou Tang, and Jianxiong Xiao.
\newblock 3d shapenets: {A} deep representation for volumetric shapes.
\newblock In \emph{{IEEE} Conference on Computer Vision and Pattern Recognition, {CVPR} 2015, Boston, MA, USA, June 7-12, 2015}, pages 1912--1920, 2015.

\bibitem[Liang et~al.(2024)Liang, Feng, Zhou, Zhang, Zou, and Bai]{liang2024pointgst}
Dingkang Liang, Tianrui Feng, Xin Zhou, Yumeng Zhang, Zhikang Zou, and Xiang Bai.
\newblock Parameter-efficient fine-tuning in spectral domain for point cloud learning.
\newblock \emph{arXiv preprint arXiv:2410.08114}, 2024.

\bibitem[Jain et~al.(2024)Jain, Katara, Gkanatsios, Harley, Sarch, Aggarwal, Chaudhary, and Fragkiadaki]{jain2024odin}
Ayush Jain, Pushkal Katara, Nikolaos Gkanatsios, Adam~W. Harley, Gabriel Sarch, Kriti Aggarwal, Vishrav Chaudhary, and Katerina Fragkiadaki.
\newblock Odin: A single model for 2d and 3d perception, 2024.

\bibitem[Behley et~al.(2019)Behley, Garbade, Milioto, Quenzel, Behnke, Stachniss, and Gall]{BehleyGMQBSG19}
Jens Behley, Martin Garbade, Andres Milioto, Jan Quenzel, Sven Behnke, Cyrill Stachniss, and J{\"{u}}rgen Gall.
\newblock Semantickitti: {A} dataset for semantic scene understanding of lidar sequences.
\newblock In \emph{2019 {IEEE/CVF} International Conference on Computer Vision, {ICCV} 2019, Seoul, Korea (South), October 27 - November 2, 2019}, pages 9296--9306, 2019.

\bibitem[Team(2020)]{openpcdet2020}
OpenPCDet~Development Team.
\newblock Openpcdet: An open-source toolbox for 3d object detection from point clouds.
\newblock \url{https://github.com/open-mmlab/OpenPCDet}, 2020.

\bibitem[Sangyun~Shin(2024)]{shin2024spherical}
Madhu Vankadari Andrew Markham Niki~Trigoni Sangyun~Shin, Kaichen~Zhou.
\newblock Spherical mask: Coarse-to-fine 3d point cloud instance segmentation with spherical representation.
\newblock In \emph{Proceedings of the IEEE/CVF Conference on Computer Vision and Pattern Recognition (CVPR)}, 2024.

\bibitem[Dellenbach et~al.(2021)Dellenbach, Deschaud, Jacquet, and Goulette]{dellenbach2021cticp}
Pierre Dellenbach, Jean-Emmanuel Deschaud, Bastien Jacquet, and François Goulette.
\newblock Ct-icp: Real-time elastic lidar odometry with loop closure, 2021.

\bibitem[Chang et~al.(2015)Chang, Funkhouser, Guibas, Hanrahan, Huang, Li, Savarese, Savva, Song, Su, Xiao, Yi, and Yu]{ChangFGHHLSSSSX15}
Angel~X. Chang, Thomas~A. Funkhouser, Leonidas~J. Guibas, Pat Hanrahan, Qi{-}Xing Huang, Zimo Li, Silvio Savarese, Manolis Savva, Shuran Song, Hao Su, Jianxiong Xiao, Li~Yi, and Fisher Yu.
\newblock Shapenet: An information-rich 3d model repository.
\newblock \emph{CoRR}, abs/1512.03012, 2015.

\bibitem[Park et~al.(2023)Park, Lee, Kim, Xiong, and Kim]{ParkLKXK23}
Jinyoung Park, Sanghyeok Lee, Sihyeon Kim, Yunyang Xiong, and Hyunwoo~J. Kim.
\newblock Self-positioning point-based transformer for point cloud understanding.
\newblock In \emph{{IEEE/CVF} Conference on Computer Vision and Pattern Recognition, {CVPR} 2023, Vancouver, BC, Canada, June 17-24, 2023}, pages 21814--21823, 2023.

\bibitem[Yin et~al.(2021)Yin, Zhou, and Kr{\"{a}}henb{\"{u}}hl]{YinZK21}
Tianwei Yin, Xingyi Zhou, and Philipp Kr{\"{a}}henb{\"{u}}hl.
\newblock Center-based 3d object detection and tracking.
\newblock In \emph{{IEEE} Conference on Computer Vision and Pattern Recognition, {CVPR} 2021, virtual, June 19-25, 2021}, pages 11784--11793, 2021.

\bibitem[Guerrero et~al.(2018)Guerrero, Kleiman, Ovsjanikov, and Mitra]{GuerreroEtAl-PCPNet}
Paul Guerrero, Yanir Kleiman, Maks Ovsjanikov, and Niloy~J. Mitra.
\newblock {PCPNet}: Learning local shape properties from raw point clouds.
\newblock \emph{Computer Graphics Forum}, 37\penalty0 (2):\penalty0 75--85, 2018.

\bibitem[Li et~al.(2023{\natexlab{h}})Li, Feng, Shi, Gao, Fang, Liu, and Han]{LiFSGFLH23}
Qing Li, Huifang Feng, Kanle Shi, Yue Gao, Yi~Fang, Yu{-}Shen Liu, and Zhizhong Han.
\newblock Shs-net: Learning signed hyper surfaces for oriented normal estimation of point clouds.
\newblock In \emph{{IEEE/CVF} Conference on Computer Vision and Pattern Recognition, {CVPR} 2023, Vancouver, BC, Canada, June 17-24, 2023}, pages 13591--13600, 2023{\natexlab{h}}.

\bibitem[Liu et~al.(2023{\natexlab{f}})Liu, Tang, Amini, Yang, Mao, Rus, and Han]{LiuTAYMRH23}
Zhijian Liu, Haotian Tang, Alexander Amini, Xinyu Yang, Huizi Mao, Daniela~L. Rus, and Song Han.
\newblock Bevfusion: Multi-task multi-sensor fusion with unified bird's-eye view representation.
\newblock In \emph{{IEEE} International Conference on Robotics and Automation, {ICRA} 2023, London, UK, May 29 - June 2, 2023}, pages 2774--2781, 2023{\natexlab{f}}.

\bibitem[Azuma et~al.(2022)Azuma, Miyanishi, Kurita, and Kawanabe]{AzumaMKK22}
Daichi Azuma, Taiki Miyanishi, Shuhei Kurita, and Motoaki Kawanabe.
\newblock Scanqa: 3d question answering for spatial scene understanding.
\newblock In \emph{{IEEE/CVF} Conference on Computer Vision and Pattern Recognition, {CVPR} 2022, New Orleans, LA, USA, June 18-24, 2022}, pages 19107--19117, 2022.

\bibitem[Man et~al.(2024)Man, Gui, and Wang]{man2024situational}
Yunze Man, Liang-Yan Gui, and Yu-Xiong Wang.
\newblock Situational awareness matters in 3d vision language reasoning.
\newblock In \emph{Proceedings of the IEEE/CVF Conference on Computer Vision and Pattern Recognition}, pages 13678--13688, 2024.

\bibitem[Ma et~al.(2023)Ma, Yong, Zheng, Li, Liang, Zhu, and Huang]{MaYZ0LZH23}
Xiaojian Ma, Silong Yong, Zilong Zheng, Qing Li, Yitao Liang, Song{-}Chun Zhu, and Siyuan Huang.
\newblock {SQA3D:} situated question answering in 3d scenes.
\newblock In \emph{The Eleventh International Conference on Learning Representations, {ICLR} 2023, Kigali, Rwanda, May 1-5, 2023}, 2023.

\bibitem[Plappert et~al.(2016)Plappert, Mandery, and Asfour]{PlappertMA16}
Matthias Plappert, Christian Mandery, and Tamim Asfour.
\newblock The {KIT} motion-language dataset.
\newblock \emph{Big Data}, 4\penalty0 (4):\penalty0 236--252, 2016.

\bibitem[Radouane et~al.(2024)Radouane, Tchechmedjiev, Ranwez, and Lagarde]{radouane2024guided}
Karim Radouane, Andon Tchechmedjiev, Sylvie Ranwez, and Julien Lagarde.
\newblock Guided attention for interpretable motion captioning.
\newblock In \emph{Proceedings of the 35th British Machine Vision Conference}, 2024.

\bibitem[Yuan et~al.(2018)Yuan, Khot, Held, Mertz, and Hebert]{yuan2018pcn}
Wentao Yuan, Tejas Khot, David Held, Christoph Mertz, and Martial Hebert.
\newblock Pcn: Point completion network.
\newblock In \emph{2018 International Conference on 3D Vision (3DV)}, pages 728--737, 2018.

\bibitem[Yu et~al.(2021)Yu, Rao, Wang, Liu, Lu, and Zhou]{YuRWLL021}
Xumin Yu, Yongming Rao, Ziyi Wang, Zuyan Liu, Jiwen Lu, and Jie Zhou.
\newblock Pointr: Diverse point cloud completion with geometry-aware transformers.
\newblock In \emph{2021 {IEEE/CVF} International Conference on Computer Vision, {ICCV} 2021, Montreal, QC, Canada, October 10-17, 2021}, pages 12478--12487, 2021.

\bibitem[Peng et~al.(2020)Peng, Niemeyer, Mescheder, Pollefeys, and Geiger]{Peng2020ECCV}
Songyou Peng, Michael Niemeyer, Lars Mescheder, Marc Pollefeys, and Andreas Geiger.
\newblock Convolutional occupancy networks.
\newblock In \emph{European Conference on Computer Vision (ECCV)}, 2020.

\bibitem[Nichol et~al.(2022)Nichol, Jun, Dhariwal, Mishkin, and Chen]{Point-E-abs-2212-08751}
Alex Nichol, Heewoo Jun, Prafulla Dhariwal, Pamela Mishkin, and Mark Chen.
\newblock Point-e: {A} system for generating 3d point clouds from complex prompts.
\newblock \emph{CoRR}, abs/2212.08751, 2022.

\bibitem[Yang et~al.(2024{\natexlab{e}})Yang, Shi, Zhang, Yang, Wang, Zhao, Liu, Wang, Lin, Yu, Wang, Chen, Liu, Liu, Yang, Wang, Jiang, and Guo]{yang2024tencent}
Xianghui Yang, Huiwen Shi, Bowen Zhang, Fan Yang, Jiacheng Wang, Hongxu Zhao, Xinhai Liu, Xinzhou Wang, Qingxiang Lin, Jiaao Yu, Lifu Wang, Zhuo Chen, Sicong Liu, Yuhong Liu, Yong Yang, Di~Wang, Jie Jiang, and Chunchao Guo.
\newblock Tencent hunyuan3d-1.0: A unified framework for text-to-3d and image-to-3d generation, 2024{\natexlab{e}}.

\bibitem[Deitke et~al.(2023)Deitke, Schwenk, Salvador, Weihs, Michel, VanderBilt, Schmidt, Ehsani, Kembhavi, and Farhadi]{DeitkeSSWMVSEKF23}
Matt Deitke, Dustin Schwenk, Jordi Salvador, Luca Weihs, Oscar Michel, Eli VanderBilt, Ludwig Schmidt, Kiana Ehsani, Aniruddha Kembhavi, and Ali Farhadi.
\newblock Objaverse: {A} universe of annotated 3d objects.
\newblock In \emph{{IEEE/CVF} Conference on Computer Vision and Pattern Recognition, {CVPR} 2023, Vancouver, BC, Canada, June 17-24, 2023}, pages 13142--13153, 2023.

\bibitem[Zhou et~al.(2018)Zhou, Park, and Koltun]{Zhou2018}
Qian-Yi Zhou, Jaesik Park, and Vladlen Koltun.
\newblock {Open3D}: {A} modern library for {3D} data processing.
\newblock \emph{arXiv:1801.09847}, 2018.

\bibitem[Guo et~al.(2024{\natexlab{b}})Guo, Mu, Javed, Wang, and Cheng]{GuoMJW024}
Chuan Guo, Yuxuan Mu, Muhammad~Gohar Javed, Sen Wang, and Li~Cheng.
\newblock Momask: Generative masked modeling of 3d human motions.
\newblock In \emph{{IEEE/CVF} Conference on Computer Vision and Pattern Recognition, {CVPR} 2024, Seattle, WA, USA, June 16-22, 2024}, pages 1900--1910, 2024{\natexlab{b}}.

\bibitem[Talmor et~al.(2019)Talmor, Herzig, Lourie, and Berant]{atca-naacl-19}
Alon Talmor, Jonathan Herzig, Nicholas Lourie, and Jonathan Berant.
\newblock Commonsenseqa: {A} question answering challenge targeting commonsense knowledge.
\newblock In \emph{Proceedings of NAACL}, pages 4149--4158, 2019.

\bibitem[Chung et~al.(2022)Chung, Hou, Longpre, Zoph, Tay, Fedus, Li, Wang, Dehghani, Brahma, Webson, Gu, Dai, Suzgun, Chen, Chowdhery, Narang, Mishra, Yu, Zhao, Huang, Dai, Yu, Petrov, Chi, Dean, Devlin, Roberts, Zhou, Le, and Wei]{abs-2210-11416}
Hyung~Won Chung, Le~Hou, Shayne Longpre, Barret Zoph, Yi~Tay, William Fedus, Eric Li, Xuezhi Wang, Mostafa Dehghani, Siddhartha Brahma, Albert Webson, Shixiang~Shane Gu, Zhuyun Dai, Mirac Suzgun, Xinyun Chen, Aakanksha Chowdhery, Sharan Narang, Gaurav Mishra, Adams Yu, Vincent~Y. Zhao, Yanping Huang, Andrew~M. Dai, Hongkun Yu, Slav Petrov, Ed~H. Chi, Jeff Dean, Jacob Devlin, Adam Roberts, Denny Zhou, Quoc~V. Le, and Jason Wei.
\newblock Scaling instruction-finetuned language models.
\newblock \emph{CoRR}, abs/2210.11416, 2022.

\bibitem[Jin et~al.(2024)Jin, Liu, Lyu, Poff, Sachan, Mihalcea, Diab, and Sch{\"{o}}lkopf]{zjcl-iclr-24}
Zhijing Jin, Jiarui Liu, Zhiheng Lyu, Spencer Poff, Mrinmaya Sachan, Rada Mihalcea, Mona~T. Diab, and Bernhard Sch{\"{o}}lkopf.
\newblock Can large language models infer causation from correlation?
\newblock In \emph{Proceedings of ICLR}, 2024.

\bibitem[Pe{\~{n}}as et~al.(2013)Pe{\~{n}}as, Hovy, Forner, Rodrigo, Sutcliffe, and Morante]{apq2-clef-13}
Anselmo Pe{\~{n}}as, Eduard~H. Hovy, Pamela Forner, {\'{A}}lvaro Rodrigo, Richard F.~E. Sutcliffe, and Roser Morante.
\newblock {QA4MRE} 2011-2013: Overview of question answering for machine reading evaluation.
\newblock In \emph{Proceedings of Information Access Evaluation. Multilinguality, Multimodality, and Visualization - 4th International Conference of the CLEF Initiative, CLEF 2013, Valencia, Spain, September 23-26, 2013. Proceedings}, pages 303--320, 2013.

\bibitem[crc()]{crcllm}
Counterfactual reasoning capacity of large language models.
\newblock URL \url{https://github.com/SushovitNanda/Counterfactual-Reasoning-Capacity-of-Large-Language-Models}.

\bibitem[Gladkova et~al.(2016)Gladkova, Drozd, and Matsuoka]{agab-naacl-16}
Anna Gladkova, Aleksandr Drozd, and Satoshi Matsuoka.
\newblock Analogy-based detection of morphological and semantic relations with word embeddings: what works and what doesn't.
\newblock In \emph{Proceedings of NAACL}, pages 8--15, 2016.

\bibitem[Yang et~al.(2018)Yang, Qi, Zhang, Bengio, Cohen, Salakhutdinov, and Manning]{yang2018hotpotqa}
Zhilin Yang, Peng Qi, Saizheng Zhang, Yoshua Bengio, William~W. Cohen, Ruslan Salakhutdinov, and Christopher~D. Manning.
\newblock {HotpotQA}: A dataset for diverse, explainable multi-hop question answering.
\newblock In \emph{Conference on Empirical Methods in Natural Language Processing ({EMNLP})}, 2018.

\bibitem[Qin et~al.(2021)Qin, Gupta, Upadhyay, He, Choi, and Faruqui]{qin-etal-2021-timedial}
Lianhui Qin, Aditya Gupta, Shyam Upadhyay, Luheng He, Yejin Choi, and Manaal Faruqui.
\newblock {TimeDial: Temporal Commonsense Reasoning in Dialog}.
\newblock In \emph{Proc. of ACL}, 2021.

\bibitem[Shi et~al.(2022{\natexlab{b}})Shi, Zhang, and Lipani]{stepGame2022shi}
Zhengxiang Shi, Qiang Zhang, and Aldo Lipani.
\newblock Stepgame: A new benchmark for robust multi-hop spatial reasoning in texts.
\newblock In \emph{Proceedings of the AAAI Conference on Artificial Intelligence}, volume~36, pages 11321--11329, Jun. 2022{\natexlab{b}}.

\bibitem[Ling et~al.(2017)Ling, Yogatama, Dyer, and Blunsom]{wlpi-acl-17}
Wang Ling, Dani Yogatama, Chris Dyer, and Phil Blunsom.
\newblock Program induction by rationale generation: Learning to solve and explain algebraic word problems.
\newblock In \emph{Proceedings of ACL}, pages 158--167, 2017.

\bibitem[Hendrycks et~al.(2021)Hendrycks, Burns, Basart, Critch, Li, Song, and Steinhardt]{dhaa-iclr-21}
Dan Hendrycks, Collin Burns, Steven Basart, Andrew Critch, Jerry Li, Dawn Song, and Jacob Steinhardt.
\newblock Aligning {AI} with shared human values.
\newblock In \emph{Proceedings of ICLR}, 2021.

\bibitem[Zhong et~al.(2020)Zhong, Xiao, Tu, Zhang, Liu, and Sun]{hzjq-aaai-20}
Haoxi Zhong, Chaojun Xiao, Cunchao Tu, Tianyang Zhang, Zhiyuan Liu, and Maosong Sun.
\newblock {JEC-QA:} {A} legal-domain question answering dataset.
\newblock In \emph{Proceedings of AAAI}, pages 9701--9708, 2020.

\bibitem[Pryzant et~al.(2020)Pryzant, Martinez, Dass, Kurohashi, Jurafsky, and Yang]{rpan-aaai-20}
Reid Pryzant, Richard~Diehl Martinez, Nathan Dass, Sadao Kurohashi, Dan Jurafsky, and Diyi Yang.
\newblock Automatically neutralizing subjective bias in text.
\newblock In \emph{Proceedings of AAAI}, pages 480--489, 2020.

\bibitem[Liu(2019)]{liu2019roberta}
Yinhan Liu.
\newblock Roberta: A robustly optimized bert pretraining approach.
\newblock \emph{arXiv preprint arXiv:1907.11692}, 364, 2019.

\bibitem[ock()]{ockaggle}
Offensive classification.
\newblock URL \url{https://www.kaggle.com/datasets/mrmorj/hate-speech-and-offensive-language-dataset}.

\bibitem[spa()]{spamdkaggle}
Spam detection.
\newblock URL \url{https://www.kaggle.com/datasets/jackksoncsie/spam-email-dataset}.

\bibitem[fak()]{fakekaggle}
Fake news.
\newblock URL \url{https://kaggle.com/competitions/fake-news}.

\bibitem[O'Neill et~al.(2021)O'Neill, Rozenshtein, Kiryo, Kubota, and Bollegala]{joiw-emnlp-21}
James O'Neill, Polina Rozenshtein, Ryuichi Kiryo, Motoko Kubota, and Danushka Bollegala.
\newblock I wish {I} would have loved this one, but {I} didn't - {A} multilingual dataset for counterfactual detection in product review.
\newblock In \emph{Proceedings of EMNLP}, pages 7092--7108, 2021.

\bibitem[bqb()]{bqbioasq}
Biomedical question answering (biomedical qa).
\newblock URL \url{http://participants-area.bioasq.org/datasets/}.

\bibitem[Li et~al.(2023{\natexlab{i}})Li, Li, Zhang, Dan, Jiang, and Zhang]{li2023chatdoctor}
Yunxiang Li, Zihan Li, Kai Zhang, Ruilong Dan, Steve Jiang, and You Zhang.
\newblock Chatdoctor: A medical chat model fine-tuned on a large language model meta-ai (llama) using medical domain knowledge.
\newblock \emph{Cureus}, 15\penalty0 (6), 2023{\natexlab{i}}.

\bibitem[sod()]{sodkaggle}
Stack overflow data, booktitle = {kaggle}, url ={https://www.kaggle.com/datasets/stackoverflow/stackoverflow/data}.

\bibitem[eqa()]{eqabath}
Engineering question answering, booktitle = {bath}, url ={https://researchportal.bath.ac.uk/en/organisations/department-of-mechanical-engineering/datasets/}.

\bibitem[Welbl et~al.(2017)Welbl, Liu, and Gardner]{jwcm-aclnut-17}
Johannes Welbl, Nelson~F. Liu, and Matt Gardner.
\newblock Crowdsourcing multiple choice science questions.
\newblock In \emph{Proceedings of EMNLP}, pages 94--106, 2017.

\bibitem[Rajpurkar et~al.(2018)Rajpurkar, Jia, and Liang]{prkw-acl-18}
Pranav Rajpurkar, Robin Jia, and Percy Liang.
\newblock Know what you don't know: Unanswerable questions for squad.
\newblock In \emph{Proceedings of ACL}, pages 784--789, 2018.

\bibitem[Sap et~al.(2019)Sap, Rashkin, Chen, Bras, and Choi]{mssi-emnlp-19}
Maarten Sap, Hannah Rashkin, Derek Chen, Ronan~Le Bras, and Yejin Choi.
\newblock Social iqa: Commonsense reasoning about social interactions.
\newblock In \emph{Proceedings of EMNLP}, pages 4462--4472, 2019.

\bibitem[pqa()]{pqahugging}
Philosophical question answering, booktitle = {huggingface}, url ={https://huggingface.co/datasets/sayhan/strix-philosophy-qaa}.

\bibitem[Kenter et~al.(2018)Kenter, Jones, and Hewlett]{tkbl-aaai-18}
Tom Kenter, Llion Jones, and Daniel Hewlett.
\newblock Byte-level machine reading across morphologically varied languages.
\newblock In \emph{Proceedings of AAAI}, pages 5820--5827, 2018.

\bibitem[Xiong et~al.(2019)Xiong, Wu, Wang, Kulkarni, Yu, Chang, Guo, and Wang]{wxta-acl-19}
Wenhan Xiong, Jiawei Wu, Hong Wang, Vivek Kulkarni, Mo~Yu, Shiyu Chang, Xiaoxiao Guo, and William~Yang Wang.
\newblock {TWEETQA:} {A} social media focused question answering dataset.
\newblock In \emph{Proceedings of ACL}, pages 5020--5031, 2019.

\bibitem[Joshi et~al.(2017)Joshi, Choi, Weld, and Zettlemoyer]{mjta-acl-17}
Mandar Joshi, Eunsol Choi, Daniel~S. Weld, and Luke Zettlemoyer.
\newblock Triviaqa: {A} large scale distantly supervised challenge dataset for reading comprehension.
\newblock In \emph{Proceedings of ACL}, pages 1601--1611, 2017.

\bibitem[Reddy et~al.(2019)Reddy, Chen, and Manning]{srca-tacl-19}
Siva Reddy, Danqi Chen, and Christopher~D. Manning.
\newblock Coqa: {A} conversational question answering challenge.
\newblock \emph{Trans. Assoc. Comput. Linguistics}, 7:\penalty0 249--266, 2019.

\bibitem[Pasupat and Liang(2015)]{ppcs-acl-15}
Panupong Pasupat and Percy Liang.
\newblock Compositional semantic parsing on semi-structured tables.
\newblock In \emph{Proceedings of ACL}, pages 1470--1480, 2015.

\bibitem[mqa()]{mqahg}
Multilingual question answering.
\newblock URL \url{https://huggingface.co/datasets/crodri/multilingual_qa}.

\bibitem[Xue et~al.(2021)Xue, Constant, Roberts, Kale, Al-Rfou, Siddhant, Barua, and Raffel]{xue-etal-2021-mt5}
Linting Xue, Noah Constant, Adam Roberts, Mihir Kale, Rami Al-Rfou, Aditya Siddhant, Aditya Barua, and Colin Raffel.
\newblock m{T}5: A massively multilingual pre-trained text-to-text transformer.
\newblock In \emph{Proceedings of the NAACL}, pages 483--498, 2021.

\bibitem[Amini et~al.(2019)Amini, Gabriel, Lin, Koncel{-}Kedziorski, Choi, and Hajishirzi]{aamt-naacl-19}
Aida Amini, Saadia Gabriel, Shanchuan Lin, Rik Koncel{-}Kedziorski, Yejin Choi, and Hannaneh Hajishirzi.
\newblock Mathqa: Towards interpretable math word problem solving with operation-based formalisms.
\newblock In \emph{Proceedings of NAACL}, pages 2357--2367, 2019.

\bibitem[Lu et~al.(2021{\natexlab{b}})Lu, Guo, Ren, Huang, Svyatkovskiy, Blanco, Clement, Drain, Jiang, Tang, Li, Zhou, Shou, Zhou, Tufano, Gong, Zhou, Duan, Sundaresan, Deng, Fu, and Liu]{slca-nips-21}
Shuai Lu, Daya Guo, Shuo Ren, Junjie Huang, Alexey Svyatkovskiy, Ambrosio Blanco, Colin~B. Clement, Dawn Drain, Daxin Jiang, Duyu Tang, Ge~Li, Lidong Zhou, Linjun Shou, Long Zhou, Michele Tufano, Ming Gong, Ming Zhou, Nan Duan, Neel Sundaresan, Shao~Kun Deng, Shengyu Fu, and Shujie Liu.
\newblock Codexglue: {A} machine learning benchmark dataset for code understanding and generation.
\newblock In \emph{Proceedings of NeurIPS}, 2021{\natexlab{b}}.

\bibitem[Wang et~al.(2021)Wang, Wang, Joty, and Hoi]{wang2021codet5}
Yue Wang, Weishi Wang, Shafiq Joty, and Steven~CH Hoi.
\newblock Codet5: Identifier-aware unified pre-trained encoder-decoder models for code understanding and generation.
\newblock \emph{arXiv preprint arXiv:2109.00859}, 2021.

\bibitem[mbp()]{mbppgithub}
Mostly basic python problems dataset.
\newblock URL \url{https://github.com/google-research/google-research/tree/master/mbpp}.

\bibitem[Goyal et~al.(2022)Goyal, Gao, Chaudhary, Chen, Wenzek, Ju, Krishnan, Ranzato, Guzm{\'{a}}n, and Fan]{ngtf-tacl-22}
Naman Goyal, Cynthia Gao, Vishrav Chaudhary, Peng{-}Jen Chen, Guillaume Wenzek, Da~Ju, Sanjana Krishnan, Marc'Aurelio Ranzato, Francisco Guzm{\'{a}}n, and Angela Fan.
\newblock The flores-101 evaluation benchmark for low-resource and multilingual machine translation.
\newblock \emph{Trans. Assoc. Comput. Linguistics}, 10:\penalty0 522--538, 2022.

\bibitem[See et~al.(2017)See, Liu, and Manning]{asgt-acl-17}
Abigail See, Peter~J. Liu, and Christopher~D. Manning.
\newblock Get to the point: Summarization with pointer-generator networks.
\newblock In \emph{Proceedings of ACL}, pages 1073--1083, 2017.

\bibitem[Lewis(2019)]{lewis2019bart}
Mike Lewis.
\newblock Bart: Denoising sequence-to-sequence pre-training for natural language generation, translation, and comprehension.
\newblock \emph{arXiv preprint arXiv:1910.13461}, 2019.

\bibitem[Narayan et~al.(2018)Narayan, Cohen, and Lapata]{sndt-emnlp-18}
Shashi Narayan, Shay~B. Cohen, and Mirella Lapata.
\newblock Don't give me the details, just the summary! topic-aware convolutional neural networks for extreme summarization.
\newblock In \emph{Proceedings of EMNLP}, pages 1797--1807, 2018.

\bibitem[Fabbri et~al.(2019)Fabbri, Li, She, Li, and Radev]{armn-acl-19}
Alexander~R. Fabbri, Irene Li, Tianwei She, Suyi Li, and Dragomir~R. Radev.
\newblock Multi-news: {A} large-scale multi-document summarization dataset and abstractive hierarchical model.
\newblock In \emph{Proceedings of ACL}, pages 1074--1084, 2019.

\bibitem[Li et~al.(2017)Li, Su, Shen, Li, Cao, and Niu]{ylda-ijcnlp-17}
Yanran Li, Hui Su, Xiaoyu Shen, Wenjie Li, Ziqiang Cao, and Shuzi Niu.
\newblock Dailydialog: {A} manually labelled multi-turn dialogue dataset.
\newblock In \emph{Proceedings of Proceedings of the Eighth International Joint Conference on Natural Language Processing, IJCNLP 2017, Taipei, Taiwan, November 27 - December 1, 2017 - Volume 1: Long Papers}, pages 986--995, 2017.

\bibitem[Zhang(2019)]{zhang2019dialogpt}
Y~Zhang.
\newblock Dialogpt: Large-scale generative pre-training for conversational response generation.
\newblock \emph{arXiv preprint arXiv:1911.00536}, 2019.

\bibitem[Parikh et~al.(2020)Parikh, Wang, Gehrmann, Faruqui, Dhingra, Yang, and Das]{apta-emnlp-20}
Ankur~P. Parikh, Xuezhi Wang, Sebastian Gehrmann, Manaal Faruqui, Bhuwan Dhingra, Diyi Yang, and Dipanjan Das.
\newblock Totto: {A} controlled table-to-text generation dataset.
\newblock In \emph{Proceedings of EMNLP}, pages 1173--1186, 2020.

\bibitem[Krishna et~al.(2020)Krishna, Wieting, and Iyyer]{kkru-emnlp-20}
Kalpesh Krishna, John Wieting, and Mohit Iyyer.
\newblock Reformulating unsupervised style transfer as paraphrase generation.
\newblock In \emph{Proceedings of EMNLP}, pages 737--762, 2020.

\bibitem[sto()]{storyhg}
Story generation.
\newblock URL \url{https://huggingface.co/datasets/qwedsacf/story-generation}.

\bibitem[gra()]{grammarckgggle}
Grammar correction.
\newblock URL \url{https://www.kaggle.com/datasets/satishgunjal/grammar-correction}.

\bibitem[Das et~al.(2023)Das, Kong, Sen, and Zhou]{das2023decoder}
Abhimanyu Das, Weihao Kong, Rajat Sen, and Yichen Zhou.
\newblock A decoder-only foundation model for time-series forecasting.
\newblock \emph{arXiv preprint arXiv:2310.10688}, 2023.

\bibitem[top()]{topickg}
Topic classification.
\newblock URL \url{https://www.kaggle.com/datasets/thedevastator/the-trec-question-classification-dataset-a-longi/data}.

\bibitem[Bowman et~al.(2015)Bowman, Angeli, Potts, and Manning]{sral-emnlp-15}
Samuel~R. Bowman, Gabor Angeli, Christopher Potts, and Christopher~D. Manning.
\newblock A large annotated corpus for learning natural language inference.
\newblock In \emph{Proceedings of EMNLP}, pages 632--642, 2015.

\bibitem[sen()]{sentkg}
Sentence similarity detection.
\newblock URL \url{https://www.kaggle.com/competitions/quora-question-pairs/data}.

\bibitem[int()]{intentgithub}
Intent detection.
\newblock URL \url{https://github.com/sz128/slot_filling_and_intent_detection_of_SLU/tree/master/data/atis-2}.

\bibitem[Mohammad et~al.(2016)Mohammad, Kiritchenko, Sobhani, Zhu, and Cherry]{sms2-semeval-16}
Saif~M. Mohammad, Svetlana Kiritchenko, Parinaz Sobhani, Xiaodan Zhu, and Colin Cherry.
\newblock Semeval-2016 task 6: Detecting stance in tweets.
\newblock In \emph{Proceedings of SemEval}, pages 31--41, 2016.

\bibitem[per()]{pergithub}
Personality analysis.
\newblock URL \url{https://github.com/hjian42/automatic-personality-prediction/tree/master/data/Essays}.

\bibitem[Weller and Seppi(2019)]{owhd-emnlp-19}
Orion Weller and Kevin~D. Seppi.
\newblock Humor detection: {A} transformer gets the last laugh.
\newblock In \emph{Proceedings of EMNLP}, pages 3619--3623, 2019.

\bibitem[Misra and Arora(2023)]{rmsd-aiopen-23}
Rishabh Misra and Prahal Arora.
\newblock Sarcasm detection using news headlines dataset.
\newblock \emph{AI Open}, 4:\penalty0 13--18, 2023.

\bibitem[Socher et~al.(2013)Socher, Perelygin, Wu, Chuang, Manning, Ng, and Potts]{rsrd-emnlp-13}
Richard Socher, Alex Perelygin, Jean Wu, Jason Chuang, Christopher~D. Manning, Andrew~Y. Ng, and Christopher Potts.
\newblock Recursive deep models for semantic compositionality over a sentiment treebank.
\newblock In \emph{Proceedings of EMNLP}, pages 1631--1642, 2013.

\bibitem[men()]{mentalgithub}
Mental health toxicity detection.
\newblock URL \url{https://github.com/hjian42/automatic-personality-prediction/tree/master/data/Essays}.

\bibitem[Malo et~al.(2014)Malo, Sinha, Korhonen, Wallenius, and Takala]{pmgd-jasis-14}
Pekka Malo, Ankur Sinha, Pekka~J. Korhonen, Jyrki Wallenius, and Pyry Takala.
\newblock Good debt or bad debt: Detecting semantic orientations in economic texts.
\newblock \emph{J. Assoc. Inf. Sci. Technol.}, 65\penalty0 (4):\penalty0 782--796, 2014.

\bibitem[Choi et~al.(2021)Choi, Lee, Choi, Park, Lee, Lee, and Lee]{mcmm-naacl-21}
Minjin Choi, Sunkyung Lee, Eunseong Choi, Heesoo Park, Junhyuk Lee, Dongwon Lee, and Jongwuk Lee.
\newblock Melbert: Metaphor detection via contextualized late interaction using metaphorical identification theories.
\newblock In \emph{Proceedings of NAACL}, pages 1763--1773, 2021.

\bibitem[Pontiki et~al.(2015)Pontiki, Galanis, Papageorgiou, Manandhar, and Androutsopoulos]{mps2-semeval-15}
Maria Pontiki, Dimitris Galanis, Haris Papageorgiou, Suresh Manandhar, and Ion Androutsopoulos.
\newblock Semeval-2015 task 12: Aspect based sentiment analysis.
\newblock In \emph{Proceedings of SemEval}, pages 486--495, 2015.

\bibitem[Fan et~al.(2019{\natexlab{b}})Fan, Wu, Dai, Huang, and Chen]{zfto-naacl-19}
Zhifang Fan, Zhen Wu, Xin{-}Yu Dai, Shujian Huang, and Jiajun Chen.
\newblock Target-oriented opinion words extraction with target-fused neural sequence labeling.
\newblock In \emph{Proceedings of NAACL}, pages 2509--2518, 2019{\natexlab{b}}.

\bibitem[Chen et~al.(2020)Chen, Liu, Wang, Zhang, and Chi]{scsd-acl-20}
Shaowei Chen, Jie Liu, Yu~Wang, Wenzheng Zhang, and Ziming Chi.
\newblock Synchronous double-channel recurrent network for aspect-opinion pair extraction.
\newblock In \emph{Proceedings of ACL}, pages 6515--6524, 2020.

\bibitem[Xu et~al.(2020)Xu, Li, Lu, and Bing]{lxpa-emnlp-20}
Lu~Xu, Hao Li, Wei Lu, and Lidong Bing.
\newblock Position-aware tagging for aspect sentiment triplet extraction.
\newblock In \emph{Proceedings of EMNLP}, pages 2339--2349, 2020.

\bibitem[Pontiki et~al.(2016)Pontiki, Galanis, Papageorgiou, Androutsopoulos, Manandhar, Al{-}Smadi, Al{-}Ayyoub, Zhao, Qin, Clercq, Hoste, Apidianaki, Tannier, Loukachevitch, Kotelnikov, Bel, Jim{\'{e}}nez{-}Zafra, and Eryigit]{mps2-semeval-16}
Maria Pontiki, Dimitris Galanis, Haris Papageorgiou, Ion Androutsopoulos, Suresh Manandhar, Mohammad Al{-}Smadi, Mahmoud Al{-}Ayyoub, Yanyan Zhao, Bing Qin, Orph{\'{e}}e~De Clercq, V{\'{e}}ronique Hoste, Marianna Apidianaki, Xavier Tannier, Natalia~V. Loukachevitch, Evgeniy~V. Kotelnikov, N{\'{u}}ria Bel, Salud~Mar{\'{\i}}a Jim{\'{e}}nez{-}Zafra, and G{\"{u}}lsen Eryigit.
\newblock Semeval-2016 task 5: Aspect based sentiment analysis.
\newblock In \emph{Proceedings of SemEval}, pages 19--30, 2016.

\bibitem[Li et~al.(2023{\natexlab{j}})Li, Fei, Li, Wu, Zhang, Wu, Li, Liu, Liao, Chua, and Ji]{blda-acl-23}
Bobo Li, Hao Fei, Fei Li, Yuhan Wu, Jinsong Zhang, Shengqiong Wu, Jingye Li, Yijiang Liu, Lizi Liao, Tat{-}Seng Chua, and Donghong Ji.
\newblock Diaasq: {A} benchmark of conversational aspect-based sentiment quadruple analysis.
\newblock In \emph{Findings of ACL}, pages 13449--13467, 2023{\natexlab{j}}.

\bibitem[soc()]{soccergithub}
Soccer sentiment classification.
\newblock URL \url{https://github.com/Mr-Chang95/FIFA-Sentiment-Analysis}.

\bibitem[Liu et~al.(2021)Liu, Xia, and Yu]{zlco-emnlp-21}
Ziheng Liu, Rui Xia, and Jianfei Yu.
\newblock Comparative opinion quintuple extraction from product reviews.
\newblock In \emph{Proceedings of EMNLP}, pages 3955--3965, 2021.

\bibitem[Zhang et~al.(2024{\natexlab{h}})Zhang, Chen, Pan, Caragea, Latecki, and Dragut]{qzsa-emnlp-24}
Qi~Zhang, Zhijia Chen, Huitong Pan, Cornelia Caragea, Longin~Jan Latecki, and Eduard Dragut.
\newblock Scier: An entity and relation extraction dataset for datasets, methods, and tasks in scientific documents.
\newblock In \emph{Proceedings of EMNLP}, pages 13083--13100, 2024{\natexlab{h}}.

\bibitem[Li et~al.(2022{\natexlab{e}})Li, Fei, Liu, Wu, Zhang, Teng, Ji, and Li]{li2022unified}
Jingye Li, Hao Fei, Jiang Liu, Shengqiong Wu, Meishan Zhang, Chong Teng, Donghong Ji, and Fei Li.
\newblock Unified named entity recognition as word-word relation classification.
\newblock In \emph{proceedings of the AAAI conference on artificial intelligence}, volume~36, pages 10965--10973, 2022{\natexlab{e}}.

\bibitem[tem()]{temgithub}
Temporal ner.
\newblock URL \url{https://github.com/paramitamirza/Causal-TimeBank/tree/main}.

\bibitem[pat()]{pathnergit}
Pathology ner.
\newblock URL \url{https://github.com/pathology-dynamics/trialsieve/tree/main}.

\bibitem[cyb()]{cybergit}
Cybersecurity ner.
\newblock URL \url{https://github.com/aiforsec/CyNER}.

\bibitem[geo()]{geonergit}
Geological ner.
\newblock URL \url{https://github.com/BritishGeologicalSurvey/geo-ner-model}.

\bibitem[Kalamkar et~al.(2022)Kalamkar, Agarwal, Tiwari, Gupta, Karn, and Raghavan]{pkne-acl-nllp-22}
Prathamesh Kalamkar, Astha Agarwal, Aman Tiwari, Smita Gupta, Saurabh Karn, and Vivek Raghavan.
\newblock Named entity recognition in indian court judgments.
\newblock In \emph{Proceedings of EMNLP}, pages 184--193, 2022.

\bibitem[Sang and Meulder(2003)]{efit-conll-03}
Erik F. Tjong~Kim Sang and Fien~De Meulder.
\newblock Introduction to the conll-2003 shared task: Language-independent named entity recognition.
\newblock In \emph{Proceedings of NAACL}, pages 142--147, 2003.

\bibitem[che({\natexlab{a}})]{chenerkaggle}
Chemical named entity recognition, {\natexlab{a}}.
\newblock URL \url{https://www.kaggle.com/datasets/abhinavwalia95/chemdner-iob-annotated-chemical-named-etities?select=training.csv}.

\bibitem[dis()]{diseasegit}
Disease-ner.
\newblock URL \url{https://github.com/Megha-Bose/Disease-NER}.

\bibitem[Zhang et~al.(2017)Zhang, Zhong, Chen, Angeli, and Manning]{yzpa-emnlp-17}
Yuhao Zhang, Victor Zhong, Danqi Chen, Gabor Angeli, and Christopher~D. Manning.
\newblock Position-aware attention and supervised data improve slot filling.
\newblock In \emph{Proceedings of EMNLP}, pages 35--45, 2017.

\bibitem[Yao et~al.(2019)Yao, Ye, Li, Han, Lin, Liu, Liu, Huang, Zhou, and Sun]{yyda-acl-19}
Yuan Yao, Deming Ye, Peng Li, Xu~Han, Yankai Lin, Zhenghao Liu, Zhiyuan Liu, Lixin Huang, Jie Zhou, and Maosong Sun.
\newblock Docred: {A} large-scale document-level relation extraction dataset.
\newblock In \emph{Proceedings of ACL}, pages 764--777, 2019.

\bibitem[Gurulingappa et~al.(2012)Gurulingappa, Rajput, Roberts, Fluck, Hofmann{-}Apitius, and Toldo]{hgdo-jbi-12}
Harsha Gurulingappa, Abdul~Mateen Rajput, Angus Roberts, Juliane Fluck, Martin Hofmann{-}Apitius, and Luca Toldo.
\newblock Development of a benchmark corpus to support the automatic extraction of drug-related adverse effects from medical case reports.
\newblock \emph{J. Biomed. Informatics}, 45\penalty0 (5):\penalty0 885--892, 2012.

\bibitem[ppi()]{ppigithub}
Protein-protein interaction extraction (ppi).
\newblock URL \url{https://github.com/BNLNLP/PPI-Relation-Extraction/tree/main/datasets/PPI/type_annotation/Typed_PPI}.

\bibitem[ddi()]{ddigithub}
Drug-drug interaction (ddi).
\newblock URL \url{https://github.com/BNLNLP/PPI-Relation-Extraction/tree/main/datasets/DDI_BLURB}.

\bibitem[gad()]{gadgit}
Genetic association datebase.
\newblock URL \url{https://github.com/BNLNLP/PPI-Relation-Extraction/tree/main/datasets/GAD_BLURB}.

\bibitem[che({\natexlab{b}})]{cheemnlpgit}
Chemical-protein relationship extraction, {\natexlab{b}}.
\newblock URL \url{https://github.com/BNLNLP/PPI-Relation-Extraction/tree/main/datasets/ChemProt_BLURB}.

\bibitem[eve()]{eventldc}
Event trigger detection.
\newblock URL \url{https://catalog.ldc.upenn.edu/LDC2020T13}.

\bibitem[srl()]{srlldc}
Semantic role labeling.
\newblock URL \url{https://catalog.ldc.upenn.edu/LDC2013T19}.

\bibitem[amr()]{amrldc}
Abstract meaning representation.
\newblock URL \url{https://catalog.ldc.upenn.edu/LDC2017T10}.

\bibitem[dpl()]{dpldc}
Dependency parsing.
\newblock URL \url{https://catalog.ldc.upenn.edu/LDC2013T19}.

\bibitem[pos()]{posgithub}
Part-of-speech.
\newblock URL \url{https://github.com/PritK99/POS-Tagging/tree/main}.

\end{thebibliography}
}

\newpage
\appendix
\onecolumn


\section{Extension on General-Bench Dataset}
\label{Extension on Benchmark Dataset}

This part provides an extension to our \texttt{General-Bench} dataset.

\subsection{Evaluation Metrics}
\label{Evaluation Metrics}

\paragraph{Metric List.}
Since all tasks in \texttt{General-Bench} retain their original task definitions without altering the output or prediction format, our evaluation methods vary according to the nature of different tasks and data. 
Table~\ref{tab:metrics-overview} summarizes the evaluation metrics and methods used across all tasks.

{
\fontsize{8.5}{11}\selectfont
\setlength{\tabcolsep}{1mm} 
\captionof{table}{
Overview of all the evaluation metrics in \texttt{General-Bench}.
$\uparrow$ means the higher the better performance, and vice versa for $\downarrow$.
}
\vspace{-3mm}
\label{tab:metrics-overview}
\begin{longtable}[!t]{
p{0.3cm}
p{1.7cm}
>{\centering\arraybackslash}p{1cm}
p{10cm}
p{3cm}}
\hline
\textbf{\#} & \textbf{Metric} & \textbf{Range} & \textbf{Calculation} & \textbf{Representative Tasks} \\
\hline
\endfirsthead
\hline
\textbf{\#} & \textbf{Metric} & \textbf{Range} & \textbf{Calculation} & \textbf{Representative Tasks} \\
\hline
\endhead
\hline
\endfoot

\rowcolor{deepgray} \multicolumn{5}{l}{$\bullet$ \textbf{General }} \\
 1 & Acc$\uparrow$ & [0,1] & Accuracy is defined as the ratio of correctly classified instances to the total number of instances. & Classification \\
\rowcolor{lightgray} 2 & Macro-Acc$\uparrow$ & [0,1] & Macro-Acc evaluates how well a model performs on average across all classes, regardless of class imbalance. & Event Relation Prediction \\
 3 & EM-Acc$\uparrow$ & [0,1] & Exact Match Accuracy evaluates the percentage of predictions that are exactly the same as their corresponding references. & QA, machine translation, or summarization \\
\rowcolor{lightgray} 4 & AP$\uparrow$ & [0,1] & AP, Average Precision, is a metric used to evaluate the performance of object detection tasks, reflecting the overall precision-recall trade-off across multiple thresholds. & Anomaly Detection \\
 5 & mAP $\uparrow$ & [0,1] & mAP, Mean Average Precision, is the mean of Average Precision values across all queries or instances: & 2D/3D Detection \\
\rowcolor{lightgray} 6 & F1$\uparrow$ & [0,1] & F1 score is the harmonic mean of Precision and Recall. & QA \\
7 & Micro-F1$\uparrow$ & [0,1] & Micro-F1 score is the harmonic mean of the Micro-averaged precision and recall. & Classification \\
\rowcolor{lightgray} 8 & AUC$\uparrow$ & [0,1] & AUC is used in binary classification tasks and measures the area under the ROC curve. It represents the model's ability to distinguish between classes. & Image Generation \\
\hline

\rowcolor{deepgray} \multicolumn{5}{l}{$\bullet$ \textbf{Ranking-related}} \\
9 & R@k$\uparrow$ & [0,1] & R@k measures the Recall rate at the top k results in tasks like image retrieval, where the true positive must appear within the top k predicted results. & Image Scene Graph Parsing \\
\rowcolor{lightgray} 10 & AP@k$\uparrow$ & [0,1] & AP@k is the Average Precision calculated at an IoU threshold of k (k<1). This metric is typically used when higher overlap between retrieved items and ground truth items is required. & Object Detection \\
 11 & mAP@k$\uparrow$ & [0,1] & mAP@k refers to the mean Average Precision where the Intersection over Union (IoU) threshold is set to k (k<1). & Object Detection \\
\rowcolor{lightgray} 12 & EM@1$\uparrow$ & [0,1] & Exact Match at 1 evaluates the proportion of instances for which the model's top prediction exactly matches the correct answer. & 3D Question Answering \\
 13 & ANLS$\uparrow$ & [0,1] & ANLS, Average Normalized Levenshtein Similarity, measures how well a model ranks items in a list based on their relevance to a query. & OCR \\
\hline

\rowcolor{deepgray} \multicolumn{5}{l}{$\bullet$ \textbf{Regression-related}} \\
\rowcolor{lightgray} 14 & MAE $\downarrow$ & [0,$\infty$) & MAE, Mean Absolute Error, measures the average of the absolute differences between the predicted values and the actual values. It's typically used in regression tasks. & Object Counting \\
 15 & RMS $\downarrow$ & [0,$\infty$) & RMS, Root Mean Square, is a metric for regression tasks that measures the square root of the average squared differences between the predicted values and true values. & Image Depth Estimation \\
\rowcolor{lightgray} 16 & MSE $\downarrow$ & [0,$\infty$) & MSE, Mean Squared Error, is commonly used for regression tasks and measures the average squared differences between predicted values and actual values. & Object Matting \\
 17 & RMSE $\downarrow$ & [0,$\infty$) & RMSE, Root Mean Squared Error. & Time Series Prediction \\
\hline

\rowcolor{deepgray} \multicolumn{5}{l}{$\bullet$ \textbf{Text Generation-related}} \\
\rowcolor{lightgray} 18 & BLEU-1$\uparrow$ & [0,1] & BLEU-1, Bilingual Evaluation Understudy (1-gram), calculates the precision of unigrams (individual words) in the generated text compared to the reference text(s). & Text Generation \\
 19 & BLEU-4$\uparrow$ & [0,1] & BLEU-4, Bilingual Evaluation Understudy (4-gram). & Text Generation \\
\rowcolor{lightgray} 20 & CodeBLEU$\uparrow$ & [0,1] & CodeBLEU is a metric designed to evaluate the quality of generated code by comparing it to reference code. CodeBLEU combines standard BLEU with additional features specific to code, such as syntax matching, data flow alignment, and weighted n-gram matching. & Code Generation \\
 21 & ROUGE-L$\uparrow$ & [0,1] & ROUGE, Recall-Oriented Understudy for Gisting Evaluation of Longest Common Subsequence, LCS, evaluates text generation tasks, which measures the overlap between the predicted text and the reference text. & Image/Video Captioning \\
\rowcolor{lightgray} 22 & ROUGE-1$\uparrow$ & [0,1] & ROUGE in 1-grams (single words). & Text Generation \\
 23 & CIDEr$\uparrow$ & [0,1] & Consensus-based Image Description Evaluation (CIDEr), evaluates the quality of generated sentences (e.g., image captions) by comparing them against a set of reference sentences. & Captioning \\
\hline

\rowcolor{deepgray} \multicolumn{5}{l}{$\bullet$ \textbf{Image-related}} \\
\rowcolor{lightgray} 24 & PSNR$\uparrow$ & [0,$\infty$) & PSNR, Peak Signal-to-Noise Ratio, measures the quality of a reconstructed or compressed signal, such as images or videos, compared to the original signal. & Image Desnowing \\
 25 & MS-SSIM$\uparrow$ & [-1,1] & MS-SSIM, Multi-Scale Structural Similarity Index Measure, is a metric used for image quality assessment that measures the structural similarity between two images across multiple scales. & Document Image Unwarping \\
\rowcolor{lightgray} 26 & CLIP-Score$\uparrow$ & [0,1] & The CLIP score measures the similarity between an image and a textual description using the CLIP model, commonly used for image-text matching tasks. & Image Editing \\
 27 & FID $\downarrow$ & [0,$\infty$) & FID, Fr´echet Inception Distance, measures the distance between two multivariate Gaussian distributions: one representing the features of real images and the other representing the features of generated images. & Text-to-Image Generation \\
\rowcolor{lightgray} 28 & SAD$\downarrow$ & [0,$\infty$) & SAD, Sum of Absolute Differences, measures the total absolute difference between the predicted and ground truth values for all pixels in an image or region. It is typically used in image matting tasks to assess how closely the model replicates fine-grained image details such as edges and textures. & Object Matting \\
\hline

\rowcolor{deepgray} \multicolumn{5}{l}{$\bullet$ \textbf{Video-related}} \\
 29 & Fram-Acc$\uparrow$ & [0,1] & Frame-level Accuracy evaluates how many frames (or time steps) in the sequence are correctly classified by comparing predictions with the ground truth on a frame-by-frame basis. & Video Translation \\
\rowcolor{lightgray} 30 & FVD $\downarrow$ & [0,$\infty$) & Fréchet Video Distance (FVD) is a metric used to evaluate the quality of generated video sequences, extending the principles of the FID to videos. FVD calculates the distance between the feature distributions of real and generated videos, taking into account both spatial and temporal dynamics. & Video Generation \\
 31 & MUSIQ$\uparrow$ & [0,1] & MUSIQ, Multi-scale Image Quality, quantifies the distance between the feature distributions of real and generated videos. & Video Superresolution \\
\rowcolor{lightgray} 32& absRel $\downarrow$ & [0,$\infty$) & absRel (Absolute Relative Error) is a metric commonly used in depth estimation tasks, which measures the average ratio of the prediction error to the ground-truth depth. & Video Depth Estimation \\
 33 & EPE$\downarrow$ & [0,$\infty$) & End-Point Error (EPE) is a metric commonly used inoptical flowtasks to evaluate the accuracy of predicted motion vectors (flow) between consecutive frames in a video or image sequence. It measures the Euclidean distance between the predicted and ground truth flow vectors at each pixel, providing an average error across the image. & Optical Flow \\
\rowcolor{lightgray} 34 & DINO-Score$\uparrow$ & [0, 1] & The DINO Score is calculated as the cosine similarity between the DINOv2 class embedding of two frames, effectively measuring the consistency of a subject's identity across frames. According to the DreamBooth paper, the DINO Score captures more detailed aspects of subject identity compared to the CLIP Score. & Subject-Driven Image Generation \\
 35 & L1-Dis$\uparrow$ & [0, 1] & L1-Dis measures the L1 distance between two consecutive frames, evaluating the model's ability to generate static (temporally stable) videos by quantifying the differences between adjacent frames. $\mathcal{L}_1 = \frac{1}{N} \sum_{i=1}^{N}(\frac{1}{T-1}\sum_{t=1}^T \frac{\mathcal{L}_1(f_i^t, f_i^{t+1}}{P})$, $\text{L1-Dis} = \frac{255-\mathcal{L}_1}{255}$. & Static Video Generation \\
\rowcolor{lightgray} 36 & OFS$\uparrow$ & [0, 1] & OFS, Optical Flow Score, captures the movement of pixels between two consecutive frames. By applying a threshold to identify "moving pixels" based on the optical flow, we calculate the ratio of these moving pixels. This ratio quantifies the degree of dynamism in the generated videos. & Dynamic Video Generation \\
 37 & Aesth-Score$\uparrow$ & [0, 1] & The aesthetic score (Aesth-score) is calculated using a ViT with a linear head, as implemented in an improved aesthetic predictor. It has a similar calculation as in CLIP score. & Artistic Content Text-to-Video Generation \\
\rowcolor{lightgray} 38 & Suc-Rate$\uparrow$ & [0, 1] & Successful Rate (Suc-Rate) measures the performance of Video Generation. Suc-Rate leverages an open-vocabulary detector to identify the subjects specified in the text prompts. If all the subjects in a text prompt are successfully detected, it classifies the corresponding frame as a successfully generated video frame, and then calculates the ratio of successful frames to the total number of generated frames. & Multi-Class-Conditioned Text-to-Video Generation, Spatial Relation Video Generation, Camera Motion Generation \\
 39 & ViCLIP-Score$\uparrow$ & [0, 1] & We calculate the cosine similarity between the ViCLIP embeddings of text prompts and videos. Compared to the CLIP Image model, ViCLIP enables a more comprehensive assessment of the video's style and its overall consistency with the text prompts. & Image-to-Video Generation \\
\rowcolor{lightgray} 40 & Avg(DINO+\newline CLIP+OFS\newline+MSS)$\uparrow$ & [0, 1] & For the image-to-video generation task, we conduct a comprehensive evaluation based on Subject Consistency, Background Consistency, Motion Smooth and Dynamic Degree. We utilize the DINO-Score and CLIP-Score to assess subject and background consistency, reflecting the model's ability to adhere to the image prompt. To evaluate the motion smoothness, we measure the Motion Smooth Score (MSS) of the frame-by-frame motion prior via video frame interpolation models. Also, we employ the Optical Flow Score to measure the dynamic degree of the generated videos, ensuring that our metrics do not favor static videos. & Image-to-Video Generation \\
\hline

\rowcolor{deepgray} \multicolumn{5}{l}{$\bullet$ \textbf{3D-related}} \\
 41 & AMOTA $\uparrow$ & [0,1] & AMOTA (Average Multi-Object Tracking Accuracy) is a performance metric used to evaluate multi-object tracking (MOT) systems. It combines detection accuracy and tracking performance by assessing how well a system detects, associates, and tracks multiple objects over time. AMOTA is calculated by averaging tracking accuracy over a range of thresholds for the Intersection over Union (IoU) or matching criteria. & 3D Tracking \\
\rowcolor{lightgray} 42 & RTE $\downarrow$ & [0,$\infty$) & Relative Translation Error (RTE) is a metric commonly used in robotics, pose estimation, and SLAM (Simultaneous Localization and Mapping) tasks. It evaluates the accuracy of a system's estimated translation (movement) compared to the ground truth, typically in scenarios where spatial accuracy is crucial. & 3D Pose Estimation \\
 43 & CD $\downarrow$ & [0,$\infty$) & Chamfer Distance (CD) is a metric widely used in 3D geometry processing, point cloud generation, and shape matching tasks. It measures the similarity between two sets of points (e.g., two point clouds) by quantifying the average closest-point distance between them. CD is particularly useful for evaluating the alignment and fidelity of reconstructed or generated 3D shapes compared to ground truth data. & Point Cloud Generation \\
\hline

\rowcolor{deepgray} \multicolumn{5}{l}{$\bullet$ \textbf{Segmentation\&Detection-related}} \\
\rowcolor{lightgray} 44 & mIoU$\uparrow$ & [0,1] & mIoU, Mean Intersection over Union, calculates the ratio of the intersection area to the union area between the predicted and ground truth segmentation masks for a single class. & Image Semantic Segmentation \\
 45 & m\_vIoU$\uparrow$ & [0,1] & m\_vIoU measures the spatiotemporal overlap between predicted and ground-truth object regions across multiple video frames & Temporal Action Detection \\
\rowcolor{lightgray} 46 & Inst-mIoU$\uparrow$ & [0,1] & Inst-mIoU computes the average IoU score for all part instances across a dataset. It ensures that both over-segmentation and under-segmentation errors are penalized, focusing on instance-level segmentation quality. & 3D Part Segmentation \\
 47 & PQ$\uparrow$ & [0,1] & PQ, Panoptic Quality, is used for panoptic segmentation tasks, combining both segmentation quality and detection quality. It is a comprehensive metric that evaluates both pixel-level segmentation and object detection quality. & Panoptic Segmentation \\
\rowcolor{lightgray} 48 & DICE$\uparrow$ & [0,1] & DICE, Dice Similarity Coefficient, is a metric commonly used in image segmentation tasks, measuring the similarity between the predicted segmentation and the ground truth segmentation. & Bone Fracture Detection \\
 49 & S-measure$\uparrow$ & [0,1] & Structure Measure (S-measure) is a metric designed to evaluate the structural similarity between a predicted binary map (e.g., an object mask) and a ground truth binary map. It balances both region-level and boundary-level consistency, ensuring that the evaluation captures holistic structural integrity and fine-grained details. & Video Object Detection \\
\hline

\rowcolor{deepgray} \multicolumn{5}{l}{$\bullet$ \textbf{Audio-related}} \\
\rowcolor{lightgray} 50 & CLAP$\uparrow$ & [0,1] & CLAP (Contrastive Language-Audio Pretraining) evaluates the alignment between generated audio and text. It is derived from a contrastive learning framework where embeddings of audio and text are trained to be close in a shared latent space if they are semantically related. & Audio Editing \\
 51 & Style-CLAP  $\uparrow$ & [0,1] & Style-CLAP calculates the CLAP cosine similarity between the generated Mel spectrograms and the corresponding textual description of the style to evaluate style fit. & Music Style Transfer \\
\rowcolor{lightgray} 52 & MCD $\downarrow$ & [0,$\infty$) & Mel-cepstral distortion (MCD) measures the spectral distance between the mel-cepstral coefficients (MCCs) of generated speech and reference speech, providing an indication of how closely the generated speech resembles the reference in terms of acoustic characteristics. & Speech Synthesis \\
 53 & WER $\downarrow$ & [0,1] & WER (Word Error Rate) measures the percentage of errors in the transcribed output compared to the reference transcription. & TTS \\
\rowcolor{lightgray} 54 & FAD $\downarrow$ & [0,$\infty$) & Frechet audio distance (FAD) evaluates the quality and realism of generated audio, and measures the similarity between the distribution of features obtained by VGGish in generated audio and those in a set of real (reference) audio samples.  & Video-to-Audio \\
 55 & PCC  $\uparrow$ & [0,1] & Pitch-Class Consistency (PCC) is a metric used in the evaluation of generated music to assess how consistent the pitch classes (e.g., notes) are across pairs of bars in a piece of music. It measures the overlapping area between the pitch-class histograms of different bars, ensuring that the generated music maintains harmonic coherence. & Music Generation \\
\hline

\rowcolor{deepgray} \multicolumn{5}{l}{$\bullet$ \textbf{Human-aware Evaluation}} \\
\rowcolor{lightgray} 56 & UPR $\uparrow$ & [0,1] & UPR, User Preference Rates, UPR measures the proportion of times a particular system or model is preferred over alternatives in a set of user evaluations. It reflects the subjective preferences of users and is often derived from pairwise comparisons or ranking experiments. & Video Style Transfer \\
 57 & MOS $\uparrow$ & [1,5] & Mean Opinion Score (MOS), in which human raters listen to synthesized speech and assess its naturalness, quality, and intelligibility using a 5-point Likert scale. & Speech Generation \\
\rowcolor{lightgray} 58 & GPT-Score $\uparrow$ & [0,1] & GPT-Score evaluates the instruction following rate with GPT assistance, as an alternative to human evaluation. & Audio Question Answering \\
\hline
\end{longtable}
}

\paragraph{Mapping Functions of Scoring Metric.}
Most task evaluation scores, despite utilizing different metrics, fall within a 0-100\% range, such as F1, Accuracy (Acc), and ROUGE-L, and follow a monotonically increasing trend.  
However, certain task metrics produce scores outside this range. 
For example, regression-related metrics, as well as FID, FVD, and similar metrics, range from 0 to infinity and follow a monotonically decreasing trend. 
In contrast, MOS scores are represented as a discrete 5-point scale.  
Due to these varying score ranges across tasks, it becomes intractable to normalize them to a unified scale for level score calculations.  
Thus, we design the following mapping functions to standardize these metrics into a 1-100\% range, thereby streamlining the computation of level scoring algorithms.

\begin{compactitem}

    \item Normalizing \textbf{MAE}:
    \[
    y = 2 \times \text{sigmoid}\left(\frac{50}{x}\right) -1 , \quad \text{where } x \in [0, +\infty), \quad y \in (0, 1).
    \]

    \item Normalizing \textbf{RMS}:
    \[
    y = 2 \times \text{sigmoid}\left(\frac{50}{x}\right) -1, \quad \text{where } x \in [0, +\infty), \quad y \in (0, 1).
    \]

    \item Normalizing \textbf{MSE}:
    \[
    y = 2 \times \text{sigmoid}\left(\frac{5}{x}\right) -1, \quad \text{where } x \in [0, +\infty), \quad y \in (0, 1).
    \]

    \item Normalizing \textbf{RMSE}:
    \[
    y = 2 \times \text{sigmoid}\left(\frac{5}{x}\right) -1, \quad \text{where } x \in [0, +\infty), \quad y \in (0, 1).
    \]

    \item Normalizing \textbf{absRel}:
    \[
    y = 2 \times \text{sigmoid}\left(\frac{0.1}{x}\right) -1, \quad \text{where } x \in [0, +\infty), \quad y \in (0, 1).
    \]

    \item Normalizing \textbf{EPE}:
    \[
    y = 2 \times \text{sigmoid}\left(\frac{1}{x}\right) -1, \quad \text{where } x \in [0, +\infty), \quad y \in (0, 1).
    \]

    \item Normalizing \textbf{FID}:
    \[
    y = 2 \times \text{sigmoid}\left(\frac{25}{x}\right) -1, \quad \text{where } x \in [0, +\infty), \quad y \in (0, 1).
    \]

    \item Normalizing \textbf{FVD}:
    \[
    y = 2 \times \text{sigmoid}\left(\frac{100}{x}\right) -1, \quad \text{where } x \in [0, +\infty), \quad y \in (0, 1).
    \]

    \item Normalizing \textbf{FAD}:
    \[
    y = 2 \times \text{sigmoid}\left(\frac{10}{x}\right) -1, \quad \text{where } x \in [0, +\infty), \quad y \in (0, 1).
    \]

    \item Normalizing \textbf{PSNR}:
    \[
    y = \text{tanh}\left( \frac{x}{20} \right), \quad \text{where } x \in [0, +\infty), \quad y \in [0, 1).
    \]

    \item Normalizing \textbf{SAD}:
    \[
    y = 2 \times \text{sigmoid}\left(\frac{10}{x}\right) -1, \quad \text{where } x \in [0, +\infty), \quad y \in (0, 1).
    \]

    \item Normalizing \textbf{RTE}:
    \[
    y = 2 \times \text{sigmoid}\left(\frac{0.5}{x}\right) -1, \quad \text{where } x \in [0, +\infty), \quad y \in (0, 1).
    \]

    \item Normalizing \textbf{CD}:
    \[
    y = 2 \times \text{sigmoid}\left(\frac{1}{x}\right) -1, \quad \text{where } x \in [0, +\infty), \quad y \in (0, 1).
    \]

    \item Normalizing \textbf{MCD}:
    \[
    y = 2 \times \text{sigmoid}\left(\frac{5}{x}\right) -1, \quad \text{where } x \in [0, +\infty), \quad y \in (0, 1).
    \]

    \item Normalizing \textbf{WER}:
    \[
    y = 1- x, \quad \text{where } x \in [0,1], \quad y \in [0, 1].
    \]

    \item Normalizing \textbf{MS-SSIM}:
    \[
    y = \frac{\left(x + 1\right)}{2}, \quad \text{where } x \in [-1,1], \quad y \in [0, 1].
    \]

    \item Normalizing \textbf{MOS}:
    \[
    y = \frac{x-1}{4}, \quad \text{where } x \in [1,5], \quad y \in [0,1].
    \]

\end{compactitem}

\clearpage

\subsection{Data Format}

\begin{figure*}[!h]
\centering
\includegraphics[width=0.96\linewidth]{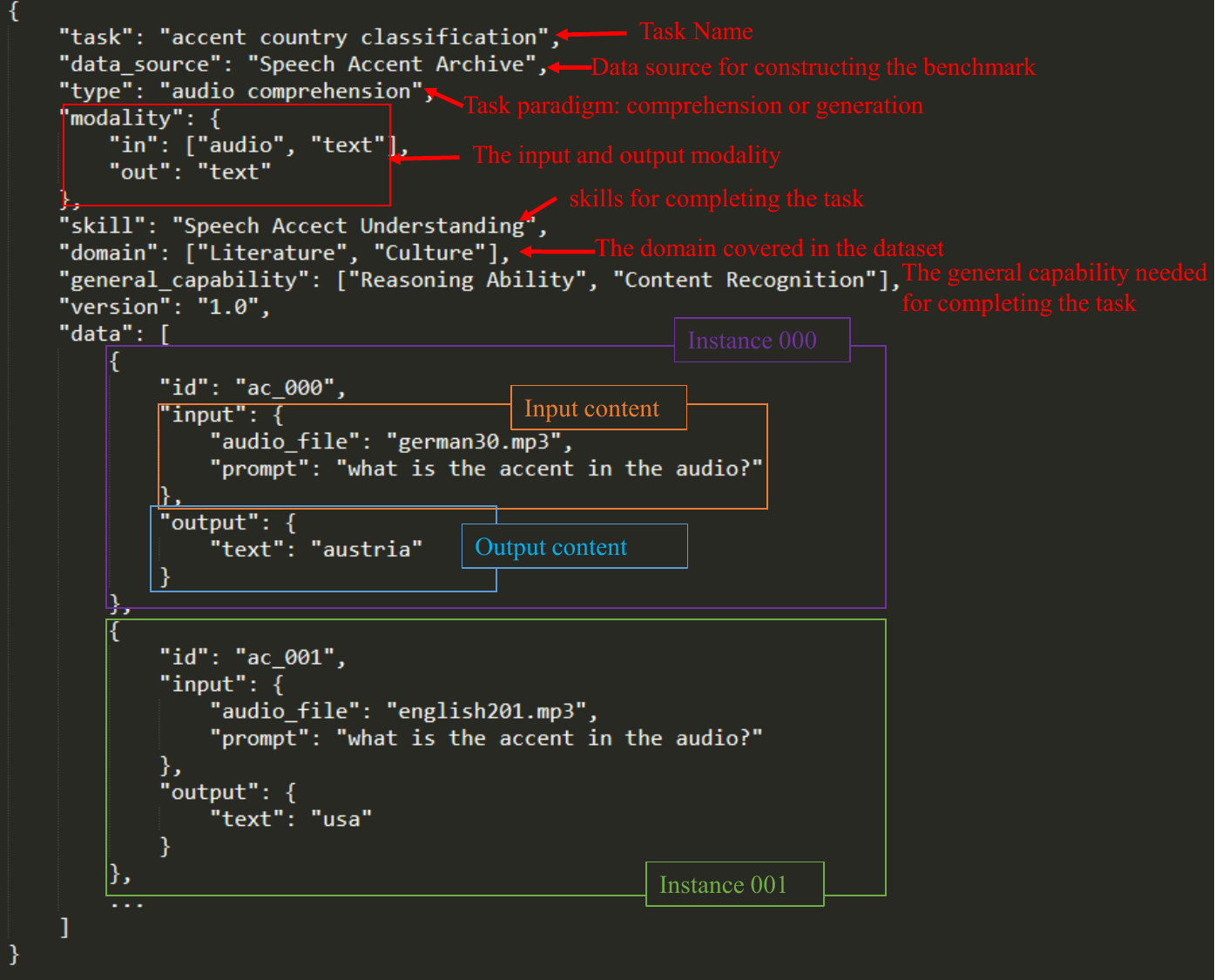}
\caption{An illustrative example of file formats.}
\label{fig:data-format-1}
\vspace{-2mm}
\end{figure*}

\begin{figure*}[!h]
\centering
\includegraphics[width=0.60\linewidth]{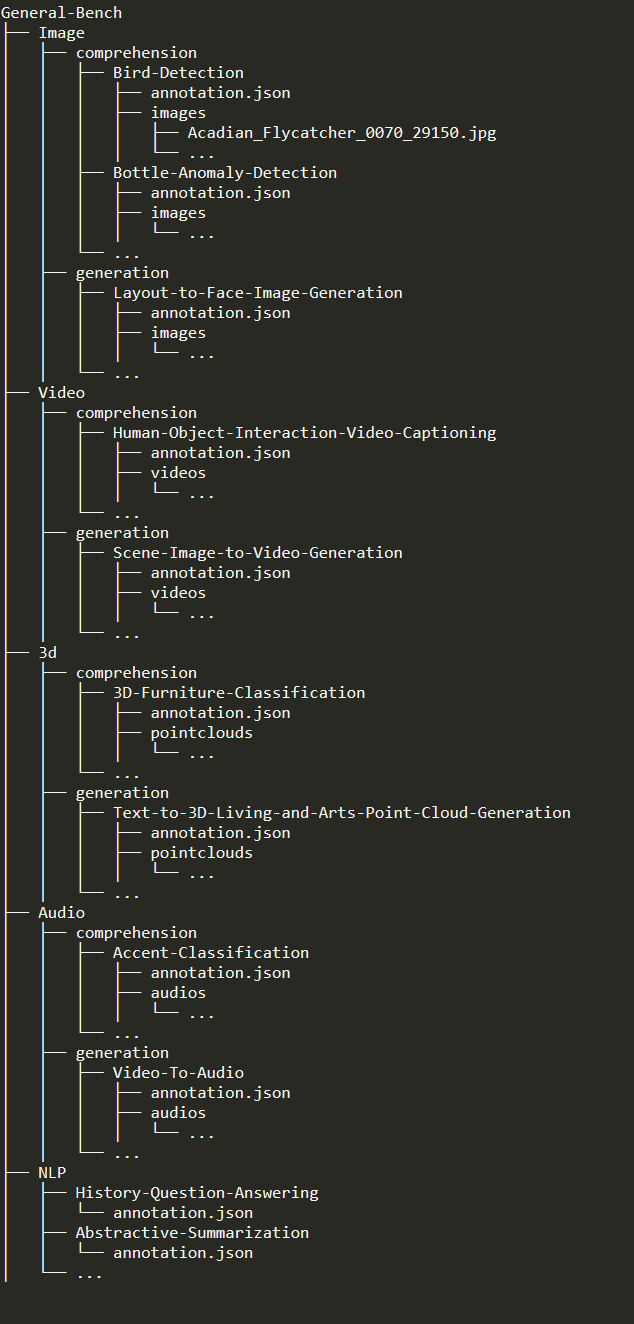}
\caption{The organization structure of the file system.}
\label{fig:data-format-2}
\vspace{-2mm}
\end{figure*}

To provide a comprehensive understanding of how to utilize our benchmark data, we present examples illustrating how the data files are stored and organized. 
Figure~\ref{fig:data-format-1} displays the code snippets showcasing the structures of some representative tasks.
Figure~\ref{fig:data-format-2} illustrates how we organize the benchmark datasets in the file system.

\clearpage

\subsection{Data Taxonomy and Hierarchy}
\label{Overall Data Taxonomy and Hierarchy}

\begin{figure*}[!t]
\centering
\includegraphics[width=0.99\linewidth]{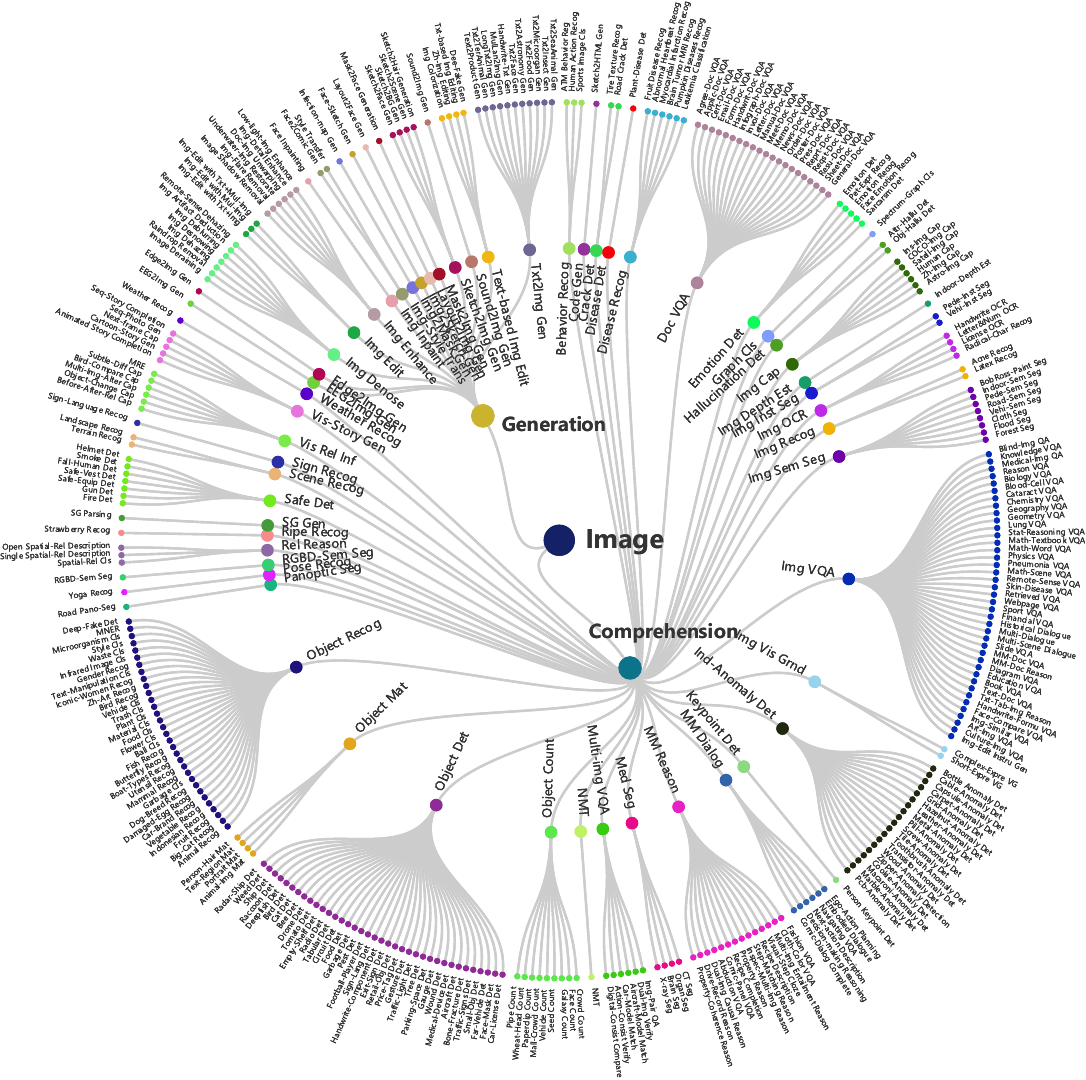}
\vspace{-2mm}
\caption{Taxonomy and hierarchy of data in terms of Image modality.}
\label{fig:tree-map-img}
\vspace{-4mm}
\end{figure*}

\begin{figure*}[!t]
\centering
\includegraphics[width=0.99\linewidth]{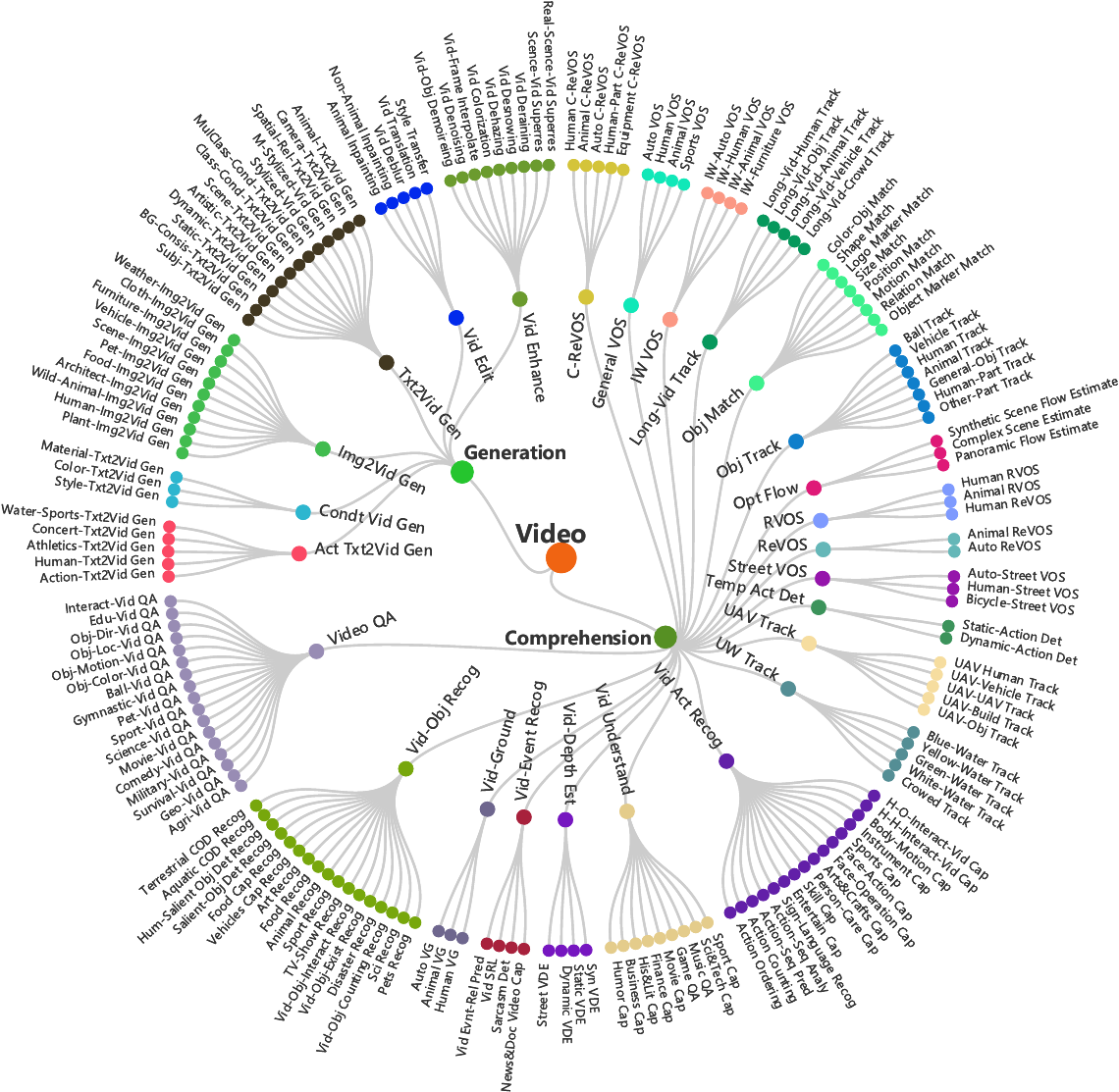}
\vspace{-2mm}
\caption{Taxonomy and hierarchy of data in terms of Video modality.}
\label{fig:tree-map-vid}
\vspace{-2mm}
\end{figure*}

\begin{figure*}[!t]
\centering
\includegraphics[width=0.94\linewidth]{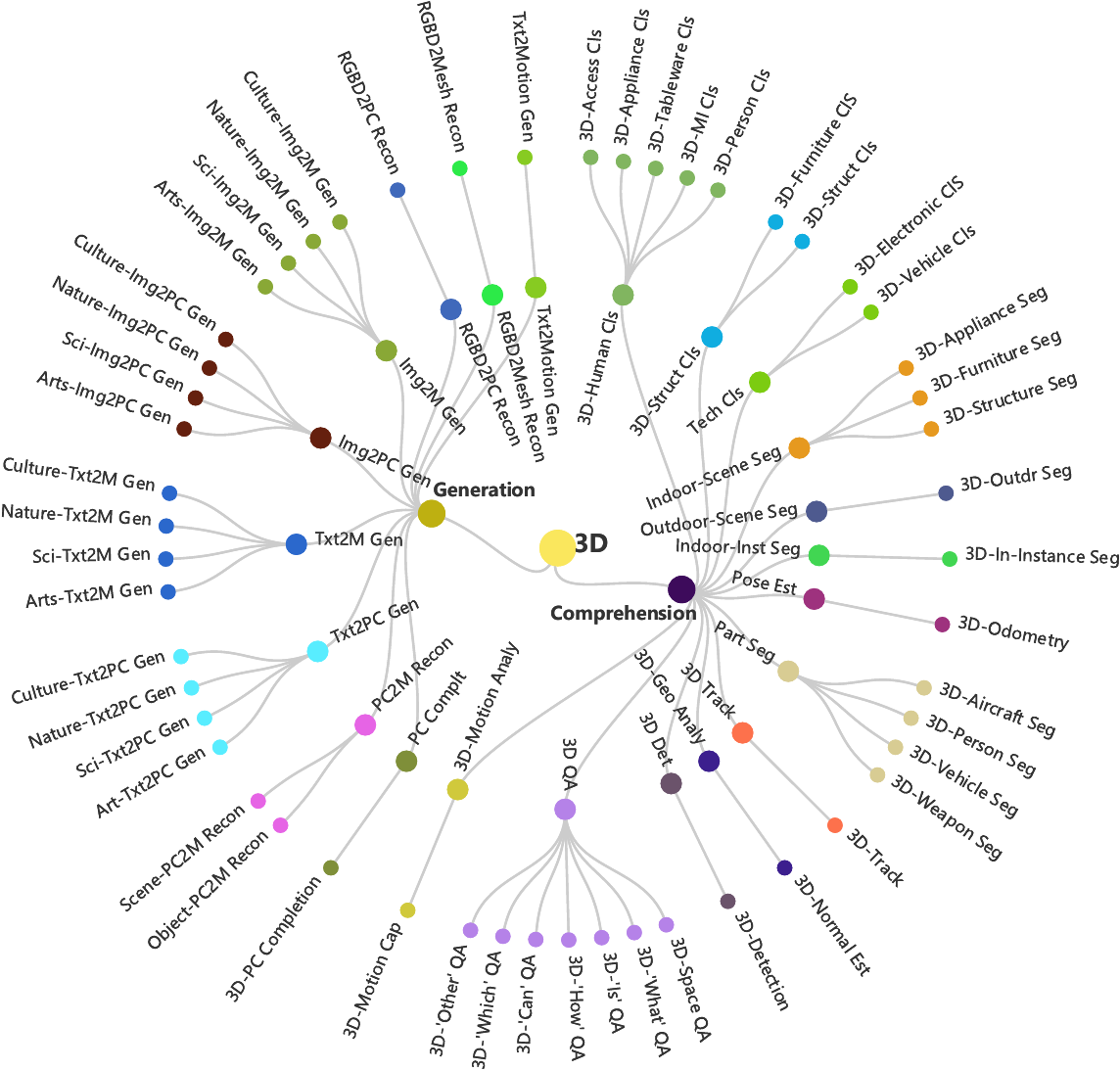}
\vspace{-2mm}
\caption{Taxonomy and hierarchy of data in terms of 3D modality.}
\label{fig:tree-map-3d}
\vspace{-2mm}
\end{figure*}

\begin{figure*}[!t]
\centering
\includegraphics[width=0.95\linewidth]{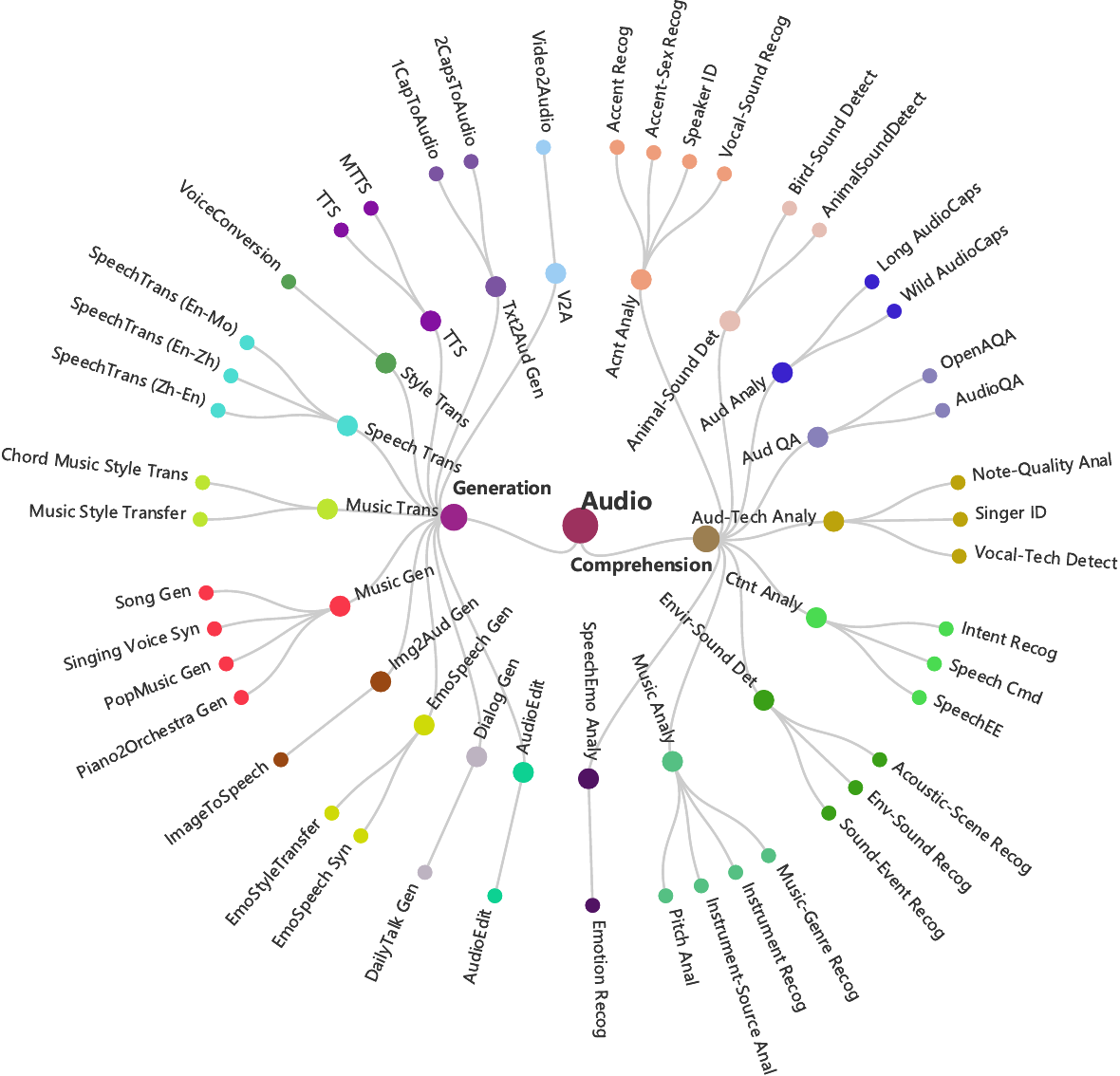}
\vspace{-2mm}
\caption{Taxonomy and hierarchy of data in terms of Audio modality.}
\label{fig:tree-map-aud}
\vspace{-2mm}
\end{figure*}

\begin{figure*}[!t]
\centering
\includegraphics[width=0.98\linewidth]{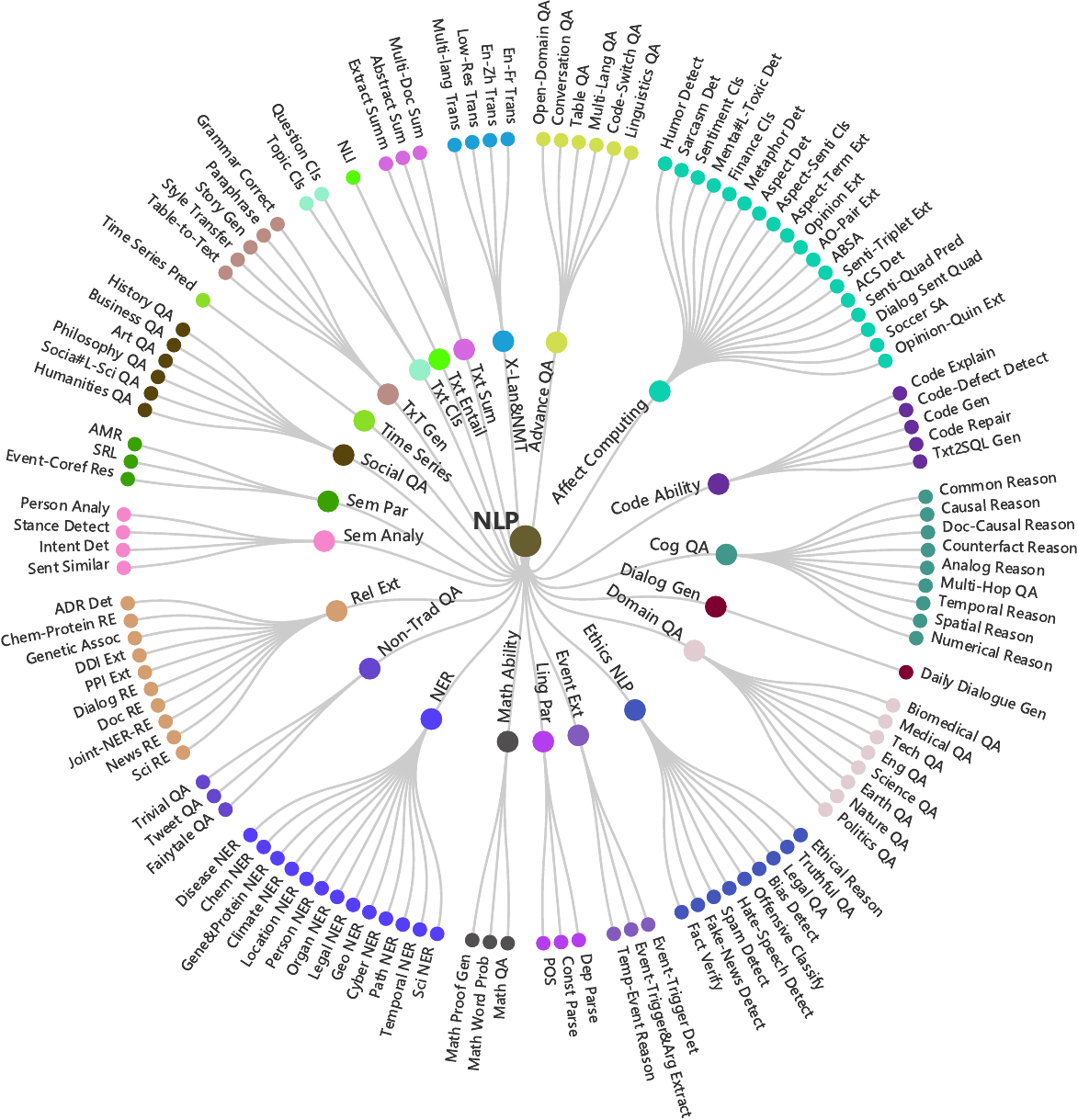}
\vspace{-2mm}
\caption{Taxonomy and hierarchy of data in terms of Language modality.}
\label{fig:tree-map-nlp}
\vspace{-2mm}
\end{figure*}

We visualize a comprehensive hierarchical taxonomy of our benchmark. 
Due to space constraints, we have separately illustrated the taxonomy for five major modalities in Figure~\ref{fig:tree-map-img}, Figure~\ref{fig:tree-map-vid}, Figure~\ref{fig:tree-map-3d}, Figure~\ref{fig:tree-map-aud}, and Figure~\ref{fig:tree-map-nlp}, respectively.
Each visualization includes comprehension and generation paradigms, skills (meta-tasks), and specific tasks for the respective modality.

\clearpage

\clearpage

\clearpage

\subsection{Data Distributions}
\label{Data Specification and Distributions}

\paragraph{Capability.}

\begin{figure*}[!h]
\centering
\includegraphics[width=0.9\linewidth]{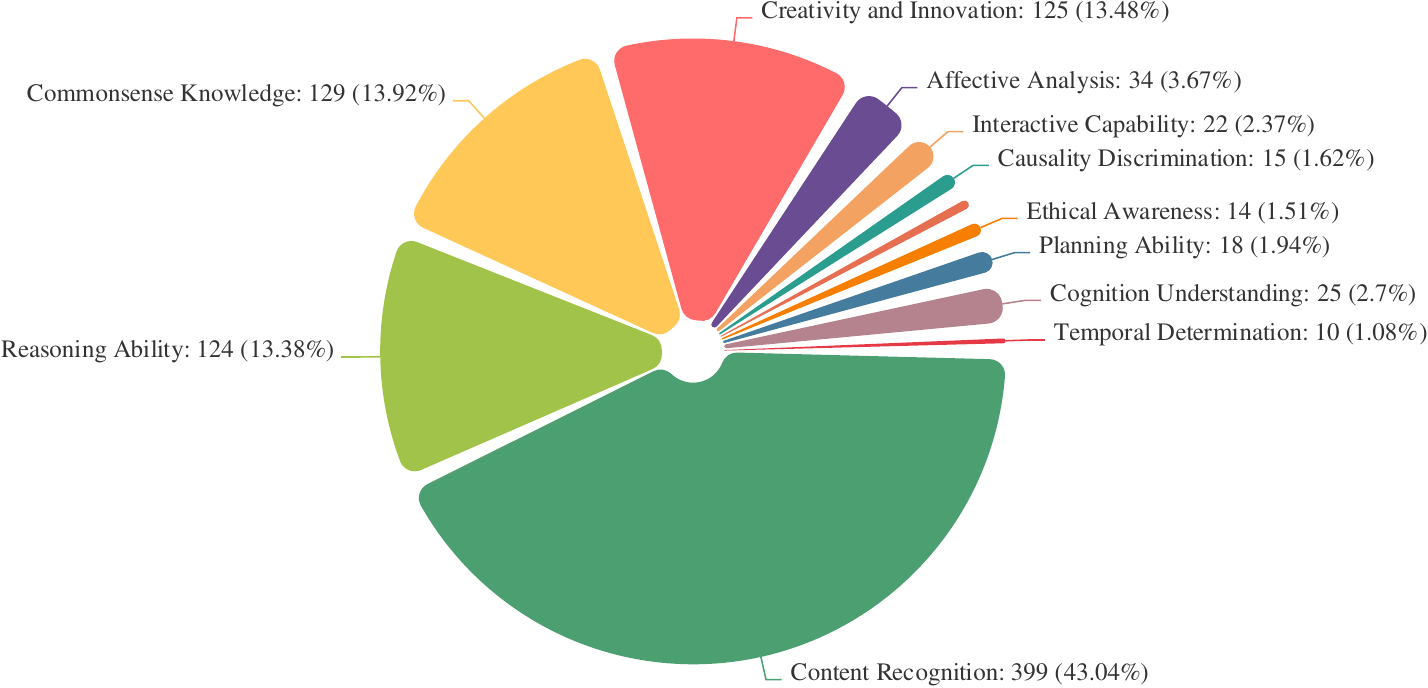}
\caption{Distribution of various capabilities evaluated in \texttt{General-Bench}.}
\label{fig:static-capability}
\vspace{-2mm}
\end{figure*}

In Figure~\ref{fig:static-capability}, we present the distribution of capability evaluations across all tasks in \texttt{General-Bench}. These capabilities include:  
Content Recognition, Commonsense Understanding, Reasoning Ability, Causality Discrimination, Affective Analysis, Problem Solving, Creativity and Innovation, Interactive Capability, and others.
As observed, the majority of tasks focus on Content Recognition or perception-related abilities. 
This emphasis aligns with the current stage of MLLM development, where models are not yet equipped with highly advanced cognitive capabilities. 
We plan to continuously update the benchmark in the future to accommodate the evolving strengths of more powerful models.

\begin{figure*}[!h]
\centering
\includegraphics[width=0.99\linewidth]{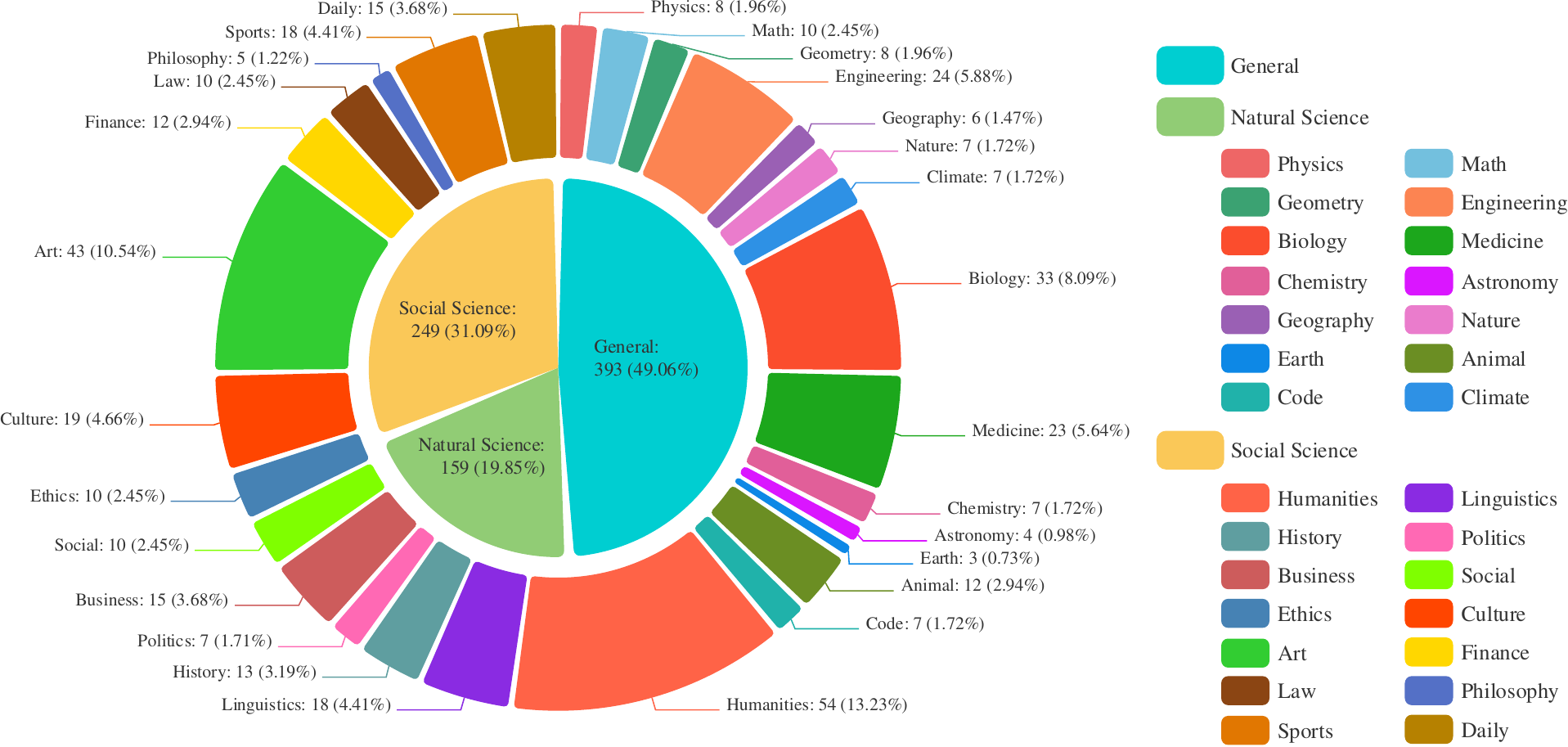}
\caption{Distribution of various domains and disciplines covered by \texttt{General-Bench}.}
\label{fig:static-domain}
\vspace{-2mm}
\end{figure*}

\paragraph{Domains and Discipline.}

Figure~\ref{fig:static-domain} illustrates the domains and disciplines covered by our benchmark. While the majority of tasks belong to the general domain, the benchmark also encompasses significant fields from both Physical Sciences and Social Sciences.
For Physical Sciences, the benchmark includes disciplines such as Physics, Geometry, Biology, Medicine, Chemistry, Astronomy, and Geography.
For Social Sciences, it spans areas including Humanities, Linguistics, History, Politics, Culture, Art, and Economics.
In other words, our benchmark is designed to evaluate the capabilities of MLLMs across a wide range of scientific fields and domains. 
This ensures the broad evaluative advantage of our benchmark, enabling comprehensive assessment of multimodal generalist models.

\paragraph{Comprehension vs. Generation.}

\begin{figure*}[!h]
\centering
\includegraphics[width=0.78\linewidth]{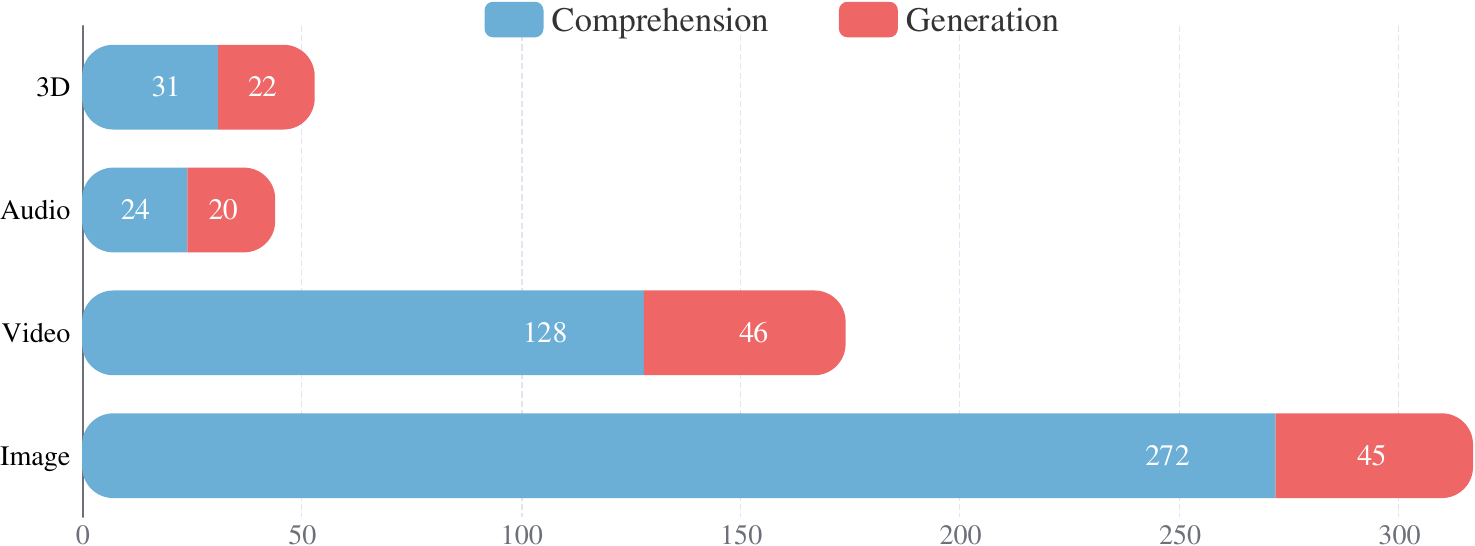}
\caption{Distribution of various domains and discipline covered by \texttt{General-Bench}.}
\label{fig:static-comp-gen}
\vspace{-2mm}
\end{figure*}

We illustrate the task distribution across the two critical paradigms, Comprehension and Generation, in Figure~\ref{fig:static-comp-gen}. 
Currently, the majority of tasks are centered on comprehension, which aligns with the present capabilities of most MLLMs.

\paragraph{Modality.}

\begin{figure*}[!h]
\centering
\includegraphics[width=0.72\linewidth]{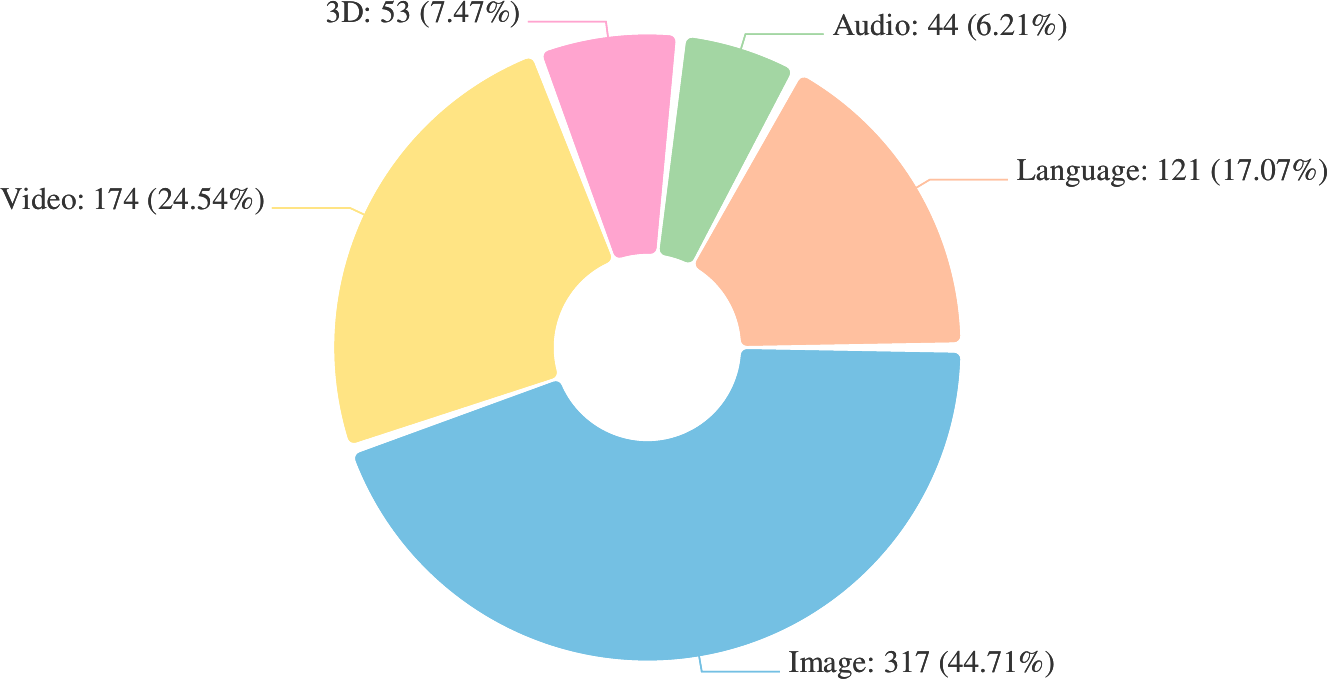}
\caption{Distribution of different modalities covered in \texttt{General-Bench}.}
\label{fig:static-modality-distribution}
\vspace{-2mm}
\end{figure*}

Finally, we present the distribution of tasks across different modalities in Figure~\ref{fig:static-modality-distribution}.
Overall, the image modality constitutes the largest proportion of tasks.
We note that beyond the five major modalities—Image, Video, 3D (3D-RGB and Point-Cloud), Audio, and Language—our benchmark also includes tasks in other modalities such as Time Series, Depth, Infrared, Spectrogram, Radar, Code, Document, and Graph. These additional modalities play important roles in specific domains. However, due to the limited number of tasks in these modalities, we have merged and classified their data under broader categories like Image and Language for ease of management.

\newpage

\subsection{Comparisons with Existing Benchmarks}
\label{Full Data Statistics}

Table~\ref{tab:data-statistics-transposed-complete} provides a comprehensive comparison of \texttt{General-Bench} with existing MLLM benchmarks. Compared to these benchmarks, \texttt{General-Bench} demonstrates absolute superiority across all comparison aspects. 
For instance, it is the first MLLM benchmark to cover nearly all commonly used modalities while emphasizing both Comprehension and Generation as critical evaluation paradigms. Most notably, \texttt{General-Bench} boasts the largest dataset, encompassing 550 tasks with over 275K instances. 
Also, it evaluates the largest number of MLLMs tested to date, setting a new standard for benchmark comprehensiveness and scale.

{
\fontsize{8}{9}\selectfont
\setlength{\tabcolsep}{0.4mm}
\centering
\captionof{table}{
A full comparison of \texttt{\textbf{General-Bench}} with other popular MLLM benchmarks.
}
\vspace{-3mm}
\label{tab:data-statistics-transposed-complete}
\begin{longtable}[!t]{lccccccccc>{\centering\arraybackslash}p{1.5cm}}
\hline
\diagbox{\textbf{Bench.}}{\textbf{Aspect}}
& \textbf{Modality} & \makecell{\bf Task\\\textbf{Scheme}} & \textbf{\#Domain} & \textbf{\#Skill} & \textbf{\#Task} & \textbf{\#Sample} & \makecell{\bf Answer\\\textbf{Form}} & \textbf{\#Metric} & \textbf{Annotation} & \makecell{\bf \#Tested\\\textbf{Models}} \\
\midrule
\endfirsthead
\hline
\diagbox{\textbf{Bench.}}{\textbf{Aspect}}
& \textbf{Modality} & \makecell{\bf Task\\\textbf{Scheme}} & \textbf{\#Domain} & \textbf{\#Skill} & \textbf{\#Task} & \textbf{\#Sample} & \makecell{\bf Answer\\\textbf{Form}} & \textbf{\#Metric} & \textbf{Annotation} & \makecell{\bf \#Tested\\\multicolumn{1}{c}{\textbf{Model}} } \\
\midrule
\endhead
\hline
\endfoot

\rowcolor{mydeepblue} \multicolumn{11}{l}{$\bullet$ \textbf{\large Science\&Discipline}}\\
\addlinespace[4pt]

\makecell{\textbf{ScienceQA}\\ \cite{lu2022learn}} & Txt,Img & Comp. & 12 & 1 & / & 21K & MC-QA & Acc. & Repurposed & 10 \\
\addlinespace[4pt]

\rowcolor{lightgraygrey} \makecell{\textbf{VisAidMath}\\ \cite{ma2024visaidmath}} & Txt,Img & Comp. & 1 & 1 & 1 & 1.2K & MC-QA & Acc. & Repurposed & 10 \\
\addlinespace[4pt]

\makecell{\textbf{MMMU}\\ \cite{yue2024mmmu}} & Txt, Img & Comp. & 6 & 6 & 30 & 11.5K & MC-QA & Acc. & Manual & 24 \\
\addlinespace[4pt]

\rowcolor{lightgraygrey} \makecell{\textbf{NaturalBench}\\ \cite{li2024naturalbench}} & Txt,Img & Comp. & 1 & / & 27 & 7.6K & MC-QA & Acc. & Repurposed & 32 \\
\addlinespace[4pt]

\hline
\rowcolor{bg-tb-heavey-audio} \multicolumn{11}{l}{$\bullet$ \textbf{\large Audio}}\\
\addlinespace[4pt]

\makecell{\textbf{AudioBench} \\ \cite{AudioBench-abs-2406-16020}} & Txt, Aud & Comp. & / & 3 & 8 & 100K  & Free-Form & \makecell{WER,METEOR\\Llama-score} & Repurposed & 4 \\
\addlinespace[4pt]

\rowcolor{lightgraygrey} \makecell{\textbf{MARBLE} \\ \cite{MARBLE-YuanMLZCYZLHTDW23}} & Txt, Aud & Comp. & / & 12 & 18 & / & Free-Form & Origin(19) & Repurposed & 0 \\
\addlinespace[4pt]

\makecell{\textbf{MMAU} \\ \cite{MMAU-abs-2410-19168}} & Txt, Aud & Comp. & / & 12 &  27 & 10K & MC-QA & Acc. & \makecell{Repurposed \\ Manual } & 14 \\
\addlinespace[4pt]

\hline
\rowcolor{bg-tb-heavey-3D} \multicolumn{11}{l}{$\bullet$ \textbf{\large 3D}}\\
\addlinespace[4pt]

\makecell{\textbf{T$^3$Bench} \\ \cite{T3Bench-abs-2310-02977}} & Txt, 3D & Gen. & / & 3 & 3 & 300 & Free-From & Origin(2) & Repurposed & 0 \\
\addlinespace[4pt]

\makecell{\textbf{3DBench} \\ \cite{3DBench-0002HHG024}} & Txt, 3D & Comp. & / & 12 & 12 & 8K & Free-From & Origin(6) & Repurposed & 3 \\
\addlinespace[4pt]

\hline
\rowcolor{bg-tb-heavey-video} \multicolumn{11}{l}{$\bullet$ \textbf{\large Video}}\\
\addlinespace[4pt]

\makecell{\textbf{Video-Bench}  \\ \cite{Video-Bench-abs-2311-16103}} & Txt,Vid & Comp. & / & 3 & 7 & 17K  & MC-QA & Acc. & Repurposed & 8 \\
\addlinespace[4pt]

\rowcolor{lightgraygrey} \makecell{\textbf{VideoMME}\\ \cite{Video-MME}} & Txt,Vid & Comp. & 6 & 12 & 12 & 900 & MC-QA & Acc. & Manual & 13 \\
\addlinespace[4pt]

\makecell{\textbf{MLVU} \\ \cite{MLVU-video}} & Txt, Vid & Comp. & 7 & 9 & 9 & 2593 & \makecell{MC-QA \\ Open} & \makecell{Acc. \\ GPT-Ranking} & Manual & 20 \\
\addlinespace[4pt]

\rowcolor{lightgraygrey} \makecell{\textbf{MVBench}  \\ \cite{MVBench-video}} & Txt,Vid & Comp. & /& 20 & 20 & 4k & MC-QA & Acc. & Repurposed & 14 \\
\addlinespace[4pt]

\makecell{\textbf{AutoEval-Video}  \\ \cite{AutoEval-Video-ChenLZH24}} & Txt,Vid & Comp. & 12 & 9 & 9 & 327 & Open & GPT-score & Manual & 11 \\
\addlinespace[4pt]

\rowcolor{lightgraygrey} \makecell{\textbf{VBench}  \\ \cite{VBench-video}} & {Txt,Vid} & Gen. & 8 & 16 & 16 & 100 & Open & / & Manual & 4 \\
\addlinespace[4pt]

\hline
\rowcolor{bg-tb-heavey-vision} \multicolumn{11}{l}{$\bullet$ \textbf{\large Image}}\\
\addlinespace[4pt]

\makecell{\textbf{MME}  \\ \cite{MME-abs-2306-13394}} & Txt,Img & Comp. & / & 14 & 14 & 2.2K  & MC-QA & Acc. & Repurposed & 30 \\
\addlinespace[4pt]

\rowcolor{lightgraygrey} \makecell{\textbf{LVLM-eHub}  \\ \cite{LVLM-eHub-abs-2306-09265}} & Txt, Img & Comp. & / & 6 & 47 & 2.1K & \makecell{MC-QA \\ Open} & \makecell{Acc. CIDEr \\ top-1 Acc.} & Repurposed & 8 \\
\addlinespace[4pt]

\makecell{\textbf{MM-Vet}  \\ \cite{MM-Vet-YuYLWL0WW24}} & Txt,Img & Comp. & / & 6 & 1 & 205 & MC-QA & GPT-score & Repurposed & 16 \\
\addlinespace[4pt]

\rowcolor{lightgraygrey} \makecell{\textbf{V*Bench} \\ \cite{V*-bench-WuX24a}} & {Txt,Img} & Comp. & 1 & 2 & 2 &191 & MC-QA & Acc. & Manual & 13 \\
\addlinespace[4pt]

\makecell{\textbf{MMIE}  \\ \cite{MMIE-abs-2410-10139} } & Txt, Img & \makecell{Comp. \\ Gen.} & 10 & 4 & 4 & 20K & \makecell{MC-QA \\ Open} & / & Manual & 8 \\
\addlinespace[4pt]

\rowcolor{lightgraygrey} \makecell{\textbf{Mia-bench}  \\ \cite{MIA-Bench-abs-2407-01509}} & Txt, Img & Comp. & 15 & 8 & 8 & 400 & Open &  GPT-score  & Manual & 29 \\
\addlinespace[4pt]

\makecell{ \textbf{MME-RealWorld} \\ \cite{MME-RealWorld-abs-2408-13257}} & Txt, Img & Comp. & 5 & 43 & 43 & 29.4K  & MC-QA & Acc. & Manual & 29 \\
\addlinespace[4pt]

\rowcolor{lightgraygrey} \makecell{ \textbf{MLLM-Bench} \\ \cite{MLLM-Bench}} & Txt, Img & Comp. & / & 6 & 6 & 420 & \makecell{MC-QA \\ Open} & GPT-score & Manual & 21 \\
\addlinespace[4pt]

\makecell{\textbf{Q-Bench}  \\ \cite{Q-Bench-0001Z0CLWLSYZL24}} & Txt, Img & Comp. & 1 & 3 & 3 & 84.7K  & \makecell{MC-QA \\ Open} & GPT-score & \makecell{Manual+ \\ Repurposed} & 15 \\
\addlinespace[4pt]

\rowcolor{lightgraygrey} \makecell{\textbf{MUIRBench}  \\ \cite{MuirBench-abs-2406-09411}} & Txt,Img & Comp. & 12 & 12 & 12 & 2.6K  & MC-QA & Acc. & \makecell{Manual+ \\ Repurposed} & 20 \\
\addlinespace[4pt]

\makecell{\textbf{MileBench}  \\ \cite{MileBench-abs-2404-18532}} & Txt,Img & Comp. & / & 2 & 2 &  6.4K  & \makecell{MC-QA \\ Open} & \makecell{Acc. \\ ROUGE-L} & \makecell{Manual+ \\ Repurposed} & 22 \\
\addlinespace[4pt]

\rowcolor{lightgraygrey} \makecell{\textbf{MMBench}  \\ \cite{MMBench-LiuDZLZZYWHLCL24}} & Txt,Img & Comp. & / & 2 & 20 & 3K  & MC-QA & Acc. & Repurposed & 21 \\
\addlinespace[4pt]

\hline
\rowcolor{mypinky} \multicolumn{11}{l}{$\bullet$ \textbf{\large Omini}}\\
\addlinespace[4pt]

\makecell{\textbf{SEED-Bench}  \\ \cite{SEED-Bench-abs-2307-16125}} & Txt,Img,Vid & Comp. & / & 12 & 12 & 19K & MC-QA & Acc. & Manual & 18 \\
\addlinespace[4pt]

\rowcolor{lightgraygrey} \makecell{\textbf{SEED-Bench-2}  \\ \cite{SEED-Bench-2-abs-2311-17092}} & Txt,Img,Vid & \makecell{Comp. \\ Gen.} & 3 & 22 & 22 & 24K  & MC-QA & Acc. & \makecell{Manual+ \\ Repurposed} & 23 \\
\addlinespace[4pt]

\makecell{\textbf{CV-Bench}  \\ \cite{Cambrian-1-abs-2406-16860}} & Txt, Img, 3D & Comp. & 2 & 4 & 4 & 2.6K  & MC-QA & Acc. & Repurposed & 15 \\
\addlinespace[4pt]

\rowcolor{lightgraygrey} \makecell{\textbf{MMIU} \\ \cite{MMIU-abs-2408-02718}} & \makecell{Txt,Img,Vid,\\Point-Cloud,Depth} & Comp. & / & 7 & 52 & 11.7K  & MC-QA & Acc. & Repurposed & 22 \\
\addlinespace[4pt]

\makecell{\textbf{MMT-Bench}  \\ \cite{MMT-Bench-YingMWLLYZZLLLL24}} & \makecell{Txt,Img,Vid,\\Point-Cloud} & Comp. & / & 32 & 162 & 31K & MC-QA & Acc. & Repurposed & 30 \\
\addlinespace[4pt]

\rowcolor{lightgraygrey} \makecell{\textbf{MEGA-Bench}  \\ \cite{MEGA-Bench-abs-2410-10563}} & Txt,Img,Vid & Comp. & / & 10 & 505 & 8K & Free-Form & Origin (45) & Manual & 22 \\
\addlinespace[4pt]

\cdashline{1-11}

\makecell{\textbf{General-Bench}  \\ (\textbf{Ours})} & \makecell{Txt,Img,Vid,Aud,\\Time,Depth,3D-RGB,\\Point-Cloud,Infrared,\\Spectrogram,Radar,\\Code,Doc,Graph,$\cdots$} & Comp.+Gen. & \color{red}{\bf 29} & \color{red}{\bf 145} & \color{red}{\textbf{702}} & \color{red}{\textbf{325.8K}} & Free-Form & Origin (\color{red}{\textbf{58}}) & \makecell{Reannotated\\ +Manual} & \parbox{1.5cm}{\centering\color{red}{\bf 172 $\times$\newline Specialists\newline \&\newline 102 $\times$\newline Generalists}} \\

\bottomrule
\end{longtable}
}



\vspace{5mm}

\subsection{Complete List of Tasks and Skills (Meta-Tasks)}
\label{Tasks-and-Skills}

We present all the tasks and their specific details, including the tasks, datasets, and corresponding SoTA specialists, in this section. 
Table \ref{tab:task-specification-img-comp} presents image-related comprehension tasks, while Table \ref{tab:task-specification-img-gen} lists image-related generation tasks. 
Similarly, Table \ref{tab:task-specification-vid-comp} summarize video-related comprehension tasks, and Table \ref{tab:task-specification-vid-gen} focuses on video-related generation tasks. 
Audio-related comprehension tasks are detailed in Table \ref{tab:task-specification-aud-comp}, with generation tasks in Table \ref{tab:task-specification-aud-gen}. 
For 3D-related tasks, Table \ref{tab:task-specification-3d-comp} highlights comprehension tasks, while Table \ref{tab:task-specification-3d-gen} covers generation tasks. 
Finally, Table \ref{tab:task-specification-nlp} outlines language-related (NLP) tasks.

\clearpage

{\centering
\fontsize{8}{8}\selectfont
\setlength{\tabcolsep}{0.7mm}
\renewcommand{\arraystretch}{0.9} 
\captionof{table}{
Detailed list of all tasks and skills (meta-tasks) under image and comprehension category.
}
\vspace{-3mm}
\label{tab:task-specification-img-comp}
\begin{longtable}[!t]{p{2.2cm}%
                >{\centering\arraybackslash}p{0.2cm}
                p{4cm}%
                >{\centering\arraybackslash}p{1.3cm}%
                p{2.3cm}%
                p{2cm}%
                >{\centering\arraybackslash}p{0.7cm}%
                p{2cm}%
                >{\centering\arraybackslash}p{1.3cm}}


\addlinespace[3pt]
\rowcolor{bg-tb-heavey-vision} \multicolumn{9}{c}{\bf \textcolor{greenCom}{\large Image Comprehension (\#I-C) Group}} \\ [4pt]
\hline

\addlinespace[4pt]
\multicolumn{5}{c}{\bf Task} & \multicolumn{2}{c}{\bf Data}  & \multicolumn{2}{c}{\bf \textbf{SoTA Specialist}} \\
\cmidrule(r){1-5}  \cmidrule(r){6-7}  \cmidrule(r){8-9} 
\textbf{Skill (Meta-Task)} & \textbf{\#}& \textbf{Task (Short Name)} &  \textbf{Domain} & \textbf{Capability} & \textbf{Data Source} & \textbf{Number} & \multicolumn{1}{c}{\textbf{Model}}  & \textbf{Metrics} \\ 
\addlinespace[4pt]
\hline
\endfirsthead
\hline
\addlinespace[4pt]
\multicolumn{5}{c}{\bf Task} & \multicolumn{2}{c}{\bf Data}  & \multicolumn{2}{c}{\bf \textbf{SoTA Specialist}} \\
\cmidrule(r){1-5}  \cmidrule(r){6-7}  \cmidrule(r){8-9} 
\textbf{Skill (Meta-Task)} & \textbf{\#}& \textbf{Task (Short Name)} &  \textbf{Domain} & \textbf{Capability} & \textbf{Data Source} & \textbf{Number} & \multicolumn{1}{c}{\textbf{Model}}  & \textbf{Metrics} \\ 
\addlinespace[4pt]
\hline
\endhead
\hline
\endfoot


\addlinespace[4pt]
\multirow{7}{*}{\shortstack[l]{\#I-C-1 \\ Behavior Recognition \\ (Behavior Recog)}} 
& 1 & ATM Booths Suspicious Behavior Recognition (ATM Behavior Reg) & General & Content Recognition & ATM Booths Anomaly Recognition \cite{furtado2023atmpositions} & 1000 & CLIP \cite{radford2021learning} & Acc \\

\addlinespace[4pt]
& \cellcolor{lightgray}2 & \cellcolor{lightgray}Human Action Recognition (Human Action Recog) & \cellcolor{lightgray}General & \cellcolor{lightgray}Content Recognition, Commonsense Knowledge & \cellcolor{lightgray}Human Action Recognition \cite{Human-Action-Recognition} & \cellcolor{lightgray}3000 & \cellcolor{lightgray}CLIP \cite{radford2021learning} & \cellcolor{lightgray}Acc \\

\addlinespace[4pt]
& 3 & Sports Image Classification (Sports Image Cls) & Sports & Content Recognition, Commonsense Knowledge & Manual Construction \cite{Sports-Image-Classification} & 500 & CaSED \cite{CaSED-conti2023vocabularyfree} & Acc \\

\addlinespace[4pt]\hline

\addlinespace[4pt]
\multirow{1}{*}{\shortstack[l]{\#I-C-2 \\ Code Generation \\ (Code Gen)}} 
& 1 &  Sketch-to-HTML Code Generation (Sketch2HTML Gen) & Code &  Creativity and Innovation, Reasoning Ability & Sketch2Code \cite{Sketch2Code} &  500 &  sketch-code \cite{sketchcode}  &  BLEU-1 \\

\addlinespace[4pt]\hline

\addlinespace[4pt]
\multirow{4}{*}{\shortstack[l]{\#I-C-3 \\ Crack Detection \\ (Crack Det)}} 
& 1 & Tire Crack Detection\newline (Tire Texture Recog) & General & Content Recognition, Commonsense Knowledge & Tire Texture Image Recognition \cite{DVN/Z3ZYLI_2021} & 325 & CLIP \cite{radford2021learning} & Acc \\

\addlinespace[4pt]
& \cellcolor{lightgray}2 & \cellcolor{lightgray}Road Crack Detection\newline (Road Crack Det) & \cellcolor{lightgray}General & \cellcolor{lightgray}Content Recognition, Commonsense Knowledge & \cellcolor{lightgray}CrackForest \cite{shi2016automatic} & \cellcolor{lightgray}118 & \cellcolor{lightgray}OWLVIT \cite{OWLVIT-abs-2205-06230} & \cellcolor{lightgray}mIoU \\

\addlinespace[4pt]\hline

\addlinespace[4pt]
\#I-C-4 \newline Disease Detection \newline (Disease Det)
&  1 &  Plant Disease Detection (Plant-Disease Det) &  Biology &  Content Recognition  &  Plant Disease Detection \cite{Plant-Disease-Detection} &  500 &  GLIP-T \cite{GLIP-li2021grounded} &  mAP@0.5 \\

\addlinespace[4pt]\hline

\addlinespace[4pt]
\multirow{12}{*}{\shortstack[l]{\#I-C-5 \\ Disease Recognition \\ (Disease Recog)}} 
& 1 & Fruit Disease Recognition\newline (Fruit Disease Recog) & General & Content Recognition & Guava Fruit Disease Dataset \cite{Guava-Fruit-Disease} & 348 & CLIP \cite{radford2021learning} & Acc \\

\addlinespace[4pt]
& \cellcolor{lightgray}2 & \cellcolor{lightgray}Abnormal Heartbeat Recognition\newline (Abnormal Heartbeat Recog) & \cellcolor{lightgray}Medicine & \cellcolor{lightgray}Content Recognition & \cellcolor{lightgray}National Heart Foundation 2023 ECG Dataset \cite{ECG-Dataset} & \cellcolor{lightgray}814 & \cellcolor{lightgray}CLIP \cite{radford2021learning} & \cellcolor{lightgray}Acc \\

\addlinespace[4pt]
& 3 & Myocardial Infarction Recognition (Myocardial Infarction Recog) & Medicine & Content Recognition & National Heart Foundation 2023 ECG Dataset \cite{ECG-Dataset} & 716 & CLIP \cite{radford2021learning} & Acc \\

\addlinespace[4pt]
& \cellcolor{lightgray}4 & \cellcolor{lightgray}Brain Tumor MRI Recognition (Brain Tumor MRI Recognition) & \cellcolor{lightgray}Medicine & \cellcolor{lightgray}Content Recognition & \cellcolor{lightgray}Brain Tumor MRI Dataset \cite{Brain-Tumor-MRI-Dataset} & \cellcolor{lightgray}505 & \cellcolor{lightgray}CLIP \cite{radford2021learning} & \cellcolor{lightgray}Acc \\

\addlinespace[4pt]
& 5 & Pumpkin Leaf Diseases Recognition (Pumpkin Diseases Recog) & Biology & Content Recognition & Pumpkin Leaf Diseases Dataset \cite{Pumpkin-Leaf-Diseases-Dataset} & 500 & CLIP \cite{radford2021learning} & Acc \\

\addlinespace[4pt]
& \cellcolor{lightgray}6 & \cellcolor{lightgray}Leukemia Classification (Leukemia Clas) & \cellcolor{lightgray}Medicine & \cellcolor{lightgray}Content Recognition & \cellcolor{lightgray}Leukemia Classification \cite{Leukemia-Classification} & \cellcolor{lightgray}500 & \cellcolor{lightgray}CLIP \cite{radford2021learning} & \cellcolor{lightgray}Acc \\

\addlinespace[4pt]\hline

\addlinespace[4pt]
\multirow{1}{*}{\shortstack[l]{\#I-C-6 \\ Document Visual \\ Question Answering \\ (Doc VQA)}} 
& 1 & Multi-Page Agreement Document Visual Question Answering\newline (Agree-Doc VQA)\newline \color{white}{1} & General & Content Recognition & MP-DocVQA \cite{MP-DocVQA-abs-2212-05935} & 272 & DocOwl \cite{DocOwl-abs-2307-02499} & Acc \\

\addlinespace[4pt]
& \cellcolor{lightgray}2 & \cellcolor{lightgray}Multi-Page Application Document Visual Question Answering\newline (Applic-Doc VQA) & \cellcolor{lightgray}General & \cellcolor{lightgray}Content Recognition & \cellcolor{lightgray}MP-DocVQA \cite{MP-DocVQA-abs-2212-05935} & \cellcolor{lightgray}282 & \cellcolor{lightgray}Donut \cite{Donut-kim2022donut} & \cellcolor{lightgray}Acc \\

\addlinespace[4pt]
& 3 & Multi-Page Certificate Document Visual Question Answering\newline (Cert-Doc VQA) & General & Content Recognition & MP-DocVQA \cite{MP-DocVQA-abs-2212-05935} & 101 & Donut \cite{Donut-kim2022donut} & Acc \\

\addlinespace[4pt]
& \cellcolor{lightgray}4 & \cellcolor{lightgray}Multi-Page E-mail Document Visual Question Answering\newline (Email-Doc VQA) & \cellcolor{lightgray}General & \cellcolor{lightgray}Content Recognition & \cellcolor{lightgray}MP-DocVQA \cite{MP-DocVQA-abs-2212-05935} & \cellcolor{lightgray}183 & \cellcolor{lightgray}DocOwl \cite{DocOwl-abs-2307-02499} & \cellcolor{lightgray}Acc \\

\addlinespace[4pt]
& 5 & Multi-Page Form Document Visual Question Answering\newline (Form-Doc VQA) & General & Content Recognition & MP-DocVQA \cite{MP-DocVQA-abs-2212-05935} & 297 & DocOwl \cite{DocOwl-abs-2307-02499} & Acc \\

\addlinespace[4pt]
& \cellcolor{lightgray}6 & \cellcolor{lightgray}Multi-Page Handwritten Document Visual Question Answering\newline (Handwrit-Doc VQA) & \cellcolor{lightgray}General & \cellcolor{lightgray}Content Recognition & \cellcolor{lightgray}MP-DocVQA \cite{MP-DocVQA-abs-2212-05935} & \cellcolor{lightgray}146 & \cellcolor{lightgray}DocOwl \cite{DocOwl-abs-2307-02499} & \cellcolor{lightgray}Acc \\

\addlinespace[4pt]
& 7 & Multi-Page Infographic Document Visual Question Answering\newline (Infograph-Doc VQA) & General & Content Recognition & MP-DocVQA \cite{MP-DocVQA-abs-2212-05935} & 239 & DocOwl \cite{DocOwl-abs-2307-02499} & Acc \\

\addlinespace[4pt]
& \cellcolor{lightgray}8 & \cellcolor{lightgray}Multi-Page Invoice Document Visual Question Answering\newline (Invoi-Doc VQA) & \cellcolor{lightgray}General & \cellcolor{lightgray}Content Recognition & \cellcolor{lightgray}MP-DocVQA \cite{MP-DocVQA-abs-2212-05935} & \cellcolor{lightgray}362 & \cellcolor{lightgray}DocOwl \cite{DocOwl-abs-2307-02499} & \cellcolor{lightgray}Acc \\

\addlinespace[4pt]
& 9 & Multi-Page Letter Document Visual Question Answering\newline (Letter-Doc VQA) & General & Content Recognition & MP-DocVQA \cite{MP-DocVQA-abs-2212-05935} & 391 & DocQuery \cite{docquery} & Acc \\

\addlinespace[4pt]
& \cellcolor{lightgray}10 & \cellcolor{lightgray}Multi-Page Manual Document Visual Question Answering\newline (Manual-Doc VQA) & \cellcolor{lightgray}General & \cellcolor{lightgray}Content Recognition & \cellcolor{lightgray}MP-DocVQA \cite{MP-DocVQA-abs-2212-05935} & \cellcolor{lightgray}619 & \cellcolor{lightgray}DocOwl \cite{DocOwl-abs-2307-02499} & \cellcolor{lightgray}Acc \\

\addlinespace[4pt]
& 11 & Multi-Page Meeting Document Visual Question Answering\newline (Meet-Doc VQA) & General & Content Recognition & MP-DocVQA \cite{MP-DocVQA-abs-2212-05935} & 229 & Donut \cite{Donut-kim2022donut} & Acc \\

\addlinespace[4pt]
& \cellcolor{lightgray}12 & \cellcolor{lightgray}Multi-Page Memo Document Visual Question Answering\newline (Memo-Doc VQA) & \cellcolor{lightgray}General & \cellcolor{lightgray}Content Recognition & \cellcolor{lightgray}MP-DocVQA \cite{MP-DocVQA-abs-2212-05935} & \cellcolor{lightgray}68 & \cellcolor{lightgray}DocQuery \cite{docquery} & \cellcolor{lightgray}Acc \\

\addlinespace[4pt]
& 13 & Multi-Page News Document Visual Question Answering\newline (News-Doc VQA) & General & Content Recognition & MP-DocVQA \cite{MP-DocVQA-abs-2212-05935} & 185 & DocOwl \cite{DocOwl-abs-2307-02499} & Acc \\

\addlinespace[4pt]
& \cellcolor{lightgray}14 & \cellcolor{lightgray}Multi-Page Order Document Visual Question Answering\newline (Order-Doc VQA) & \cellcolor{lightgray}General & \cellcolor{lightgray}Content Recognition & \cellcolor{lightgray}MP-DocVQA \cite{MP-DocVQA-abs-2212-05935} & \cellcolor{lightgray}151 & \cellcolor{lightgray}DocQuery \cite{docquery} & \cellcolor{lightgray}Acc \\

\addlinespace[4pt]
& 15 & Multi-Page Poster Document Visual Question Answering\newline (Poster-Doc VQA) & General & Content Recognition & MP-DocVQA \cite{MP-DocVQA-abs-2212-05935} & 320 & DocOwl \cite{DocOwl-abs-2307-02499} & Acc \\

\addlinespace[4pt]
& \cellcolor{lightgray}16 & \cellcolor{lightgray}Multi-Page Presentation Document Visual Question Answering\newline (Pres-Doc VQA) & \cellcolor{lightgray}General & \cellcolor{lightgray}Content Recognition & \cellcolor{lightgray}MP-DocVQA \cite{MP-DocVQA-abs-2212-05935} & \cellcolor{lightgray}310 & \cellcolor{lightgray}DocQuery \cite{docquery} & \cellcolor{lightgray}Acc \\

\addlinespace[4pt]
& 17 & Multi-Page Report Document Visual Question Answering\newline (Reprt-Doc VQA) & General & Content Recognition & MP-DocVQA \cite{MP-DocVQA-abs-2212-05935} & 648 & DocQuery \cite{docquery} & Acc \\

\addlinespace[4pt]
& \cellcolor{lightgray}18 & \cellcolor{lightgray}Multi-Page Request Document Visual Question Answering\newline (Reqst-Doc VQA) & \cellcolor{lightgray}General & \cellcolor{lightgray}Content Recognition & \cellcolor{lightgray}MP-DocVQA \cite{MP-DocVQA-abs-2212-05935} & \cellcolor{lightgray}43 & \cellcolor{lightgray}DocQuery \cite{docquery} & \cellcolor{lightgray}Acc \\

\addlinespace[4pt]
& 19 & Multi-Page Resume Document Visual Question Answering\newline (Resu-Doc VQA) & General & Content Recognition & MP-DocVQA \cite{MP-DocVQA-abs-2212-05935} & 210 & DocQuery \cite{docquery} & Acc \\

\addlinespace[4pt]
& \cellcolor{lightgray}20 & \cellcolor{lightgray}Multi-Page Sheet Document Visual Question Answering\newline (Sheet-Doc VQA) & \cellcolor{lightgray}General & \cellcolor{lightgray}Content Recognition, Commonsense Knowledge & \cellcolor{lightgray}MP-DocVQA \cite{MP-DocVQA-abs-2212-05935} & \cellcolor{lightgray}245 & \cellcolor{lightgray}Donut \cite{Donut-kim2022donut} & \cellcolor{lightgray}Acc \\

\addlinespace[4pt]
& 21 & General Document Visual Question Answering\newline (General-Doc VQA) & General & Content Recognition & MP-DocVQA \cite{MP-DocVQA-abs-2212-05935} & 500 & DocOwl 1.5-Chat \cite{DocOwl-abs-2307-02499} & Acc \\

\addlinespace[4pt]\hline

\addlinespace[4pt]
\multirow{1}{*}{\shortstack[l]{\#I-C-7 \\ Emotion Detection \\ (Emotion Det)}} 
& 1 & Person Emotion Detection (Emotion Det) & Humanities & Affective Analysis & Emotion Detection \cite{Emotion-Detection} & 707 & Deepface \cite{deepface-serengil2020lightface} & Acc \\

\addlinespace[4pt]
& \cellcolor{lightgray}2 & \cellcolor{lightgray}Pet Face Expression Recognition (Pet-Expr Recog) & \cellcolor{lightgray}General & \cellcolor{lightgray}Affective Analysis & \cellcolor{lightgray}Pet's Facial Expression Image Dataset \cite{Pet-Facial-Expression-Image-Dataset} & \cellcolor{lightgray}249 & \cellcolor{lightgray}CLIP \cite{radford2021learning} & \cellcolor{lightgray}Acc \\

\addlinespace[4pt]
& 3 & Emotion Recognition\newline (Emotion Recog) & Humanities & Affective Analysis & FER \cite{Emotion-Detection} & 717 & CLIP \cite{radford2021learning} & Acc \\

\addlinespace[4pt]
& \cellcolor{lightgray}4 & \cellcolor{lightgray}Face Emotion Recognition (Face Emotion Recog) & \cellcolor{lightgray}General & \cellcolor{lightgray}Content Recognition, Commonsense Knowledge & \cellcolor{lightgray}Manual Construction \cite{Facial-Emotion-Recognition} & \cellcolor{lightgray}540 & \cellcolor{lightgray}DDAMFN++ \cite{DDAMFN-electronics12173595} & \cellcolor{lightgray}Acc \\

\addlinespace[4pt]\hline

\addlinespace[4pt]
\multirow{1}{*}{\shortstack[l]{\#I-C-8 \\ Graph Classification \\ (Graph Cls)}} 
& 1 & Spectrum Graph Car Classification (Spectrum-Graph Cls) & General & Content Recognition & Emergency Vehicle Siren Sounds \cite{Emergency-Vehicle-Siren-Sounds} & 600 & Mantis \cite{MANTIS-Jiang2024MANTISIM} & Acc \\

\addlinespace[4pt]\hline

\addlinespace[4pt]
\multirow{4}{*}{\shortstack[l]{\#I-C-9 \\ Hallucination Detection \\ (Hallucination Det)}} 
& 1 & Attribute Hallucination Detection (Attr-Hallu Det) & Ethics & Content Recognition & PhD \cite{PhD-liu2024phd} & 500 & SAC3 \cite{zhang2023sac} & Acc \\

\addlinespace[4pt]
& \cellcolor{lightgray}2 & \cellcolor{lightgray}Object Hallucination Detection (Obj-Hallu Det) & \cellcolor{lightgray}Ethics & \cellcolor{lightgray}Content Recognition & \cellcolor{lightgray}PhD \cite{PhD-liu2024phd} & \cellcolor{lightgray}500 & \cellcolor{lightgray}SAC3 \cite{zhang2023sac}  & \cellcolor{lightgray}Acc \\

\addlinespace[4pt]\hline

\addlinespace[4pt]
\multirow{18}{*}{\shortstack[l]{\#I-C-10 \\ Image Caption \\ (Img Cap)}} 
& 1 & Instagram Image Caption (Ins-Img Cap) & Humanities, Social & Content Recognition, Reasoning Ability & Instagram Images with Captions \cite{Instagram-Images-with-Captions} & 557 & GRIT \cite{Grit-nguyen2022} & ROUGE \\

\addlinespace[4pt]
& \cellcolor{lightgray}2 & \cellcolor{lightgray}COCO Image Captioning (COCO-Img Cap) & \cellcolor{lightgray}General & \cellcolor{lightgray}Content Recognition, Reasoning Ability & \cellcolor{lightgray}COCO \cite{COCO-LinMBHPRDZ14} & \cellcolor{lightgray}500 & \cellcolor{lightgray}GRIT \cite{Grit-nguyen2022} & \cellcolor{lightgray}ROUGE \\

\addlinespace[4pt]
& 3 & Satellite Image Caption (Satell-Img Cap) & General & Content Recognition, Reasoning Ability & Satellite Image Caption Generation \cite{Satellite-Image-Caption-Generation-LuWZL18} & 646 & GRIT \cite{Grit-nguyen2022} & ROUGE \\

\addlinespace[4pt]
& \cellcolor{lightgray}4 & \cellcolor{lightgray}Human Related Image Captioning (Human Cap) & \cellcolor{lightgray}General & \cellcolor{lightgray}Content Recognition, Reasoning Ability & \cellcolor{lightgray}Human Related Image Captioning \cite{Human-Related-Image-Captioning} & \cellcolor{lightgray}500 & \cellcolor{lightgray}GRIT \cite{Grit-nguyen2022} & \cellcolor{lightgray}BLEU-1 \\

\addlinespace[4pt]
& 5 & Chinese Image Captioning (Zh-Img Cap) & General & Content Recognition, Reasoning Ability, Linguistics & Chinese-Image-Captioning \cite{Chinese-Image-Captioning} & 500 & GRIT \cite{Grit-nguyen2022} & BLEU-1 \\

\addlinespace[4pt]
& \cellcolor{lightgray}6 & \cellcolor{lightgray}Astronomy Image Captioning (Astro-Img Cap) & \cellcolor{lightgray}Astronomy & \cellcolor{lightgray}Content Recognition, Reasoning Ability & \cellcolor{lightgray}Manual Construction \cite{SpaceNet-abs-2405-13267} & \cellcolor{lightgray}500 & \cellcolor{lightgray}llava-v1.5 \cite{LLava-15-LiuLLL24} & \cellcolor{lightgray}ROUGE-L \\

\addlinespace[4pt]\hline

\addlinespace[4pt]
\#I-C-11  \newline  Image Depth  \newline  Estimation  \newline  (Img Depth Est)
& 1 & Indoor Depth Estimation (Indoor-Depth Est) & General & Content Recognition & Nyudv2 \cite{Nyudv2-SilbermanHKF12} & 654 & AdaBins \cite{AdaBins-BhatAW21} & RMS \\

\addlinespace[4pt]\hline

\addlinespace[4pt]
\multirow{7}{*}{\shortstack[l]{\#I-C-12 \\ Image Instance \\ Segmentation \\ (Img Inst Seg)}} 
& 1 & Pedestrian Instance Segmentation (Pede-Inst Seg) & General & Content Recognition, Reasoning Ability & Cityscapes \cite{Cityscapes-CordtsORREBFRS16} & 398 & BPR \cite{BPR-TangCLLZH21} & AP@0.5 \\

\addlinespace[4pt]
& \cellcolor{lightgray}2 & \cellcolor{lightgray}Vehicle Instance Segmentation (Vehi-Inst Seg) & \cellcolor{lightgray}General & \cellcolor{lightgray}Content Recognition, Commonsense Knowledge & \cellcolor{lightgray}Cityscapes \cite{Cityscapes-CordtsORREBFRS16} & \cellcolor{lightgray}500 & \cellcolor{lightgray}BPR \cite{BPR-TangCLLZH21} & \cellcolor{lightgray}AP@0.5 \\

\addlinespace[4pt]\hline

\addlinespace[4pt]
\multirow{12}{*}{\shortstack[l]{\#I-C-13 \\ Image OCR \\ (Img OCR)}} 
& 1 & Handwriting OCR (Handwrite OCR) & General & Content Recognition & Handwriting Recognition (OCR) \cite{Handwriting-Recognition(OCR)} & 500 &  Tesseract \cite{tesseract} & ANLS \\

\addlinespace[4pt]
& \cellcolor{lightgray}2 & \cellcolor{lightgray}Letter and Number OCR\newline (Letter\&Num OCR) & \cellcolor{lightgray}General & \cellcolor{lightgray}Content Recognition & \cellcolor{lightgray}standard OCR dataset \cite{standard-OCR-dataset} & \cellcolor{lightgray}720 & \cellcolor{lightgray}Tesseract \cite{tesseract} & \cellcolor{lightgray}ANLS \\

\addlinespace[4pt]
& 3 & License Plate OCR (License OCR) & General & Content Recognition & License Plate Text Recognition Dataset \cite{Car-License-Plate-Detection} & 533 & Tesseract \cite{tesseract} & ANLS \\

\addlinespace[4pt]
& \cellcolor{lightgray}4 & \cellcolor{lightgray}Radical-Level Oracle Bone Character Recognition (Radical-Char Recog) & \cellcolor{lightgray}History & \cellcolor{lightgray}Content Recognition & \cellcolor{lightgray}Radical-Level Oracle Bone Character Dataset \cite{Radical-Level-Oracle-Bone-Character-Dataset} & \cellcolor{lightgray}500 & \cellcolor{lightgray}Tesseract \cite{tesseract} & \cellcolor{lightgray}ANLS \\

\addlinespace[4pt]\hline

\addlinespace[4pt]
\multirow{7}{*}{\shortstack[l]{\#I-C-14 \\ Image Recognition \\ (Img Recog)}} 
& 1 & Acne Recognition (Acne Recog) & Medicine & Content Recognition, Commonsense Knowledge & Acne \cite{Acne-Recognition-Dataset} & 600 & CLIP \cite{radford2021learning} & Acc \\

\addlinespace[4pt]
& \cellcolor{lightgray}2 & \cellcolor{lightgray}Latex Code Recognition (Latex Recog) & \cellcolor{lightgray}Code & \cellcolor{lightgray}Content Recognition & \cellcolor{lightgray}Latex OCR \cite{LaTeX-OCR} & \cellcolor{lightgray}500 & \cellcolor{lightgray}pix2tex \cite{pix2tex}  & \cellcolor{lightgray}BLEU-1 \\

\addlinespace[4pt]\hline

\addlinespace[4pt]
\multirow{1}{*}{\shortstack[l]{\#I-C-15 \\ Image Semantic \\ Segmentation \\ (Img Sem Seg)}} 
& 1 & BobRoss Painting Segmentation \newline (BobRoss-Paint seg) \newline \color{white}{1} \newline \color{white}{1} \newline \color{white}{1}  & Art & Content Recognition & Segmented Bob Ross Paintings \cite{Segmented-Bob-Ross-Images} & 250 & CLIPSeg \cite{CLIPSeg-lueddecke22} & mIoU \\

\addlinespace[4pt]
& \cellcolor{lightgray}2 & \cellcolor{lightgray}Indoor Semantic Segmentation (Indoor-Sem Seg) & \cellcolor{lightgray}General & \cellcolor{lightgray}Content Recognition, Commonsense Knowledge & \cellcolor{lightgray}Nyudv2 \cite{Nyudv2-SilbermanHKF12} & \cellcolor{lightgray}654 & \cellcolor{lightgray}Cerberus \cite{Cerberus-ChenL0ZZ22} & \cellcolor{lightgray}mIoU \\

\addlinespace[4pt]
& 3 & Pedestrian Semantic Segmentation (Pede-Sem Seg) & General & Content Recognition, Commonsense Knowledge & Cityscapes \cite{Cityscapes-CordtsORREBFRS16} & 398 & DeepLabV3+ \cite{deeplabv3plus2018} & mIoU \\

\addlinespace[4pt]
& \cellcolor{lightgray}4 & \cellcolor{lightgray}Road Semantic Segmentation (Road-Sem Seg) & \cellcolor{lightgray}General & \cellcolor{lightgray}Content Recognition, Commonsense Knowledge & \cellcolor{lightgray}Cityscapes \cite{Cityscapes-CordtsORREBFRS16} & \cellcolor{lightgray}500 & \cellcolor{lightgray}DeepLabV3+ \cite{deeplabv3plus2018} & \cellcolor{lightgray}mIoU \\

\addlinespace[4pt]
& 5 & Vehicle Semantic Segmentation (Vehi-Sem Seg) & General & Content Recognition, Commonsense Knowledge & Cityscapes \cite{Cityscapes-CordtsORREBFRS16} & 500 & DeepLabV3+ \cite{deeplabv3plus2018} & mIoU \\

\addlinespace[4pt]
& \cellcolor{lightgray}6 & \cellcolor{lightgray}People Clothing Segmentation (Cloth Seg) & \cellcolor{lightgray}General & \cellcolor{lightgray}Content Recognition & \cellcolor{lightgray}People Clothing Segmentation \cite{yang2014clothing} & \cellcolor{lightgray}500 & \cellcolor{lightgray}CLIPseg \cite{CLIPSeg-lueddecke22} & \cellcolor{lightgray}mIoU \\

\addlinespace[4pt]
& 7 & Flood Area Segmentation\newline (Flood Seg) & General & Content Recognition & Flood Area Segmentation \cite{Flood-Area-Segmentation} & 290 & CLIPseg \cite{CLIPSeg-lueddecke22} & mIoU \\

\addlinespace[4pt]
& \cellcolor{lightgray}8 & \cellcolor{lightgray}Forest Aerial Images for Segmentation (Forest Seg) & \cellcolor{lightgray}General & \cellcolor{lightgray}Content Recognition & \cellcolor{lightgray}Forest Aerial Images for Segmentation \cite{DeepGlobe18} & \cellcolor{lightgray}465 & \cellcolor{lightgray}CLIPseg \cite{CLIPSeg-lueddecke22} & \cellcolor{lightgray}mIoU \\

\addlinespace[4pt]\hline

\addlinespace[4pt]
\multirow{6}{*}{\shortstack[l]{\#I-C-16 \\ Image Visual \\ Grounding \\ (Img Vis Grnd)}} 
& 1 & Complex Expression Visual Grounding (Complex-Expre VG) & General & Content Recognition, Reasoning Ability & Refcocog \cite{Refcocog-YuPYBB16} & 990 & Polygon-Transformer \cite{PolyFormer-liu2023} & AP@0.5 \\

\addlinespace[4pt]
& \cellcolor{lightgray}2 & \cellcolor{lightgray}Short Expression Visual Grounding (Short-Expre VG) & \cellcolor{lightgray}General & \cellcolor{lightgray}Content Recognition, Reasoning Ability & \cellcolor{lightgray}Refcoco \cite{Refcoco-KazemzadehOMB14} & \cellcolor{lightgray}665 & \cellcolor{lightgray}Polygon-Transformer \cite{PolyFormer-liu2023} & \cellcolor{lightgray}AP@0.5 \\

\addlinespace[4pt]\hline

\addlinespace[4pt]
\multirow{48}{*}{\shortstack[l]{\#I-C-17 \\ Image Visual \\ Question \\ Answering \\ (Img VQA)}} 
& 1 & Blind Image Question Answering (Blind VQA) & General & Content Recognition & vizwiz \cite{VizWiz-Gurari0SGLGLB18} & 550 & GIT \cite{GIT-WangYHLLGLLW22} & Acc \\

\addlinespace[4pt]
& \cellcolor{lightgray}2 & \cellcolor{lightgray}External Knowledge Required Image Question Answering\newline (Knowledge VQA) & \cellcolor{lightgray}General & \cellcolor{lightgray}Commonsense Knowledge, Reasoning Ability & \cellcolor{lightgray}OKVQA \cite{okvqa} & \cellcolor{lightgray}550 & \cellcolor{lightgray}GIT \cite{GIT-WangYHLLGLLW22} & \cellcolor{lightgray}Acc \\

\addlinespace[4pt]
& 3 & Medical Image Question Answering (Medical VQA) & Medicine & Commonsense Knowledge & Medical Visual Question Answering \cite{medical-visual-question-answering} & 316 & PMC-VQA \cite{PMC-VQA-zhang2023} & BLEU-1 \\

\addlinespace[4pt]
& \cellcolor{lightgray}4 & \cellcolor{lightgray}Reasoning and Compositional Image Question Answering (Reason VQA) & \cellcolor{lightgray}General & \cellcolor{lightgray}Content Recognition, Reasoning Ability & \cellcolor{lightgray}GQA \cite{GQA-HudsonM19} & \cellcolor{lightgray}519 & \cellcolor{lightgray}GIT \cite{GIT-WangYHLLGLLW22} & \cellcolor{lightgray}Acc \\

\addlinespace[4pt]
& 5 & Biological Commonsense Visual Question Answering (Biology VQA) & Biology & Commonsense Knowledge & ScienceQA \cite{ScienceQA-LuMX0CZTCK22} & 428 & MMCoT \cite{MMCoT-0001Z00KS24} & Acc \\

\addlinespace[4pt]
& \cellcolor{lightgray}6 & \cellcolor{lightgray}Blood Cell Visual Question Answering (Blood-Cell VQA) & \cellcolor{lightgray}Biology & \cellcolor{lightgray}Commonsense Knowledge & \cellcolor{lightgray}Blood Cell Images \cite{Blood-Cell-Images} & \cellcolor{lightgray}352 & \cellcolor{lightgray}PMC-VQA \cite{PMC-VQA-zhang2023} & \cellcolor{lightgray}Acc \\

\addlinespace[4pt]
& 7 & Cataract Diagnosis Visual Question Answering (Cataract VQA) & Medicine & Commonsense Knowledge & Cataract dataset \cite{cataract-dataset} & 300 & PMC-VQA \cite{PMC-VQA-zhang2023} & Acc \\

\addlinespace[4pt]
& \cellcolor{lightgray}8 & \cellcolor{lightgray}Chemistry Visual Question Answering (Chemistry VQA) & \cellcolor{lightgray}Chemistry & \cellcolor{lightgray}Reasoning Ability & \cellcolor{lightgray}ScienceQA \cite{ScienceQA-LuMX0CZTCK22} & \cellcolor{lightgray}110 & \cellcolor{lightgray}MMCoT \cite{MMCoT-0001Z00KS24} & \cellcolor{lightgray}Acc \\

\addlinespace[4pt]
& 9 & Geography Visual Question Answering (Geography VQA) & Geography & Reasoning Ability & ScienceQA \cite{ScienceQA-LuMX0CZTCK22} & 594 & MMCoT \cite{MMCoT-0001Z00KS24} & Acc \\

\addlinespace[4pt]
& \cellcolor{lightgray}10 & \cellcolor{lightgray}Geometry Visual Question Answering (Geometry VQA) & \cellcolor{lightgray}Geometry & \cellcolor{lightgray}Reasoning Ability & \cellcolor{lightgray}Inter-GPS \cite{Inter-GPS-LuGJQHLZ20} & \cellcolor{lightgray}300 & \cellcolor{lightgray}Inter-GPS \cite{Inter-GPS-LuGJQHLZ20} & \cellcolor{lightgray}Acc \\

\addlinespace[4pt]
& 11 & Lung Nodule Diagnosis Visual Question Answering (Lung VQA) & Medicine & Reasoning Ability & Lung cancer CAM attempting \cite{Lung-cance-CAM-attempting} & 300 & PMC-VQA \cite{PMC-VQA-zhang2023} & Acc \\

\addlinespace[4pt]
& \cellcolor{lightgray}12 & \cellcolor{lightgray}Mathematical Statistical Reasoning Visual Question Answering\newline (Stat-Reasoning VQA) & \cellcolor{lightgray}Mathematics & \cellcolor{lightgray}Reasoning Ability & \cellcolor{lightgray}MathVista \cite{MathVista-lu2024} & \cellcolor{lightgray}116 & \cellcolor{lightgray}Chimera \cite{Chimera-peng2024} & \cellcolor{lightgray}Acc \\

\addlinespace[4pt]
& 13 & Mathematical Textbook Visual Question Answering (Math-Textbook VQA) & Mathematics & Reasoning Ability & MathVista \cite{MathVista-lu2024} & 107 & Chimera \cite{Chimera-peng2024} & Acc \\

\addlinespace[4pt]
& \cellcolor{lightgray}14 & \cellcolor{lightgray}Mathematical Word Problem Visual Question Answering (Math-Word VQA) & \cellcolor{lightgray}Mathematics & \cellcolor{lightgray}Reasoning Ability & \cellcolor{lightgray}MathVista \cite{MathVista-lu2024} & \cellcolor{lightgray}22 & \cellcolor{lightgray}Chimera \cite{Chimera-peng2024} & \cellcolor{lightgray}Acc \\

\addlinespace[4pt]
& 15 & Physics Visual Question Answering (Physics VQA) & Physics & Reasoning Ability & ScienceQA \cite{ScienceQA-LuMX0CZTCK22} & 449 & MMCoT \cite{MMCoT-0001Z00KS24} & Acc \\

\addlinespace[4pt]
& \cellcolor{lightgray}16 & \cellcolor{lightgray}Pneumonia Diagnosis Visual Question Answering (Pneumonia VQA) & \cellcolor{lightgray}Medicine & \cellcolor{lightgray}Reasoning Ability & \cellcolor{lightgray}COVIDx CXR-4 \cite{COVID-Net-Wang2020} & \cellcolor{lightgray}400 & \cellcolor{lightgray}PMC-VQA \cite{PMC-VQA-zhang2023} & \cellcolor{lightgray}Acc \\

\addlinespace[4pt]
& 17 & Mathematical Synthetic Scene Visual Question Answering (Math-Scene VQA) & Mathematics & Reasoning Ability & MathVista \cite{MathVista-lu2024} & 92 & Chimera \cite{Chimera-peng2024} & Acc \\

\addlinespace[4pt]
& \cellcolor{lightgray}18 & \cellcolor{lightgray}Remote Sense Visual Question Answering (Remote-Sense VQA) & \cellcolor{lightgray}Earth & \cellcolor{lightgray}Reasoning Ability & \cellcolor{lightgray}EarthVQA \cite{EarthVQA-WangZCMZ24} & \cellcolor{lightgray}300 & \cellcolor{lightgray}RSVQA \cite{RSVQA-LobryMMT20} & \cellcolor{lightgray}Acc \\

\addlinespace[4pt]
& 19 & Skin Disease Visual Question Answering (Skin-Disease VQA) & Medicine & Reasoning Ability & Skin Cancer \cite{Skin-Cancer-Dateset} & 300 & PMC-VQA \cite{PMC-VQA-zhang2023} & Acc \\

\addlinespace[4pt]
& \cellcolor{lightgray}20 & \cellcolor{lightgray}Internet-Retrieved-Information Visual Question Answering (Retrieved VQA) & \cellcolor{lightgray}General & \cellcolor{lightgray}Content Recognition & \cellcolor{lightgray}WebQA \cite{WebQA-ChangCNGSB22} & \cellcolor{lightgray}500 & \cellcolor{lightgray}MLoEM \cite{MLoEM-WeiSXZLW24} & \cellcolor{lightgray}Acc \\

\addlinespace[4pt]
& 21 & Webpage Content Visual Question Answering (Webpage VQA) & General & Content Recognition & WebQA \cite{WebQA-ChangCNGSB22} & 500 & MLoEM \cite{MLoEM-WeiSXZLW24} & Acc \\

\addlinespace[4pt]
& \cellcolor{lightgray}22 & \cellcolor{lightgray}Sporting-related VQA (Sport VQA) & \cellcolor{lightgray}Sports & \cellcolor{lightgray}Content Recognition & \cellcolor{lightgray}MMCoQA \cite{MMCoQA-LiLN22} & \cellcolor{lightgray}500 & \cellcolor{lightgray}LLaVA-NeXT-Interleave \cite{li2024llava} & \cellcolor{lightgray}ROUGE-L \\

\addlinespace[4pt]
& 23 & Financial Dialogue VQA (Financial VQA) & Finance & Content Recognition & MMCoQA \cite{MMCoQA-LiLN22} & 500 & VPG-C \cite{VPG-C-li2023fine} & ROUGE-L \\

\addlinespace[4pt]
& \cellcolor{lightgray}24 & \cellcolor{lightgray}Historical Multi-Source Dialogue (Historical Dialogue) & \cellcolor{lightgray}General & \cellcolor{lightgray}Content Recognition & \cellcolor{lightgray}MMCoQA \cite{MMCoQA-LiLN22} & \cellcolor{lightgray}500 & \cellcolor{lightgray}VPG-C \cite{VPG-C-li2023fine} & \cellcolor{lightgray}ROUGE-L \\

\addlinespace[4pt]
& 25 & Multi-Turn Interactive Dialogue (Multi-Dialogue) & General & Content Recognition & MMCoQA \cite{MMCoQA-LiLN22} & 500 & MLoEM \cite{MLoEM-WeiSXZLW24} & ROUGE-L \\

\addlinespace[4pt]
& \cellcolor{lightgray}26 & \cellcolor{lightgray}Complex Multi-Scenario Dialogue (Multi-Scene Dialogue) & \cellcolor{lightgray}General & \cellcolor{lightgray}Content Recognition & \cellcolor{lightgray}MMCoQA \cite{MMCoQA-LiLN22} & \cellcolor{lightgray}500 & \cellcolor{lightgray}Mantis \cite{MANTIS-Jiang2024MANTISIM} & \cellcolor{lightgray}ROUGE-L \\

\addlinespace[4pt]
& 27 & Multimodal Slide Question Answering (Slide VQA) & General & Content Recognition & SlideVQA \cite{SlideVQA-TanakaNNHSS23} & 500 & Mantis \cite{MANTIS-Jiang2024MANTISIM} & Acc \\

\addlinespace[4pt]
& \cellcolor{lightgray}28 & \cellcolor{lightgray}Multimodal Document VQA (MM-Doc VQA) & \cellcolor{lightgray}General & \cellcolor{lightgray}Content Recognition & \cellcolor{lightgray}DocVQA \cite{DocVQA-MathewKJ21} & \cellcolor{lightgray}500 & \cellcolor{lightgray}MLoEM \cite{MLoEM-WeiSXZLW24} & \cellcolor{lightgray}Acc \\

\addlinespace[4pt]
& 29 & Multimodal Document Fine-grained Reasoning (MM-Doc Reason) & General & Content Recognition & DocVQA \cite{DocVQA-MathewKJ21} & 500 & Mantis \cite{MANTIS-Jiang2024MANTISIM} & Acc \\

\addlinespace[4pt]
& \cellcolor{lightgray}30 & \cellcolor{lightgray}Textbook Diagram VQA (Diagram VQA) & \cellcolor{lightgray}Education & \cellcolor{lightgray}Content Recognition & \cellcolor{lightgray}TQA \cite{TQA-KembhaviSSCFH17} & \cellcolor{lightgray}500 & \cellcolor{lightgray}MLoEM \cite{MLoEM-WeiSXZLW24} & \cellcolor{lightgray}Acc \\

\addlinespace[4pt]
& 31 & Education-Related VQA (Education VQA) & Education & Content Recognition & TQA \cite{TQA-KembhaviSSCFH17} & 500 & MLoEM \cite{MLoEM-WeiSXZLW24} & Acc \\

\addlinespace[4pt]
& \cellcolor{lightgray}32 & \cellcolor{lightgray}Book Information VQA (Book VQA) & \cellcolor{lightgray}General & \cellcolor{lightgray}Content Recognition & \cellcolor{lightgray}OCR-VQA \cite{OCR-VQA-0001SSC19} & \cellcolor{lightgray}500 & \cellcolor{lightgray}VPG-C \cite{VPG-C-li2023fine} & \cellcolor{lightgray}Acc \\

\addlinespace[4pt]
& 33 & Textual Document VQA (Text-Doc VQA) & General & Content Recognition & OCR-VQA \cite{OCR-VQA-0001SSC19} & 500 & VPG-C \cite{VPG-C-li2023fine} & Acc \\

\addlinespace[4pt]
& \cellcolor{lightgray}34 & \cellcolor{lightgray}Text-Table-Image Joint Reasoning (Txt-Tab-Img Reason) & \cellcolor{lightgray}General & \cellcolor{lightgray}Content Recognition & \cellcolor{lightgray}MultiModalQA \cite{} & \cellcolor{lightgray}500 & \cellcolor{lightgray}MLoEM \cite{MLoEM-WeiSXZLW24} & \cellcolor{lightgray}Acc \\

\addlinespace[4pt]
& 35 & Handwritten Formula Visual Question Answering\newline (Handwrite-Formu VQA) & Mathematics & Reasoning Ability & Chrome2016 \cite{Handwritten-Formula-VQA} & 500 & Mantis \cite{MANTIS-Jiang2024MANTISIM} & Acc \\

\addlinespace[4pt]
& \cellcolor{lightgray}36 & \cellcolor{lightgray}Face Comparison Visual Question Answering (Face-Compare VQA) & \cellcolor{lightgray}General & \cellcolor{lightgray}Content Recognition & \cellcolor{lightgray}LFW \cite{Face-Comparison-VQA} & \cellcolor{lightgray}500 & \cellcolor{lightgray}mPLUG-Docowl2 \cite{mPLUG-DocOwl2-abs-2409-03420} & \cellcolor{lightgray}Acc \\

\addlinespace[4pt]
& 37 & Image Similarity Visual Question Answering (Img-Similar VQA) & General & Reasoning Ability & TTL \cite{ttl-RosenfeldST18} & 500 & Mantis \cite{MANTIS-Jiang2024MANTISIM} & Acc \\

\addlinespace[4pt]
& \cellcolor{lightgray}38 & \cellcolor{lightgray}Art Image Visual Question Answering (Art-Img VQA) & \cellcolor{lightgray}Art & \cellcolor{lightgray}Content Recognition, Reasoning Ability & \cellcolor{lightgray}AQUA \cite{garcia2020AQUA} & \cellcolor{lightgray}500 & \cellcolor{lightgray}mPLUG-1B \cite{mPLUG-LiXTWYBYCXCZHHZ22} & \cellcolor{lightgray}Acc \\

\addlinespace[4pt]
& 39 & Culture Image Visual Question Answering (Culture-Img VQA) & Culture & Content Recognition, Reasoning Ability & CII-Bench \cite{CII-Bench-abs-2410-13854} & 499 & mPLUG-1B \cite{mPLUG-LiXTWYBYCXCZHHZ22} & Acc \\

\addlinespace[4pt]
& \cellcolor{lightgray}40 & \cellcolor{lightgray}Imge Editing Instruction Generation (Img-Edit Instru Gen) & \cellcolor{lightgray}General & \cellcolor{lightgray}Content Recognition, Creativity and Innovation & \cellcolor{lightgray}Manual Construction \cite{StyleBooth-abs-2404-12154} & \cellcolor{lightgray}500 & \cellcolor{lightgray}llava-v1.5 \cite{LLava-15-LiuLLL24} & \cellcolor{lightgray}ROUGE-L \\

\addlinespace[4pt]
\hline

\addlinespace[4pt]
\multirow{8}{*}{\shortstack[l]{\#I-C-18 \\ Industrial\\ Anomaly Detection \\ (Ind-Anomaly Det)}} 
& 1 & Bottle Anomaly Detection (Bottle Anomaly Det) & Engineering & Content Recognition & MVTec \cite{MVTec-BergmannBFSS21} & 63 & DeSTSeg \cite{DeSTSeg-ZhangL0HSC23} & AP \\ 

\addlinespace[4pt]
& \cellcolor{lightgray}2 & \cellcolor{lightgray}Cable Anomaly Detection (Cable-Anomaly Det) & \cellcolor{lightgray}Engineering & \cellcolor{lightgray}Content Recognition & \cellcolor{lightgray}MVTec \cite{MVTec-BergmannBFSS21} & \cellcolor{lightgray}92 & \cellcolor{lightgray}DeSTSeg \cite{DeSTSeg-ZhangL0HSC23} & \cellcolor{lightgray}AP \\

\addlinespace[4pt]
& 3 & Capsule Anomaly Detection (Capsule-Anomaly Det) & Engineering & Content Recognition & MVTec \cite{MVTec-BergmannBFSS21} & 109 & DeSTSeg \cite{DeSTSeg-ZhangL0HSC23} & AP \\

\addlinespace[4pt]
& \cellcolor{lightgray}4 & \cellcolor{lightgray}Carpet Anomaly Detection (Carpet-Anomaly Det) & \cellcolor{lightgray}Engineering & \cellcolor{lightgray}Content Recognition & \cellcolor{lightgray}MVTec \cite{MVTec-BergmannBFSS21} & \cellcolor{lightgray}89 & \cellcolor{lightgray}DeSTSeg \cite{DeSTSeg-ZhangL0HSC23} & \cellcolor{lightgray}AP \\

\addlinespace[4pt]
& 5 & Grid Anomaly Detection (Grid-Anomaly Det) & Engineering & Content Recognition & MVTec \cite{MVTec-BergmannBFSS21} & 57 & DeSTSeg \cite{DeSTSeg-ZhangL0HSC23} & AP \\

\addlinespace[4pt]
& \cellcolor{lightgray}6 & \cellcolor{lightgray}Hazelnut Anomaly Detection (Hazelnut-Anomaly Det) & \cellcolor{lightgray}Engineering & \cellcolor{lightgray}Content Recognition & \cellcolor{lightgray}MVTec \cite{MVTec-BergmannBFSS21} & \cellcolor{lightgray}70 & \cellcolor{lightgray}DeSTSeg \cite{DeSTSeg-ZhangL0HSC23} & \cellcolor{lightgray}AP \\

\addlinespace[4pt]
& 7 & Leather Anomaly Detection (Leather-Anomaly Det) & Engineering & Content Recognition & MVTec \cite{MVTec-BergmannBFSS21} & 92 & DeSTSeg \cite{DeSTSeg-ZhangL0HSC23} & AP \\

\addlinespace[4pt]
& \cellcolor{lightgray}8 & \cellcolor{lightgray}Metal Nut Anomaly Detection (Metal-Anomaly Det) & \cellcolor{lightgray}Engineering & \cellcolor{lightgray}Content Recognition & \cellcolor{lightgray}MVTec \cite{MVTec-BergmannBFSS21} & \cellcolor{lightgray}93 & \cellcolor{lightgray}DeSTSeg \cite{DeSTSeg-ZhangL0HSC23} & \cellcolor{lightgray}AP \\

\addlinespace[4pt]
& 9 & Pill Anomaly Detection (Pill-Anomaly Det) & Engineering & Content Recognition & MVTec \cite{MVTec-BergmannBFSS21} & 141 & DeSTSeg \cite{DeSTSeg-ZhangL0HSC23} & AP \\

\addlinespace[4pt]
& \cellcolor{lightgray}10 & \cellcolor{lightgray}Screw Anomaly Detection (Screw-Anomaly Det) & \cellcolor{lightgray}Engineering & \cellcolor{lightgray}Content Recognition & \cellcolor{lightgray}MVTec \cite{MVTec-BergmannBFSS21} & \cellcolor{lightgray}119 & \cellcolor{lightgray}DeSTSeg \cite{DeSTSeg-ZhangL0HSC23} & \cellcolor{lightgray}AP \\

\addlinespace[4pt]
& 11 & Tile Anomaly Detection (Tile-Anomaly Det) & Engineering & Content Recognition & MVTec \cite{MVTec-BergmannBFSS21} & 84 & DeSTSeg \cite{DeSTSeg-ZhangL0HSC23} & AP \\

\addlinespace[4pt]
& \cellcolor{lightgray}12 & \cellcolor{lightgray}Toothbrush-Anomaly Detection (Toothbrush Anomaly Det) & \cellcolor{lightgray}Engineering & \cellcolor{lightgray}Content Recognition & \cellcolor{lightgray}MVTec \cite{MVTec-BergmannBFSS21} & \cellcolor{lightgray}30 & \cellcolor{lightgray}DeSTSeg \cite{DeSTSeg-ZhangL0HSC23} & \cellcolor{lightgray}AP \\

\addlinespace[4pt]
& 13 & Transistor Anomaly Detection (Transistor-Anomaly Det) & Engineering & Content Recognition & MVTec \cite{MVTec-BergmannBFSS21} & 40 & DeSTSeg \cite{DeSTSeg-ZhangL0HSC23} & AP \\

\addlinespace[4pt]
& \cellcolor{lightgray}14 & \cellcolor{lightgray}Wood Anomaly Detection (Wood-Anomaly Det) & \cellcolor{lightgray}Engineering & \cellcolor{lightgray}Content Recognition & \cellcolor{lightgray}MVTec \cite{MVTec-BergmannBFSS21} & \cellcolor{lightgray}60 & \cellcolor{lightgray}DeSTSeg \cite{DeSTSeg-ZhangL0HSC23} & \cellcolor{lightgray}AP \\

\addlinespace[4pt]
& 15 & Zipper Anomaly Detection (Zipper-Anomaly Detection) & Engineering & Content Recognition & MVTec \cite{MVTec-BergmannBFSS21} & 119 & DeSTSeg \cite{DeSTSeg-ZhangL0HSC23} & AP \\

\addlinespace[4pt]
& \cellcolor{lightgray}16 & \cellcolor{lightgray}Cookie Anomaly Detection (Cookie-Anomaly Det) & \cellcolor{lightgray}Engineering & \cellcolor{lightgray}Content Recognition & \cellcolor{lightgray}Industry Biscuit (Cookie) dataset \cite{Industry-Biscuit} & \cellcolor{lightgray}800 & \cellcolor{lightgray}CLIP \cite{radford2021learning} & \cellcolor{lightgray}Acc \\

\addlinespace[4pt]
& 17 & Macaroni Anomaly Detection (Macaroni-Anomaly Det) & Engineering & Content Recognition & Visual Anomaly Dataset (VisA) \cite{VisA-ZouJPZD22} & 280 & WinCLIP \cite{radford2021learning} & Acc \\

\addlinespace[4pt]
& \cellcolor{lightgray}18 & \cellcolor{lightgray}Marble Surface Anomaly Detection (Marble-Anomaly Det) & \cellcolor{lightgray}Engineering & \cellcolor{lightgray}Content Recognition & \cellcolor{lightgray}Marble Surface Anomaly Detection \cite{Marble-Surface-Anomaly-Detection} & \cellcolor{lightgray}502 & \cellcolor{lightgray}CLIP \cite{radford2021learning} & \cellcolor{lightgray}Acc \\

\addlinespace[4pt]
& 19 & Pcb Anomaly Detection (Pcb-Anomaly Det) & Engineering & Content Recognition & Visual Anomaly Dataset (VisA) \cite{VisA-ZouJPZD22} & 280 & WinCLIP \cite{radford2021learning} & Acc \\

\addlinespace[4pt]
\hline

\addlinespace[4pt]
\multirow{10}{*}{\shortstack[l]{
\#I-C-19  \\  Medical\\ Segmentation  \\  (Med Seg)
} }

& 1 & COVID-19 CT Scan Lesion Segmentation (COVID-19 CT Scan Lesion Seg) & Medicine & Content Recognition & COVID-19 CT scan lesion segmentation dataset \cite{COVID-19-CT-scan-lesion-segmentation} & 290 & Unet \cite{COVID-19-CT-scan-lesion-segmentation-Unet} & mIoU \\

\addlinespace[4pt]
& \cellcolor{lightgray}2 & \cellcolor{lightgray}Hubmap Organ Segmentation (Hubmap Organ Seg) & \cellcolor{lightgray}Medicine & \cellcolor{lightgray}Content Recognition & \cellcolor{lightgray}hubmap\_organ\_512\*512 \cite{Hubmap-Organ-Segmentation} & \cellcolor{lightgray}331 & \cellcolor{lightgray}CLIPseg \cite{CLIPSeg-lueddecke22} & \cellcolor{lightgray}mIoU \\

\addlinespace[4pt]
&3 & Lung segmentation\newline (Lung Seg) & Medicine & Content Recognition & Lung Mask Image Dataset \cite{Lung-Mask-Image-Dataset} & 500 & CLIPseg \cite{CLIPSeg-lueddecke22} & mIoU \\

\addlinespace[4pt]
& \cellcolor{lightgray}4 & \cellcolor{lightgray}Bacteria Segmentation (Bacteria Seg) & \cellcolor{lightgray}Medicine & \cellcolor{lightgray}Content Recognition & \cellcolor{lightgray}Bacteria detection with darkfield microscopy \cite{Bacteria-detection} & \cellcolor{lightgray}366 & \cellcolor{lightgray}CLIPseg \cite{CLIPSeg-lueddecke22} & \cellcolor{lightgray}mIoU \\

\addlinespace[4pt]
& 5 & Brain FLAIR Abnormality Segmentation (Brain Seg) & Medicine & Content Recognition & Brain MRI segmentation \cite{BudaSM19} & 788 & BrainUNet \cite{BrainUNet-buda2019association} & mIoU \\

\addlinespace[4pt]
& \cellcolor{lightgray}6 & \cellcolor{lightgray}Tuberculosis X-ray Segmentation (X-ray Seg) & \cellcolor{lightgray}General & \cellcolor{lightgray}Content Recognition & \cellcolor{lightgray}Chest X-ray Dataset for Tuberculosis Segmentation \cite{Chest-X-ray-Segmentation} & \cellcolor{lightgray}704 & \cellcolor{lightgray}CLIPSeg \cite{CLIPSeg-lueddecke22} & \cellcolor{lightgray}mIoU \\

\addlinespace[4pt]
\hline

\addlinespace[4pt]
\#I-C-20  \newline  Keypoint Detection  \newline  (Keypoint Det)
& 1 & Person Keypoint Detection (Person Keypoint Det) & General & Content Recognition & OCHuman \cite{OCHuman-ZhangLDRCHYHH19} & 600 & ViTPose \cite{ViTPose-XuZZT22} & mAP@0.5 \\

\addlinespace[4pt]
\hline

\addlinespace[4pt]
\multirow{1}{*}{\shortstack[l]{\#I-C-21  \\  Multi-image  \\ Visual Question \\  Answering  \\  (Multi-img VQA)}} 
& 1 & Image Pair Multi-Attribute Question Answering (Img-Pair QA) & General & Content Recognition & NLVR2 \cite{nlvr2-SuhrZZZBA19} & 500 & Mantis \cite{MANTIS-Jiang2024MANTISIM} & Acc \\

\addlinespace[4pt]
& \cellcolor{lightgray}2 & \cellcolor{lightgray}Numerical Dual Image Description Verification (Dual-Img Verify) & \cellcolor{lightgray}General & \cellcolor{lightgray}Reasoning Ability & \cellcolor{lightgray}NLVR2 \cite{nlvr2-SuhrZZZBA19} & \cellcolor{lightgray}500 & \cellcolor{lightgray}Mantis \cite{MANTIS-Jiang2024MANTISIM} & \cellcolor{lightgray}Acc \\

\addlinespace[4pt]
& 3 & Aircraft Model Matching (Aircraft-Model Match) & General & Content Recognition & FGVC-Aircraft \cite{fgvc-aircraft} & 500 & Mantis \cite{MANTIS-Jiang2024MANTISIM} & Acc \\

\addlinespace[4pt]
& \cellcolor{lightgray}4 & \cellcolor{lightgray}Car Model Matching (Car-Model Match) & \cellcolor{lightgray}General & \cellcolor{lightgray}Content Recognition & \cellcolor{lightgray}Stanford-car \cite{stanford-car} & \cellcolor{lightgray}500 & \cellcolor{lightgray}Mantis \cite{MANTIS-Jiang2024MANTISIM} & \cellcolor{lightgray}Acc \\

\addlinespace[4pt]
& 5 & Pose and Activity Consistency Verification (Action-Consist Verify) & General & Action Affective Analysis & CUHK03 \cite{cuhk03} & 500 & mPLUG-Docowl2 \cite{mPLUG-DocOwl2-abs-2409-03420} & Acc \\

\addlinespace[4pt]
& \cellcolor{lightgray}6 & \cellcolor{lightgray}Digital Consistency Comparison (Digital-Consist Compare) & \cellcolor{lightgray}General & \cellcolor{lightgray}Reasoning Ability & \cellcolor{lightgray}MNIST \cite{mnist} & \cellcolor{lightgray}500 & \cellcolor{lightgray}Mantis \cite{MANTIS-Jiang2024MANTISIM} & \cellcolor{lightgray}Acc \\

\addlinespace[4pt]
\hline
\addlinespace[4pt]
\multirow{8}{*}{\shortstack[l]{\#I-C-22 \\ Multimodal \\ Dialogue \\ (MM Dialog)}} 
& 1 & Egocentric Daily Tasks Action Planning (Ego-Action Planning) & General & Planning Ability, Interactive Capability & ALFRED \cite{ALFRED20} & 500 & VPG-C \cite{VPG-C-li2023fine} & ROUGE-L \\

\addlinespace[4pt]
& \cellcolor{lightgray}2 & \cellcolor{lightgray}Context-Aware Embodied Agent Dialogue (Embodied Dialogue) & \cellcolor{lightgray}General & \cellcolor{lightgray}Interactive Capability & \cellcolor{lightgray}ALFRED \cite{ALFRED20} & \cellcolor{lightgray}500 & \cellcolor{lightgray}VPG-C \cite{VPG-C-li2023fine} & \cellcolor{lightgray}ROUGE-L \\

\addlinespace[4pt]
& 3 & Embodied Navigating Visual Question Answering (Navigating VQA) & General & Interactive Capability & ALFRED \cite{ALFRED20} & 500 & MLoEM \cite{MLoEM-WeiSXZLW24} & ROUGE-L \\

\addlinespace[4pt]
& \cellcolor{lightgray}4 & \cellcolor{lightgray}Environment-Based Next-action Description (Next-action Description) & \cellcolor{lightgray}Geography, Earth & \cellcolor{lightgray}Interactive Capability & \cellcolor{lightgray}ALFRED \cite{ALFRED20} & \cellcolor{lightgray}500 & \cellcolor{lightgray}LLaVA-NeXT-Interleave \cite{li2024llava} & \cellcolor{lightgray}ROUGE-L \\

\addlinespace[4pt]
& 5 & Multimodal Decision-making Reasoning (Decision-making Reasoning) & General & Interactive Capability & ALFRED \cite{ALFRED20} & 500 & VPG-C \cite{VPG-C-li2023fine} & ROUGE-L \\

\addlinespace[4pt]
& \cellcolor{lightgray}6 & \cellcolor{lightgray}Comic Dialogue Completion (Comic-Dialog Complete) & \cellcolor{lightgray}Art & \cellcolor{lightgray}Interactive Capability & \cellcolor{lightgray}COMICS-Dialogue \cite{IyyerComics2016} & \cellcolor{lightgray}500 & \cellcolor{lightgray}Mantis \cite{MANTIS-Jiang2024MANTISIM} & \cellcolor{lightgray}Acc \\

\addlinespace[4pt]
\hline

\addlinespace[4pt]
\multirow{1}{*}{\shortstack[l]{\#I-C-23 \\ Multimodal \\ Reasoning \\ (MM Reason)}} 
& 1 & Fashion Concept Visual Question Answering (Fashion VQA) & Fashion & Content Recognition & Fashion200K \cite{Fashion200K-HanWHZZLZD17} & 500 & MLoEM \cite{MLoEM-WeiSXZLW24} & Acc \\

\addlinespace[4pt]
& \cellcolor{lightgray}2 & \cellcolor{lightgray}Cloth Color Visual Question Answering (Cloth-Color VQA) & \cellcolor{lightgray}Fashion & \cellcolor{lightgray}Reasoning Ability & \cellcolor{lightgray}Fashion200K \cite{Fashion200K-HanWHZZLZD17} & \cellcolor{lightgray}500 & \cellcolor{lightgray}MLoEM \cite{MLoEM-WeiSXZLW24} & \cellcolor{lightgray}Acc \\

\addlinespace[4pt]
& 3 & Multi-Image Visual Entailment Reasoning (Multi-Img Entailment Reason) & General & Reasoning Ability & NLVR2 \cite{nlvr2-SuhrZZZBA19} & 500 & MLoEM \cite{MLoEM-WeiSXZLW24} & Acc \\

\addlinespace[4pt]
& \cellcolor{lightgray}4 & \cellcolor{lightgray}Visual Step Completion (Visual-Step Cloze) & \cellcolor{lightgray}General & \cellcolor{lightgray}Reasoning Ability & \cellcolor{lightgray}RecipeQA-VisualCloze \cite{RecipeQA-YagciogluEEI18} & \cellcolor{lightgray}500 & \cellcolor{lightgray}LLaVA-NeXT-Interleave \cite{li2024llava} & \cellcolor{lightgray}Acc \\

\addlinespace[4pt]
& 5 & Recipe Ingredient Description (Recipe Description) & General & Reasoning Ability & RecipeQA-VisualCloze \cite{RecipeQA-YagciogluEEI18} & 500 & Mantis \cite{MANTIS-Jiang2024MANTISIM} & Acc \\

\addlinespace[4pt]
& \cellcolor{lightgray}6 & \cellcolor{lightgray}Visual Step Matching Reasoning (Step-Matching Reason) & \cellcolor{lightgray}General & \cellcolor{lightgray}Reasoning Ability & \cellcolor{lightgray}RecipeQA-ImageCoherence \cite{RecipeQA-YagciogluEEI18} & \cellcolor{lightgray}500 & \cellcolor{lightgray}MLoEM \cite{MLoEM-WeiSXZLW24} & \cellcolor{lightgray}Acc \\

\addlinespace[4pt]
& 7 & Industrial Inspection Multi-Image Reasoning\newline (Inspec-Multi-Img Reason) & Engineering & Reasoning Ability & VISION \cite{VISION-ShullaniFISP17} & 500 & MLoEM \cite{MLoEM-WeiSXZLW24} & Acc \\

\addlinespace[4pt]
& \cellcolor{lightgray}8 & \cellcolor{lightgray}Property Coherence Reasoning (Property Reason) & \cellcolor{lightgray}General & \cellcolor{lightgray}Reasoning Ability & \cellcolor{lightgray}MIT-States-PropertyCoherence \cite{MIT-States-IsolaLA15} & \cellcolor{lightgray}500 & \cellcolor{lightgray}MLoEM \cite{MLoEM-WeiSXZLW24} & \cellcolor{lightgray}Acc \\

\addlinespace[4pt]
& 9 & Recipe Completion (Recipe Completion) & General & Reasoning Ability & RecipeQA-TextCloze \cite{RecipeQA-YagciogluEEI18} & 500 & VPG-C \cite{VPG-C-li2023fine} & Acc \\

\addlinespace[4pt]
& \cellcolor{lightgray}10 & \cellcolor{lightgray}Comic Panel VQA (Comic-Panel VQA) & \cellcolor{lightgray}Art & \cellcolor{lightgray}Reasoning Ability & \cellcolor{lightgray}COMICS-Panel \cite{IyyerComics2016} & \cellcolor{lightgray}500 & \cellcolor{lightgray}VPG-C \cite{VPG-C-li2023fine} & \cellcolor{lightgray}Acc \\

\addlinespace[4pt]
& 11 & Abduction-based VQA (Abduction VQA) & General & Reasoning Ability & VizWiz \cite{VizWiz-Gurari0SGLGLB18} & 500 & VPG-C \cite{VPG-C-li2023fine} & Acc \\

\addlinespace[4pt]
& \cellcolor{lightgray}12 & \cellcolor{lightgray}Dual-Image Causal-and-Effect Reasoning (Dual-Img Causal Reason) & \cellcolor{lightgray}General & \cellcolor{lightgray}Reasoning Ability & \cellcolor{lightgray}VizWiz \cite{VizWiz-Gurari0SGLGLB18} & \cellcolor{lightgray}500 & \cellcolor{lightgray}MLoEM \cite{MLoEM-WeiSXZLW24} & \cellcolor{lightgray}Acc \\

\addlinespace[4pt]
& 13 & Driving Recording Reasoning (Drive-Record Reason) & Geography & Reasoning Ability & nuScenes \cite{nuScenes-CaesarBLVLXKPBB20} & 500 & VPG-C \cite{VPG-C-li2023fine} & Acc \\

\addlinespace[4pt]
& \cellcolor{lightgray}14 & \cellcolor{lightgray}Attribute Congruity Reasoning (Attri-Congruity Reason) & \cellcolor{lightgray}General & \cellcolor{lightgray}Reasoning Ability & \cellcolor{lightgray}MIT-States-StateCoherence \cite{MIT-States-IsolaLA15} & \cellcolor{lightgray}500 & \cellcolor{lightgray}MLoEM \cite{MLoEM-WeiSXZLW24} & \cellcolor{lightgray}Acc \\

\addlinespace[4pt]
\hline

\addlinespace[4pt]
\multirow{1}{*}{\shortstack[l]{\#I-C-24 \\ Object Counting \\ Object Count}} 
& 1 & Crowd Counting (Crowd Count) & General & Content Recognition, Reasoning Ability & Crowd Counting \cite{Crowd-Counting-LoyGX13} & 650 & CLIPCount \cite{CLIP-Count-JiangLC23} & MAE \\

\addlinespace[4pt]
& \cellcolor{lightgray}2 & \cellcolor{lightgray}Face Counting (Face Count) & \cellcolor{lightgray}General & \cellcolor{lightgray}Reasoning Ability & \cellcolor{lightgray}Count the number of Faces present in an Image \cite{Count-number-Faces} & \cellcolor{lightgray}650 & \cellcolor{lightgray}CLIPCount \cite{CLIP-Count-JiangLC23} & \cellcolor{lightgray}MAE \\

\addlinespace[4pt]
& 3 & Galaxy Counting (Galaxy Count) & Physics & Reasoning Ability & FITS Images for Object Counting and Detection \cite{FITS-Images} & 597 & CLIPCount \cite{CLIP-Count-JiangLC23} & MAE \\

\addlinespace[4pt]
& \cellcolor{lightgray}4 & \cellcolor{lightgray}Seed Counting (Seed Count) & \cellcolor{lightgray}General & \cellcolor{lightgray}Content Recognition, Reasoning Ability & \cellcolor{lightgray}seeds counting \cite{Seeds-Counting} & \cellcolor{lightgray}141 & \cellcolor{lightgray}CLIPCount \cite{CLIP-Count-JiangLC23} & \cellcolor{lightgray}MAE \\

\addlinespace[4pt]
& 5 & Vehicle Counting (Vehicle Count) & General & Content Recognition, Reasoning Ability & CARPK \cite{CARPK-HsiehLH17} & 459 & CLIPCount \cite{CLIP-Count-JiangLC23} & MAE \\

\addlinespace[4pt]
& \cellcolor{lightgray}6 & \cellcolor{lightgray}Mall Crowd Counting (Mall-Crowd Count) & \cellcolor{lightgray}General & \cellcolor{lightgray}Content Recognition, Reasoning Ability & \cellcolor{lightgray}Mall Dataset \cite{Mall-Dataset-LoyGX13} & \cellcolor{lightgray}500 & \cellcolor{lightgray}CLIP \cite{radford2021learning} & \cellcolor{lightgray}Acc \\

\addlinespace[4pt]
& 7 & Paperclips Counting (Paperclip Count) & General & Content Recognition, Reasoning Ability & Count the Paperclips \cite{Count-Paperclips} & 500 & CLIPCount \cite{CLIP-Count-JiangLC23} & MAE \\

\addlinespace[4pt]
& \cellcolor{lightgray}8 & \cellcolor{lightgray}Wheat Head Counting (Wheat-Head Count) & \cellcolor{lightgray}General & \cellcolor{lightgray}Content Recognition, Reasoning Ability & \cellcolor{lightgray}Global Wheat Head Dataset 2021 \cite{david2020global} & \cellcolor{lightgray}500 & \cellcolor{lightgray}CLIPCount \cite{CLIP-Count-JiangLC23} & \cellcolor{lightgray}MAE \\

\addlinespace[4pt]
& 9 & Pipe Counting (Pipe Count) & General & Content Recognition, Reasoning Ability & Object Counting Using Pipes Dataset \cite{Counting-Pipe} & 240 & CLIPCount \cite{CLIP-Count-JiangLC23} & MAE \\

\addlinespace[4pt]
\hline

\addlinespace[4pt]
\#I-C-25  \newline  Multimodal Neural \newline  Translation  \newline  (NMT)
& 1 & Multimodal Neural Translation (NMT) & Linguistics & Reasoning Ability & Multi30k \cite{Multimodal-Neural-Translation} & 500 & mPLUG-Docowl2 \cite{mPLUG-DocOwl2-abs-2409-03420} & Acc \\

\addlinespace[4pt]
\hline

\addlinespace[4pt]
\multirow{1}{*}{\shortstack[l]{\#I-C-26 \\ Object Detection \\ (Object Det)}} 
& 1 & Car License Plate Detection (Car-License Det) & General & Content Recognition & Car License Plate Detection \cite{Car-License-Plate-Detection} & 433 & OWL-VIT \cite{OWL-VIT-abs-2205-06230} & mAP@0.5 \\

\addlinespace[4pt]
& \cellcolor{lightgray}2 & \cellcolor{lightgray}Face Mask Detection (Face-Mask Det) & \cellcolor{lightgray}General & \cellcolor{lightgray}Content Recognition & \cellcolor{lightgray}Face Mask Detection \cite{make-ml} & \cellcolor{lightgray}500 & \cellcolor{lightgray}YOLO-World \cite{YOLO-World-ChengSG0WS24} & \cellcolor{lightgray}mAP@0.5 \\

\addlinespace[4pt]
& 3 & Far Vehicle Detection (Far-Vehicle Det) & General & Content Recognition & Far Vehicle Detection \cite{Far-Vehicle-Det} & 221 & OWL-VIT \cite{OWL-VIT-abs-2205-06230} & mAP@0.5 \\

\addlinespace[4pt]
& \cellcolor{lightgray}4 & \cellcolor{lightgray}Small Object Detection (Small-Obj Det) & \cellcolor{lightgray}General & \cellcolor{lightgray}Content Recognition, Commonsense Knowledge & \cellcolor{lightgray}Grand Dataset for small object detection \cite{small-object-detection} & \cellcolor{lightgray}512 & \cellcolor{lightgray}OWL-VIT \cite{OWL-VIT-abs-2205-06230} & \cellcolor{lightgray}mAP@0.5 \\

\addlinespace[4pt]
& 5 & Traffic Signs Detection (Traffic-Signs Det) & General & Content Recognition, Commonsense Knowledge & Traffic Signs Detection \cite{Traffic-Signs-Detection} & 634 & OWL-VIT \cite{OWL-VIT-abs-2205-06230} & mAP@0.5 \\

\addlinespace[4pt]
& \cellcolor{lightgray}6 & \cellcolor{lightgray}Bone Fracture Detection (Bone-Fracture Det) & \cellcolor{lightgray}Medicine & \cellcolor{lightgray}Content Recognition & \cellcolor{lightgray}Bone Fracture Detection \cite{Bone-Fracture-Detection} & \cellcolor{lightgray}500 & \cellcolor{lightgray}OWLVIT \cite{OWLVIT-abs-2205-06230} & \cellcolor{lightgray}DICE \\

\addlinespace[4pt]
& 7 & Aircraft Detection (Aircraft Det) & General & Content Recognition & Aircraft Detection \cite{drone-detection-new-peksv_dataset} & 603 & OWLVIT \cite{OWLVIT-abs-2205-06230} & DICE \\

\addlinespace[4pt]
& \cellcolor{lightgray}8 & \cellcolor{lightgray}Medical Device Detection (Medical-Device Det) & \cellcolor{lightgray}Medicine & \cellcolor{lightgray}Content Recognition & \cellcolor{lightgray}Medical Device Detection \cite{hospital-i5qoz_dataset} & \cellcolor{lightgray}500 & \cellcolor{lightgray}GLIP-T \cite{GLIP-li2021grounded} & \cellcolor{lightgray}mAP@0.5 \\

\addlinespace[4pt]
& 9 & Wound Detection (Wound det) & Medicine & Content Recognition & Wound detection \cite{wound-care-oufqn_dataset} & 500 & GLIP-T \cite{GLIP-li2021grounded} & mAP@0.5 \\

\addlinespace[4pt]
& \cellcolor{lightgray}10 & \cellcolor{lightgray}Gauge Detection (Gauge Det) & \cellcolor{lightgray}General & \cellcolor{lightgray}Content Recognition & \cellcolor{lightgray}Gauge Detection \cite{gauge-detection-cohbz_dataset} & \cellcolor{lightgray}500 & \cellcolor{lightgray}GLIP-T \cite{GLIP-li2021grounded} & \cellcolor{lightgray}mAP@0.5 \\

\addlinespace[4pt]
& 11 & Parking Space Detection (Parking-Space Det) & General & Content Recognition & Parking Space Detection \cite{car-parking-occupation_dataset} & 500 & GLIP-T \cite{GLIP-li2021grounded} & mAP@0.5 \\

\addlinespace[4pt]
& \cellcolor{lightgray}12 & \cellcolor{lightgray}Tree Detection (Tree det) & \cellcolor{lightgray}Geography & \cellcolor{lightgray}Content Recognition & \cellcolor{lightgray}Tree detection \cite{tre-o6dcm_dataset} & \cellcolor{lightgray}500 & \cellcolor{lightgray}GLIP-T \cite{GLIP-li2021grounded} & \cellcolor{lightgray}mAP@0.5 \\

\addlinespace[4pt]
& 13 & Traffic Light Detection (Traffic-Light Det) & General & Content Recognition & Traffic Light Detection \cite{traffic_light-28zwo_dataset} & 500 & GLIP-T \cite{GLIP-li2021grounded} & mAP@0.5 \\

\addlinespace[4pt]
& \cellcolor{lightgray}14 & \cellcolor{lightgray}rock-paper-scissors-gesture-detection (Gesture Det) & \cellcolor{lightgray}General & \cellcolor{lightgray}Content Recognition & \cellcolor{lightgray}rock-paper-scissors-gesture-detection \cite{rock-paper-scissors-sxsw_dataset} & \cellcolor{lightgray}500 & \cellcolor{lightgray}GLIP-T \cite{GLIP-li2021grounded} & \cellcolor{lightgray}mAP@0.5 \\

\addlinespace[4pt]
& 15 & Price Tag Detection and Text Grounding (Price-Tag Det) & General & Content Recognition & Price Tag Detection and Text Grounding \cite{pricetag-4d8lm_dataset} & 500 & YOLO-World \cite{YOLO-World-ChengSG0WS24} & mAP@0.5 \\

\addlinespace[4pt]
& \cellcolor{lightgray}16 & \cellcolor{lightgray}Retail Object Detection (Retail-Obj Det) & \cellcolor{lightgray}General & \cellcolor{lightgray}Content Recognition & \cellcolor{lightgray}Retail object detection \cite{retail-object-detection-bqtzf_dataset} & \cellcolor{lightgray}500 & \cellcolor{lightgray}YOLO-World \cite{YOLO-World-ChengSG0WS24} & \cellcolor{lightgray}mAP@0.5 \\

\addlinespace[4pt]
& 17 & Exit Sign Detection (Exit-Sign Det) & General & Content Recognition & Exit Sign Detection \cite{exit-sign-extended-version_dataset} & 500 & YOLO-World \cite{YOLO-World-ChengSG0WS24} & mAP@0.5 \\

\addlinespace[4pt]
& \cellcolor{lightgray}18 & \cellcolor{lightgray}Handwriting Component Detection (Handwrite-Component Det) & \cellcolor{lightgray}General & \cellcolor{lightgray}Content Recognition & \cellcolor{lightgray}Handwriting Component Detection \cite{handwritten-text-recognition-wb2o4_dataset} & \cellcolor{lightgray}500 & \cellcolor{lightgray}YOLO-World \cite{YOLO-World-ChengSG0WS24} & \cellcolor{lightgray}mAP@0.5 \\

\addlinespace[4pt]
& 19 & Sign Language Detection (Sign-Lang Det) & Linguistics & Content Recognition & Sign Language Detection \cite{signlanguage-5kmtf_dataset} & 500 & YOLO-World \cite{YOLO-World-ChengSG0WS24} & mAP@0.5 \\

\addlinespace[4pt]
& \cellcolor{lightgray}20 & \cellcolor{lightgray}football player detection (Football-Player Det) & \cellcolor{lightgray}General & \cellcolor{lightgray}Content Recognition & \cellcolor{lightgray}football player detection \cite{football-video-tracking-project_dataset} & \cellcolor{lightgray}500 & \cellcolor{lightgray}YOLO-World \cite{YOLO-World-ChengSG0WS24} & \cellcolor{lightgray}mAP@0.5 \\

\addlinespace[4pt]
& 21 & Pest Detection (Pest Det) & Biology & Content Recognition & Pest Detection \cite{pest-detection-m0inx_dataset} & 500 & YOLO-World \cite{YOLO-World-ChengSG0WS24} & mAP@0.5 \\

\addlinespace[4pt]
& \cellcolor{lightgray}22 & \cellcolor{lightgray}Underwater garbage Detection (Garbage Det) & \cellcolor{lightgray}General & \cellcolor{lightgray}Content Recognition & \cellcolor{lightgray}Underwater garbage Detection \cite{underwater-garbage-detection-d4dnn_dataset} & \cellcolor{lightgray}500 & \cellcolor{lightgray}YOLO-World \cite{YOLO-World-ChengSG0WS24} & \cellcolor{lightgray}mAP@0.5 \\

\addlinespace[4pt]
& 23 & Food Detection (Food Det) & General & Content Recognition & Food Detection \cite{ingredient-detection-d6nwz_dataset} & 500 & YOLO-World \cite{YOLO-World-ChengSG0WS24} & mAP@0.5 \\

\addlinespace[4pt]
& \cellcolor{lightgray}24 & \cellcolor{lightgray}Circuit Elements Detection (Circuit Det) & \cellcolor{lightgray}General & \cellcolor{lightgray}Content Recognition & \cellcolor{lightgray}Circuit elements \cite{circuit-elements_dataset} & \cellcolor{lightgray}500 & \cellcolor{lightgray}YOLO-World \cite{YOLO-World-ChengSG0WS24} & \cellcolor{lightgray}mAP@0.5 \\

\addlinespace[4pt]
& 25 & Tabular Data Detection (Tabular Det) & General & Content Recognition & Tabular data detection \cite{tabular-data-wf9uh_dataset} & 500 & OWL-VIT \cite{OWL-VIT-abs-2205-06230} & mAP@0.5 \\

\addlinespace[4pt]
& \cellcolor{lightgray}26 & \cellcolor{lightgray}Radio-Spectrogram Detection (Radio Det) & \cellcolor{lightgray}General & \cellcolor{lightgray}Content Recognition & \cellcolor{lightgray}Radio spectrogram detection \cite{radio-signal_dataset} & \cellcolor{lightgray}500 & \cellcolor{lightgray}OWL-VIT \cite{OWL-VIT-abs-2205-06230} & \cellcolor{lightgray}mAP@0.5 \\

\addlinespace[4pt]
& 27 & Empty Shelf Detection (Empty-Shelf Det) & General & Content Recognition & empty-shelf-detection \cite{empty-shelf-detector-wt4im_dataset} & 500 & OWL-VIT \cite{OWL-VIT-abs-2205-06230} & mAP@0.5 \\

\addlinespace[4pt]
& \cellcolor{lightgray}28 & \cellcolor{lightgray}Tomato Detection (Tomato Det) & \cellcolor{lightgray}General & \cellcolor{lightgray}Content Recognition & \cellcolor{lightgray}Tomato Detection \cite{Tomato-Dataset} & \cellcolor{lightgray}895 & \cellcolor{lightgray}GroundingDINO \cite{Grounding-DINO-LiuZRLZYJLYSZZ24} & \cellcolor{lightgray}AP@0.6 \\

\addlinespace[4pt]
& 29 & Drone Detection (Drone Det) & General & Content Recognition & Drone Detection \cite{Drone-Detection} & 596 & GroundingDINO \cite{Grounding-DINO-LiuZRLZYJLYSZZ24} & AP@0.7 \\

\addlinespace[4pt]
& \cellcolor{lightgray}30 & \cellcolor{lightgray}Bee Detection (Bee Det) & \cellcolor{lightgray}General & \cellcolor{lightgray}Content Recognition & \cellcolor{lightgray}Bee Detection \cite{Bee-Detection-Dataset} & \cellcolor{lightgray}836 & \cellcolor{lightgray}GroundingDINO \cite{Grounding-DINO-LiuZRLZYJLYSZZ24} & \cellcolor{lightgray}AP@0.8 \\

\addlinespace[4pt]
& 31 & Cat Faces Detection (Cat Det) & General & Content Recognition & Cat Faces Detection \cite{Cat-Faces-Detection} & 500 & GroundingDINO \cite{Grounding-DINO-LiuZRLZYJLYSZZ24} & AP@0.9 \\

\addlinespace[4pt]
& \cellcolor{lightgray}32 & \cellcolor{lightgray}Bird Detection (Bird Det) & \cellcolor{lightgray}General & \cellcolor{lightgray}Content Recognition & \cellcolor{lightgray}CUB 200 Bird Species XML Detection Dataset \cite{CUB-200-Bird-Species} & \cellcolor{lightgray}500 & \cellcolor{lightgray}GroundingDINO \cite{Grounding-DINO-LiuZRLZYJLYSZZ24} & \cellcolor{lightgray}AP@0.75 \\

\addlinespace[4pt]
& 33 & Deepfish Detection (Deepfish Det) & General & Content Recognition & Deep Fish Object Detection \cite{Deep-Fish-Object-Detection} & 500 & GroundingDINO \cite{Grounding-DINO-LiuZRLZYJLYSZZ24} & AP@0.5 \\

\addlinespace[4pt]
& \cellcolor{lightgray}34 & \cellcolor{lightgray}Raccoon Detection (Raccoon Det) & \cellcolor{lightgray}General & \cellcolor{lightgray}Content Recognition & \cellcolor{lightgray}Trash Panda Detection \cite{Trash-Panda-Detection} & \cellcolor{lightgray}196 & \cellcolor{lightgray}GroundingDINO \cite{Grounding-DINO-LiuZRLZYJLYSZZ24} & \cellcolor{lightgray}AP@0.5 \\

\addlinespace[4pt]
& 35 & Ship Detection (Ship Det) & General & Content Recognition & Ship Detection from Aerial Images \cite{Ships-Data} & 621 & GroundingDINO \cite{Grounding-DINO-LiuZRLZYJLYSZZ24} & AP@0.5 \\

\addlinespace[4pt]
& \cellcolor{lightgray}36 & \cellcolor{lightgray}Weed Detection (Weed Det) & \cellcolor{lightgray}General & \cellcolor{lightgray}Content Recognition & \cellcolor{lightgray}Weed Detection \cite{Weed-Detection} & \cellcolor{lightgray}500 & \cellcolor{lightgray}GroundingDINO \cite{Grounding-DINO-LiuZRLZYJLYSZZ24} & \cellcolor{lightgray}AP@0.5 \\

\addlinespace[4pt]
& 37 & Radar Ship Detection (Radar-Ship Det) & General & Content Recognition & SARscope \cite{SARscope} & 672 & GLIP \cite{GLIP-li2021grounded} & mAP@0.5 \\

\addlinespace[4pt]
\hline

\addlinespace[4pt]
\multirow{9}{*}{\shortstack[l]{\#I-C-27 \\ Object Matting \\ (Object Mat)}} 
& 1 & Animal Image Matting (Animal-Img Mat) & Animal & Content Recognition & AM2K \cite{li2022bridging} & 200 & MAM \cite{MAM-li2023matting} & SAD \\

\addlinespace[4pt]
& \cellcolor{lightgray}2 & \cellcolor{lightgray}Portrait Matting (Portrait Mat) & \cellcolor{lightgray}General & \cellcolor{lightgray}Content Recognition & \cellcolor{lightgray}PM-10K \cite{PM-10K} & \cellcolor{lightgray}500 & \cellcolor{lightgray}MAM \cite{MAM-li2023matting} & \cellcolor{lightgray}SAD \\

\addlinespace[4pt]
& 3 & Text Manipulation Region Matting (Text-Region Mat) & Ethics & Content Recognition & Text Manipulation Detection \cite{Text-Manipulation-Detection} & 500 & CLIPseg \cite{CLIPSeg-lueddecke22} & mIoU \\

\addlinespace[4pt]
& \cellcolor{lightgray}4 & \cellcolor{lightgray}Person Hair Matting (Person-Hair Mat) & \cellcolor{lightgray}General & \cellcolor{lightgray}Content Recognition & \cellcolor{lightgray}Manual Construction \cite{SketchHairSalon-xiao2021} & \cellcolor{lightgray}500 & \cellcolor{lightgray}MAM \cite{MAM-li2023matting} & \cellcolor{lightgray}MSE \\

\addlinespace[4pt]
\hline

\addlinespace[4pt]
\multirow{10}{*}{\shortstack[l]{\#I-C-28 \\ Object Recognition \\ (Object Recog) }} 
& 1 & Animal Recognition (Animal Recog) & Animal & Content Recognition & Animals-10 \cite{Animals-10} & 510 & CLIP \cite{radford2021learning} & Acc \\

\addlinespace[4pt]
& \cellcolor{lightgray}2 & \cellcolor{lightgray}Big Cat Image Recognition (Big Cat Image Recog) & \cellcolor{lightgray}General & \cellcolor{lightgray}Content Recognition & \cellcolor{lightgray}Big Cats Image Classification Dataset \cite{Big-Cats-Image} & \cellcolor{lightgray}738 & \cellcolor{lightgray}CLIP \cite{radford2021learning} & \cellcolor{lightgray}Acc \\

\addlinespace[4pt]
& 3 & Fruit Recognition (Fruit Recog) & General & Content Recognition & Fruit Recognition \cite{Fruit-Recognition} & 660 & CLIP \cite{radford2021learning} & Acc \\

\addlinespace[4pt]
& \cellcolor{lightgray}4 & \cellcolor{lightgray}Indonesian Wayang Type Recognition (Indonesian Recog) & \cellcolor{lightgray}Culture & \cellcolor{lightgray}Content Recognition & \cellcolor{lightgray}Indonesian Wayang Types \cite{Indonesian-Wayang-Types} & \cellcolor{lightgray}233 & \cellcolor{lightgray}CLIP \cite{radford2021learning} & \cellcolor{lightgray}Acc \\

\addlinespace[4pt]
& 5 & Vegetable Recognition (Vegetable Recog) & General & Content Recognition, Commonsense Knowledge & Vegetable Image Dataset \cite{Vegetable-Image-Dataset} & 600 & CLIP \cite{radford2021learning} & Acc \\

\addlinespace[4pt]
& \cellcolor{lightgray}6 & \cellcolor{lightgray}Car Brands Recognition (Car-Brand Recog) & \cellcolor{lightgray}General & \cellcolor{lightgray}Content Recognition & \cellcolor{lightgray}Car-Logo-Dataset \cite{Car_Logo_Dataset} & \cellcolor{lightgray}480 & \cellcolor{lightgray}CLIP \cite{radford2021learning} & \cellcolor{lightgray}Acc \\

\addlinespace[4pt]
& 7 & Damaged Egg Recognition (Damaged-Egg Recog) & General & Content Recognition & Egg Image Dataset \cite{Egg-Image-Dataset} & 412 & OWLVIT \cite{OWLVIT-abs-2205-06230} & Acc \\

\addlinespace[4pt]
& \cellcolor{lightgray}8 & \cellcolor{lightgray}Dog Breeds Recognition (Dog-Breed Recog) & \cellcolor{lightgray}General & \cellcolor{lightgray}Content Recognition & \cellcolor{lightgray}Dog Breeds \cite{Dog-Breeds} & \cellcolor{lightgray}467 & \cellcolor{lightgray}CLIP \cite{radford2021learning} & \cellcolor{lightgray}Acc \\

\addlinespace[4pt]
& 9 & Garbage Classification (Garbage Cls) & General & Content Recognition & Garbage Classification \cite{Garbage-Classification} & 498 & CLIP \cite{radford2021learning} & Acc \\

\addlinespace[4pt]
& \cellcolor{lightgray}10 & \cellcolor{lightgray}Mammal Recognition (Mammal Recog) & \cellcolor{lightgray}General & \cellcolor{lightgray}Content Recognition & \cellcolor{lightgray}Mammal Recognition \cite{Mammals-classification} & \cellcolor{lightgray}495 & \cellcolor{lightgray}OWLVIT \cite{OWLVIT-abs-2205-06230} & \cellcolor{lightgray}Acc \\

\addlinespace[4pt]
& 11 & Utensil Recognition (Utensil Recog) & General & Content Recognition & Utensil Image Recognition \cite{Utensil-Image-Recognition} & 475 & CLIP \cite{radford2021learning} & Acc \\

\addlinespace[4pt]
& \cellcolor{lightgray}12 & \cellcolor{lightgray}Boat Types Recognition (Boat-Types Recog) & \cellcolor{lightgray}General & \cellcolor{lightgray}Content Recognition & \cellcolor{lightgray}Boat Types Recognition \cite{Boat-Types-Recognition} & \cellcolor{lightgray}640 & \cellcolor{lightgray}CLIP \cite{radford2021learning} & \cellcolor{lightgray}Acc \\

\addlinespace[4pt]
& 13 & Butterfly Recognition (Butterfly Recog) & General & Content Recognition & Butterfly Dataset \cite{Wang09} & 832 & CLIPCount \cite{CLIP-Count-JiangLC23} & Acc \\

\addlinespace[4pt]
& \cellcolor{lightgray}14 & \cellcolor{lightgray}Fish Recognition (Fish Recog) & \cellcolor{lightgray}General & \cellcolor{lightgray}Content Recognition, Commonsense Knowledge & \cellcolor{lightgray}Fish Recognition Ground-Truth data \cite{Fish-Recognition} & \cellcolor{lightgray}456 & \cellcolor{lightgray}CLIP \cite{radford2021learning} & \cellcolor{lightgray}Acc \\

\addlinespace[4pt]
& 15 & Ball Image Classification (Ball Cls) & General & Content Recognition, Commonsense Knowledge & Ball Classification \cite{Ball_Classification} & 300 & CaSED \cite{CaSED-conti2023vocabularyfree} & Acc \\

\addlinespace[4pt]
& \cellcolor{lightgray}16 & \cellcolor{lightgray}Flower Image Classification (Flower Cls) & \cellcolor{lightgray}General & \cellcolor{lightgray}Content Recognition, Commonsense Knowledge & \cellcolor{lightgray}Flower Classification \cite{flower_classification} & \cellcolor{lightgray}300 & \cellcolor{lightgray}CaSED \cite{CaSED-conti2023vocabularyfree} & \cellcolor{lightgray}Acc \\

\addlinespace[4pt]
& 17 & Food Image Classification (Food Cls) & General & Content Recognition, Commonsense Knowledge & Data Food Category Classification \cite{data-food-category-classification} & 300 & CaSED \cite{CaSED-conti2023vocabularyfree} & Acc \\

\addlinespace[4pt]
& \cellcolor{lightgray}18 & \cellcolor{lightgray}Material Image Classification (Material Cls) & \cellcolor{lightgray}General & \cellcolor{lightgray}Content Recognition, Commonsense Knowledge & \cellcolor{lightgray}Material Classification \cite{Material_Classification} & \cellcolor{lightgray}150 & \cellcolor{lightgray}CaSED \cite{CaSED-conti2023vocabularyfree} & \cellcolor{lightgray}Acc \\

\addlinespace[4pt]
& 19 & Plant Image Classification (Plant Cls) & General & Content Recognition, Commonsense Knowledge & Plant Classification \cite{plant_classification_dataset} & 300 & CaSED \cite{CaSED-conti2023vocabularyfree} & Acc \\

\addlinespace[4pt]
& 20 & Trash Image Classification (Trash Cls) & General & Content Recognition, Commonsense Knowledge & Trash Classification \cite{trash_classification} & 300 & CaSED \cite{CaSED-conti2023vocabularyfree} & Acc \\

\addlinespace[4pt]
& \cellcolor{lightgray}21 & \cellcolor{lightgray}Vehicle Image Classification (Vehicle Cls) & \cellcolor{lightgray}General & \cellcolor{lightgray}Content Recognition, Commonsense Knowledge & \cellcolor{lightgray}Vehicle Classification \cite{vehicle-classification} & \cellcolor{lightgray}300 & \cellcolor{lightgray}CaSED \cite{CaSED-conti2023vocabularyfree} & \cellcolor{lightgray}Acc \\

\addlinespace[4pt]
& 22 & Bird Species Recognition (Bird Recog) & General & Content Recognition, Commonsense Knowledge & Bird Species Classification \cite{Bird-Species-Classification} & 307 & CLIP \cite{radford2021learning} & Acc \\

\addlinespace[4pt]
& \cellcolor{lightgray}23 & \cellcolor{lightgray}Chinese Fine Art Recognition (Zh-Art Recog) & \cellcolor{lightgray}Art & \cellcolor{lightgray}Commonsense Knowledge & \cellcolor{lightgray}Chinese Fine Art \cite{Chinese-Fine-Art} & \cellcolor{lightgray}398 & \cellcolor{lightgray}CLIP \cite{radford2021learning} & \cellcolor{lightgray}Acc \\

\addlinespace[4pt]
& 24 & Famous Iconic Women Recognition (Iconic-Women Recog) & Humanities, History & Commonsense Knowledge & Famous Iconic Women \cite{Famous-Iconic-Women} & 225 & CLIP \cite{radford2021learning} & Acc \\

\addlinespace[4pt]
& \cellcolor{lightgray}25 & \cellcolor{lightgray}Text Manipulation Classification (Text-Manipulation Cls) & \cellcolor{lightgray}Ethics & \cellcolor{lightgray}Content Recognition & \cellcolor{lightgray}Text Manipulation Classification \cite{Text-Manipulation-Detection} & \cellcolor{lightgray}500 & \cellcolor{lightgray}CLIP \cite{radford2021learning} & \cellcolor{lightgray}Acc \\

\addlinespace[4pt]
& 26 & Gender Recognition (Gender Recog) & General & Content Recognition & Gender Detection \& Classification - Face Dataset \cite{Face-Dataset} & 300 & CLIP \cite{radford2021learning} & Acc \\

\addlinespace[4pt]
& \cellcolor{lightgray}27 & \cellcolor{lightgray}Infrared/Thermal Image Classification (Infrared Image Cls) & \cellcolor{lightgray}Infrared & \cellcolor{lightgray}Content Recognition & \cellcolor{lightgray}Radar Threat Object Classification \cite{Radar-Threat-Object} & \cellcolor{lightgray}521 & \cellcolor{lightgray}Mantis \cite{MANTIS-Jiang2024MANTISIM} & \cellcolor{lightgray}Acc \\

\addlinespace[4pt]
& 28 & Waste Item Image Classification (Waste Cls) & General & Content Recognition, Commonsense Knowledge & Manual Construction \cite{RealWaste-Image-Classification} & 540 & CLIP \cite{radford2021learning} & Acc \\

\addlinespace[4pt]
& \cellcolor{lightgray}29 & \cellcolor{lightgray}Image Style Classification\newline (Style Cls) & \cellcolor{lightgray}Art & \cellcolor{lightgray}Content Recognition & \cellcolor{lightgray}Stylebooth \cite{StyleBooth-abs-2404-12154} & \cellcolor{lightgray}536 & \cellcolor{lightgray}CaSED \cite{CaSED-conti2023vocabularyfree} & \cellcolor{lightgray}Acc \\

\addlinespace[4pt]
& 30 & Microorganism Image Classification (Microorganism Cls) & Biology & Content Recognition & Manual Construction \cite{Micro-Organism-Image-Classification} & 504 & CaSED \cite{CaSED-conti2023vocabularyfree} & Acc \\

\addlinespace[4pt]
& \cellcolor{lightgray}31 & \cellcolor{lightgray}Multimodal Named Entity Recognition (MNER) & \cellcolor{lightgray}General & \cellcolor{lightgray}Content Recognition & \cellcolor{lightgray}Twitter-17 \cite{Twitter-17-YuJYX20} & \cellcolor{lightgray}500 & \cellcolor{lightgray}UMT \cite{UMT-YuJYX20} & \cellcolor{lightgray}F1 \\

\addlinespace[4pt]
& 32 & Deep Fake Detection\newline (Deep-Fake Det) & Ethics & Content Recognition & DF40 \cite{DF40-yan2024} & 500 & NPR \cite{tan2024rethinking}& Acc \\

\addlinespace[4pt]
\hline

\addlinespace[4pt]
\multirow{1}{*}{\shortstack[l]{\#I-C-29 \\ Panoptic Segmentation \\ (Panoptic Seg)}} 
& 1 & Road Scene Panoptic Segmentation (Road Pano-Seg) & General & Content Recognition, Commonsense Knowledge & Cityscapes \cite{Cityscapes-CordtsORREBFRS16} & 500 & Panoptic-Deeplab \cite{Panoptic-DeepLab-ChengCZ0HAC20} & PQ \\

\addlinespace[4pt]
\hline

\addlinespace[4pt]
\#I-C-30 \newline Pose Recognition \newline (Pose Recog)
& 1 & Yoga Pose Recognition (Yoga Recog) & Sports & Content Recognition & Yoga Pose Classification \cite{Yoga-Pose-Classification} & 988 & CLIP \cite{radford2021learning} & Acc \\

\addlinespace[4pt]
\hline

\addlinespace[4pt]
\multirow{3}{*}{\shortstack[l]{\#I-C-31 \\ Relation Reasoning \\ (Rel Reason)}} 
& 1 & Classification of Visual Spatial Relationship (Spatial-Rel Cls) & General & Content Recognition, Spatial Perception & VSD \cite{VSD-ZhaoWLSZ022} & 500 & VidVRD \cite{yang2024multi}  & F1 \\

\addlinespace[4pt]
& \cellcolor{lightgray}2 & \cellcolor{lightgray}Description of Single Spatial Relationship (Single Spatial-Rel Description) & \cellcolor{lightgray}General & \cellcolor{lightgray}Content Recognition, Spatial Perception & \cellcolor{lightgray}VSD \cite{VSD-ZhaoWLSZ022} & \cellcolor{lightgray}500 & \cellcolor{lightgray}VidVRD \cite{yang2024multi}   & \cellcolor{lightgray}BLEU-4 \& SPICE \\

\addlinespace[4pt]
& 3 & Description of Open-ended Spatial Relationship (Open Spatial-Rel Description) & General & Content Recognition, Spatial Perception & VSD \cite{VSD-ZhaoWLSZ022} & 500 & VidVRD \cite{yang2024multi}   & BLEU-4 \& SPICE \\

\addlinespace[4pt]
\hline

\addlinespace[4pt]
\multirow{1}{*}{\shortstack[l]{\#I-C-32 \\ RGBD Semantic\\ Segmentation \\ (RGBD-Sem Seg)}} 
& 1 & RGBD Semantic Segmentation (RGBD-Sem Seg) \newline \color{white}{1} \newline \color{white}{1} & General & Content Recognition, Commonsense Knowledge & Nyudv2 \cite{Nyudv2-SilbermanHKF12} & 654 & DFormer-L \cite{DFormer-YinZLLCH24} & mIoU \\

\addlinespace[4pt]
\hline

\addlinespace[4pt]
\#I-C-33 \newline Ripeness \newline Recognition \newline (Ripe Recog)
& 1 & Strawberry Ripeness Recognition (Strawberry Recog) & General & Content Recognition & Strawberry Classification \cite{Strawberry-Classification} & 500 & OWLVIT \cite{OWLVIT-abs-2205-06230} & Acc \\

\addlinespace[4pt]
\hline

\addlinespace[4pt]
\multirow{15}{*}{\shortstack[l]{\#I-C-34 \\ Safety Detection \\ (Safe Det)}} 
& 1 & Fire Detection (Fire Det) & General & Content Recognition & Fire Detection \cite{fire-smoke-detection-eozii-nfuzs-vgo4c_dataset} & 500 & GLIP-T \cite{GLIP-li2021grounded} & mAP@0.5 \\

\addlinespace[4pt]
& \cellcolor{lightgray}2 & \cellcolor{lightgray}Gun Detection (Gun Det) & \cellcolor{lightgray}General & \cellcolor{lightgray}Content Recognition & \cellcolor{lightgray}Gun Detection \cite{gun-with-webcam-views_dataset} & \cellcolor{lightgray}500 & \cellcolor{lightgray}YOLO-World \cite{YOLO-World-ChengSG0WS24} & \cellcolor{lightgray}mAP@0.5 \\

\addlinespace[4pt]
& 3 & Construction Safety Equipment Detection (Safe-Equip Det) & General & Content Recognition & Construction Safety Equipment Detection \cite{construction-safety-gsnvb_dataset} & 500 & OWL-VIT \cite{OWL-VIT-abs-2205-06230} & mAP@0.5 \\

\addlinespace[4pt]
& \cellcolor{lightgray}4 & \cellcolor{lightgray}Safety Vests Detection (Safe-Vest Det) & \cellcolor{lightgray}General & \cellcolor{lightgray}Content Recognition & \cellcolor{lightgray}Safety Vests Detection \cite{safety-vests_dataset} & \cellcolor{lightgray}500 & \cellcolor{lightgray}OWL-VIT \cite{OWL-VIT-abs-2205-06230} & \cellcolor{lightgray}mAP@0.5 \\

\addlinespace[4pt]
& 5 & Fall Human Detection (Fall-Human Det) & General & Content Recognition & Fall Human Detection \cite{fall_person_dataset} & 500 & OWL-VIT \cite{OWL-VIT-abs-2205-06230} & mAP@0.5 \\

\addlinespace[4pt]
& \cellcolor{lightgray}6 & \cellcolor{lightgray}Smoke Detection (Smoke Det) & \cellcolor{lightgray}General & \cellcolor{lightgray}Content Recognition & \cellcolor{lightgray}Smoke Detection \cite{smoke-detection1-owusa_dataset} & \cellcolor{lightgray}500 & \cellcolor{lightgray}OWL-VIT \cite{OWL-VIT-abs-2205-06230} & \cellcolor{lightgray}mAP@0.5 \\

\addlinespace[4pt]
& 7 & Helmet Detection (Helmet Det) & General & Content Recognition & Helmet Detection \cite{Bikes-Helmets-Dataset} & 764 & Grounding DINO \cite{liu2023grounding} & AP@0.5 \\

\addlinespace[4pt]
\hline

\addlinespace[4pt]
\#I-C-35 \newline Scene Graph\newline Generation \newline (SG Gen)
& 1 & Image Scene Graph Parsing (SG Parsing) & General & Reasoning Ability & Visual Genome \cite{Visual-Genome-KrishnaZGJHKCKL17} & 767 & Graph-RCNN \cite{Graph-RCNN-YangLWWQHC22} & R@50 \\

\addlinespace[4pt]
\hline

\addlinespace[4pt]
\multirow{1}{*}{\shortstack[l]{\#I-C-36 \\ Scene Recognition \\ (Scene Recog)}} 
& 1 & Terrain Recognition (Terrain Recog)\newline \color{white}{1} \newline \color{white}{1} \newline \color{white}{1} & Nature & Content Recognition & Terrain Recognition \cite{Terrain-Recognition} & 500 & CLIP \cite{radford2021learning} & Acc \\

\addlinespace[4pt]
& \cellcolor{lightgray}2 & \cellcolor{lightgray}Landscape Recognition (Landscape Recog) & \cellcolor{lightgray}Nature & \cellcolor{lightgray}Content Recognition, Commonsense Knowledge & \cellcolor{lightgray}Landscape Recognition \cite{Landscape-Recognition} & \cellcolor{lightgray}500 & \cellcolor{lightgray}CLIP \cite{radford2021learning} & \cellcolor{lightgray}Acc \\

\addlinespace[4pt]
\hline

\addlinespace[4pt]
\#I-C-37 \newline Sign Language \newline Recognition \newline (Sign Recog)
& 1 & American Sign Language Recognition (Sign-Language Recog) & Linguistics & Content Recognition & American Sign Language Recognition \cite{American-Sign-Language-Dataset} & 540 & YOLOV5 \cite{yolov5} & F1 \\

\addlinespace[4pt]
\hline

\addlinespace[4pt]
\multirow{16}{*}{\shortstack[l]{\#I-C-38 \\ Visual Relation\\ Inference \\ (Vis Rel Inf)}} 
& 1 & Before-After Relationship Caption (Before-After-Rel Cap) & Art & Content Recognition & IEdit \cite{iEdit-BodurGB0DB22} & 500 & VPG-C \cite{VPG-C-li2023fine} & ROUGE-L \\

\addlinespace[4pt]
& \cellcolor{lightgray}2 & \cellcolor{lightgray}Surveillance Object-Level Change Description (Object-Change Cap) & \cellcolor{lightgray}General & \cellcolor{lightgray}Content Recognition & \cellcolor{lightgray}Spot-the-Diff \cite{jhamtani2018learning} & \cellcolor{lightgray}500 & \cellcolor{lightgray}MLoEM \cite{MLoEM-WeiSXZLW24} & \cellcolor{lightgray}ROUGE-L \\

\addlinespace[4pt]
& 3 & Multi-Image Alteration Description (Multi-Img-Alter Cap) & General & Content Recognition & CLEVR-Change \cite{CLEVR-JohnsonHMFZG17} & 500 & MLoEM \cite{MLoEM-WeiSXZLW24} & ROUGE-L \\

\addlinespace[4pt]
& \cellcolor{lightgray}4 & \cellcolor{lightgray}Bird Variation Comparison Description (Bird-Compare Cap) & \cellcolor{lightgray}Biology & \cellcolor{lightgray}Content Recognition & \cellcolor{lightgray}Birds-to-Words \cite{Birds-to-Words-ForbesKSB19} & \cellcolor{lightgray}500 & \cellcolor{lightgray}VPG-C \cite{VPG-C-li2023fine} & \cellcolor{lightgray}ROUGE-L \\

\addlinespace[4pt]
& 5 & Subtle Difference Description (Subtle-Diff Cap) & Biology & Content Recognition & Birds-to-Words \cite{Birds-to-Words-ForbesKSB19} & 500 & LLaVA-NeXT-Interleave \cite{li2024llava} & ROUGE-L \\

\addlinespace[4pt]
& \cellcolor{lightgray}6 & \cellcolor{lightgray}Multimodal Relation Extraction (MRE) & \cellcolor{lightgray}General & \cellcolor{lightgray}Content Recognition & \cellcolor{lightgray}MNRE \cite{MNRE-ZhengWFF021} & \cellcolor{lightgray}500 & \cellcolor{lightgray}BertNRE \cite{BertNRE-SoaresFLK19} & \cellcolor{lightgray}F1 \\

\addlinespace[4pt]
\hline

\addlinespace[4pt]
\multirow{15}{*}{\shortstack[l]{\#I-C-39 \\ Visual Storytelling \\ Generation\\ (Vis-Story Gen)}} 
& 1 & Animated Story Completion (Animated Story Completion) & General & Commonsense Understanding & AESOP \cite{Ravi_2018_CVPR} & 500 & LLaVA-NeXT-Interleave \cite{li2024llava} & ROUGE-L \\

\addlinespace[4pt]
& \cellcolor{lightgray}2 & \cellcolor{lightgray}Cartoon Story Telling (Cartoon-Story Gen) & \cellcolor{lightgray}General & \cellcolor{lightgray}Commonsense Understanding & \cellcolor{lightgray}PororoSV \cite{StoryGAN-LiGSLCWCCG19} & \cellcolor{lightgray}500 & \cellcolor{lightgray}LLaVA-NeXT-Interleave \cite{li2024llava} & \cellcolor{lightgray}ROUGE-L \\

\addlinespace[4pt]
& 3 & Multi-image Next-frame Description (Next-frame Cap) & Humanities & Commonsense Understanding & FlintstonesSV \cite{FlintstonesSV-GuptaSFHK18} & 500 & MLoEM \cite{MLoEM-WeiSXZLW24} & ROUGE-L \\

\addlinespace[4pt]
& \cellcolor{lightgray}4 & \cellcolor{lightgray}Sequential Photo Storytelling (Seq-Photo Gen) & \cellcolor{lightgray}Humanities & \cellcolor{lightgray}Commonsense Understanding & \cellcolor{lightgray}VIST \cite{VIST-HuangFMMADGHKBZ16} & \cellcolor{lightgray}500 & \cellcolor{lightgray}VPG-C \cite{VPG-C-li2023fine} & \cellcolor{lightgray}ROUGE-L \\

\addlinespace[4pt]
& 5 & Sequential Story Completion (Seq-Story Completion) & Humanities & Commonsense Understanding & DiDeMoSV \cite{DiDeMoSV-MaharanaHB22} & 500 & VPG-C \cite{VPG-C-li2023fine} & ROUGE-L \\

\addlinespace[4pt]
\hline

\addlinespace[4pt]
\#I-C-40 \newline Weather Recognition \newline (Weather Recog)
& 1 & Weather Recognition (Weather Recog) & Climate & Content Recognition & Weather Image Recognition \cite{DVN/M8JQCR_2021} & 495 & CLIP \cite{radford2021learning} & Acc \\

\addlinespace[4pt]
\hline

\end{longtable}

}

\clearpage

{\centering
\fontsize{8}{8}\selectfont
\setlength{\tabcolsep}{0.7mm}
\renewcommand{\arraystretch}{0.9} 
\captionof{table}{
Detailed list of all tasks and skills (meta-tasks) under image and generation category.
}
\vspace{-3mm}
\label{tab:task-specification-img-gen}
\begin{longtable}[!t]{p{2.2cm}%
                >{\centering\arraybackslash}p{0.2cm}
                p{4cm}%
                >{\centering\arraybackslash}p{1.3cm}%
                p{2.3cm}%
                p{2cm}%
                >{\centering\arraybackslash}p{0.7cm}%
                p{2cm}%
                >{\centering\arraybackslash}p{1.3cm}}

\addlinespace[3pt]
\rowcolor{bg-tb-heavey-vision} \multicolumn{9}{c}{\bf \textcolor{blueGen}{\large Image Generation Group}} \\ [4pt]

\hline

\addlinespace[4pt]
\multicolumn{5}{c}{\bf Task} & \multicolumn{2}{c}{\bf Data}  & \multicolumn{2}{c}{\bf \textbf{SoTA Specialist}} \\

\cmidrule(r){1-5}  \cmidrule(r){6-7}  \cmidrule(r){8-9} 

\textbf{Skill (Meta-Task)} & \textbf{\#}& \textbf{Task (Short Name)} &  \textbf{Domain} & \textbf{Capability} & \textbf{Data Source} & \textbf{Number} & \multicolumn{1}{c}{\textbf{Model}}  & \textbf{Metrics} \\ 

\addlinespace[4pt]
\hline
\endfirsthead
\hline
\addlinespace[4pt]
\multicolumn{5}{c}{\bf Task} & \multicolumn{2}{c}{\bf Data}  & \multicolumn{2}{c}{\bf \textbf{SoTA Specialist}} \\
\cmidrule(r){1-5}  \cmidrule(r){6-7}  \cmidrule(r){8-9} 
\textbf{Skill (Meta-Task)} & \textbf{\#}& \textbf{Task (Short Name)} &  \textbf{Domain} & \textbf{Capability} & \textbf{Data Source} & \textbf{Number} & \multicolumn{1}{c}{\textbf{Model}}  & \textbf{Metrics} \\ 
\addlinespace[4pt]
\hline
\endhead
\hline
\endfoot

\addlinespace[4pt]
\#I-G-1 \newline Edge-to-Image \newline Generation \newline (Edge2Img Gen)
& 1 & Edge-to-Image Generation (Edge2Img Gen) & General & Creativity and Innovation & SketchyCOCO \cite{SketchyCOCO-GaoLXWLZ20} & 1,340 & PITI \cite{PITI-wang2022pretraining} & FID \\

\addlinespace[4pt]
\hline

\addlinespace[4pt]
\#I-G-2 \newline EEG-to-Image \newline Generation \newline (EEG2Img Gen)
& 1 & EEG-based Image Generation (EEG2Img Gen) & Medicine & Creativity and Innovation & DreamDiffusion \cite{DreamDiffusion-BaiWCGYS24} & 500 & DreamDiffusion \cite{DreamDiffusion-BaiWCGYS24} & Acc \\

\addlinespace[4pt]
\hline

\addlinespace[4pt]
\multirow{16}{*}{\shortstack[l]{\#I-G-3 \\ Image Denosing \\ (Img Denose)}} 
& 1 & Image Deraining (Image Deraining) & General & Creativity and Innovation & Rain100H \cite{Rain100H-Yang2017RainRemoval} & 500 & GOUB \cite{GOUB-YuePMDWZ24} & PSNR \\

\addlinespace[4pt]
& \cellcolor{lightgray}2 & \cellcolor{lightgray}Image Raindrop Removal (Raindrop Removal) & \cellcolor{lightgray}General & \cellcolor{lightgray}Creativity and Innovation & \cellcolor{lightgray}Raindrop \cite{Raindrop-Qian-2018} & \cellcolor{lightgray}500 & \cellcolor{lightgray}TransWeather \cite{TransWeather-valanarasu2021} & \cellcolor{lightgray}PSNR \\

\addlinespace[4pt]
& 3 & Image Dehazing (Img Dehazing) & General & Creativity and Innovation & RESIDE-6K \cite{RESIDE-6K-LiRFTFZW19} & 500 & DehazeFormer \cite{DehazeFormer-SongHQD23} & PSNR \\

\addlinespace[4pt]
& \cellcolor{lightgray}4 & \cellcolor{lightgray}Image Desnowing (Img Desnowing) & \cellcolor{lightgray}General & \cellcolor{lightgray}Creativity and Innovation & \cellcolor{lightgray}Snow100K \cite{Snow100K-LiuJHH18} & \cellcolor{lightgray}500 & \cellcolor{lightgray}ConvIR \cite{ConvIR-CuiRCK24a} & \cellcolor{lightgray}PSNR \\

\addlinespace[4pt]
& 5 & Image Deblurring (Img Deblurring) & General & Creativity and Innovation & GoPro \cite{GoPro-Nah-2017R} & 500 & AdaRevD \cite{AdaRevD-xintm2024} & PSNR \\

\addlinespace[4pt]
& \cellcolor{lightgray}6 & \cellcolor{lightgray}JPEG Image Artifact Deduction (Img Artifact Deduction) & \cellcolor{lightgray}General & \cellcolor{lightgray}Creativity and Innovation & \cellcolor{lightgray}DIV2K+Flickr2K \cite{Agustsson_2017_CVPR_Workshops} & \cellcolor{lightgray}29 & \cellcolor{lightgray}DA-CLIP \cite{DA-CLIP-LuoG0SS24} & \cellcolor{lightgray}FID \\

\addlinespace[4pt]
& 7 & Remote Sensing Image Dehazing (Remote-Sense Dehazing) & Geography & Creativity and Innovation & OD-GAN \cite{OD-GAN-HuangLYSS20} & 536 & DehazeFormer \cite{DehazeFormer-SongHQD23} & PSNR \\

\addlinespace[4pt]
\hline

\addlinespace[4pt]
\multirow{12}{*}{\shortstack[l]{\#I-G-4 \\ Image Enhancement \\ (Img Enhance)}} 
& 1 & Image Shadow Removal (Image Shadow Removal) & General & Creativity and Innovation & SRD \cite{SRD-guo2024single} & 408 & DSC \cite{DSC-HuFZQH20} & RMSE \\

\addlinespace[4pt]
& \cellcolor{lightgray}2 & \cellcolor{lightgray}Image Flare Removal (Img-Flare Removal) & \cellcolor{lightgray}General & \cellcolor{lightgray}Creativity and Innovation & \cellcolor{lightgray}Flare7K \cite{Flare7K-DaiLZFL22} & \cellcolor{lightgray}200 & \cellcolor{lightgray}FF-Former \cite{FF-Former-ZhangOLWKJ23} & \cellcolor{lightgray}PSNR \\

\addlinespace[4pt]
& 3 & Underwater Image Restoration (Underwater-Img Restorate) & General & Creativity and Innovation & LSUI \cite{LSUI-10129222} & 500 & CE-VAE \cite{CE-VAE-pucci2024} & PSNR \\

\addlinespace[4pt]
& \cellcolor{lightgray}4 & \cellcolor{lightgray}Document Image Unwarping (Doc-Img Unwarping) & \cellcolor{lightgray}General & \cellcolor{lightgray}Creativity and Innovation & \cellcolor{lightgray}DocUNet \cite{DocUNet-MaSBWS18} & \cellcolor{lightgray}65 & \cellcolor{lightgray}FDRNet \cite{FDRNet-Zhu0KL21} & \cellcolor{lightgray}MS-SSIM \\

\addlinespace[4pt]
& 5 & Image Detail Enhancement (Img-Detail Enhance) & General & Creativity and Innovation & FUnIE-GAN \cite{FUnIE-GAN-IslamXS20} & 500 & FUnIE-GAN \cite{FUnIE-GAN-IslamXS20} & PSNR \\

\addlinespace[4pt]
& \cellcolor{lightgray}6 & \cellcolor{lightgray}Low-light Image Enhancement (Low-light-Img Enhance) & \cellcolor{lightgray}General & \cellcolor{lightgray}Creativity and Innovation & \cellcolor{lightgray}SCI \cite{SCI-ma2022toward} & \cellcolor{lightgray}500 & \cellcolor{lightgray}SCI \cite{SCI-ma2022toward} & \cellcolor{lightgray}PSNR \\

\addlinespace[4pt]
\hline

\addlinespace[4pt]
\#I-G-5 \newline Image Inpainting \newline (Img Inpaint)
& 1 & Face Image Inpainting (Face Inpainting) & General & Creativity and Innovation & CelebaHQ-256 \cite{CelebaHQ-256-KarrasALL18} & 500 & MAT \cite{MAT-li2022} & FID \\

\addlinespace[4pt]
\hline

\addlinespace[4pt]
\multirow{2}{*}{\shortstack[l]{\#I-G-6 \\ Image Style Transfer \\ (Img-Style Trans)}} 
& 1 & Face Image Comic Style Transfer (Face2Comic Gen) & Art & Creativity and Innovation & Manual Construction \cite{StyleBooth-abs-2404-12154} & 500 & FLUX \cite{flux2023} & FID \\

\addlinespace[4pt]
& \cellcolor{lightgray}2 & \cellcolor{lightgray}Text-based Style Transfer (Style Transfer) & \cellcolor{lightgray}Art & \cellcolor{lightgray}Creativity and Innovation & \cellcolor{lightgray}StyleBooth \cite{StyleBooth-abs-2404-12154} & \cellcolor{lightgray}536 & \cellcolor{lightgray}StyleBooth \cite{StyleBooth-abs-2404-12154} & \cellcolor{lightgray}PSNR \\

\addlinespace[4pt]
\hline

\addlinespace[4pt]
\#I-G-7 \newline Imge-to-Mask \newline Generation \newline (Img2Mask Gen)
& 1 & Infection-map Generation (Infection-map Gen) & Medicine & Creativity and Innovation & QaTa-COV19 \cite{QaTa-COV19-DegerliKCG22} & 500 & DDLA \cite{DDLA-DegerliAYKCHHMG21} & AUC \\

\addlinespace[4pt]
\hline

\addlinespace[4pt]
\#I-G-8 \newline Image-to-Sketch \newline Generation \newline (Img2Sketch Gen)
& 1 & Face Sketch Synthesis (Face-Sketch Gen) & General & Creativity and Innovation & FS2K \cite{FS2K-Fan2022} & 500 & HIDA \cite{HIDA-gao2023human} & FID \\

\addlinespace[4pt]
\hline

\addlinespace[4pt]
\multirow{1}{*}{\shortstack[l]{\#I-G-9 \\ In-context Image \\ Editing (Img Edit)}} 
& 1 & Image Editing with Text and Image Prompt (Img-Edit with Txt+Img) & General & Creativity and Innovation & InstructPix2Pix \cite{InstructPix2Pix-BrooksHE23} & 500 & InstructPix2Pix \cite{InstructPix2Pix-BrooksHE23} & CLIP-Score \\

\addlinespace[4pt]
& \cellcolor{lightgray}2 & \cellcolor{lightgray}Image Editing with Multi-image Prompt (Img-Edit with Mul-Img) & \cellcolor{lightgray}General & \cellcolor{lightgray}Creativity and Innovation & \cellcolor{lightgray}StyleBooth \cite{StyleBooth-abs-2404-12154} & \cellcolor{lightgray}500 & \cellcolor{lightgray}StyleBooth \cite{StyleBooth-abs-2404-12154} & \cellcolor{lightgray}CLIP-Score \\

\addlinespace[4pt]
& 3 & Image Editing with Text and Multi-image Prompt (Img-Edit with Txt+Mul-Img) & General & Creativity and Innovation & StyleBooth \cite{StyleBooth-abs-2404-12154} & 500 & StyleBooth \cite{StyleBooth-abs-2404-12154} & CLIP-Score \\

\addlinespace[4pt]
\hline

\addlinespace[4pt]
\#I-G-10 \newline Layout-to-Image \newline Generation \newline (Layout2Img Gen)
& 1 & Layout-to-Face Image Generation (Layout2Face Gen) & Ethics & Creativity and Innovation & Manual Construction \cite{Pretty-Face} & 500 & LayoutDiffusion \cite{LayoutDiffusion-ZhengZLQSL23} & FID \\

\addlinespace[4pt]
\hline

\addlinespace[4pt]
\#I-G-11 \newline Mask-to-Image \newline Generation \newline (Mask2Img Gen)
& 1 & Mask-to-Image Face Generation (Mask2Face Generation) & General & Creativity and Innovation & UniControl \cite{UniControl-QinZYFYZWNXSE0X23} & 500 & UniControl \cite{UniControl-QinZYFYZWNXSE0X23} & PSNR \\

\addlinespace[4pt]
\hline

\addlinespace[4pt]
\multirow{9}{*}{\shortstack[l]{\#I-G-12 \\ Sketch-to-Image \\ Generation \\ (Sketch2Img Gen)}} 
& 1 & Sketch-to-Face Image Generation (Sketch2Face Gen) & General & Creativity and Innovation & Manual Construction \cite{Pretty-Face} & 500 & PITI \cite{PITI-wang2022pretraining} & FID \\

\addlinespace[4pt]
& \cellcolor{lightgray}2 & \cellcolor{lightgray}Sketch-to-Background Image Generation (Sketch2BG Gen) & \cellcolor{lightgray}General & \cellcolor{lightgray}Creativity and Innovation, Planning Ability & \cellcolor{lightgray}SketchyCOCO \cite{SketchyCOCO-GaoLXWLZ20} & \cellcolor{lightgray}1,033 & \cellcolor{lightgray}PITI \cite{PITI-wang2022pretraining} & \cellcolor{lightgray}FID \\

\addlinespace[4pt]
& 3 & Sketch-to-Scene Image Generation (Sketch2Scene Gen) & General & Creativity and Innovation & SketchyCOCO \cite{SketchyCOCO-GaoLXWLZ20} & 542 & PITI \cite{PITI-wang2022pretraining} & FID \\

\addlinespace[4pt]
& \cellcolor{lightgray}4 & \cellcolor{lightgray}Sketch-to-Image Hair Generation (Sketch2Hair Generation) & \cellcolor{lightgray}General & \cellcolor{lightgray}Creativity and Innovation & \cellcolor{lightgray}HairCLIP \cite{HairCLIP-Wei0Z0TY0Y22} & \cellcolor{lightgray}500 & \cellcolor{lightgray}HairCLIP \cite{HairCLIP-Wei0Z0TY0Y22} & \cellcolor{lightgray}PSNR \\

\addlinespace[4pt]
\hline

\addlinespace[4pt]
\#I-G-13 \newline Sound-to-Image \newline Generation \newline (Sound2Img Gen)
& 1 & Sound-to-Image Generation (Sound2Img Gen) & Physics & Creativity and Innovation & GlueGen \cite{GlueGen-QinYXZCE0XX23} & 500 & GlueGen \cite{GlueGen-QinYXZCE0XX23} & CLIP-Score \\

\addlinespace[4pt]
\hline

\addlinespace[4pt]
\multirow{8}{*}{\shortstack[l]{\#I-G-14 \\ Text-based Image\\ Editing (Text-based \\ Img Edit)}} 
& 1 & Image Colorization (Img Colorization) & General & Creativity and Innovation & Manual Construction \cite{Image-Colorization} & 500 & DDNM \cite{DDNM-wang2022zero} & FID \\

\addlinespace[4pt]
& \cellcolor{lightgray}2 & \cellcolor{lightgray}Chinese-based Image Editing (Zh-Img Editing) & \cellcolor{lightgray}Linguistics & \cellcolor{lightgray}Creativity and Innovation & \cellcolor{lightgray}SEED-Data-Edit \cite{SEED-Data-Edit-ge2024seed} & \cellcolor{lightgray}500 & \cellcolor{lightgray}SEED-X \cite{SEED-X-ge2024seed} & \cellcolor{lightgray}CLIP-Score \\

\addlinespace[4pt]
& 3 & Text-based Image Editing (Txt-based Img Editing) & General & Creativity and Innovation & StyleBooth \cite{StyleBooth-abs-2404-12154} & 500 & StyleBooth \cite{StyleBooth-abs-2404-12154} & CLIP-Score \\

\addlinespace[4pt]
& \cellcolor{lightgray}4 & \cellcolor{lightgray}Deep Fake Generation (Dee-Fake Gen) & \cellcolor{lightgray}Ethics & \cellcolor{lightgray}Creativity and Innovation & \cellcolor{lightgray}DF40 \cite{DF40-yan2024} & \cellcolor{lightgray}500 & \cellcolor{lightgray}Collaborative Diffusion \cite{huang2023collaborative} & \cellcolor{lightgray}FID \\

\addlinespace[4pt]
\hline

\addlinespace[4pt]
\multirow{20}{*}{\shortstack[l]{\#I-G-15 \\ Text-to-Image\\ Generation \\ (Txt2Img Gen)}} 
& 1 & Text-based E-commerce Product Image Generation (Text2Product Gen) & General & Creativity and Innovation & Manual  & 2,907 & FLUX \cite{flux2023} & FID \\

\addlinespace[4pt]
& \cellcolor{lightgray}2 & \cellcolor{lightgray}Text-based Terrestrial Animal Image Generation (Txt2TerAnimal Gen) & \cellcolor{lightgray}Animal & \cellcolor{lightgray}Creativity and Innovation & \cellcolor{lightgray}Manual  & \cellcolor{lightgray}1,944 & \cellcolor{lightgray}FLUX \cite{flux2023} & \cellcolor{lightgray}PSNR \\

\addlinespace[4pt]
& 3 & Long-text-based Image generation (LongTxt2Img Gen) & General & Creativity and Innovation & ParaDiffusion \cite{ParaDiffusion-abs-2311-14284} & 500 & ParaDiffusion \cite{ParaDiffusion-abs-2311-14284} & FID \\

\addlinespace[4pt]
& \cellcolor{lightgray}4 & \cellcolor{lightgray}Multi-language-based Image Generation (MulLan2Img Gen) & \cellcolor{lightgray}General & \cellcolor{lightgray}Creativity and Innovation & \cellcolor{lightgray}GlueGen \cite{GlueGen-QinYXZCE0XX23} & \cellcolor{lightgray}500 & \cellcolor{lightgray}GlueGen \cite{GlueGen-QinYXZCE0XX23} & \cellcolor{lightgray}FID \\

\addlinespace[4pt]
& 5 & Handwriting Text Generation (Handwrite-Txt Gen) & Humanity & Creativity and Innovation & Manual  & 500 & FLUX \cite{flux2023} & FID \\

\addlinespace[4pt]
& \cellcolor{lightgray}6 & \cellcolor{lightgray}Text-based Face Generation (Txt2Face Gen) & \cellcolor{lightgray}General & \cellcolor{lightgray}Creativity and Innovation & \cellcolor{lightgray}Manual  & \cellcolor{lightgray}500 & \cellcolor{lightgray}FLUX \cite{flux2023} & \cellcolor{lightgray}FID \\

\addlinespace[4pt]
& 7 & Text-based Astronomical Image Generation (Txt2Astronomy Gen) & Astronomy & Creativity and Innovation & Manual & 500 & FLUX \cite{flux2023} & PSNR \\

\addlinespace[4pt]
& \cellcolor{lightgray}8 & \cellcolor{lightgray}Text-to-Image Indian Food Generation (Txt2Food Gen) & \cellcolor{lightgray}Humanities & \cellcolor{lightgray}Creativity and Innovation & \cellcolor{lightgray}Manual  & \cellcolor{lightgray}560 & \cellcolor{lightgray}FLUX \cite{flux2023} & \cellcolor{lightgray}PSNR \\

\addlinespace[4pt]
& 9 & Text-to-Image Microorganism Generation (Txt2Microorgan Gen) & Biology & Creativity and Innovation & Manual  & 504 & FLUX \cite{flux2023} & PSNR \\

\addlinespace[4pt]
& \cellcolor{lightgray}10 & \cellcolor{lightgray}Text-to-Image Insect Generation (Txt2Insect Gen) & \cellcolor{lightgray}Biology & \cellcolor{lightgray}Creativity and Innovation & \cellcolor{lightgray}Manual & \cellcolor{lightgray}500 & \cellcolor{lightgray}FLUX \cite{flux2023} & \cellcolor{lightgray}FID \\

\addlinespace[4pt]
& 11 & Text-to-Image Sea Animal Generation (Txt2SeaAnimal Gen) & Biology, Animal & Creativity and Innovation & Manual  & 506 & FLUX \cite{flux2023} & PSNR \\

\addlinespace[4pt]
\hline

\end{longtable}

}


{\centering
\fontsize{8}{8}\selectfont
\setlength{\tabcolsep}{0.7mm}
\renewcommand{\arraystretch}{0.9} 
\captionof{table}{
Detailed list of all tasks and skills (meta-tasks) under video and comprehension category.
}
\vspace{-3mm}
\label{tab:task-specification-vid-comp}
\begin{longtable}[!t]{p{2.2cm}%
                >{\centering\arraybackslash}p{0.2cm}
                p{4cm}%
                >{\centering\arraybackslash}p{1.3cm}%
                p{2.3cm}%
                p{2cm}%
                >{\centering\arraybackslash}p{0.7cm}%
                p{2cm}%
                >{\centering\arraybackslash}p{1.3cm}}

\addlinespace[3pt]
\rowcolor{bg-tb-heavey-video} \multicolumn{9}{c}{\bf \textcolor{greenCom}{\large Video Comprehension Group}} \\ [4pt]

\hline

\addlinespace[4pt]
\multicolumn{5}{c}{\bf Task} & \multicolumn{2}{c}{\bf Data}  & \multicolumn{2}{c}{\bf \textbf{SoTA Specialist}} \\

\cmidrule(r){1-5}  \cmidrule(r){6-7}  \cmidrule(r){8-9} 

\textbf{Skill (Meta-Task)} & \textbf{\#}& \textbf{Task (Short Name)} &  \textbf{Domain} & \textbf{Capability} & \textbf{Data Source} & \textbf{Number} & \multicolumn{1}{c}{\textbf{Model}} & \textbf{Metrics} \\ 

\addlinespace[4pt]
\hline
\endfirsthead
\hline
\addlinespace[4pt]
\multicolumn{5}{c}{\bf Task} & \multicolumn{2}{c}{\bf Data}  & \multicolumn{2}{c}{\bf \textbf{SoTA Specialist}} \\
\cmidrule(r){1-5}  \cmidrule(r){6-7}  \cmidrule(r){8-9} 
\textbf{Skill (Meta-Task)} & \textbf{\#}& \textbf{Task (Short Name)} &  \textbf{Domain} & \textbf{Capability} & \textbf{Data Source} & \textbf{Number} & \multicolumn{1}{c}{\textbf{Model}}  & \textbf{Metrics} \\ 
\addlinespace[4pt]
\hline
\endhead
\hline
\endfoot

\addlinespace[4pt]
\multirow{32}{*}{\shortstack[l]{\#V-C-1 \\ Video Question\\ Answering \\ (Video QA)}} 
& 1 & Agriculture Video Question Answering (Agri-Vid QA) & Culture & Content Recognition, Affective Analysis & WildQA \cite{WildQA-2022-in-the-wild} & 75 & Qwen-2-VL-72B \cite{wang2024qwen2} & BLEU-1 \\

\addlinespace[4pt]
& \cellcolor{lightgray}2 & \cellcolor{lightgray}Geography Video Question Answering (Geo-Vid QA) & \cellcolor{lightgray}General & \cellcolor{lightgray}Content Recognition, Affective Analysis & \cellcolor{lightgray}WildQA \cite{WildQA-2022-in-the-wild} & \cellcolor{lightgray}81 & \cellcolor{lightgray}Qwen-2-VL-72B \cite{wang2024qwen2} & \cellcolor{lightgray}BLEU-1 \\

\addlinespace[4pt]
& 3 & Human Survival Video Question Answering (Survival-Vid QA) & Humanity & Content Recognition, Affective Analysis & WildQA \cite{WildQA-2022-in-the-wild} & 218 & LLaVA-Video-72B-Qwen2 \cite{LLaVA-Video-72B-Qwen2} & BLEU-1 \\

\addlinespace[4pt]
& \cellcolor{lightgray}4 & \cellcolor{lightgray}Military Video Question Answering (Military-Vid QA) & \cellcolor{lightgray}General & \cellcolor{lightgray}Content Recognition & \cellcolor{lightgray}WildQA \cite{WildQA-2022-in-the-wild} & \cellcolor{lightgray}145 & \cellcolor{lightgray}LLaVA-Video-72B-Qwen2 \cite{LLaVA-Video-72B-Qwen2} & \cellcolor{lightgray}BLEU \\

\addlinespace[4pt]
& 5 & Comedy Video Question Answering (Comedy-Vid QA) & General & Content Recognition, Commonsense Knowledge & VCGBench \cite{Maaz2024VideoGPT} & 158 & Aria \cite{aria-dongxu} & BLEU-1 \\

\addlinespace[4pt]
& \cellcolor{lightgray}6 & \cellcolor{lightgray}Movie Video Question Answering (Movie-Vid QA) & \cellcolor{lightgray}General & \cellcolor{lightgray}Content Recognition & \cellcolor{lightgray}VCGBench \cite{Maaz2024VideoGPT} & \cellcolor{lightgray}156 & \cellcolor{lightgray}Aria \cite{aria-dongxu} & \cellcolor{lightgray}Acc \\

\addlinespace[4pt]
& 7 & Science Video Question Answering (Science-Vid QA) & Humanity, Biology, Geometry & Content Recognition, Commonsense Knowledge & VCGBench \cite{Maaz2024VideoGPT} & 341 & Aria \cite{aria-dongxu} & Acc \\

\addlinespace[4pt]
& \cellcolor{lightgray}8 & \cellcolor{lightgray}Sports Video Question Answering (Sport-Vid QA) & \cellcolor{lightgray}Humanity & \cellcolor{lightgray}Content Recognition & \cellcolor{lightgray}VCGBench \cite{Maaz2024VideoGPT} & \cellcolor{lightgray}400 & \cellcolor{lightgray}Aria \cite{aria-dongxu} & \cellcolor{lightgray}Acc \\

\addlinespace[4pt]
& 9 & Pets Video Question Answering (Pet-Vid QA) & General & Content Recognition & VCGBench \cite{Maaz2024VideoGPT} & 72 & Aria \cite{aria-dongxu} & Acc \\

\addlinespace[4pt]
& \cellcolor{lightgray}10 & \cellcolor{lightgray}Gymnastics Video Question Answering (Gymnastic-Vid QA) & \cellcolor{lightgray}Humanity, Sports & \cellcolor{lightgray}Content Recognition & \cellcolor{lightgray}SportsQA \cite{Sports-QA-abs-2401-01505} & \cellcolor{lightgray}202 & \cellcolor{lightgray}Aria \cite{aria-dongxu} & \cellcolor{lightgray}BLEU-1 \\

\addlinespace[4pt]
& 11 & Ball Game Video Question Answering (Ball-Vid QA) & Humanity & Content Recognition, Commonsense Knowledge & SportsQA \cite{Sports-QA-abs-2401-01505} & 350 & Aria \cite{aria-dongxu} & BLEU-1 \\

\addlinespace[4pt]
& \cellcolor{lightgray}12 & \cellcolor{lightgray}Object Color Video Question Answering (Obj-Color-Vid QA) & \cellcolor{lightgray}General & \cellcolor{lightgray}Content Recognition & \cellcolor{lightgray}ActivityNetQA \cite{ActivityNet-QA-YuXYYZZT19} & \cellcolor{lightgray}800 & \cellcolor{lightgray}Qwen-2-VL-7B \cite{Qwen2-VL-abs-2409-12191} & \cellcolor{lightgray}Acc \\

\addlinespace[4pt]
& 13 & Object Motion Video Question Answering (Obj-Motion-Vid QA) & General & Content Recognition & ActivityNetQA \cite{ActivityNet-QA-YuXYYZZT19} & 800 & Qwen-2-VL-7B \cite{Qwen2-VL-abs-2409-12191} & Acc \\

\addlinespace[4pt]
& \cellcolor{lightgray}14 & \cellcolor{lightgray}Object Location Video Question Answering (Obj-Loc-Vid QA) & \cellcolor{lightgray}General & \cellcolor{lightgray}Content Recognition & \cellcolor{lightgray}ActivityNetQA \cite{ActivityNet-QA-YuXYYZZT19} & \cellcolor{lightgray}800 & \cellcolor{lightgray}Qwen-2-VL-7B \cite{Qwen2-VL-abs-2409-12191} & \cellcolor{lightgray}Acc \\

\addlinespace[4pt]
& 15 & Object Direction Video Question Answering (Obj-Dir-Vid QA) & General & Content Recognition & MVBench \cite{MVBench-0002WH00LWX0L0024} & 200 & Qwen-2-VL-7B \cite{Qwen2-VL-abs-2409-12191} & BLEU-1 \\

\addlinespace[4pt]
& \cellcolor{lightgray}16 & \cellcolor{lightgray}Education Video Question Answering (Edu-Vid QA) & \cellcolor{lightgray}Knowledge & \cellcolor{lightgray}Content Recognition & \cellcolor{lightgray}VideoMME \cite{Video-MME-abs-2405-21075} & \cellcolor{lightgray}300 & \cellcolor{lightgray}Oryx-34B \cite{Oryx-abs-2409-12961} & \cellcolor{lightgray}Acc \\

\addlinespace[4pt]
& 17 & Human-Object Interaction Video Question Answering (Interact-Vid QA) & Humanity & Content Recognition & MMBench-Video \cite{MMBench-Video-abs-2406-14515} & 111 & Aria \cite{aria-dongxu} & Acc \\

\addlinespace[4pt]
\hline

\addlinespace[4pt]
\multirow{4}{*}{\shortstack[l]{\#V-C-2 \\ Video Object\\ Recognition \\ (Vid-Obj Recog)}} 
& 1 & Pets Video Recognition (Pets Video Recog) & General & Content Recognition & MMBench-Video \cite{MMBench-Video-abs-2406-14515} & 87 & Aria \cite{aria-dongxu} & Acc \\

\addlinespace[4pt]
& \cellcolor{lightgray}2 & \cellcolor{lightgray}Science Video Recognition (Sci-Vid Recog) & \cellcolor{lightgray}Knowledge & \cellcolor{lightgray}Content Recognition & \cellcolor{lightgray}MMBench-Video \cite{MMBench-Video-abs-2406-14515} & \cellcolor{lightgray}100 & \cellcolor{lightgray}Aria \cite{aria-dongxu} & \cellcolor{lightgray}Acc \\

\addlinespace[4pt]
& 3 & Video Object Counting (Vid-Obj Counting) & General & Content Recognition, Causality Discrimination & MVBench \cite{MVBench-0002WH00LWX0L0024} & 200 & Qwen-2-VL-7B \cite{Qwen2-VL-abs-2409-12191} & BLEU-1 \\

\addlinespace[4pt]
& \cellcolor{lightgray}4 & \cellcolor{lightgray}Natural Disaster Video Recognition (Disaster-Vid Recog) & \cellcolor{lightgray}General & \cellcolor{lightgray}Content Recognition & \cellcolor{lightgray}WildQA \cite{WildQA-2022-in-the-wild} & \cellcolor{lightgray}133 & \cellcolor{lightgray}Qwen-2-VL-72B \cite{} & \cellcolor{lightgray}BLEU \\

\addlinespace[4pt]
& 5 & Video Object Existence Recognition (Vid-Obj-Exist Recog) & General & Content Recognition & MVBench \cite{MVBench-0002WH00LWX0L0024} & 200 & Qwen-2-VL-7B \cite{Qwen2-VL-abs-2409-12191} & BLEU-1 \\

\addlinespace[4pt]
& \cellcolor{lightgray}6 & \cellcolor{lightgray}Video Object Interaction Recognition (Vid-Obj-Interact Recog) & \cellcolor{lightgray}General & \cellcolor{lightgray}Content Recognition & \cellcolor{lightgray}MVBench \cite{MVBench-0002WH00LWX0L0024} & \cellcolor{lightgray}200 & \cellcolor{lightgray}Qwen-2-VL-7B \cite{Qwen2-VL-abs-2409-12191} & \cellcolor{lightgray}BLEU-1 \\

\addlinespace[4pt]
& 7 & TV-Show Recognition (TV-Show Recog) & Humanity, Art & Content Recognition & VideoMME \cite{Video-MME-abs-2405-21075} & 100 & Oryx-34B \cite{Oryx-abs-2409-12961} & Acc \\

\addlinespace[4pt]
& \cellcolor{lightgray}8 & \cellcolor{lightgray}Video Sports Recognition (Sport Recog) & \cellcolor{lightgray}Humanity & \cellcolor{lightgray}Content Recognition & \cellcolor{lightgray}VideoMME \cite{Video-MME-abs-2405-21075} & \cellcolor{lightgray}100 & \cellcolor{lightgray}Oryx-34B \cite{Oryx-abs-2409-12961} & \cellcolor{lightgray}Acc \\

\addlinespace[4pt]
& 9 & Video Animal Recognition (Animal Recog) & General & Content Recognition & VideoMME \cite{Video-MME-abs-2405-21075} & 90 & Oryx-34B \cite{Oryx-abs-2409-12961} & Acc \\

\addlinespace[4pt]
& \cellcolor{lightgray}10 & \cellcolor{lightgray}Video Food Recognition (Food Recog) & \cellcolor{lightgray}General & \cellcolor{lightgray}Content Recognition & \cellcolor{lightgray}VideoMME \cite{Video-MME-abs-2405-21075} & \cellcolor{lightgray}90 & \cellcolor{lightgray}Oryx-34B \cite{Oryx-abs-2409-12961} & \cellcolor{lightgray}Acc \\

\addlinespace[4pt]
& 11 & Art Recognition (Art Recog) & Art, Culture & Content Recognition, Commonsense Knowledge & VideoMME \cite{Video-MME-abs-2405-21075} & 90 & Oryx-34B \cite{Oryx-abs-2409-12961} & Acc \\

\addlinespace[4pt]
& \cellcolor{lightgray}12 & \cellcolor{lightgray}Autos and Vehicles Video Captioning (Vehicles Cap) & \cellcolor{lightgray}General & \cellcolor{lightgray}Content Recognition, Commonsense Knowledge & \cellcolor{lightgray}MMBench-Video \cite{MMBench-Video-abs-2406-14515} & \cellcolor{lightgray}33 & \cellcolor{lightgray}LLaVA-Video-72B-Qwen2 \cite{LLaVA-Video-72B-Qwen2} & \cellcolor{lightgray}BLEU-1 \\

\addlinespace[4pt]
& 13 & Food Video Captioning (Food Cap) & General & Content Recognition, Commonsense Knowledge & MMBench-Video \cite{MMBench-Video-abs-2406-14515} & 41 & LLaVA-Video-72B-Qwen2 \cite{LLaVA-Video-72B-Qwen2} & BLEU-1 \\

\addlinespace[4pt]
& \cellcolor{lightgray}14 & \cellcolor{lightgray}Video Salient Object Detection (Salient-Obj Det) & \cellcolor{lightgray}General & \cellcolor{lightgray}Content Recognition & \cellcolor{lightgray}SegTrack, FBMS, vidsod\_100 \cite{LiKHTR13,OchsMB14,ViDSOD-100-LinZSFZW24} & \cellcolor{lightgray}69 & \cellcolor{lightgray}RealFlow \cite{cho2024transforming} & \cellcolor{lightgray}S-measure \\

\addlinespace[4pt]
& 15 & Human Video Salient Detection ( Hum-Salient Obj Det) & Humanities & Content Recognition & SegTrack, FBMS, vidsod\_101 \cite{LiKHTR13,OchsMB14,ViDSOD-100-LinZSFZW24} & 14 & RealFlow \cite{cho2024transforming} & S-measure \\

\addlinespace[4pt]
& \cellcolor{lightgray}16 & \cellcolor{lightgray}Aquatic Video Camouflaged Object Detection (Aquatic COD) & \cellcolor{lightgray}Animal & \cellcolor{lightgray}Content Recognition & \cellcolor{lightgray}MoCA \cite{LamdouarYXZ20} & \cellcolor{lightgray}49 & \cellcolor{lightgray}SAM-PM \cite{SAM-PM-MeeranTM22} & \cellcolor{lightgray}S-measure \\

\addlinespace[4pt]
& 17 & Terrestrial Video Camouflaged Object Detection (Terrestrial COD) & Animal & Content Recognition & MoCA \cite{LamdouarYXZ20} & 92 & SAM-PM \cite{SAM-PM-MeeranTM22} & S-measure \\

\addlinespace[4pt]
\hline

\addlinespace[4pt]
\multirow{15}{*}{\shortstack[l]{\#V-C-3 \\ Video Action\\ Recognition \\ (Vid Act Recog)}}
& 1 & Human-Object Interaction Video Captioning (H-O-Interact-Vid Cap) & Humanity & Content Recognition, Commonsense Knowledge & ucf101, hmdb51 \cite{UCF101-abs-1212-0402,HMDB-KuehneJGPS11} & 1,000 & LLaVA-Video-72B-Qwen2 \cite{LLaVA-Video-72B-Qwen2} & BLEU-1 \\

\addlinespace[4pt]
& \cellcolor{lightgray}2 & \cellcolor{lightgray}Human-Human Interaction Video Captioning (H-H-Interact-Vid Cap) & \cellcolor{lightgray}Humanity & \cellcolor{lightgray}Content Recognition, Commonsense Knowledge & \cellcolor{lightgray}ucf101, hmdb52 \cite{UCF101-abs-1212-0402,HMDB-KuehneJGPS11} & \cellcolor{lightgray}500 & \cellcolor{lightgray}LLaVA-Video-72B-Qwen2 \cite{LLaVA-Video-72B-Qwen2} & \cellcolor{lightgray}BLEU-1 \\

\addlinespace[4pt]
& 3 & BodyMotion Video Captioning (Body-Motion Cap) & Humanity & Content Recognition, Commonsense Knowledge & ucf101, hmdb53 \cite{UCF101-abs-1212-0402,HMDB-KuehneJGPS11} & 1,000 & LLaVA-Video-72B-Qwen2 \cite{LLaVA-Video-72B-Qwen2} & BLEU-1 \\

\addlinespace[4pt]
& \cellcolor{lightgray}4 & \cellcolor{lightgray}Musical Instruments Video Captioning (Instrument Cap) & \cellcolor{lightgray}Art & \cellcolor{lightgray}Content Recognition, Commonsense Knowledge & \cellcolor{lightgray}ucf101 \cite{UCF101-abs-1212-0402} & \cellcolor{lightgray}401 & \cellcolor{lightgray}LLaVA-Video-72B-Qwen2 \cite{LLaVA-Video-72B-Qwen2} & \cellcolor{lightgray}BLEU-1 \\

\addlinespace[4pt]
& 5 & Sports and Exercise Video Captioning (Sports Cap) & Humanity & Content Recognition, Commonsense Knowledge & ucf101, k600 \cite{UCF101-abs-1212-0402,Kinetics-600-abs-1808-01340} & 3,000 & LLaVA-Video-72B-Qwen2 \cite{LLaVA-Video-72B-Qwen2} & BLEU-1 \\

\addlinespace[4pt]
& \cellcolor{lightgray}6 & \cellcolor{lightgray}Facial Action Video Captioning (Face-Action Cap) & \cellcolor{lightgray}Humanity & \cellcolor{lightgray}Content Recognition, Commonsense Knowledge & \cellcolor{lightgray}hmdb51 \cite{HMDB-KuehneJGPS11} & \cellcolor{lightgray}300 & \cellcolor{lightgray}LLaVA-Video-72B-Qwen2 \cite{LLaVA-Video-72B-Qwen2} & \cellcolor{lightgray}BLEU-1 \\

\addlinespace[4pt]
& 7 & Facial-Object Operations Video Captioning (Face-Operation Cap) & Humanity & Content Recognition, Commonsense Knowledge & hmdb51 \cite{HMDB-KuehneJGPS11} & 257 & LLaVA-Video-72B-Qwen2 \cite{LLaVA-Video-72B-Qwen2} & BLEU-1 \\

\addlinespace[4pt]
& \cellcolor{lightgray}8 & \cellcolor{lightgray}Arts and Crafts Video Captioning (Arts\&Crafts Cap) & \cellcolor{lightgray}Art & \cellcolor{lightgray}Content Recognition, Commonsense Knowledge & \cellcolor{lightgray}k600 \cite{Kinetics-600-abs-1808-01340} & \cellcolor{lightgray}2,000 & \cellcolor{lightgray}VideoChat2-7B \cite{li2023videochat} & \cellcolor{lightgray}BLEU-1 \\

\addlinespace[4pt]
& 9 & Personal Care Video Captioning (Person-Care Cap) & Humanity & Content Recognition, Commonsense Knowledge & k600 \cite{Kinetics-600-abs-1808-01340} & 2,000 & VideoChat2-7B \cite{li2023videochat} & BLEU-1 \\

\addlinespace[4pt]
& \cellcolor{lightgray}10 & \cellcolor{lightgray}DailyLife and Skills Video Captioning (Skill Cap) & \cellcolor{lightgray}Humanity & \cellcolor{lightgray}Content Recognition, Commonsense Knowledge & \cellcolor{lightgray}k600 \cite{Kinetics-600-abs-1808-01340} & \cellcolor{lightgray}2,000 & \cellcolor{lightgray}VideoChat2-7B \cite{li2023videochat} & \cellcolor{lightgray}BLEU-1 \\

\addlinespace[4pt]
& 11 & Entertainment-related Video Captioning (Entertain Cap) & Humanity & Content Recognition, Commonsense Knowledge & k600 \cite{Kinetics-600-abs-1808-01340} & 2,000 & VideoChat2-7B \cite{li2023videochat} & BLEU-1 \\

\addlinespace[4pt]
& \cellcolor{lightgray}12 & \cellcolor{lightgray}Sign Language Video Recognition (Sign-Language Recog) & \cellcolor{lightgray}Linguistics & \cellcolor{lightgray}Content Recognition & \cellcolor{lightgray}WLASL \cite{li2020word} & \cellcolor{lightgray}1,404 & \cellcolor{lightgray}NLA-SLR \cite{NLA-SLR-ZuoWM23} & \cellcolor{lightgray}Acc \\

\addlinespace[4pt]
& 13 & Video Action Sequence Understanding (Action-Seq Analy) & General & Content Recognition, Causality Discrimination & MVBench \cite{MVBench-0002WH00LWX0L0024} & 188 & Video\_XL \cite{Video-XL-abs-2409-14485} & BLEU-1 \\

\addlinespace[4pt]
& \cellcolor{lightgray}14 & \cellcolor{lightgray}Video Action Sequence Prediction (Action-Seq Pred) & \cellcolor{lightgray}General & \cellcolor{lightgray}Content Recognition, Causality Discrimination & \cellcolor{lightgray}MVBench \cite{MVBench-0002WH00LWX0L0024} & \cellcolor{lightgray}200 & \cellcolor{lightgray}Video\_XL \cite{Video-XL-abs-2409-14485} & \cellcolor{lightgray}BLEU-1 \\

\addlinespace[4pt]
& 15 & Video Action Counting (Action Counting) & General & Content Recognition, Causality Discrimination & MVLU \cite{MLVU-zhou} & 189 & Video\_XL \cite{Video-XL-abs-2409-14485} & Acc \\

\addlinespace[4pt]
& \cellcolor{lightgray}16 & \cellcolor{lightgray}Video Action Ordering (Action Ordering) & \cellcolor{lightgray}General & \cellcolor{lightgray}Content Recognition & \cellcolor{lightgray}MVLU \cite{MLVU-zhou} & \cellcolor{lightgray}125 & \cellcolor{lightgray}Video\_XL \cite{Video-XL-abs-2409-14485} & \cellcolor{lightgray}Acc \\

\addlinespace[4pt]
\hline

\addlinespace[4pt]
\multirow{30}{*}{\shortstack[l]{
\#V-C-4 \\ Video\\ Understanding\\ (Vid Understand)
}}
& 1 & Ball Sports Video Captioning (Sport Cap) & Humanity & Content Recognition, Commonsense Knowledge & VideoMME \cite{Video-MME-abs-2405-21075} & 60 & LLaVA-Video-72B-Qwen2 \cite{LLaVA-Video-72B-Qwen2} & BLEU-1 \\


\addlinespace[4pt]

& \cellcolor{lightgray}2 & \cellcolor{lightgray}Science and Technology Video Captioning (Sci\&Tech Cap) & \cellcolor{lightgray}Science, Engineering & \cellcolor{lightgray}Content Recognition, Commonsense Knowledge & \cellcolor{lightgray}VideoMME \cite{Video-MME-abs-2405-21075} & \cellcolor{lightgray}60 & \cellcolor{lightgray}LLaVA-Video-72B-Qwen2 \cite{LLaVA-Video-72B-Qwen2} & \cellcolor{lightgray}BLEU-1 \\


\addlinespace[4pt]
& 3 & Music Video Question Answering (Music QA) & Humanity, Art & Content Recognition, Reasoning Ability & VCGBench \cite{Maaz2024VideoGPT} & 50 & Aria \cite{aria-dongxu} & Acc \\

\addlinespace[4pt]
& \cellcolor{lightgray}4 & \cellcolor{lightgray}Game Video Question Answering (Game QA) & \cellcolor{lightgray}Art, Culture & \cellcolor{lightgray}Content Recognition, Reasoning Ability & \cellcolor{lightgray}VCGBench \cite{Maaz2024VideoGPT} & \cellcolor{lightgray}224 & \cellcolor{lightgray}Aria \cite{aria-dongxu} & \cellcolor{lightgray}Acc \\

\addlinespace[4pt]
& 5 & Movie and Show Video Captioning (Movie Cap) & Art & Content Recognition, Commonsense Knowledge, Reasoning Ability & VideoMME \cite{Video-MME-abs-2405-21075} & 60 & LLaVA-Video-72B-Qwen2 \cite{LLaVA-Video-72B-Qwen2} & BLEU-1 \\


\addlinespace[4pt]
& \cellcolor{lightgray}6 & \cellcolor{lightgray}Finance Video Captioning (Finance Cap) & \cellcolor{lightgray}Finance & \cellcolor{lightgray}Content Recognition, Commonsense Knowledge & \cellcolor{lightgray}VideoMME \cite{Video-MME-abs-2405-21075} & \cellcolor{lightgray}60 & \cellcolor{lightgray}LLaVA-Video-72B-Qwen2 \cite{LLaVA-Video-72B-Qwen2} & \cellcolor{lightgray}BLEU-1 \\


\addlinespace[4pt]
& 7 & History and Literature Video Captioning (His\&Lit Cap) & History, Humanities, Culture & Content Recognition, Commonsense Knowledge & VideoMME \cite{Video-MME-abs-2405-21075} & 30 & LLaVA-Video-72B-Qwen2 \cite{LLaVA-Video-72B-Qwen2} & BLEU-1 \\


\addlinespace[4pt]
& \cellcolor{lightgray}8 & \cellcolor{lightgray}Business Video Captioning (Business Cap) & \cellcolor{lightgray}Business & \cellcolor{lightgray}Content Recognition, Commonsense Knowledge & \cellcolor{lightgray}MMBench-Video \cite{MMBench-Video-abs-2406-14515} & \cellcolor{lightgray}56 & \cellcolor{lightgray}LLaVA-Video-72B-Qwen2 \cite{LLaVA-Video-72B-Qwen2} & \cellcolor{lightgray}BLEU-1 \\


\addlinespace[4pt]
& 9 & Humor Video Captioning (Humor Cap) & Humanity, Social & Content Recognition, Affective Analysis, Cognition Understanding & MMBench-Video \cite{MMBench-Video-abs-2406-14515} & 33 & LLaVA-Video-72B-Qwen2 \cite{LLaVA-Video-72B-Qwen2} & BLEU-1 \\

\addlinespace[4pt]
\hline

\addlinespace[4pt]
\multirow{1}{*}{\shortstack[l]{\#V-C-5 \\ In-the-Wild\\ Video Object \\ Segmentation\\ (IW VOS)}}
& 1 & In the Wild Automobile Video Object Segmentation (IW-Auto VOS) \newline \color{white}{1} \newline \color{white}{1} \newline \color{white}{1}  \newline \color{white}{1}& General & Content Recognition, Interactive Capability & VIPSeg \cite{MiaoWWLZWY22} & 262 & SAM2 \cite{SAM2-abs-2408-00714} & mIoU \\

\addlinespace[4pt]
& \cellcolor{lightgray}2 & \cellcolor{lightgray}In the Wild Human Video Object Segmentation (IW-Human VOS) & \cellcolor{lightgray}Humanity & \cellcolor{lightgray}Content Recognition, Interactive Capability & \cellcolor{lightgray}VIPSeg \cite{MiaoWWLZWY22} & \cellcolor{lightgray}500 & \cellcolor{lightgray}SAM2 \cite{SAM2-abs-2408-00714} & \cellcolor{lightgray}mIoU \\

\addlinespace[4pt]
& 3 & In the Wild Animal Video Object Segmentation (IW-Animal VOS) & Biology & Content Recognition, Interactive Capability & VIPSeg \cite{MiaoWWLZWY22} & 70 & SAM2 \cite{SAM2-abs-2408-00714} & mIoU \\

\addlinespace[4pt]
& \cellcolor{lightgray}4 & \cellcolor{lightgray}In the Wild Furniture Video Object Segmentation (IW-Furniture VOS) & \cellcolor{lightgray}General & \cellcolor{lightgray}Content Recognition, Interactive Capability & \cellcolor{lightgray}VIPSeg \cite{MiaoWWLZWY22} & \cellcolor{lightgray}401 & \cellcolor{lightgray}SAM2 \cite{SAM2-abs-2408-00714} & \cellcolor{lightgray}mIoU \\

\addlinespace[4pt]
\hline

\addlinespace[4pt]
\multirow{16}{*}{\shortstack[l]{\#V-C-6 \\ General Video\\ Object Segmentation\\ (General VOS)}}
& 1 & Automobile Video Object Segmentation (Auto VOS) & General & Content Recognition, Interactive Capability & YouTube-VOS \cite{YouTube-VOS-abs-1809-03327} & 266 & SAM2 \cite{SAM2-abs-2408-00714} & mIoU \\

\addlinespace[4pt]
& \cellcolor{lightgray}2 & \cellcolor{lightgray}Human Video Object Segmentation (Human VOS) & \cellcolor{lightgray}Humanity & \cellcolor{lightgray}Content Recognition, Interactive Capability & \cellcolor{lightgray}YouTube-VOS \cite{YouTube-VOS-abs-1809-03327} & \cellcolor{lightgray}500 & \cellcolor{lightgray}SAM2 \cite{SAM2-abs-2408-00714} & \cellcolor{lightgray}mIoU \\

\addlinespace[4pt]
& 3 & Animal Video Object Segmentation (Animal VOS) & Biology & Content Recognition, Interactive Capability & YouTube-VOS \cite{YouTube-VOS-abs-1809-03327} & 500 & SAM2 \cite{SAM2-abs-2408-00714} & mIoU \\

\addlinespace[4pt]
& \cellcolor{lightgray}4 & \cellcolor{lightgray}Sports Video Object Segmentation (Sports VOS) & \cellcolor{lightgray}Culture & \cellcolor{lightgray}Content Recognition, Interactive Capability & \cellcolor{lightgray}YouTube-VOS \cite{YouTube-VOS-abs-1809-03327} & \cellcolor{lightgray}133 & \cellcolor{lightgray}SAM2 \cite{SAM2-abs-2408-00714} & \cellcolor{lightgray}mIoU \\

\addlinespace[4pt]
\hline

\addlinespace[4pt]
\multirow{11}{*}{\shortstack[l]{\#V-C-7 \\ Street-Scene Video\\ Object Segmentation\\ (Street VOS)}}
& 1 & Automobile Street-Scene Video Object Segmentation (Auto-Street VOS) & General & Content Recognition, Interactive Capability & Cityscapes \cite{Cityscapes-CordtsORREBFRS16} & 473 & SAM2 \cite{SAM2-abs-2408-00714} & mIoU \\

\addlinespace[4pt]
& \cellcolor{lightgray}2 & \cellcolor{lightgray}Human Street-Scene Video Object Segmentation (Human-Street VOS) & \cellcolor{lightgray}Humanity & \cellcolor{lightgray}Content Recognition, Interactive Capability & \cellcolor{lightgray}Cityscapes \cite{Cityscapes-CordtsORREBFRS16} & \cellcolor{lightgray}325 & \cellcolor{lightgray}SAM2 \cite{SAM2-abs-2408-00714} & \cellcolor{lightgray}mIoU \\

\addlinespace[4pt]
& 3 & Bicycle Street-Scene Video Object Segmentation (Bicycle-Street VOS) & General & Content Recognition,Interactive Capability & Cityscapes \cite{Cityscapes-CordtsORREBFRS16} & 108 & SAM2 \cite{SAM2-abs-2408-00714} & mIoU\\

\addlinespace[4pt]
\hline

\addlinespace[4pt]
\multirow{11}{*}{\shortstack[l]{\#V-C-8 \\ Referring Video\\ Object Segmentation\\ (RVOS)}}
& 1 & Human Referring Video Object Segmentation (Human RVOS) & Humanity & Content Recognition, Interactive Capability & Ref-DAVIS 2017 \cite{URVOS-SeoLH20} & 69 & UniRef++ \cite{UniRef-abs-2312-15715} & mIoU \\

\addlinespace[4pt]
& \cellcolor{lightgray}2 & \cellcolor{lightgray}Animal Referring Video Object Segmentation (Animal RVOS) & \cellcolor{lightgray}Biology & \cellcolor{lightgray}Content Recognition, Interactive Capability & \cellcolor{lightgray}Ref-DAVIS 2017 \cite{URVOS-SeoLH20} & \cellcolor{lightgray}25 & \cellcolor{lightgray}UniRef++ \cite{UniRef-abs-2312-15715} & \cellcolor{lightgray}mIoU \\

\addlinespace[4pt]
& 3 & Human Reasoning Video Object Segmentation (Human ReVOS) & Humanity & Reasoning Ability, Interactive Capability & ReVOS \cite{yan2024visa} & 500 & UniRef++ \cite{UniRef-abs-2312-15715} & mIoU \\

\addlinespace[4pt]
\hline

\addlinespace[4pt]
\multirow{6}{*}{\shortstack[l]{\#V-C-9 \\ Reasoning\\ Video Object\\ Segmentation\\ (ReVOS)}}
& 1 & Animal Reasoning Video Object Segmentation (Animal ReVOS) & Biology & Reasoning Ability, Interactive Capability & ReVOS \cite{yan2024visa} & 500 & UniRef++ \cite{UniRef-abs-2312-15715} & mIoU \\

\addlinespace[4pt]
& \cellcolor{lightgray}2 & \cellcolor{lightgray}Automobile Reasoning Video Object Segmentation (Auto ReVOS) & \cellcolor{lightgray}General & \cellcolor{lightgray}Reasoning Ability, Commonsense Knowledge & \cellcolor{lightgray}ReVOS \cite{yan2024visa} & \cellcolor{lightgray}259 & \cellcolor{lightgray}UniRef++ \cite{UniRef-abs-2312-15715} & \cellcolor{lightgray}mIoU \\

\addlinespace[4pt]
\hline

\addlinespace[4pt]
\multirow{6}{*}{\shortstack[l]{\#V-C-10 \\ Temporal Action\\ Detection \\ (Temp Act Det)}}
& 1 & Spatio-Temporal Static Action Detection (Static-Action Det) & General & Reasoning Ability, Causality Discrimination & HC-STVG2 \cite{HC-STVG2-TangLLLJJYX22} & 200 & TubeDETR \cite{TubeDETR-YangMSLS22} & m\_vIoU \\

\addlinespace[4pt]
& \cellcolor{lightgray}2 & \cellcolor{lightgray}Spatio-Temporal Dynamic Action Detection (Dynamic-Action Det) & \cellcolor{lightgray}General & \cellcolor{lightgray}Reasoning Ability, Causality Discrimination & \cellcolor{lightgray}HC-STVG2 \cite{HC-STVG2-TangLLLJJYX22} & \cellcolor{lightgray}200 & \cellcolor{lightgray}TubeDETR \cite{TubeDETR-YangMSLS22} & \cellcolor{lightgray}m\_vIoU \\

\addlinespace[4pt]
\hline

\addlinespace[4pt]
\multirow{1}{*}{\shortstack[l]{\#V-C-11 \\ Complex-Scene\\ Reasoning\\ Video Object \\ Segmentation\\ (C-ReVOS)}}
& 1 & Human Complex-Scene Reasoning Video Object Segmentation (Human C-ReVOS)  & General & Reasoning Ability, Commonsense Knowledge & SA-V \cite{SAM2-abs-2408-00714} & 500 & UniRef++ \cite{UniRef-abs-2312-15715} & mIoU \\

\addlinespace[4pt]
& \cellcolor{lightgray}2 & \cellcolor{lightgray}Animal Complex-Scene Reasoning Video Object Segmentation (Animal C-ReVOS) & \cellcolor{lightgray}Animal, Biology  & \cellcolor{lightgray}Reasoning Ability, Commonsense Knowledge & \cellcolor{lightgray}SA-V \cite{SAM2-abs-2408-00714} & \cellcolor{lightgray}108 & \cellcolor{lightgray}UniRef++ \cite{UniRef-abs-2312-15715} & \cellcolor{lightgray}mIoU \\

\addlinespace[4pt]
& 3 & Automobile Complex-Scene Reasoning Video Object Segmentation (Auto C-ReVOS) & General & Reasoning Ability, Commonsense Knowledge & SA-V \cite{SAM2-abs-2408-00714} & 117 & UniRef++ \cite{UniRef-abs-2312-15715} & mIoU \\

\addlinespace[4pt]
& \cellcolor{lightgray}4 & \cellcolor{lightgray}Human and Part Complex-Scene Reasoning Video Object Segmentation (Human-Part C-ReVOS) & \cellcolor{lightgray}General & \cellcolor{lightgray}Reasoning Ability, Commonsense Knowledge & \cellcolor{lightgray}SA-V \cite{SAM2-abs-2408-00714} & \cellcolor{lightgray}50 & \cellcolor{lightgray}UniRef++ \cite{UniRef-abs-2312-15715} & \cellcolor{lightgray}mIoU \\

\addlinespace[4pt]
& 5 & Equipment Complex-Scene Reasoning Video Object Segmentation (Equipment C-ReVOS) & Humanity & Reasoning Ability, Commonsense Knowledge & SA-V \cite{SAM2-abs-2408-00714} & 50 & UniRef++ \cite{UniRef-abs-2312-15715} & mIoU \\

\addlinespace[4pt]
\hline

\addlinespace[4pt]
\multirow{9}{*}{\shortstack[l]{\#V-C-12 \\ Video Grounding \\ (Vid-Ground)}}
& 1 & Human Video Grounding\newline (Human VG) & General & Reasoning Ability, Commonsense Knowledge, Causality Discrimination & VidSTG \cite{VidSTG-ZhangZZWLG20} & 200 & TubeDETR \cite{TubeDETR-YangMSLS22} & m\_vIoU \\

\addlinespace[4pt]
& \cellcolor{lightgray}2 & \cellcolor{lightgray}Animal Video Grounding\newline (Animal VG) & \cellcolor{lightgray}Biology & \cellcolor{lightgray}Reasoning Ability, Commonsense Knowledge, Causality Discrimination & \cellcolor{lightgray}VidSTG \cite{VidSTG-ZhangZZWLG20} & \cellcolor{lightgray}200 & \cellcolor{lightgray}TubeDETR \cite{TubeDETR-YangMSLS22} & \cellcolor{lightgray}m\_vIoU \\

\addlinespace[4pt]
& 3 & Automobile Video Grounding\newline (Auto VG) & General & Reasoning Ability, Commonsense Knowledge, Causality Discrimination & VidSTG \cite{VidSTG-ZhangZZWLG20} & 200 & TubeDETR \cite{TubeDETR-YangMSLS22} & m\_vIoU \\

\addlinespace[4pt]
\hline

\addlinespace[4pt]
\multirow{8}{*}{\shortstack[l]{\#V-C-13 \\ Video Depth\\ Estimation \\ (Vid-Depth Est)}}
& 1 & Sythetic Video Depth Estimation (Syn VDE) & General & Content Recognition & Sintel \cite{ButlerWSB12} & 23 & DepthCrafter \cite{DepthCrafter-abs-2409-02095} & absRel \\

\addlinespace[4pt]
& \cellcolor{lightgray}2 & \cellcolor{lightgray}Indoor Static Video Depth Estimation (Static VDE) & \cellcolor{lightgray}General & \cellcolor{lightgray}Content Recognition & \cellcolor{lightgray}Scannet \cite{DaiCSHFN17} & \cellcolor{lightgray}50 & \cellcolor{lightgray}DepthCrafter \cite{DepthCrafter-abs-2409-02095} & \cellcolor{lightgray}absRel \\

\addlinespace[4pt]
& 3 & Indoor Dynamic Video Depth Estimation (Dynamic VDE) & General & Content Recognition & Bonn \cite{PalazzoloBLGS19} & 50 & DepthCrafter \cite{DepthCrafter-abs-2409-02095} & absRel \\

\addlinespace[4pt]
& \cellcolor{lightgray}4 & \cellcolor{lightgray}Street Scene Video Depth Estimation (Street VDE) & \cellcolor{lightgray}General & \cellcolor{lightgray}Content Recognition & \cellcolor{lightgray}KITTI \cite{GeigerLU12} & \cellcolor{lightgray}13 & \cellcolor{lightgray}DepthCrafter \cite{DepthCrafter-abs-2409-02095} & \cellcolor{lightgray}absRel \\

\addlinespace[4pt]
\hline

\addlinespace[4pt]
\multirow{30}{*}{\shortstack[l]{\#V-C-14 \\ Object Matching \\ (Obj Match)}}
& 1 & Color Aware Object Matching (Color-Obj Match) & General & Content Recognition, Commonsense Knowledge, Reasoning Ability & MMVM \cite{MMVMBench} & 500 & CoLVA \cite{zhou2025CoLVA} & ACC \\

\addlinespace[4pt]
& \cellcolor{lightgray}2 & \cellcolor{lightgray}Shape or Posture Aware Matching (Shape Match) & \cellcolor{lightgray}General & \cellcolor{lightgray}Content Recognition, Commonsense Knowledge, Reasoning Ability & \cellcolor{lightgray}MMVM \cite{MMVMBench} & \cellcolor{lightgray}135 & \cellcolor{lightgray}CoLVA \cite{zhou2025CoLVA} & \cellcolor{lightgray}ACC \\

\addlinespace[4pt]
& 3 & Textual or Logo Markers Aware Object Matching (Logo Marker Match) & General & Content Recognition, Commonsense Knowledge, Reasoning Ability & MMVM \cite{MMVMBench} & 500 & CoLVA \cite{zhou2025CoLVA} & ACC \\

\addlinespace[4pt]
& \cellcolor{lightgray}4 & \cellcolor{lightgray}Size Aware Object Matching (Size Match) & \cellcolor{lightgray}General & \cellcolor{lightgray}Content Recognition, Commonsense Knowledge, Reasoning Ability & \cellcolor{lightgray}MMVM \cite{MMVMBench} & \cellcolor{lightgray}385 & \cellcolor{lightgray}CoLVA \cite{zhou2025CoLVA} & \cellcolor{lightgray}ACC \\

\addlinespace[4pt]
& 5 & Relative Position Aware Object Matching (Position Match) & General & Content Recognition, Commonsense Knowledge, Reasoning Ability & MMVM \cite{MMVMBench} & 500 & CoLVA \cite{zhou2025CoLVA} & ACC \\

\addlinespace[4pt]
& \cellcolor{lightgray}6 & \cellcolor{lightgray}Orientation and Movement Aware Object Matching (Motion Match) & \cellcolor{lightgray}General & \cellcolor{lightgray}Content Recognition, Commonsense Knowledge, Reasoning Ability & \cellcolor{lightgray}MMVM \cite{MMVMBench} & \cellcolor{lightgray}500 & \cellcolor{lightgray}CoLVA \cite{zhou2025CoLVA} & \cellcolor{lightgray}ACC \\

\addlinespace[4pt]
& 7 & Binding Relationship Aware Object Matching (Relation Match) & General & Content Recognition, Commonsense Knowledge, Reasoning Ability & MMVM \cite{MMVMBench} & 500 & CoLVA \cite{zhou2025CoLVA} & ACC \\

\addlinespace[4pt]
& \cellcolor{lightgray}8 & \cellcolor{lightgray}Object Markers Aware Object Matching (Object Marker Match) & \cellcolor{lightgray}General & \cellcolor{lightgray}Content Recognition, Commonsense Knowledge, Reasoning Ability & \cellcolor{lightgray}MMVM \cite{MMVMBench} & \cellcolor{lightgray}500 & \cellcolor{lightgray}CoLVA \cite{zhou2025CoLVA} & \cellcolor{lightgray}ACC \\

\addlinespace[4pt]
\hline

\addlinespace[4pt]
\multirow{1}{*}{\shortstack[l]{\#V-C-15 \\ Object Tracking \\ (Obj Track)}}
& 1 & Ball Tracking (Ball Track) & General & Content Recognition, Commonsense Knowledge & VOT2018 \cite{VOT2018-KristanLMFPZVBL18} & 255 & UniNext \cite{UniNext-0004JWWLYL23} & AUC \\

\addlinespace[4pt]
& \cellcolor{lightgray}2 & \cellcolor{lightgray}Vehicle Tracking (Vehicle Track) & \cellcolor{lightgray}General & \cellcolor{lightgray}Content Recognition, Commonsense Knowledge & \cellcolor{lightgray}VOT2018 \cite{VOT2018-KristanLMFPZVBL18} & \cellcolor{lightgray}95 & \cellcolor{lightgray}UniNext \cite{UniNext-0004JWWLYL23} & \cellcolor{lightgray}AUC \\

\addlinespace[4pt]
& 3 & Human Tracking (Human Track) & Humanity & Content Recognition, Commonsense Knowledge & VOT2018 \cite{VOT2018-KristanLMFPZVBL18} & 500 & UniNext \cite{UniNext-0004JWWLYL23} & AUC \\

\addlinespace[4pt]
& \cellcolor{lightgray}4 & \cellcolor{lightgray}Animal Tracking (Animal Track) & \cellcolor{lightgray}Biology & \cellcolor{lightgray}Content Recognition, Commonsense Knowledge & \cellcolor{lightgray}VOT2018 \cite{VOT2018-KristanLMFPZVBL18} & \cellcolor{lightgray}500 & \cellcolor{lightgray}UniNext \cite{UniNext-0004JWWLYL23} & \cellcolor{lightgray}AUC \\

\addlinespace[4pt]
& 5 & General Objects Tracking (General-Obj Track) & General & Content Recognition, Commonsense Knowledge & VOT2018 \cite{VOT2018-KristanLMFPZVBL18} & 500 & UniNext \cite{UniNext-0004JWWLYL23} & AUC \\

\addlinespace[4pt]
& \cellcolor{lightgray}6 & \cellcolor{lightgray}Human Part Tracking (Human-Part Track) & \cellcolor{lightgray}Humanity & \cellcolor{lightgray}Content Recognition, Commonsense Knowledge & \cellcolor{lightgray}LaSOT \cite{LaSOT-FanLYCDYBXLL19} & \cellcolor{lightgray}500 & \cellcolor{lightgray}UniNext \cite{UniNext-0004JWWLYL23} & \cellcolor{lightgray}AUC \\

\addlinespace[4pt]
& 7 & General Objects Part Tracking (Other-Part Track) & General & Content Recognition, Commonsense Knowledge & LaSOT \cite{LaSOT-FanLYCDYBXLL19} & 500 & UniNext \cite{UniNext-0004JWWLYL23} & AUC \\

\addlinespace[4pt]
\hline

\addlinespace[4pt]
\multirow{20}{*}{\shortstack[l]{\#V-C-16 \\ Long Video \\ Tracking \\ (Long-Vid Track)}}
& 1 & Long Video Human Tracking (Long-Vid-Human Track) & Humanity & Content Recognition, Commonsense Knowledge & LaSOT \cite{LaSOT-FanLYCDYBXLL19} & 10 & UniNext \cite{UniNext-0004JWWLYL23} & AUC \\

\addlinespace[4pt]
& \cellcolor{lightgray}2 & \cellcolor{lightgray}Long Video General Object Tracking (Long-Vid-Obj Track) & \cellcolor{lightgray}General & \cellcolor{lightgray}Content Recognition, Commonsense Knowledge & \cellcolor{lightgray}LaSOT \cite{LaSOT-FanLYCDYBXLL19} & \cellcolor{lightgray}10 & \cellcolor{lightgray}UniNext \cite{UniNext-0004JWWLYL23} & \cellcolor{lightgray}AUC \\

\addlinespace[4pt]
& 3 & Long Video Animal Tracking (Long-Vid-Animal Track) & Biology & Content Recognition, Commonsense Knowledge & LaSOT \cite{LaSOT-FanLYCDYBXLL19} & 10 & UniNext \cite{UniNext-0004JWWLYL23} & AUC \\

\addlinespace[4pt]
& \cellcolor{lightgray}4 & \cellcolor{lightgray}Long Video Vehicle Tracking (Long-Vid-Vehicle Track) & \cellcolor{lightgray}General & \cellcolor{lightgray}Content Recognition, Commonsense Knowledge & \cellcolor{lightgray}LaSOT \cite{LaSOT-FanLYCDYBXLL19} & \cellcolor{lightgray}10 & \cellcolor{lightgray}UniNext \cite{UniNext-0004JWWLYL23} & \cellcolor{lightgray}AUC \\

\addlinespace[4pt]
& 5 & Long Video Object Tracking in Crowd Sense (Long-Vid-Crowd Track) & Humanity, Culture & Content Recognition, Commonsense Knowledge & DanceTrack \cite{DanceTrack-SunCJYBKL22} & 10 & UniNext \cite{UniNext-0004JWWLYL23} & AUC \\

\addlinespace[4pt]
\hline

\addlinespace[4pt]
\multirow{15}{*}{\shortstack[l]{\#V-C-17 \\ UAV Object \\ Tracking \\ (UAV Track)}}
& 1 & UAV Video Human Tracking (UAV Human Track) & Humanity & Content Recognition, Commonsense Knowledge & UAV123 \cite{UAV123-MuellerSG16} & 500 & UniNext \cite{UniNext-0004JWWLYL23} & AUC \\

\addlinespace[4pt]
& \cellcolor{lightgray}2 & \cellcolor{lightgray}UAV Video Vehicle Tracking (UAV-Vehicle Track) & \cellcolor{lightgray}General & \cellcolor{lightgray}Content Recognition, Commonsense Knowledge & \cellcolor{lightgray}UAV123 \cite{UAV123-MuellerSG16} & \cellcolor{lightgray}500 & \cellcolor{lightgray}UniNext \cite{UniNext-0004JWWLYL23} & \cellcolor{lightgray}AUC \\

\addlinespace[4pt]
& 3 & UAV Video UAV Tracking (UAV-UAV Track) & General & Content Recognition, Commonsense Knowledge & UAV123 \cite{UAV123-MuellerSG16} & 500 & UniNext \cite{UniNext-0004JWWLYL23} & AUC \\

\addlinespace[4pt]
& \cellcolor{lightgray}4 & \cellcolor{lightgray}UAV Video Building Tracking (UAV-Build Track) & \cellcolor{lightgray}Culture & \cellcolor{lightgray}Content Recognition, Commonsense Knowledge & \cellcolor{lightgray}UAV123 \cite{UAV123-MuellerSG16} & \cellcolor{lightgray}500 & \cellcolor{lightgray}UniNext \cite{UniNext-0004JWWLYL23} & \cellcolor{lightgray}AUC \\

\addlinespace[4pt]
& 5 & UAV Video General Object Tracking (UAV-Obj Track) & General & Content Recognition, Commonsense Knowledge & UAV123 \cite{UAV123-MuellerSG16} & 500 & UniNext \cite{UniNext-0004JWWLYL23} & AUC \\

\addlinespace[4pt]
\hline

\addlinespace[4pt]
\multirow{15}{*}{\shortstack[l]{\#V-C-18 \\ Underwater Object\\ Tracking \\ (UW Track)}}
& 1 & Underwater Video Object Tracking in Blue Water (Blue-Water Track) & Biology & Content Recognition, Commonsense Knowledge & UTB180 \cite{UTB180-alawode2022utb180} & 500 & UniNext \cite{UniNext-0004JWWLYL23} & AUC \\

\addlinespace[4pt]
& \cellcolor{lightgray}2 & \cellcolor{lightgray}Underwater Video Object Tracking in Yellow Water (Yellow-Water Track) & \cellcolor{lightgray}Biology & \cellcolor{lightgray}Content Recognition, Commonsense Knowledge & \cellcolor{lightgray}UTB180 \cite{UTB180-alawode2022utb180} & \cellcolor{lightgray}410 & \cellcolor{lightgray}UniNext \cite{UniNext-0004JWWLYL23} & \cellcolor{lightgray}AUC \\

\addlinespace[4pt]
& 3 & Underwater Video Object Tracking in Green Water (Green-Water Track) & Biology & Content Recognition, Commonsense Knowledge & UTB180 \cite{UTB180-alawode2022utb180} & 500 & UniNext \cite{UniNext-0004JWWLYL23} & AUC \\

\addlinespace[4pt]
& \cellcolor{lightgray}4 & \cellcolor{lightgray}Underwater Video Object Tracking in White Water (White-Water Track) & \cellcolor{lightgray}Biology & \cellcolor{lightgray}Content Recognition, Commonsense Knowledge & \cellcolor{lightgray}UTB180 \cite{UTB180-alawode2022utb180} & \cellcolor{lightgray}500 & \cellcolor{lightgray}UniNext \cite{UniNext-0004JWWLYL23} & \cellcolor{lightgray}AUC \\

\addlinespace[4pt]
& 5 & Video Object Tracking in Crowd Sense (Crowed Track) & Humanity, Culture & Content Recognition, Commonsense Knowledge & DanceTrack \cite{DanceTrack-SunCJYBKL22} & 500 & UniNext \cite{UniNext-0004JWWLYL23} & AUC \\

\addlinespace[4pt]
\hline

\addlinespace[4pt]
\multirow{1}{*}{\shortstack[l]{\#V-C-19 \\ Optical Flow \\ (Opt Flow)}}
& 1 & Optical Flow in Simple Synthetic Scene (Synthetic Scene Flow Estimate) & General & Content Recognition & FlyingChairs \cite{DosovitskiyFIHH15} & 150 & VideoFlow \cite{deArmas2019videoflow} & EPE \\

\addlinespace[4pt]
& \cellcolor{lightgray}2 & \cellcolor{lightgray}Optical Flow in Complex Scene Estimation (Complex Scene Estimate) & \cellcolor{lightgray}General & \cellcolor{lightgray}Content Recognition & \cellcolor{lightgray}ChairsSDHom \cite{IMKDB17} & \cellcolor{lightgray}150 & \cellcolor{lightgray}VideoFlow \cite{deArmas2019videoflow} & \cellcolor{lightgray}EPE \\

\addlinespace[4pt]
& 3 & Panoramic Optical Flow Estimation (Panoramic Flow Estimate) & General & Content Recognition & OmniFlow \cite{seidel2021omniflow} & 150 & PanoFlow \cite{shi2022panoflow} & EPE \\

\hline

\addlinespace[4pt]
\multirow{10}{*}{\shortstack[l]{\#V-C-20 \\ Video Event\\ Recognition \\ (Vid-Event Recog)}}
& 1 & News and Documentary Video Captioning (News\&Doc Video Cap) & General & Content Recognition, Commonsense Knowledge & VideoMME \cite{Video-MME-abs-2405-21075} & 60 & LLaVA-Video-72B-Qwen2 \cite{LLaVA-Video-72B-Qwen2} & BLEU-1 \\

\addlinespace[4pt]
& \cellcolor{lightgray}2 & \cellcolor{lightgray}Multimodal SarcaS-measure Detection (SarcaS-measure Det) & \cellcolor{lightgray}Humanities & \cellcolor{lightgray}Content Recognition & \cellcolor{lightgray}MUStARD \cite{CastroHPZMP19} & \cellcolor{lightgray}500 & \cellcolor{lightgray}SVM \cite{CastroHPZMP19} & \cellcolor{lightgray}F1 \\

\addlinespace[4pt]
& 3 & Video Semantic Role Labeling (Vid SRL) & General & Content Recognition & VidSitu \cite{Sadhu_2021_CVPR} & 500 & TxEnc-Dec using I3D features \cite{Sadhu_2021_CVPR} & CIDEr \\

\addlinespace[4pt]
& \cellcolor{lightgray}4 & \cellcolor{lightgray}Video Event Relation Prediction (Vid Evnt-Rel Pred) & \cellcolor{lightgray}General & \cellcolor{lightgray}Content Recognition & \cellcolor{lightgray}VidSitu \cite{Sadhu_2021_CVPR} & \cellcolor{lightgray}500 & \cellcolor{lightgray}Roberta + I3D features \cite{Sadhu_2021_CVPR} & \cellcolor{lightgray}Macro-Acc \\

\addlinespace[4pt]
\hline

\end{longtable}
}


{\centering
\fontsize{8}{8}\selectfont
\setlength{\tabcolsep}{0.7mm}
\renewcommand{\arraystretch}{0.9} 
\captionof{table}{
Detailed list of all tasks and skills (meta-tasks) under video and generation category.
}
\vspace{-3mm}
\label{tab:task-specification-vid-gen}
\begin{longtable}[!t]{p{2.2cm}%
                >{\centering\arraybackslash}p{0.2cm}
                p{4cm}%
                >{\centering\arraybackslash}p{1.3cm}%
                p{2.3cm}%
                p{2cm}%
                >{\centering\arraybackslash}p{0.7cm}%
                p{2cm}%
                >{\centering\arraybackslash}p{1.3cm}}

\addlinespace[3pt]
\rowcolor{bg-tb-heavey-video} \multicolumn{9}{c}{\bf \textcolor{blueGen}{\large Video Generation Group}} \\ [4pt]

\hline

\addlinespace[4pt]
\multicolumn{5}{c}{\bf Task} & \multicolumn{2}{c}{\bf Data}  & \multicolumn{2}{c}{\bf \textbf{SoTA Specialist}} \\

\cmidrule(r){1-5}  \cmidrule(r){6-7}  \cmidrule(r){8-9} 

\textbf{Skill (Meta-Task)} & \textbf{\#}& \textbf{Task (Short Name)} &  \textbf{Domain} & \textbf{Capability} & \textbf{Data Source} & \textbf{Number} & \multicolumn{1}{c}{\textbf{Model}}  & \textbf{Metrics} \\ 

\addlinespace[4pt]
\hline
\endfirsthead
\hline
\addlinespace[4pt]
\multicolumn{5}{c}{\bf Task} & \multicolumn{2}{c}{\bf Data}  & \multicolumn{2}{c}{\bf \textbf{SoTA Specialist}} \\
\cmidrule(r){1-5}  \cmidrule(r){6-7}  \cmidrule(r){8-9} 
\textbf{Skill (Meta-Task)} & \textbf{\#}& \textbf{Task (Short Name)} &  \textbf{Domain} & \textbf{Capability} & \textbf{Data Source} & \textbf{Number} & \multicolumn{1}{c}{\textbf{Model}}  & \textbf{Metrics} \\ 
\addlinespace[4pt]
\hline
\endhead
\hline
\endfoot

\addlinespace[4pt]
\multirow{30}{*}{\shortstack[l]{\#V-G-1 \\ Text-to-Video\\ Generation \\ (Txt2Vid Gen)}}
& 1 & Subject-Driven Text to Video Generation (Subj-Txt2Vid Gen) & General & Creativity and lnnovation & VBench \cite{VBench-video} & 500 & Zeroscope \cite{video-diff-zeroscope} & DINO-Score \\

\addlinespace[4pt]
& \cellcolor{lightgray}2 & \cellcolor{lightgray}Background-Consistency Text to Video Geneation (BG-Consis-Txt2Vid Gen) & \cellcolor{lightgray}General & \cellcolor{lightgray}Creativity and lnnovation & \cellcolor{lightgray}VBench \cite{VBench-video} & \cellcolor{lightgray}500 & \cellcolor{lightgray}LaVie \cite{LAVIE-abs-2309-15103} & \cellcolor{lightgray}CLIP-Score \\

\addlinespace[4pt]
& 3 & Static Video Generation (Static-Txt2Vid Gen) & General & Creativity and lnnovation & VBench \cite{VBench-video} & 500 & LaVie \cite{LAVIE-abs-2309-15103} & L1-Dis \\

\addlinespace[4pt]
& \cellcolor{lightgray}4 & \cellcolor{lightgray}Dynamic Video Generation (Dynamic-Txt2Vid Gen) & \cellcolor{lightgray}General & \cellcolor{lightgray}Creativity and lnnovation & \cellcolor{lightgray}VBench \cite{VBench-video} & \cellcolor{lightgray}500 & \cellcolor{lightgray}CogVideoX-5b \cite{CogVideoX-abs-2408-06072} & \cellcolor{lightgray}OFS \\

\addlinespace[4pt]
& 5 & Artistic Content Text to Video Generation (Artistic-Txt2Vid Gen) & Art & Creativity and lnnovation & laion-aesthetics \cite{Laion-aesthetics} & 500 & CogVideoX-5b \cite{CogVideoX-abs-2408-06072} & Aesth-Score \\

\addlinespace[4pt]
& \cellcolor{lightgray}6 & \cellcolor{lightgray}Scene Text to Video Generation (Scene-Txt2Vid Gen) & \cellcolor{lightgray}General & \cellcolor{lightgray}Creativity and lnnovation & \cellcolor{lightgray}laion-aesthetics \cite{Laion-aesthetics} & \cellcolor{lightgray}500 & \cellcolor{lightgray}CogVideoX-5b \cite{CogVideoX-abs-2408-06072} & \cellcolor{lightgray}MUSIQ \\

\addlinespace[4pt]
& 7 & Class-Conditioned Text to Video Generation (Class-Cond-Txt2Vid Gen) & General & Creativity and lnnovation & VBench \cite{VBench-video} & 500 & LaVie \cite{LAVIE-abs-2309-15103} & Suc-Rate \\

\addlinespace[4pt]
& \cellcolor{lightgray}8 & \cellcolor{lightgray}Multi-Class-Conditioned Text to Video Generation (MulClass-Cond-Txt2Vid Gen) & \cellcolor{lightgray}General & \cellcolor{lightgray}Creativity and lnnovation & \cellcolor{lightgray}VBench \cite{VBench-video} & \cellcolor{lightgray}500 & \cellcolor{lightgray}CogVideoX-2b \cite{CogVideoX-abs-2408-06072} & \cellcolor{lightgray}Suc-Rate \\

\addlinespace[4pt]
& 9 & Stylized Video Generation-Single Reference (Stylized-Vid Gen) & Art & Creativity and lnnovation & StyleCrafter \cite{StyleCrafter-abs-2312-00330} & 38 & StyleCrafter \cite{StyleCrafter-abs-2312-00330} & CLIP-Score \\

\addlinespace[4pt]
& \cellcolor{lightgray}10 & \cellcolor{lightgray}Stylized Video Generation-Multiple Reference (M-Stylized-Vid Gen) & \cellcolor{lightgray}Art & \cellcolor{lightgray}Creativity and lnnovation & \cellcolor{lightgray}StyleCrafter \cite{StyleCrafter-abs-2312-00330} & \cellcolor{lightgray}38 & \cellcolor{lightgray}StyleCrafter \cite{StyleCrafter-abs-2312-00330} & \cellcolor{lightgray}CLIP-Score \\

\addlinespace[4pt]
& 11 & Spatial Relation Video Generation (Spatial-Rel-Txt2Vid Gen) & General & Creativity and lnnovation & VBench \cite{VBench-video} & 500 & Zeroscope \cite{} & Suc-Rate \\

\addlinespace[4pt]
& \cellcolor{lightgray}12 & \cellcolor{lightgray}Camera Motion Video Generation (Camera-Txt2Vid Gen) & \cellcolor{lightgray}General & \cellcolor{lightgray}Creativity and lnnovation & \cellcolor{lightgray}VBench \cite{VBench-video} & \cellcolor{lightgray}500 & \cellcolor{lightgray}CogVideoX-5b \cite{CogVideoX-abs-2408-06072} & \cellcolor{lightgray}Suc-Rate \\

\addlinespace[4pt]
& 13 & Terrestrial Animal Video Generation (Animal-Txt2Vid Gen) & Animal & Creativity and lnnovation & WebVid10M \cite{Bain21} & 500 & CogVideoX-5b \cite{CogVideoX-abs-2408-06072} & FVD \\

\addlinespace[4pt]
\hline

\addlinespace[4pt]
\multirow{6}{*}{\shortstack[l]{\#V-G-2 \\ Conditional Video\\ Generation \\ (Condt Vid Gen)}}
& 1 & Style-Specific Text to Video Generation (Style-Txt2Vid Gen) & Art & Creativity and Innovation & Manual \cite{VBench-video} & 500 & VideoCrafterv2 \cite{chen2024videocrafter2} & CLIP-Score \\

\addlinespace[4pt]
& \cellcolor{lightgray}2 & \cellcolor{lightgray}Color-Specific Text to Video Generation (Color-Txt2Vid Gen) & \cellcolor{lightgray}General & \cellcolor{lightgray}Creativity and Innovation & \cellcolor{lightgray}Manual \cite{VBench-video} & \cellcolor{lightgray}500 & \cellcolor{lightgray}LaVie \cite{LAVIE-abs-2309-15103} & \cellcolor{lightgray}Suc-Rate \\

\addlinespace[4pt]
& 3 & Material-Specific Text to Video Generation (Material-Txt2Vid Gen) & General & Creativity and Innovation & Manual \cite{VBench-video} & 500 & CogVideoX-5b \cite{CogVideoX-abs-2408-06072} & Suc-Rate \\

\addlinespace[4pt]
\hline

\addlinespace[4pt]
\multirow{5}{*}{\shortstack[l]{\#V-G-3 \\ Video Action\\ Generation \\ (Act Txt2Vid Gen)}}
& 1 & Action-Specific Text to Video Generation (Action-Txt2Vid Gen) & Sports & Creativity and lnnovation & Kinetics700 \cite{Kinetics-700-abs-2010-10864} & 500 & LaVie \cite{LAVIE-abs-2309-15103} & Suc-Rate \\

\addlinespace[4pt]
& \cellcolor{lightgray}2 & \cellcolor{lightgray}Human Video Generation (Human-Txt2Vid Gen) & \cellcolor{lightgray}Humanities & \cellcolor{lightgray}Creativity and lnnovation & \cellcolor{lightgray}WebVid10M \cite{Bain21} & \cellcolor{lightgray}500 & \cellcolor{lightgray}CogVideoX-5b \cite{CogVideoX-abs-2408-06072} & \cellcolor{lightgray}FVD \\

\addlinespace[4pt]
& 3 & Athletics Video Generation (Athletics-Txt2Vid Gen) & Sports & Creativity and lnnovation & UCF101 \cite{soomro2012ucf101} & 500 & CogVideoX-5b \cite{CogVideoX-abs-2408-06072} & FVD \\

\addlinespace[4pt]
& \cellcolor{lightgray}4 & \cellcolor{lightgray}Concert Video Generation (Concert-Txt2Vid Gen) & \cellcolor{lightgray}Sports & \cellcolor{lightgray}Creativity and lnnovation & \cellcolor{lightgray}UCF101 \cite{soomro2012ucf101} & \cellcolor{lightgray}500 & \cellcolor{lightgray}CogVideoX-5b \cite{CogVideoX-abs-2408-06072} & \cellcolor{lightgray}FVD \\

\addlinespace[4pt]
& 5 & Water Sports Video Generation (Water-Sports-Txt2Vid Gen) & Sports & Creativity and lnnovation & UCF101 \cite{soomro2012ucf101} & 500 & CogVideoX-5b \cite{CogVideoX-abs-2408-06072} & FVD \\

\addlinespace[4pt]
\hline

\addlinespace[4pt]
\multirow{25}{*}{\shortstack[l]{\#V-G-4 \\ Image-to-Video\\ Generation \\ (Img2Vid Gen)}}
& 1 & Plant Image to Video Generation (Plant-Img2Vid Gen) \newline \color{white}{1} \newline \color{white}{1} \newline \color{white}{1} & Biology & Creativity and lnnovation & VBench \cite{VBench-video} & 500 & CogVideoX-5b-I2V \cite{CogVideoX-abs-2408-06072} & Avg({\tiny DINO+CLIP}\newline{\tiny+OFS+MSS}) \\

\addlinespace[4pt]
& \cellcolor{lightgray}2 & \cellcolor{lightgray}Human Image to Video Generation (Human-Img2Vid Gen) & \cellcolor{lightgray}Humanities & \cellcolor{lightgray}Creativity and lnnovation & \cellcolor{lightgray}VBench \cite{VBench-video} & \cellcolor{lightgray}500 & \cellcolor{lightgray}CogVideoX-5b-I2V \cite{CogVideoX-abs-2408-06072} & \cellcolor{lightgray}Avg({\tiny DINO+CLIP}\newline{\tiny+OFS+MSS}) \\

\addlinespace[4pt]
& 3 & Wild Animal Image to Video Generation (Wild-Animal-Img2Vid Gen) & Animal & Creativity and lnnovation & VBench \cite{VBench-video} & 500 & CogVideoX-5b-I2V \cite{CogVideoX-abs-2408-06072} & Avg({\tiny DINO+CLIP}\newline{\tiny+OFS+MSS}) \\

\addlinespace[4pt]
& \cellcolor{lightgray}4 & \cellcolor{lightgray}Architecture Image to Video Generation (Architect-Img2Vid Gen) & \cellcolor{lightgray}Engineering & \cellcolor{lightgray}Creativity and lnnovation & \cellcolor{lightgray}VBench \cite{VBench-video} & \cellcolor{lightgray}500 & \cellcolor{lightgray}CogVideoX-5b-I2V \cite{CogVideoX-abs-2408-06072} & \cellcolor{lightgray}Avg({\tiny DINO+CLIP}\newline{\tiny+OFS+MSS}) \\

\addlinespace[4pt]
& 5 & Food Image to Video Generation (Food-Img2Vid Gen) & Daily & Creativity and lnnovation & VBench \cite{VBench-video} & 500 & CogVideoX-5b-I2V \cite{CogVideoX-abs-2408-06072} & Avg({\tiny DINO+CLIP}\newline{\tiny+OFS+MSS}) \\

\addlinespace[4pt]
& \cellcolor{lightgray}6 & \cellcolor{lightgray}Pet Image to Video Generation (Pet-Img2Vid Gen) & \cellcolor{lightgray}Animal & \cellcolor{lightgray}Creativity and lnnovation & \cellcolor{lightgray}VBench \cite{VBench-video} & \cellcolor{lightgray}500 & \cellcolor{lightgray}CogVideoX-5b-I2V \cite{CogVideoX-abs-2408-06072} & \cellcolor{lightgray}Avg({\tiny DINO+CLIP}\newline{\tiny+OFS+MSS}) \\

\addlinespace[4pt]
& 7 & Scene Image to Video Generation (Scene-Img2Vid Gen) & Art & Creativity and lnnovation & VBench \cite{VBench-video} & 500 & CogVideoX-5b-I2V \cite{CogVideoX-abs-2408-06072} & Avg({\tiny DINO+CLIP}\newline{\tiny+OFS+MSS}) \\

\addlinespace[4pt]
& \cellcolor{lightgray}8 & \cellcolor{lightgray}Vehicle Image to Video Generation (Vehicle-Img2Vid Gen) & \cellcolor{lightgray}Daily & \cellcolor{lightgray}Creativity and lnnovation & \cellcolor{lightgray}VBench \cite{VBench-video} & \cellcolor{lightgray}500 & \cellcolor{lightgray}CogVideoX-5b-I2V \cite{CogVideoX-abs-2408-06072} & \cellcolor{lightgray}Avg({\tiny DINO+CLIP}\newline{\tiny+OFS+MSS}) \\

\addlinespace[4pt]
& 9 & Furniture Image to Video Generation (Furniture-Img2Vid Gen) & Daily & Creativity and lnnovation & Manual & 500 & CogVideoX-5b-I2V \cite{CogVideoX-abs-2408-06072} &Avg({\tiny DINO+CLIP}\newline{\tiny+OFS+MSS})\\

\addlinespace[4pt]
& \cellcolor{lightgray}10 & \cellcolor{lightgray}Cloth Image to Video Generation (Cloth-Img2Vid Gen) & \cellcolor{lightgray}Daily & \cellcolor{lightgray}Creativity and lnnovation & \cellcolor{lightgray}Manual & \cellcolor{lightgray}500 & \cellcolor{lightgray}CogVideoX-5b-I2V \cite{CogVideoX-abs-2408-06072} & \cellcolor{lightgray}Avg({\tiny DINO+CLIP}\newline{\tiny+OFS+MSS}) \\

\addlinespace[4pt]
& 11 & Weather Image to Video Generation (Weather-Img2Vid Gen) & Climate & Creativity and lnnovation & Manual  & 500 & CogVideoX-5b-I2V \cite{CogVideoX-abs-2408-06072} & Avg({\tiny DINO+CLIP}\newline{\tiny+OFS+MSS}) \\

\addlinespace[4pt]
\hline

\addlinespace[4pt]
\multirow{21}{*}{\shortstack[l]{\#V-G-5 \\ Video \\ Enhancement \\ (Vid Enhance)}}
& 1 & Video Object Demoireing (Vid-Obj Demoireing) & General & Creativity and lnnovation & Vdmoire\_iphonev2, RawVDemoire \cite{dai2022video,yue2023recaptured} & 93 & DTNet \cite{XuSCZ24} & PSNR \\

\addlinespace[4pt]
& \cellcolor{lightgray}2 & \cellcolor{lightgray}Video Denoising (Vid Denoising) & \cellcolor{lightgray}General & \cellcolor{lightgray}Creativity and lnnovation & \cellcolor{lightgray}CRVD, DAVIS\_2017 \cite{yue2020supervised,Pont-Tuset_arXiv_2017} & \cellcolor{lightgray}51 & \cellcolor{lightgray}BSVD \cite{qi2022BSVD} & \cellcolor{lightgray}PSNR \\

\addlinespace[4pt]
& 3 & Video Frame Interpolation (Vid-Frame Interpolate) & General & Reasoning Ability, Creativity and Innovation & Vimeo-90k \cite{XueCWWF19} & 79 & EMA-VFI \cite{zhang2023extracting} & PSNR \\

\addlinespace[4pt]
& \cellcolor{lightgray}4 & \cellcolor{lightgray}Video Colorization (Vid Colorization) & \cellcolor{lightgray}General & \cellcolor{lightgray}Creativity and Innovation & \cellcolor{lightgray}MSU Video Colorization Benchmark \cite{MSU-Video-Colorization-Benchmark} & \cellcolor{lightgray}36 & \cellcolor{lightgray}LVVCP \cite{HofingerKobler22} & \cellcolor{lightgray}PSNR \\

\addlinespace[4pt]
& 5 & Video Dehazing (Vid Dehazing) & General & Content Recognition & Real Haze Video Database \cite{Real-Haze-Video-Database} & 403 & MAP-Net \cite{xu2023map} & PSNR \\

\addlinespace[4pt]
& \cellcolor{lightgray}6 & \cellcolor{lightgray}Video Desnowing (Vid Desnowing) & \cellcolor{lightgray}General & \cellcolor{lightgray}Content Recognition & \cellcolor{lightgray}RVSD, realsnow85 \cite{WuYARCCZ24,ChenRGWLCZ23} & \cellcolor{lightgray}55 & \cellcolor{lightgray}Turtle \cite{ghasemabadilearning} & \cellcolor{lightgray}PSNR \\

\addlinespace[4pt]
& 7 & Video Deraining (Vid Deraining) & General & Content Recognition & waterdrop removal(VWRD), VRDS \cite{wen2023video,wu2023mask} & 45 & RainMamba \cite{wu2024rainmamba} & PSNR \\

\addlinespace[4pt]
& \cellcolor{lightgray}8 & \cellcolor{lightgray}Road Scene Video Superresolution (Scence-Vid Superres) & \cellcolor{lightgray}Daily & \cellcolor{lightgray}Content Recognition & \cellcolor{lightgray}VideoLQ \cite{chan2022investigating} & \cellcolor{lightgray}34 & \cellcolor{lightgray}Upscale-A-Video \cite{zhou2024upscaleavideo} & \cellcolor{lightgray}MUSIQ \\

\addlinespace[4pt]
& 9 & Real World Video Superresolution (Real-Scence-Vid Superres) & General & Content Recognition & VideoLQ \cite{chan2022investigating} & 15 & Upscale-A-Video \cite{zhou2024upscaleavideo} & MUSIQ \\

\addlinespace[4pt]
\hline

\addlinespace[4pt]
\multirow{12}{*}{\shortstack[l]{\#V-G-6 \\ Video Editing \\ (Vid Edit)}}
& 1 & Video Animal Inpainting (Animal Inpainting) & Animal & Reasoning Ability, Creativity and Innovation & YouTube-VOS \cite{YouTube-VOS-abs-1809-03327} & 250 & ProPainter \cite{zhou2023propainter} & PSNR \\

\addlinespace[4pt]
& \cellcolor{lightgray}2 & \cellcolor{lightgray}Video Non-Animal Inpainting (Non-Animal Inpainting) & \cellcolor{lightgray}General & \cellcolor{lightgray}Reasoning Ability, Creativity and Innovation & \cellcolor{lightgray}YouTube-VOS \cite{YouTube-VOS-abs-1809-03327} & \cellcolor{lightgray}195 & \cellcolor{lightgray}ProPainter \cite{zhou2023propainter} & \cellcolor{lightgray}PSNR \\

\addlinespace[4pt]
& 3 & Video Deblurring (Vid Deblur) & General & Content Recognition & REDS \cite{Nah_2019_CVPR_Workshops_REDS} & 60 & VRT \cite{liang2022vrt} & PSNR \\

\addlinespace[4pt]
& \cellcolor{lightgray}4 & \cellcolor{lightgray}Video Translation (Vid Translation) & \cellcolor{lightgray}Art & \cellcolor{lightgray}Creativity and lnnovation & \cellcolor{lightgray}FRESCO \cite{yang2024fresco} & \cellcolor{lightgray}19 & \cellcolor{lightgray}FRESCO \cite{yang2024fresco} & \cellcolor{lightgray}Fram-Acc \\

\addlinespace[4pt]
& 5 & Portrait Video Style Transfer (Style Transfer) & Art & Creativity and lnnovation & VToonify \cite{yang2022Vtoonify} & 19 & VToonify \cite{yang2022Vtoonify} & UPR \\

\addlinespace[4pt]
\hline

\end{longtable}
}

\clearpage

{\centering
\fontsize{8}{8}\selectfont
\setlength{\tabcolsep}{0.7mm}
\renewcommand{\arraystretch}{0.9} 
\captionof{table}{
Detailed list of all tasks and skills (meta-tasks) under audio and comprehension category.
}
\vspace{-3mm}
\label{tab:task-specification-aud-comp}
\begin{longtable}[!t]{p{2.2cm}%
                >{\centering\arraybackslash}p{0.2cm}
                p{4cm}%
                >{\centering\arraybackslash}p{1.3cm}%
                p{2.3cm}%
                p{2cm}%
                >{\centering\arraybackslash}p{0.7cm}%
                p{2cm}%
                >{\centering\arraybackslash}p{1.3cm}}

\addlinespace[3pt]
\rowcolor{bg-tb-heavey-audio} \multicolumn{9}{c}{\bf \textcolor{greenCom}{\large Audio Comprehension Group}} \\ [4pt]

\hline

\addlinespace[4pt]
\multicolumn{5}{c}{\bf Task} & \multicolumn{2}{c}{\bf Data}  & \multicolumn{2}{c}{\bf \textbf{SoTA Specialist}} \\

\cmidrule(r){1-5}  \cmidrule(r){6-7}  \cmidrule(r){8-9} 

\textbf{Skill (Meta-Task)} & \textbf{\#}& \textbf{Task (Short Name)} &  \textbf{Domain} & \textbf{Capability} & \textbf{Data Source} & \textbf{Number} & \multicolumn{1}{c}{\textbf{Model}}  & \textbf{Metrics} \\ 

\addlinespace[4pt]
\hline
\endfirsthead
\hline
\addlinespace[4pt]
\multicolumn{5}{c}{\bf Task} & \multicolumn{2}{c}{\bf Data}  & \multicolumn{2}{c}{\bf \textbf{SoTA Specialist}} \\
\cmidrule(r){1-5}  \cmidrule(r){6-7}  \cmidrule(r){8-9} 
\textbf{Skill (Meta-Task)} & \textbf{\#}& \textbf{Task (Short Name)} &  \textbf{Domain} & \textbf{Capability} & \textbf{Data Source} & \textbf{Number} & \multicolumn{1}{c}{\textbf{Model}}  & \textbf{Metrics} \\ 
\addlinespace[4pt]
\hline
\endhead
\hline
\endfoot

\addlinespace[4pt]
\multirow{8}{*}{\shortstack[l]{\#A-C-1 \\ Speech Accent\\ Understanding \\ (Acnt Analy)}}
& 1 & Accent Classification\newline (Accent Recog) & Linguistics & Reasoning Ability & Speech Accent Archive \cite{Speech-Accent-Archive} & 500 & CLAP \cite{CLAP-ElizaldeDIW23} & Acc \\

\addlinespace[4pt]
& \cellcolor{lightgray}2 & \cellcolor{lightgray}Accent Sex Classification\newline (Accent-Sex Recog) & \cellcolor{lightgray}Linguistics & \cellcolor{lightgray}Reasoning Ability, Commonsense Knowledge & \cellcolor{lightgray}Speech Accent Archive \cite{Speech-Accent-Archive} & \cellcolor{lightgray}500 & \cellcolor{lightgray}CLAP \cite{CLAP-ElizaldeDIW23} & \cellcolor{lightgray}Acc \\

\addlinespace[4pt]
& 3 & Speaker Identification\newline (Speaker ID) & Linguistics & Reasoning Ability & VoxCeleb \cite{VoxCeleb-NagraniCZ17} & 279 & HuBERT Large \cite{HuBERT-HsuBTLSM21} & Acc \\

\addlinespace[4pt]
& \cellcolor{lightgray}4 & \cellcolor{lightgray}Vocal Sound Classification\newline (Vocal-Sound Recog) & \cellcolor{lightgray}General & \cellcolor{lightgray}Reasoning Ability & \cellcolor{lightgray}Vocal Sound \cite{Vocalsound-GongYG22} & \cellcolor{lightgray}500 & \cellcolor{lightgray}CLAP \cite{CLAP-ElizaldeDIW23} & \cellcolor{lightgray}Acc \\

\addlinespace[4pt]
\hline

\addlinespace[4pt]
\multirow{8}{*}{\shortstack[l]{\#A-C-2 \\ Speech Content\\ Understanding \\ (Ctnt Analy)}}
& 1 & Intent Classification\newline (Intent Recog) & General & Reasoning Ability & Fluent Speech Commands \cite{Flue-speech-command-LugoschRITB19} & 500 & HuBERT Large \cite{HuBERT-HsuBTLSM21} & Acc \\

\addlinespace[4pt]
& \cellcolor{lightgray}2 & \cellcolor{lightgray}Speech Command\newline (Speech Cmd) & \cellcolor{lightgray}General & \cellcolor{lightgray}Content Recognition & \cellcolor{lightgray}speech-commands \cite{Speech-Commands-abs-1804-03209} & \cellcolor{lightgray}500 & \cellcolor{lightgray}HuBERT Large \cite{HuBERT-HsuBTLSM21} & \cellcolor{lightgray}Acc \\

\addlinespace[4pt]
& 3 & Speech Event Extraction\newline (SpeechEE) & General & Content Recognition, Reasoning Ability & CASIE \cite{SpeechEE-WangZ0ZLW0Z24} & 500 & E2E (T5) \cite{SpeechEE-WangZ0ZLW0Z24} & F1 \\

\addlinespace[4pt]
\hline

\addlinespace[4pt]
\#A-C-3 \newline Speech Emotion\newline Understanding\newline (SpeechEmo Analy)
& 1 & Speech Emotion Recognition \newline (Emotion Recog) & General & Affective Analysis & IEMOCAP \cite{IEMOCAP-BussoBLKMKCLN08} & 500 & WavLM Large \cite{WavLM-ChenWCWLCLKYXWZ22} & Acc \\

\addlinespace[4pt]
\hline

\addlinespace[4pt]
\multirow{10}{*}{\shortstack[l]{\#A-C-4 \\ Music\\ Understanding\\  (Music Analy)}}
& 1 & Music Genre Classification\newline (Music-Genre Recog) & Art & Content Recognition & Music Genre \cite{Music-Genre-GTZAN} & 500 & Musicset-Sup \cite{Musicset-Sup-McCallumKOGE22} & Acc \\

\addlinespace[4pt]
& \cellcolor{lightgray}2 & \cellcolor{lightgray}Music Instrument Classification\newline (Instrument Recog) & \cellcolor{lightgray}Art & \cellcolor{lightgray}Content Recognition, Commonsense Knowledge & \cellcolor{lightgray}NS. Instruments \cite{NSynth-nsynth2017} & \cellcolor{lightgray}500 & \cellcolor{lightgray}HuBERT-base \cite{HuBERT-HsuBTLSM21} & \cellcolor{lightgray}Acc \\

\addlinespace[4pt]
& 3 & Music Instrument Source Analysis\newline (Instrument-Source Anal) & Art & Content Recognition & NS. Instruments Source \cite{NSynth-nsynth2017} & 500 & Musicset-ULarge \cite{Musicset-Sup-McCallumKOGE22} & Acc \\

\addlinespace[4pt]
& \cellcolor{lightgray}4 & \cellcolor{lightgray}Music Pitch Analysis\newline (Pitch Anal) & \cellcolor{lightgray}Art & \cellcolor{lightgray}Content Recognition & \cellcolor{lightgray}NS. Pitch \cite{NSynth-nsynth2017} & \cellcolor{lightgray}500 & \cellcolor{lightgray}Musicset-ULarge \cite{Musicset-Sup-McCallumKOGE22} & \cellcolor{lightgray}Acc \\

\addlinespace[4pt]
\hline

\addlinespace[4pt]
\multirow{6}{*}{\shortstack[l]{\#A-C-5 \\ Audio Technique\\ Understanding \\ (Aud-Tech Analy)}}
& 1 & Note Qualities Analysis\newline (Note-Quality Anal) & Art & Content Recognition & NS. quality \cite{NSynth-nsynth2017} & 500 & HuBERT-base \cite{HuBERT-HsuBTLSM21} & Acc \\

\addlinespace[4pt]
& \cellcolor{lightgray}2 & \cellcolor{lightgray}Singer Identification\newline (Singer ID) & \cellcolor{lightgray}Art & \cellcolor{lightgray}Reasoning Ability & \cellcolor{lightgray}VocalSet \cite{VocalSet-WilkinsSWP18} & \cellcolor{lightgray}500 & \cellcolor{lightgray}HuBERT-base \cite{HuBERT-HsuBTLSM21} & \cellcolor{lightgray}Acc \\

\addlinespace[4pt]
& 3 & Vocal Technique Detection\newline (Vocal-Tech Detect) & Art & Reasoning Ability & VocalSet \cite{VocalSet-WilkinsSWP18} & 500 & HuBERT-base \cite{HuBERT-HsuBTLSM21} & Acc \\

\addlinespace[4pt]
\hline

\addlinespace[4pt]
\multirow{1}{*}{\shortstack[l]{\#A-C-6 \\ Audio Content\\ Understanding\\  (Aud Analy)}}
& 1 & Long Audio Captioning\newline (Long AudioCaps) \newline \color{white}{newline} & General & Reasoning Ability & Clotho Caption \cite{Clotho-cap-DrossosLV20} & 500 & PTAAC \cite{PTAAC-KimSO23} & BLUE-1 \\

\addlinespace[4pt]
& \cellcolor{lightgray}2 & \cellcolor{lightgray}Wild Audio Captioning\newline (Wild AudioCaps) & \cellcolor{lightgray}General & \cellcolor{lightgray}Reasoning Ability & \cellcolor{lightgray}AudioCaps \cite{AudioCaps-KimKLK19} & \cellcolor{lightgray}500 & \cellcolor{lightgray}PTAAC \cite{PTAAC-KimSO23} & \cellcolor{lightgray}BLUE-1 \\

\addlinespace[4pt]
\hline

\addlinespace[4pt]
\multirow{6}{*}{\shortstack[l]{\#A-C-7 \\ General Audio\\ Question\\ Answering \\ (Aud QA)}}
& 1 & Open Audio Question Answering\newline (OpenAQA)\newline \color{white}{1}\newline \color{white}{1}\newline \color{white}{1}\newline \color{white}{1} & General & Reasoning Ability, Commonsense Knowledge & OpenAQA (LTU) \cite{LTU-0001LLKG24} & 500 & LTU \cite{LTU-0001LLKG24} & GPT-Score \\

\addlinespace[4pt]
& \cellcolor{lightgray}2 & \cellcolor{lightgray}Audio Question Answering\newline (AudioQA) & \cellcolor{lightgray}General & \cellcolor{lightgray}Reasoning Ability, Commonsense Knowledge & \cellcolor{lightgray}ClothoAQA \cite{Clotho-AQA-LippingSDV22} & \cellcolor{lightgray}500 & \cellcolor{lightgray}MWAFM \cite{MWAFM-LiX023} & \cellcolor{lightgray}Acc \\

\addlinespace[4pt]
\hline

\addlinespace[4pt]
\multirow{8}{*}{\shortstack[l]{\#A-C-8 \\ Animal Sound\\ Analysis \\ (Animal-Sound Det)}}
& 1 & Bird Sound Detection\newline (Bird-Sound Detect) & Biology, Animal & Content Recognition, Commonsense Knowledge & Birdsong \cite{Bird-Sound-Detection-abs-1807-05812} & 500 & HuBERT Large \cite{HuBERT-HsuBTLSM21} & Acc \\

\addlinespace[4pt]
& \cellcolor{lightgray}2 & \cellcolor{lightgray}Animal Sound Detection \newline(AnimalSoundDetect) & \cellcolor{lightgray}Biology, Animal & \cellcolor{lightgray}Content Recognition, Commonsense Knowledge & \cellcolor{lightgray}Animal-Sound Classification \cite{Animal-Sound-Detection} & \cellcolor{lightgray}500 & \cellcolor{lightgray}HuBERT Large \cite{HuBERT-HsuBTLSM21} & \cellcolor{lightgray}Acc \\

\addlinespace[4pt]
\hline

\addlinespace[4pt]
\multirow{1}{*}{\shortstack[l]{\#A-C-9 \\ Environment Sound\\ Understanding \\ (Envir-Sound Det)}}
& 1 & Acoustic Scene Recognition\newline (Acoustic-Scene Recog)\newline \color{white}{1}\newline \color{white}{1} & General & Reasoning Ability & TUT 2017 \cite{TUT17-MesarosHV16} & 468 & CLAP \cite{CLAP-ElizaldeDIW23} & Acc \\

\addlinespace[4pt]
& \cellcolor{lightgray}2 & \cellcolor{lightgray}Environment Sound Recognition\newline (Env-Sound Recog) & \cellcolor{lightgray}General & \cellcolor{lightgray}Reasoning Ability, Commonsense Knowledge & \cellcolor{lightgray}ESC50 \cite{ESC50-Piczak15} & \cellcolor{lightgray}500 & \cellcolor{lightgray}CLAP \cite{CLAP-ElizaldeDIW23} & \cellcolor{lightgray}Acc \\

\addlinespace[4pt]
& 3 & Sound Event Recognition\newline (Sound-Event Recog) & General & Reasoning Ability & ESC50 \cite{ESC50-Piczak15} & 500 & CLAP \cite{CLAP-ElizaldeDIW23} & Acc \\

\addlinespace[4pt]
\hline

\end{longtable}
}

{\centering
\fontsize{8}{8}\selectfont
\setlength{\tabcolsep}{0.7mm}
\renewcommand{\arraystretch}{0.9} 
\captionof{table}{
Detailed list of all tasks and skills (meta-tasks) under audio and generation category.
}
\vspace{-3mm}
\label{tab:task-specification-aud-gen}
\begin{longtable}[!t]{p{2.2cm}%
                >{\centering\arraybackslash}p{0.2cm}
                p{4cm}%
                >{\centering\arraybackslash}p{1.3cm}%
                p{2.3cm}%
                p{2cm}%
                >{\centering\arraybackslash}p{0.7cm}%
                p{2cm}%
                >{\centering\arraybackslash}p{1.3cm}}

\addlinespace[3pt]
\rowcolor{bg-tb-heavey-audio} \multicolumn{9}{c}{\bf \textcolor{blueGen}{\large Audio Generation Group}} \\ [4pt]

\hline

\addlinespace[4pt]
\multicolumn{5}{c}{\bf Task} & \multicolumn{2}{c}{\bf Data}  & \multicolumn{2}{c}{\bf \textbf{SoTA Specialist}} \\

\cmidrule(r){1-5}  \cmidrule(r){6-7}  \cmidrule(r){8-9} 

\textbf{Skill (Meta-Task)} & \textbf{\#}& \textbf{Task (Short Name)} &  \textbf{Domain} & \textbf{Capability} & \textbf{Data Source} & \textbf{Number} & \multicolumn{1}{c}{\textbf{Model}}  & \textbf{Metrics} \\ 

\addlinespace[4pt]
\hline
\endfirsthead
\hline
\addlinespace[4pt]
\multicolumn{5}{c}{\bf Task} & \multicolumn{2}{c}{\bf Data}  & \multicolumn{2}{c}{\bf \textbf{SoTA Specialist}} \\
\cmidrule(r){1-5}  \cmidrule(r){6-7}  \cmidrule(r){8-9} 
\textbf{Skill (Meta-Task)} & \textbf{\#}& \textbf{Task (Short Name)} &  \textbf{Domain} & \textbf{Capability} & \textbf{Data Source} & \textbf{Number} & \multicolumn{1}{c}{\textbf{Model}}  & \textbf{Metrics} \\ 
\addlinespace[4pt]
\hline
\endhead
\hline
\endfoot

\addlinespace[4pt]
\multirow{1}{*}{\shortstack[l]{\#A-G-1 \\ Audio Edit \\ (Audio Edit)}}
& 1 & AudioEdit (AudioEdit) & Art & Creativity and Innovation & Song Describer Dataset \cite{Song-Describer-Dataset-manco2023thesong} & 16 & ControlNet-DiffuTrans \cite{Control-diff-abs-2410-05151} & CLAP \\

\addlinespace[4pt]
\hline

\addlinespace[4pt]
\#A-G-2 \newline Dialogue Speech\newline Generation \newline (Dialog Gen)
& 1 & Daily Talk Generation\newline (DailyTalk Gen) & General & Creativity and Innovation & DailyTalk \cite{DailyTalk-LeePK23} & 500 & FastSpeech2 \cite{FastSpeech-0006H0QZZL21} & MOS \\

\addlinespace[4pt]
\hline

\addlinespace[4pt]
\multirow{2}{*}{\shortstack[l]{\#A-G-3 \\ Emotional Speech\\ Generation \\ (EmoSpeech Gen)}}
& 1 & Emotional Speech Synthesis\newline (EmoSpeech Syn) & General & Affective Analysis & Emotional Speech Data \cite{emotion-speech-zhou2021seen} & 500 & DeepEST \cite{emotion-speech-zhou2021seen} & MCD \\

\addlinespace[4pt]
& \cellcolor{lightgray}2 & \cellcolor{lightgray}Emotion Style Transfer\newline (EmoStyleTransfer) & \cellcolor{lightgray}General & \cellcolor{lightgray}Affective Analysis & \cellcolor{lightgray}EmotionalSpeechDataset \cite{emotion-speech-zhou2021seen} & \cellcolor{lightgray}500 & \cellcolor{lightgray}DeepEST \cite{emotion-speech-zhou2021seen} & \cellcolor{lightgray}MOS \\

\addlinespace[4pt]
\hline

\addlinespace[4pt]
\multirow{6}{*}{\shortstack[l]{\#A-G-4 \\ Text-To-Speech\\ Synthesis \\ (TTS)}}
& 1 & Text-To-Speech (TTS) & Linguistics & Commonsense Knowledge, Creativity and Innovation & Librispeech test-clean \cite{Librispeech-PanayotovCPK15} & 500 & USLM \cite{SpeechTokenizer-zhang2023} & WER  \\

\addlinespace[4pt]
& \cellcolor{lightgray}2 & \cellcolor{lightgray}Multimodal TTS\newline (MTTS) & \cellcolor{lightgray}General & \cellcolor{lightgray}Creativity and Innovation & \cellcolor{lightgray}MEAD-TTS \cite{MM-TTS-GuanLLHWLHLH24} & \cellcolor{lightgray}500 & \cellcolor{lightgray}MM-TTS \cite{MM-TTS-GuanLLHWLHLH24} & \cellcolor{lightgray}MOS \\

\addlinespace[4pt]
\hline

\addlinespace[4pt]
\multirow{4}{*}{\shortstack[l]{\#A-G-5 \\ Text-to-Audio\\ Synthesis\newline (Txt2Aud Gen)}}
& 1 & Single Caption To Audio Generation\newline (1CapToAudio) & General & Creativity and Innovation & AudioCap \cite{AudioCaps-KimKLK19} & 500 & AudioLDM2 \cite{AudioLDM2-LiuYLMKTWWWP24} & CLAP \\

\addlinespace[4pt]
& \cellcolor{lightgray}2 & \cellcolor{lightgray}Two Captions To Audio Generation\newline (2CapsToAudio) & \cellcolor{lightgray}General & \cellcolor{lightgray}Creativity and Innovation & \cellcolor{lightgray}AudioCap \cite{AudioCaps-KimKLK19} & \cellcolor{lightgray}500 & \cellcolor{lightgray}AudioLDM2 \cite{AudioLDM2-LiuYLMKTWWWP24} & \cellcolor{lightgray}CLAP \\

\addlinespace[4pt]
\hline

\addlinespace[4pt]
\#A-G-6\newline Image-to-Audio\newline Synthesis (Img2Aud Gen)
& 1 & Image-to-Speech\newline (ImageToSpeech) & General & Content Recognition & Flickr8k Audio \cite{Flicker8k} & 500 & Im2Sp \cite{Img2Sp} & CIDEr \\

\addlinespace[4pt]
\hline

\addlinespace[4pt]
\multirow{1}{*}{\shortstack[l]{\#A-G-7 \\ Video-to-Audio\\ Synthesis  (V2A)}}
& 1 & Video-to-Audio\newline (Video2Audio) & General & Commonsense Knowledge & AVSync15 \cite{AVSync15} & 500 & Diff-Foley \cite{Diff-Foley-luo2023} & FAD \\

\addlinespace[4pt]
\hline

\addlinespace[4pt]
\multirow{1}{*}{\shortstack[l]{\#A-G-8 \\ Speech Style Transfer \\ (Style Trans)}}
& 1 & Voice Conversion\newline (VoiceConversion)\newline \color{white}{newline} & General & Content Recognition & VCTK Corpus \cite{VCTK-Corpus} & 500 & USLM \cite{SpeechTokenizer-zhang2023} & WER \\

\addlinespace[4pt]
\hline

\addlinespace[4pt]
\multirow{8}{*}{\shortstack[l]{\#A-G-9 \\ Speech Translation \\ (Speech Trans)}}
& 1 & Chinese-to-English Speech Translation (SpeechTrans (Zh-En)) & Multi-domain & Content Recognition & GigaST \cite{GigaST-ye2023} & 500 & USLM \cite{SpeechTokenizer-zhang2023} & WER \\

\addlinespace[4pt]
& \cellcolor{lightgray}2 & \cellcolor{lightgray}English-to-Chinese Speech Translation (SpeechTrans (En-Zh)) & \cellcolor{lightgray}General & \cellcolor{lightgray}Content Recognition & \cellcolor{lightgray}CoVoST v1 \cite{CoVoST-wang-etal-2020-covost} & \cellcolor{lightgray}500 & \cellcolor{lightgray}USLM \cite{SpeechTokenizer-zhang2023} & \cellcolor{lightgray}WER\\

\addlinespace[4pt]
& 3 & English-to-Mogolian Speech Translation (SpeechTrans (En-Mo)) & General & Content Recognition & CoVoST v1 \cite{CoVoST-wang-etal-2020-covost} & 500 & USLM \cite{SpeechTokenizer-zhang2023} & WER  \\

\addlinespace[4pt]
\hline

\addlinespace[4pt]
\multirow{4}{*}{\shortstack[l]{\#A-G-10 \\ Music Synthesis \\ (Music Gen)}}
& 1 & Piano-to-Orchestration Generation\newline (Piano2Orchestra Gen) & Art & Creativity and Innovation & LOP datasets \cite{crestel2017liveorchestralpianorealtime} & 500 & FGcRBM \cite{crestel2017liveorchestralpianorealtime} & Acc \\

\addlinespace[4pt]
& \cellcolor{lightgray}2 & \cellcolor{lightgray}Pop Music Generation\newline (PopMusic Gen) & \cellcolor{lightgray}Art & \cellcolor{lightgray}Creativity and Innovation & \cellcolor{lightgray}POP909 \cite{pop909-ismir2020} & \cellcolor{lightgray}500 & \cellcolor{lightgray}ThemeTransformer \cite{Theme-Transformer-ShihWZMY23} & \cellcolor{lightgray}PCC \\

\addlinespace[4pt]
& 3 & Singing Voice Synthesis\newline (Singing Voice Syn) & Art & Commonsense Knowledge, Creativity and Innovation & M4Singer \cite{msinger-zhang2022} & 500 & Diffsinger \cite{DiffSinger-Liu00CZ22} & MOS \\

\addlinespace[4pt]
& \cellcolor{lightgray}4 & \cellcolor{lightgray}Song Generation\newline (Song Gen) & \cellcolor{lightgray}Art & \cellcolor{lightgray}Creativity and Innovation & \cellcolor{lightgray}Song Describer Dataset \cite{Song-Describer-Dataset-manco2023thesong} & \cellcolor{lightgray}500 & \cellcolor{lightgray}Riffusion \cite{Riffusion-Forsgren_Martiros_2022} & \cellcolor{lightgray}FAD \\

\addlinespace[4pt]
\hline

\addlinespace[4pt]
\multirow{8}{*}{\shortstack[l]{\#A-G-11 \\ Music Style Transfer \\ (Music Trans)}}
& 1 & Music Style Transfer & Art & Commonsense Knowledge, Creativity and Innovation & MusicTI \cite{MusicIT-LiZTMDX24} & 500 & TVI-Diff \cite{MusicIT-LiZTMDX24} & Style-CLAP \\

\addlinespace[4pt]
& \cellcolor{lightgray}2 & \cellcolor{lightgray}Chord-based Music Style Transfer\newline (Chord Music Style Trans) & \cellcolor{lightgray}Art & \cellcolor{lightgray}Commonsense Knowledge, Creativity and Innovation & \cellcolor{lightgray}Groove2Groove \cite{Groove2Groove-CifkaSR20} & \cellcolor{lightgray}500 & \cellcolor{lightgray}TVI-Diff \cite{MusicIT-LiZTMDX24} & \cellcolor{lightgray}Style-CLAP \\

\addlinespace[4pt]
\hline

\end{longtable}
}

\clearpage

{\centering
\fontsize{8}{8}\selectfont
\setlength{\tabcolsep}{0.7mm}
\renewcommand{\arraystretch}{0.9} 
\captionof{table}{
Detailed list of all tasks and skills (meta-tasks) under 3D and comprehension category.
}
\vspace{-3mm}
\label{tab:task-specification-3d-comp}
\begin{longtable}[!t]{p{2.3cm}%
                >{\centering\arraybackslash}p{0.2cm}
                p{4cm}%
                >{\centering\arraybackslash}p{1.3cm}%
                p{2.3cm}%
                p{2cm}%
                >{\centering\arraybackslash}p{0.7cm}%
                p{2cm}%
                >{\centering\arraybackslash}p{1.3cm}}

\addlinespace[3pt]
\rowcolor{bg-tb-heavey-3D} \multicolumn{9}{c}{\bf \textcolor{greenCom}{\large 3D Comprehension Group}} \\ [4pt]

\hline

\addlinespace[4pt]
\multicolumn{5}{c}{\bf Task} & \multicolumn{2}{c}{\bf Data}  & \multicolumn{2}{c}{\bf \textbf{SoTA Specialist}} \\

\cmidrule(r){1-5}  \cmidrule(r){6-7}  \cmidrule(r){8-9} 

\textbf{Skill (Meta-Task)} & \textbf{\#}& \textbf{Task (Short Name)} &  \textbf{Domain} & \textbf{Capability} & \textbf{Data Source} & \textbf{Number} & \multicolumn{1}{c}{\textbf{Model}}  & \textbf{Metrics} \\ 

\addlinespace[4pt]
\hline
\endfirsthead
\hline
\addlinespace[4pt]
\multicolumn{5}{c}{\bf Task} & \multicolumn{2}{c}{\bf Data}  & \multicolumn{2}{c}{\bf \textbf{SoTA Specialist}} \\
\cmidrule(r){1-5}  \cmidrule(r){6-7}  \cmidrule(r){8-9} 
\textbf{Skill (Meta-Task)} & \textbf{\#}& \textbf{Task (Short Name)} &  \textbf{Domain} & \textbf{Capability} & \textbf{Data Source} & \textbf{Number} & \multicolumn{1}{c}{\textbf{Model}}  & \textbf{Metrics} \\ 
\addlinespace[4pt]
\hline
\endhead
\hline
\endfoot

\addlinespace[4pt]
\multirow{10}{*}{\shortstack[l]{\#D-C-1 \\ 3D Human-related \\ Object Classification \\ (3D-Human Cls)}}
& 1 & 3D Accessory Classification\newline (3D-Access Cls) & General & Content Recognition & ModelNet \cite{WuSKYZTX15} & 340 & PointGST \cite{liang2024pointgst} & Acc \\

\addlinespace[4pt]
& \cellcolor{lightgray}2 & \cellcolor{lightgray}3D Appliance Classification\newline (3D-Appliance Cls) & \cellcolor{lightgray}General & \cellcolor{lightgray}Content Recognition & \cellcolor{lightgray}ModelNet \cite{WuSKYZTX15} & \cellcolor{lightgray}240 & \cellcolor{lightgray}PointGST \cite{liang2024pointgst} & \cellcolor{lightgray}Acc \\

\addlinespace[4pt]
& 3 & 3D Tableware Classification\newline (3D-Tableware Cls) & General & Content Recognition & ModelNet \cite{WuSKYZTX15} & 180 & PointGST \cite{liang2024pointgst} & Acc \\

\addlinespace[4pt]
& \cellcolor{lightgray}4 & \cellcolor{lightgray}3D Musical Instrument Classification (3D-MI Cls) & \cellcolor{lightgray}General & \cellcolor{lightgray}Content Recognition & \cellcolor{lightgray}ModelNet \cite{WuSKYZTX15} & \cellcolor{lightgray}500 & \cellcolor{lightgray}PointGST \cite{liang2024pointgst} & \cellcolor{lightgray}Acc \\

\addlinespace[4pt]
& 5 & 3D Person Classification\newline (3D-Person Cls) & General & Content Recognition & ModelNet \cite{WuSKYZTX15} & 200 & PointGST \cite{liang2024pointgst} & Acc \\

\addlinespace[4pt]
\hline

\addlinespace[4pt]
\multirow{5}{*}{\shortstack[l]{\#D-C-2 \\ 3D Structure\\ and Environment\\ Classification \\ (3D-Struct Cls)}}
& 1 & 3D Furniture Classification\newline (3D-Furniture ClS)\newline \color{white}{1} & General & Content Recognition & ModelNet \cite{WuSKYZTX15} & 20 & PointGST \cite{liang2024pointgst} & Acc \\

\addlinespace[4pt]
& \cellcolor{lightgray}2 & \cellcolor{lightgray}3D Structure Classification\newline (3D-Struct Cls) & \cellcolor{lightgray}General & \cellcolor{lightgray}Content Recognition & \cellcolor{lightgray}ModelNet \cite{WuSKYZTX15} & \cellcolor{lightgray}40 & \cellcolor{lightgray}PointGST \cite{liang2024pointgst} & \cellcolor{lightgray}Acc \\

\addlinespace[4pt]
\hline

\addlinespace[4pt]
\multirow{5}{*}{\shortstack[l]{\#D-C-3 \\ Transportation and \\ Technology Object\\ Classification \\ (Tech Cls)}}
& 1 & 3D Electronic Classification\newline (3D-Electronic ClS)\newline \color{white}{1} & General & Content Recognition & ModelNet \cite{WuSKYZTX15} & 140 & PointGST \cite{liang2024pointgst} & Acc \\

\addlinespace[4pt]
& \cellcolor{lightgray}2 & \cellcolor{lightgray}3D Vehicle Classification (3D-Vehicle Cls) & \cellcolor{lightgray}General & \cellcolor{lightgray}Content Recognition & \cellcolor{lightgray}ModelNet \cite{WuSKYZTX15} & \cellcolor{lightgray}200 & \cellcolor{lightgray}PointGST \cite{liang2024pointgst} & \cellcolor{lightgray}Acc \\

\addlinespace[4pt]
\hline

\addlinespace[4pt]
\multirow{1}{*}{\shortstack[l]{\#D-C-4 \\ 3D Indoor Scene \\ Semantic\\ Segmentation \\ (Indoor-Scene Seg)}}
& 1 & 3D Indoor Appliance Semantic Segmentation (3D-Appliance Seg) \newline \color{white}{1}\newline \color{white}{1}\newline \color{white}{1}& General & Commonsense Knowledge & ScanNet \cite{DaiCSHFN17} & 142 & ODIN \cite{jain2024odin} & mIoU \\

\addlinespace[4pt]
\hline

\addlinespace[4pt]
\multirow{1}{*}{\shortstack[l]{\#D-C-5 \\ 3D Outdoor Scene \\ Semantic\\ Segmentation \\ (Outdoor-Scene Seg)}}
& 1 & 3D Outdoor Semantic Segmentation (3D-Outdr Seg) \newline \color{white}{1}\newline \color{white}{1}\newline \color{white}{1}& General & Commonsense Knowledge & Semantic KITTI \cite{BehleyGMQBSG19} & 4,071 & OpenPCSeg \cite{openpcdet2020} & mIoU \\

\addlinespace[4pt]
\hline

\addlinespace[4pt]
\#D-C-6 \newline 3D Indoor Scene \newline Instance Segmentation (Indoor-Inst Seg)
& 1 & 3D Indoor Instance Segmentation\newline (3D-In-Instance Seg) & General & Commonsense Knowledge & ScanNet \cite{DaiCSHFN17} & 142 & SphericalMask \cite{shin2024spherical} & mIoU \\

\addlinespace[4pt]
\hline

\addlinespace[4pt]
\#D-C-7 \newline 3D Pose Estimation \newline (Pose Est)
& 1 & 3D Odometry (3D Odometry) & Geometry & Problem Solving & KITTI \cite{GeigerLU12} & 10 & CT-ICP \cite{dellenbach2021cticp} & RTE  \\

\addlinespace[4pt]
\hline

\addlinespace[4pt]
\multirow{8}{*}{\shortstack[l]{\#D-C-8 \\ 3D Part\\ Segmentation \\ (Part Seg)}}
& 1 & 3D Aircrafts Part Segmentation (3D-Aircraft Seg) & General & Commonsense Knowledge & Shape Net Part \cite{ChangFGHHLSSSSX15} & 523 & SPOTR \cite{ParkLKXK23} & Instance mIoU \\

\addlinespace[4pt]
& \cellcolor{lightgray}2 & \cellcolor{lightgray}3D Personal Item Part Segmentation (3D-Person Seg) & \cellcolor{lightgray}General & \cellcolor{lightgray}Commonsense Knowledge & \cellcolor{lightgray}Shape Net Part \cite{ChangFGHHLSSSSX15} & \cellcolor{lightgray}346 & \cellcolor{lightgray}SPOTR \cite{ParkLKXK23} & \cellcolor{lightgray}Instance mIoU \\

\addlinespace[4pt]
& 3 & 3D Vehicle Part Segmentation \newline (3D-Vehicle Seg) & General & Commonsense Knowledge & Shape Net Part \cite{ChangFGHHLSSSSX15} & 288 & SPOTR \cite{ParkLKXK23} & Instance mIoU \\

\addlinespace[4pt]
& \cellcolor{lightgray}4 & \cellcolor{lightgray}3D Furniture Part Segmentation	 (3D-Furniture Seg) & \cellcolor{lightgray}General & \cellcolor{lightgray}Commonsense Knowledge & \cellcolor{lightgray}Shape Net Part \cite{ChangFGHHLSSSSX15} & \cellcolor{lightgray}2,128 & \cellcolor{lightgray}SPOTR \cite{ParkLKXK23} & \cellcolor{lightgray}Instance mIoU \\

\addlinespace[4pt]
& 5 & 3D Tableware Part Segmentation \newline (3D-Tableware Seg) & General & Commonsense Knowledge & Shape Net Part \cite{ChangFGHHLSSSSX15} & 377 & SPOTR \cite{ParkLKXK23} & Instance mIoU \\

\addlinespace[4pt]
& \cellcolor{lightgray}6 & \cellcolor{lightgray}3D Weapon Part Segmentation \newline (3D-Weapon Seg) & \cellcolor{lightgray}General & \cellcolor{lightgray}Commonsense Knowledge & \cellcolor{lightgray}Shape Net Part \cite{ChangFGHHLSSSSX15} & \cellcolor{lightgray}133 & \cellcolor{lightgray}SPOTR \cite{ParkLKXK23} & \cellcolor{lightgray}Instance mIoU \\

\addlinespace[4pt]
\hline

\addlinespace[4pt]
\multirow{4}{*}{\shortstack[l]{\#D-C-9 \\ 3D Tracking \\ (3D Track)}}
& 1 & 3D Tracking \newline (3D Track)  \newline \color{white}{1} \newline \color{white}{1} & General & Commonsense Knowledge & NuScenes \cite{nuScenes-CaesarBLVLXKPBB20} & 500 & CenterPoint \cite{YinZK21} & AMOTA \\

\addlinespace[4pt]
\hline

\addlinespace[4pt]
\#D-C-10  \newline 3D Geometry Feature Analysis  \newline (3D-Geo Analy)
& 1 & 3D Normal Estimation \newline (3D-Normal Est) & Geometry & Problem Solving & PCPNet dataset \cite{GuerreroEtAl-PCPNet} & 108 & SHS-Net \cite{LiFSGFLH23} & RMSE \\

\addlinespace[4pt]
\hline

\addlinespace[4pt]
\multirow{1}{*}{\shortstack[l]{\#D-C-11 \\ 3D Detection\\ (3D Det)}}
& 1 & 3D Detection \newline (3D Det) \newline \color{white}{1} & General & Content Recognition & NuScenes \cite{nuScenes-CaesarBLVLXKPBB20} & 500 & BEVFusion \cite{LiuTAYMRH23} & mAP \\

\addlinespace[4pt]
\hline

\addlinespace[4pt]
\multirow{2}{*}{\shortstack[l]{\#D-C-12 \\ 3D Question \\ Answering \\ (3D QA)}}
& 1 & 3D Spatial Scene Question Answering \newline (3D-Space QA) & General & Commonsense Knowledge & ScanQA \cite{AzumaMKK22} & 4,675 & SIG3D \cite{man2024situational} & BLEU@4 \\

\addlinespace[4pt]
& \cellcolor{lightgray}2 & \cellcolor{lightgray}3D Situated Question Answering on "What" (3D-"What" QA) & \cellcolor{lightgray}General & \cellcolor{lightgray}Commonsense Knowledge & \cellcolor{lightgray}SQA3D \cite{MaYZ0LZH23} & \cellcolor{lightgray}1,147 & \cellcolor{lightgray}SIG3D \cite{man2024situational} & \cellcolor{lightgray}EM@1 \\

\addlinespace[4pt]
& 3 & 3D Situated Question Answering on "Is" (3D-"Is" QA) & General & Commonsense Knowledge & ScanQA \cite{AzumaMKK22} & 652 & SIG3D \cite{man2024situational} & EM@1 \\

\addlinespace[4pt]
& \cellcolor{lightgray}4 & \cellcolor{lightgray}3D Situated Question Answering on "How" (3D-"How" QA) & \cellcolor{lightgray}General & \cellcolor{lightgray}Commonsense Knowledge & \cellcolor{lightgray}SQA3D \cite{MaYZ0LZH23} & \cellcolor{lightgray}465 & \cellcolor{lightgray}SIG3D \cite{man2024situational} & \cellcolor{lightgray}EM@1 \\

\addlinespace[4pt]
& 5 & 3D Situated Question Answering on "Can" (3D-"Can" QA) & General & Commonsense Knowledge & ScanQA \cite{AzumaMKK22} & 338 & SIG3D \cite{man2024situational} & EM@1 \\

\addlinespace[4pt]
& \cellcolor{lightgray}6 & \cellcolor{lightgray}3D Situated Question Answering on "Which" (3D-"Which" QA) & \cellcolor{lightgray}General & \cellcolor{lightgray}Commonsense Knowledge & \cellcolor{lightgray}SQA3D \cite{MaYZ0LZH23} & \cellcolor{lightgray}351 & \cellcolor{lightgray}SIG3D \cite{man2024situational} & \cellcolor{lightgray}EM@1 \\

\addlinespace[4pt]
& 7 & 3D Situated Question Answering on "Other" (3D-"Other" QA) & General & Commonsense Knowledge & ScanQA \cite{AzumaMKK22} & 566 & SIG3D \cite{man2024situational} & EM@1 \\

\addlinespace[4pt]
\hline

\addlinespace[4pt]
\#D-C-13  \newline 3D Motion  \newline Understanding  \newline (3D-Motion Analy)
& 1 & 3D Motion Captioning \newline (3D-Motion Cap) & General & Commonsense Knowledge & KIT-ML \cite{PlappertMA16} & 4,383 & M2T-Interpretable \cite{radouane2024guided} & BLEU-4 \\

\addlinespace[4pt]
\hline

\end{longtable}
}

{\centering
\fontsize{8}{8}\selectfont
\setlength{\tabcolsep}{0.7mm}
\renewcommand{\arraystretch}{0.9} 
\captionof{table}{
Detailed list of all tasks and skills (meta-tasks) under 3D and generation category.
}
\vspace{-3mm}
\label{tab:task-specification-3d-gen}
\begin{longtable}[!t]{p{2.3cm}%
                >{\centering\arraybackslash}p{0.2cm}
                p{4cm}%
                >{\centering\arraybackslash}p{1.3cm}%
                p{2.3cm}%
                p{2cm}%
                >{\centering\arraybackslash}p{0.7cm}%
                p{2cm}%
                >{\centering\arraybackslash}p{1.3cm}}

\addlinespace[3pt]
\rowcolor{bg-tb-heavey-3D} \multicolumn{9}{c}{\bf \textcolor{blueGen}{\large 3D Generation Group}} \\ [4pt]

\hline

\addlinespace[4pt]
\multicolumn{5}{c}{\bf Task} & \multicolumn{2}{c}{\bf Data}  & \multicolumn{2}{c}{\bf \textbf{SoTA Specialist}} \\

\cmidrule(r){1-5}  \cmidrule(r){6-7}  \cmidrule(r){8-9} 

\textbf{Skill (Meta-Task)} & \textbf{\#}& \textbf{Task (Short Name)} &  \textbf{Domain} & \textbf{Capability} & \textbf{Data Source} & \textbf{Number} & \multicolumn{1}{c}{\textbf{Model}}  & \textbf{Metrics} \\ 

\addlinespace[4pt]
\hline
\endfirsthead
\hline
\addlinespace[4pt]
\multicolumn{5}{c}{\bf Task} & \multicolumn{2}{c}{\bf Data}  & \multicolumn{2}{c}{\bf \textbf{SoTA Specialist}} \\
\cmidrule(r){1-5}  \cmidrule(r){6-7}  \cmidrule(r){8-9} 
\textbf{Skill (Meta-Task)} & \textbf{\#}& \textbf{Task (Short Name)} &  \textbf{Domain} & \textbf{Capability} & \textbf{Data Source} & \textbf{Number} & \multicolumn{1}{c}{\textbf{Model}}  & \textbf{Metrics} \\ 
\addlinespace[4pt]
\hline
\endhead
\hline
\endfoot

\addlinespace[4pt]
\#D-G-1 \newline Point Cloud \newline Completion \newline (PC Complt)
& 1 & 3D Point Cloud Completion\newline (3D-PC Completion) & General & Creativity and Innovation & PCN data \cite{yuan2018pcn} & 500 & PointTr \cite{YuRWLL021} & CD \\

\addlinespace[4pt]
\hline

\addlinespace[4pt]
\multirow{4}{*}{\shortstack[l]{
\#D-G-2 \\ Point Cloud to Mesh \\ Reconstruction \\ (PC2M Recon)
}}
& 1 & Point Cloud to Mesh Object Reconstruction (Object-PC2M Recon) & General & Creativity and Innovation & ShapeNet \cite{ChangFGHHLSSSSX15} & 500 & ConvONet \cite{Peng2020ECCV} & CD \\

\addlinespace[4pt]
& \cellcolor{lightgray}2 & \cellcolor{lightgray}Point Cloud to Mesh Scene Reconstruction (Scene-PC2M Recon) & \cellcolor{lightgray}General & \cellcolor{lightgray}Creativity and Innovation & \cellcolor{lightgray}Synthetic Indoor Scene Dataset \cite{Peng2020ECCV} & \cellcolor{lightgray}500 & \cellcolor{lightgray}ConvONet \cite{Peng2020ECCV} & \cellcolor{lightgray}CD \\

\addlinespace[4pt]
\hline

\addlinespace[4pt]
\multirow{10}{*}{\shortstack[l]{\#D-G-3 \\ Text to Point \\ Cloud Generation \\ (Txt2PC Gen)}}
& 1 & Text to 3D Living and Arts Point Cloud Generation (Art-Txt2PC Gen) & Art & Creativity and Innovation & Manual & 500 & Point-E \cite{Point-E-abs-2212-08751} & CLIP-Score \\

\addlinespace[4pt]
& \cellcolor{lightgray}2 & \cellcolor{lightgray}Text to 3D Science and Technology Point Cloud Generation\newline (Sci-Txt2PC Gen) & \cellcolor{lightgray}General & \cellcolor{lightgray}Creativity and Innovation & \cellcolor{lightgray}Manual & \cellcolor{lightgray}500 & \cellcolor{lightgray}Point-E \cite{Point-E-abs-2212-08751} & \cellcolor{lightgray}CLIP-Score \\

\addlinespace[4pt]
& 3 & Text to 3D Nature and Biology Point Cloud Generation (Nature-Txt2PC Gen) & General & Creativity and Innovation & Manual & 500 & Point-E \cite{Point-E-abs-2212-08751} & CLIP-Score \\

\addlinespace[4pt]
& \cellcolor{lightgray}4 & \cellcolor{lightgray}Text to 3D Culture and Structure Point Cloud Generation\newline (Culture-Txt2PC Gen) & \cellcolor{lightgray}General & \cellcolor{lightgray}Creativity and Innovation & \cellcolor{lightgray}Manual & \cellcolor{lightgray}500 & \cellcolor{lightgray}Point-E \cite{Point-E-abs-2212-08751} & \cellcolor{lightgray}CLIP-Score \\

\addlinespace[4pt]
\hline

\addlinespace[4pt]
\multirow{4}{*}{\shortstack[l]{\#D-G-4 \\ Text to Mesh \\ Generation \\ (Txt2M Gen)}}
& 1 & Text to 3D Living and Arts Mesh Generation (Arts-Txt2M Gen) \newline \color{white}{1} \newline \color{white}{1} & Art & Creativity and Innovation & Manual & 500 & Hunyuan3D-1.0 \cite{yang2024tencent} & CLIP-Score \\

\addlinespace[4pt]
& \cellcolor{lightgray}2 & \cellcolor{lightgray}Text to 3D Science and Technology Mesh Generation (Sci-Txt2M Gen) & \cellcolor{lightgray}General & \cellcolor{lightgray}Creativity and Innovation & \cellcolor{lightgray}Manual & \cellcolor{lightgray}500 & \cellcolor{lightgray}Hunyuan3D-1.0 \cite{yang2024tencent} & \cellcolor{lightgray}CLIP-Score \\

\addlinespace[4pt]
& 3 & Text to 3D Nature and Biology Mesh Generation (Nature-Txt2M Gen) & General & Creativity and Innovation & Manual & 500 & Hunyuan3D-1.0 \cite{yang2024tencent} & CLIP-Score \\

\addlinespace[4pt]
& \cellcolor{lightgray}4 & \cellcolor{lightgray}Text to 3D Culture and Structure Mesh Generation (Culture-Txt2M Gen) & \cellcolor{lightgray}General & \cellcolor{lightgray}Creativity and Innovation & \cellcolor{lightgray}Manual & \cellcolor{lightgray}500 & \cellcolor{lightgray}Hunyuan3D-1.0 \cite{yang2024tencent} & \cellcolor{lightgray}CLIP-Score \\

\addlinespace[4pt]
\hline

\addlinespace[4pt]
\multirow{1}{*}{\shortstack[l]{\#D-G-5 \\ Image to Point \\ Cloud Generation \\ (Img2PC Gen)}}
& 1 & Living and Arts Image to 3D Point Cloud Generation (Arts-Img2PC Gen) \newline \color{white}{1} \newline \color{white}{1} \newline \color{white}{1} & Art & Creativity and Innovation & Objaverse \cite{DeitkeSSWMVSEKF23} & 500 & Point-E \cite{Point-E-abs-2212-08751} & CLIP-Score \\

\addlinespace[4pt]
& \cellcolor{lightgray}2 & \cellcolor{lightgray}Science and Technology Image to 3D Point Cloud Generation\newline (Sci-Img2PC Gen) & \cellcolor{lightgray}General & \cellcolor{lightgray}Creativity and Innovation & \cellcolor{lightgray}Objaverse \cite{DeitkeSSWMVSEKF23} & \cellcolor{lightgray}500 & \cellcolor{lightgray}Point-E \cite{Point-E-abs-2212-08751} & \cellcolor{lightgray}CLIP-Score \\

\addlinespace[4pt]
& 3 & Nature and Biology Image to 3D Point Cloud Generation\newline (Nature-Img2PC Gen) & General & Creativity and Innovation & Objaverse \cite{DeitkeSSWMVSEKF23} & 500 & Point-E \cite{Point-E-abs-2212-08751} & CLIP-Score \\

\addlinespace[4pt]
& \cellcolor{lightgray}4 & \cellcolor{lightgray}Culture and Structure Image to 3D Point Cloud Generation\newline (Culture-Img2PC Gen) & \cellcolor{lightgray}General & \cellcolor{lightgray}Creativity and Innovation & \cellcolor{lightgray}Objaverse \cite{DeitkeSSWMVSEKF23} & \cellcolor{lightgray}500 & \cellcolor{lightgray}Point-E \cite{Point-E-abs-2212-08751} & \cellcolor{lightgray}CLIP-Score \\

\addlinespace[4pt]
\hline

\addlinespace[4pt]
\multirow{8}{*}{\shortstack[l]{\#D-G-6 \\ Image to Mesh \\ Generation \\ (Img2M Gen)}}
& 1 & Living and Arts Image to 3D Mesh Generation (Arts-Img2M  Gen) & Art & Creativity and Innovation & Objaverse \cite{DeitkeSSWMVSEKF23} & 500 & Hunyuan3D-1.0 \cite{yang2024tencent} & CLIP-Score \\

\addlinespace[4pt]
& \cellcolor{lightgray}2 & \cellcolor{lightgray}Science and Technology Image to 3D Mesh Generation (Sci-Img2M  Gen) & \cellcolor{lightgray}General & \cellcolor{lightgray}Creativity and Innovation & \cellcolor{lightgray}Objaverse \cite{DeitkeSSWMVSEKF23} & \cellcolor{lightgray}500 & \cellcolor{lightgray}Hunyuan3D-1.0 \cite{yang2024tencent} & \cellcolor{lightgray}CLIP-Score \\

\addlinespace[4pt]
& 3 & Nature and Biology Image to 3D Mesh Generation (Nature-Img2M  Gen) & General & Creativity and Innovation & Objaverse \cite{DeitkeSSWMVSEKF23} & 500 & Hunyuan3D-1.0 \cite{yang2024tencent} & CLIP-Score \\

\addlinespace[4pt]
& \cellcolor{lightgray}4 & \cellcolor{lightgray}Culture and Structure Image to 3D Mesh Generation (Culture-Img2M Gen) & \cellcolor{lightgray}General & \cellcolor{lightgray}Creativity and Innovation & \cellcolor{lightgray}Objaverse \cite{DeitkeSSWMVSEKF23} & \cellcolor{lightgray}500 & \cellcolor{lightgray}Hunyuan3D-1.0 \cite{yang2024tencent} & \cellcolor{lightgray}CLIP-Score \\

\addlinespace[4pt]
\hline

\addlinespace[4pt]
\#D-G-7 \newline RGB-D to Point \newline Cloud Reconstruction  (RGBD2PC Recon)
& 1 & RGB-D to Point Cloud Reconstruction (RGBD2PC Recon) & General & Creativity and Innovation & ScanNet \cite{DaiCSHFN17} & 142 & open3d \cite{Zhou2018} & CD \\

\addlinespace[4pt]
\hline

\addlinespace[4pt]
\#D-G-8 \newline RGB-D to Mesh \newline Reconstruction \newline (RGBD2Mesh Recon)
& 1 & RGB-D to Mesh Reconstruction (RGBD2Mesh Recon) & General & Creativity and Innovation & ScanNet \cite{DaiCSHFN17} & 142 & open3d \cite{Zhou2018} & CD \\

\addlinespace[4pt]
\hline

\addlinespace[4pt]
\#D-G-9 \newline Text to 3D \newline Motion Generation \newline (Txt2Motion Gen)
& 1 & Text to 3D Motion Generation (Txt2Motion Gen) & General & Creativity and Innovation & KIT-ML \cite{PlappertMA16} & 830 & MoMask \cite{GuoMJW024} & FID \\

\addlinespace[4pt]
\hline

\end{longtable}
}

\clearpage

{\centering
\fontsize{8}{8}\selectfont
\setlength{\tabcolsep}{0.7mm}
\renewcommand{\arraystretch}{0.9} 
\captionof{table}{
Detailed list of all NLP tasks and skills (meta-tasks).
}
\vspace{-3mm}
\label{tab:task-specification-nlp}
\begin{longtable}[!t]{p{2.2cm}%
                >{\centering\arraybackslash}p{0.2cm}
                p{4cm}%
                >{\centering\arraybackslash}p{1.3cm}%
                p{2.3cm}%
                p{2cm}%
                >{\centering\arraybackslash}p{0.7cm}%
                p{2cm}%
                >{\centering\arraybackslash}p{1.3cm}}

\addlinespace[3pt]
\rowcolor{bg-tb-heavey-nlp} \multicolumn{9}{c}{\bf \textcolor{redNLP}{\large NLP Group}} \\ [4pt]

\hline

\addlinespace[4pt]
\multicolumn{5}{c}{\bf Task} & \multicolumn{2}{c}{\bf Data}  & \multicolumn{2}{c}{\bf \textbf{SoTA Specialist}} \\

\cmidrule(r){1-5}  \cmidrule(r){6-7}  \cmidrule(r){8-9} 

\textbf{Skill (Meta-Task)} & \textbf{\#}& \textbf{Task (Short Name)} &  \textbf{Domain} & \textbf{Capability} & \textbf{Data Source} & \textbf{Number} & \multicolumn{1}{c}{\textbf{Model}}  & \textbf{Metrics} \\ 

\addlinespace[4pt]
\hline
\endfirsthead
\hline
\addlinespace[4pt]
\multicolumn{5}{c}{\bf Task} & \multicolumn{2}{c}{\bf Data}  & \multicolumn{2}{c}{\bf \textbf{SoTA Specialist}} \\
\cmidrule(r){1-5}  \cmidrule(r){6-7}  \cmidrule(r){8-9} 
\textbf{Skill (Meta-Task)} & \textbf{\#}& \textbf{Task (Short Name)} &  \textbf{Domain} & \textbf{Capability} & \textbf{Data Source} & \textbf{Number} & \multicolumn{1}{c}{\textbf{Model}}  & \textbf{Metrics} \\ 
\addlinespace[4pt]
\hline
\endhead
\hline
\endfoot

\addlinespace[4pt]
\multirow{9}{*}{\shortstack[l]{\#L-1 \\ Cognitive Question Answering \\ (Cog QA)}}
& 1 & Commonsense Reasoning (Common Reason) & General & Commonsense Knowledge & commonsense\_qa \cite{atca-naacl-19} & 510 & Flan-T5-Large \cite{abs-2210-11416} & Acc \\

\addlinespace[4pt]
& \cellcolor{lightgray}2 & \cellcolor{lightgray}Causal Reasoning (Causal Reason) & \cellcolor{lightgray}General & \cellcolor{lightgray}Causality Discrimination & \cellcolor{lightgray}corr2cause \cite{zjcl-iclr-24} & \cellcolor{lightgray}500 & \cellcolor{lightgray}Flan-T5-Large \cite{abs-2210-11416} & \cellcolor{lightgray}Acc \\

\addlinespace[4pt]
& 3 & Document-Level Causal Reasoning (Doc-Causal Reason) & General & Reasoning Ability & qa4mre \cite{apq2-clef-13} & 564 & Flan-T5-Large \cite{abs-2210-11416} & BLEU-1 \\

\addlinespace[4pt]
& \cellcolor{lightgray}4 & \cellcolor{lightgray}Counterfactual Reasoning (Counterfact Reason) & \cellcolor{lightgray}General & \cellcolor{lightgray}Causality Discrimination & \cellcolor{lightgray}Counterfactual-Reasoning-Capacity-of-Large-Language-Models \cite{crcllm} & \cellcolor{lightgray}501 & \cellcolor{lightgray}Flan-T5-Large \cite{abs-2210-11416} & \cellcolor{lightgray}F1 \\

\addlinespace[4pt]
& 5 & Analogical Reasoning (Analog Reason) & General & Reasoning Ability & BATS \cite{agab-naacl-16} & 500 & Flan-T5-Large \cite{abs-2210-11416} & Acc \\

\addlinespace[4pt]
& \cellcolor{lightgray}6 & \cellcolor{lightgray}Multi-Hop Question Answering (Multi-Hop QA) & \cellcolor{lightgray}General & \cellcolor{lightgray}Reasoning Ability & \cellcolor{lightgray}hotpot\_qa \cite{yang2018hotpotqa} & \cellcolor{lightgray}500 & \cellcolor{lightgray}Flan-T5-Large \cite{abs-2210-11416} & \cellcolor{lightgray}BLEU-1 \\

\addlinespace[4pt]
& 7 & Temporal Reasoning (Temporal Reason) & General & Reasoning Ability & time\_dial \cite{qin-etal-2021-timedial} & 506 & Flan-T5-Large \cite{abs-2210-11416} & BLEU-1 \\

\addlinespace[4pt]
& \cellcolor{lightgray}8 & \cellcolor{lightgray}Spatial Reasoning (Spatial Reason) & \cellcolor{lightgray}General & \cellcolor{lightgray}Spatial Perception & \cellcolor{lightgray}stepgame \cite{stepGame2022shi} & \cellcolor{lightgray}507 & \cellcolor{lightgray}Flan-T5-Large \cite{abs-2210-11416} & \cellcolor{lightgray}Acc \\

\addlinespace[4pt]
& 9 & Numerical Reasoning (Numerical Reason) & General & Reasoning Ability & aqua\_rat \cite{wlpi-acl-17} & 508 & Flan-T5-Large \cite{abs-2210-11416} & Acc \\

\addlinespace[4pt]
\hline

\addlinespace[4pt]
\multirow{9}{*}{\shortstack[l]{\#L-2 \\ Ethical NLP \\ (Ethics NLP)}}
& 1 & Ethical Reasoning (Ethical Reason) & Ethics & Ethical Awareness & ethics \cite{dhaa-iclr-21} & 500 & Flan-T5-Large \cite{abs-2210-11416} & Acc \\

\addlinespace[4pt]
& \cellcolor{lightgray}2 & \cellcolor{lightgray}Truthful Question Answering (Truthful QA) & \cellcolor{lightgray}Ethics & \cellcolor{lightgray}Ethical Awareness & \cellcolor{lightgray}truthful\_qa \cite{dhaa-iclr-21} & \cellcolor{lightgray}500 & \cellcolor{lightgray}Flan-T5-Large \cite{abs-2210-11416} & \cellcolor{lightgray}BLEU-1 \\

\addlinespace[4pt]
& 3 & Legal Question Answering (Legal QA) & Law & Ethical Awareness & JEC-QA \cite{hzjq-aaai-20} & 500 & Flan-T5-Large \cite{abs-2210-11416} & Acc \\

\addlinespace[4pt]
& \cellcolor{lightgray}4 & \cellcolor{lightgray}Bias Detection (Bias Detect) & \cellcolor{lightgray}Culture & \cellcolor{lightgray}Affective Analysis & \cellcolor{lightgray}WNC \cite{rpan-aaai-20} & \cellcolor{lightgray}500 & \cellcolor{lightgray}Roberta-Large \cite{liu2019roberta} & \cellcolor{lightgray}Micro-F1 \\

\addlinespace[4pt]
& 5 & Offensive Classification (Offensive Classify) & Culture & Affective Analysis & offensive-speech \cite{ockaggle} & 500 & Roberta-Large \cite{liu2019roberta} & Micro-F1 \\

\addlinespace[4pt]
& \cellcolor{lightgray}6 & \cellcolor{lightgray}Hate Speech Detection (Hate-Speech Detect) & \cellcolor{lightgray}Culture & \cellcolor{lightgray}Affective Analysis & \cellcolor{lightgray}hate-speech \cite{ockaggle} & \cellcolor{lightgray}500 & \cellcolor{lightgray}Roberta-Large \cite{liu2019roberta} & \cellcolor{lightgray}Micro-F1 \\

\addlinespace[4pt]
& 7 & Spam Detection (Spam Detect) & Culture & Causality Discrimination & spam-email \cite{spamdkaggle} & 500 & Roberta-Large \cite{liu2019roberta} & Micro-F1 \\

\addlinespace[4pt]
& \cellcolor{lightgray}8 & \cellcolor{lightgray}Fake News Detection (Fake-News Detect) & \cellcolor{lightgray}Humanities & \cellcolor{lightgray}Ethical Awareness & \cellcolor{lightgray}fake-news \cite{fakekaggle} & \cellcolor{lightgray}500 & \cellcolor{lightgray}Roberta-Large \cite{liu2019roberta} & \cellcolor{lightgray}Micro-F1 \\

\addlinespace[4pt]
& 9 & Fact Verification (Fact Verify) & General & Causality Discrimination & Counterfactual \cite{joiw-emnlp-21} & 500 & Roberta-Large \cite{liu2019roberta} & Micro-F1 \\

\addlinespace[4pt]
\hline

\addlinespace[4pt]
\multirow{8}{*}{\shortstack[l]{\#L-3 \\ Domain-Specific \\ QA  (Domain QA)}}
& 1 & Biomedical Question Answering (Biomedical QA) & Biomedical & Reasoning Ability & BioASQ \cite{bqbioasq} & 500 & Flan-T5-Large \cite{abs-2210-11416} & BLEU-1 \\
\addlinespace[4pt]
& \cellcolor{lightgray}2 & \cellcolor{lightgray}Medical Question Answering (Medical QA) & \cellcolor{lightgray}Medical & \cellcolor{lightgray}Reasoning Ability & \cellcolor{lightgray}ChatDoctor-HealthCareMagic-100k \cite{li2023chatdoctor} & \cellcolor{lightgray}505 & \cellcolor{lightgray}Flan-T5-Large \cite{abs-2210-11416} & \cellcolor{lightgray}BLEU-1 \\

\addlinespace[4pt]
& 3 & Technical Question Answering (Tech QA) & Engineering & Reasoning Ability & Stack Overflow Data \cite{sodkaggle} & 500 & Flan-T5-Large \cite{abs-2210-11416} & BLEU-1 \\

\addlinespace[4pt]
& \cellcolor{lightgray}4 & \cellcolor{lightgray}Engineering Question Answering (Eng QA) & \cellcolor{lightgray}Engineering & \cellcolor{lightgray}Reasoning Ability & \cellcolor{lightgray}University of Bath \cite{eqabath} & \cellcolor{lightgray}500 & \cellcolor{lightgray}Flan-T5-Large \cite{abs-2210-11416} & \cellcolor{lightgray}BLEU-1 \\

\addlinespace[4pt]
& 5 & Science Question Answering (Science QA) & Science & Reasoning Ability & sciq \cite{jwcm-aclnut-17} & 500 & Flan-T5-Large \cite{abs-2210-11416} & Acc \\

\addlinespace[4pt]
& \cellcolor{lightgray}6 & \cellcolor{lightgray}Earth Question Answering (Earth QA) & \cellcolor{lightgray}Earth & \cellcolor{lightgray}Reasoning Ability & \cellcolor{lightgray}Manual & \cellcolor{lightgray}500 & \cellcolor{lightgray}Flan-T5-Large \cite{abs-2210-11416} & \cellcolor{lightgray}Acc \\

\addlinespace[4pt]
& 7 & Nature Question Answering (Nature QA) & Nature & Reasoning Ability & Manual & 500 & Flan-T5-Large \cite{abs-2210-11416} & Acc \\

\addlinespace[4pt]
\hline

\addlinespace[4pt]
\multirow{1}{*}{\shortstack[l]{\#L-4 \\ Social  QA \\ (Social QA)}}
& 1 & Humanities Question Answering (Humanities QA) \newline \color{white}{1} & Humanities & Reasoning Ability & squad\_v2 \cite{prkw-acl-18} & 500 & Flan-T5-Large \cite{abs-2210-11416} & BLEU-1 \\

\addlinespace[4pt]
& \cellcolor{lightgray}2 & \cellcolor{lightgray}Social Science Question Answering (Social-Sci QA) & \cellcolor{lightgray}Social & \cellcolor{lightgray}Reasoning Ability & \cellcolor{lightgray}Social IQA \cite{mssi-emnlp-19} & \cellcolor{lightgray}500 & \cellcolor{lightgray}Flan-T5-Large \cite{abs-2210-11416} & \cellcolor{lightgray}Acc \\

\addlinespace[4pt]
& 3 & Philosophical Question Answering (Philosophy QA) & Philosophy & Reasoning Ability & strix-philosophy-qa \cite{pqahugging} & 500 & Flan-T5-Large \cite{abs-2210-11416} & BLEU-1 \\

\addlinespace[4pt]
& \cellcolor{lightgray}4 & \cellcolor{lightgray}Art Question Answering (Art QA) & \cellcolor{lightgray}Art & \cellcolor{lightgray}Reasoning Ability & \cellcolor{lightgray}Manual & \cellcolor{lightgray}500 & \cellcolor{lightgray}Flan-T5-Large \cite{abs-2210-11416} & \cellcolor{lightgray}Acc \\

\addlinespace[4pt]
& 5 & Business Question Answering (Business QA) & Business & Reasoning Ability & Manual & 500 & Flan-T5-Large \cite{abs-2210-11416} & Acc \\

\addlinespace[4pt]
& \cellcolor{lightgray}6 & \cellcolor{lightgray}History Question Answering (History QA) & \cellcolor{lightgray}History & \cellcolor{lightgray}Reasoning Ability & \cellcolor{lightgray}wiki-reading \cite{tkbl-aaai-18} & \cellcolor{lightgray}500 & \cellcolor{lightgray}Flan-T5-Large \cite{abs-2210-11416} & \cellcolor{lightgray}BLEU-1 \\

\addlinespace[4pt]
\hline

\addlinespace[4pt]
\multirow{3}{*}{\shortstack[l]{\#L-5 \\ Non-Traditional \\ QA  (Non-Trad QA)}}
& 1 & Fairytale Question Answering (Fairytale QA) & Culture & Reasoning Ability & FairytaleQA \cite{prkw-acl-18} & 501 & Flan-T5-Large \cite{abs-2210-11416} & BLEU-1 \\

\addlinespace[4pt]
& \cellcolor{lightgray}2 & \cellcolor{lightgray}Tweet Question Answering (Tweet QA) & \cellcolor{lightgray}Social & \cellcolor{lightgray}Reasoning Ability & \cellcolor{lightgray}tweet\_qa \cite{wxta-acl-19} & \cellcolor{lightgray}509 & \cellcolor{lightgray}Flan-T5-Large \cite{abs-2210-11416} & \cellcolor{lightgray}F1 \\

\addlinespace[4pt]
& 3 & Trivial QA (Trivial QA) & Social & Reasoning Ability & trivia\_qa \cite{mjta-acl-17} & 500 & Flan-T5-Large \cite{abs-2210-11416} & F1 \\

\addlinespace[4pt]
\hline

\addlinespace[4pt]
\multirow{6}{*}{\shortstack[l]{\#L-6 \\ Advanced QA\\ (Advance QA)}}
& 1 & Open-Domain Question Answering (Open-Domain QA) & General & Reasoning Ability & squad\_v2 \cite{prkw-acl-18} & 513 & Flan-T5-Large \cite{abs-2210-11416} & BLEU-1 \\

\addlinespace[4pt]
& \cellcolor{lightgray}2 & \cellcolor{lightgray}Conversational Question Answering (Conversation QA) & \cellcolor{lightgray}General & \cellcolor{lightgray}Reasoning Ability & \cellcolor{lightgray}coqa \cite{srca-tacl-19} & \cellcolor{lightgray}500 & \cellcolor{lightgray}Flan-T5-Large \cite{abs-2210-11416} & \cellcolor{lightgray}BLEU-1 \\

\addlinespace[4pt]
& 3 & Table Question Answering (Table QA) & General & Reasoning Ability & wikitablequestions \cite{ppcs-acl-15} & 590 & Flan-T5-Large \cite{abs-2210-11416} & F1 \\

\addlinespace[4pt]
& \cellcolor{lightgray}4 & \cellcolor{lightgray}Multilingual Question Answering (Multi-Lang QA) & \cellcolor{lightgray}Linguistics & \cellcolor{lightgray}Reasoning Ability & \cellcolor{lightgray}multilingual\_qa \cite{mqahg} & \cellcolor{lightgray}600 & \cellcolor{lightgray}mT5-Large \cite{xue-etal-2021-mt5} & \cellcolor{lightgray}BLEU-1 \\

\addlinespace[4pt]
& 5 & Code-Switch Question Answering (Code-Switch QA) & Linguistics & Reasoning Ability & Manual & 500 & mT5-Large \cite{xue-etal-2021-mt5} & Acc \\

\addlinespace[4pt]
\hline

\addlinespace[4pt]
\multirow{3}{*}{\shortstack[l]{\#L-7 \\ Math Problem \\ Solving \\ (Math Ability)}}
& 1 & Math Question Answering (Math QA) & Math & Reasoning Ability & math-QA \cite{aamt-naacl-19} & 500 & Flan-T5-Large \cite{abs-2210-11416} & Acc \\

\addlinespace[4pt]
& \cellcolor{lightgray}2 & \cellcolor{lightgray}Mathematical Word Problem Solving (Math Word Prob) & \cellcolor{lightgray}Math & \cellcolor{lightgray}Problem Solving & \cellcolor{lightgray}math-QA \cite{aamt-naacl-19} & \cellcolor{lightgray}500 & \cellcolor{lightgray}Flan-T5-Large \cite{abs-2210-11416} & \cellcolor{lightgray}Acc \\

\addlinespace[4pt]
& 3 & Mathematical Proof Generation (Math Proof Gen) & Math & Problem Solving & math-QA \cite{aamt-naacl-19} & 500 & Flan-T5-Large \cite{abs-2210-11416} & BLEU-1 \\

\addlinespace[4pt]
\hline

\addlinespace[4pt]
\multirow{5}{*}{\shortstack[l]{\#L-8 \\ Code Problem \\ Solving \\ (Code Ability)}}
& 1 & Code Explanation (Code Explain) & Code & Problem Solving & CodeXGLUE \cite{slca-nips-21} & 508 & CodeT5 \cite{wang2021codet5} & BLEU-1 \\

\addlinespace[4pt]
& \cellcolor{lightgray}2 & \cellcolor{lightgray}Code Defect Detection (Code-Defect Detect) & \cellcolor{lightgray}Code & \cellcolor{lightgray}Problem Solving & \cellcolor{lightgray}CodeXGLUE \cite{slca-nips-21} & \cellcolor{lightgray}501 & \cellcolor{lightgray}CodeT5 \cite{wang2021codet5} & \cellcolor{lightgray}Acc \\

\addlinespace[4pt]
& 3 & Code Generation (Code Gen) & Code & Problem Solving & mbpp \cite{mbppgithub} & 500 & CodeT5 \cite{wang2021codet5} & CodeBLEU \\

\addlinespace[4pt]
& \cellcolor{lightgray}4 & \cellcolor{lightgray}Code Repair (Code Repair) & \cellcolor{lightgray}Code & \cellcolor{lightgray}Problem Solving & \cellcolor{lightgray}CodeXGLUE \cite{slca-nips-21} & \cellcolor{lightgray}600 & \cellcolor{lightgray}CodeT5 \cite{wang2021codet5} & \cellcolor{lightgray}CodeBLEU \\

\addlinespace[4pt]
& 5 & Text-to-SQL Generation (Txt2SQL Gen) & Code & Problem Solving & Text-to-sql-v1 \cite{} & 600 & CodeT5 \cite{wang2021codet5} & BLEU-1 \\

\addlinespace[4pt]
\hline

\addlinespace[4pt]
\multirow{4}{*}{\shortstack[l]{\#L-9 \\ Cross-lingual/ \\ Translation \\ (X-Lan\&NMT) }}
& 1 & Multilingual Translation (Multi-lang Trans) & General & Multilingual Capability & Manual & 504 & Flan-T5-Large \cite{abs-2210-11416} & ROUGE-1 \\

\addlinespace[4pt]
& \cellcolor{lightgray}2 & \cellcolor{lightgray}Low-Resource Translation (Low-Res Trans) & \cellcolor{lightgray}General & \cellcolor{lightgray}Multilingual Capability & \cellcolor{lightgray}flores\_101 \cite{ngtf-tacl-22} & \cellcolor{lightgray}502 & \cellcolor{lightgray}Flan-T5-Large \cite{abs-2210-11416} & \cellcolor{lightgray}ROUGE-1 \\

\addlinespace[4pt]
& 3 & English-Chinese Translation (En-Zh Trans) & General & Multilingual Capability & Manual & 500 & Flan-T5-Large \cite{abs-2210-11416} & ROUGE-1 \\

\addlinespace[4pt]
& \cellcolor{lightgray}4 & \cellcolor{lightgray}English-French Translation (En-Fr Trans) & \cellcolor{lightgray}General & \cellcolor{lightgray}Multilingual Capability & \cellcolor{lightgray}Manual & \cellcolor{lightgray}501 & \cellcolor{lightgray}Flan-T5-Large \cite{abs-2210-11416} & \cellcolor{lightgray}ROUGE-1 \\

\addlinespace[4pt]
\hline

\addlinespace[4pt]
\multirow{3}{*}{\shortstack[l]{\#L-10 \\ Text \\Summarization \\ (Txt Sum)}}
& 1 & Extractive Summarization (Extract Summ) & General & Text Manipulation & cnn\_dailymail \cite{asgt-acl-17} & 503 & BART-Large \cite{lewis2019bart} & ROUGE-1 \\

\addlinespace[4pt]
& \cellcolor{lightgray}2 & \cellcolor{lightgray}Abstractive Summarization (Abstract Sum) & \cellcolor{lightgray}General & \cellcolor{lightgray}Text Manipulation & \cellcolor{lightgray}xsum \cite{sndt-emnlp-18} & \cellcolor{lightgray}501 & \cellcolor{lightgray}BART-Large \cite{lewis2019bart} & \cellcolor{lightgray}ROUGE-1 \\

\addlinespace[4pt]
& 3 & Multi-Document Summarization (Multi-Doc Sum) & General & Text Manipulation & multi\_news \cite{armn-acl-19} & 300 & BART-Large \cite{lewis2019bart} & ROUGE-1 \\

\addlinespace[4pt]
\hline

\addlinespace[4pt]
\multirow{1}{*}{\shortstack[l]{\#L-11 \\ Dialogue Generation\\ (Dialog Gen)}}
& 1 & Multi-Turn Daily Dialogue Generation (Daily Dialogue Gen) & General & Interactive Capability, Cognition Understanding & daily\_dialogue \cite{ylda-ijcnlp-17} & 553 & DialoGPT \cite{zhang2019dialogpt} & BLEU-1 \\

\addlinespace[4pt]
\hline

\addlinespace[4pt]
\multirow{5}{*}{\shortstack[l]{\#L-12 \\ Text  Generation \\ (TxT Gen)}}
& 1 & Table-to-Text Generation (Table-to-Text) & General & Text Manipulation & ToTTo \cite{apta-emnlp-20} & 500 & Flan-T5-Large \cite{abs-2210-11416} & BLEU-1 \\

\addlinespace[4pt]
& \cellcolor{lightgray}2 & \cellcolor{lightgray}Text Style Transfer (Style Transfer) & \cellcolor{lightgray}General & \cellcolor{lightgray}Creativity and lnnovation & \cellcolor{lightgray}style-transfer-paraphrase, text\_style\_transfer \cite{} & \cellcolor{lightgray}500 & \cellcolor{lightgray}Flan-T5-Large \cite{abs-2210-11416} & \cellcolor{lightgray}BLEU-1 \\

\addlinespace[4pt]
& 3 & Story Generation (Story Gen) & General & Creativity and lnnovation & story-generation \cite{kkru-emnlp-20} & 600 & mFlan-T5-Large \cite{abs-2210-11416} & BLEU-1 \\

\addlinespace[4pt]
& \cellcolor{lightgray}4 & \cellcolor{lightgray}Paraphrase Generation (Paraphrase) & \cellcolor{lightgray}General & \cellcolor{lightgray}Creativity and lnnovation & \cellcolor{lightgray}Manual \cite{storyhg} & \cellcolor{lightgray}633 & \cellcolor{lightgray}Flan-T5-Large \cite{abs-2210-11416} & \cellcolor{lightgray}BLEU-1 \\

\addlinespace[4pt]
& 5 & Grammar Correction (Grammar Correct) & General & Text Manipulation & Grammar Correction \cite{grammarckgggle} & 302 & Flan-T5-Large \cite{abs-2210-11416} & BLEU-1 \\

\addlinespace[4pt]
\hline

\addlinespace[4pt]
\multirow{1}{*}{\shortstack[l]{\#L-13 \\ Time Series  Analysis\\ (Time Series)}}
& 1 & Time Series Prediction (Time Series Pred) & General & Temporal Determination, Reasoning Ability & Manual & 510 & TimesFm-1.0-200m \cite{das2023decoder} & RMSE \\

\addlinespace[4pt]
\hline

\addlinespace[4pt]
\multirow{3}{*}{\shortstack[l]{\#L-14 \\ Content\\ Categorization\\ (Txt Cls)}}
& 1 & Topic Classification (Topic Cls) & General & Content Recognition & Topic \cite{topickg} & 500 & Roberta-Large \cite{liu2019roberta} & Micro-F1 \\

\addlinespace[4pt]
\hline

\addlinespace[4pt]
\multirow{1}{*}{\shortstack[l]{\#L-15 \\ Text Entailment\\ (Txt Entail)}}
& 1 & Natural Language Inference (NLI) & General & Reasoning Ability, Causality Discrimination & SNLI \cite{sral-emnlp-15} & 500 & Roberta-Large \cite{liu2019roberta} & Micro-F1 \\

\addlinespace[4pt]
\hline

\addlinespace[4pt]
\multirow{1}{*}{\shortstack[l]{
\#L-16 \\ Semantic Analysis \\ (Sem Analy)
}}
& 1 & Sentence Similarity Detection (Sent Similar) & General & Reasoning Ability, Causality Discrimination & QuoraQP \cite{sentkg} & 500 & Roberta-Large \cite{liu2019roberta} & Micro-F1 \\

\addlinespace[4pt]
& \cellcolor{lightgray}2 & \cellcolor{lightgray}Intent Detection (Intent Det) & \cellcolor{lightgray}General, Social, Culture & \cellcolor{lightgray}Cognition Understanding, Reasoning Ability & \cellcolor{lightgray}Intent \cite{intentgithub} & \cellcolor{lightgray}500 & \cellcolor{lightgray}Roberta-Large \cite{liu2019roberta} & \cellcolor{lightgray}Micro-F1 \\

\addlinespace[4pt]
& 3 & Stance Detection (Stance Detect) & General, Social, Humanities & Cognition Understanding, Reasoning Ability & SemEval2016 \cite{sms2-semeval-16} & 500 & Roberta-Large \cite{liu2019roberta} & Micro-F1 \\

\addlinespace[4pt]
& \cellcolor{lightgray}4 & \cellcolor{lightgray}Personality Analysis (Person Analy) & \cellcolor{lightgray}Business & \cellcolor{lightgray}Reasoning Ability, Commonsense Knowledge, Cognition Understanding & \cellcolor{lightgray}Essay2007 \cite{pergithub} & \cellcolor{lightgray}500 & \cellcolor{lightgray}Roberta-Large \cite{liu2019roberta} & \cellcolor{lightgray}Micro-F1 \\

\addlinespace[4pt]
\hline

\addlinespace[4pt]
\multirow{19}{*}{\shortstack[l]{\#L-17 \\ Affective \\ Computing\\ (Affect Computing)}}
& 1 & Humor Detection (Humor Detect) & Culture & Affective Analysis & Reddit \cite{owhd-emnlp-19} & 500 & Roberta-Large \cite{liu2019roberta} & Micro-F1 \\

\addlinespace[4pt]
& \cellcolor{lightgray}2 & \cellcolor{lightgray}Sarcasm Detection (Sarcasm Det) & \cellcolor{lightgray}Culture & \cellcolor{lightgray}Affective Analysis & \cellcolor{lightgray}Sarcasm \cite{rmsd-aiopen-23} & \cellcolor{lightgray}500 & \cellcolor{lightgray}Roberta-Large \cite{liu2019roberta} & \cellcolor{lightgray}Micro-F1 \\

\addlinespace[4pt]
& 3 & Sentiment Classification (Sentiment Cls) & General & Affective Analysis & SST5 \cite{rsrd-emnlp-13} & 500 & Roberta-Large \cite{liu2019roberta} & Micro-F1 \\

\addlinespace[4pt]
& \cellcolor{lightgray}4 & \cellcolor{lightgray}Mental Health Toxicity Detection (Mental-Toxic Det) & \cellcolor{lightgray}Humanities & \cellcolor{lightgray}Affective Analysis & \cellcolor{lightgray}MentalHealth \cite{mentalgithub} & \cellcolor{lightgray}500 & \cellcolor{lightgray}Roberta-Large \cite{liu2019roberta} & \cellcolor{lightgray}Micro-F1 \\

\addlinespace[4pt]
& 5 & Financial Sentiment Analysis (Finance Cls) & Finance & Affective Analysis & Financial Sentiment Analysis \cite{pmgd-jasis-14} & 500 & Roberta-Large \cite{liu2019roberta} & Micro-F1 \\

\addlinespace[4pt]
& \cellcolor{lightgray}6 & \cellcolor{lightgray}Metaphor Detection (Metaphor Det) & \cellcolor{lightgray}Culture & \cellcolor{lightgray}Affective Analysis & \cellcolor{lightgray}Metaphor \cite{mcmm-naacl-21} & \cellcolor{lightgray}500 & \cellcolor{lightgray}Roberta-Large \cite{liu2019roberta} & \cellcolor{lightgray}Micro-F1 \\

\addlinespace[4pt]
& 7 & Aspect Category Detection (Aspect Det) & Business & Affective Analysis & res15 \cite{mps2-semeval-15} & 500 & Roberta-Large \cite{liu2019roberta} & Micro-F1 \\

\addlinespace[4pt]
& \cellcolor{lightgray}8 & \cellcolor{lightgray}Aspect Sentiment Classification (Aspect-Senti Cls) & \cellcolor{lightgray}Business & \cellcolor{lightgray}Affective Analysis & \cellcolor{lightgray}res15 \cite{mps2-semeval-15} & \cellcolor{lightgray}500 & \cellcolor{lightgray}Flan-T5-Large \cite{abs-2210-11416} & \cellcolor{lightgray}Micro-F1 \\

\addlinespace[4pt]
& 9 & Aspect Term Extraction (Aspect-Term Ext) & Business & Affective Analysis & res15 \cite{mps2-semeval-15} & 500 & Flan-T5-Large \cite{abs-2210-11416} & Micro-F1 \\

\addlinespace[4pt]
& \cellcolor{lightgray}10 & \cellcolor{lightgray}Target-oriented Opinion Words Extraction (Opinion Ext) & \cellcolor{lightgray}Business & \cellcolor{lightgray}Affective Analysis & \cellcolor{lightgray}TOWE \cite{zfto-naacl-19} & \cellcolor{lightgray}500 & \cellcolor{lightgray}Flan-T5-Large \cite{abs-2210-11416} & \cellcolor{lightgray}Micro-F1 \\

\addlinespace[4pt]
& 11 & Aspect-Opinion Pair Extraction (AO-Pair Ext) & Business & Affective Analysis & SDRN \cite{scsd-acl-20} & 500 & Flan-T5-Large \cite{abs-2210-11416} & Micro-F1 \\

\addlinespace[4pt]
& \cellcolor{lightgray}12 & \cellcolor{lightgray}End-to-End ABSA (ABSA) & \cellcolor{lightgray}Business & \cellcolor{lightgray}Affective Analysis & \cellcolor{lightgray}ABSA \cite{mps2-semeval-15} & \cellcolor{lightgray}500 & \cellcolor{lightgray}Flan-T5-Large \cite{abs-2210-11416} & \cellcolor{lightgray}Micro-F1 \\

\addlinespace[4pt]
& 13 & Aspect Sentiment Triplet Extraction (Senti-Triplet Ext) & Business & Affective Analysis & ASTE \cite{lxpa-emnlp-20} & 500 & Flan-T5-Large \cite{abs-2210-11416} & Micro-F1 \\

\addlinespace[4pt]
& \cellcolor{lightgray}14 & \cellcolor{lightgray}Aspect-Category-Sentiment Detection (ACS Det) & \cellcolor{lightgray}Business & \cellcolor{lightgray}Affective Analysis & \cellcolor{lightgray}res15 \cite{mps2-semeval-15} & \cellcolor{lightgray}500 & \cellcolor{lightgray}Flan-T5-Large \cite{abs-2210-11416} & \cellcolor{lightgray}Micro-F1 \\

\addlinespace[4pt]
& 15 & Aspect Sentiment Quad Prediction (Senti-Quad Pred) & Business & Affective Analysis & res16 \cite{mps2-semeval-16} & 500 & Flan-T5-Large \cite{abs-2210-11416} & Micro-F1 \\

\addlinespace[4pt]
& \cellcolor{lightgray}16 & \cellcolor{lightgray}Dialogue-Level Sentiment Quadruple Extraction (Dialog Sent Quad) & \cellcolor{lightgray}Business & \cellcolor{lightgray}Affective Analysis & \cellcolor{lightgray}diaasq \cite{blda-acl-23} & \cellcolor{lightgray}500 & \cellcolor{lightgray}Flan-T5-Large \cite{abs-2210-11416} & \cellcolor{lightgray}Micro-F1 \\

& 17 & Soccer Sentiment Classification (Soccer SA) & Sports & Affective Analysis & FIFA-SA\cite{soccergithub} & 500 & Roberta-Large \cite{liu2019roberta} & Micro-F1 \\

\addlinespace[4pt]
& \cellcolor{lightgray}18 & \cellcolor{lightgray}Comparative Opinion Quintuple Extraction (Opinion-Quin Ext) & \cellcolor{lightgray}Business & \cellcolor{lightgray}Affective Analysis & \cellcolor{lightgray}Camera \cite{zlco-emnlp-21} & \cellcolor{lightgray}500 & \cellcolor{lightgray}Flan-T5-Large \cite{abs-2210-11416} & \cellcolor{lightgray}Micro-F1 \\

\addlinespace[4pt]
\hline

\addlinespace[4pt]
\multirow{10}{*}{\shortstack[l]{\#L-18 \\ Named Entity \\ Recognition (NER)}}
& 1 & Scientific NER (Sci NER) & Physical Science & Content Recognition & SciER \cite{qzsa-emnlp-24} & 500 & W2NER \cite{li2022unified} & Micro-F1 \\

\addlinespace[4pt]
& \cellcolor{lightgray}2 & \cellcolor{lightgray}Temporal NER (Temporal NER) & \cellcolor{lightgray}Engineering & \cellcolor{lightgray}Content Recognition & \cellcolor{lightgray}TimeBank \cite{temgithub} & \cellcolor{lightgray}500 & \cellcolor{lightgray}W2NER \cite{li2022unified} & \cellcolor{lightgray}Micro-F1 \\

\addlinespace[4pt]
& 3 & Pathology NER (Path NER) & Biology & Content Recognition & Pathology NER \cite{pathnergit} & 500 & W2NER \cite{li2022unified} & Micro-F1 \\

\addlinespace[4pt]
& \cellcolor{lightgray}4 & \cellcolor{lightgray}Cybersecurity NER (Cyber NER) & \cellcolor{lightgray}Physical Science & \cellcolor{lightgray}Content Recognition & \cellcolor{lightgray}CyNER \cite{cybergit} & \cellcolor{lightgray}500 & \cellcolor{lightgray}W2NER \cite{li2022unified} & \cellcolor{lightgray}Micro-F1 \\

\addlinespace[4pt]
& 5 & Geological NER (Geo NER) & Geography & Content Recognition & GEO-NER \cite{geonergit} & 500 & W2NER \cite{li2022unified} & Micro-F1 \\

\addlinespace[4pt]
& \cellcolor{lightgray}6 & \cellcolor{lightgray}Legal NER (Legal NER) & \cellcolor{lightgray}Politics & \cellcolor{lightgray}Content Recognition & \cellcolor{lightgray}InLegalNER \cite{pkne-acl-nllp-22} & \cellcolor{lightgray}500 & \cellcolor{lightgray}W2NER \cite{li2022unified} & \cellcolor{lightgray}Micro-F1 \\

\addlinespace[4pt]
& 7 & Organization Recognition (Organ NER) & Politics & Content Recognition & CoNLL2003 \cite{efit-conll-03} & 500 & W2NER \cite{li2022unified} & Micro-F1 \\

\addlinespace[4pt]
& \cellcolor{lightgray}8 & \cellcolor{lightgray}Person Recognition (Person NER) & \cellcolor{lightgray}General & \cellcolor{lightgray}Content Recognition & \cellcolor{lightgray}CoNLL2003 \cite{efit-conll-03} & \cellcolor{lightgray}500 & \cellcolor{lightgray}W2NER \cite{li2022unified} & \cellcolor{lightgray}Micro-F1 \\

\addlinespace[4pt]
& 9 & Location Recognition (Location NER) & General & Content Recognition & CoNLL2003 \cite{efit-conll-03} & 500 & W2NER \cite{li2022unified} & Micro-F1 \\

\addlinespace[4pt]

\addlinespace[4pt]
& \cellcolor{lightgray}10 & \cellcolor{lightgray}Climate Change NER (Climate NER) & \cellcolor{lightgray}Climate & \cellcolor{lightgray}Content Recognition & \cellcolor{lightgray}Climate-Change-NER \cite{efit-conll-03} & \cellcolor{lightgray}500 & \cellcolor{lightgray}W2NER \cite{li2022unified} & \cellcolor{lightgray}Micro-F1 \\

\addlinespace[4pt]

& 11 & Gene/Protein Named Entity Recognition (Gene\&Protein NER) & Biology & Content Recognition & Genia \cite{efit-conll-03} & 500 & W2NER \cite{li2022unified} & Micro-F1 \\

\addlinespace[4pt]

& \cellcolor{lightgray}12 & \cellcolor{lightgray}Chemical Named Entity Recognition (Chem NER) & \cellcolor{lightgray}Chemistry & \cellcolor{lightgray}Content Recognition & \cellcolor{lightgray}CHEMDNER \cite{chenerkaggle} & \cellcolor{lightgray}500 & \cellcolor{lightgray}W2NER \cite{li2022unified} & \cellcolor{lightgray}Micro-F1 \\

\addlinespace[4pt]
& 13 & Disease-NER (Disease NER) & Biology & Content Recognition & Disease-NER \cite{diseasegit} & 500 & W2NER \cite{li2022unified} & Micro-F1 \\

\addlinespace[4pt]
\hline

\addlinespace[4pt]
\multirow{24}{*}{\shortstack[l]{\#L-19 \\ Relation Extraction\\ (Rel Ext)}}
& 1 & Scientific Relation Extraction (Sci RE) & Physical Science & Content Recognition & SciER \cite{qzsa-emnlp-24} & 500 & W2NER \cite{li2022unified} & Micro-F1 \\

\addlinespace[4pt]
& \cellcolor{lightgray}2 & \cellcolor{lightgray}News Relation Extraction (News RE) & \cellcolor{lightgray}General & \cellcolor{lightgray}Content Recognition & \cellcolor{lightgray}TACRED \cite{yzpa-emnlp-17} & \cellcolor{lightgray}500 & \cellcolor{lightgray}Roberta-Large \cite{liu2019roberta} & \cellcolor{lightgray}Micro-F1 \\

\addlinespace[4pt]
& 3 & Joint NER and RE (Joint-NER-RE) & General & Content Recognition & TACRED \cite{yzpa-emnlp-17} & 500 & W2NER \cite{li2022unified} & Micro-F1 \\

\addlinespace[4pt]
& \cellcolor{lightgray}4 & \cellcolor{lightgray}DocRE (Doc RE) & \cellcolor{lightgray}General & \cellcolor{lightgray}Content Recognition & \cellcolor{lightgray}DocRED \cite{yyda-acl-19} & \cellcolor{lightgray}500 & \cellcolor{lightgray}W2NER \cite{li2022unified} & \cellcolor{lightgray}Micro-F1 \\

\addlinespace[4pt]
& 5 & Adverse Drug Reaction (ADR) Detection (ADR Det) & Medicine & Content Recognition & ADEv2 \cite{hgdo-jbi-12} & 500 & Roberta-Large \cite{liu2019roberta} & Micro-F1 \\

\addlinespace[4pt]
& \cellcolor{lightgray}6 & \cellcolor{lightgray}Protein-Protein Interaction Extraction (PPI Ext) & \cellcolor{lightgray}Biology & \cellcolor{lightgray}Content Recognition & \cellcolor{lightgray}PPI \cite{ppigithub} & \cellcolor{lightgray}500 & \cellcolor{lightgray}W2NER \cite{li2022unified} & \cellcolor{lightgray}Micro-F1 \\

\addlinespace[4pt]
& 7 & Drug-Drug Interaction (DDI Ext) & Medicine & Content Recognition & DDI \cite{ddigithub} & 500 & Roberta-Large \cite{liu2019roberta} & Acc \\

\addlinespace[4pt]
& \cellcolor{lightgray}8 & \cellcolor{lightgray}Genetic Association Datebase (Genetic Assoc) & \cellcolor{lightgray}Biology & \cellcolor{lightgray}Content Recognition & \cellcolor{lightgray}GAD \cite{gadgit} & \cellcolor{lightgray}500 & \cellcolor{lightgray}Roberta-Large \cite{liu2019roberta} & \cellcolor{lightgray}Micro-F1 \\

\addlinespace[4pt]
& 9 & Chemical-Protein Relationship Extraction (Chem-Protein RE) & Chemistry & Content Recognition & ChemProt\_BLURB \cite{cheemnlpgit} & 500 & Roberta-Large \cite{liu2019roberta} & Micro-F1 \\

\hline

\addlinespace[4pt]
\multirow{2}{*}{\shortstack[l]{\#L-20 \\ Event  Extraction\\ (Event Ext)}}
& 1 & Event Trigger Detection (Event-Trigger Det) & General & Content Recognition & TAC-KBP \cite{eventldc} & 500 & Flan-T5-Large \cite{abs-2210-11416} & Micro-F1 \\

\addlinespace[4pt]
& 2 & Temporal Event Reasoning (Temp-Event Reason) & General & Content Recognition, Temporal Determination & TimeBank \cite{temgithub} & 500 & Roberta-Large \cite{liu2019roberta} & Micro-F1 \\

\addlinespace[4pt]
\hline

\addlinespace[4pt]
\multirow{3}{*}{\shortstack[l]{\#L-21 \\ Semantic Parsing\\ (Sem Par)}}
& 1 & Event Coreference Resolution (Event-Coref Res) & Linguistics & Content Recognition & TAC-KBP 2015 \cite{eventldc} & 500 & Flan-T5-Large \cite{abs-2210-11416} & Acc \\

\addlinespace[4pt]
& \cellcolor{lightgray}2 & \cellcolor{lightgray}Semantic Role Labeling (SRL) & \cellcolor{lightgray}Linguistics & \cellcolor{lightgray}Content Recognition & \cellcolor{lightgray}OntoNotes \cite{srlldc} & \cellcolor{lightgray}500 & \cellcolor{lightgray}Flan-T5-Large \cite{abs-2210-11416} & \cellcolor{lightgray} Acc \\

\addlinespace[4pt]
& 3 & Abstract Meaning Representation (AMR) & Linguistics & Content Recognition & AMR 2.0 {amrldc} \cite{amrldc} & 500 & Flan-T5-Large \cite{abs-2210-11416} & Acc \\

\addlinespace[4pt]
\hline

\addlinespace[4pt]
\multirow{3}{*}{\shortstack[l]{\#L-22 \\ Linguistic Parsing\\ (Ling Par)}}
& 1 & Dependency Parsing (Dep Parse) & Linguistics & Content Recognition & Universal Dependency \cite{srlldc} & 500 & Flan-T5-Large \cite{abs-2210-11416} & Acc \\

\addlinespace[4pt]
& \cellcolor{lightgray}2 & \cellcolor{lightgray}Constituency Parsing (Const Parse) & \cellcolor{lightgray}Linguistics & \cellcolor{lightgray}Content Recognition & \cellcolor{lightgray}OntoNotes \cite{dpldc} & \cellcolor{lightgray}500 & \cellcolor{lightgray}Flan-T5-Large \cite{abs-2210-11416} & \cellcolor{lightgray} Acc \\

\addlinespace[4pt]
& 3 & Part-of-Speech (POS) & Linguistics & Content Recognition & Wall Street Journal \cite{posgithub} & 500 & Flan-T5-Large \cite{abs-2210-11416} & Micro-F1 \\

\addlinespace[4pt]
\hline

\end{longtable}
}

\subsection{Example Task Gallery}

In this part, we provide visualized examples of representative tasks under each skill to offer a clearer and more intuitive understanding of the task definitions, including their inputs and outputs. 
The examples are categorized based on different modalities and paradigms.

\subsubsection{Image-related Tasks}

\paragraph{Comprehension Tasks.}
Following we showcase 40 image-oriented comprehensive tasks, each representing a specific skill.

\begin{figure*}[!h]
\centering
\includegraphics[width=0.9\linewidth]{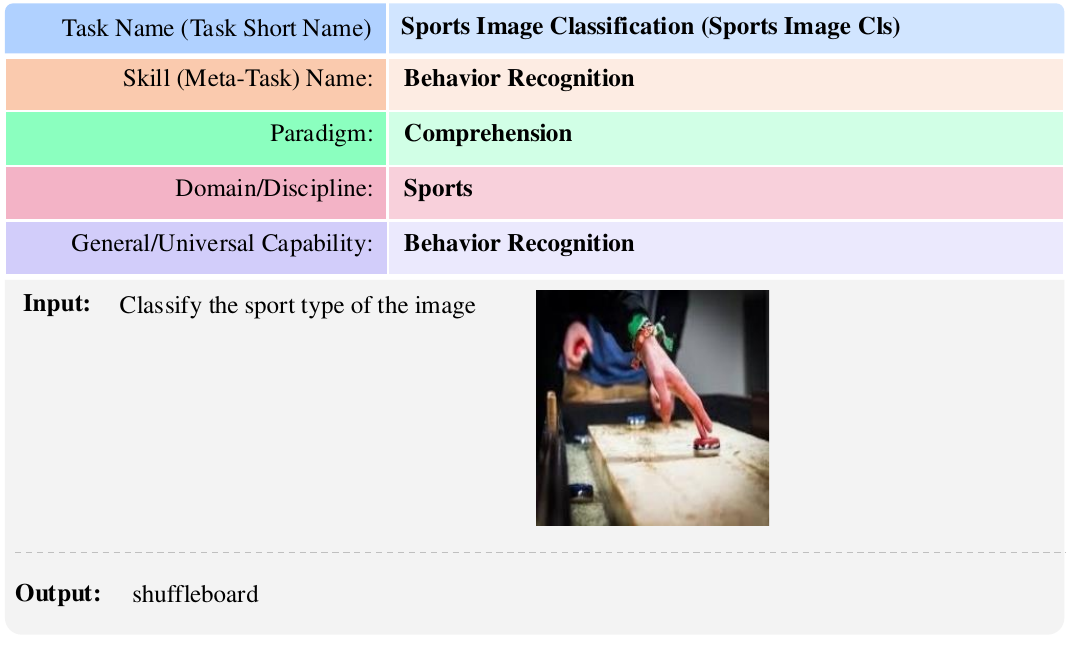}
\caption{Sport Image Classification.}
\label{fig:case-I-C-1}
\vspace{-2mm}
\end{figure*}

\begin{figure*}[!h]
\centering
\includegraphics[width=0.9\linewidth]{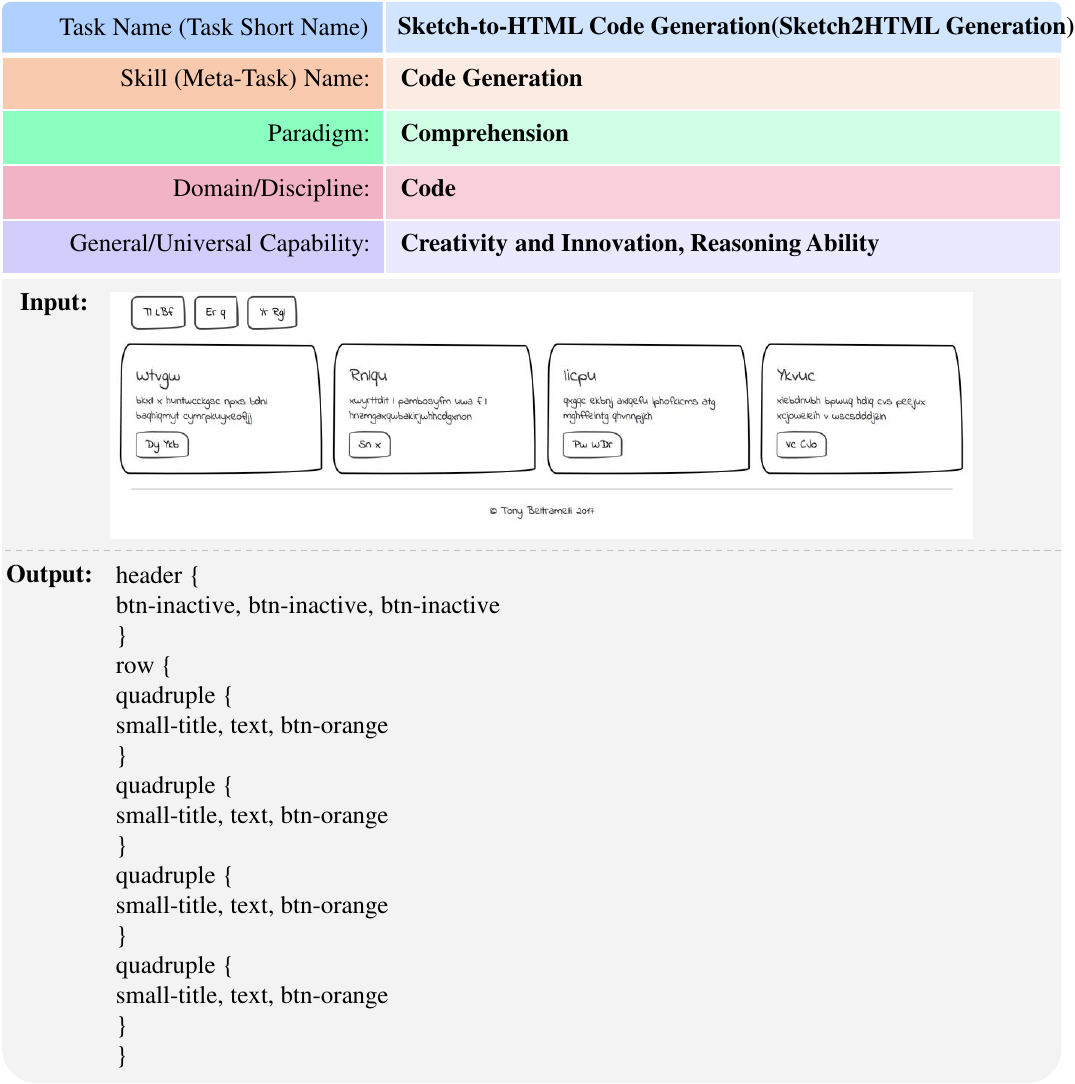}
\caption{Sketch-to-HTML Code Generation.}
\label{fig:case-I-C-2}
\vspace{-2mm}
\end{figure*}

\begin{figure*}[!h]
\centering
\includegraphics[width=0.9\linewidth]{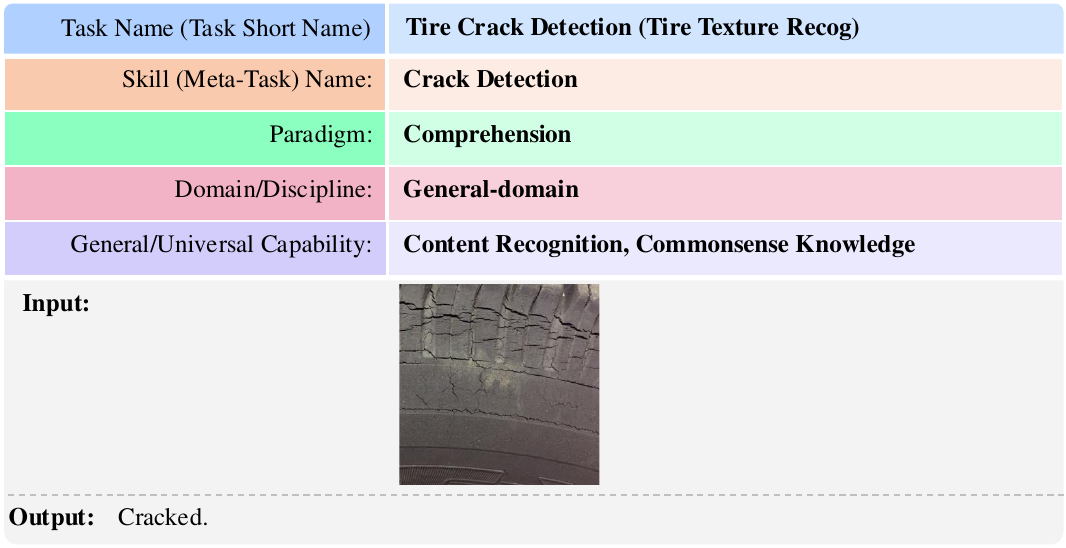}
\caption{Tire Crack Detection.}
\label{fig:case-I-C-3}
\vspace{-2mm}
\end{figure*}

\begin{figure*}[!h]
\centering
\includegraphics[width=0.9\linewidth]{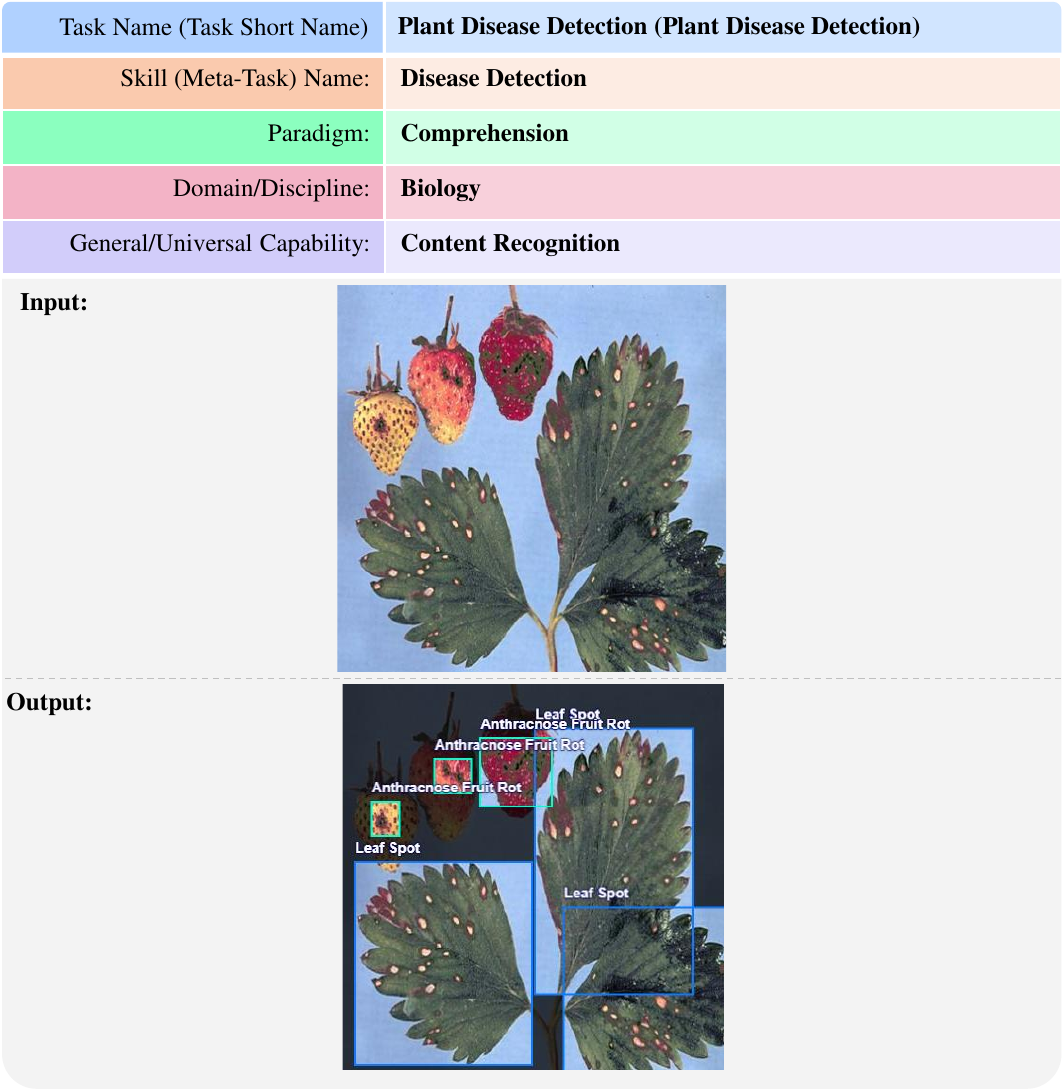}
\caption{Plant Disease Detection.}
\label{fig:case-I-C-4}
\vspace{-2mm}
\end{figure*}

\begin{figure*}[!h]
\centering
\includegraphics[width=0.9\linewidth]{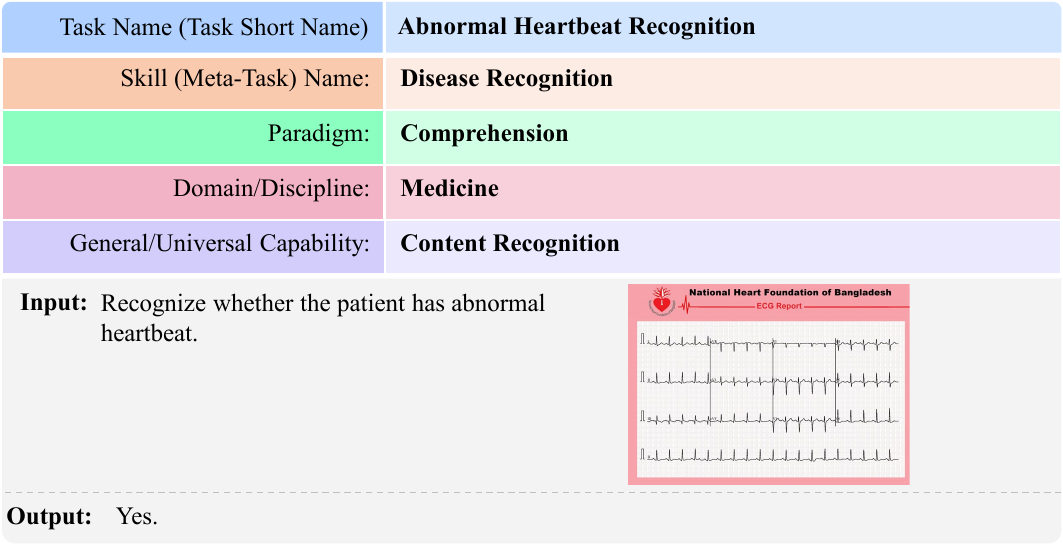}
\caption{Abnormal Heartbeat Recognition.}
\label{fig:case-I-C-5}
\vspace{-2mm}
\end{figure*}

\begin{figure*}[!h]
\centering
\includegraphics[width=0.9\linewidth]{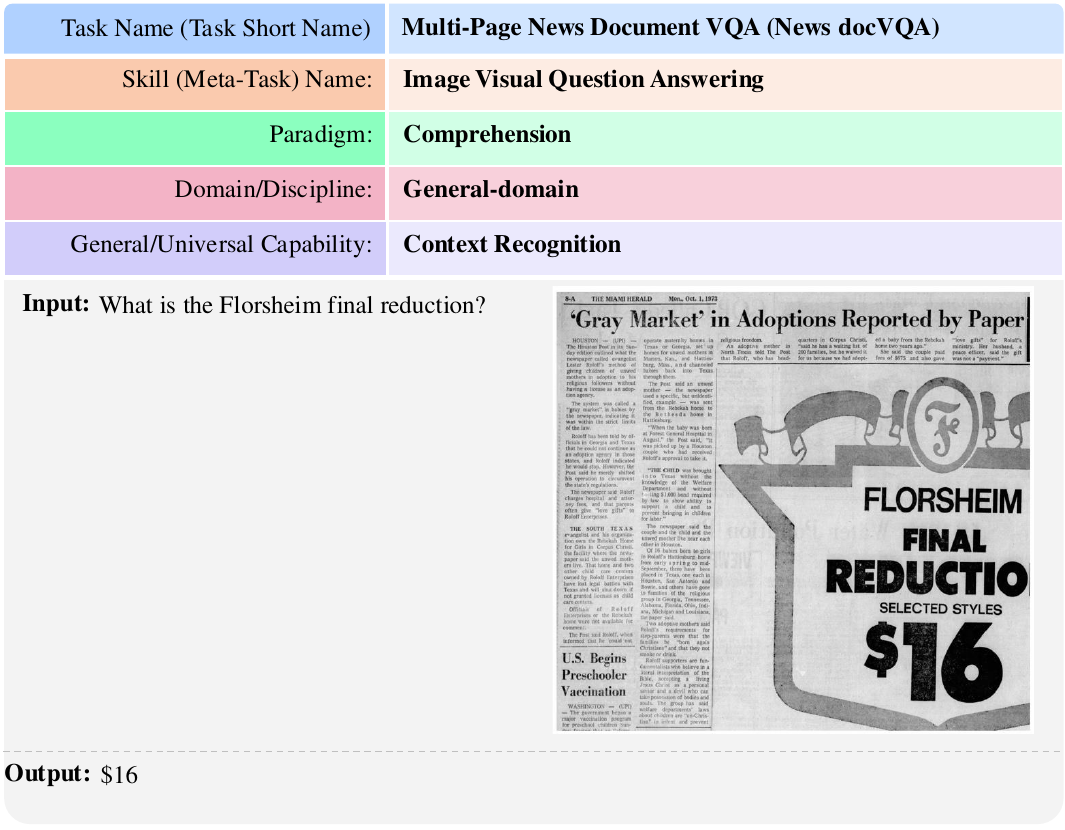}
\caption{Multi-Page News Document VQA.}
\label{fig:case-I-C-6}
\vspace{-2mm}
\end{figure*}

\begin{figure*}[!h]
\centering
\includegraphics[width=0.9\linewidth]{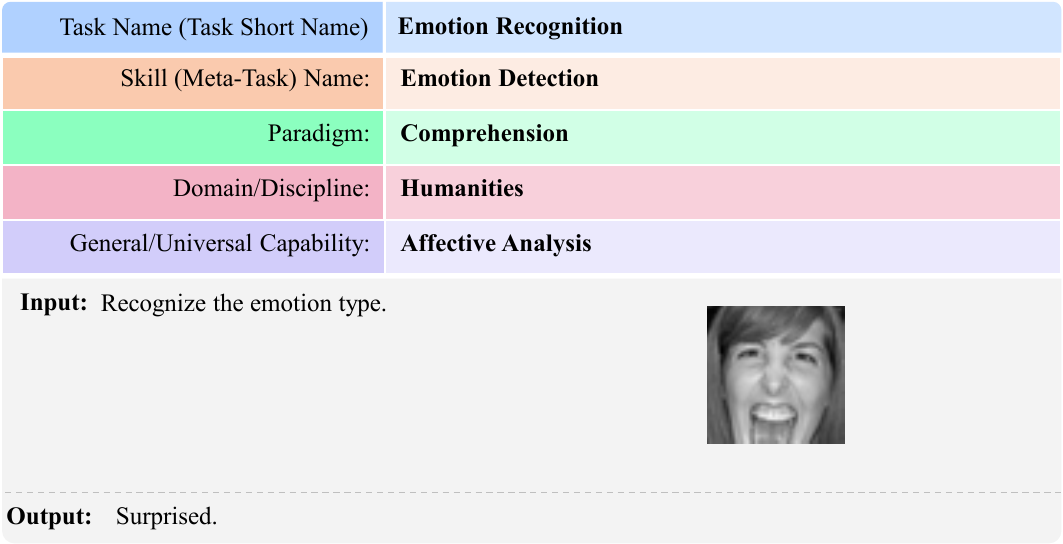}
\caption{Emotion Recognition.}
\label{fig:case-I-C-7}
\vspace{-2mm}
\end{figure*}

\begin{figure*}[!h]
\centering
\includegraphics[width=0.9\linewidth]{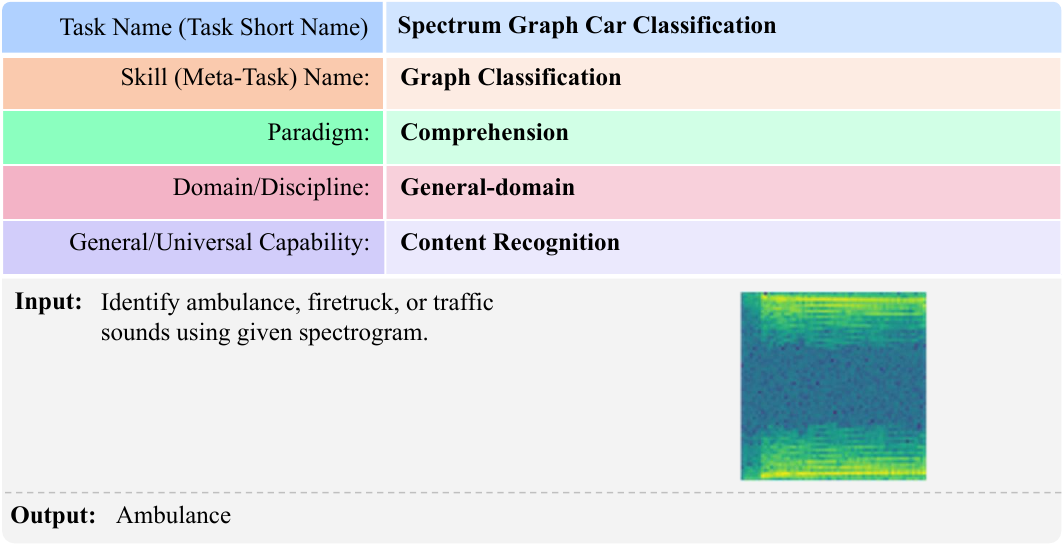}
\caption{Spectrum Graph Car Classification.
}
\label{fig:case-I-C-8}
\vspace{-2mm}
\end{figure*}

\begin{figure*}[!h]
\centering
\includegraphics[width=0.9\linewidth]{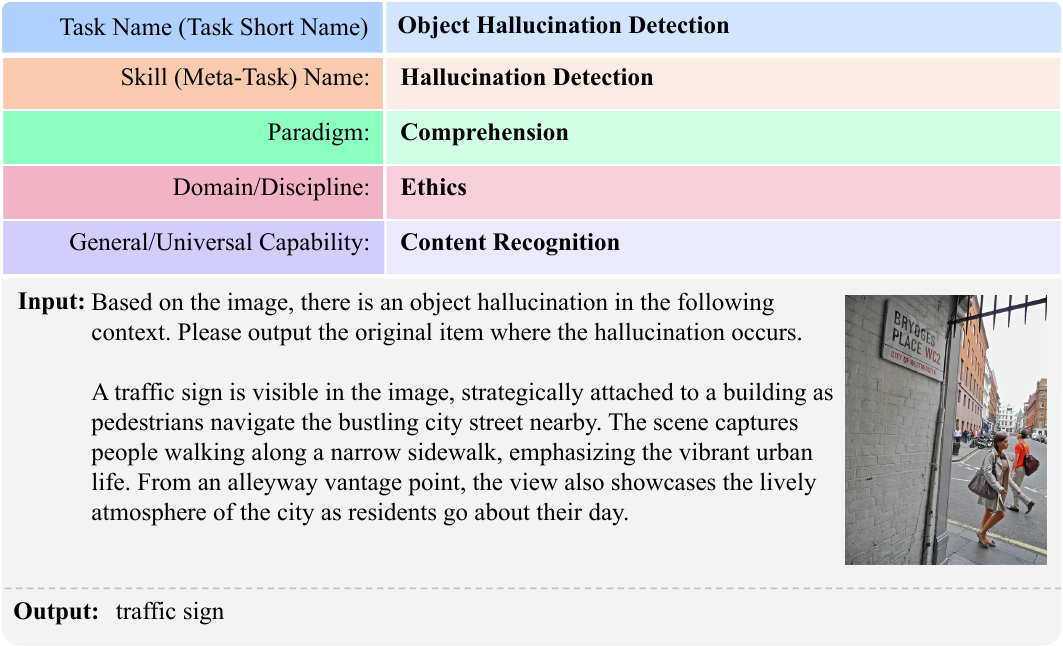}
\caption{Object Hallucination Detection}
\label{fig:case-I-C-9}
\vspace{-2mm}
\end{figure*}

\begin{figure*}[!h]
\centering
\includegraphics[width=0.9\linewidth]{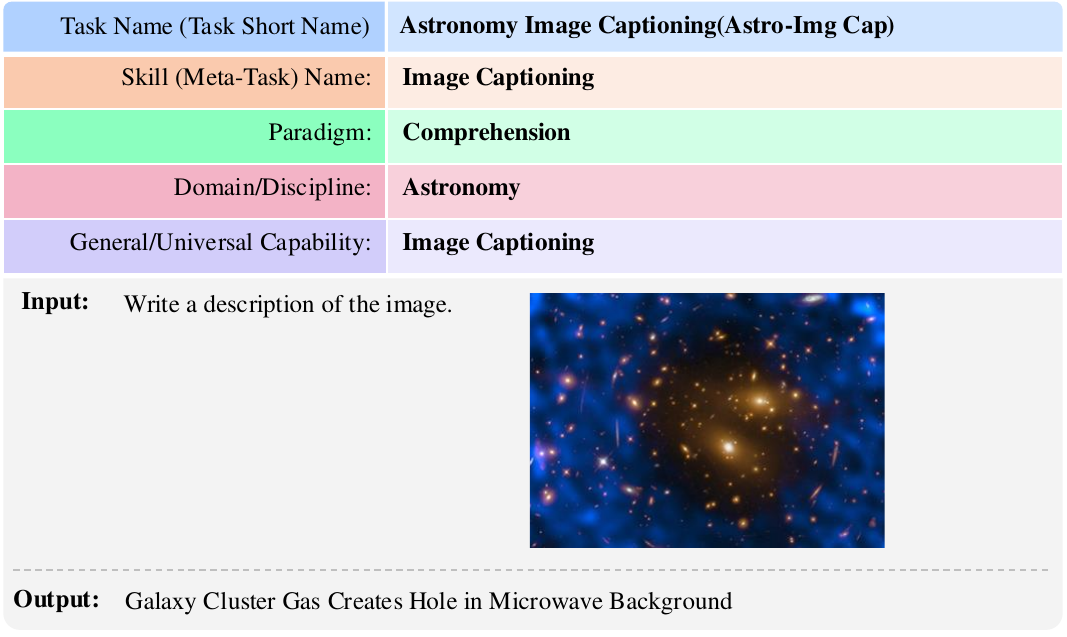}
\caption{Astronomy Image Captioning.}
\label{fig:case-I-C-10}
\vspace{-2mm}
\end{figure*}

\begin{figure*}[!h]
\centering
\includegraphics[width=0.9\linewidth]{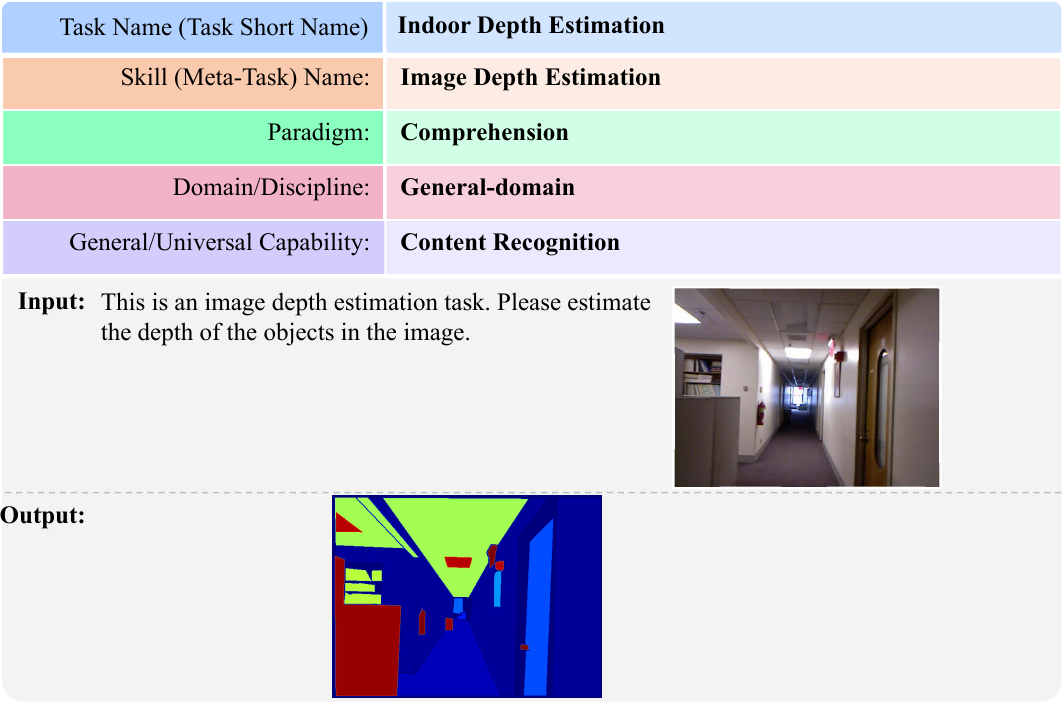}
\caption{Image Depth Estimation.}
\label{fig:case-I-C-11}
\vspace{-2mm}
\end{figure*}

\begin{figure*}[!h]
\centering
\includegraphics[width=0.9\linewidth]{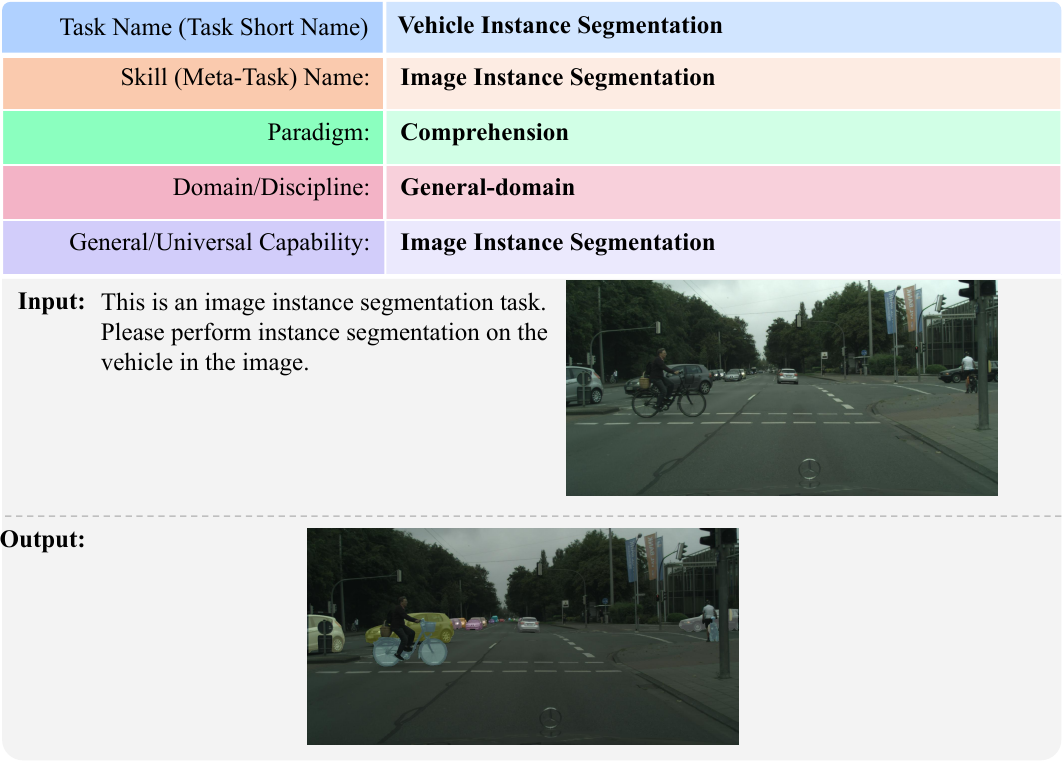}
\caption{Image Instance Segmentation.}
\label{fig:case-I-C-12}
\vspace{-2mm}
\end{figure*}

\begin{figure*}[!h]
\centering
\includegraphics[width=0.9\linewidth]{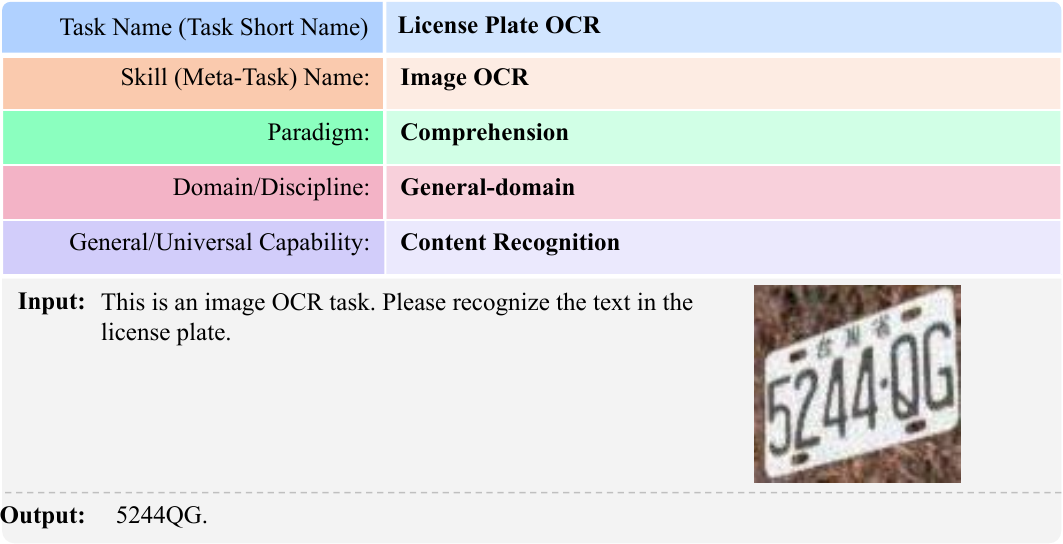}
\caption{License Plate ORC.}
\label{fig:case-I-C-13}
\vspace{-2mm}
\end{figure*}

\begin{figure*}[!h]
\centering
\includegraphics[width=0.9\linewidth]{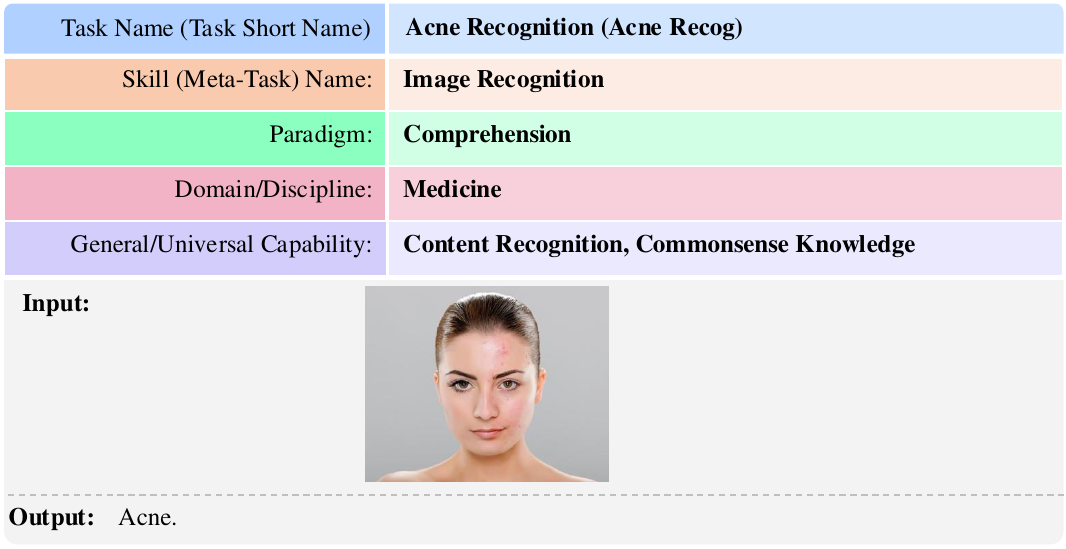}
\caption{Acne Recognition.}
\label{fig:case-I-C-14}
\vspace{-2mm}
\end{figure*}

\begin{figure*}[!h]
\centering
\includegraphics[width=0.9\linewidth]{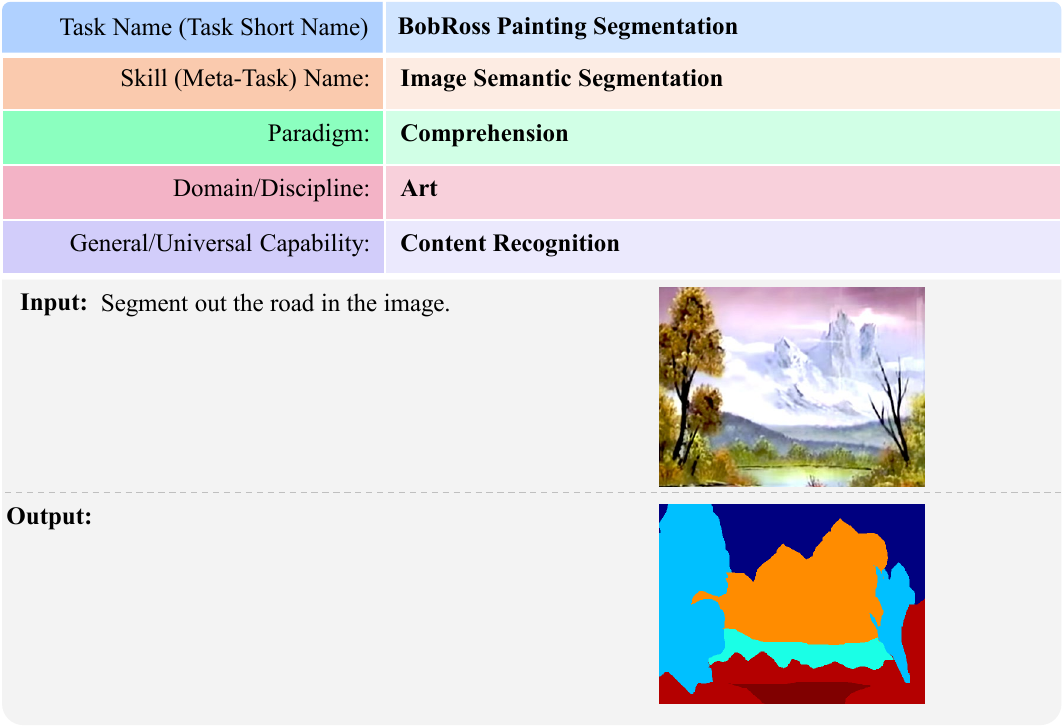}
\caption{BobRoss Painting Segmentation.}
\label{fig:case-I-C-15}
\vspace{-2mm}
\end{figure*}

\begin{figure*}[!h]
\centering
\includegraphics[width=0.9\linewidth]{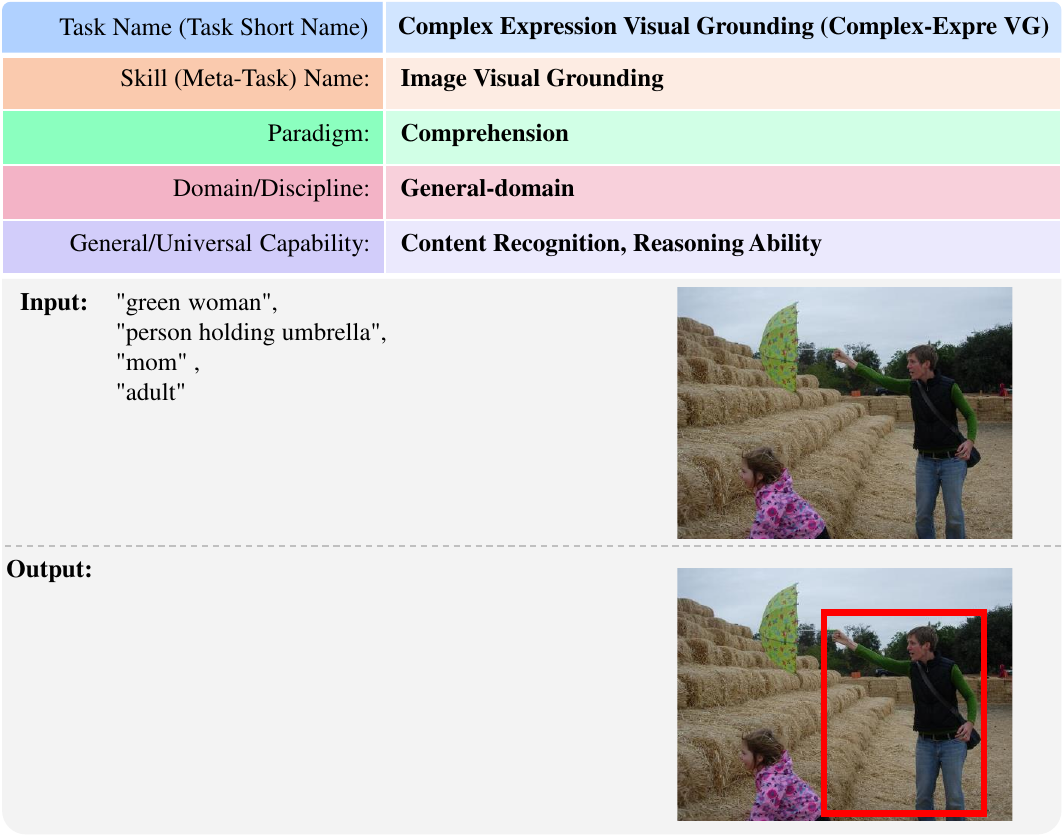}
\caption{Complex Expression Visual Grounding.}
\label{fig:case-I-C-16}
\vspace{-2mm}
\end{figure*}

\begin{figure*}[!h]
\centering
\includegraphics[width=0.9\linewidth]{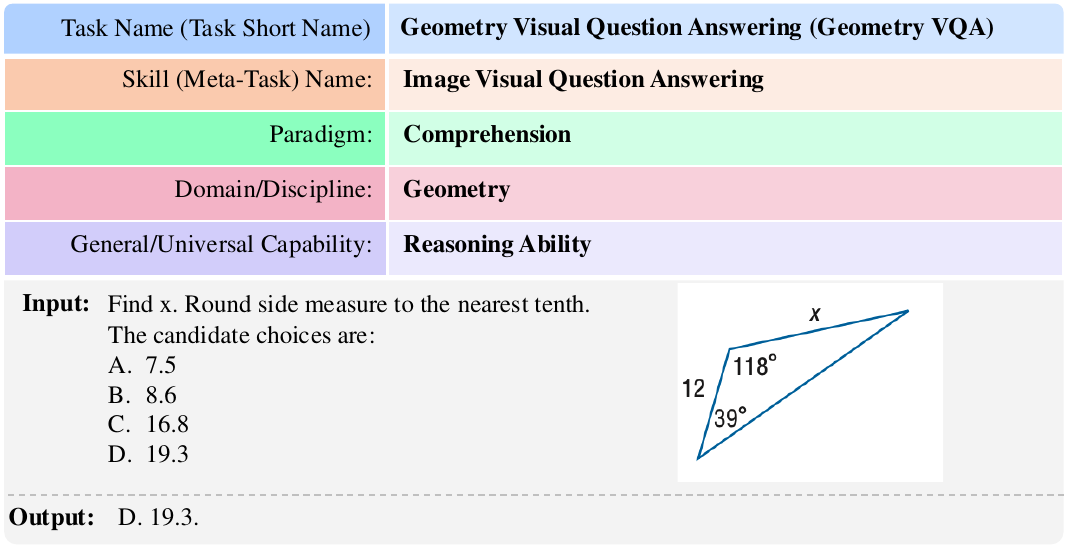}
\caption{Geometry Visual Question Answering.}
\label{fig:case-I-C-17-1}
\vspace{-2mm}
\end{figure*}

\begin{figure*}[!h]
\centering
\includegraphics[width=0.9\linewidth]{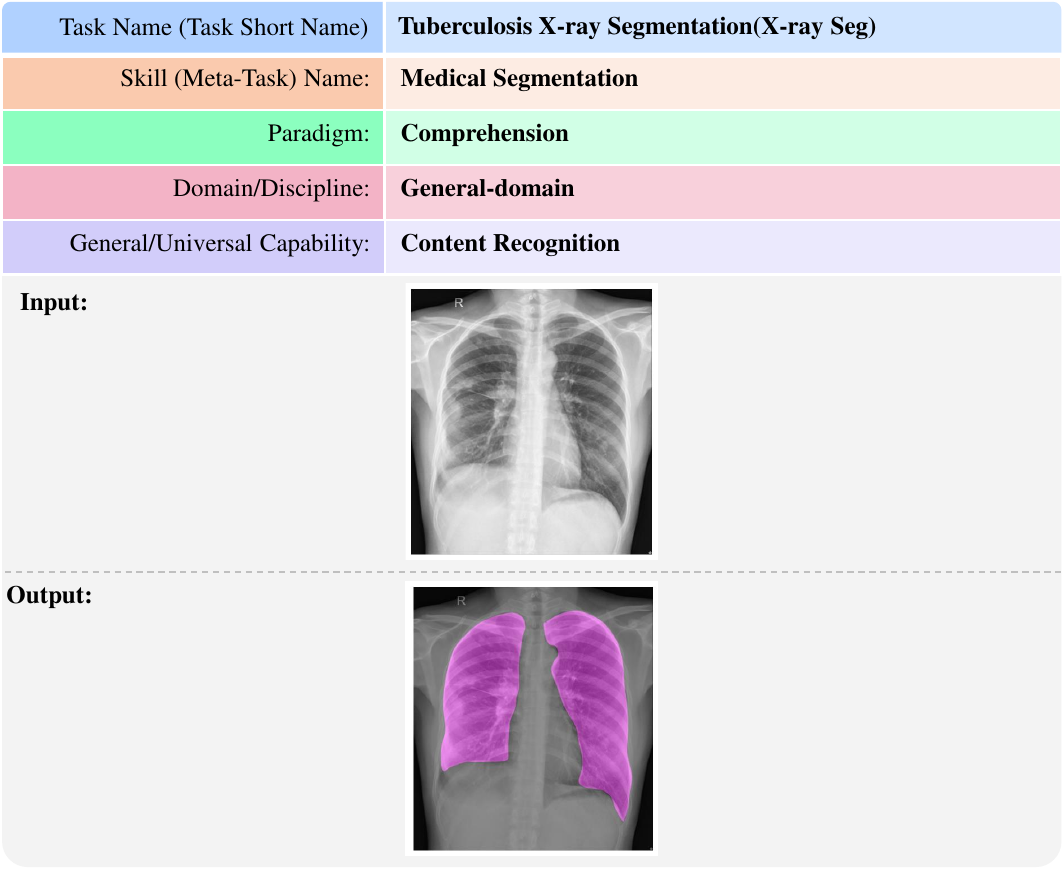}
\caption{Tuberculosis X-ray Segmentation.}
\label{fig:case-I-C-19-1}
\vspace{-2mm}
\end{figure*}

\begin{figure*}[!h]
\centering
\includegraphics[width=0.9\linewidth]{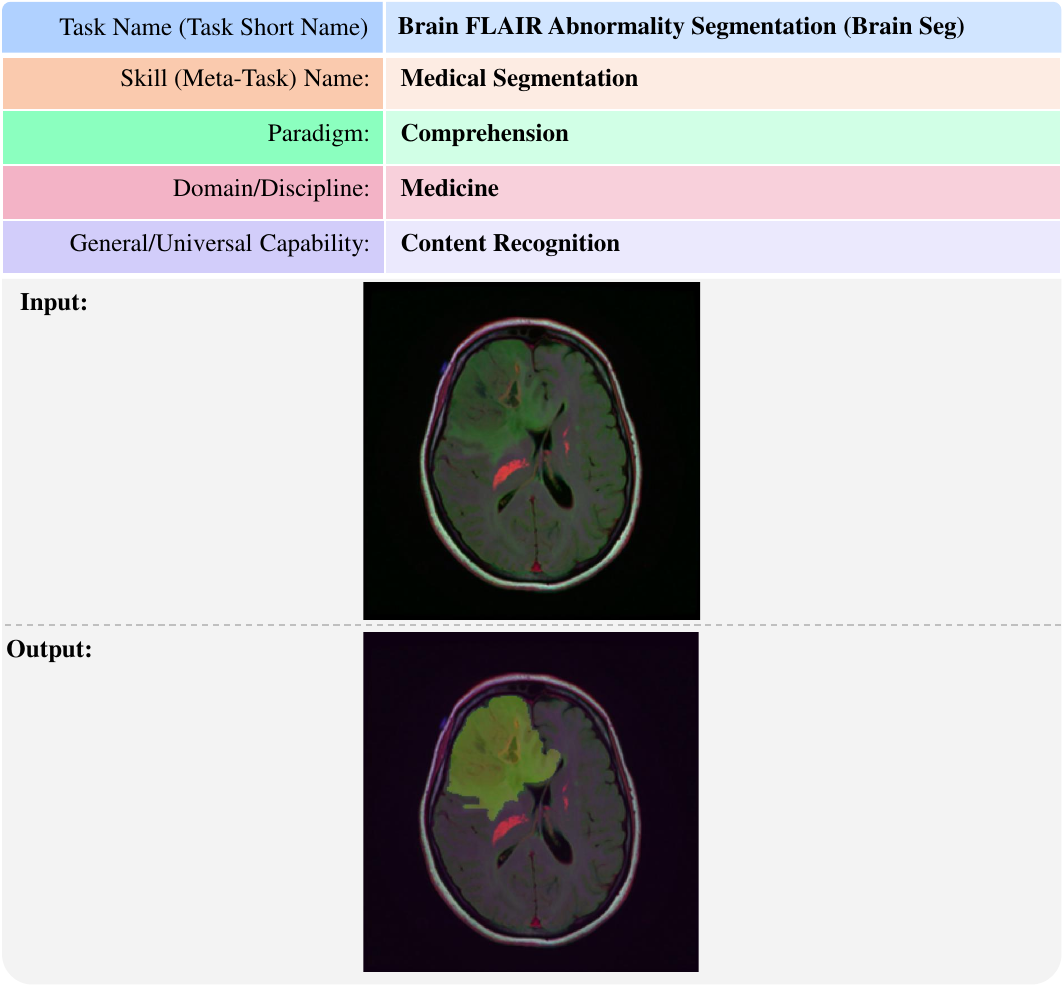}
\caption{Brain FLAIR Abnormality Segmentation.}
\label{fig:case-I-C-19-2}
\vspace{-2mm}
\end{figure*}

\begin{figure*}[!h]
\centering
\includegraphics[width=0.9\linewidth]{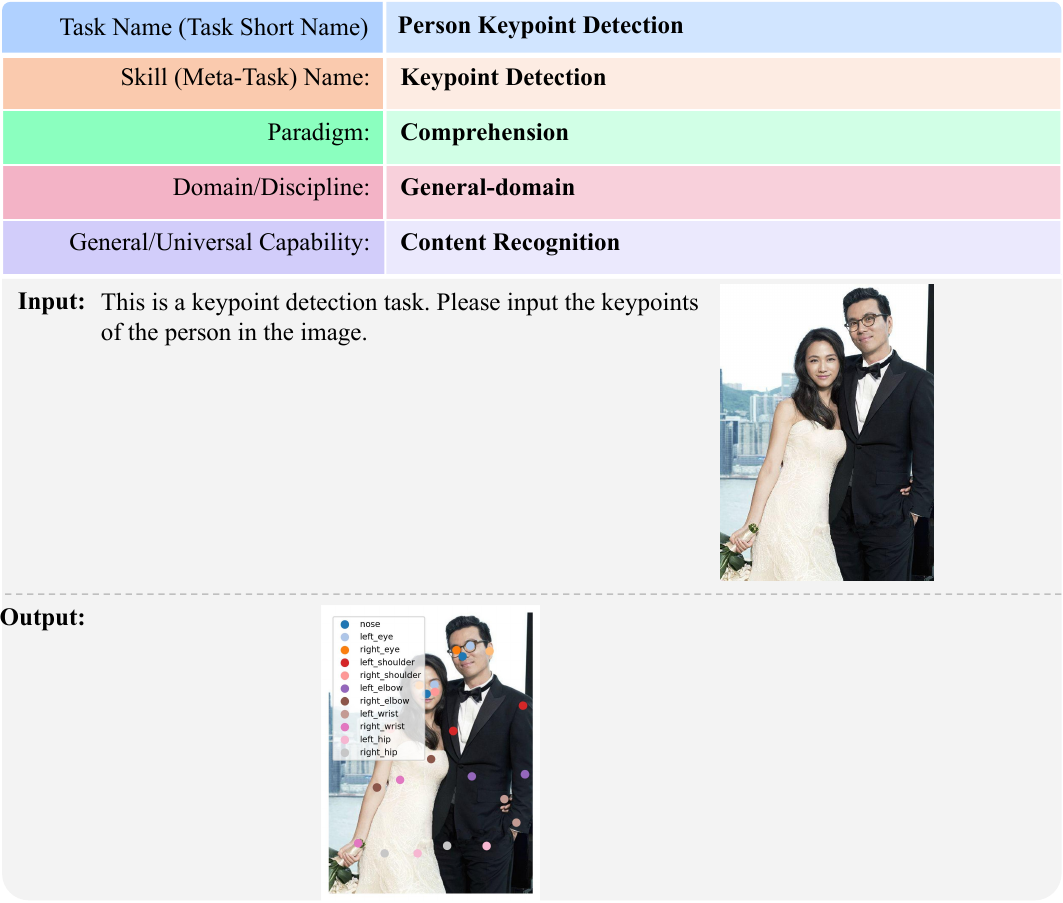}
\caption{Person Keypoint Detection.}
\label{fig:case-I-C-20}
\vspace{-2mm}
\end{figure*}

\begin{figure*}[!h]
\centering
\includegraphics[width=0.9\linewidth]{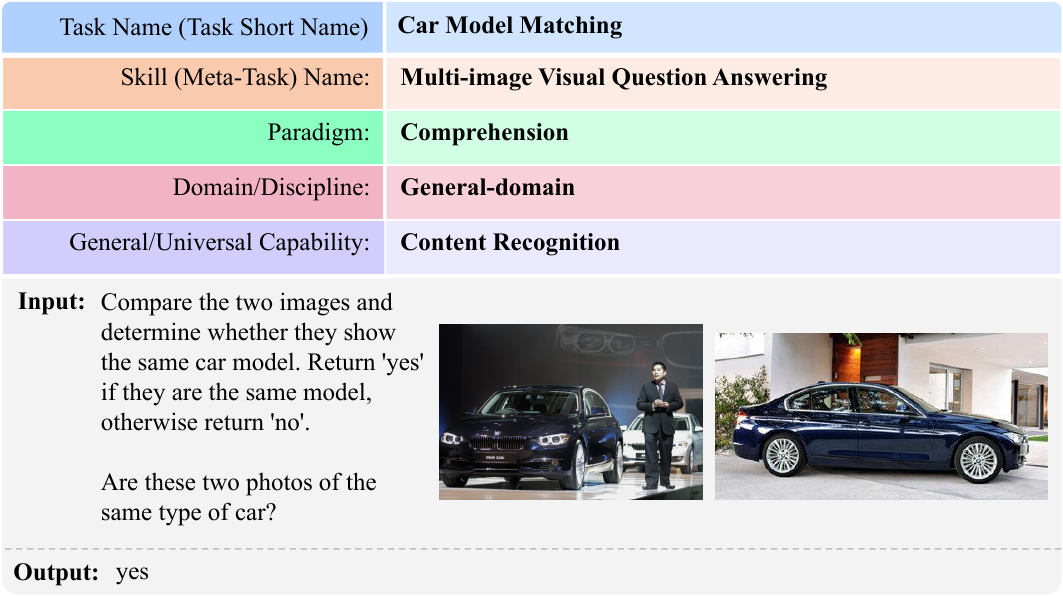}
\caption{Car Model Matching.}
\label{fig:case-I-C-21}
\vspace{-2mm}
\end{figure*}

\begin{figure*}[!h]
\centering
\includegraphics[width=0.9\linewidth]{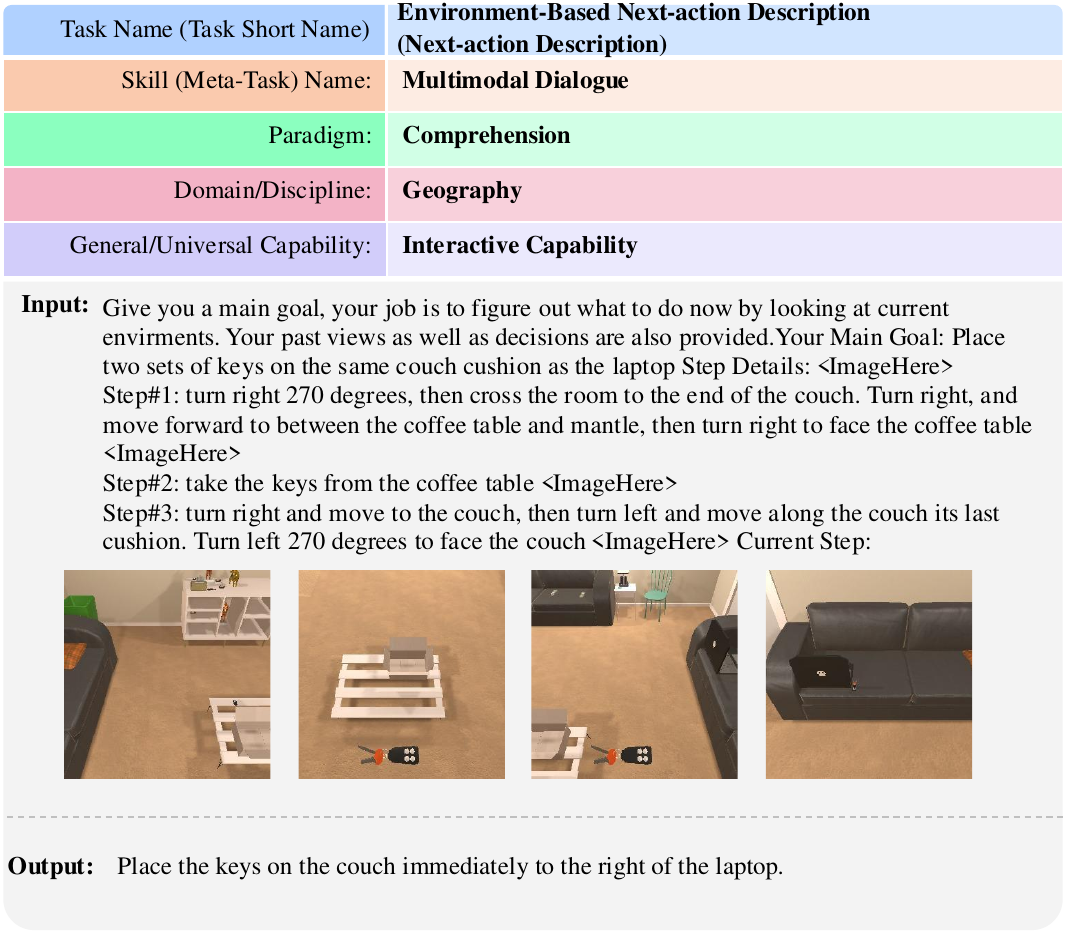}
\caption{Environment-based Next-action Description.}
\label{fig:case-I-C-22}
\vspace{-2mm}
\end{figure*}

\begin{figure*}[!h]
\centering
\includegraphics[width=0.9\linewidth]{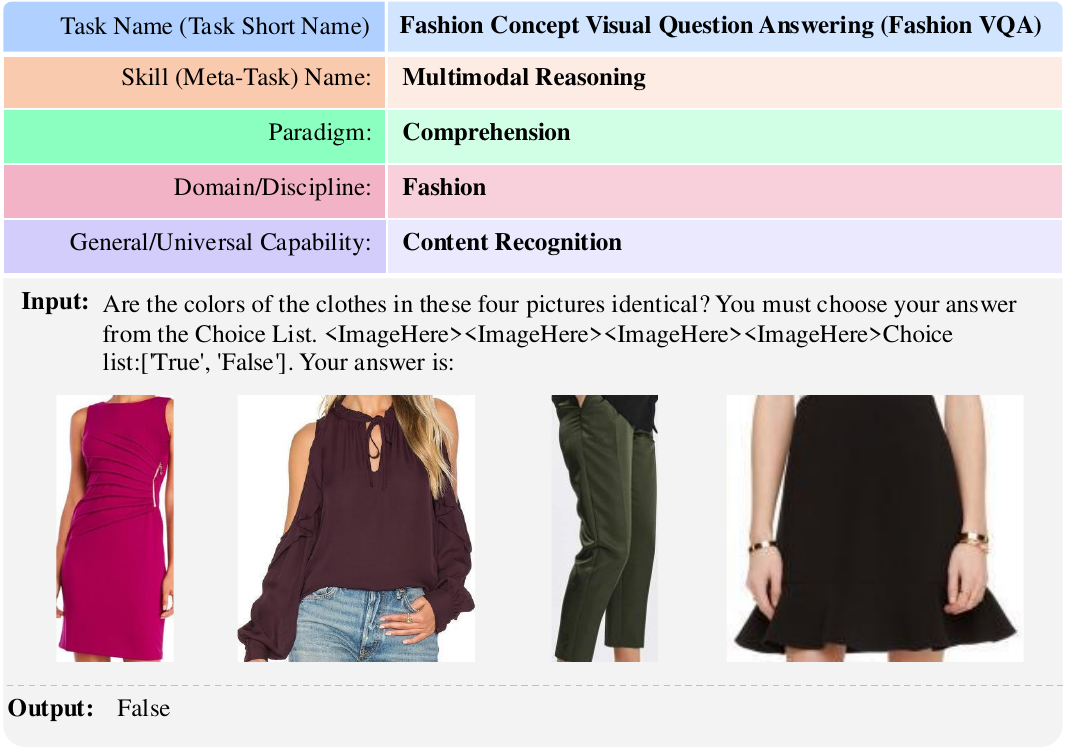}
\caption{Fashion Concept Visual Question Answering.}
\label{fig:case-I-C-23}
\vspace{-2mm}
\end{figure*}

\begin{figure*}[!h]
\centering
\includegraphics[width=0.9\linewidth]{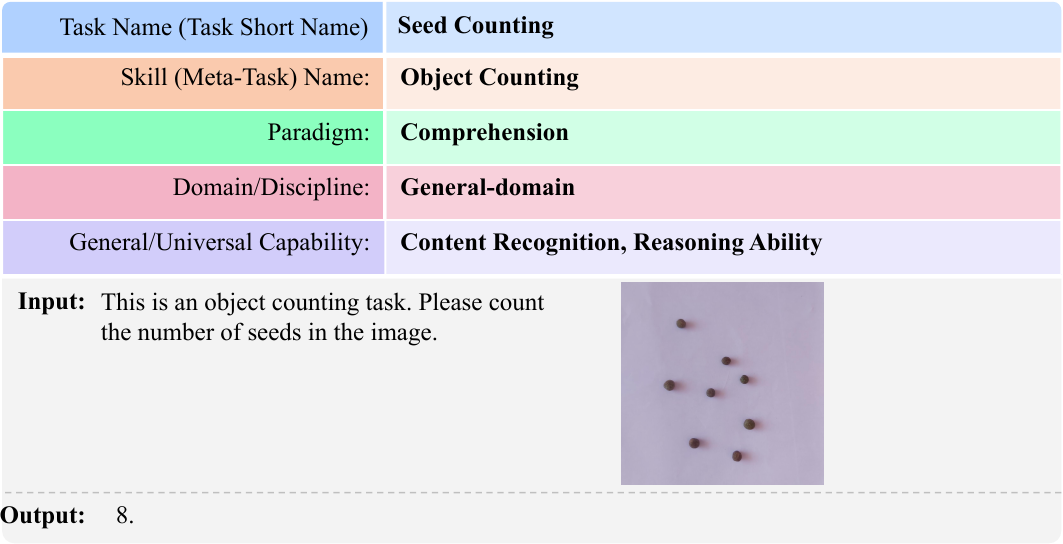}
\caption{Seed Counting.}
\label{fig:case-I-C-24}
\vspace{-2mm}
\end{figure*}

\begin{figure*}[!h]
\centering
\includegraphics[width=0.9\linewidth]{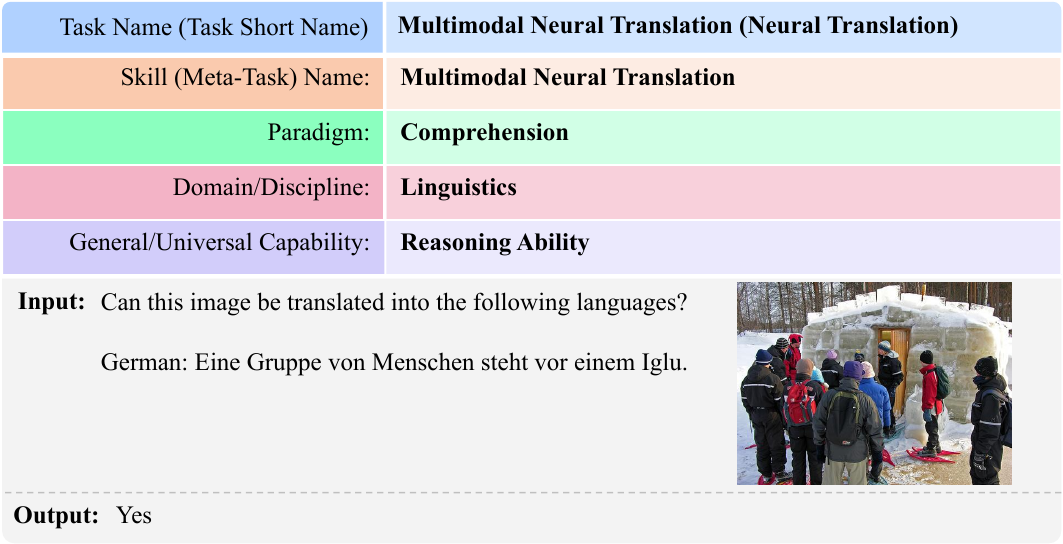}
\caption{Multimodal Neural Translation.}
\label{fig:case-I-C-25}
\vspace{-2mm}
\end{figure*}

\begin{figure*}[!h]
\centering
\includegraphics[width=0.9\linewidth]{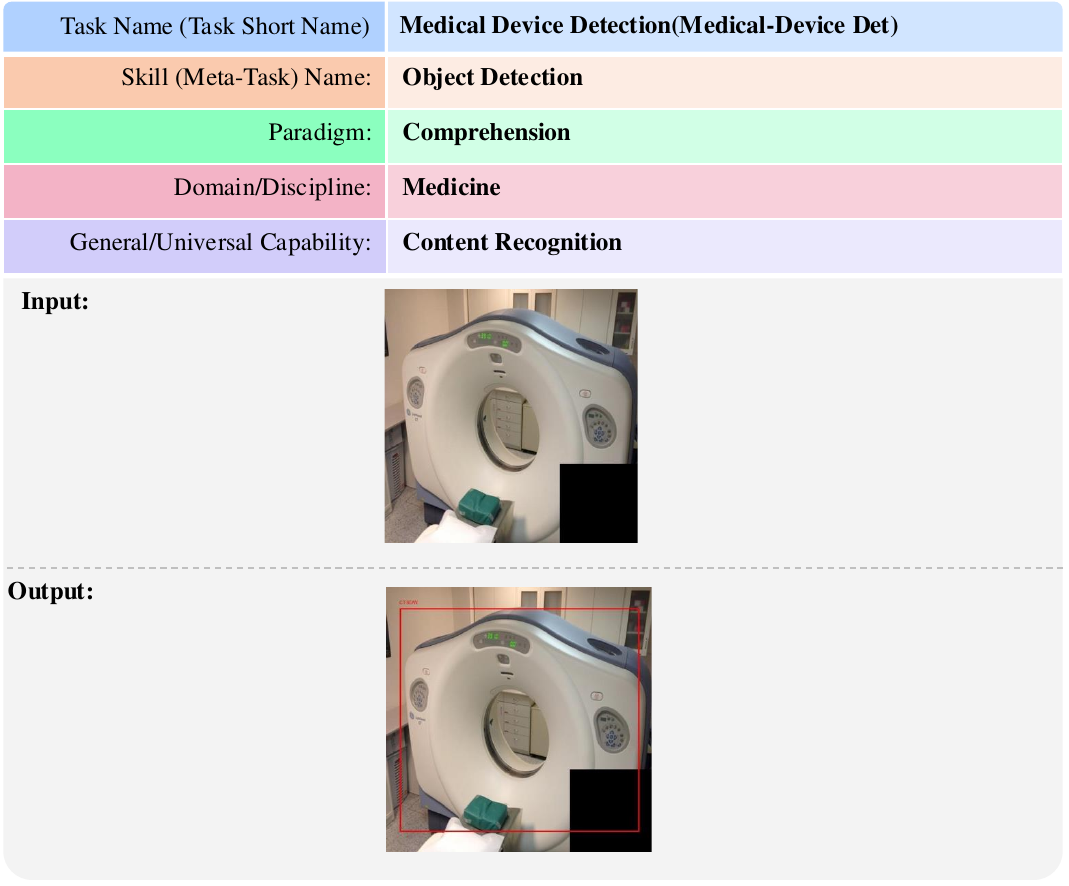}
\caption{Medical Device Detection.}
\label{fig:case-I-C-26}
\vspace{-2mm}
\end{figure*}

\begin{figure*}[!h]
\centering
\includegraphics[width=0.9\linewidth]{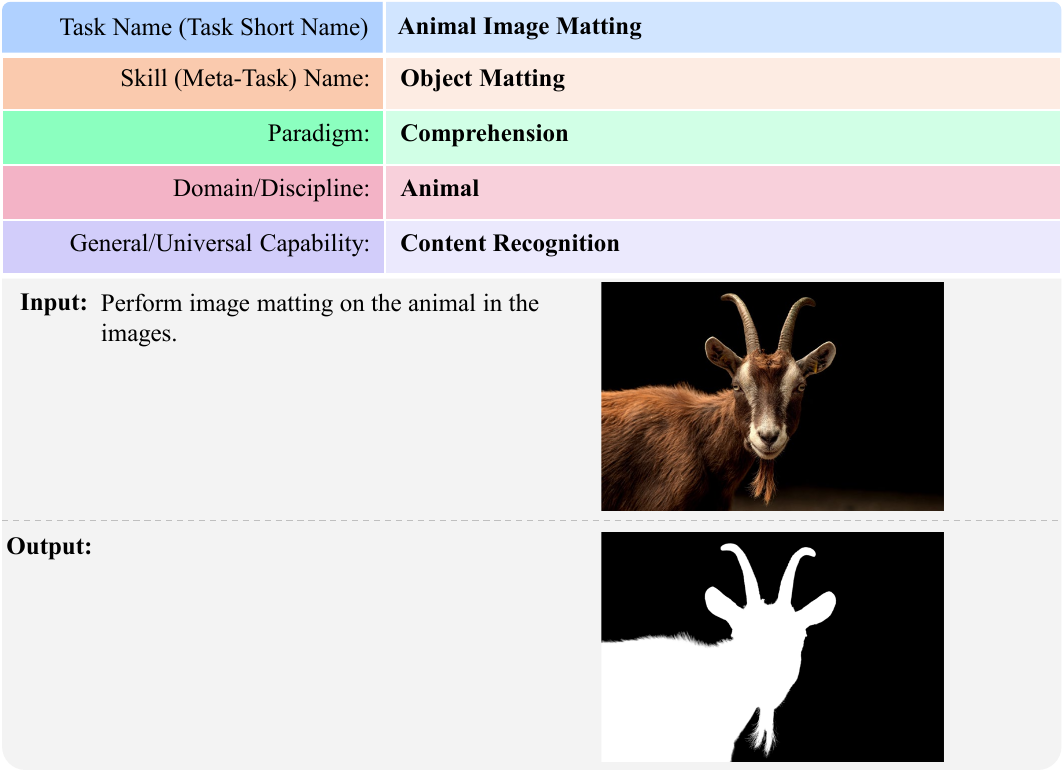}
\caption{Animal Image Matting.}
\label{fig:case-I-C-27}
\vspace{-2mm}
\end{figure*}

\begin{figure*}[!h]
\centering
\includegraphics[width=0.9\linewidth]{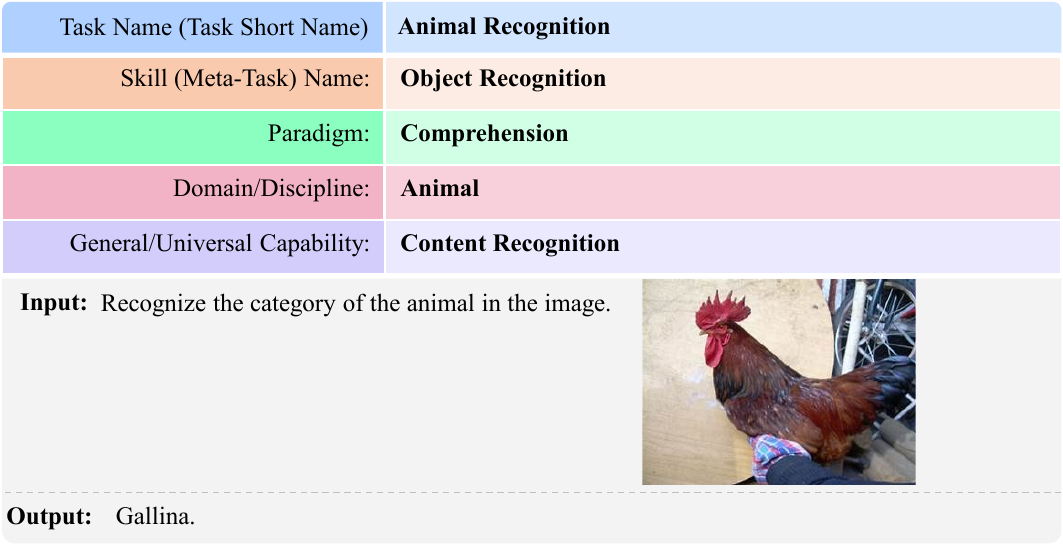}
\caption{Animal Recognition.}
\label{fig:case-I-C-28}
\vspace{-2mm}
\end{figure*}

\begin{figure*}[!h]
\centering
\includegraphics[width=0.9\linewidth]{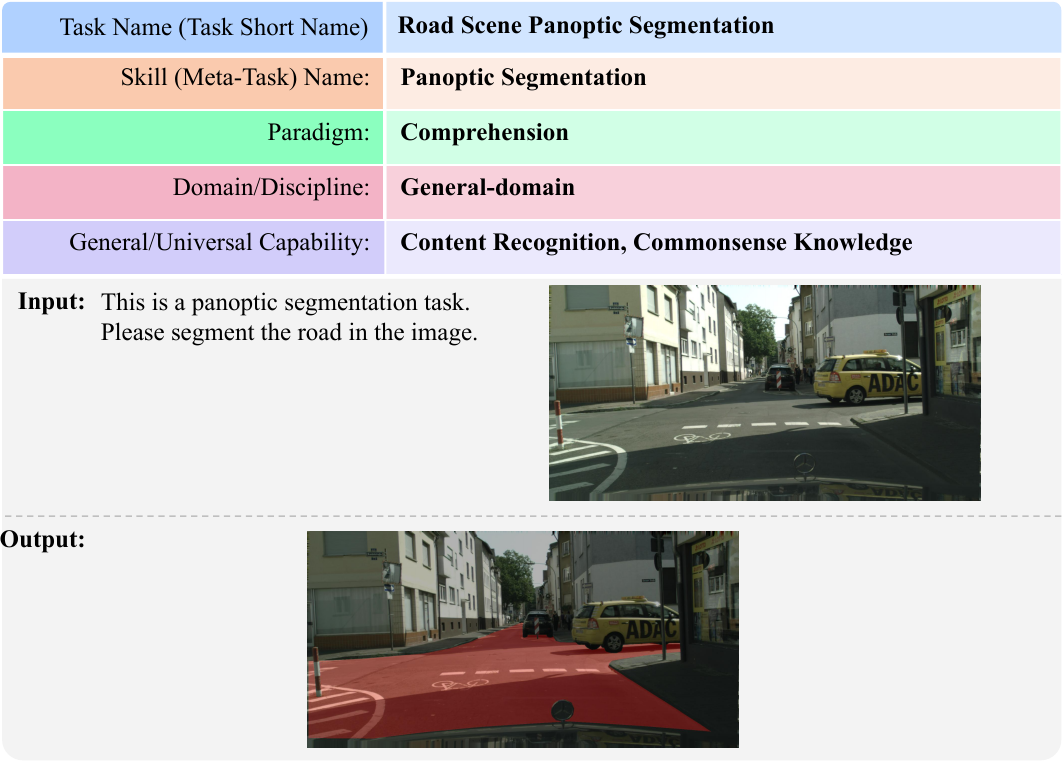}
\caption{Road Scene Panoptic Segmentation.}
\label{fig:case-I-C-29}
\vspace{-2mm}
\end{figure*}

\begin{figure*}[!h]
\centering
\includegraphics[width=0.9\linewidth]{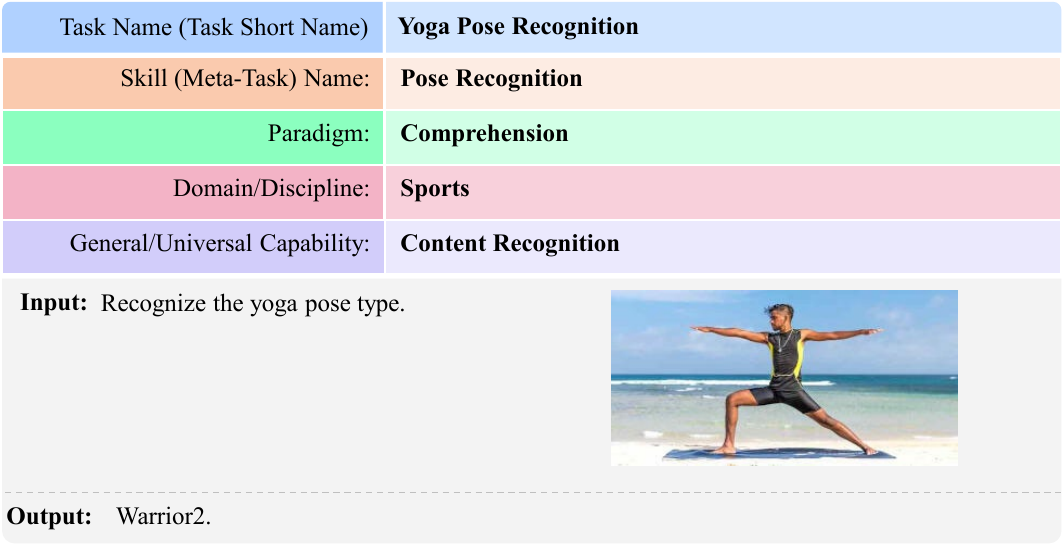}
\caption{Yoga Pose Recognition.}
\label{fig:case-I-C-30}
\vspace{-2mm}
\end{figure*}

\begin{figure*}[!h]
\centering
\includegraphics[width=0.9\linewidth]{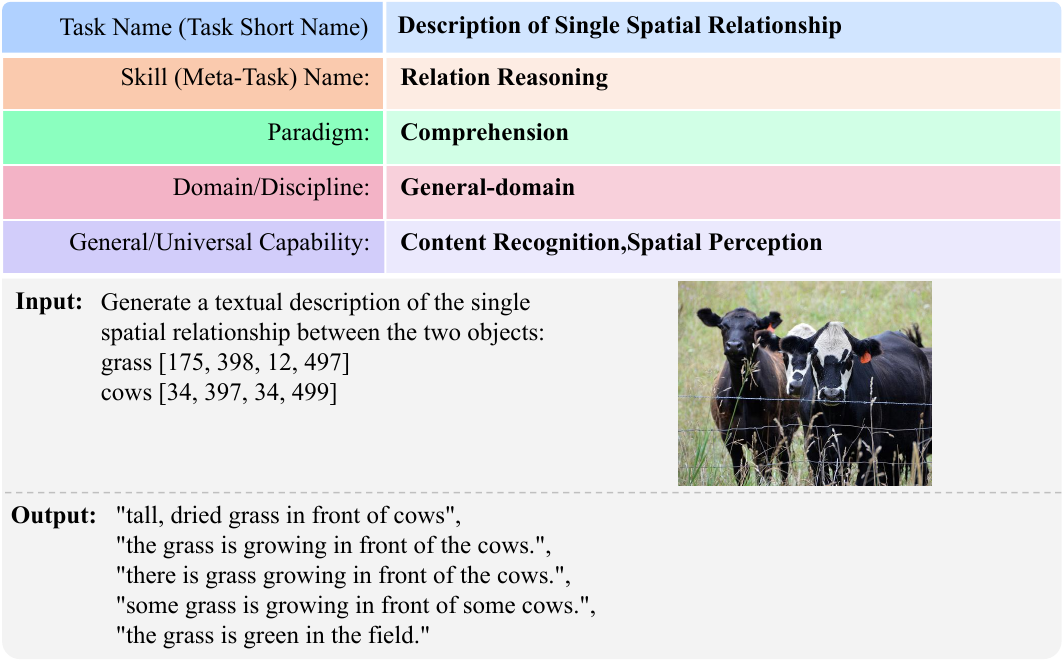}
\caption{Description of Single Spatial Relationship.}
\label{fig:case-I-C-31}
\vspace{-2mm}
\end{figure*}

\begin{figure*}[!h]
\centering
\includegraphics[width=0.9\linewidth]{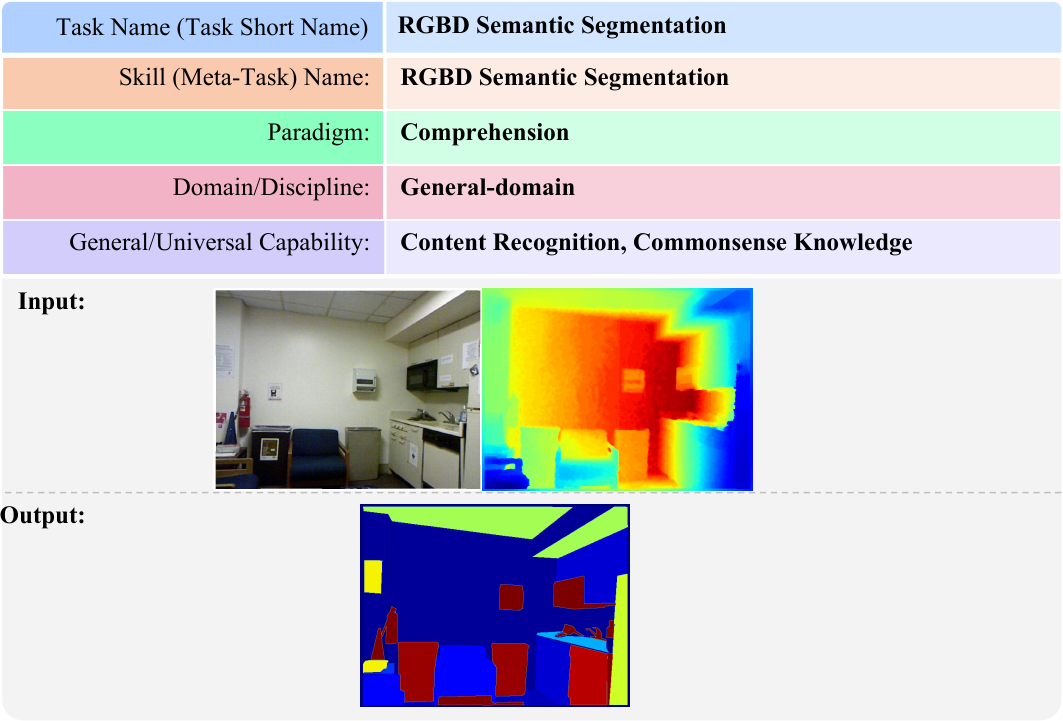}
\caption{RGBD Semantic Segmentation.}
\label{fig:case-I-C-32}
\vspace{-2mm}
\end{figure*}

\begin{figure*}[!h]
\centering
\includegraphics[width=0.9\linewidth]{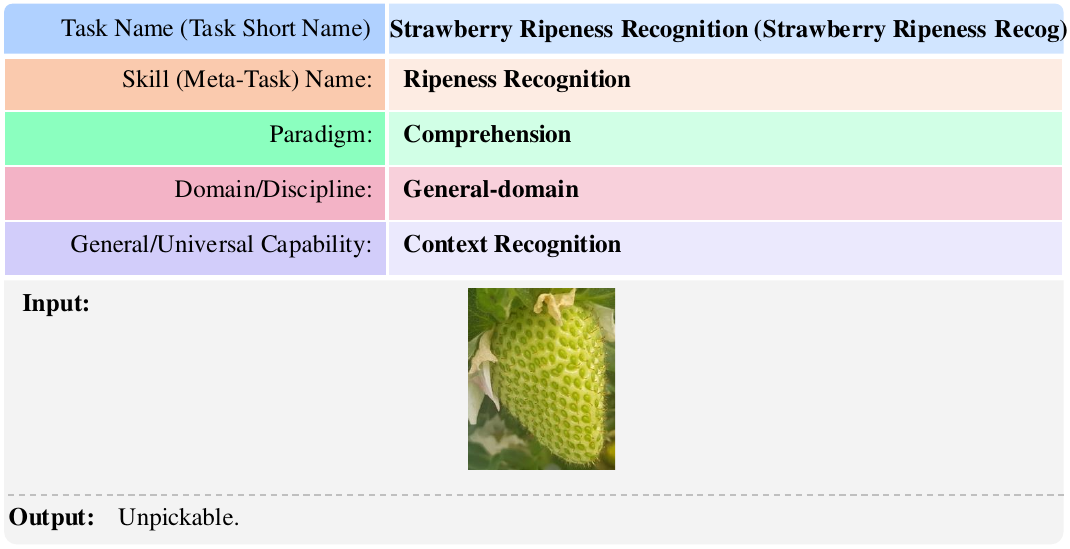}
\caption{Strawberry Ripeness Recognition.}
\label{fig:case-I-C-33}
\vspace{-2mm}
\end{figure*}

\begin{figure*}[!h]
\centering
\includegraphics[width=0.9\linewidth]{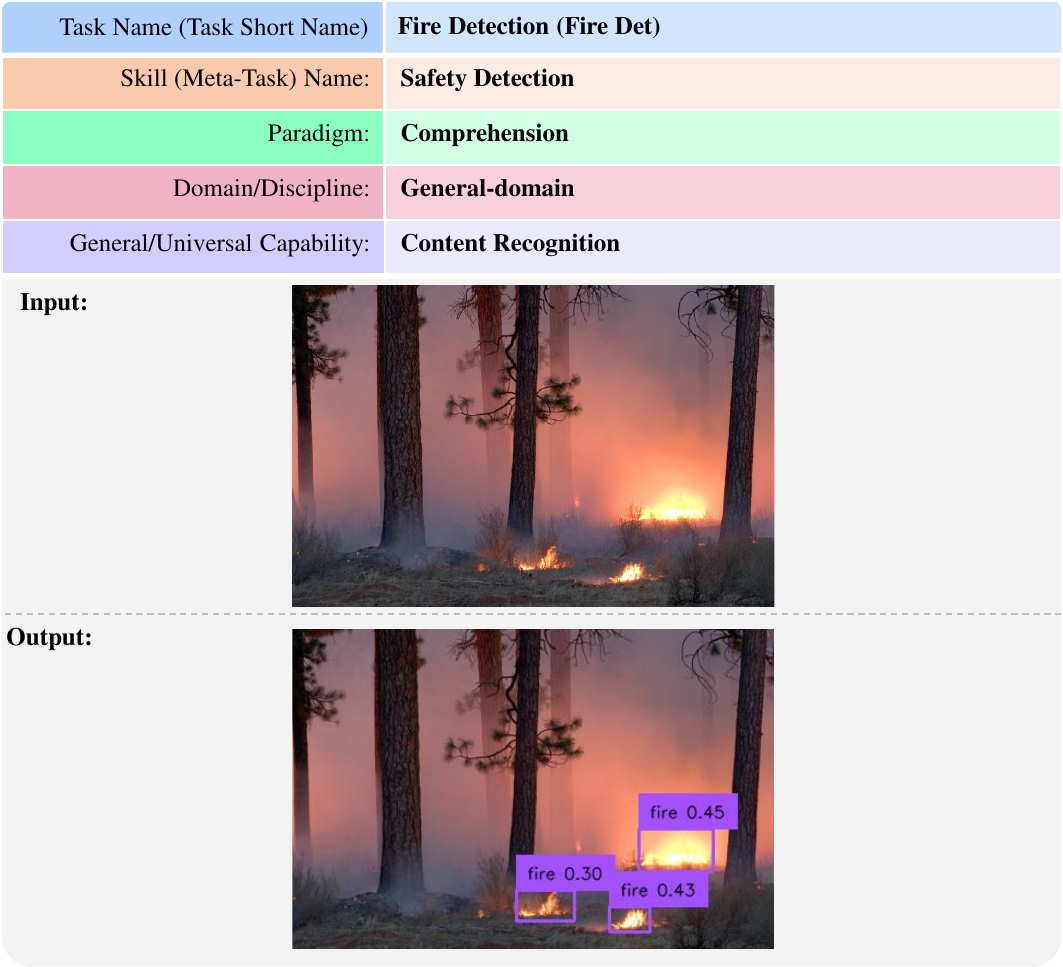}
\caption{Fire Detection.}
\label{fig:case-I-C-34}
\vspace{-2mm}
\end{figure*}

\begin{figure*}[!h]
\centering
\includegraphics[width=0.9\linewidth]{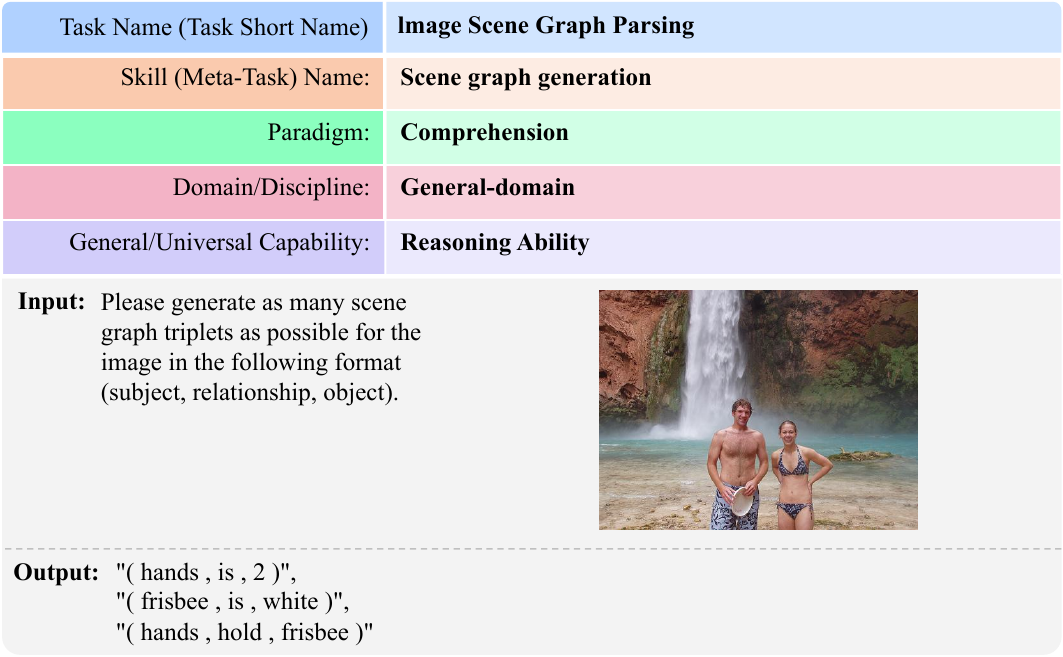}
\caption{Image Scene Graph Parsing.}
\label{fig:case-I-C-35}
\vspace{-2mm}
\end{figure*}

\begin{figure*}[!h]
\centering
\includegraphics[width=0.9\linewidth]{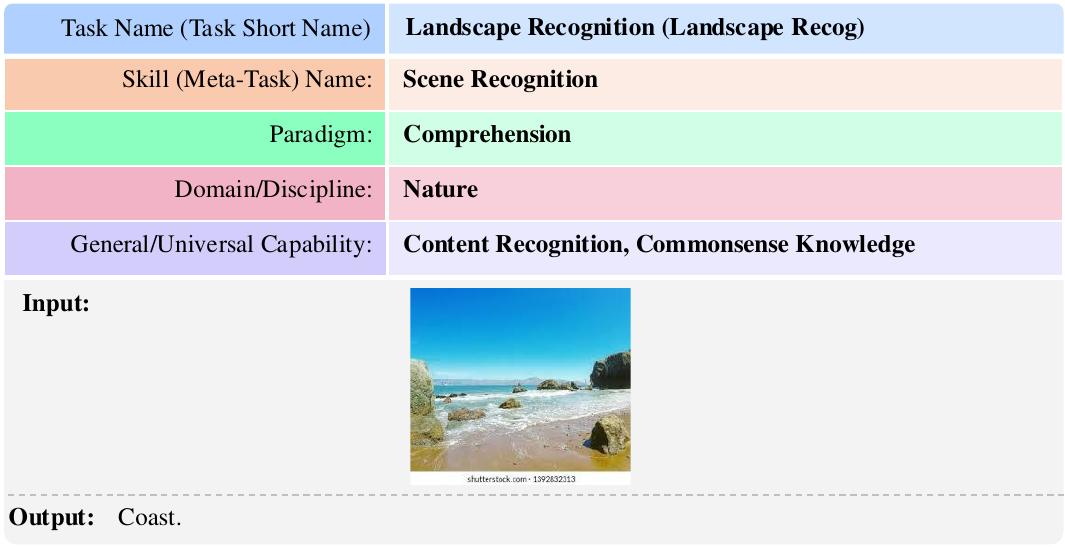}
\caption{Landscape Recognition.}
\label{fig:case-I-C-36}
\vspace{-2mm}
\end{figure*}

\begin{figure*}[!h]
\centering
\includegraphics[width=0.9\linewidth]{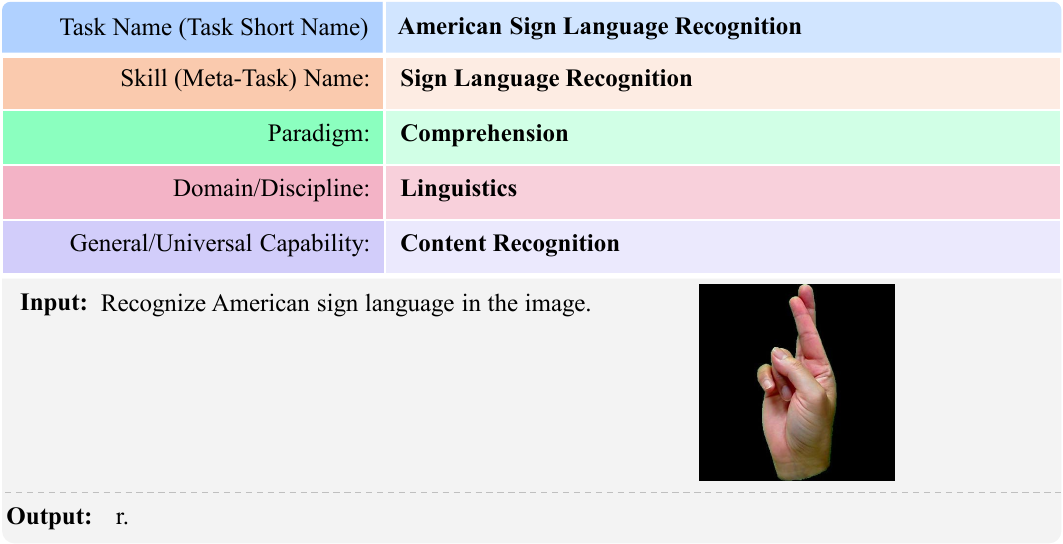}
\caption{American Sign Language Recognition.}
\label{fig:case-I-C-37}
\vspace{-2mm}
\end{figure*}

\begin{figure*}[!h]
\centering
\includegraphics[width=0.9\linewidth]{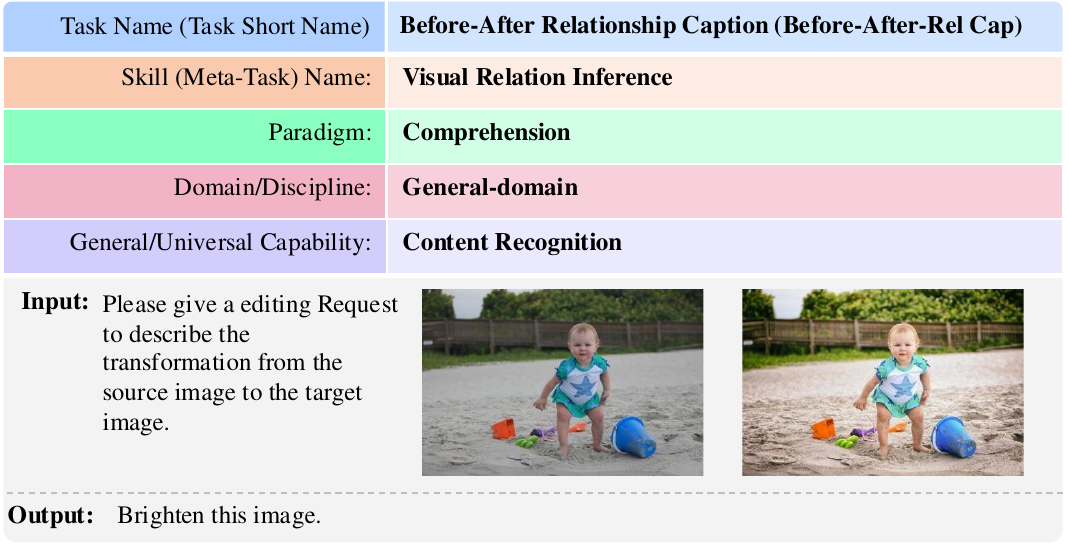}
\caption{Before-After Relationship Caption.}
\label{fig:case-I-C-38}
\vspace{-2mm}
\end{figure*}

\begin{figure*}[!h]
\centering
\includegraphics[width=0.9\linewidth]{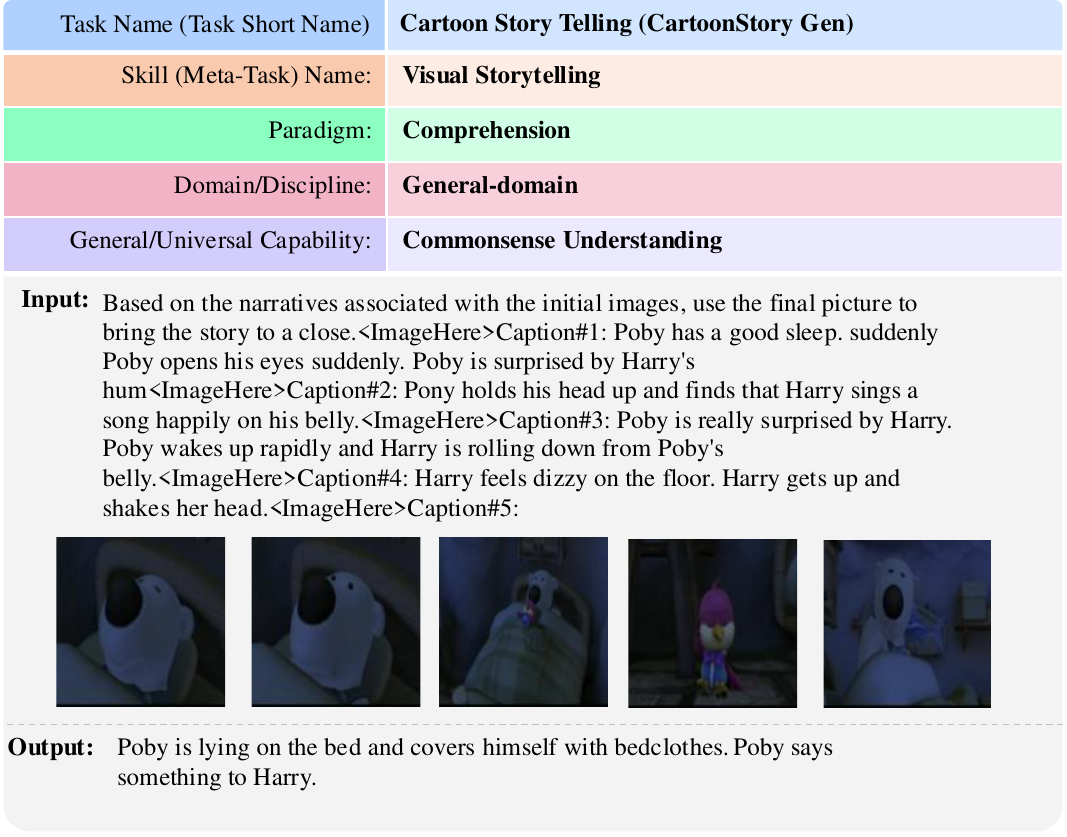}
\caption{Cartoon Storytelling.}
\label{fig:case-I-C-39}
\vspace{-2mm}
\end{figure*}

\begin{figure*}[!h]
\centering
\includegraphics[width=0.9\linewidth]{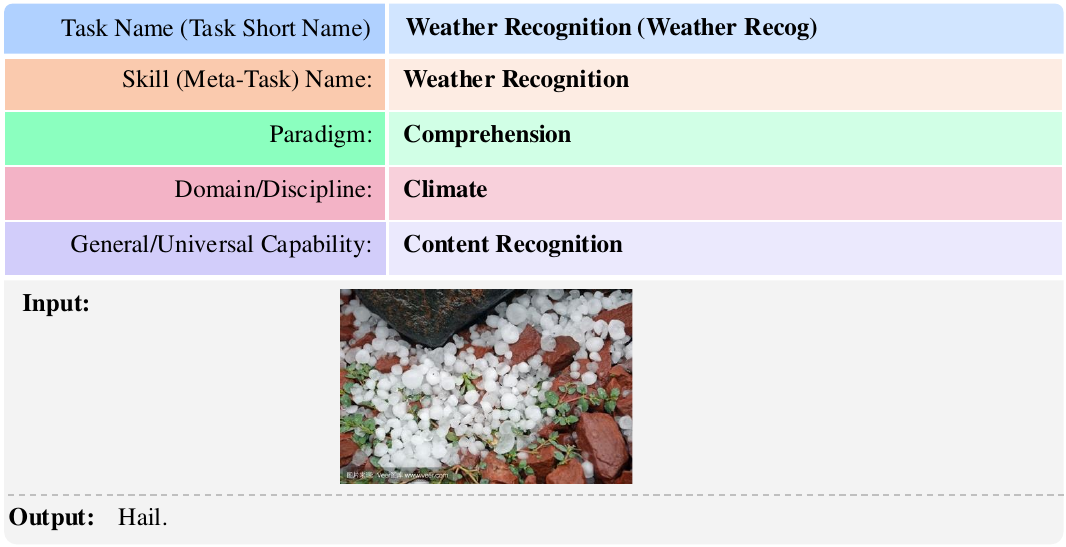}
\caption{Weather Recognition.}
\label{fig:case-I-C-40}
\vspace{-2mm}
\end{figure*}

\clearpage

\paragraph{Generation Tasks.}
Following we showcase 15 image-oriented generative tasks, each representing a specific skill.

\begin{figure*}[!h]
\centering
\includegraphics[width=0.9\linewidth]{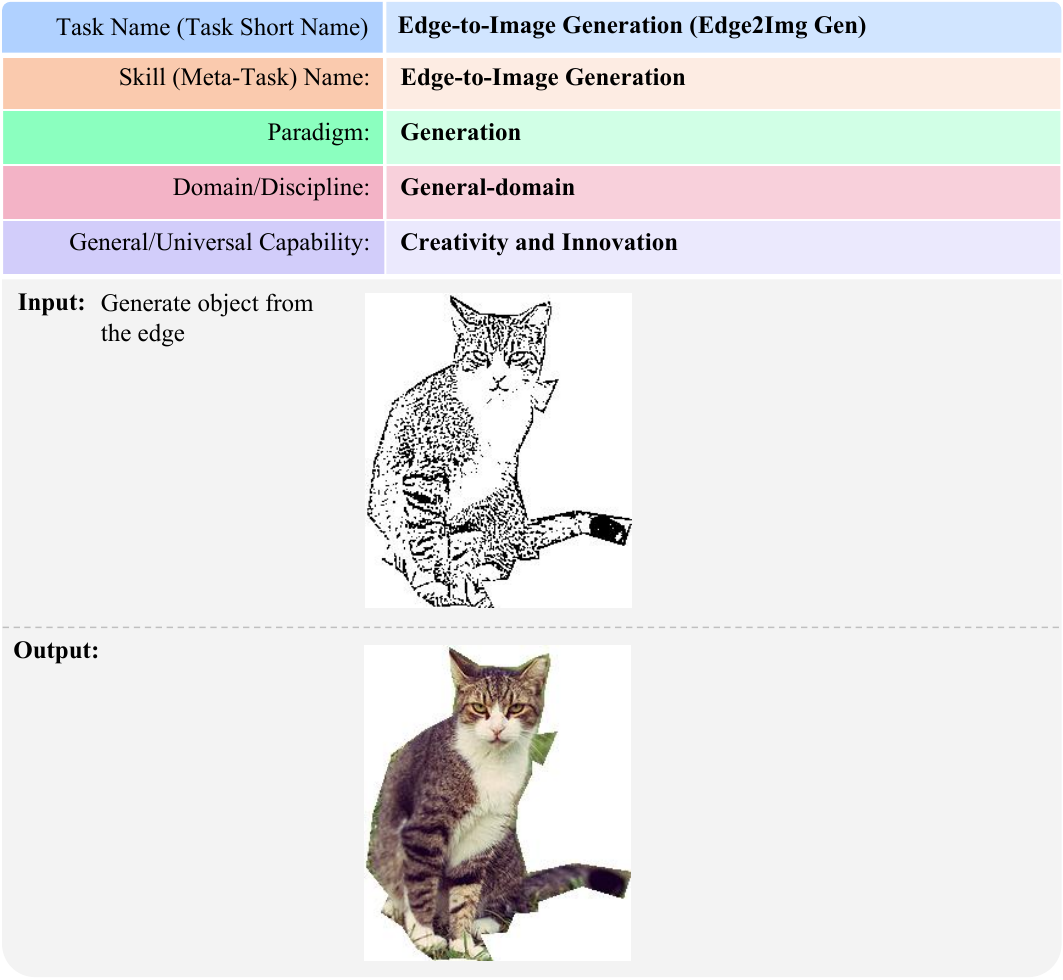}
\caption{Edge-to-Image Generation.}
\label{fig:case-I-G-1}
\vspace{-2mm}
\end{figure*}

\begin{figure*}[!h]
\centering
\includegraphics[width=0.9\linewidth]{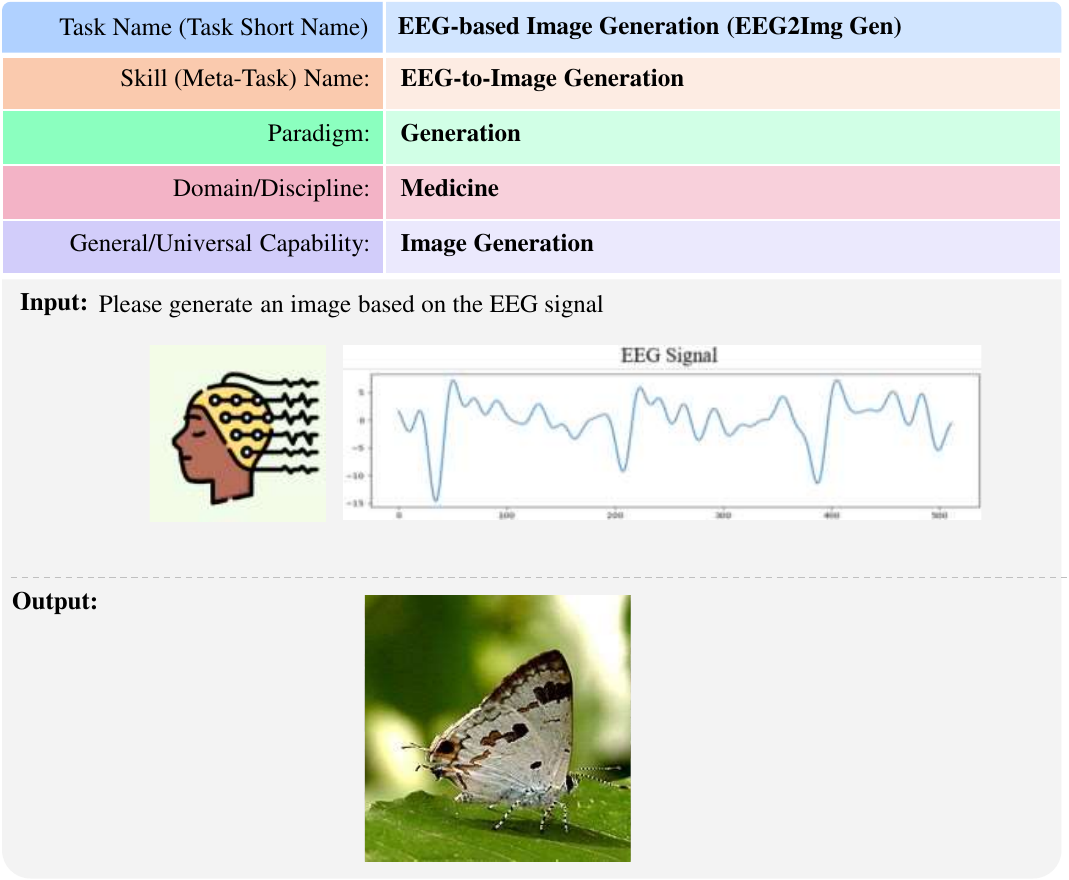}
\caption{EEG-to-Image Generation.}
\label{fig:case-I-G-2}
\vspace{-2mm}
\end{figure*}

\begin{figure*}[!h]
\centering
\includegraphics[width=0.9\linewidth]{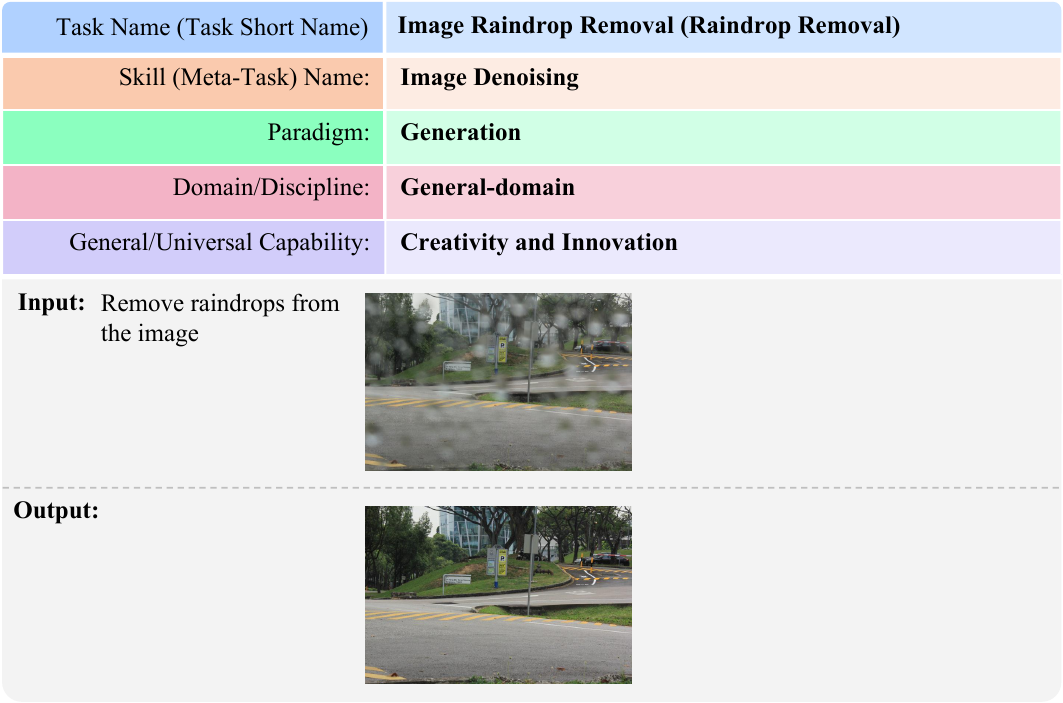}
\caption{Image Raindrop Removal.}
\label{fig:case-I-G-3}
\vspace{-2mm}
\end{figure*}

\begin{figure*}[!h]
\centering
\includegraphics[width=0.9\linewidth]{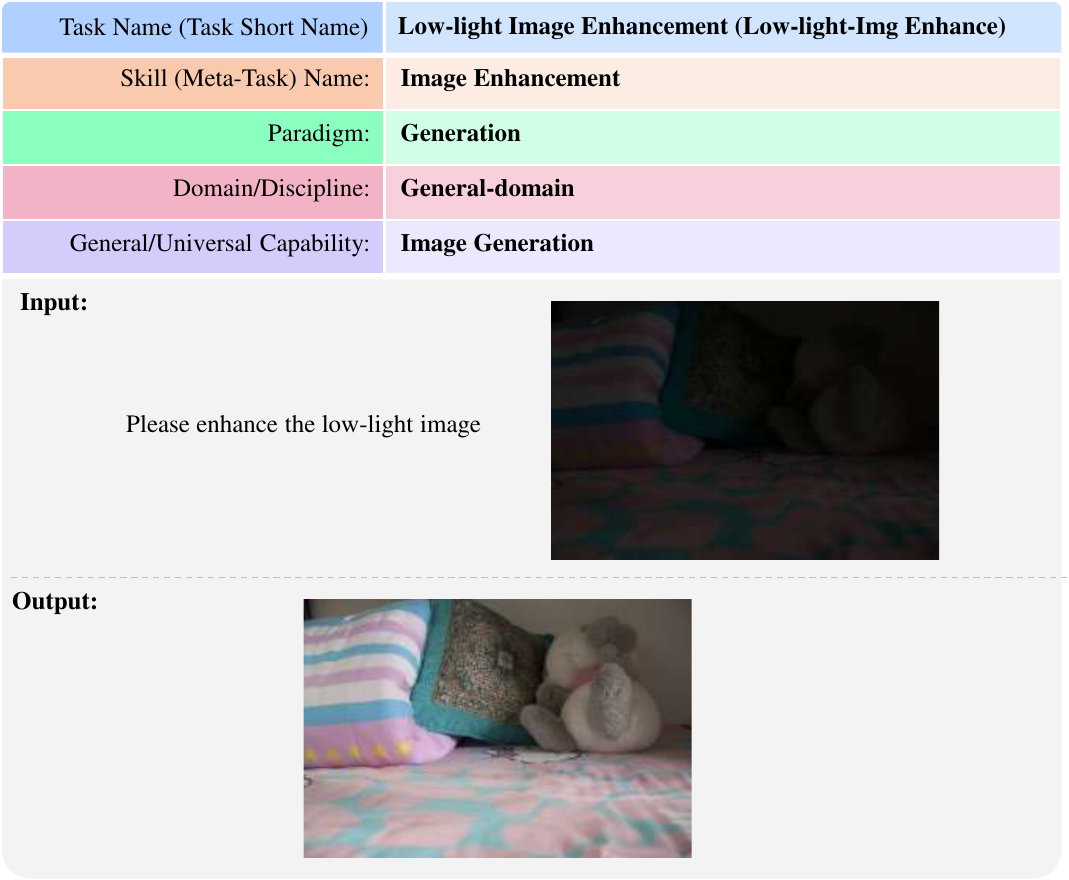}
\caption{Low-light Image Enhancement.}
\label{fig:case-I-G-4}
\vspace{-2mm}
\end{figure*}

\begin{figure*}[!h]
\centering
\includegraphics[width=0.9\linewidth]{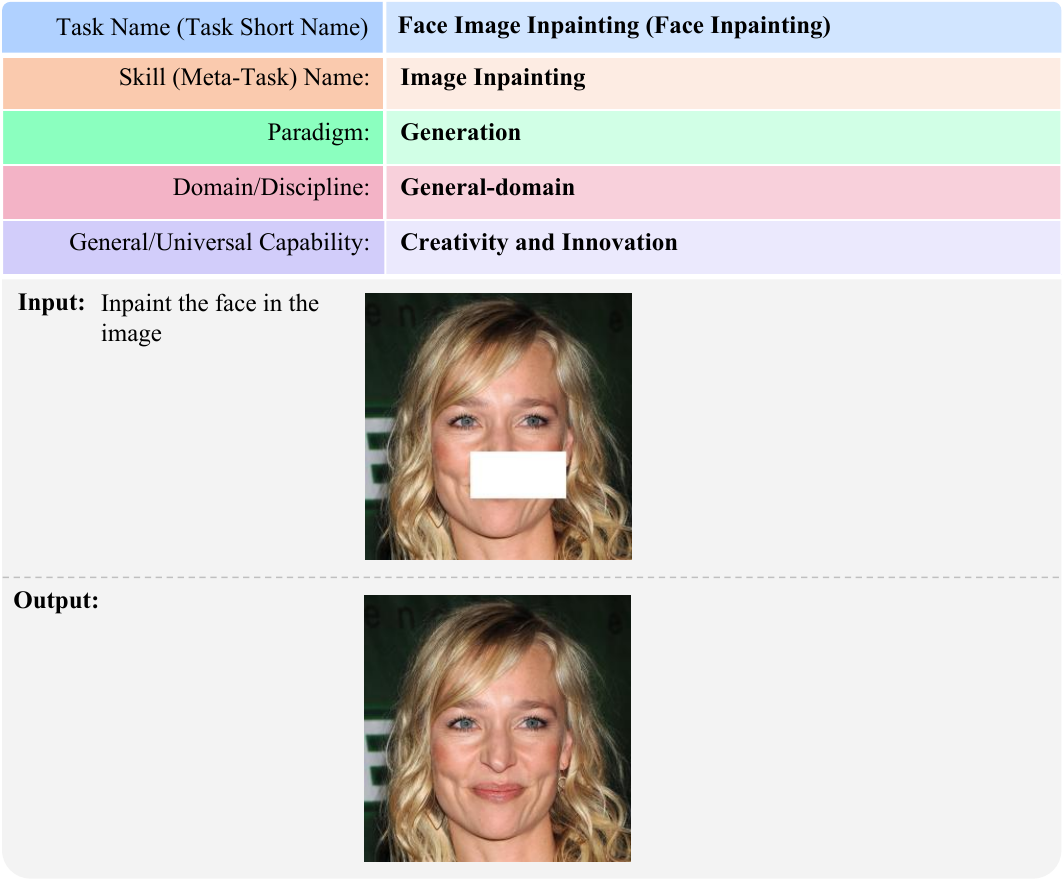}
\caption{Face Image Inpainting.}
\label{fig:case-I-G-5}
\vspace{-2mm}
\end{figure*}

\begin{figure*}[!h]
\centering
\includegraphics[width=0.9\linewidth]{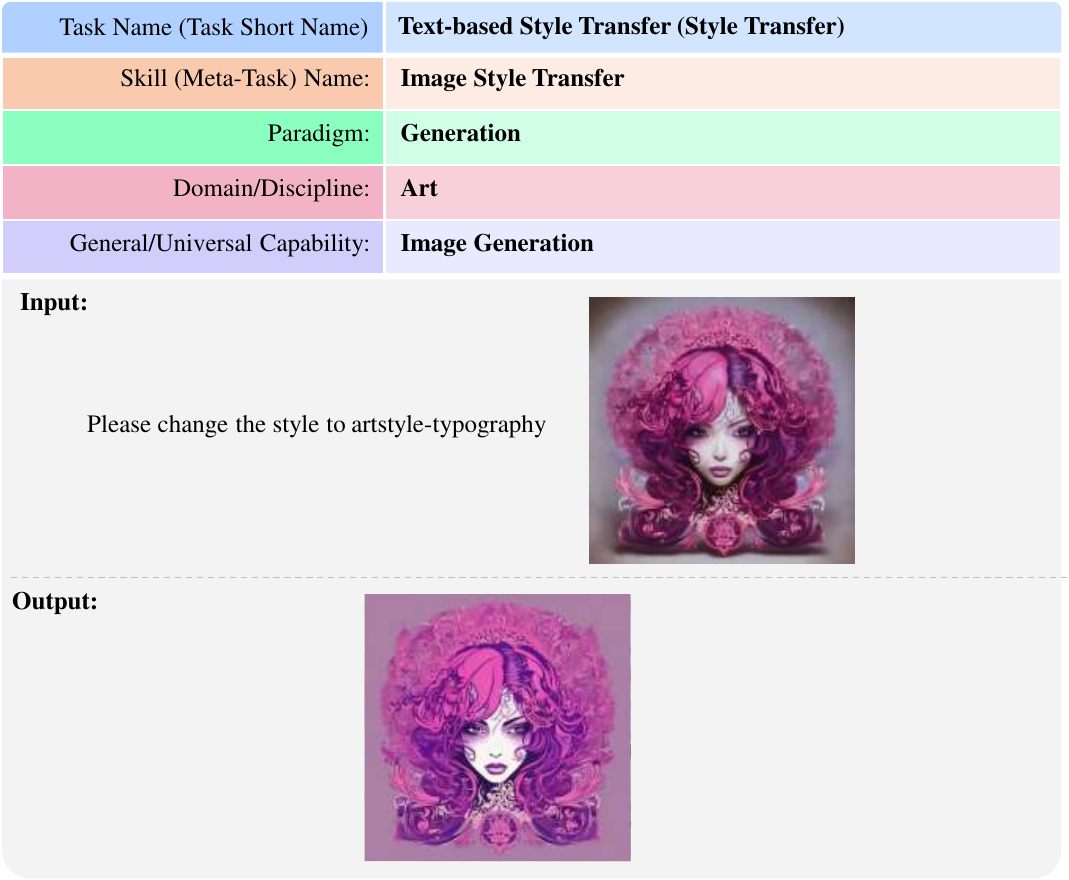}
\caption{Text-based Image Style Transfer.}
\label{fig:case-I-G-6}
\vspace{-2mm}
\end{figure*}

\begin{figure*}[!h]
\centering
\includegraphics[width=0.9\linewidth]{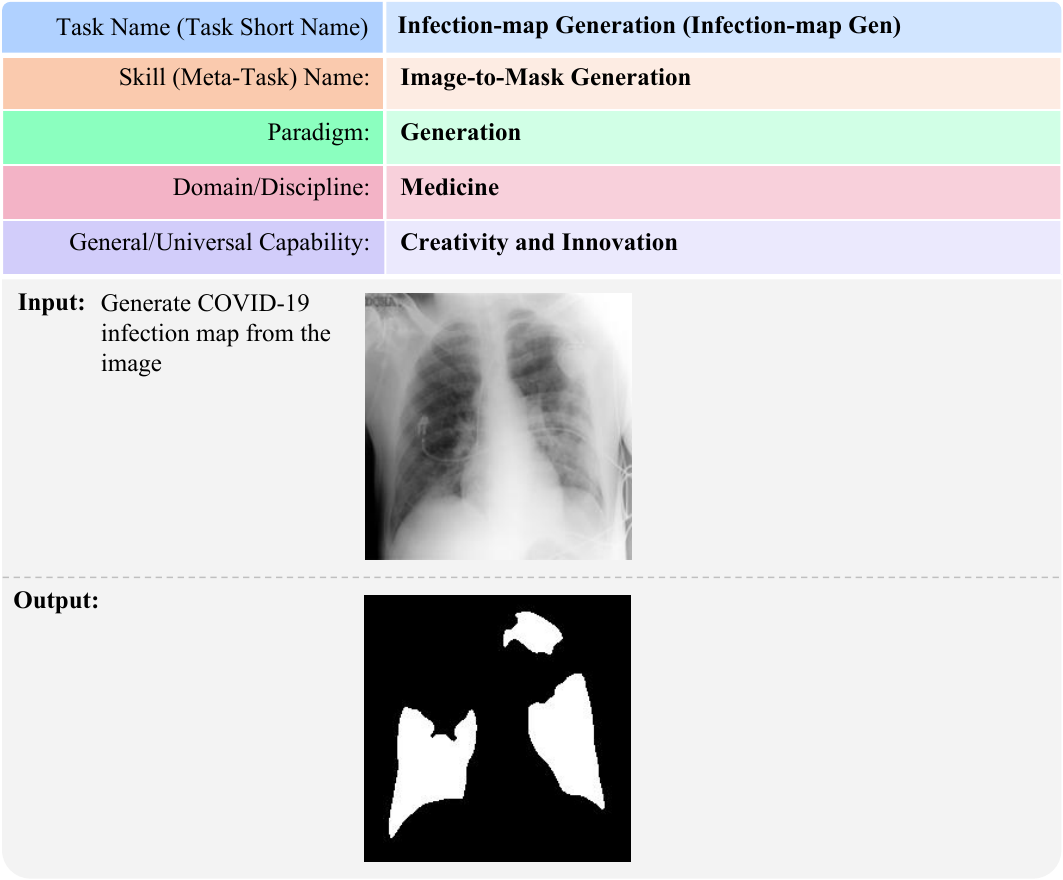}
\caption{Infection-map Generation.}
\label{fig:case-I-G-7}
\vspace{-2mm}
\end{figure*}

\begin{figure*}[!h]
\centering
\includegraphics[width=0.9\linewidth]{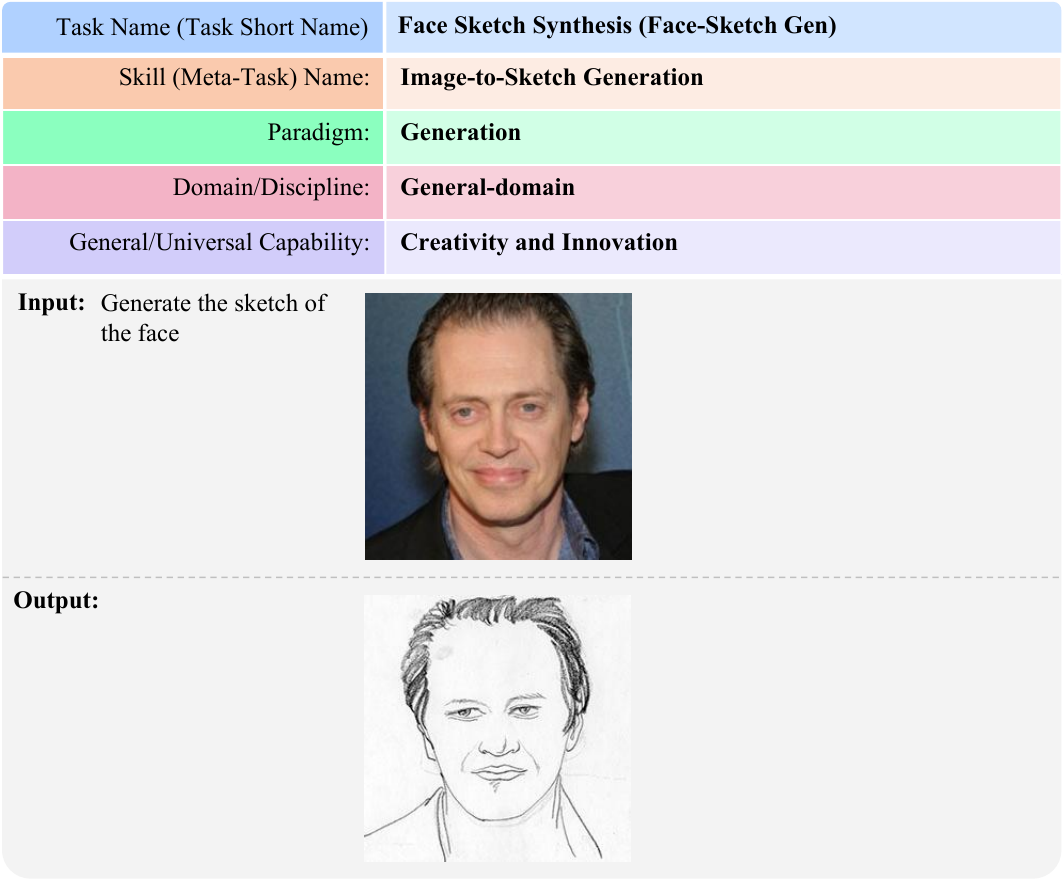}
\caption{Face Sketch Synthesis.}
\label{fig:case-I-G-8}
\vspace{-2mm}
\end{figure*}

\begin{figure*}[!h]
\centering
\includegraphics[width=0.9\linewidth]{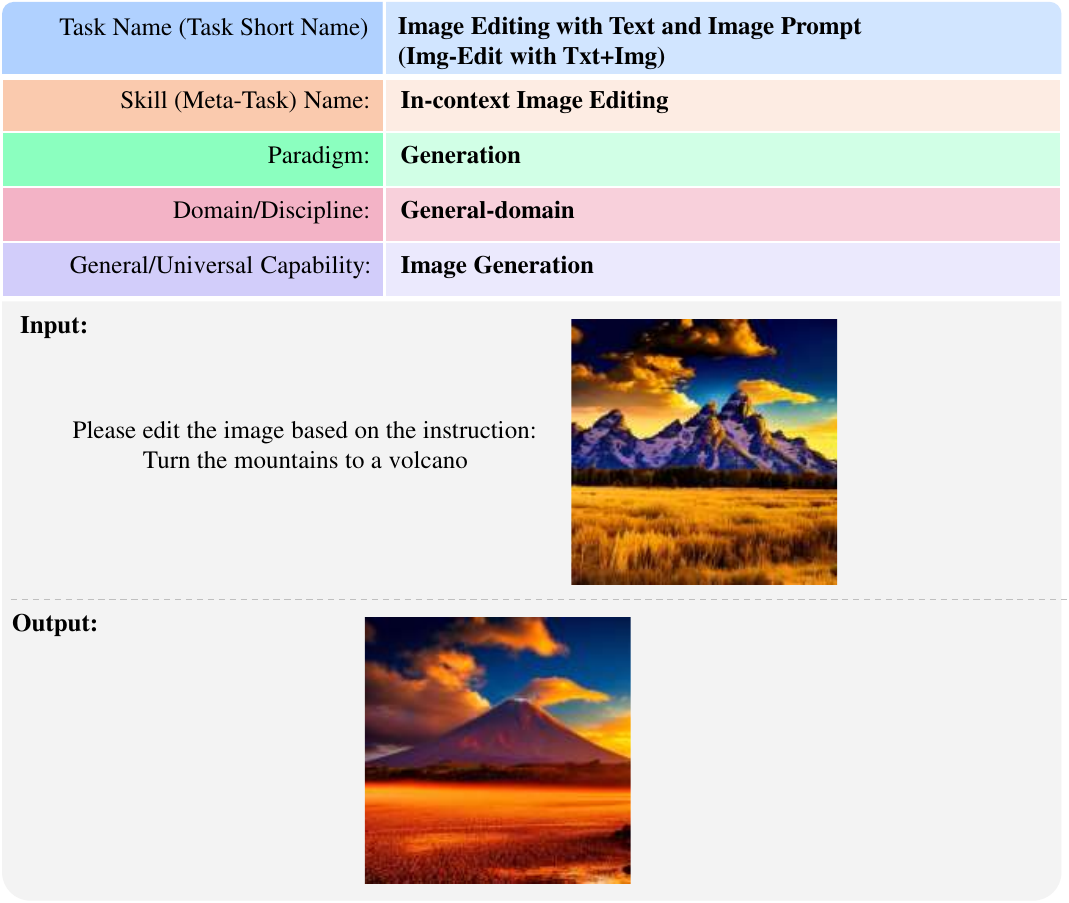}
\caption{Image Editing with Text and Image Prompt.}
\label{fig:case-I-G-9}
\vspace{-2mm}
\end{figure*}

\begin{figure*}[!h]
\centering
\includegraphics[width=0.9\linewidth]{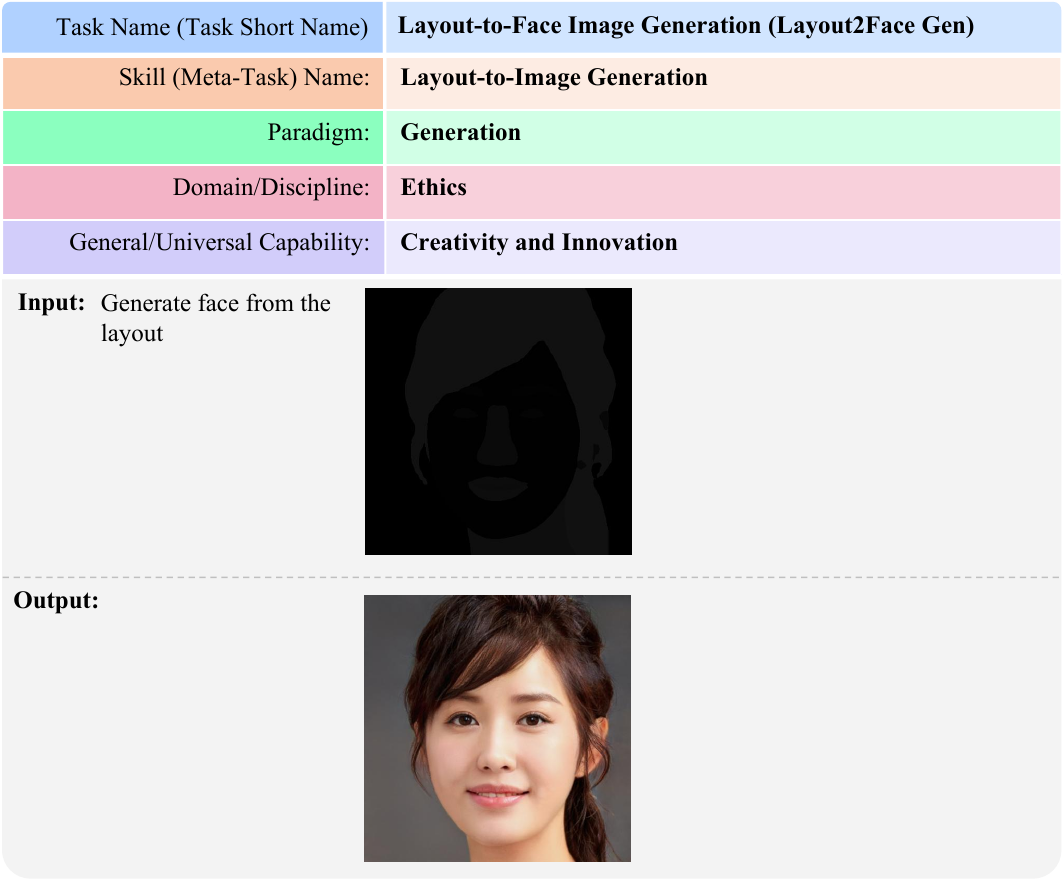}
\caption{Layout-to-Face Image Generation.}
\label{fig:case-I-G-10}
\vspace{-2mm}
\end{figure*}

\begin{figure*}[!h]
\centering
\includegraphics[width=0.9\linewidth]{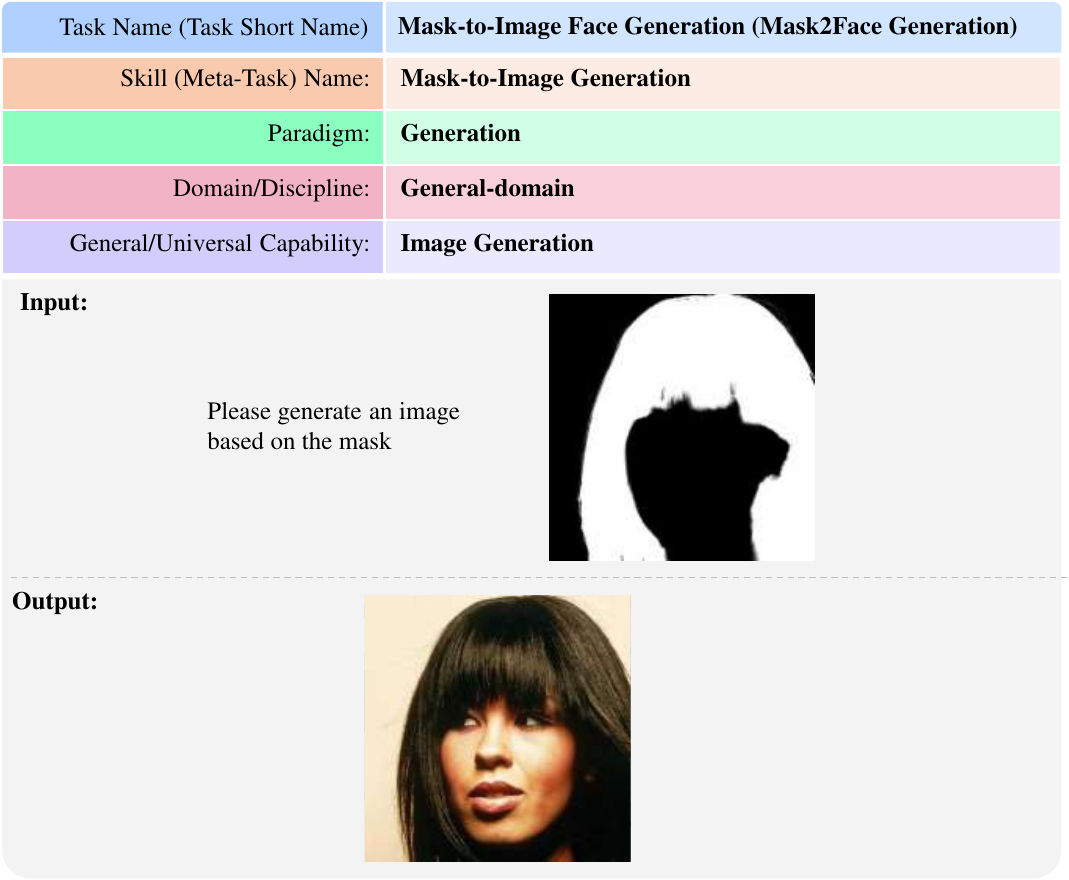}
\caption{Mask-to-Face Image Generation.}
\label{fig:case-I-G-11}
\vspace{-2mm}
\end{figure*}

\begin{figure*}[!h]
\centering
\includegraphics[width=0.9\linewidth]{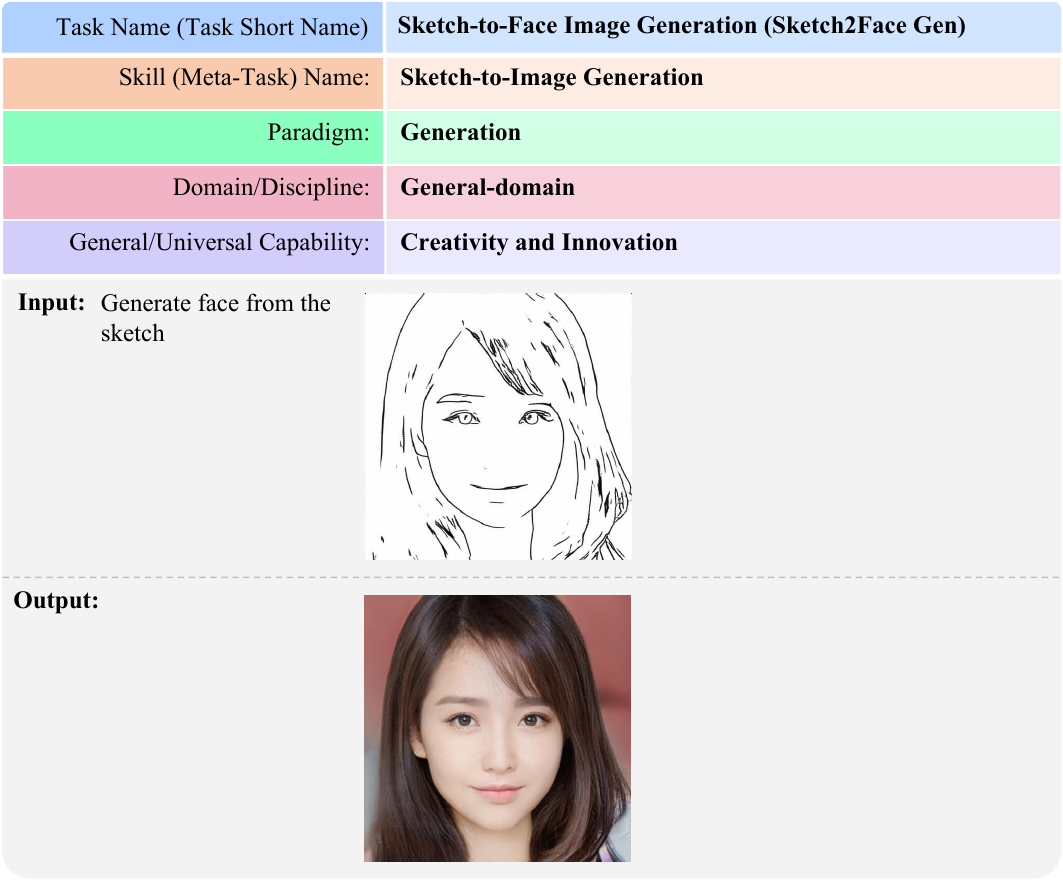}
\caption{Sketch-to-Face Image Generation.}
\label{fig:case-I-G-12}
\vspace{-2mm}
\end{figure*}

\begin{figure*}[!h]
\centering
\includegraphics[width=0.9\linewidth]{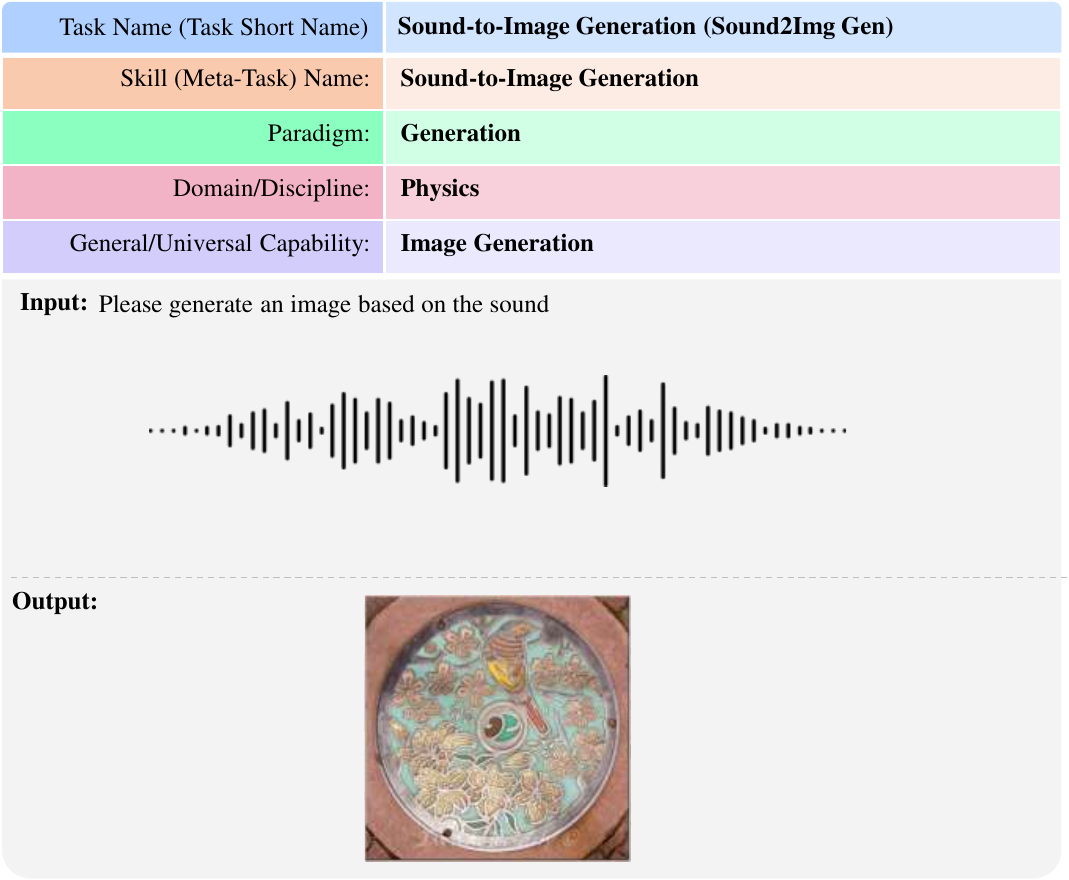}
\caption{Sound-to-Image Generation.}
\label{fig:case-I-G-13}
\vspace{-2mm}
\end{figure*}

\begin{figure*}[!h]
\centering
\includegraphics[width=0.9\linewidth]{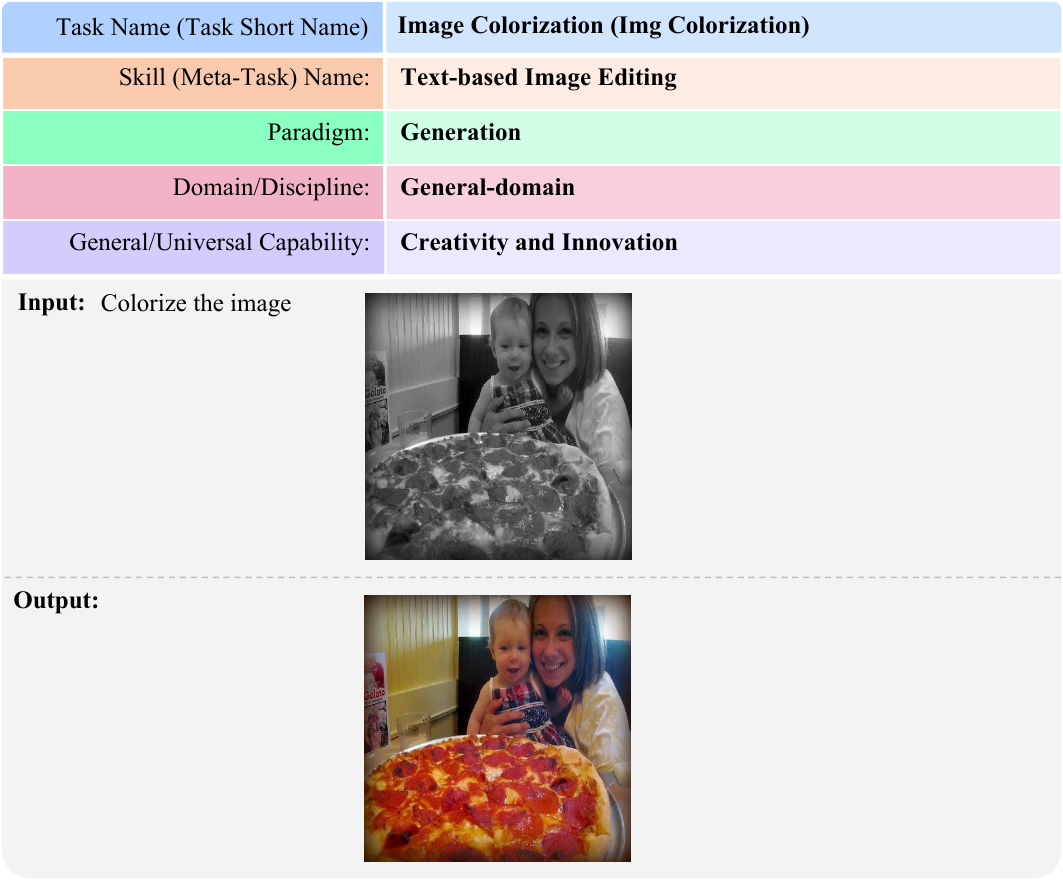}
\caption{Image Colorization.}
\label{fig:case-I-G-14}
\vspace{-2mm}
\end{figure*}

\begin{figure*}[!h]
\centering
\includegraphics[width=0.9\linewidth]{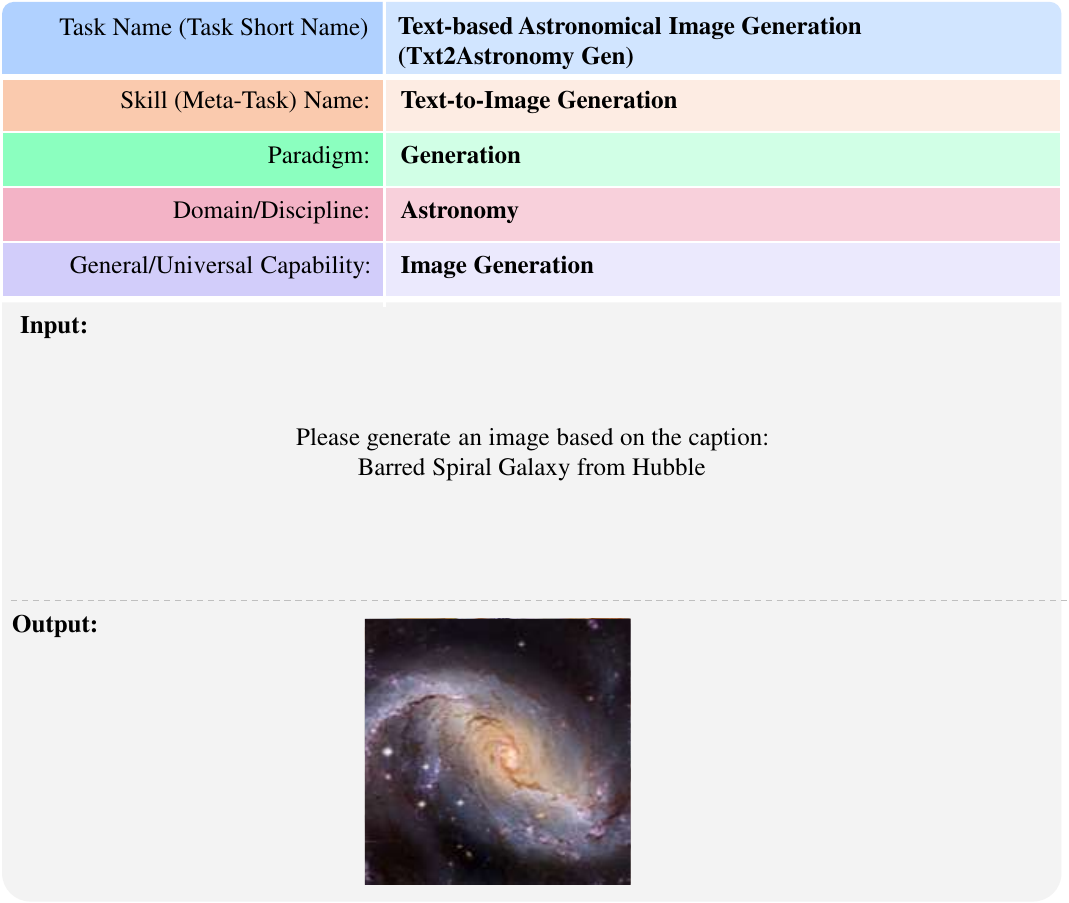}
\caption{Text-based Astronomical Image Generation.}
\label{fig:case-I-G-15}
\vspace{-2mm}
\end{figure*}

\clearpage

\subsubsection{Video-related Tasks}

\paragraph{Comprehension Tasks.}
Following, we showcase 20 video-oriented comprehensive tasks, each representing a specific skill.

\begin{figure*}[!h]
\centering
\includegraphics[width=0.9\linewidth]{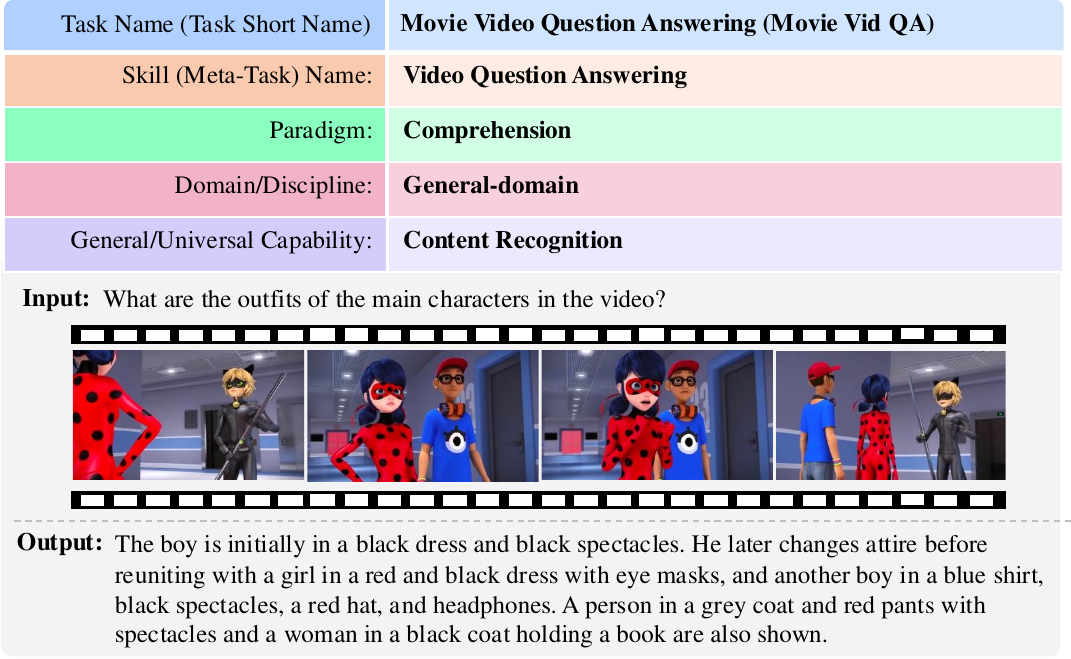}
\caption{Movie Video Question Answering.}
\label{fig:case-V-C-1}
\vspace{-2mm}
\end{figure*}

\begin{figure*}[!h]
\centering
\includegraphics[width=0.9\linewidth]{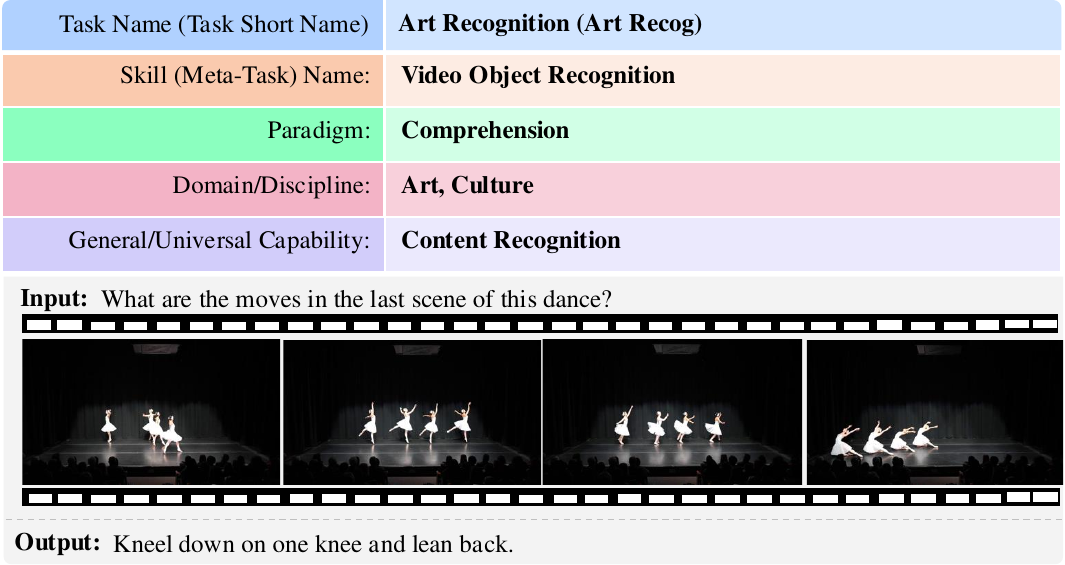}
\caption{Art Recognition.}
\label{fig:case-V-C-2}
\vspace{-2mm}
\end{figure*}

\begin{figure*}[!h]
\centering
\includegraphics[width=0.9\linewidth]{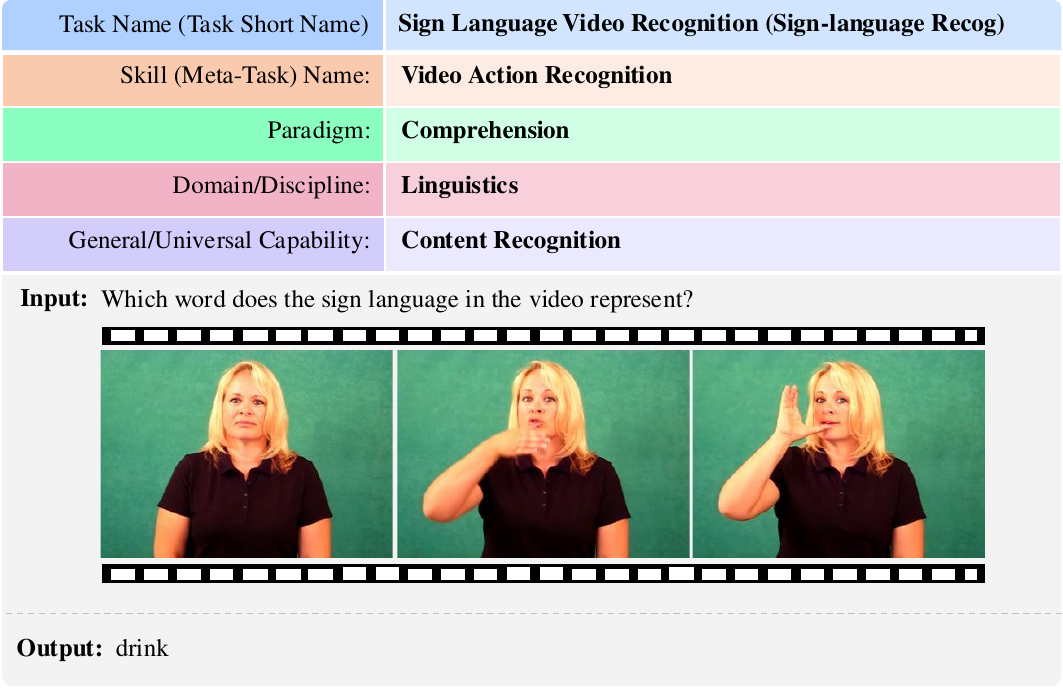}
\caption{Sign Language Video Recognition.}
\label{fig:case-V-C-3}
\vspace{-2mm}
\end{figure*}

\begin{figure*}[!h]
\centering
\includegraphics[width=0.9\linewidth]{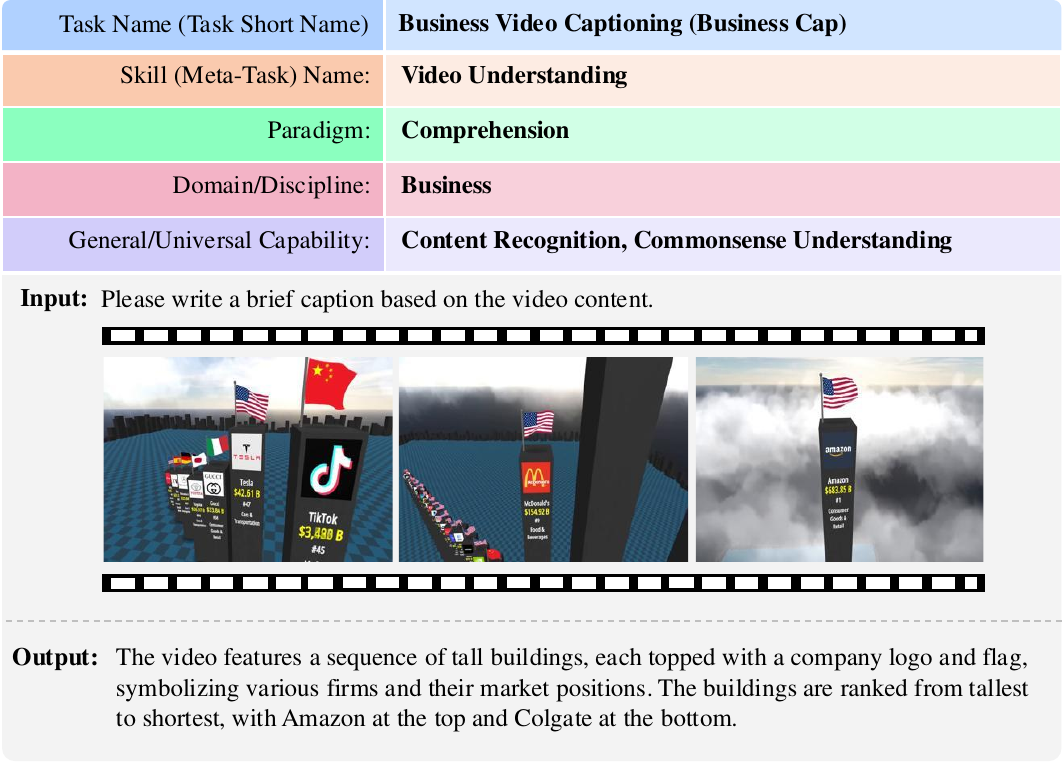}
\caption{Business Video Understanding.}
\label{fig:case-V-C-4}
\vspace{-2mm}
\end{figure*}

\begin{figure*}[!h]
\centering
\includegraphics[width=0.9\linewidth]{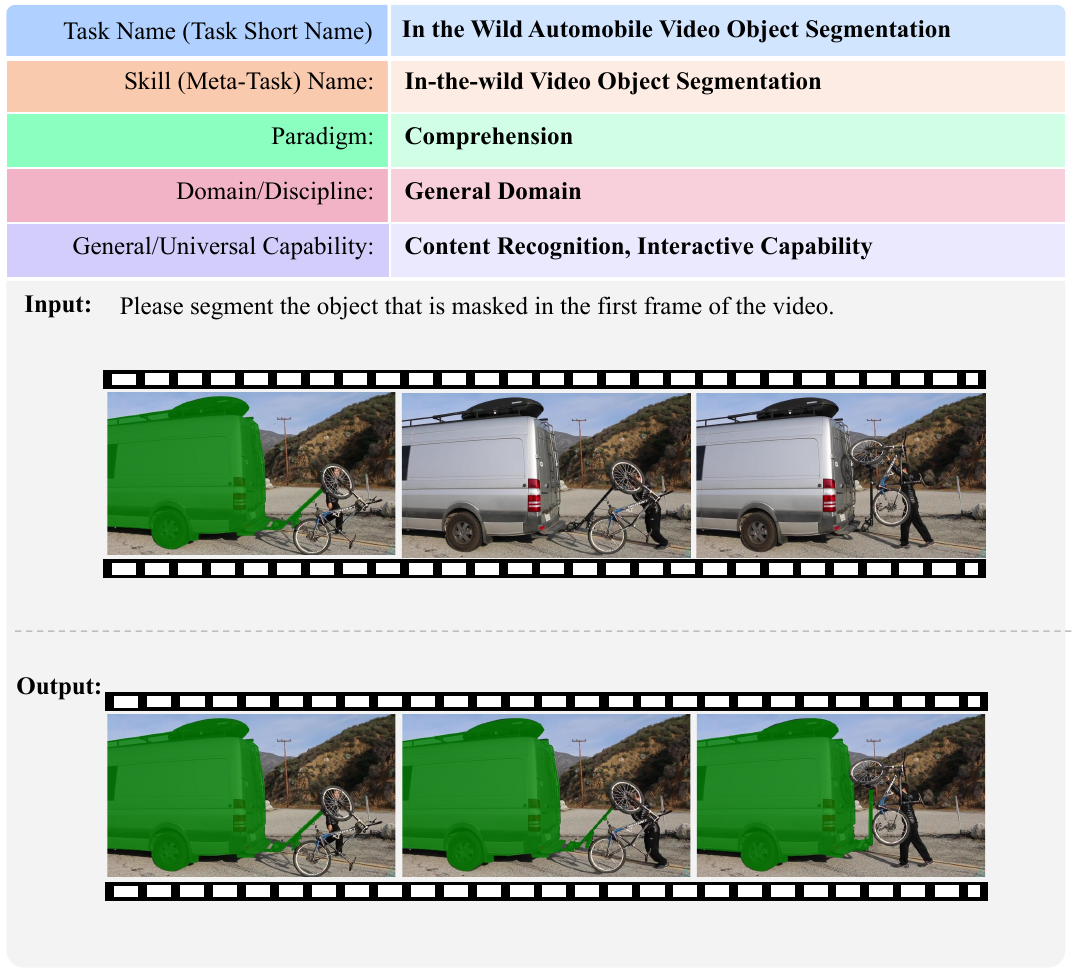}
\caption{In-the-Wild Automobile Video Object Segmentation.}
\label{fig:case-V-C-5}
\vspace{-2mm}
\end{figure*}

\begin{figure*}[!h]
\centering
\includegraphics[width=0.9\linewidth]{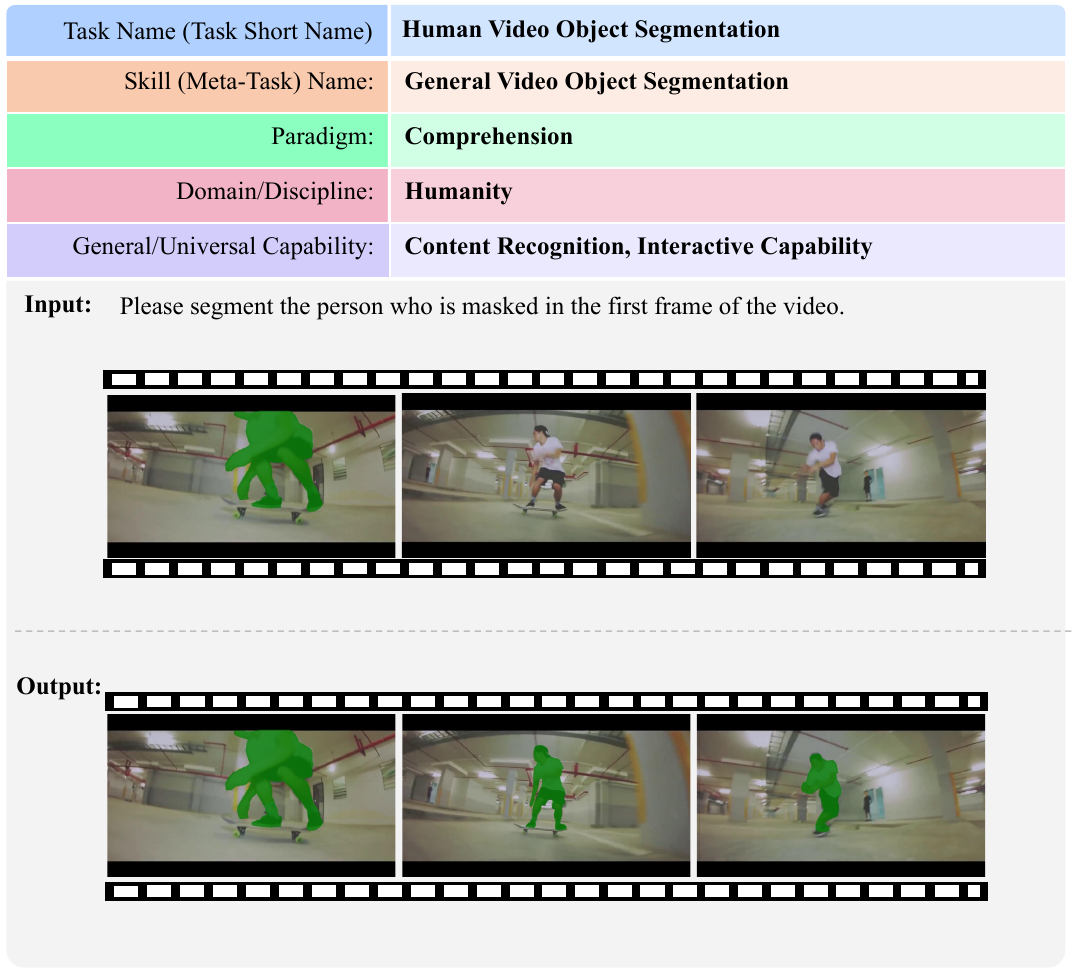}
\caption{Human Video Object Segmentation.}
\label{fig:case-V-C-6}
\vspace{-2mm}
\end{figure*}

\begin{figure*}[!h]
\centering
\includegraphics[width=0.9\linewidth]{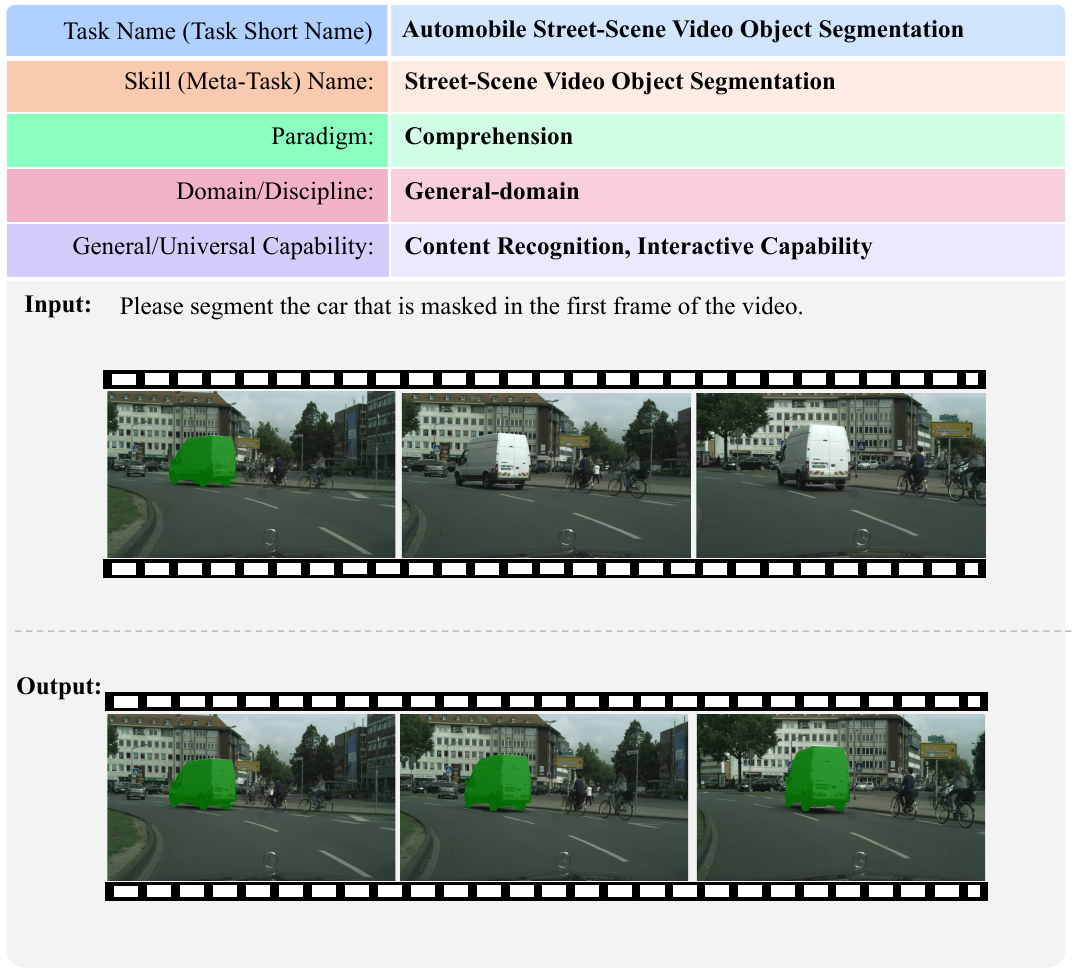}
\caption{Automobile Street-Scene Video Object Segmentation.}
\label{fig:case-V-C-7}
\vspace{-2mm}
\end{figure*}

\begin{figure*}[!h]
\centering
\includegraphics[width=0.9\linewidth]{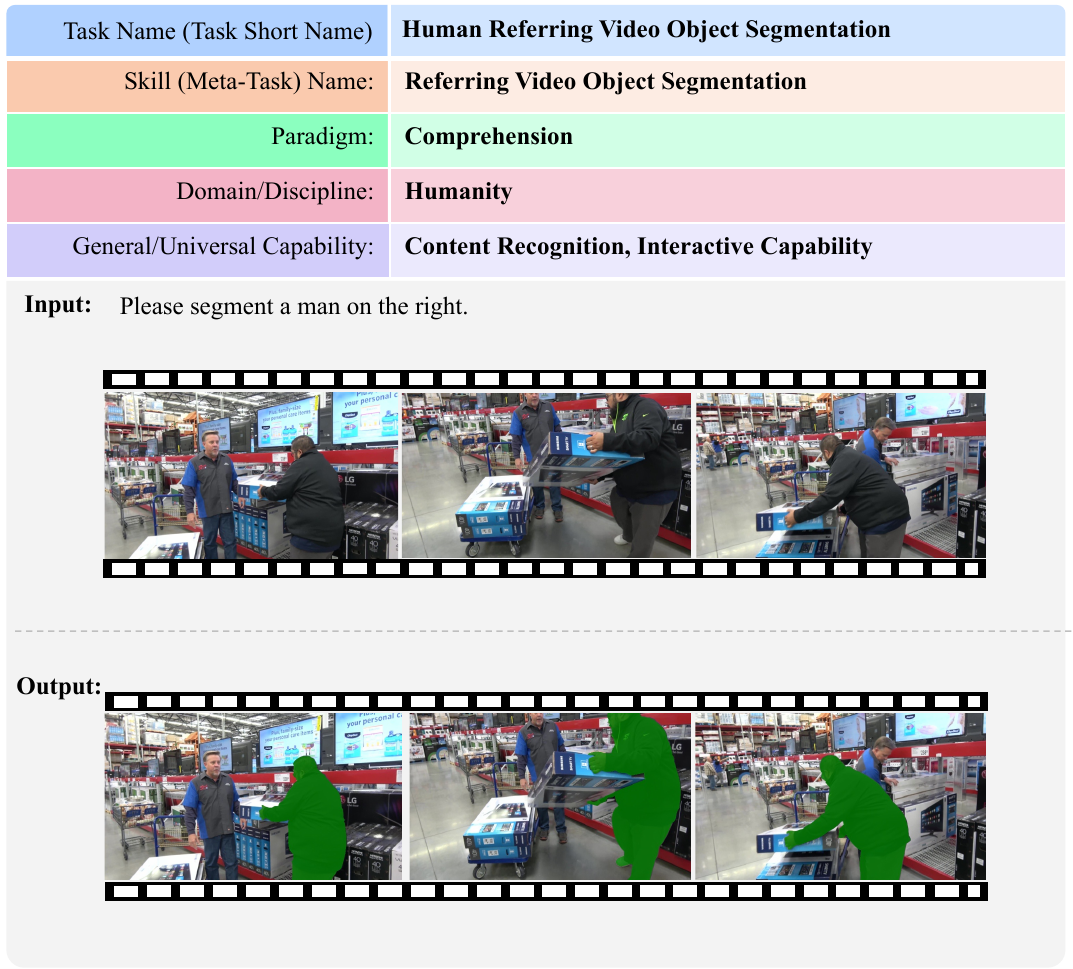}
\caption{Human Referring Video Object Segmentation.}
\label{fig:case-V-C-8}
\vspace{-2mm}
\end{figure*}

\begin{figure*}[!h]
\centering
\includegraphics[width=0.9\linewidth]{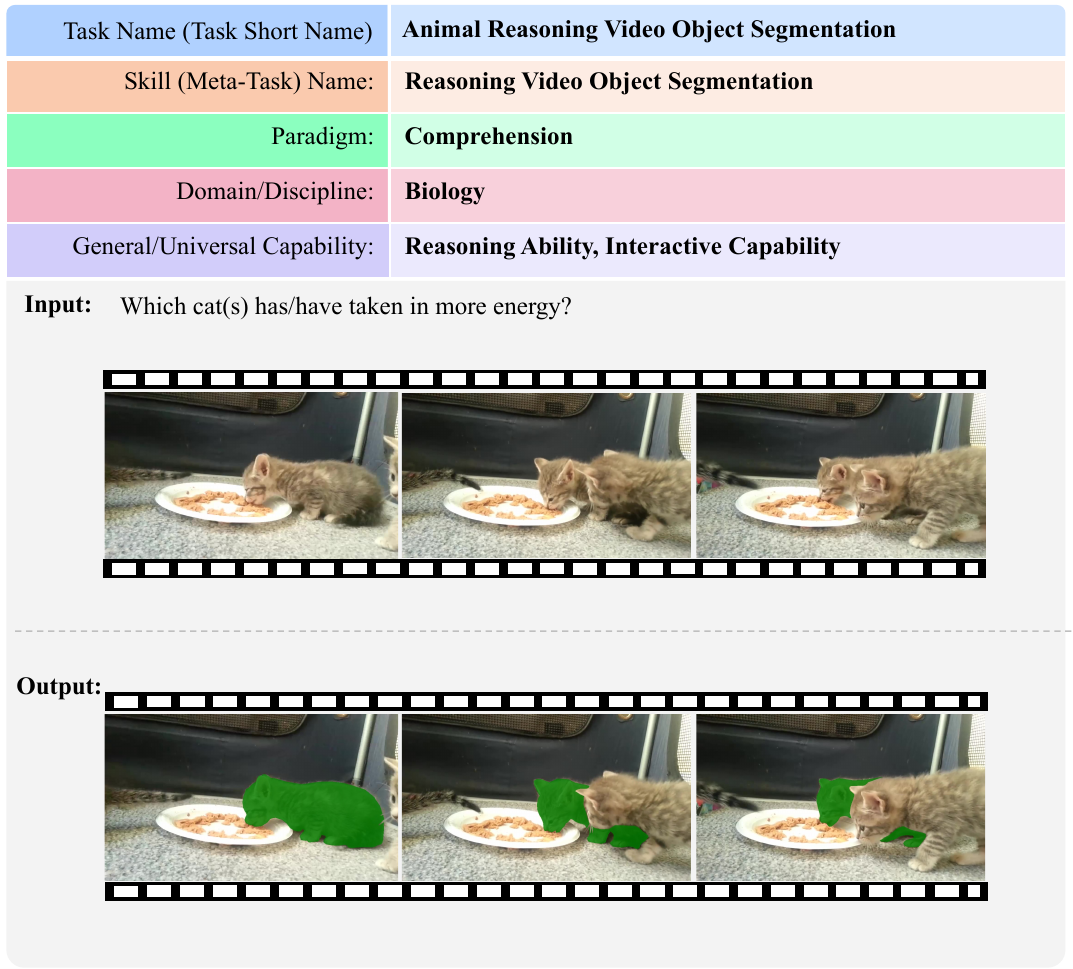}
\caption{Animal Reasoning Video Object Segmentation.}
\label{fig:case-V-C-9}
\vspace{-2mm}
\end{figure*}

\begin{figure*}[!h]
\centering
\includegraphics[width=0.9\linewidth]{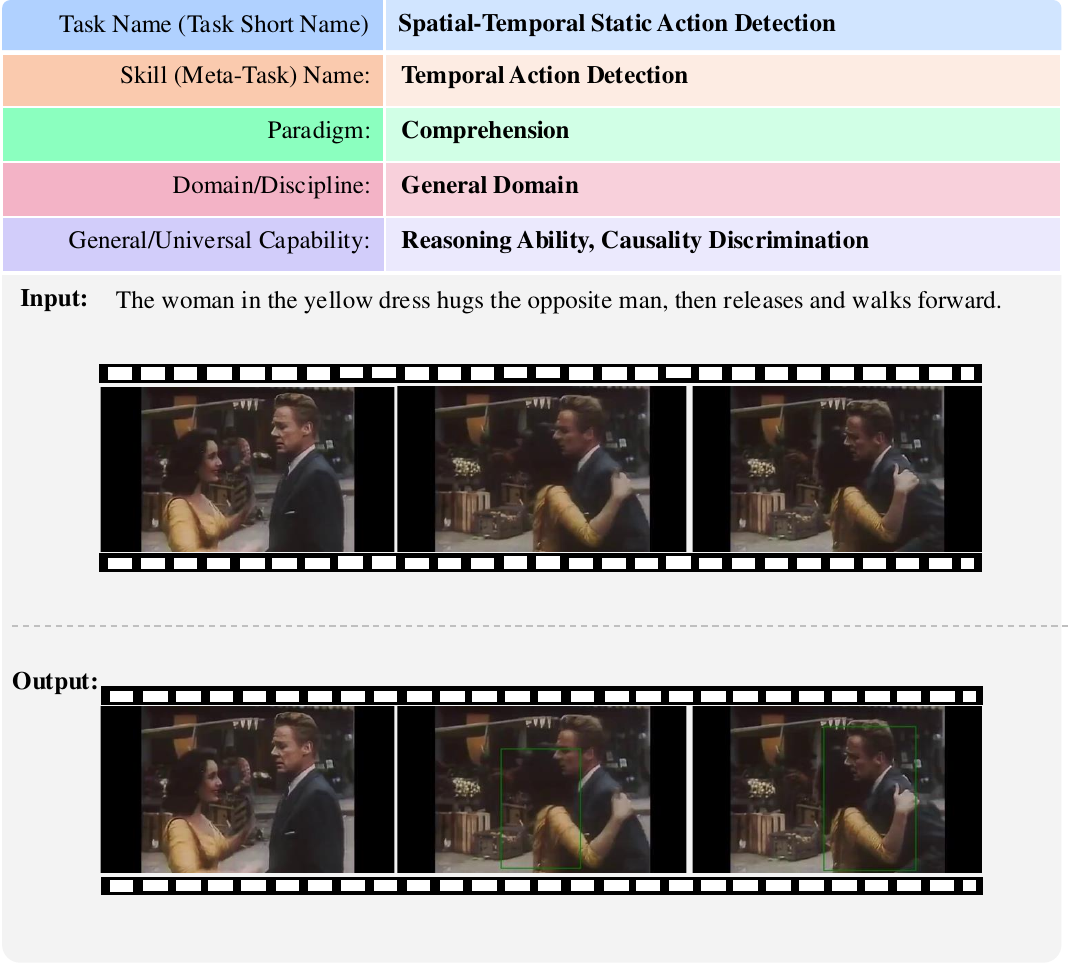}
\caption{Spatial-Temporal Static Action Detection.
}
\label{fig:case-V-C-10}
\vspace{-2mm}
\end{figure*}

\begin{figure*}[!h]
\centering
\includegraphics[width=0.9\linewidth]{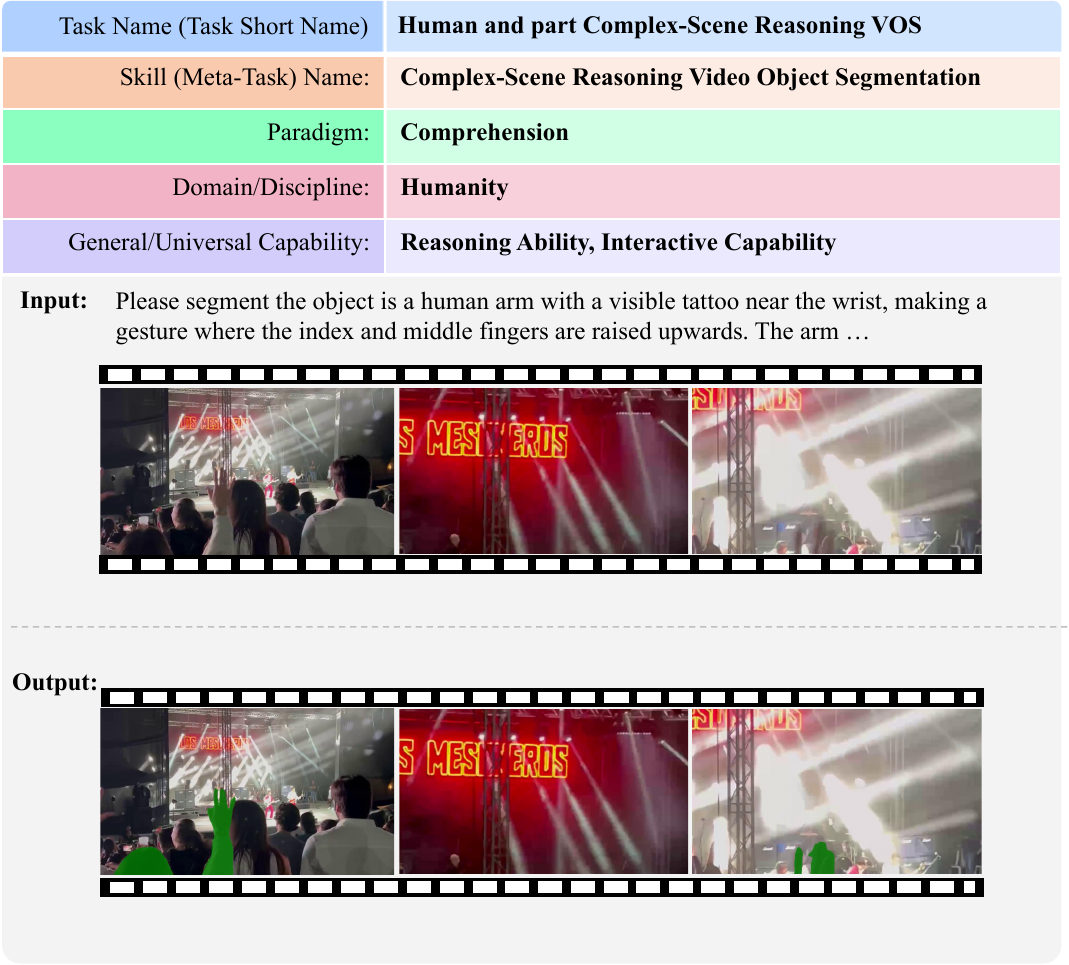}
\caption{Human-and-Part Complex-Scene Reasoning Video Object Segmentation (VOS).}
\label{fig:case-V-C-11}
\vspace{-2mm}
\end{figure*}

\begin{figure*}[!h]
\centering
\includegraphics[width=0.9\linewidth]{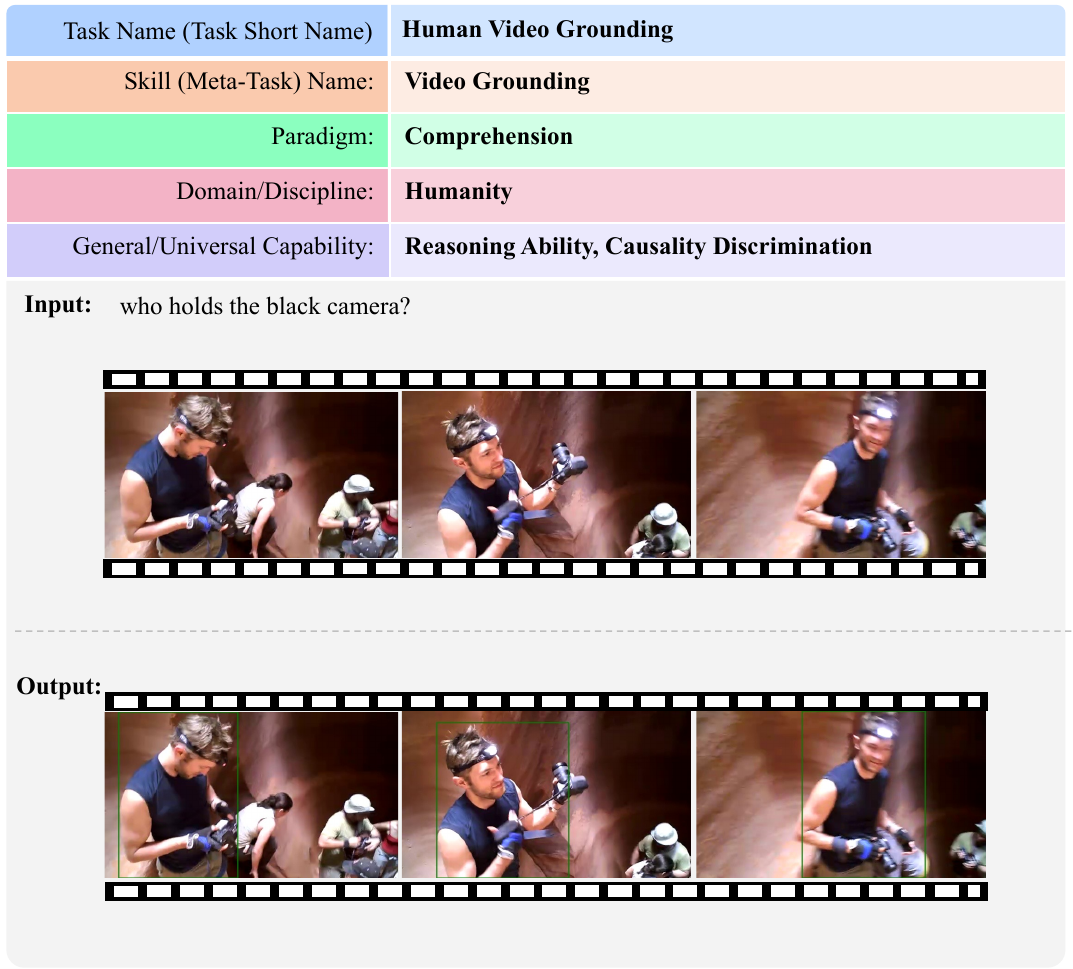}
\caption{Human Video Grounding.}
\label{fig:case-V-C-12}
\vspace{-2mm}
\end{figure*}

\begin{figure*}[!h]
\centering
\includegraphics[width=0.9\linewidth]{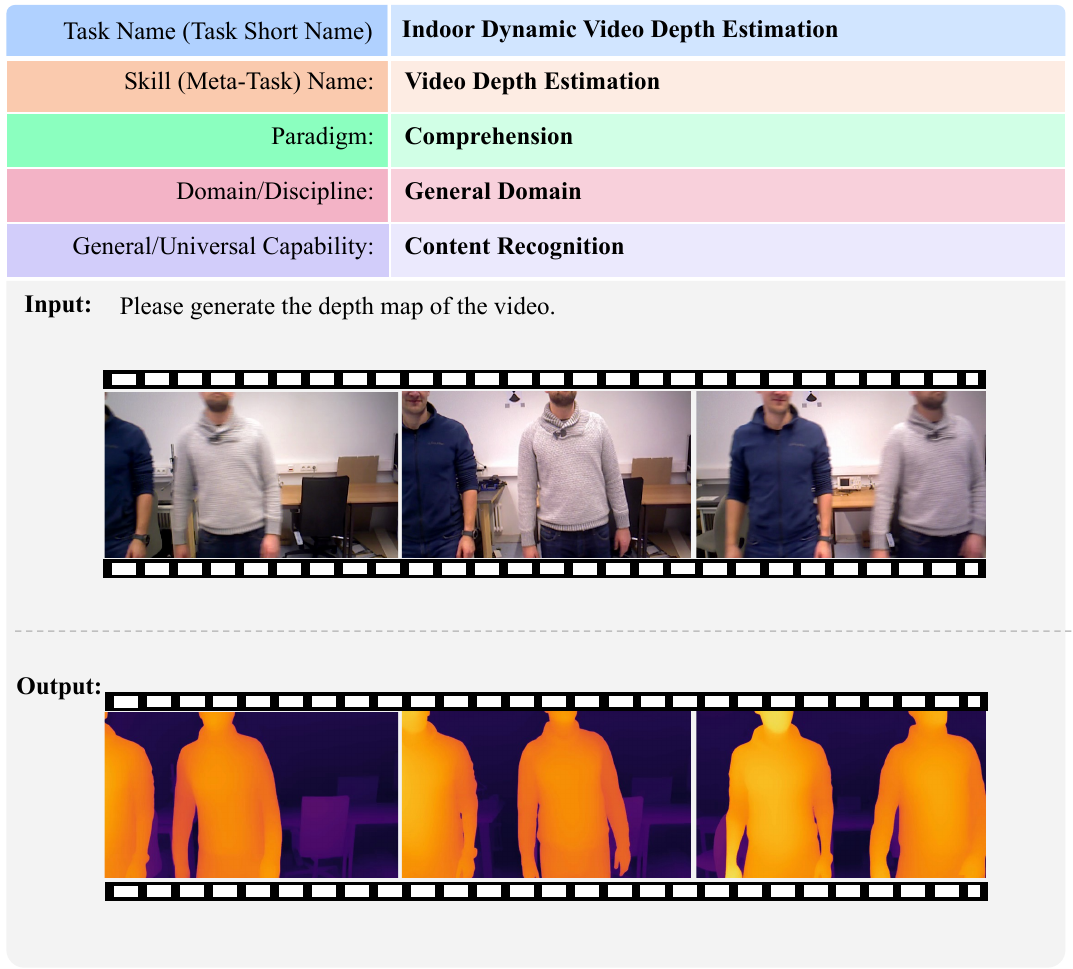}
\caption{Indoor Video Depth Estimation.}
\label{fig:case-V-C-13}
\vspace{-2mm}
\end{figure*}

\begin{figure*}[!h]
\centering
\includegraphics[width=0.9\linewidth]{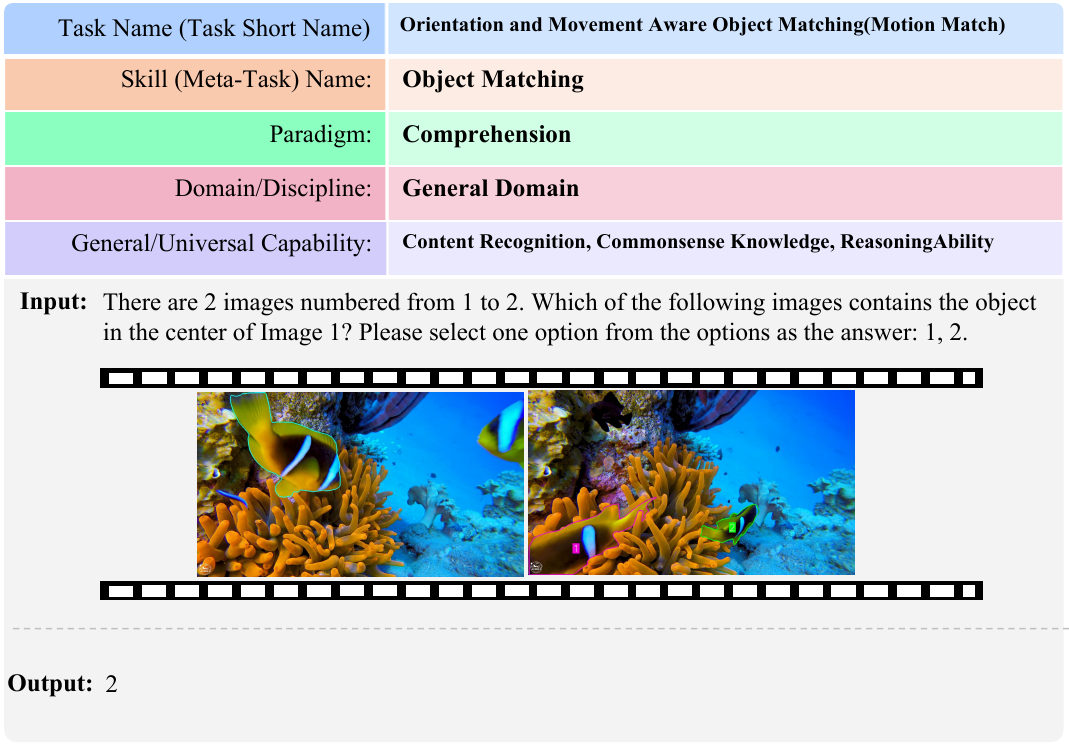}
\caption{Orientation-and-movement-aware Object Matching.}
\label{fig:case-V-C-14}
\vspace{-2mm}
\end{figure*}

\begin{figure*}[!h]
\centering
\includegraphics[width=0.9\linewidth]{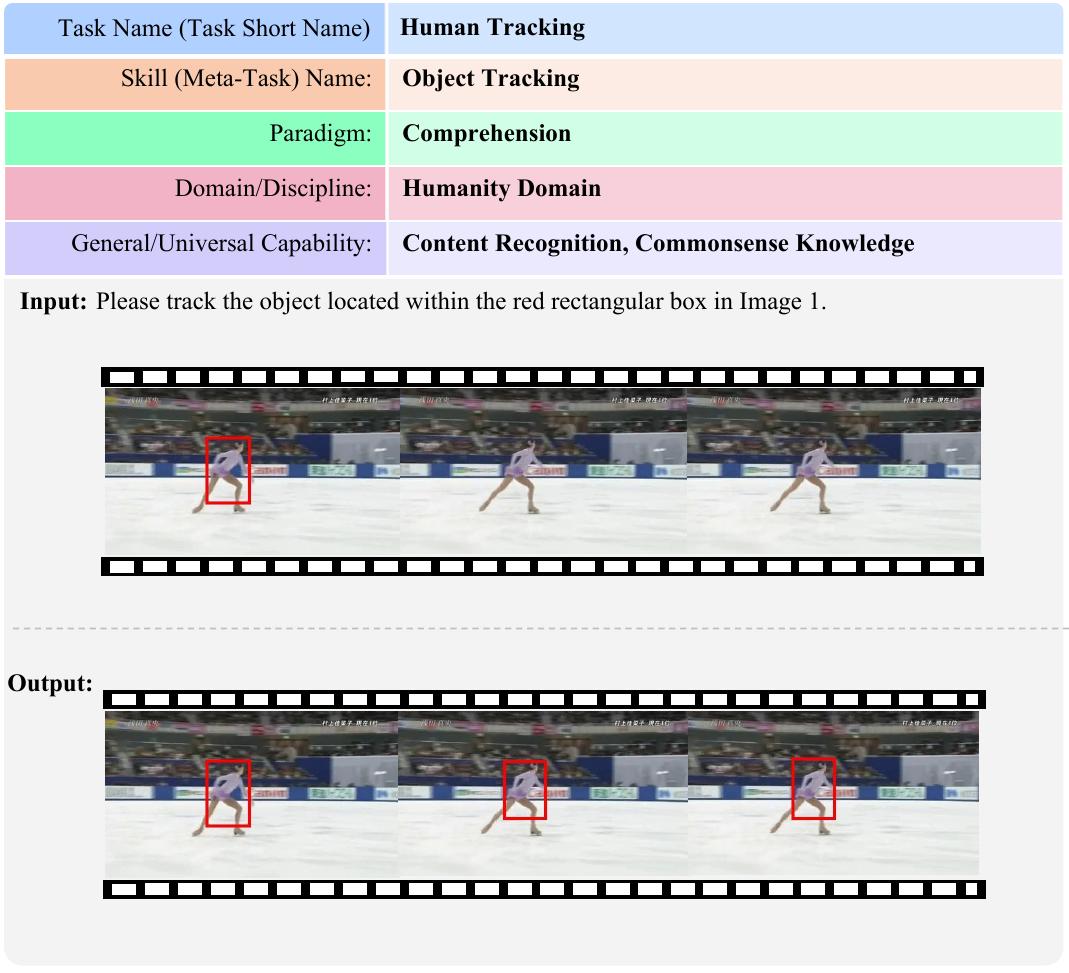}
\caption{Human Tracking.}
\label{fig:case-V-C-15}
\vspace{-2mm}
\end{figure*}

\begin{figure*}[!h]
\centering
\includegraphics[width=0.9\linewidth]{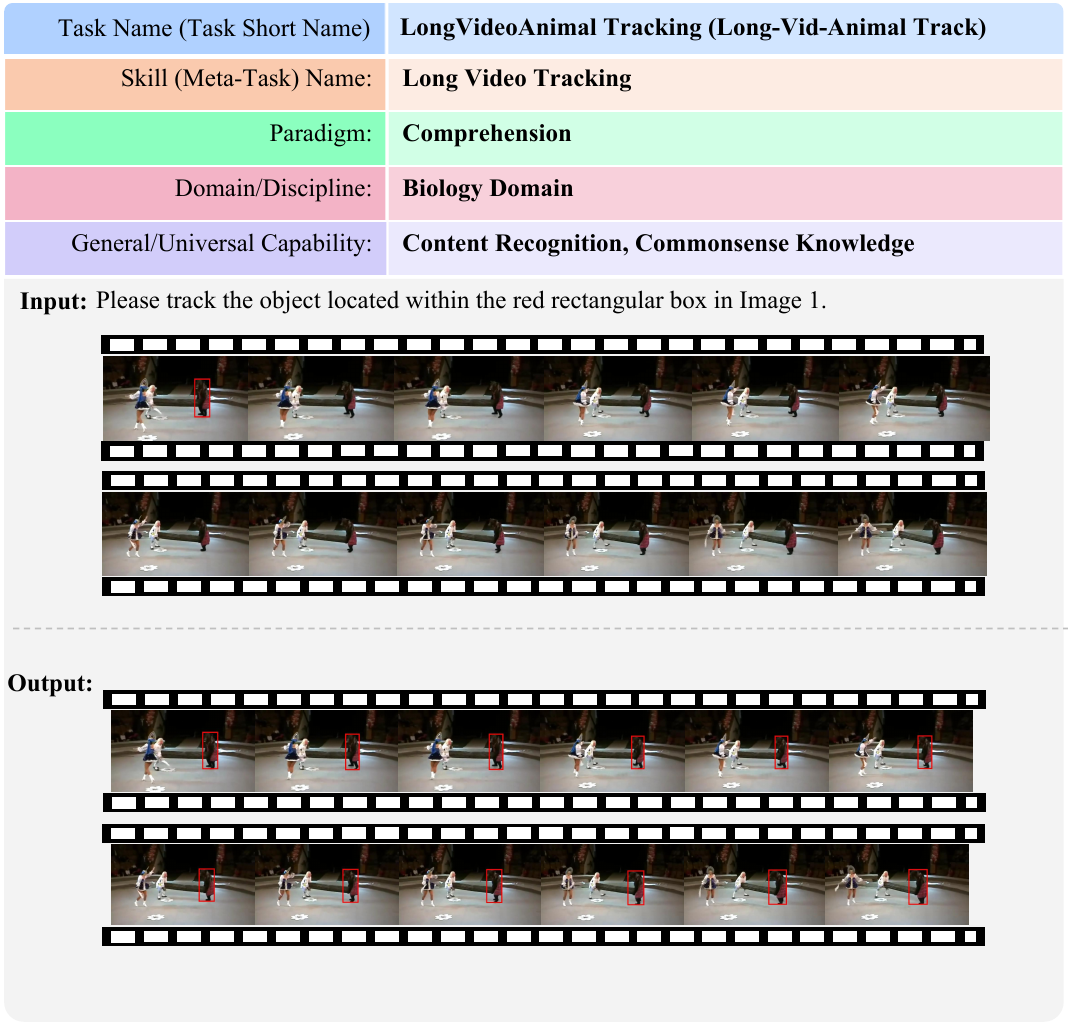}
\caption{Animal Long-Video Tracking.}
\label{fig:case-V-C-16}
\vspace{-2mm}
\end{figure*}

\begin{figure*}[!h]
\centering
\includegraphics[width=0.9\linewidth]{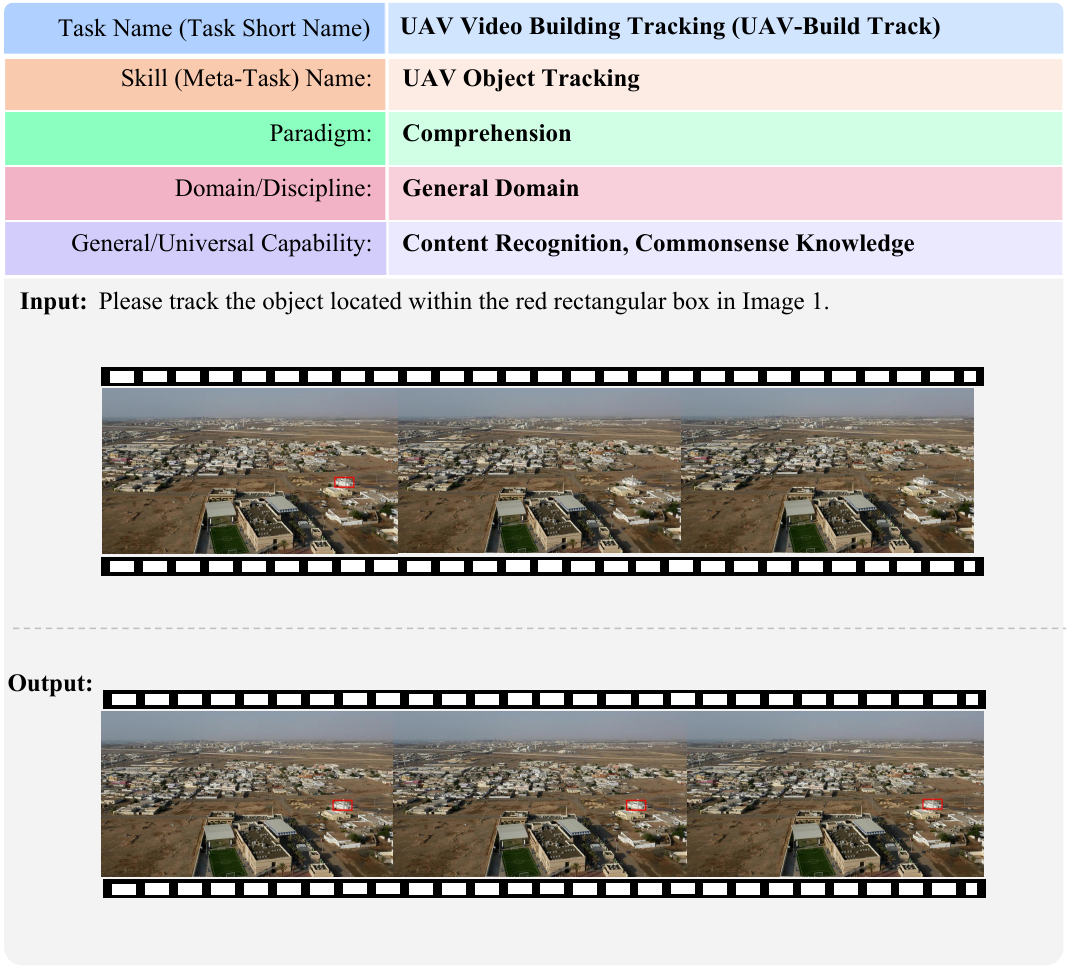}
\caption{UAV Video Building Tracking.}
\label{fig:case-V-C-17}
\vspace{-2mm}
\end{figure*}

\begin{figure*}[!h]
\centering
\includegraphics[width=0.9\linewidth]{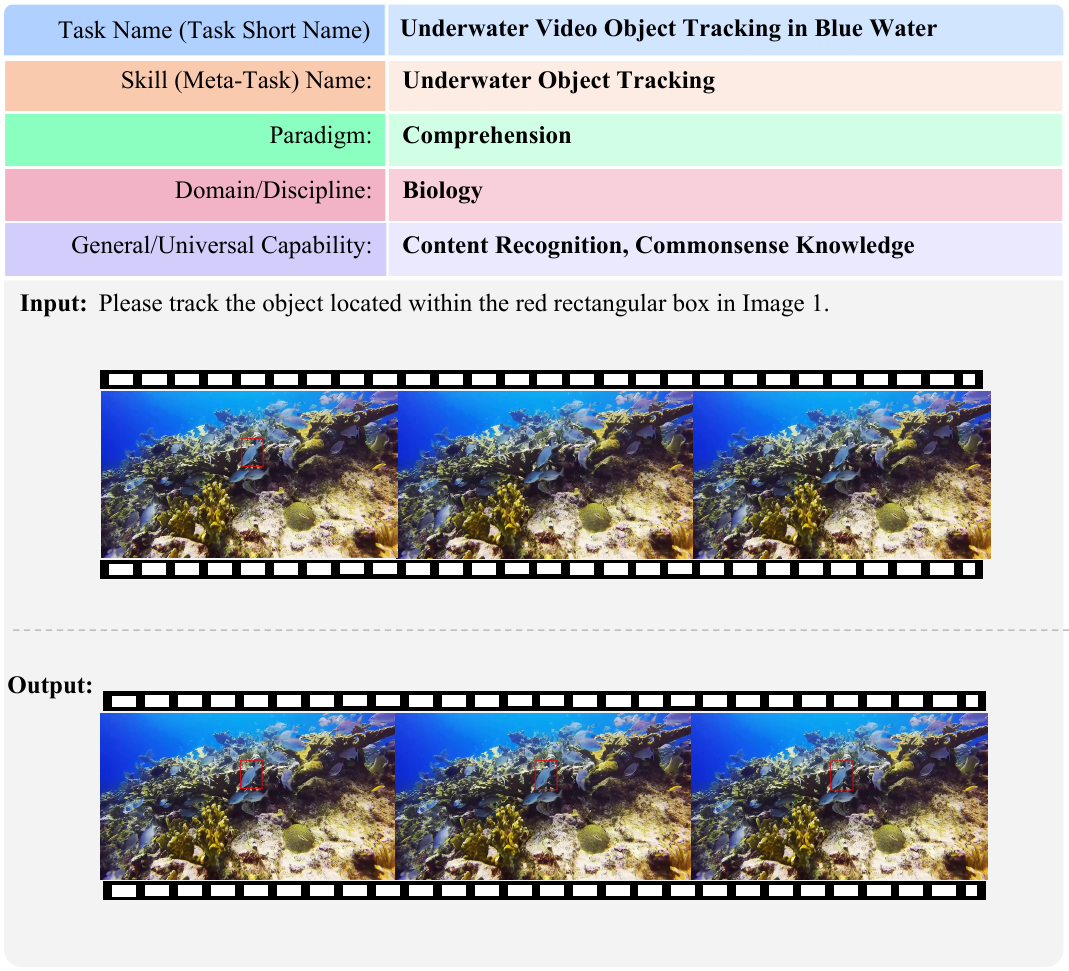}
\caption{Underwater Video Object Tracking in Blue Water.}
\label{fig:case-V-C-18}
\vspace{-2mm}
\end{figure*}

\begin{figure*}[!h]
\centering
\includegraphics[width=0.9\linewidth]{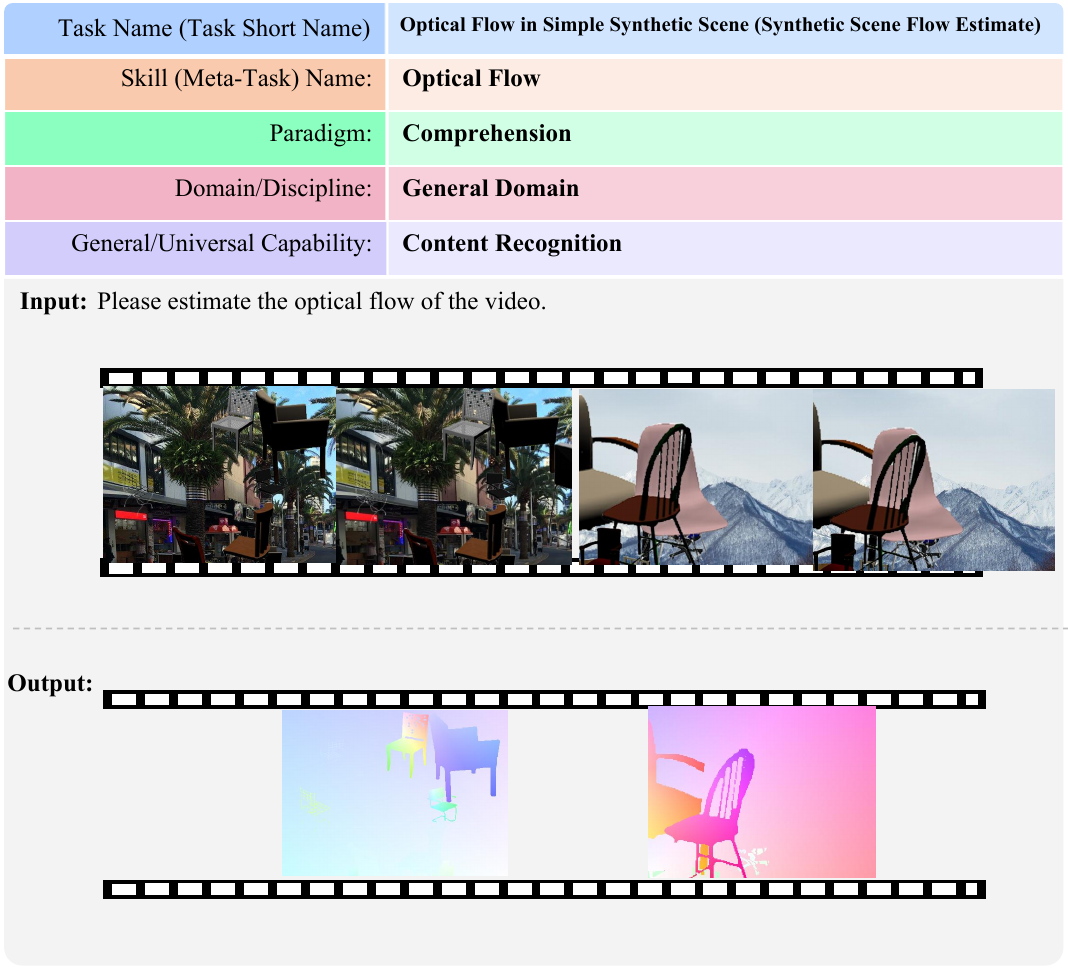}
\caption{Optical Flow in Simple Synthetic Scene.}
\label{fig:case-V-C-19}
\vspace{-2mm}
\end{figure*}

\begin{figure*}[!h]
\centering
\includegraphics[width=0.9\linewidth]{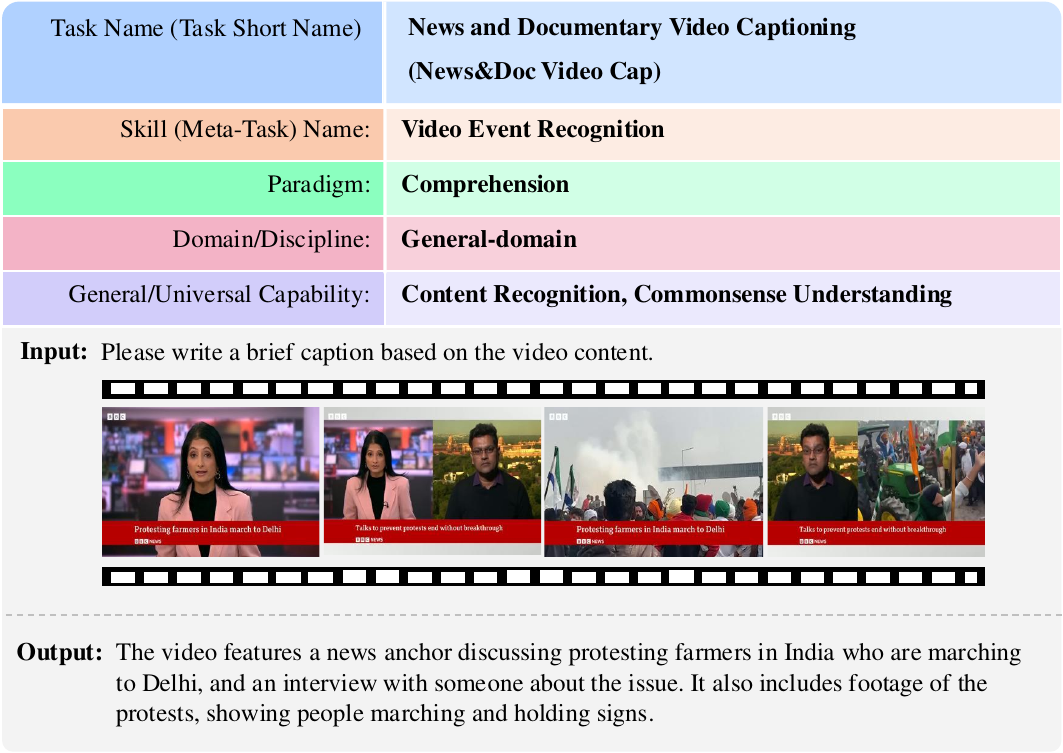}
\caption{News and Document Video Captioning.}
\label{fig:case-V-C-20}
\vspace{-2mm}
\end{figure*}

\clearpage

\paragraph{Generation Tasks.}
Following, we showcase 6 video-oriented generative tasks, each representing a specific skill.

\begin{figure*}[!h]
\centering
\includegraphics[width=0.9\linewidth]{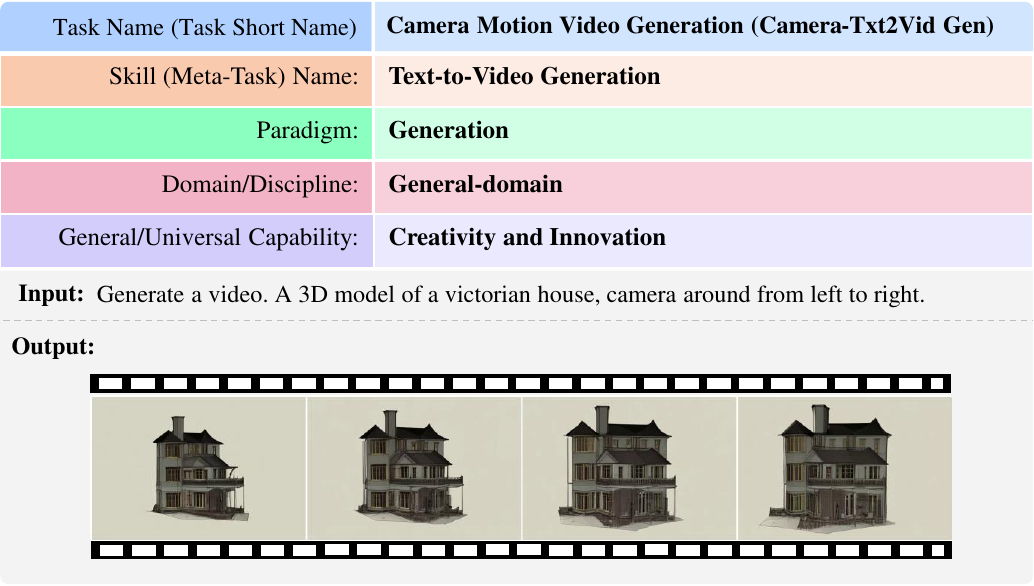}
\caption{Camera Motion Video Generation.}
\label{fig:case-V-G-1}
\vspace{-2mm}
\end{figure*}

\begin{figure*}[!h]
\centering
\includegraphics[width=0.9\linewidth]{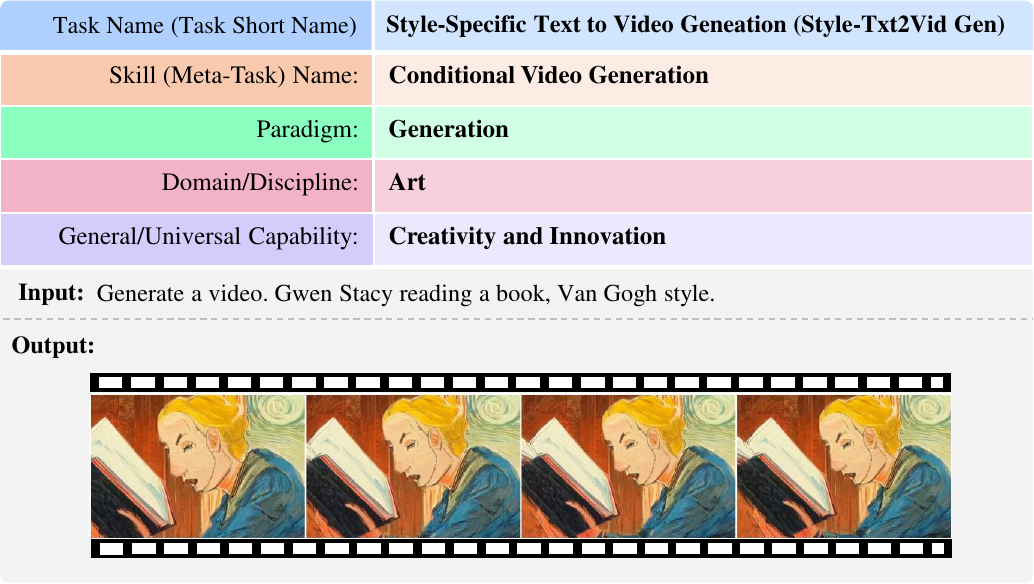}
\caption{Style-Specific Text to Video Generation.}
\label{fig:case-V-G-2}
\vspace{-2mm}
\end{figure*}

\begin{figure*}[!h]
\centering
\includegraphics[width=0.9\linewidth]{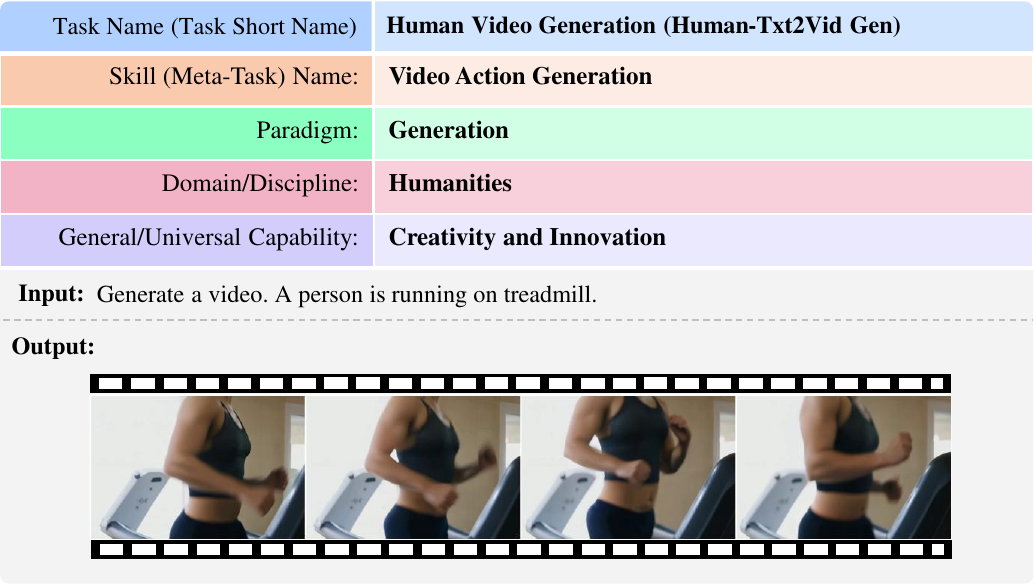}
\caption{Human Video Generation.}
\label{fig:case-V-G-3}
\vspace{-2mm}
\end{figure*}

\begin{figure*}[!h]
\centering
\includegraphics[width=0.9\linewidth]{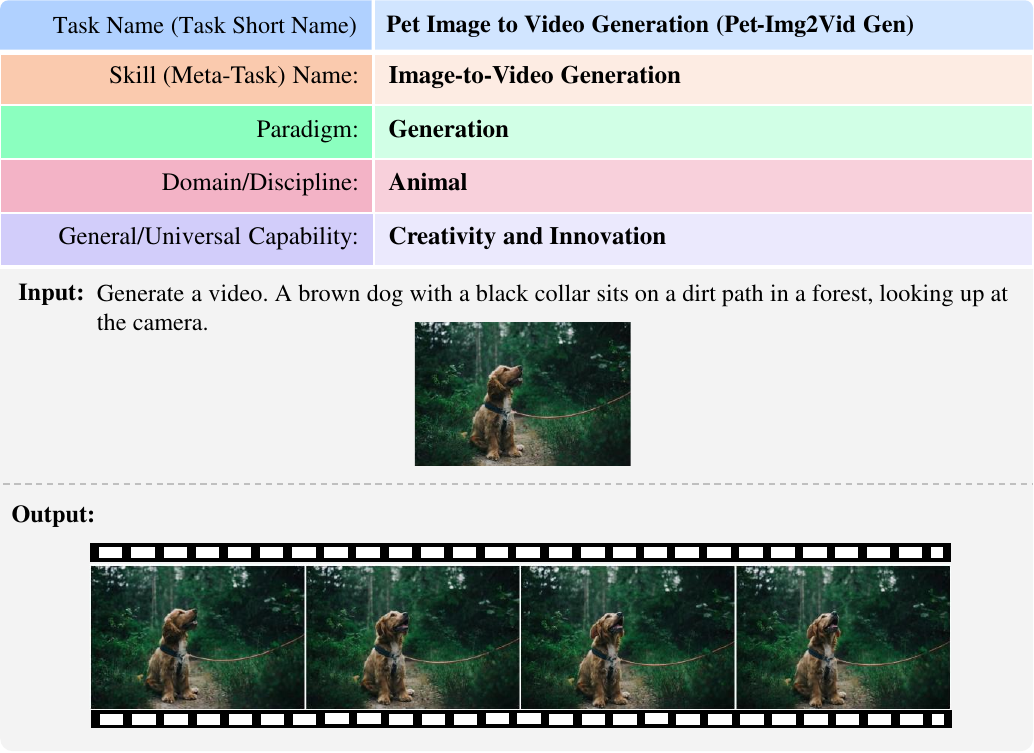}
\caption{Pet Image to Video Generation.}
\label{fig:case-V-G-4}
\vspace{-2mm}
\end{figure*}

\begin{figure*}[!h]
\centering
\includegraphics[width=0.9\linewidth]{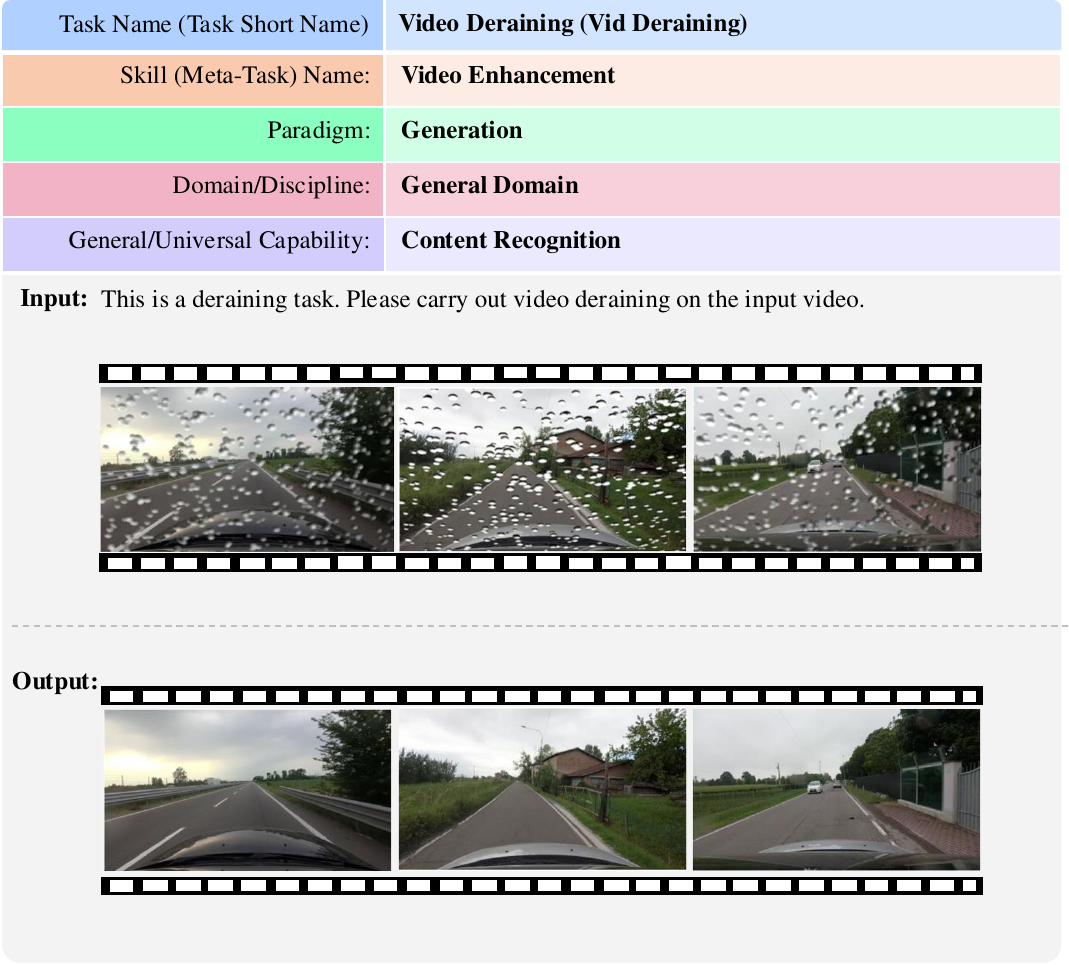}
\caption{Video Deraining.}
\label{fig:case-V-G-5}
\vspace{-2mm}
\end{figure*}

\begin{figure*}[!h]
\centering
\includegraphics[width=0.9\linewidth]{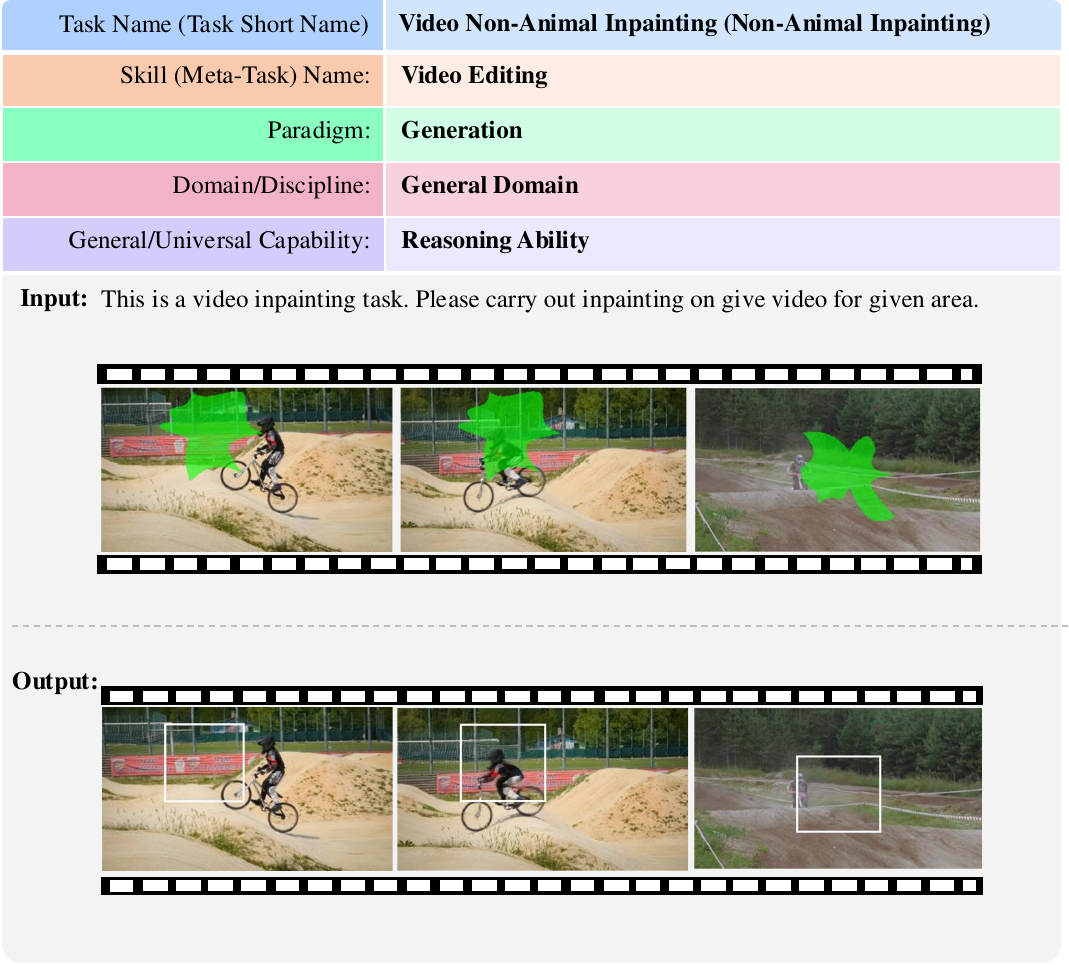}
\caption{Video Non-Animal Inpainting.}
\label{fig:case-V-G-6}
\vspace{-2mm}
\end{figure*}

\clearpage

\subsubsection{Audio-related Tasks}

\paragraph{Comprehension Tasks.}
Following we showcase 9 audio-oriented comprehensive tasks, each representing a specific skill.

\begin{figure*}[!h]
\centering
\includegraphics[width=0.9\linewidth]{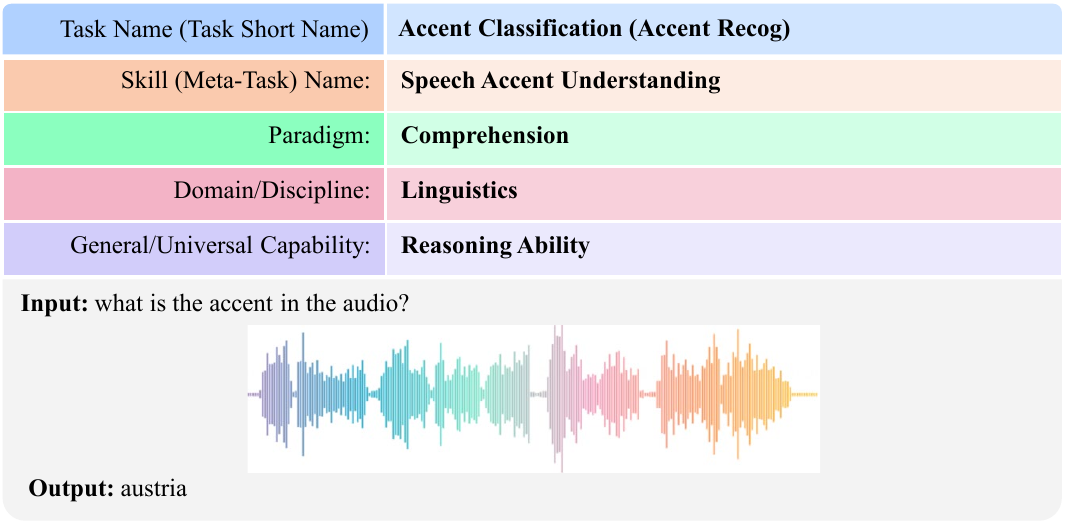}
\caption{Accent Classification.}
\label{fig:A-C-1}
\vspace{-2mm}
\end{figure*}


\begin{figure*}[!h]
\centering
\includegraphics[width=0.9\linewidth]{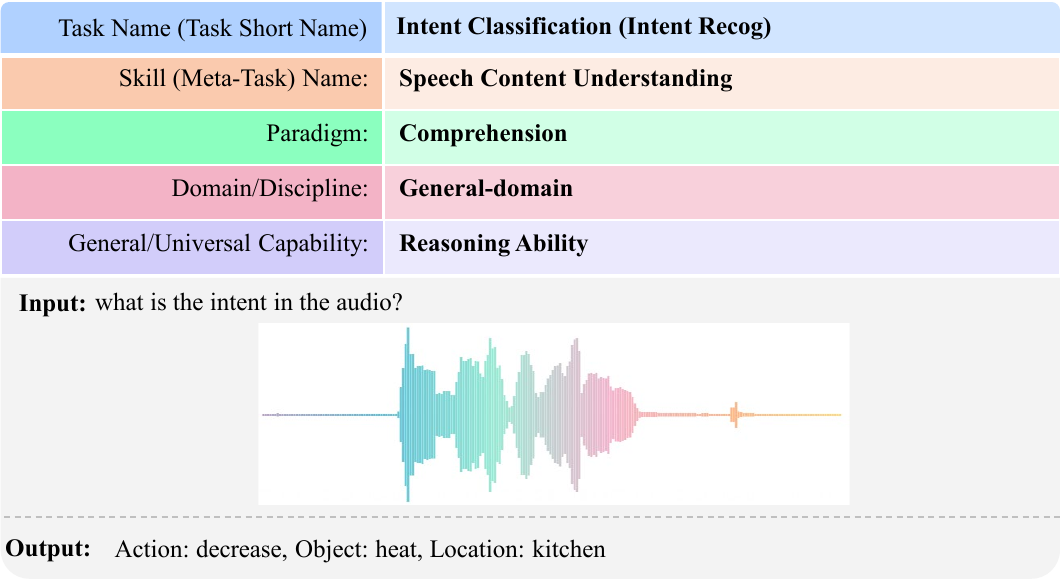}
\caption{Intent Classification.}
\label{fig:A-C-2}
\vspace{-2mm}
\end{figure*}


\begin{figure*}[!h]
\centering
\includegraphics[width=0.9\linewidth]{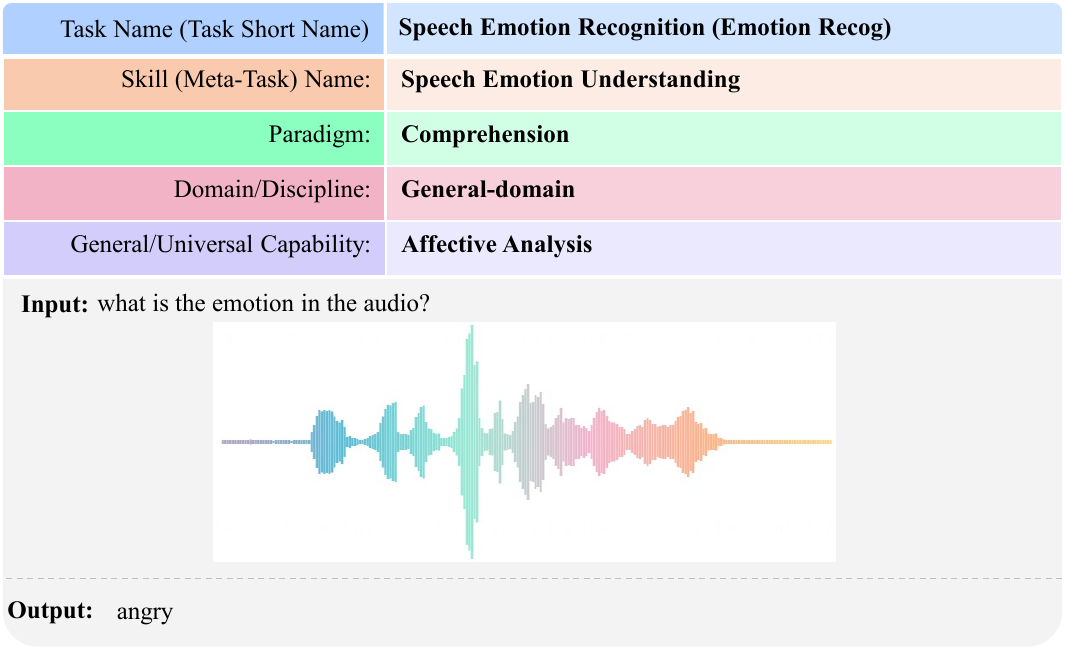}
\caption{Speech Emotion Recognition.}
\label{fig:A-C-3}
\vspace{-2mm}
\end{figure*}


\begin{figure*}[!h]
\centering
\includegraphics[width=0.9\linewidth]{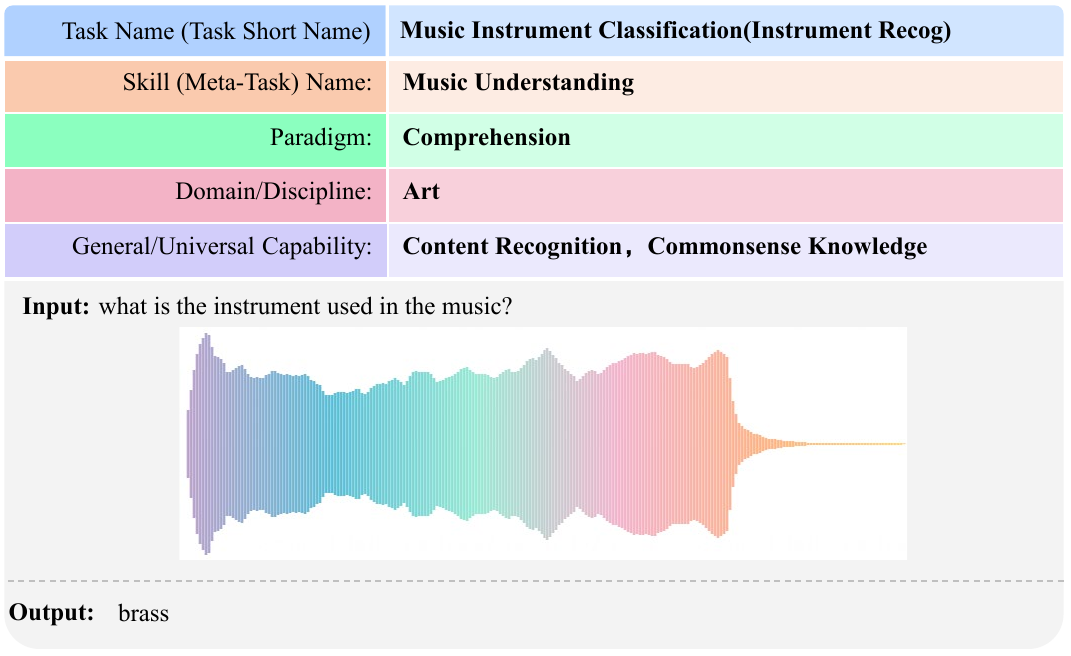}
\caption{Music Instrument Classification.}
\label{fig:A-C-4}
\vspace{-2mm}
\end{figure*}


\begin{figure*}[!h]
\centering
\includegraphics[width=0.9\linewidth]{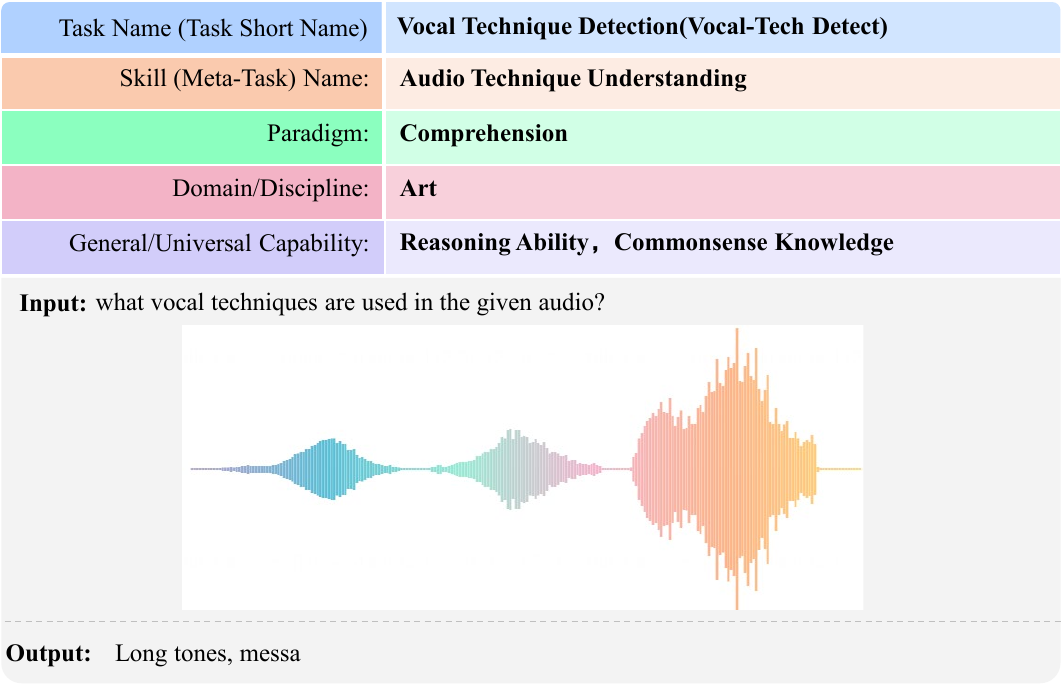}
\caption{Vocal Technique Detection.}
\label{fig:A-C-5}
\vspace{-2mm}
\end{figure*}


\begin{figure*}[!h]
\centering
\includegraphics[width=0.9\linewidth]{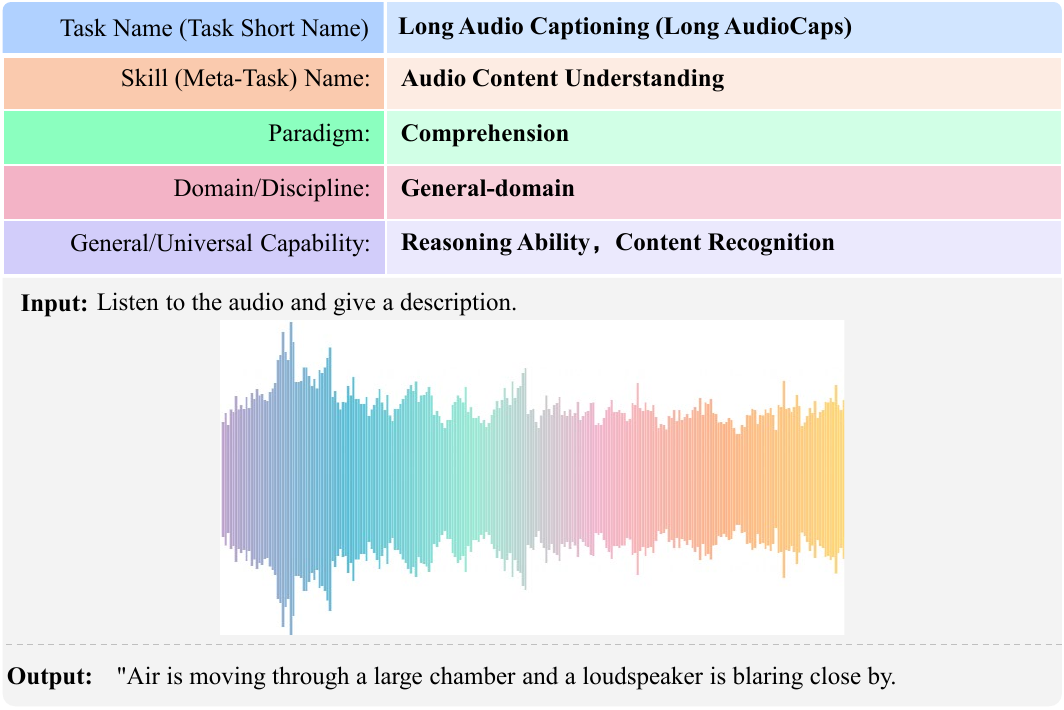}
\caption{Long Audio Captioning.}
\label{fig:A-C-6}
\vspace{-2mm}
\end{figure*}


\begin{figure*}[!h]
\centering
\includegraphics[width=0.9\linewidth]{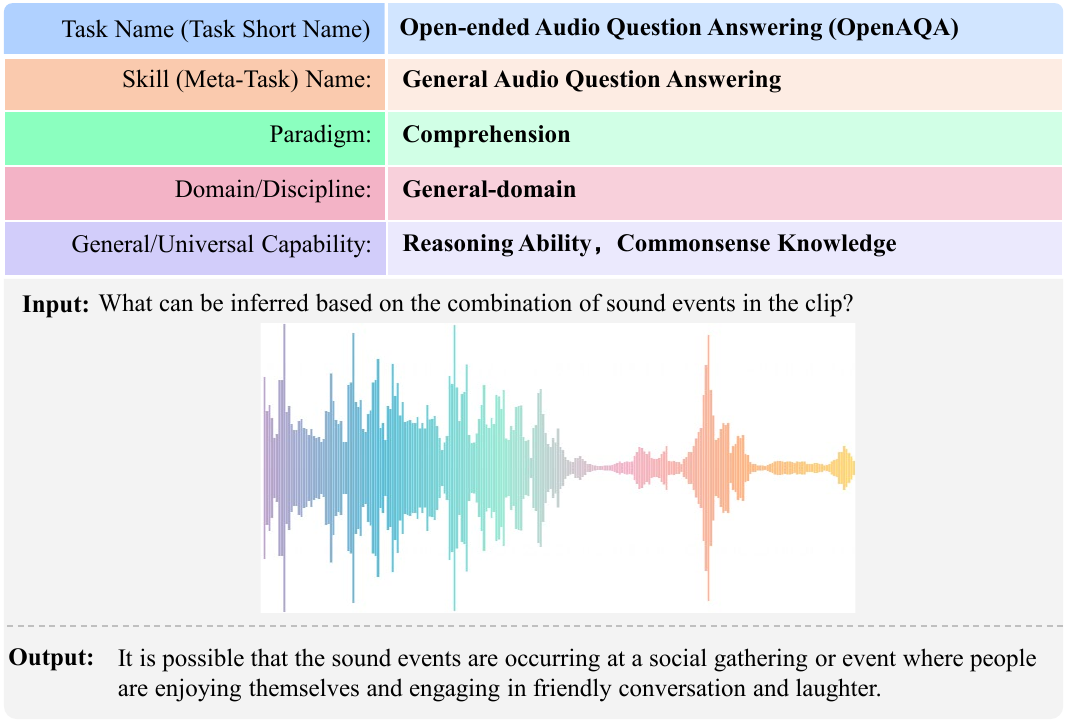}
\caption{Open-ended Audio Question Answering.}
\label{fig:A-C-7}
\vspace{-2mm}
\end{figure*}


\begin{figure*}[!h]
\centering
\includegraphics[width=0.9\linewidth]{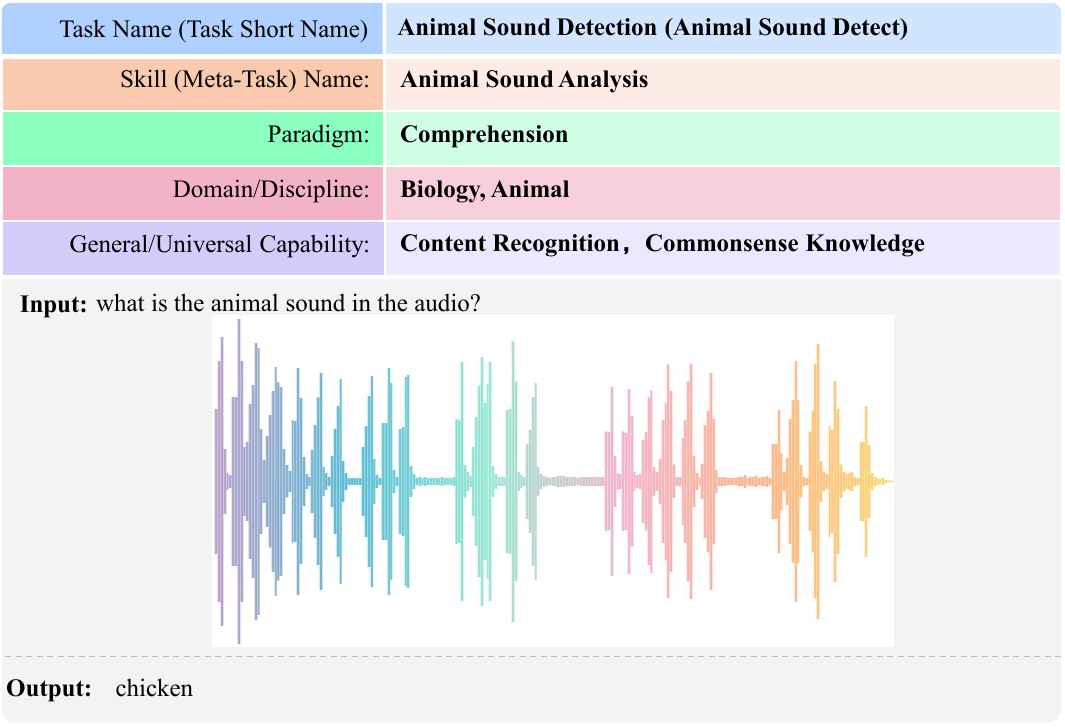}
\caption{Animal Sound Analysis.}
\label{fig:A-C-8}
\vspace{-2mm}
\end{figure*}


\begin{figure*}[!h]
\centering
\includegraphics[width=0.9\linewidth]{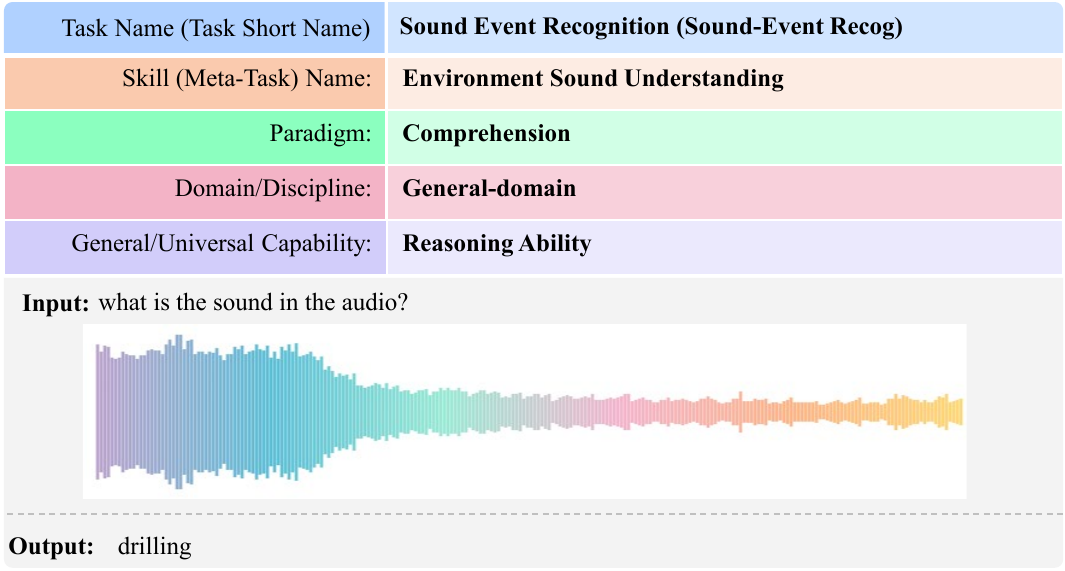}
\caption{Sound Event Recognition.}
\label{fig:A-C-9}
\vspace{-2mm}
\end{figure*}

\clearpage

\paragraph{Generation Tasks.}
Following we showcase 11 audio-oriented generative tasks, each representing a specific skill.

\begin{figure*}[!h]
\centering
\includegraphics[width=0.9\linewidth]{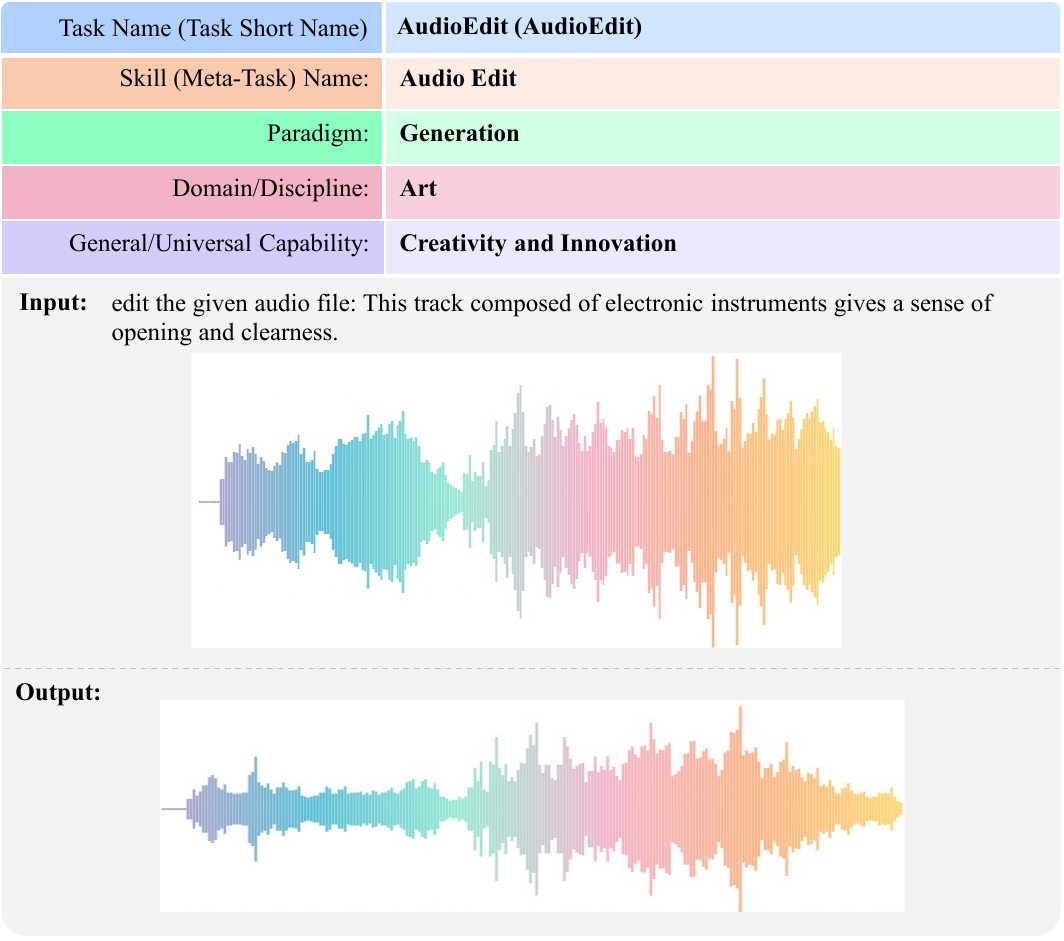}
\caption{Audio Editing.}
\label{fig:A-G-1}
\vspace{-2mm}
\end{figure*}


\begin{figure*}[!h]
\centering
\includegraphics[width=0.9\linewidth]{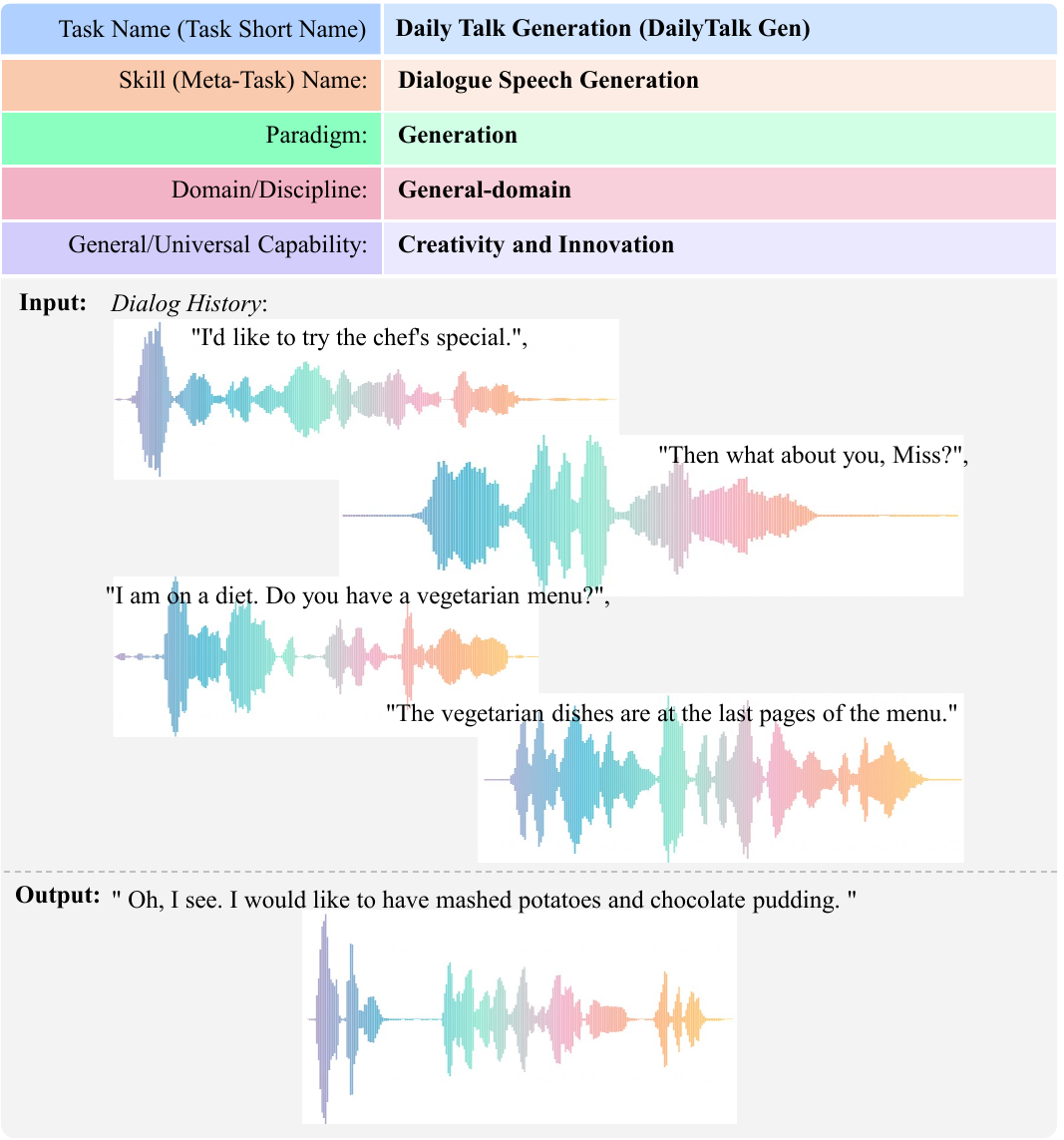}
\caption{Daily Talk Generation.}
\label{fig:A-G-2}
\vspace{-2mm}
\end{figure*}


\begin{figure*}[!h]
\centering
\includegraphics[width=0.9\linewidth]{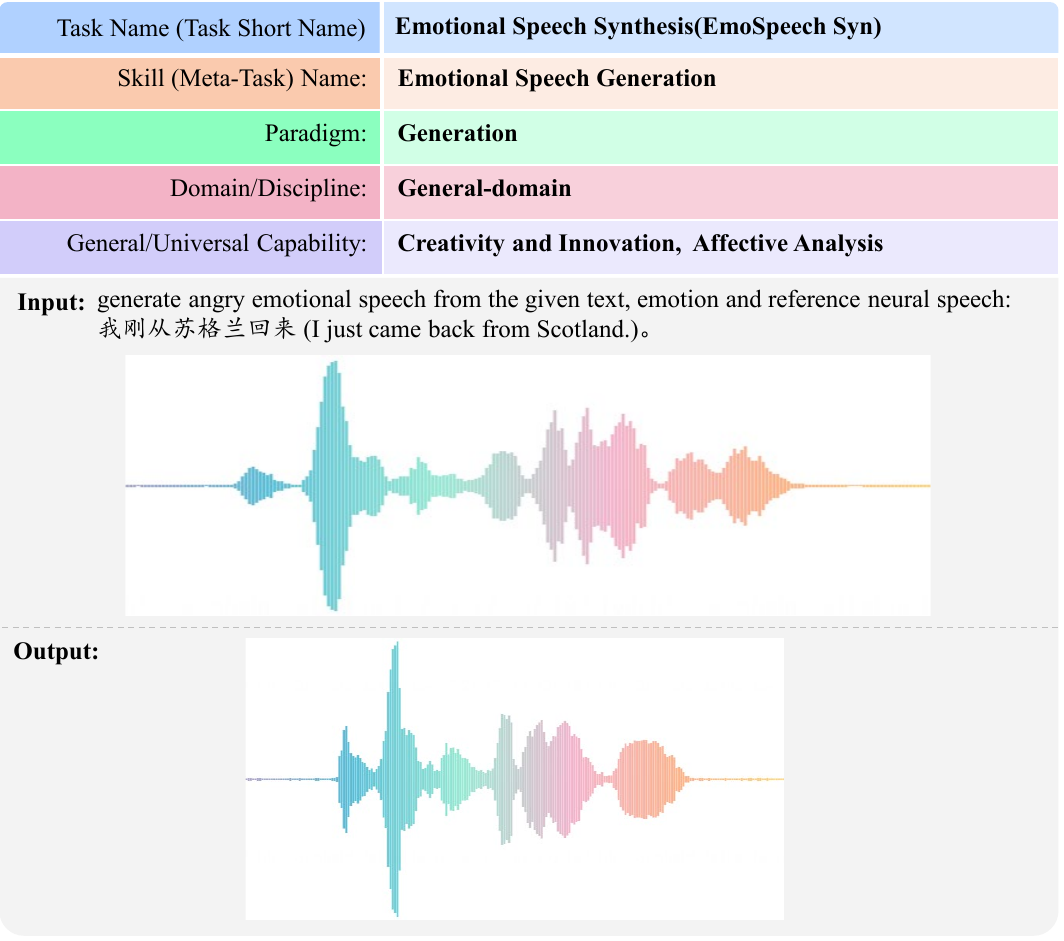}
\caption{Emotional Speech Synthesis.}
\label{fig:A-G-3}
\vspace{-2mm}
\end{figure*}


\begin{figure*}[!h]
\centering
\includegraphics[width=0.9\linewidth]{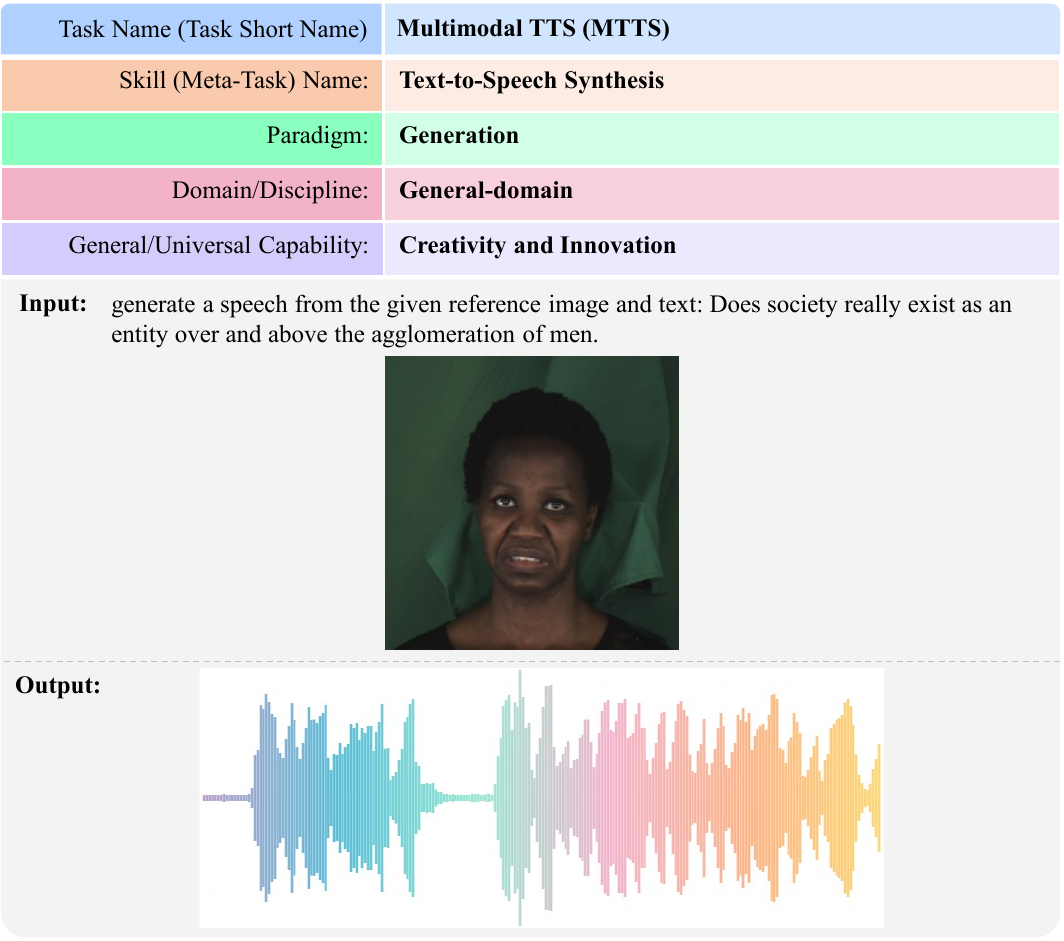}
\caption{Multimodal Text-to-Speech (TTS).}
\label{fig:A-G-4}
\vspace{-2mm}
\end{figure*}


\begin{figure*}[!h]
\centering
\includegraphics[width=0.9\linewidth]{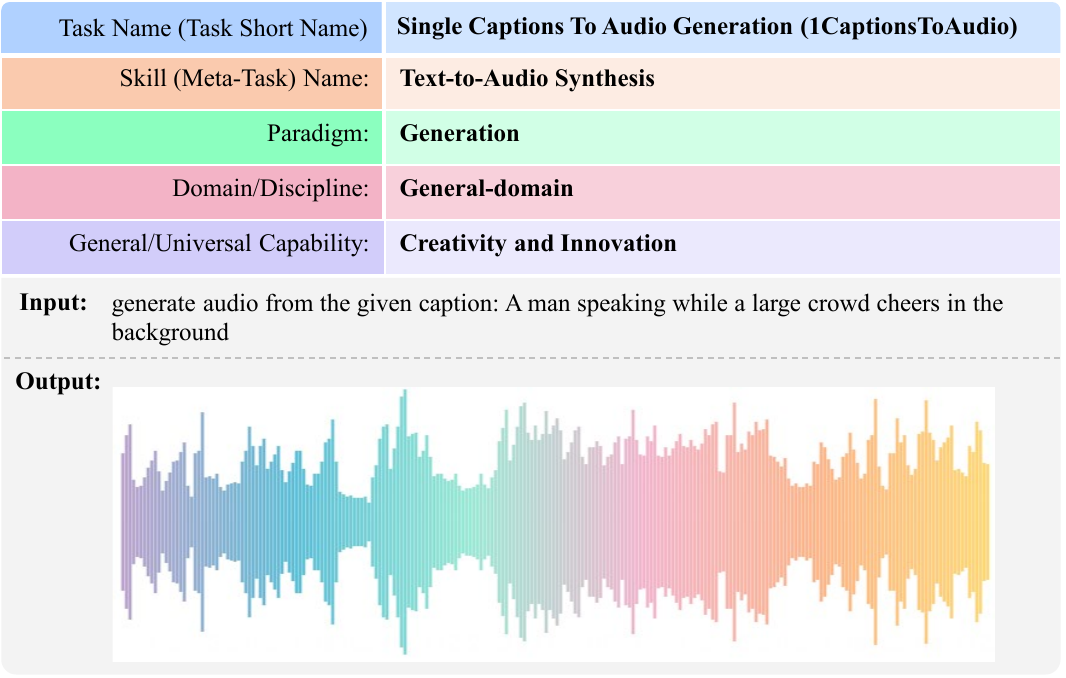}
\caption{Single Captions To Audio Generation.}
\label{fig:A-G-5}
\vspace{-2mm}
\end{figure*}


\begin{figure*}[!h]
\centering
\includegraphics[width=0.9\linewidth]{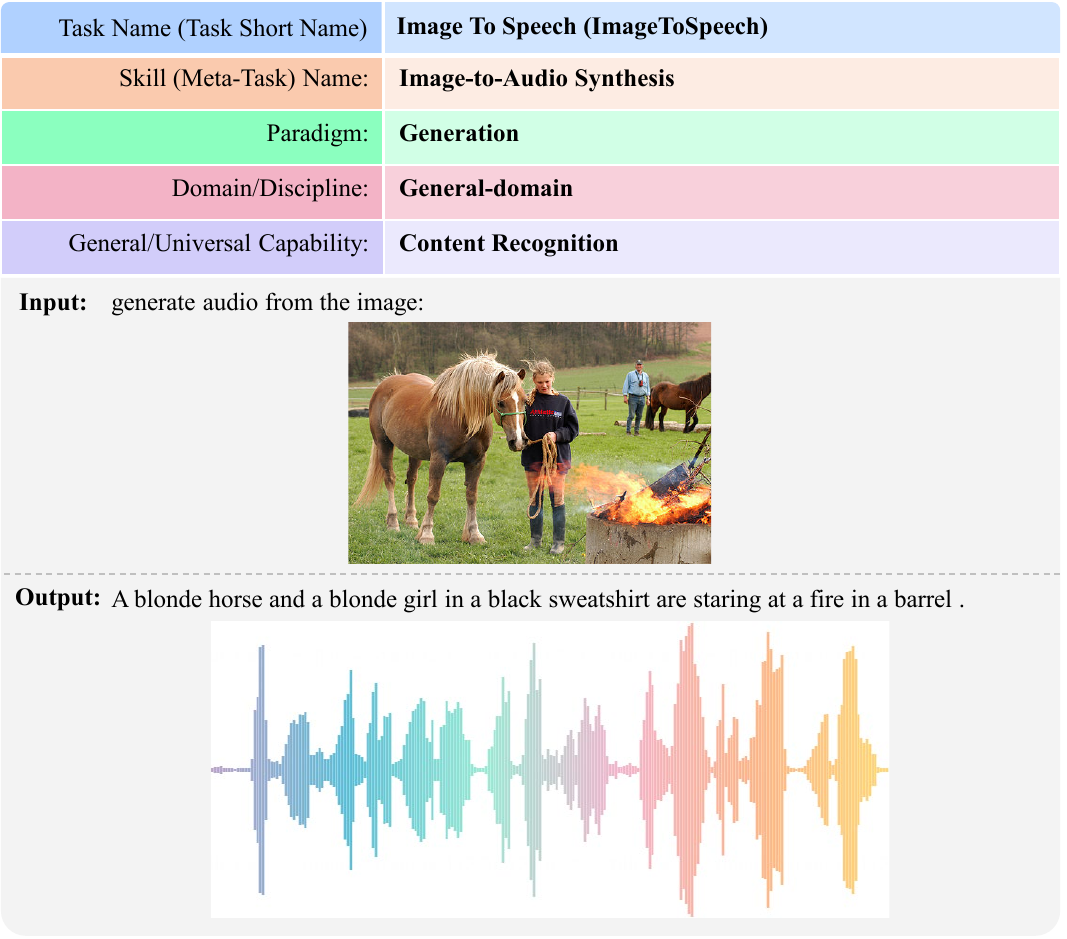}
\caption{Image-to-Speech Synthesis.}
\label{fig:A-G-6}
\vspace{-2mm}
\end{figure*}


\begin{figure*}[!h]
\centering
\includegraphics[width=0.9\linewidth]{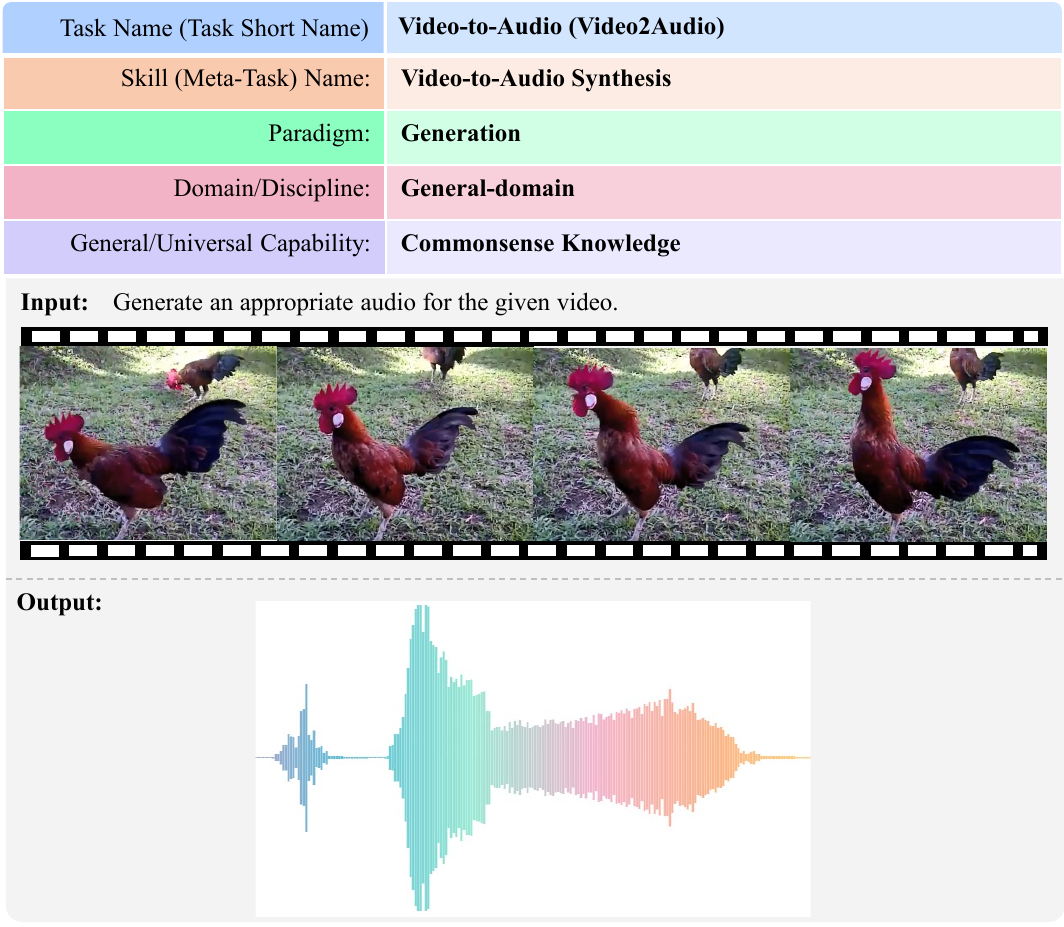}
\caption{Video-to-Audio Synthesis.}
\label{fig:A-G-7}
\vspace{-2mm}
\end{figure*}


\begin{figure*}[!h]
\centering
\includegraphics[width=0.9\linewidth]{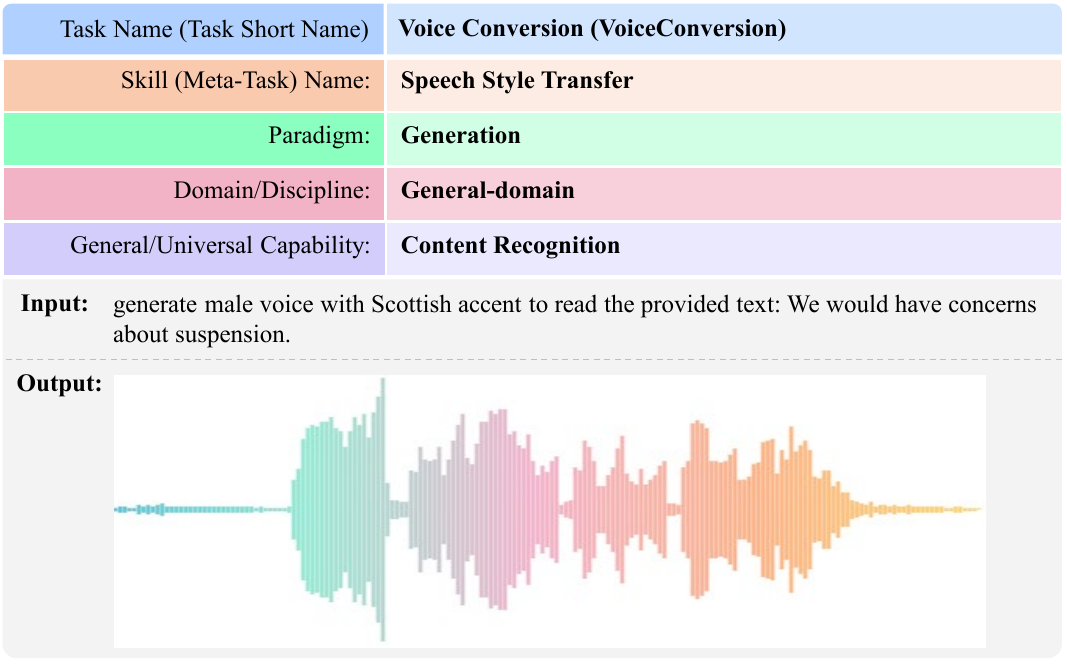}
\caption{Voice Conversion.}
\label{fig:A-G-8}
\vspace{-2mm}
\end{figure*}


\begin{figure*}[!h]
\centering
\includegraphics[width=0.9\linewidth]{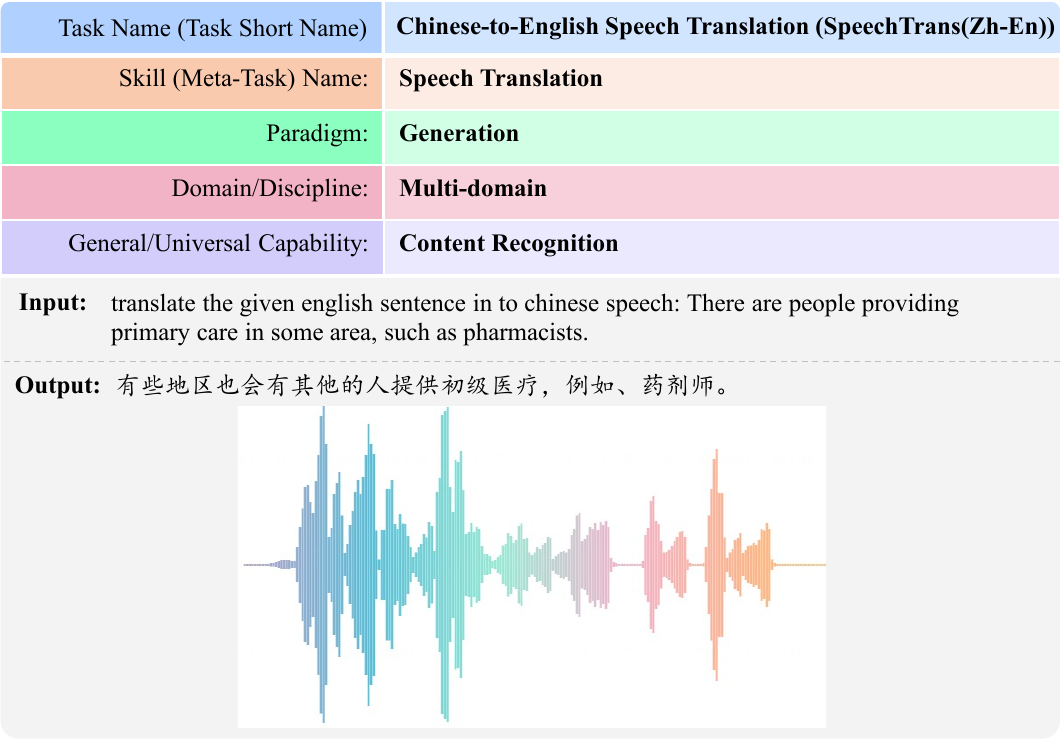}
\caption{Chinese-to-English Speech Translation.}
\label{fig:A-G-9}
\vspace{-2mm}
\end{figure*}


\begin{figure*}[!h]
\centering
\includegraphics[width=0.9\linewidth]{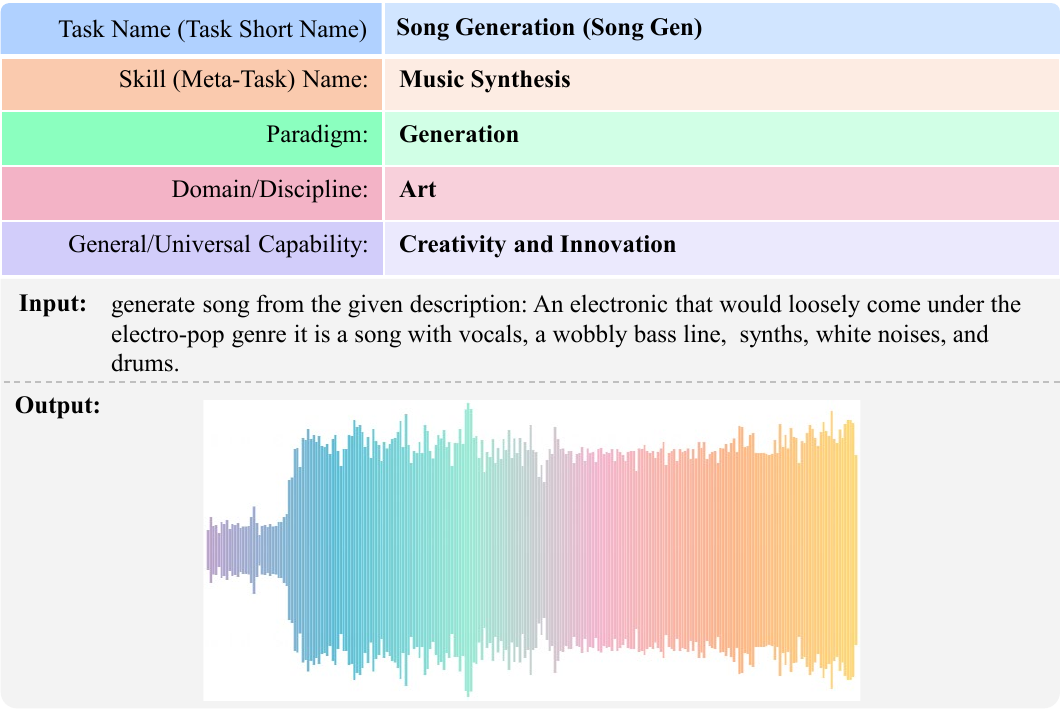}
\caption{Song Synthesis.}
\label{fig:A-G-10}
\vspace{-2mm}
\end{figure*}


\begin{figure*}[!h]
\centering
\includegraphics[width=0.9\linewidth]{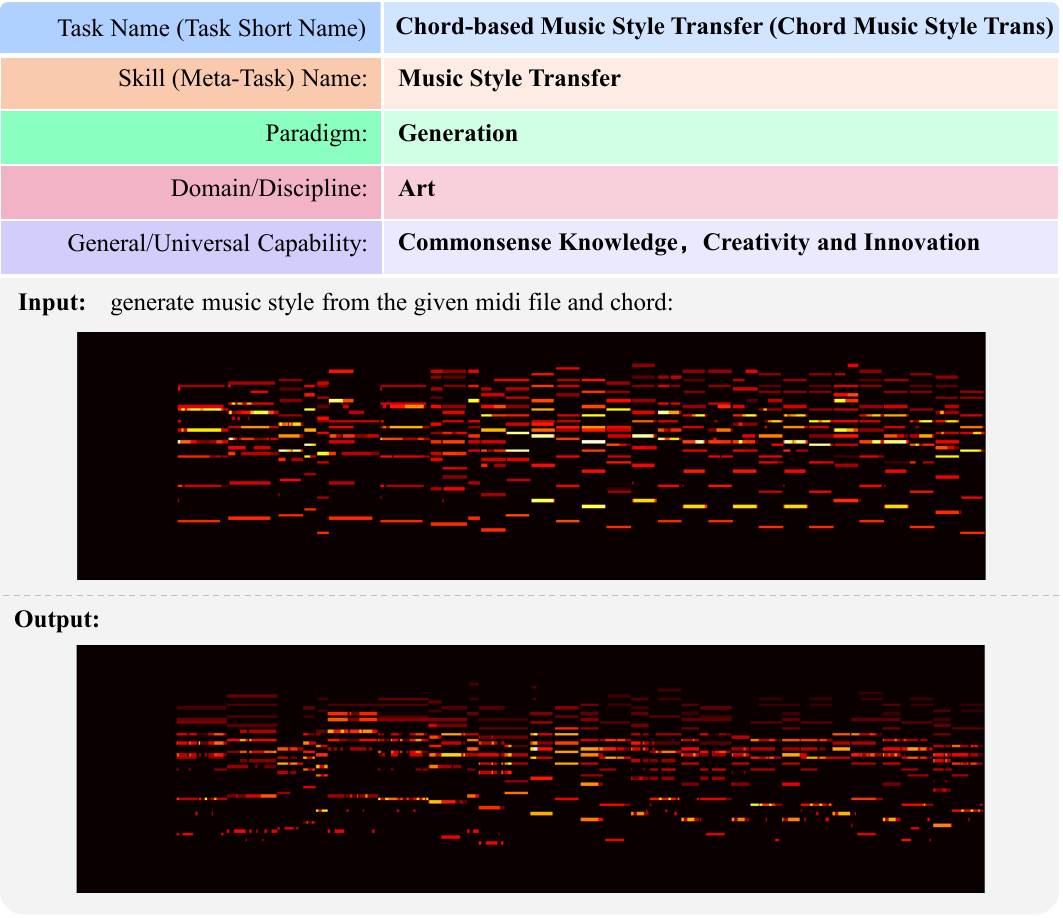}
\caption{Chord-based Music Style Transfer.}
\label{fig:A-G-11}
\vspace{-2mm}
\end{figure*}


\clearpage

\subsubsection{3D-related Tasks}

\paragraph{Comprehension Tasks.}
Following, we showcase 13 3D-oriented comprehensive tasks, each representing a specific skill.


\begin{figure*}[!h]
\centering
\includegraphics[width=0.9\linewidth]{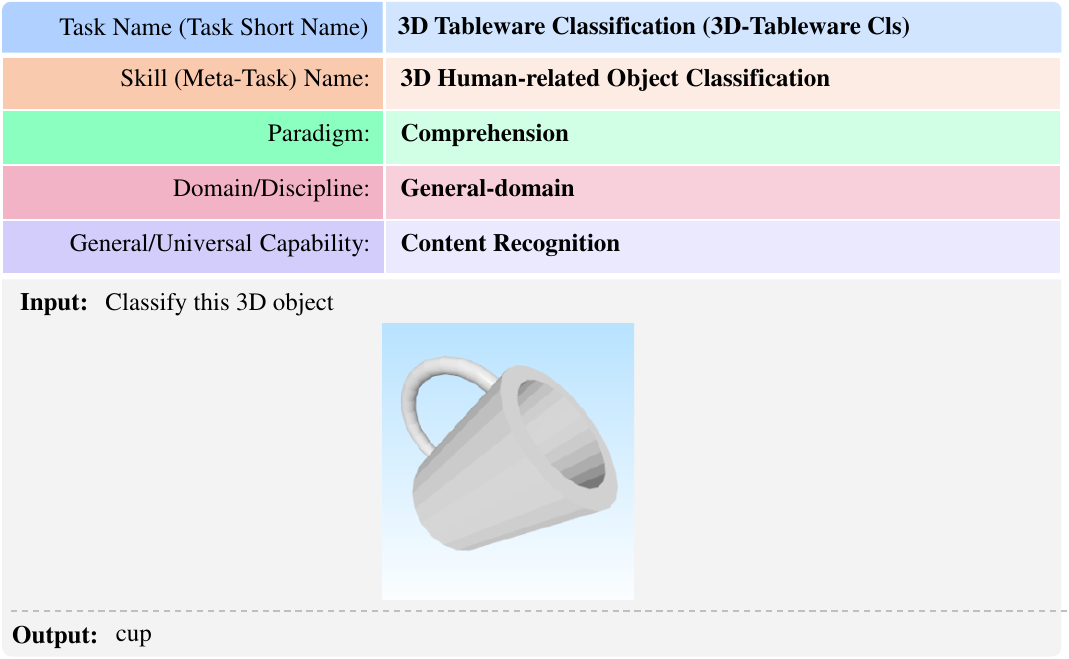}
\caption{3D Tableware Classification.}
\label{fig:D-C-1}
\vspace{-2mm}
\end{figure*}


\begin{figure*}[!h]
\centering
\includegraphics[width=0.9\linewidth]{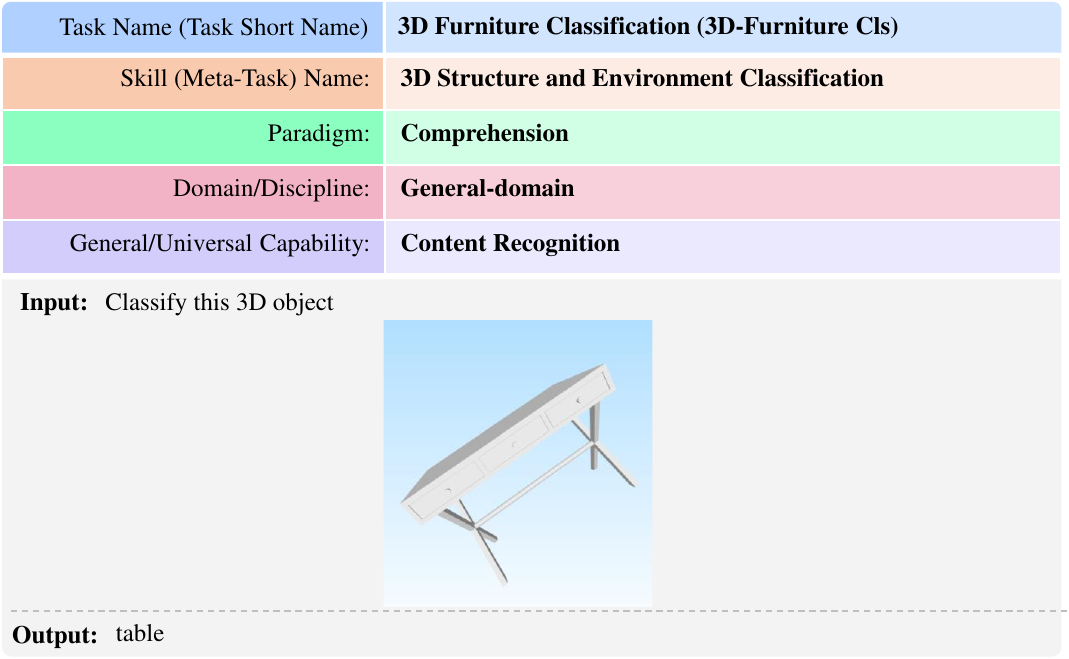}
\caption{3D Furniture Classification.}
\label{fig:D-C-2}
\vspace{-2mm}
\end{figure*}


\begin{figure*}[!h]
\centering
\includegraphics[width=0.9\linewidth]{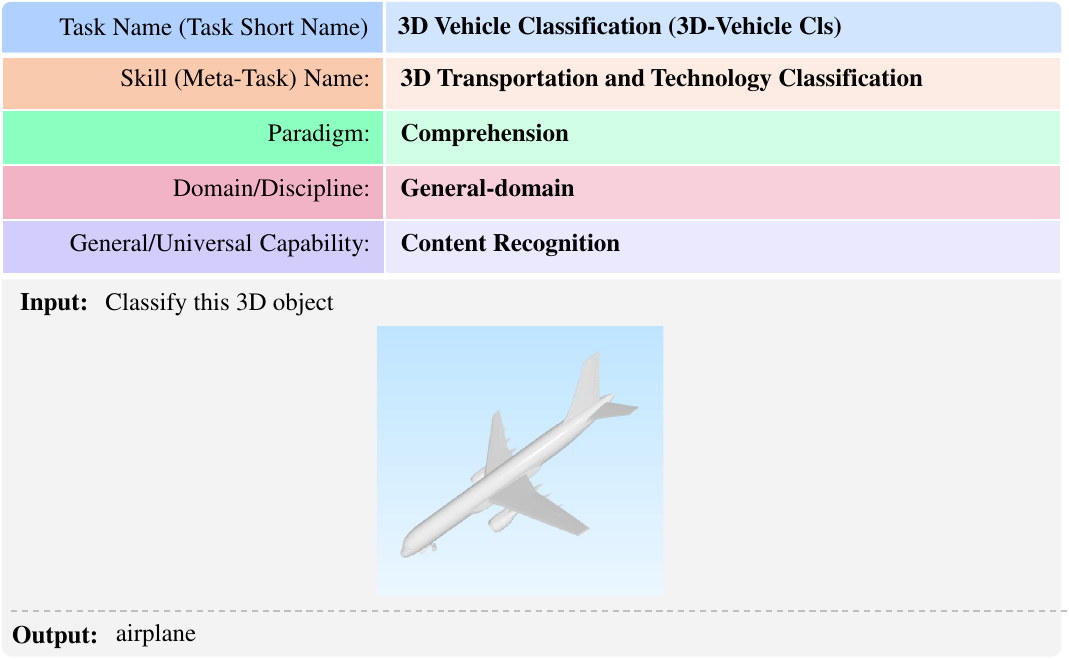}
\caption{3D Vehicle Classification.}
\label{fig:D-C-3}
\vspace{-2mm}
\end{figure*}


\begin{figure*}[!h]
\centering
\includegraphics[width=0.9\linewidth]{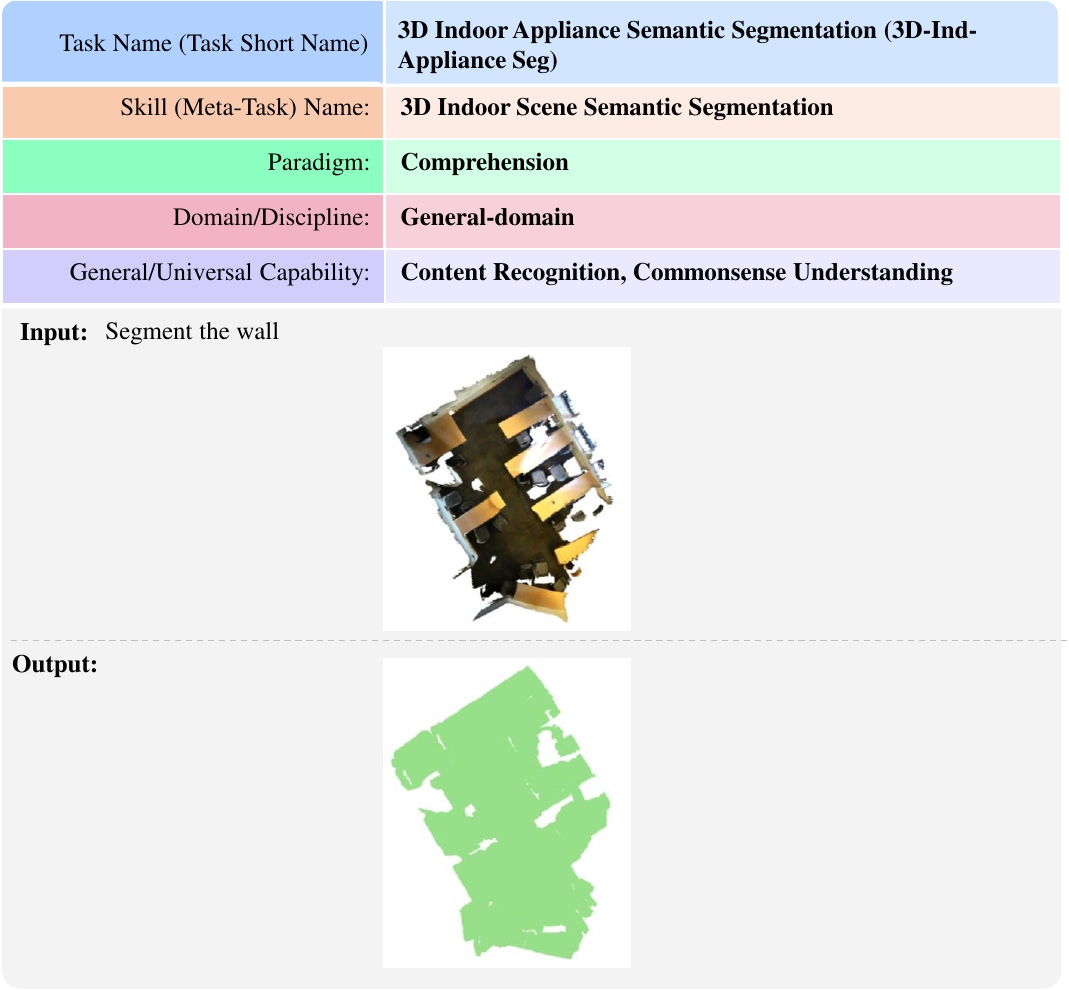}
\caption{3D Indoor Appliance Semantic Segmentation.}
\label{fig:D-C-4}
\vspace{-2mm}
\end{figure*}


\begin{figure*}[!h]
\centering
\includegraphics[width=0.9\linewidth]{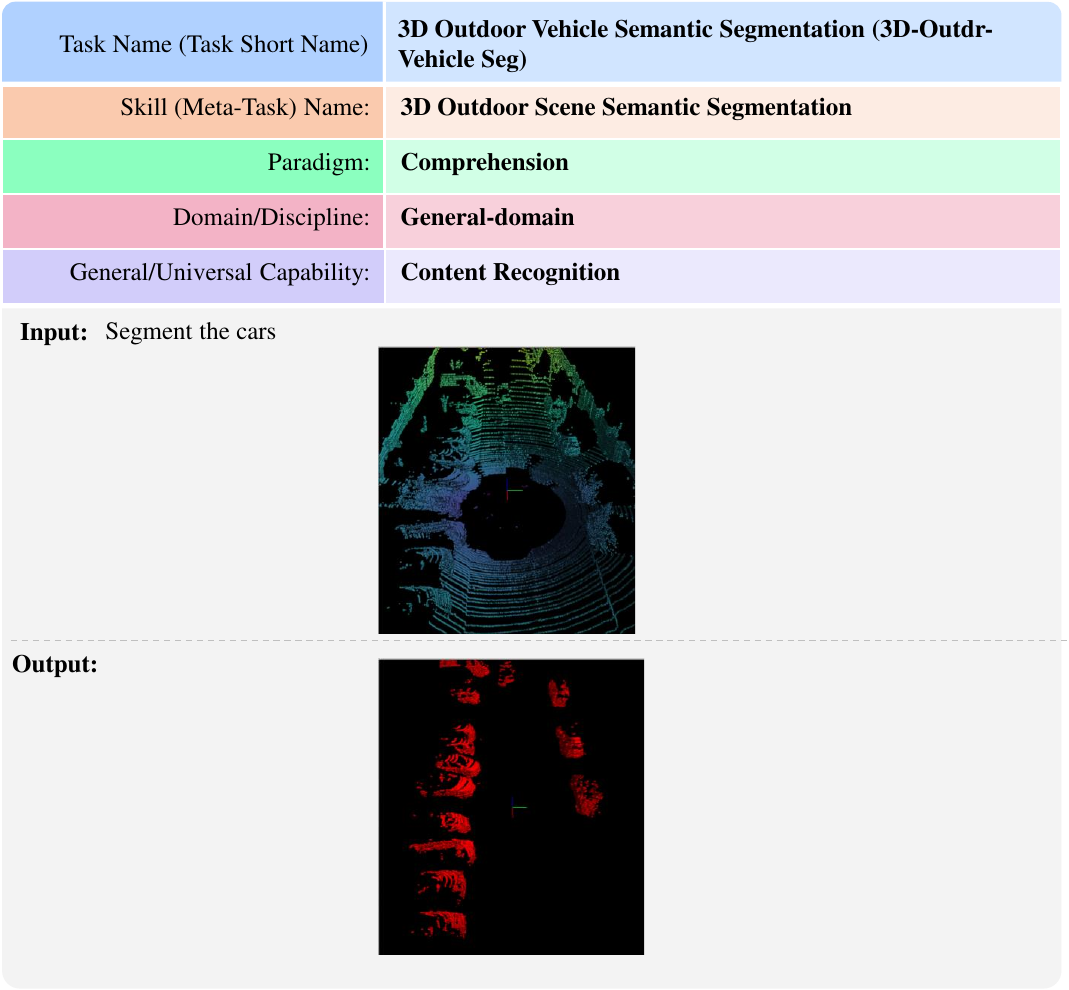}
\caption{3D Outdoor Vehicle Semantic Segmentation.}
\label{fig:D-C-5}
\vspace{-2mm}
\end{figure*}


\begin{figure*}[!h]
\centering
\includegraphics[width=0.9\linewidth]{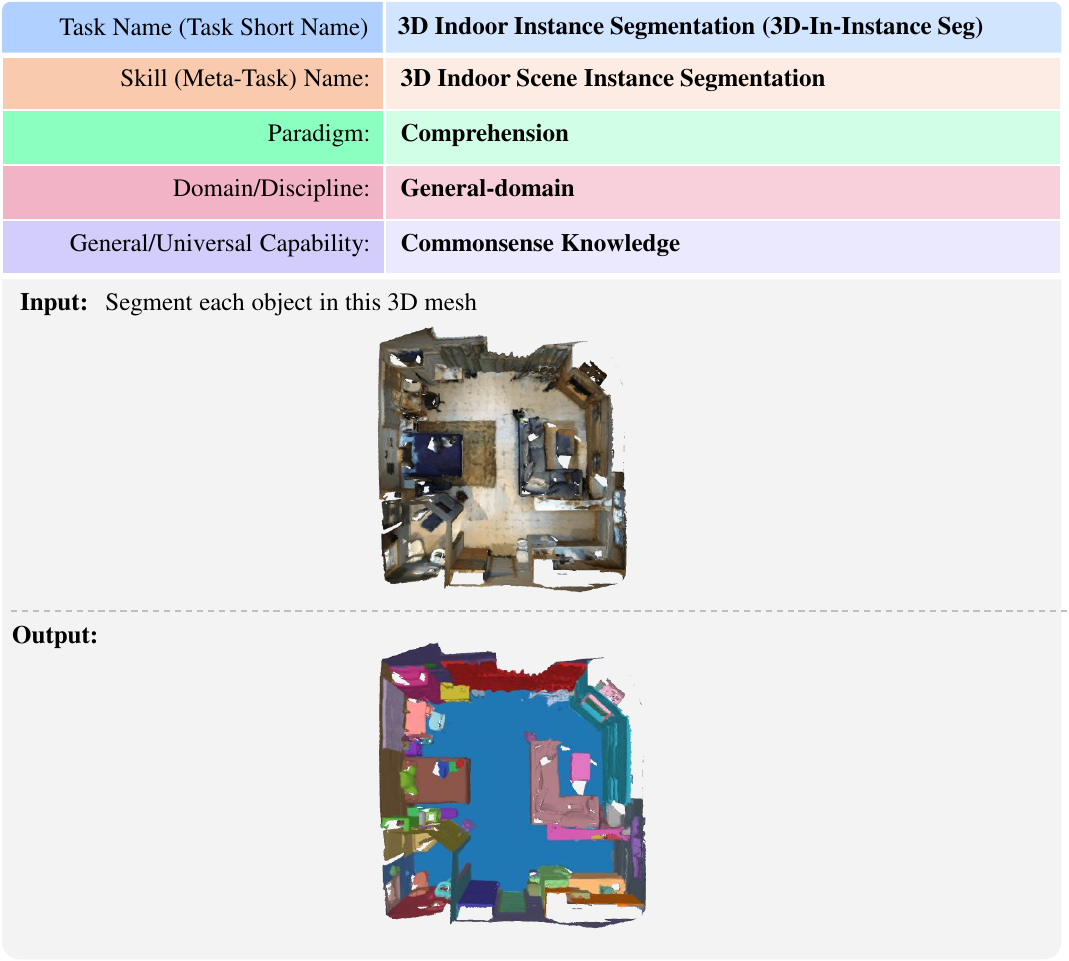}
\caption{3D Indoor Instance Segmentation.}
\label{fig:D-C-6}
\vspace{-2mm}
\end{figure*}


\begin{figure*}[!h]
\centering
\includegraphics[width=0.9\linewidth]{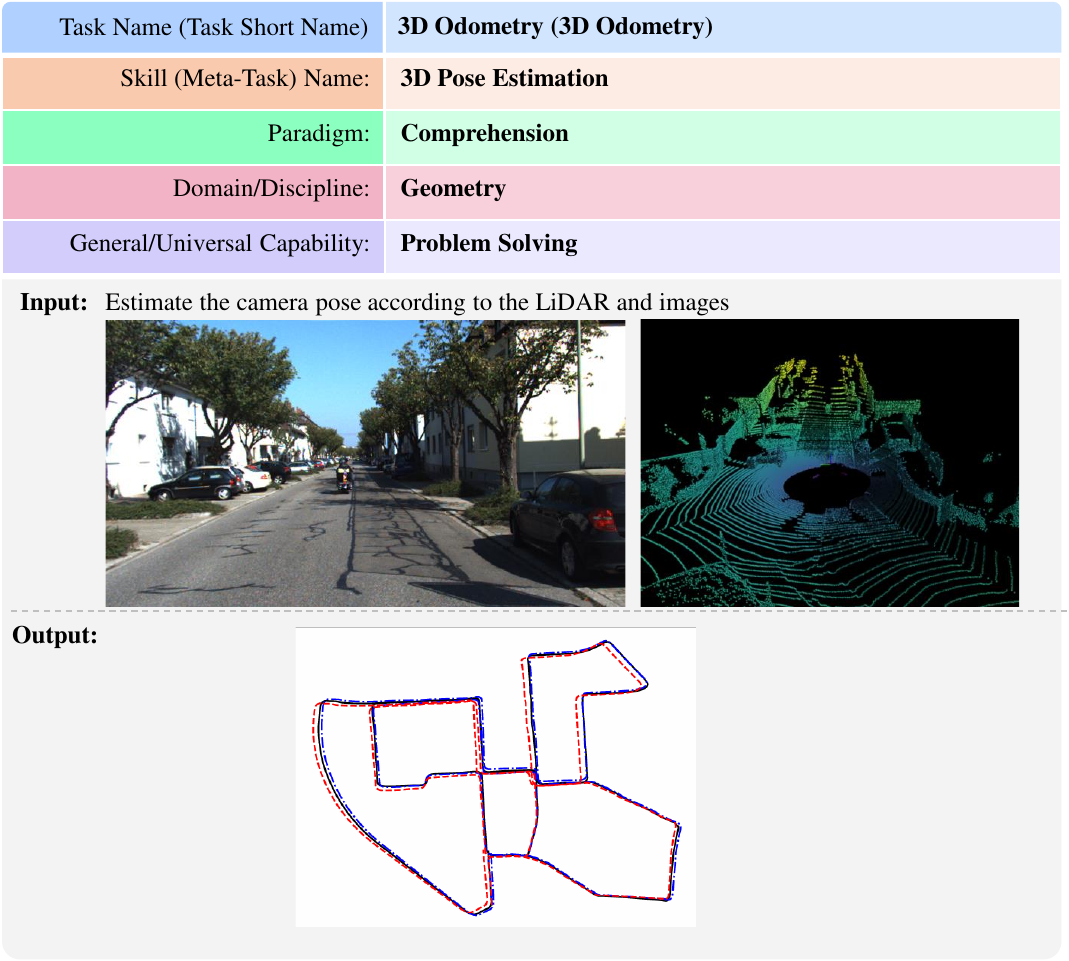}
\caption{3D Odometry.}
\label{fig:D-C-7}
\vspace{-2mm}
\end{figure*}


\begin{figure*}[!h]
\centering
\includegraphics[width=0.9\linewidth]{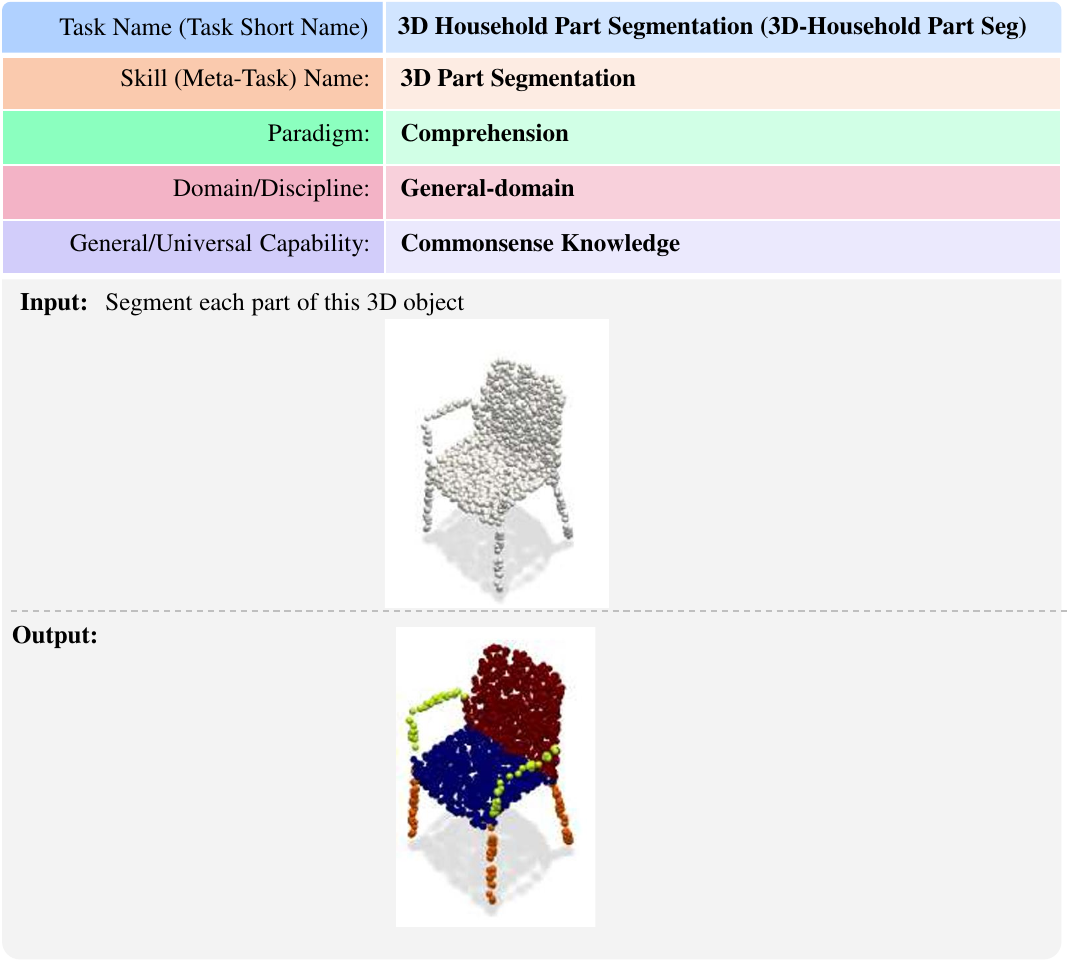}
\caption{3D Household Part Segmentation.}
\label{fig:D-C-8}
\vspace{-2mm}
\end{figure*}


\begin{figure*}[!h]
\centering
\includegraphics[width=0.9\linewidth]{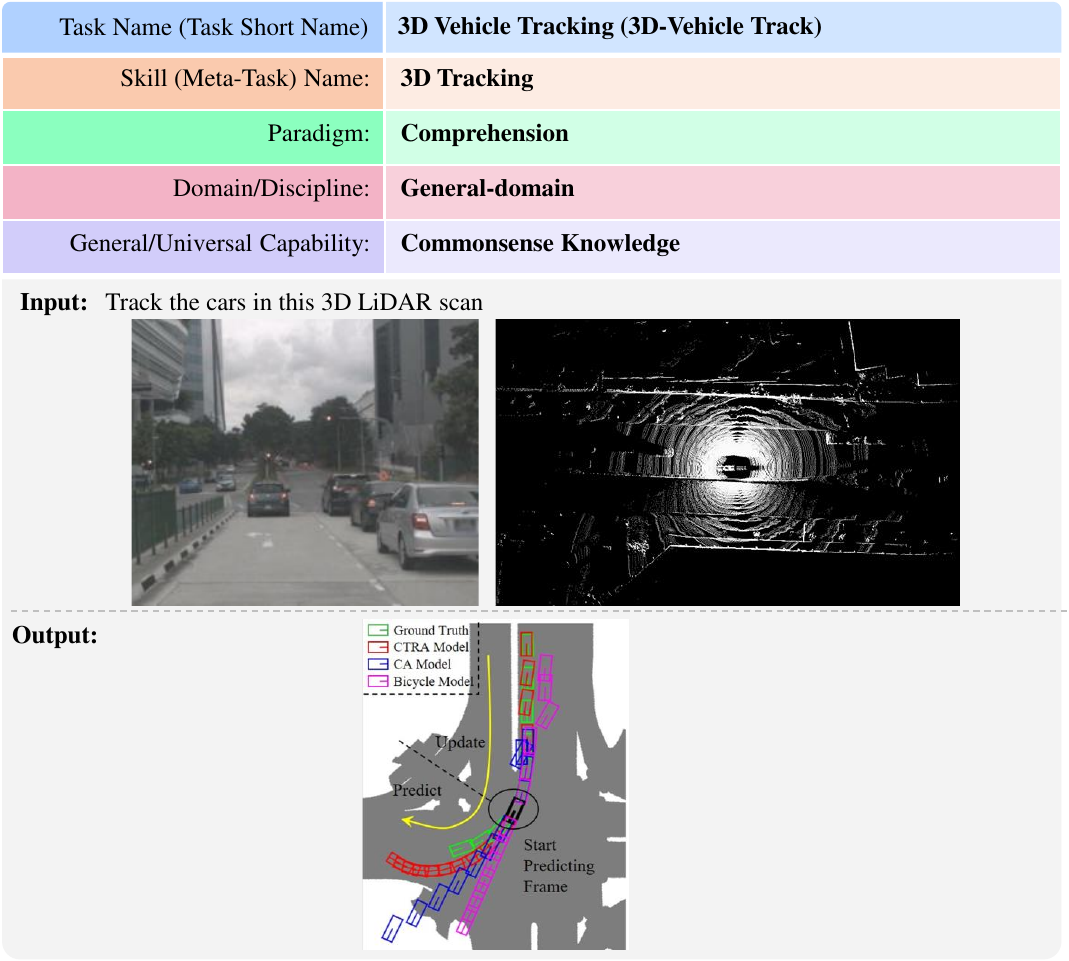}
\caption{3D Vehicle Tracking.}
\label{fig:D-C-9}
\vspace{-2mm}
\end{figure*}


\begin{figure*}[!h]
\centering
\includegraphics[width=0.9\linewidth]{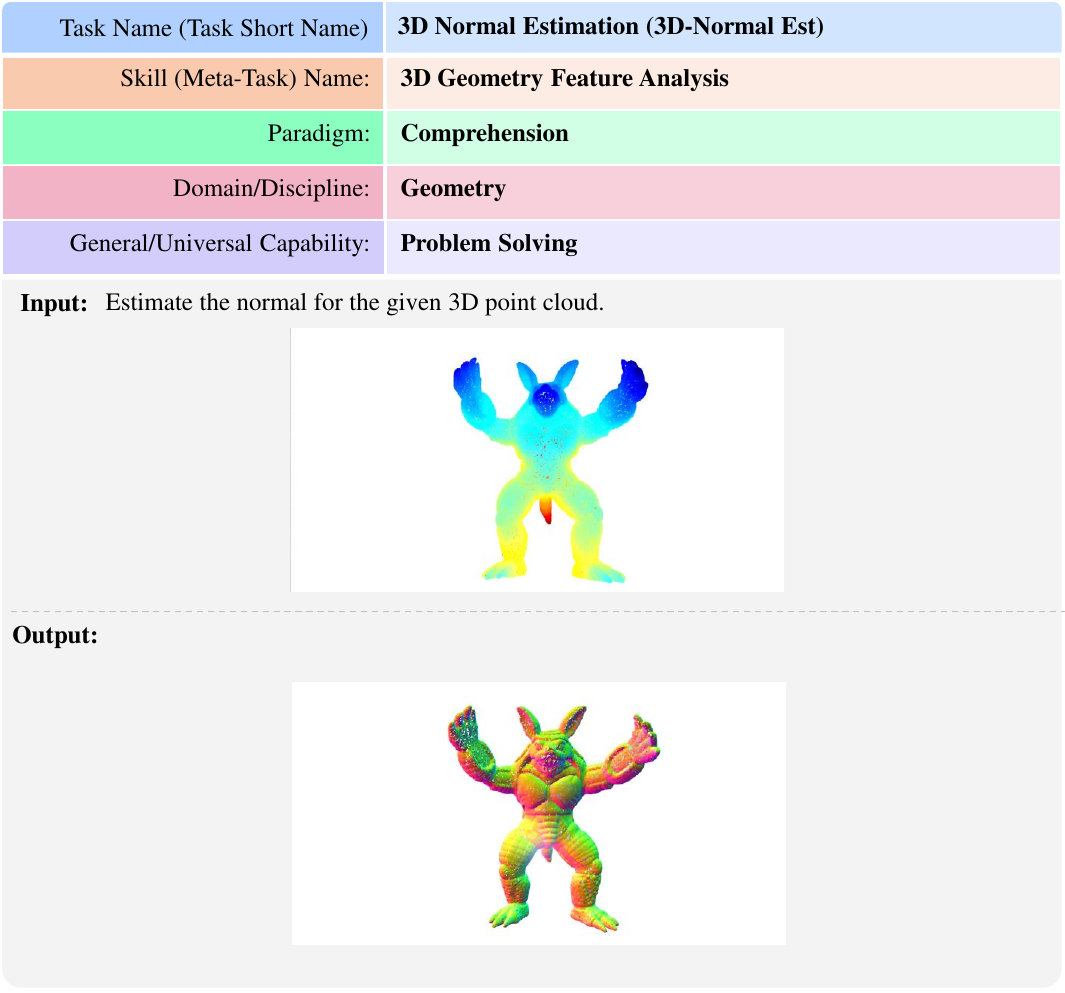}
\caption{3D Normal Estimation.}
\label{fig:D-C-10}
\vspace{-2mm}
\end{figure*}


\begin{figure*}[!h]
\centering
\includegraphics[width=0.9\linewidth]{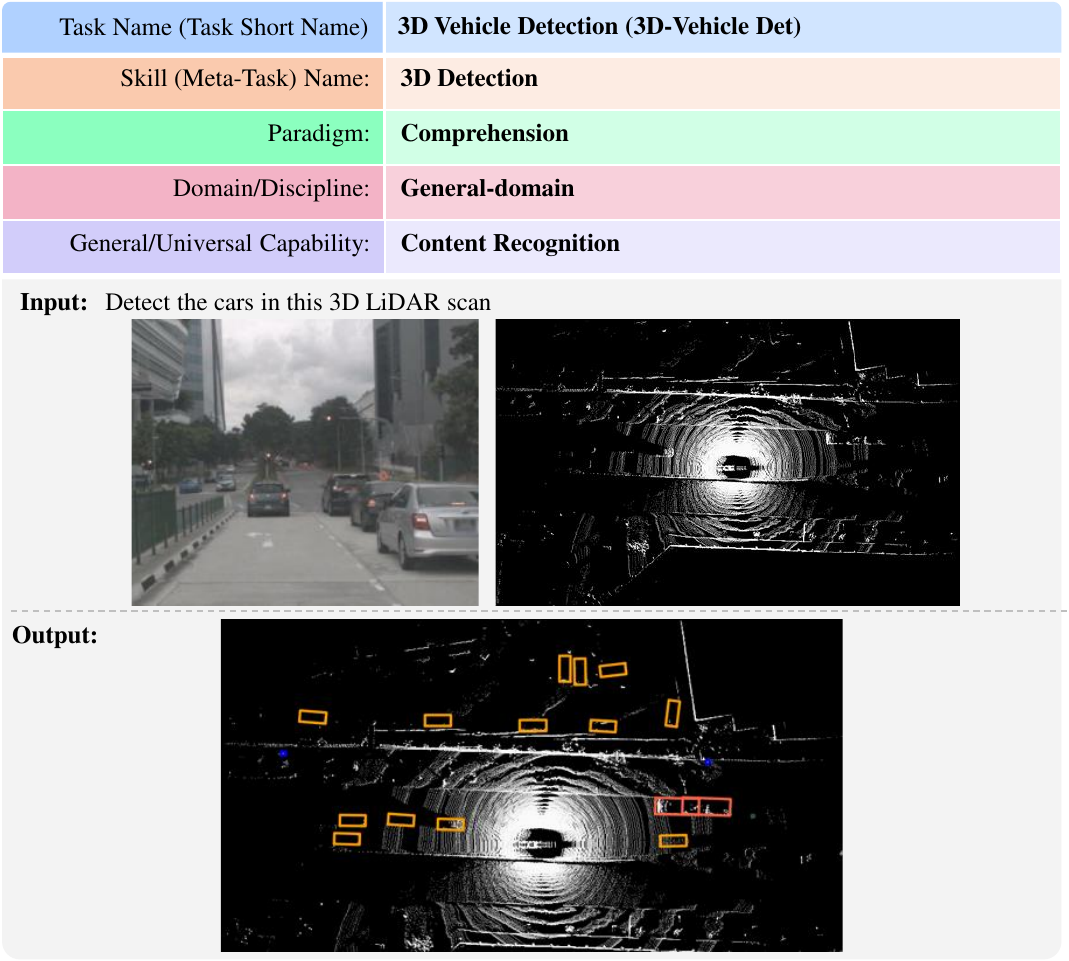}
\caption{3D Vehicle Detection.}
\label{fig:D-C-11}
\vspace{-2mm}
\end{figure*}


\begin{figure*}[!h]
\centering
\includegraphics[width=0.9\linewidth]{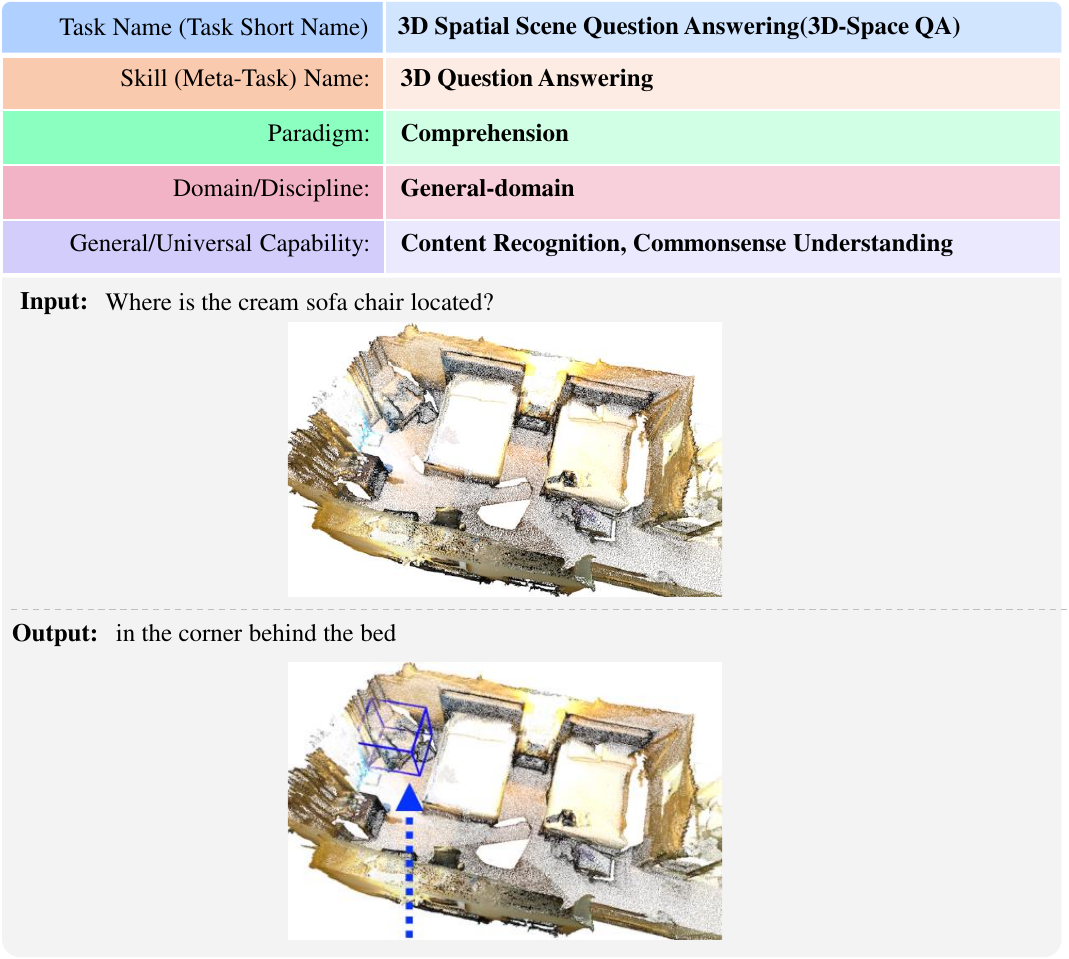}
\caption{3D QA for Spatial Scene Understanding.}
\label{fig:D-C-12}
\vspace{-2mm}
\end{figure*}


\begin{figure*}[!h]
\centering
\includegraphics[width=0.9\linewidth]{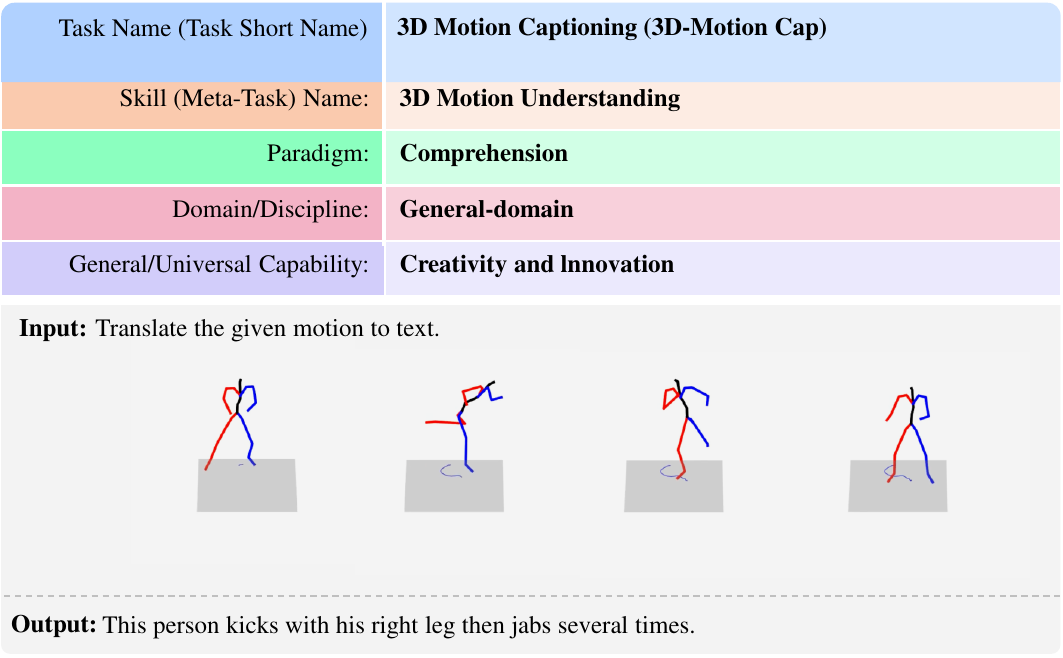}
\caption{3D Motion Captioning.}
\label{fig:D-C-13}
\vspace{-2mm}
\end{figure*}


\clearpage

\paragraph{Generation Tasks.}
Following, we showcase 9 3D-oriented generative tasks, each representing a specific skill.


\begin{figure*}[!h]
\centering
\includegraphics[width=0.9\linewidth]{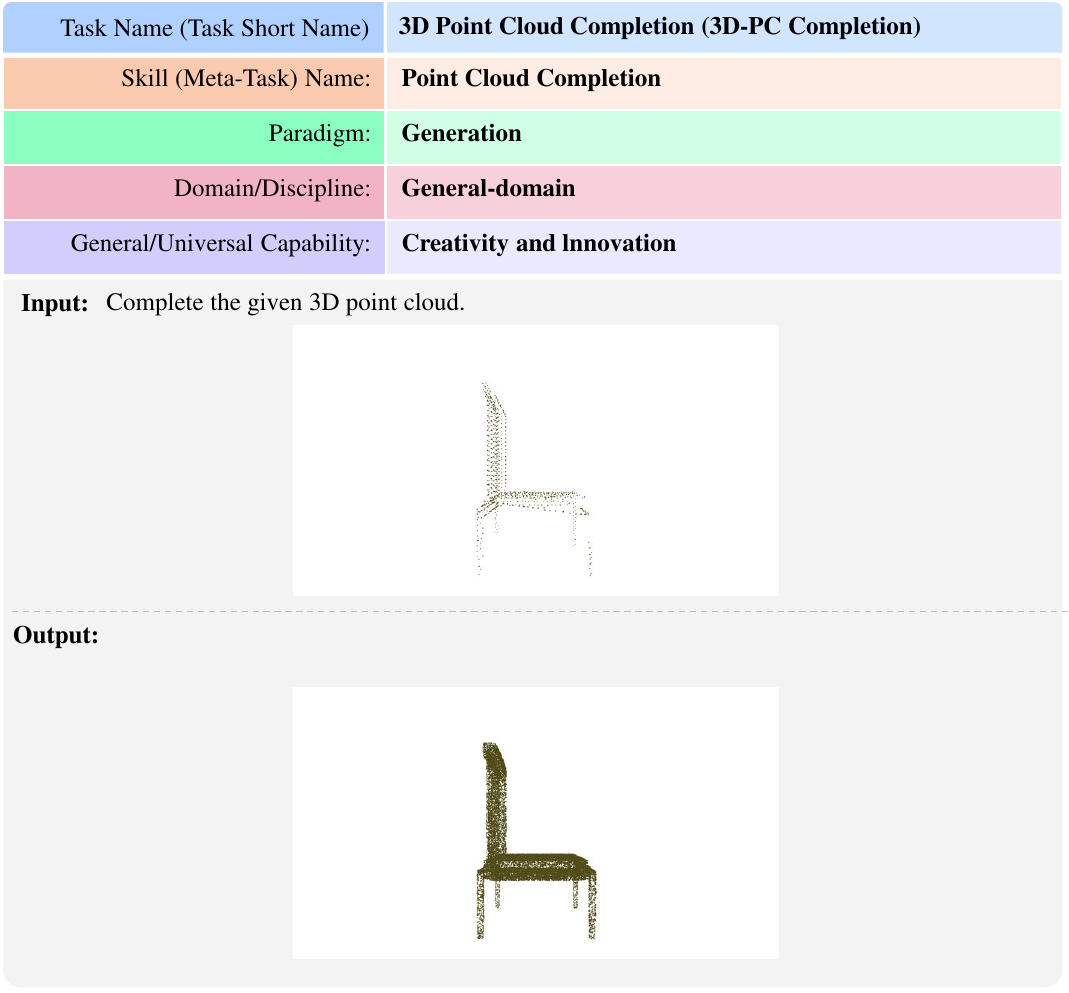}
\caption{3D-Point-Cloud Completion.}
\label{fig:D-G-1}
\vspace{-2mm}
\end{figure*}


\begin{figure*}[!h]
\centering
\includegraphics[width=0.9\linewidth]{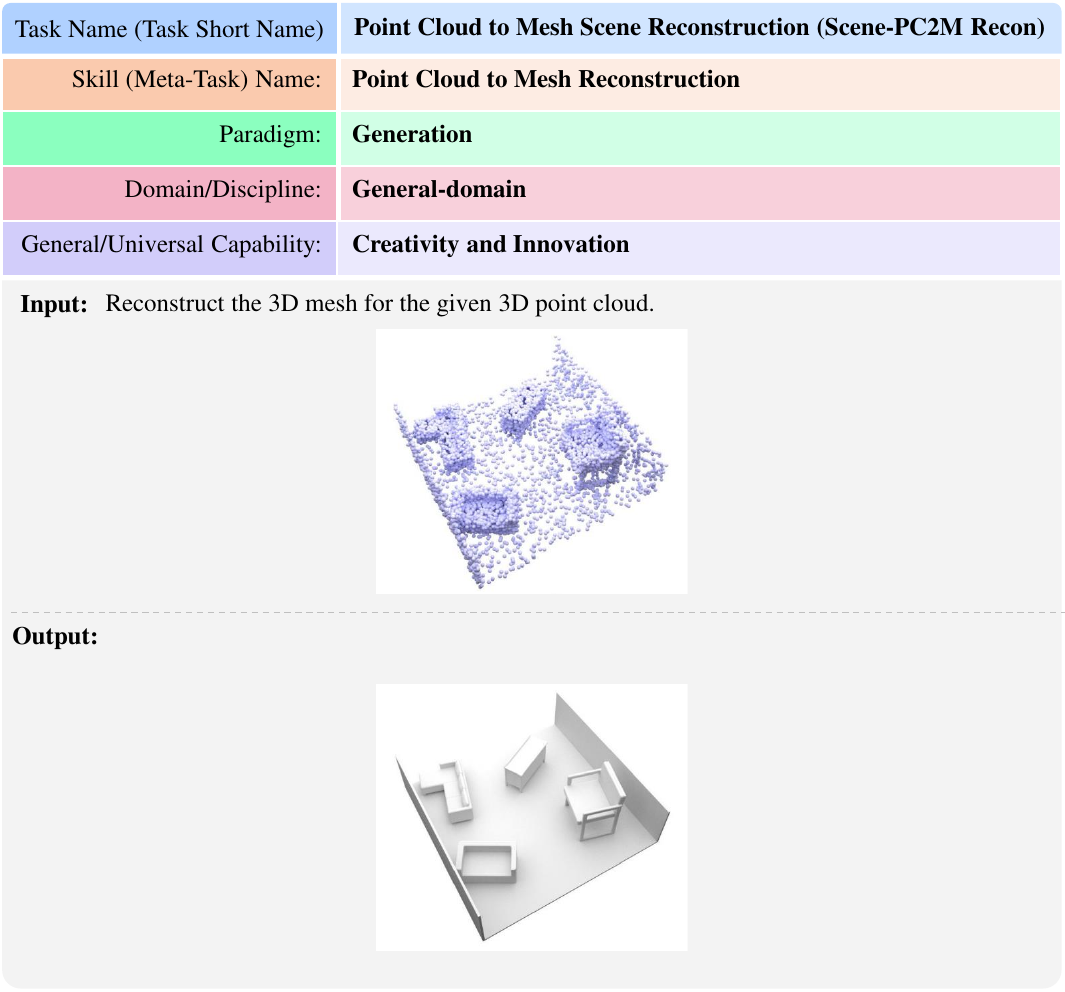}
\caption{Point-Cloud-to-Mesh Scene Reconstruction.}
\label{fig:D-G-2}
\vspace{-2mm}
\end{figure*}


\begin{figure*}[!h]
\centering
\includegraphics[width=0.9\linewidth]{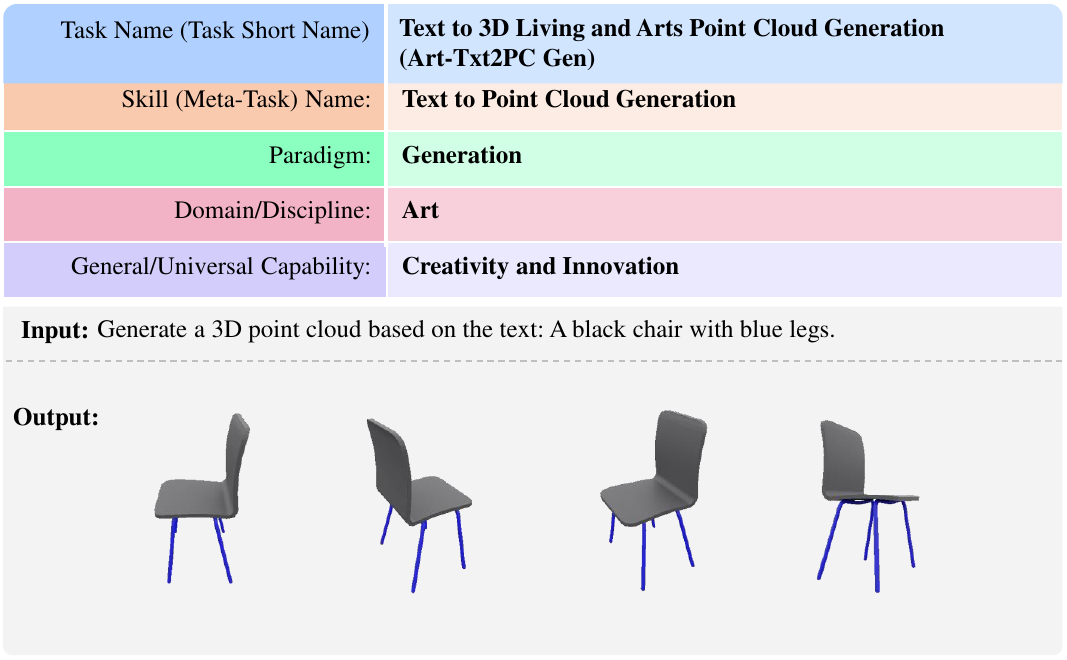}
\caption{Living-and-Arts Image-to-3D-Point-Cloud Generation.}
\label{fig:D-G-3}
\vspace{-2mm}
\end{figure*}


\begin{figure*}[!h]
\centering
\includegraphics[width=0.9\linewidth]{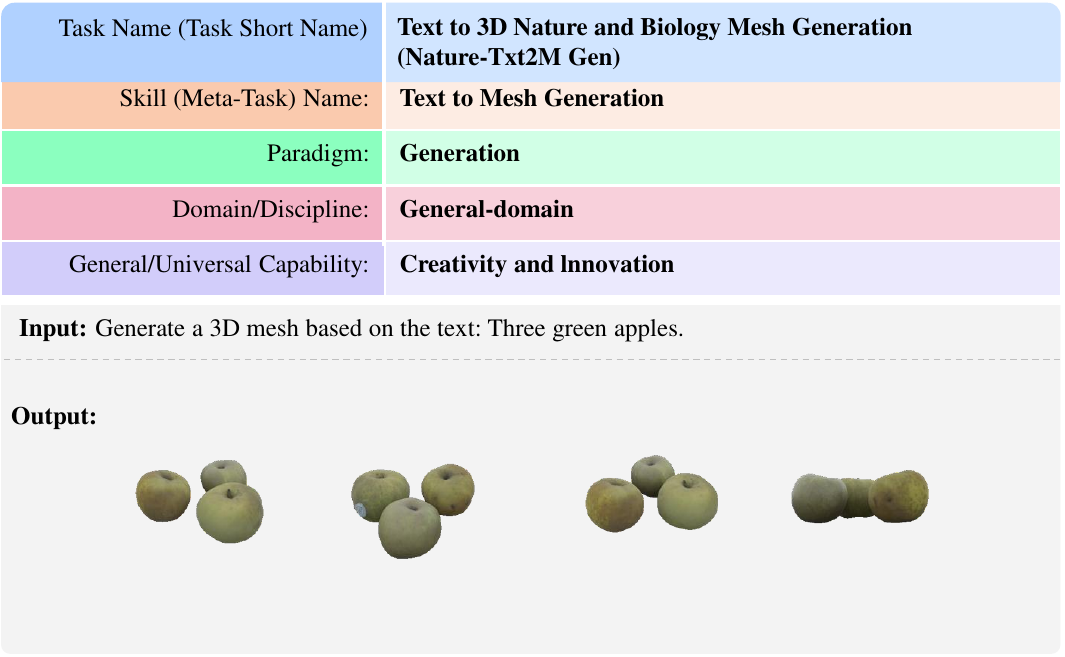}
\caption{Nature-and-Biology Text-to-3D-Mesh Generation.}
\label{fig:D-G-4}
\vspace{-2mm}
\end{figure*}


\begin{figure*}[!h]
\centering
\includegraphics[width=0.9\linewidth]{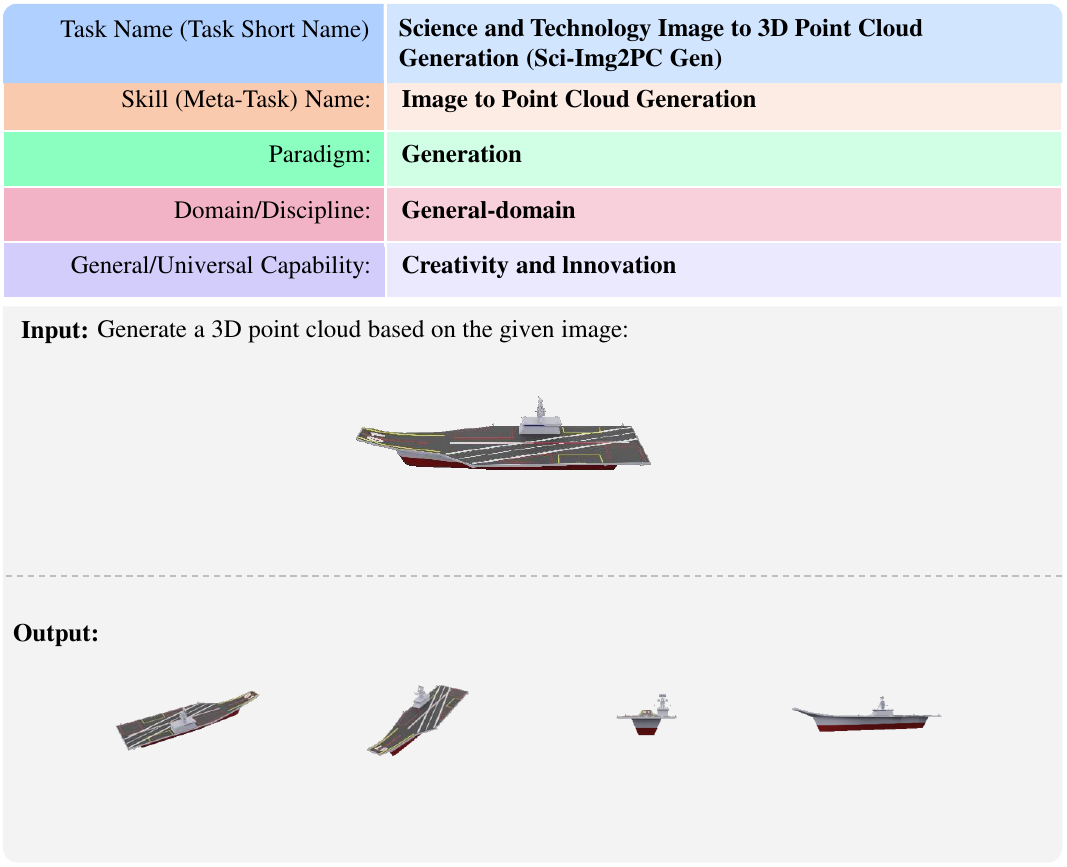}
\caption{Science-and-Technology Image-to-3D-Point-Cloud Generation.}
\label{fig:D-G-5}
\vspace{-2mm}
\end{figure*}


\begin{figure*}[!h]
\centering
\includegraphics[width=0.9\linewidth]{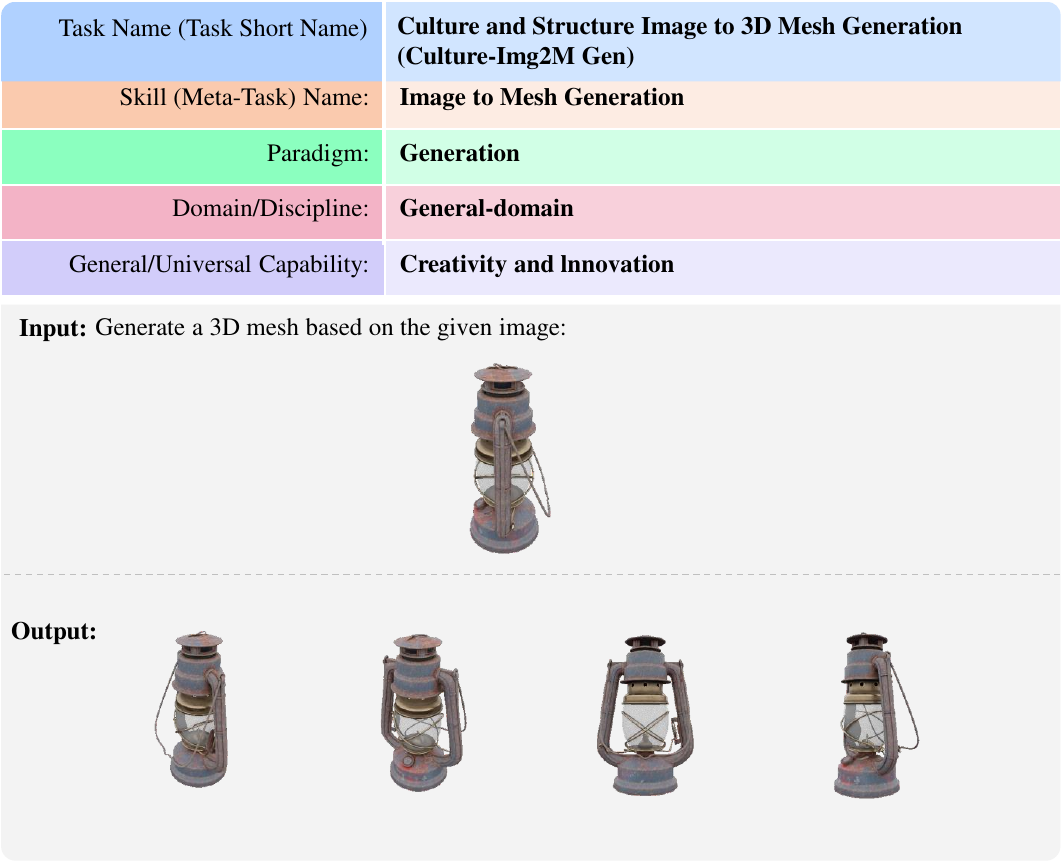}
\caption{Culture-and-Structure Image to 3D Mesh Generation.}
\label{fig:D-G-6}
\vspace{-2mm}
\end{figure*}


\begin{figure*}[!h]
\centering
\includegraphics[width=0.9\linewidth]{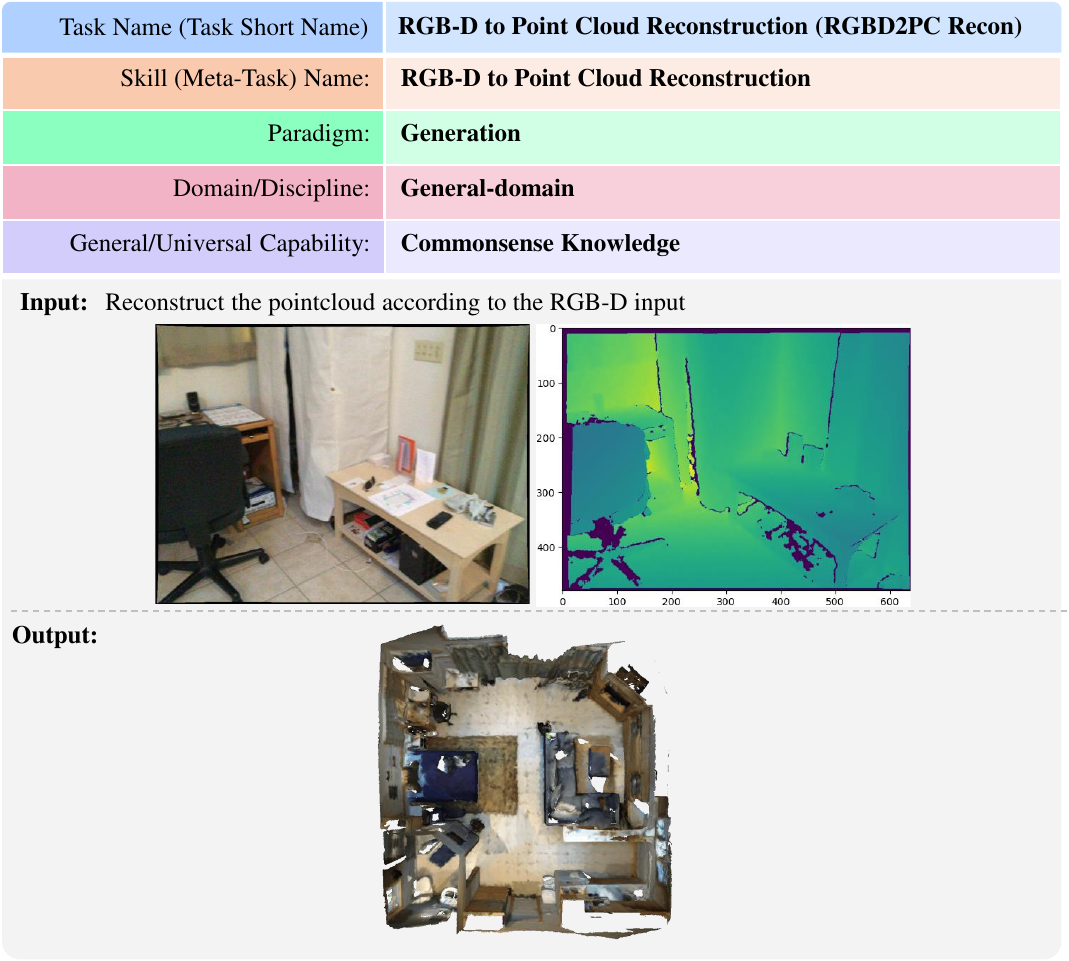}
\caption{RGBD-to-Point-Cloud Reconstruction.}
\label{fig:D-G-7}
\vspace{-2mm}
\end{figure*}


\begin{figure*}[!h]
\centering
\includegraphics[width=0.9\linewidth]{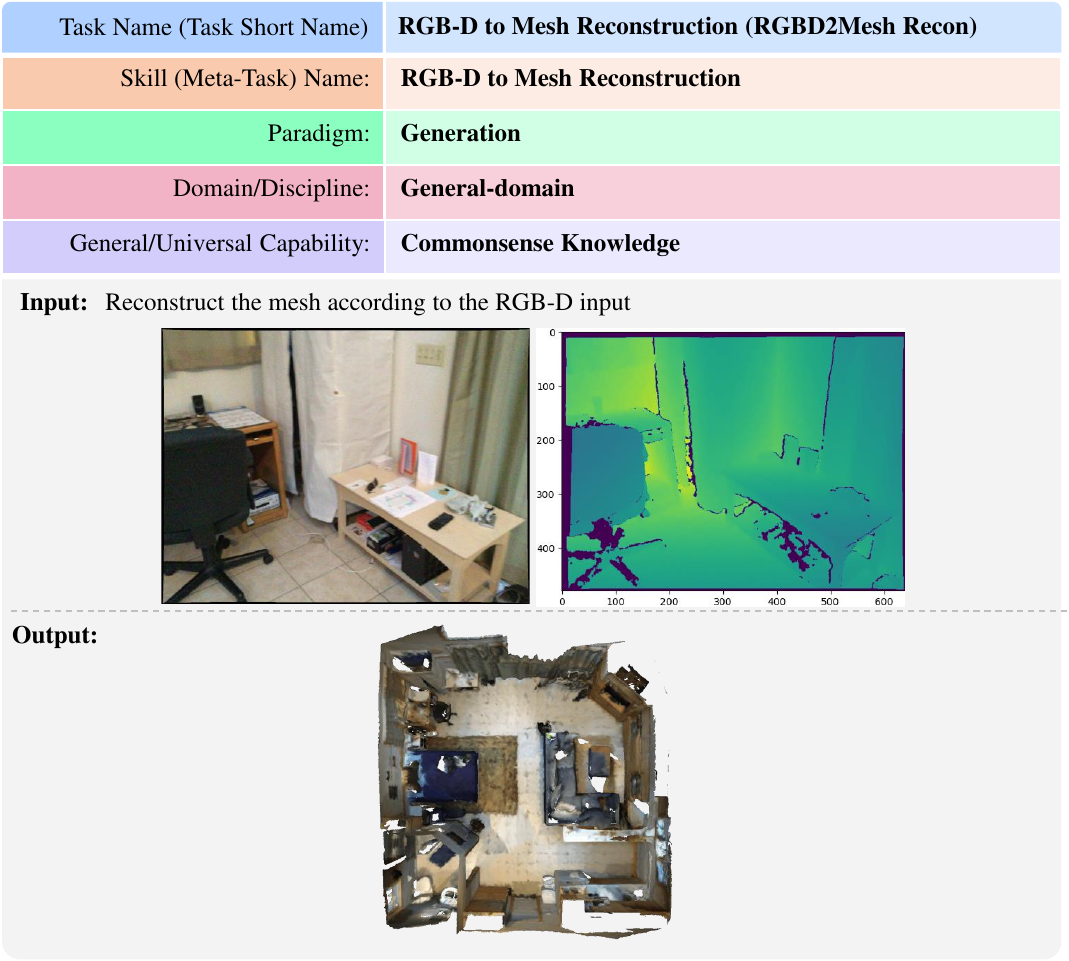}
\caption{RGBD-to-Mesh Reconstruction.}
\label{fig:D-G-8}
\vspace{-2mm}
\end{figure*}


\begin{figure*}[!h]
\centering
\includegraphics[width=0.9\linewidth]{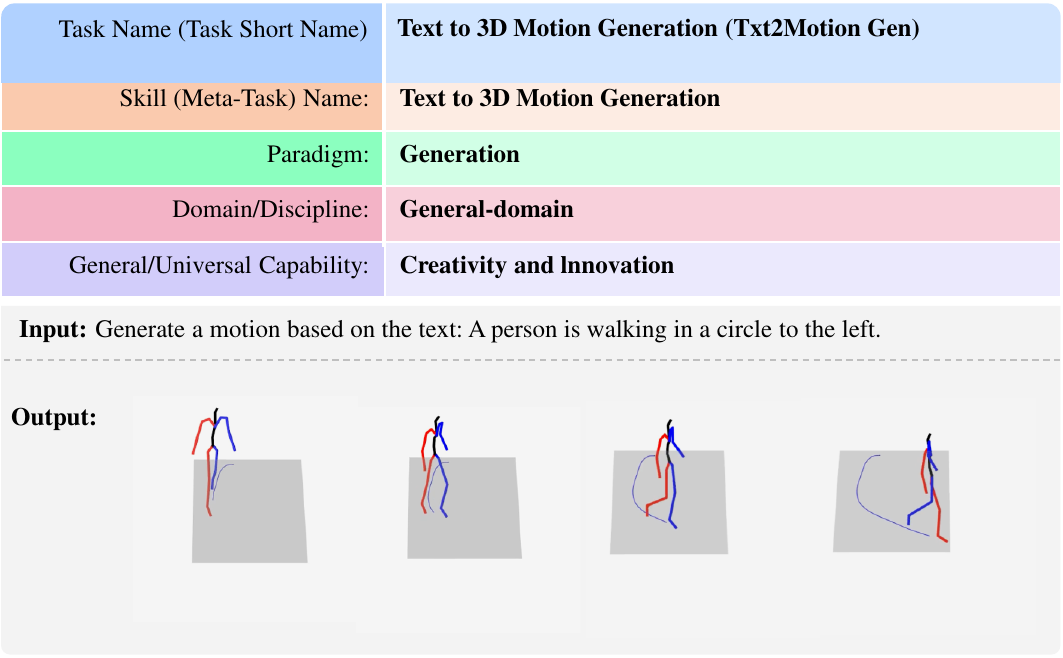}
\caption{Text-to-3D-Motion Generation.}
\label{fig:D-G-9}
\vspace{-2mm}
\end{figure*}


\clearpage

\subsubsection{Language Tasks.}
Following we showcase 22 NLP tasks, each representing a specific skill.


\begin{figure*}[!h]
\centering
\includegraphics[width=0.9\linewidth]{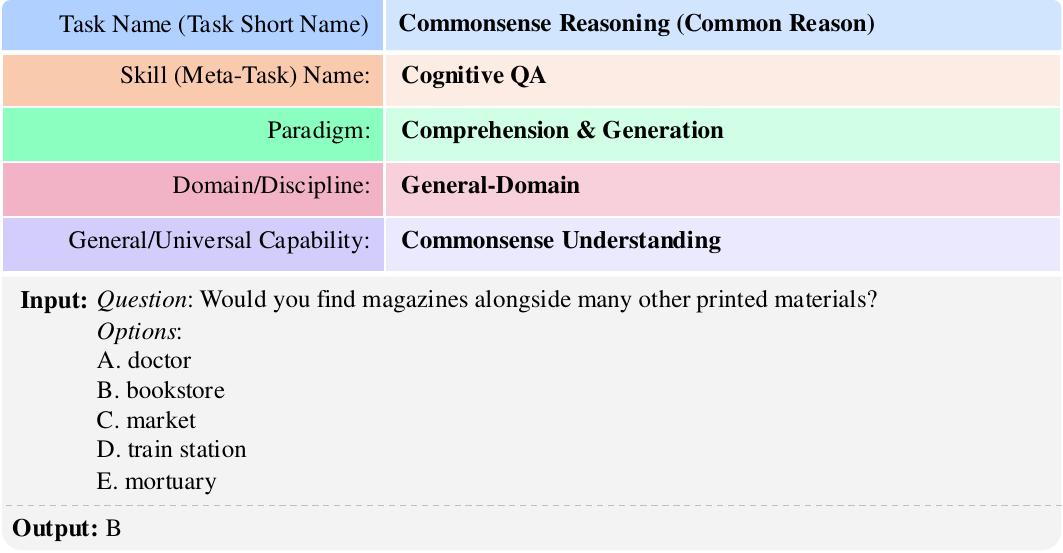}
\caption{Commonsense Reasoning.}
\label{fig:L-1}
\vspace{-2mm}
\end{figure*}


\begin{figure*}[!h]
\centering
\includegraphics[width=0.9\linewidth]{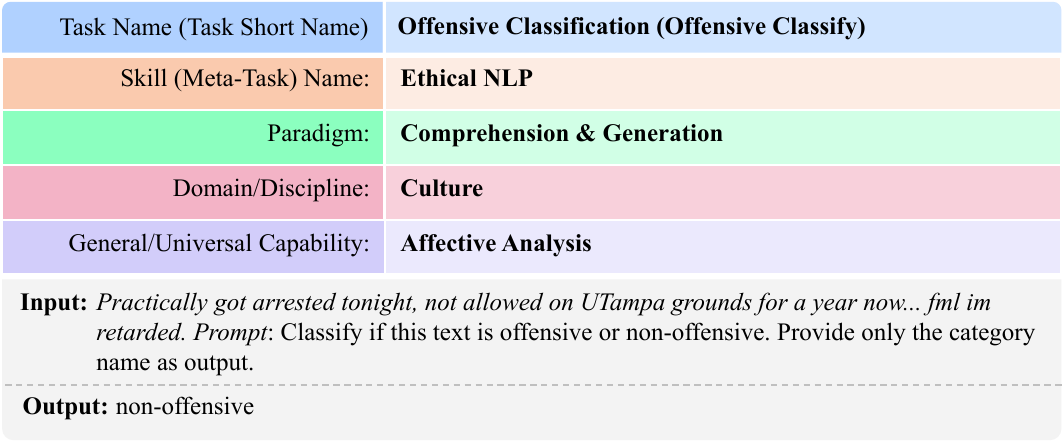}
\caption{Offensive Classification.}
\label{fig:L-2-1}
\vspace{-2mm}
\end{figure*}


\begin{figure*}[!h]
\centering
\includegraphics[width=0.9\linewidth]{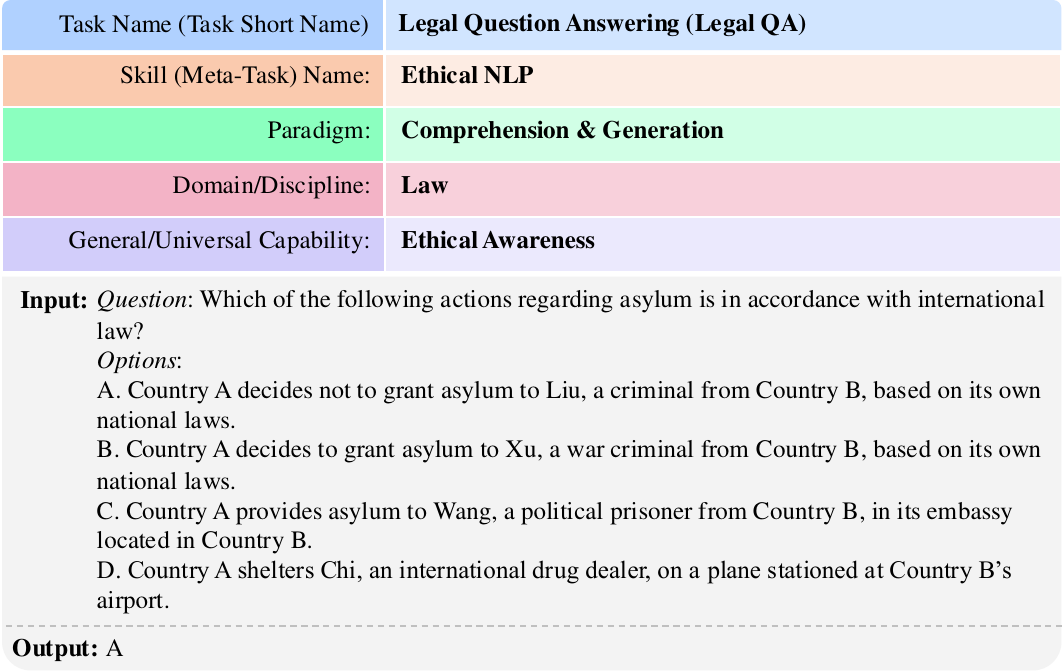}
\caption{Legal Question Answering.}
\label{fig:L-2-2}
\vspace{-2mm}
\end{figure*}


\begin{figure*}[!h]
\centering
\includegraphics[width=0.9\linewidth]{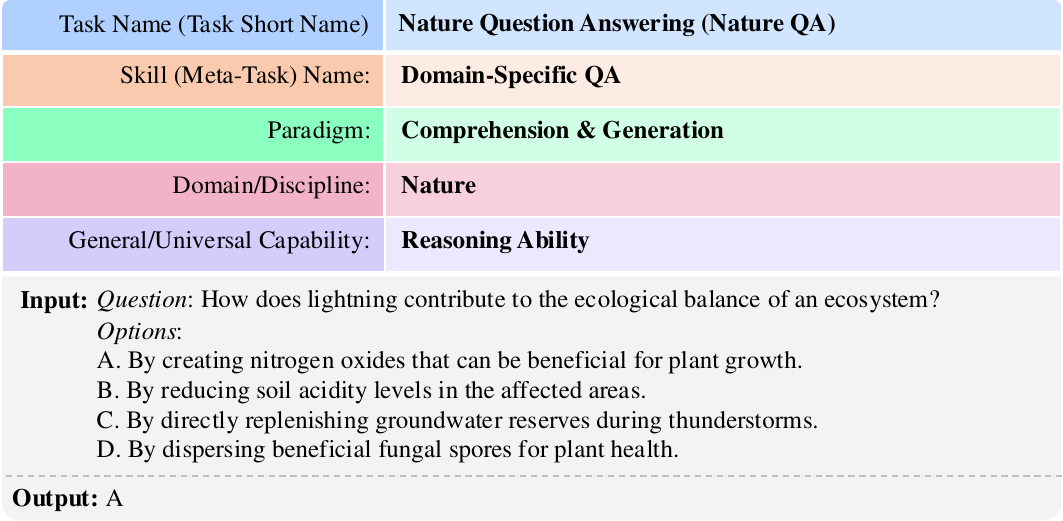}
\caption{Nature Question Answering.}
\label{fig:L-3}
\vspace{-2mm}
\end{figure*}


\begin{figure*}[!h]
\centering
\includegraphics[width=0.9\linewidth]{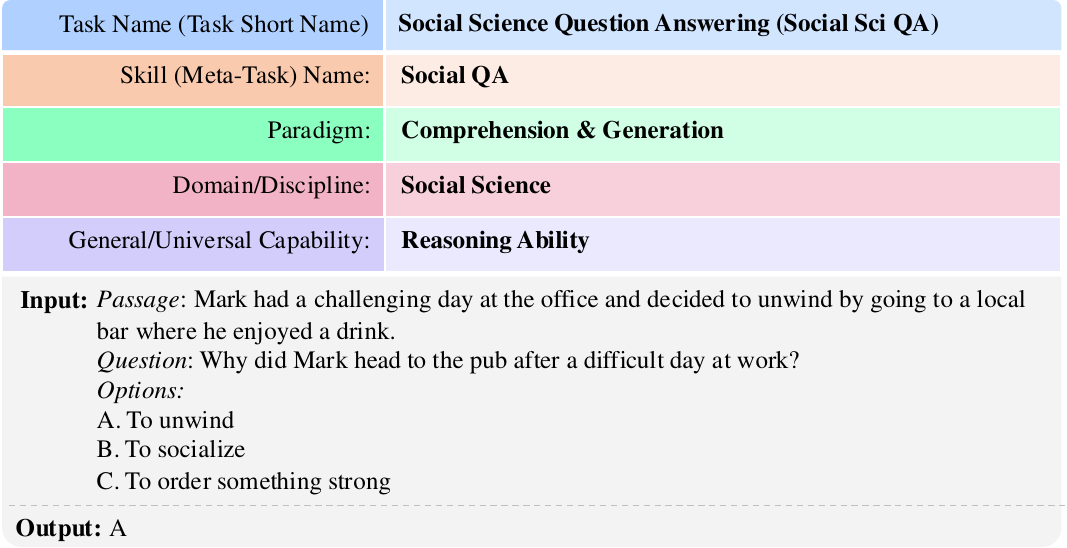}
\caption{Social Science Question Answering.}
\label{fig:L-4}
\vspace{-2mm}
\end{figure*}


\begin{figure*}[!h]
\centering
\includegraphics[width=0.9\linewidth]{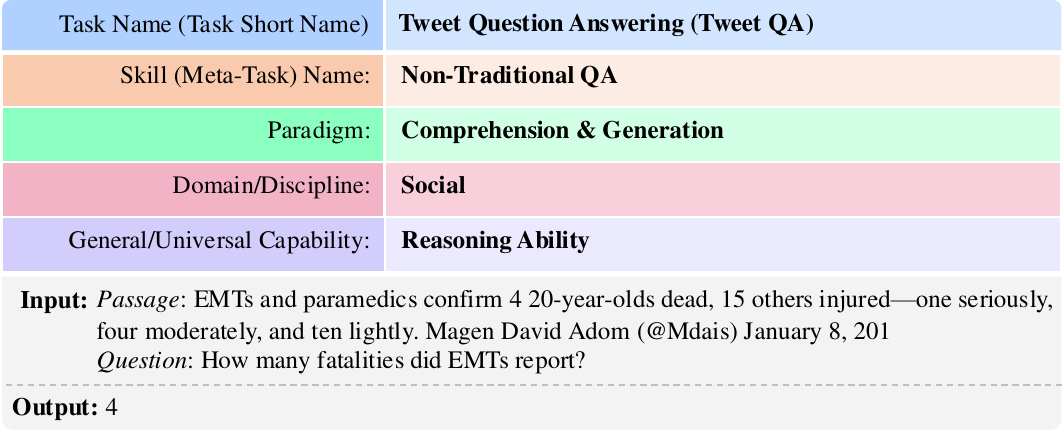}
\caption{Tweet Question Answering.}
\label{fig:L-5-1}
\vspace{-2mm}
\end{figure*}


\begin{figure*}[!h]
\centering
\includegraphics[width=0.9\linewidth]{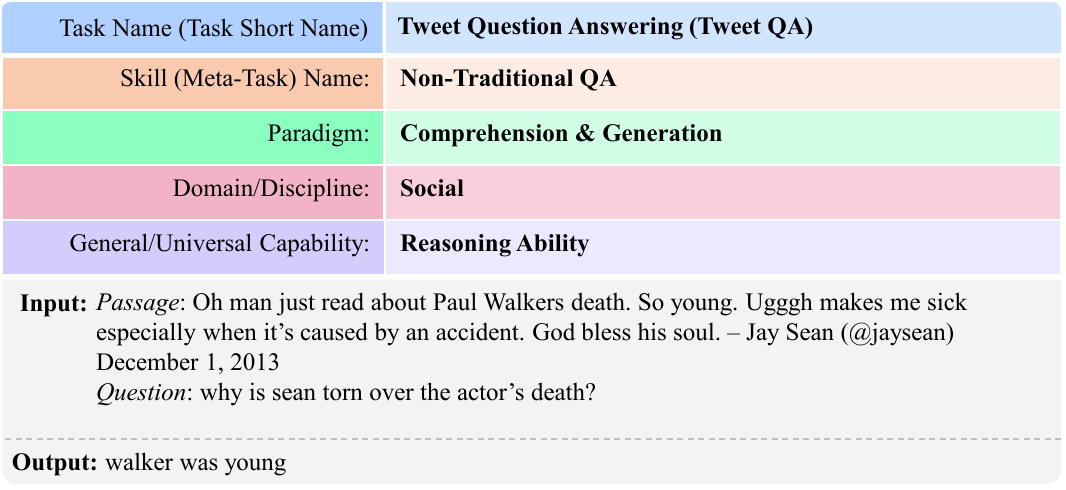}
\caption{Tweet Question Answering.}
\label{fig:L-5-2}
\vspace{-2mm}
\end{figure*}


\begin{figure*}[!h]
\centering
\includegraphics[width=0.9\linewidth]{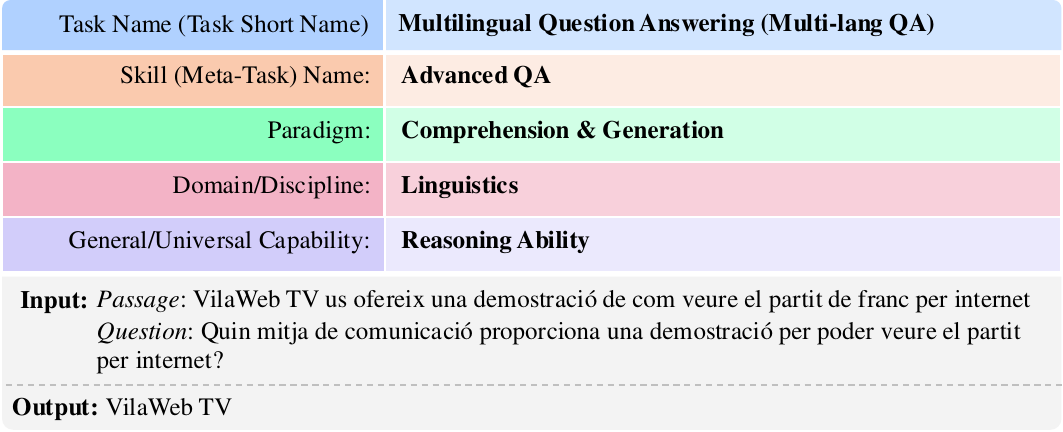}
\caption{Multilingual Question Answering.}
\label{fig:L-6}
\vspace{-2mm}
\end{figure*}


\begin{figure*}[!h]
\centering
\includegraphics[width=0.9\linewidth]{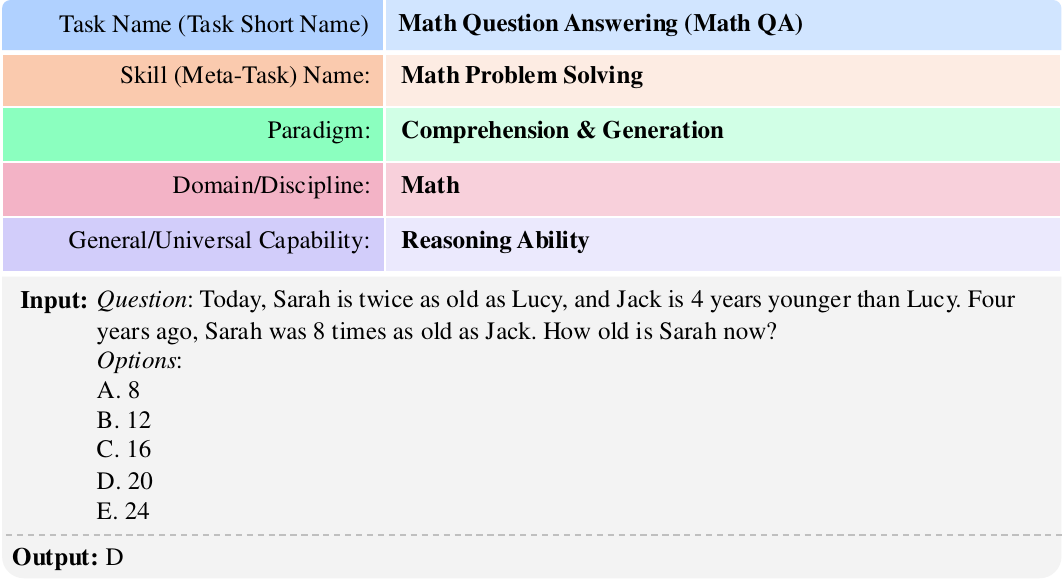}
\caption{Math Question Answering.}
\label{fig:L-7}
\vspace{-2mm}
\end{figure*}


\begin{figure*}[!h]
\centering
\includegraphics[width=0.9\linewidth]{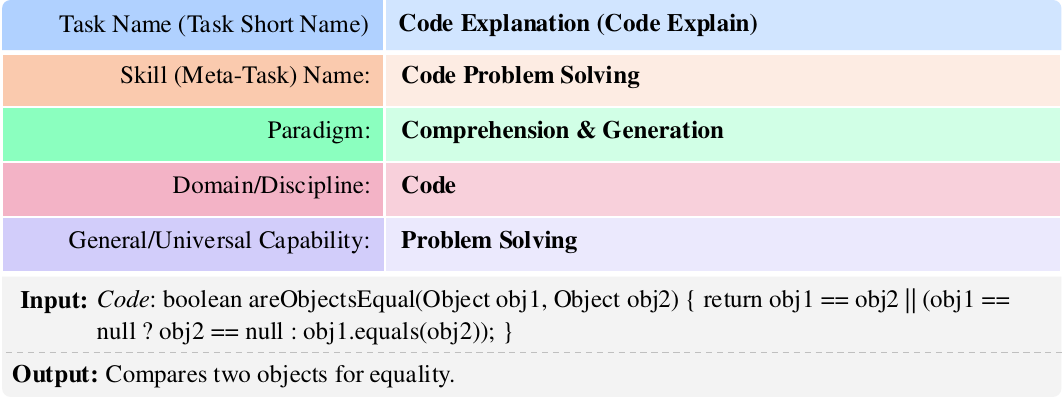}
\caption{Code Explanation.}
\label{fig:L-8}
\vspace{-2mm}
\end{figure*}


\begin{figure*}[!h]
\centering
\includegraphics[width=0.9\linewidth]{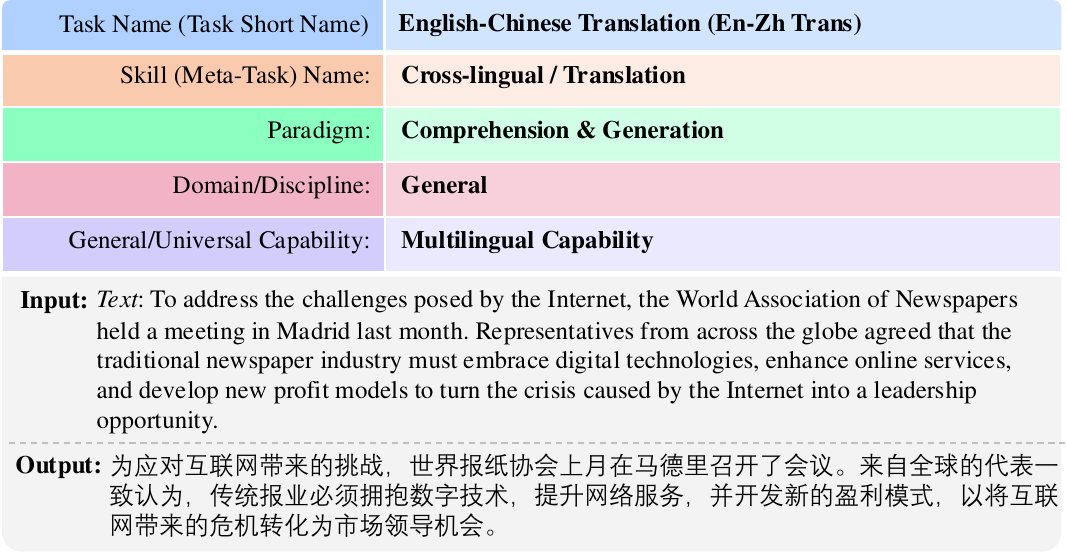}
\caption{English-Chinese Translation.}
\label{fig:L-9}
\vspace{-2mm}
\end{figure*}


\begin{figure*}[!h]
\centering
\includegraphics[width=0.9\linewidth]{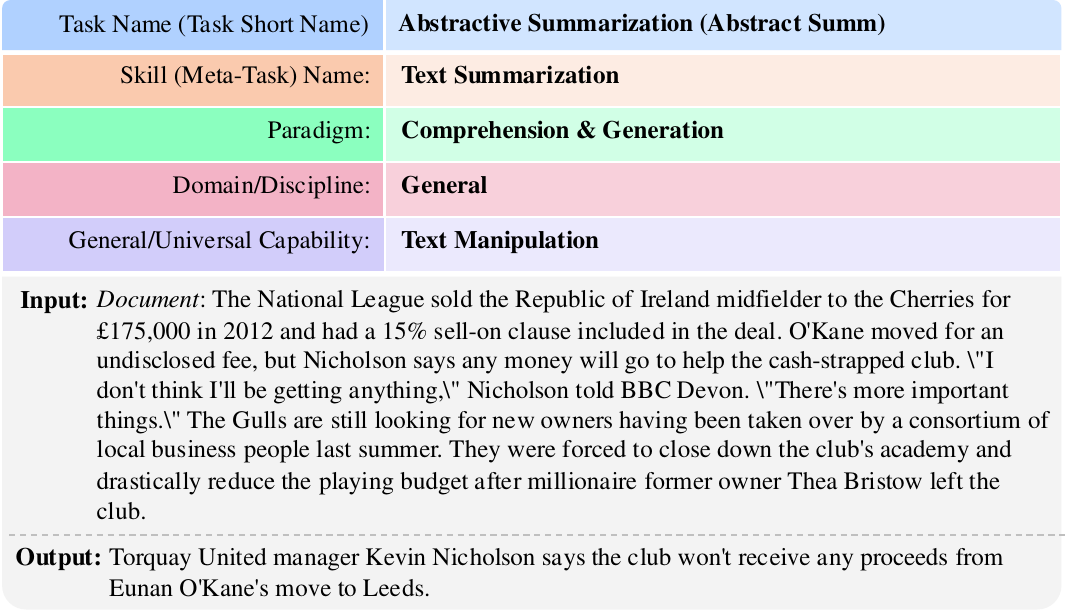}
\caption{Abstractive Summarization.}
\label{fig:L-10}
\vspace{-2mm}
\end{figure*}


\begin{figure*}[!h]
\centering
\includegraphics[width=0.9\linewidth]{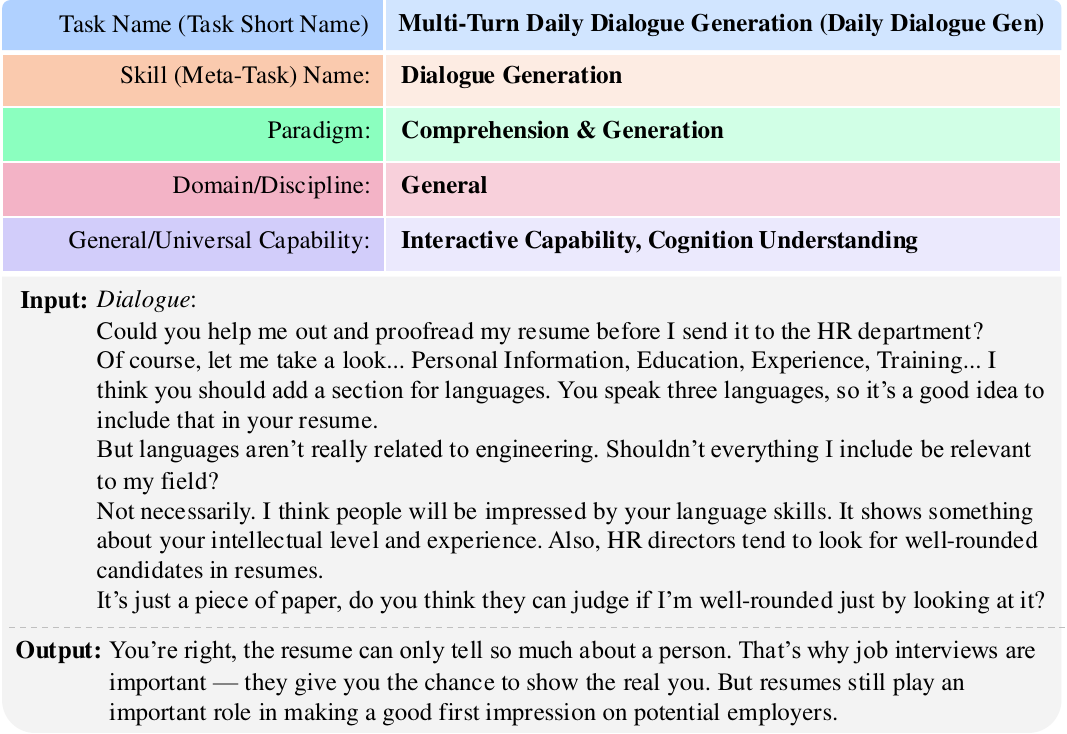}
\caption{Multi-Turn Daily Dialogue Generation.}
\label{fig:L-11}
\vspace{-2mm}
\end{figure*}


\begin{figure*}[!h]
\centering
\includegraphics[width=0.9\linewidth]{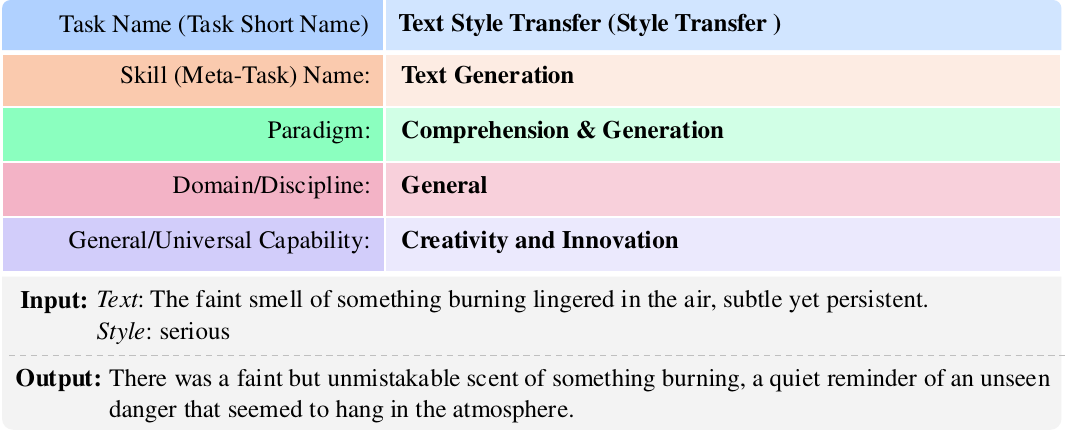}
\caption{Text Style Transfer.}
\label{fig:L-12}
\vspace{-2mm}
\end{figure*}


\begin{figure*}[!h]
\centering
\includegraphics[width=0.9\linewidth]{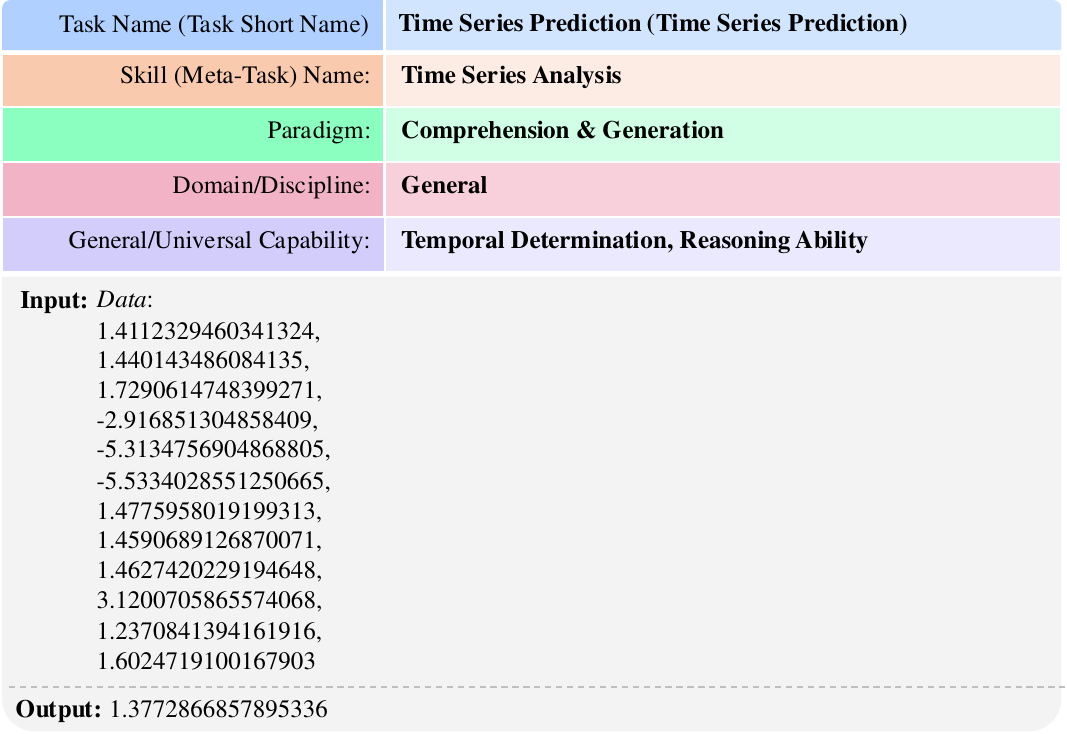}
\caption{Time Series Prediction.}
\label{fig:L-13}
\vspace{-2mm}
\end{figure*}


\begin{figure*}[!h]
\centering
\includegraphics[width=0.9\linewidth]{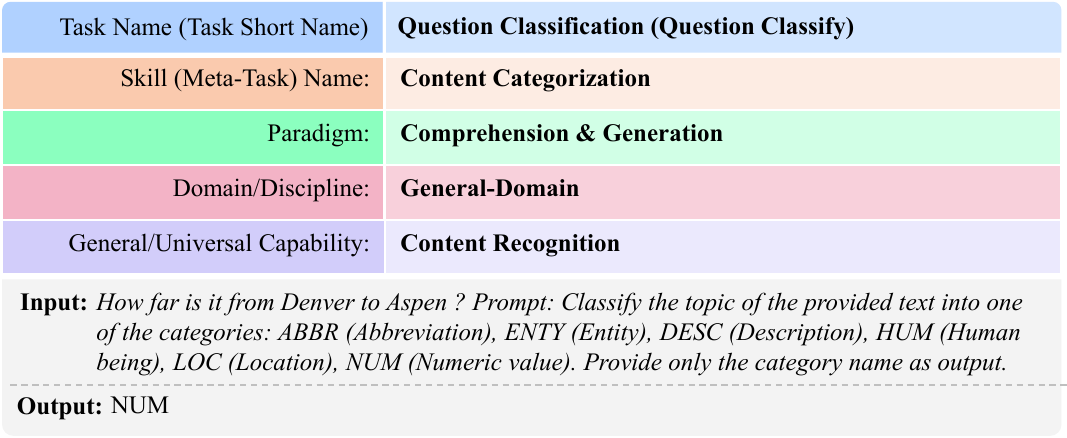}
\caption{Question Classification.}
\label{fig:L-14}
\vspace{-2mm}
\end{figure*}


\begin{figure*}[!h]
\centering
\includegraphics[width=0.9\linewidth]{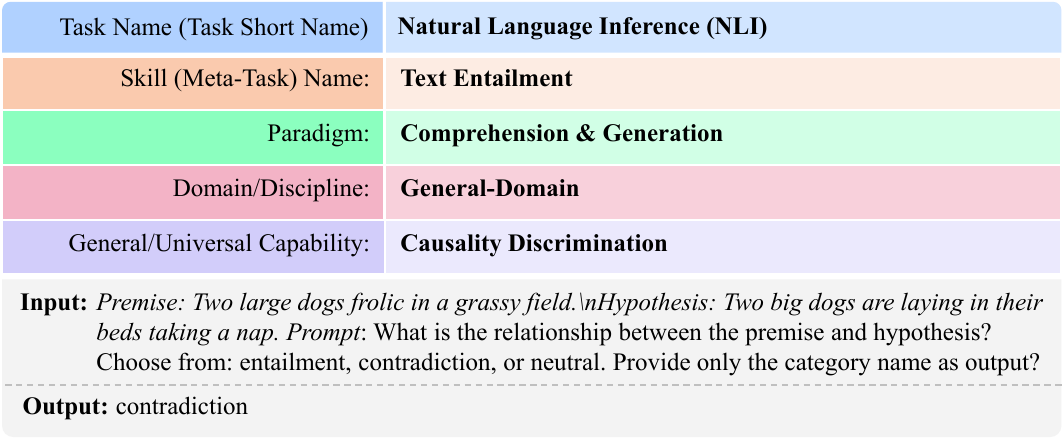}
\caption{Natural Language Inference.}
\label{fig:L-15}
\vspace{-2mm}
\end{figure*}


\begin{figure*}[!h]
\centering
\includegraphics[width=0.9\linewidth]{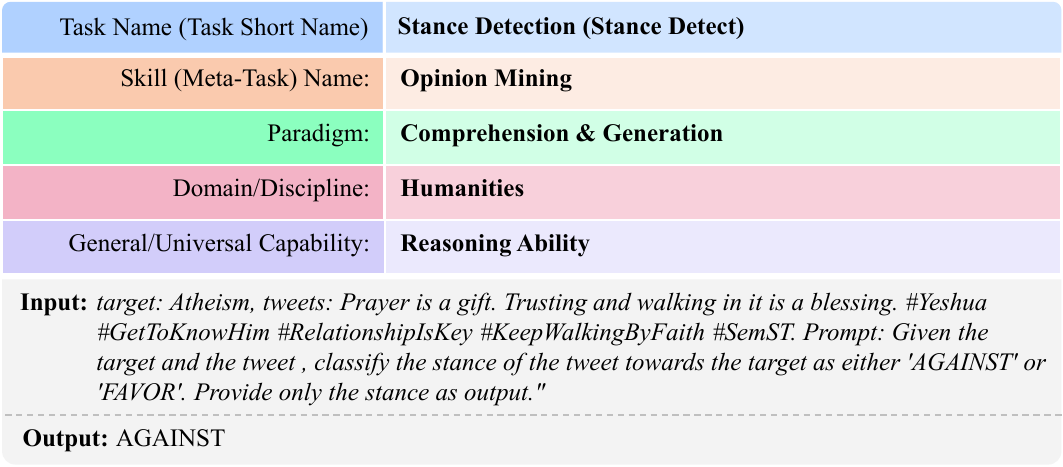}
\caption{Stance Detection.}
\label{fig:L-16-1}
\vspace{-2mm}
\end{figure*}


\begin{figure*}[!h]
\centering
\includegraphics[width=0.9\linewidth]{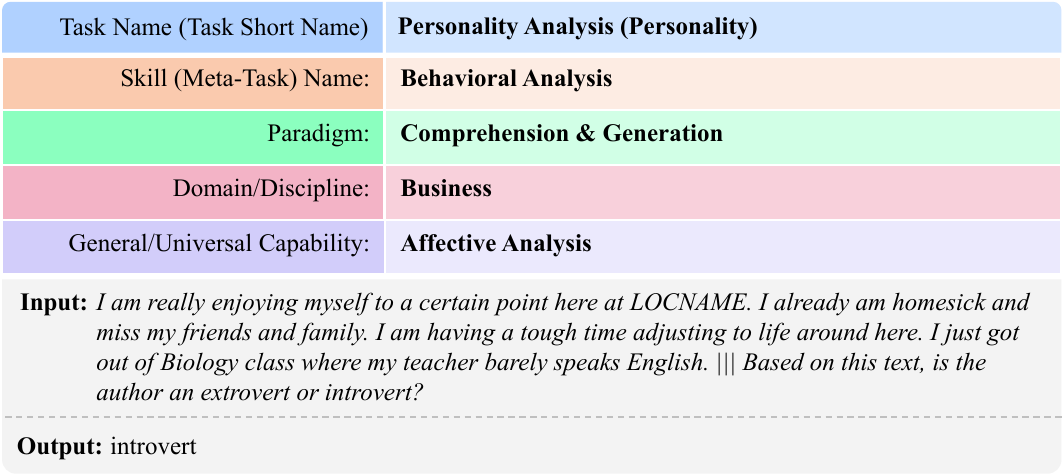}
\caption{Personality Analysis.}
\label{fig:L-16-2}
\vspace{-2mm}
\end{figure*}


\begin{figure*}[!h]
\centering
\includegraphics[width=0.9\linewidth]{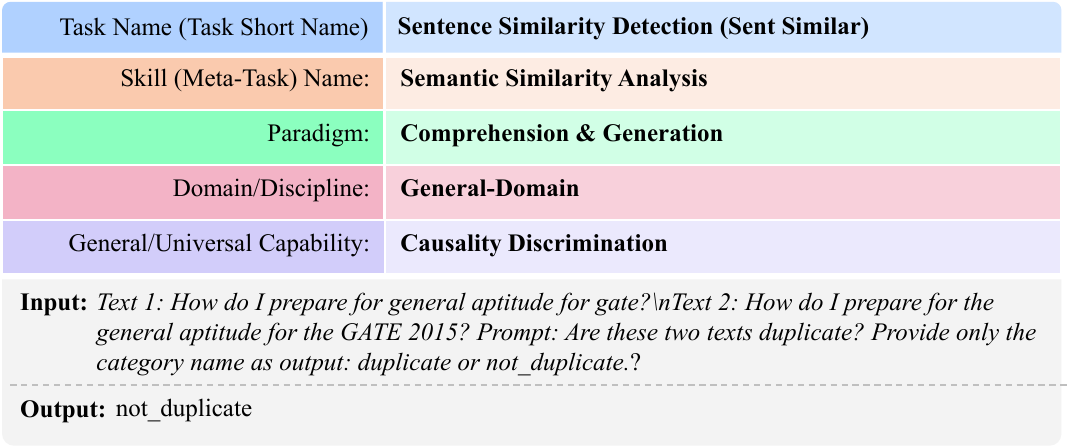}
\caption{Sentence Similarity Detection.}
\label{fig:L-16-3}
\vspace{-2mm}
\end{figure*}


\begin{figure*}[!h]
\centering
\includegraphics[width=0.9\linewidth]{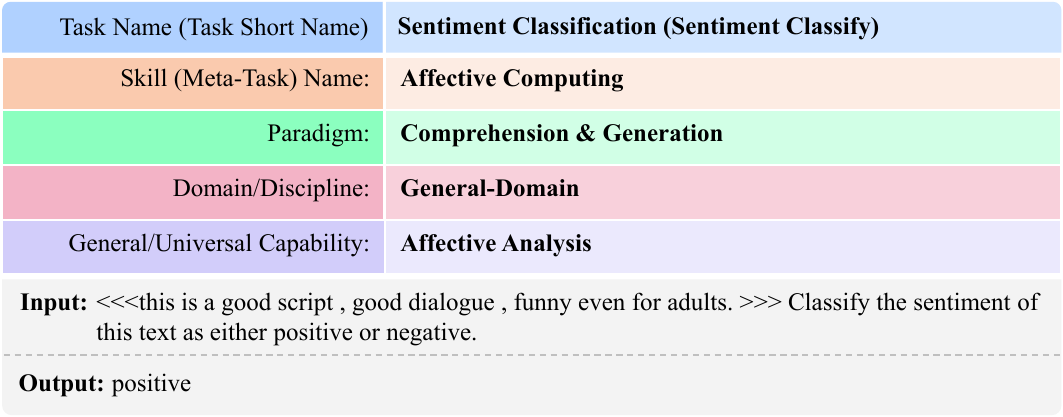}
\caption{Sentiment Classification.}
\label{fig:L-17}
\vspace{-2mm}
\end{figure*}


\begin{figure*}[!h]
\centering
\includegraphics[width=0.9\linewidth]{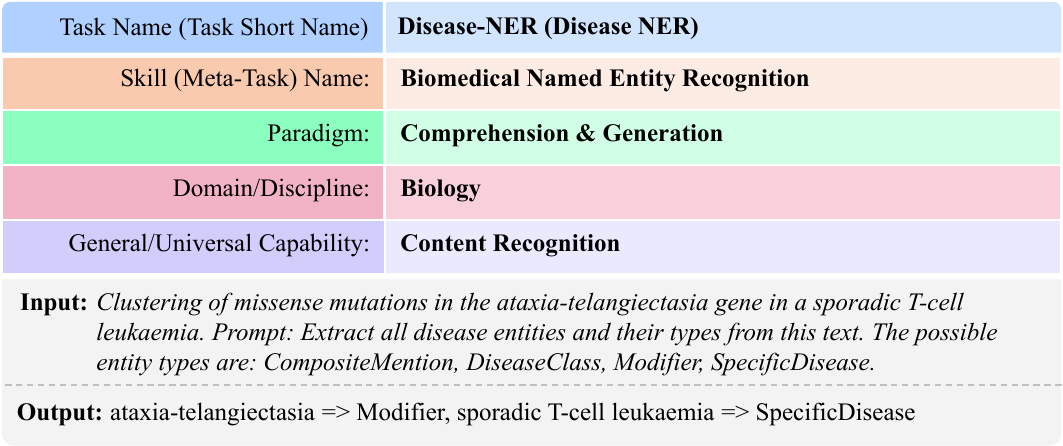}
\caption{Disease-NER.}
\label{fig:L-18-1}
\vspace{-2mm}
\end{figure*}


\begin{figure*}[!h]
\centering
\includegraphics[width=0.9\linewidth]{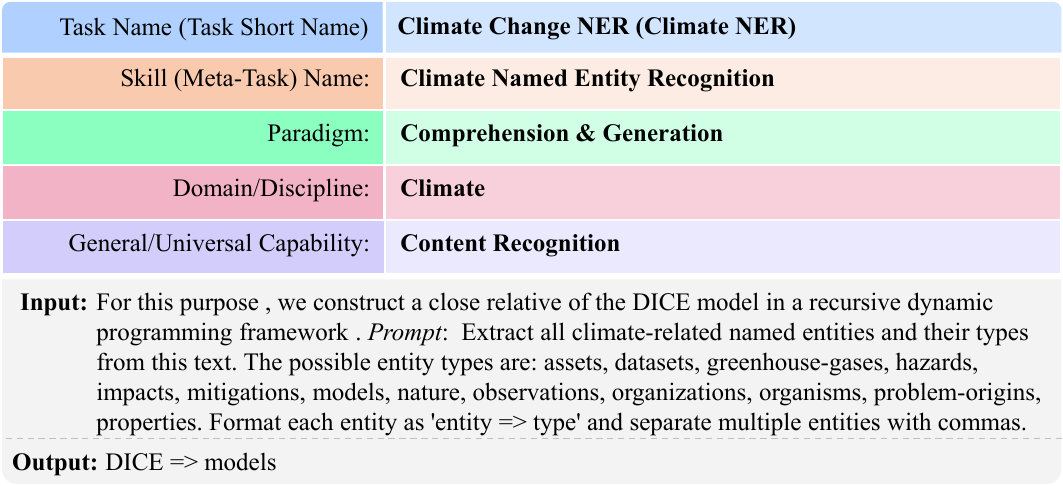}
\caption{Climate Change NER.}
\label{fig:L-18-2}
\vspace{-2mm}
\end{figure*}


\begin{figure*}[!h]
\centering
\includegraphics[width=0.9\linewidth]{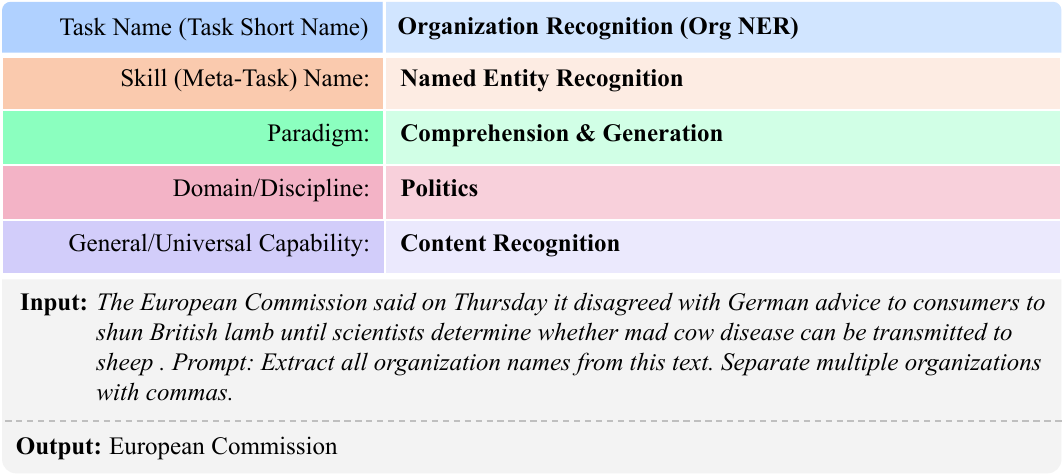}
\caption{Organization Recognition.}
\label{fig:L-18-3}
\vspace{-2mm}
\end{figure*}


\begin{figure*}[!h]
\centering
\includegraphics[width=0.9\linewidth]{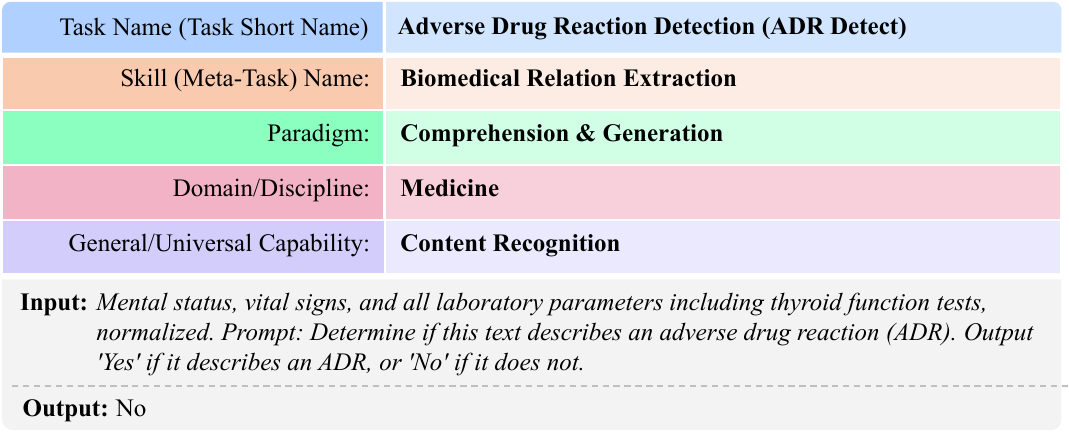}
\caption{Adverse Drug Reaction Detection.}
\label{fig:L-19-1}
\vspace{-2mm}
\end{figure*}


\begin{figure*}[!h]
\centering
\includegraphics[width=0.9\linewidth]{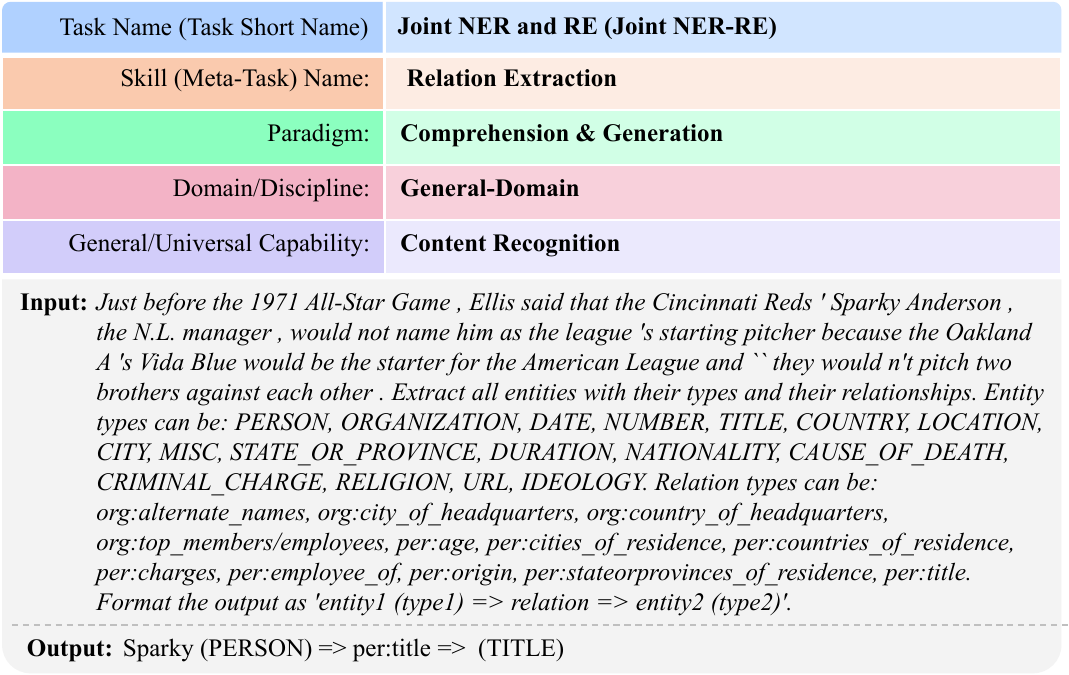}
\caption{Joint NER and RE.}
\label{fig:L-19-2}
\vspace{-2mm}
\end{figure*}


\begin{figure*}[!h]
\centering
\includegraphics[width=0.9\linewidth]{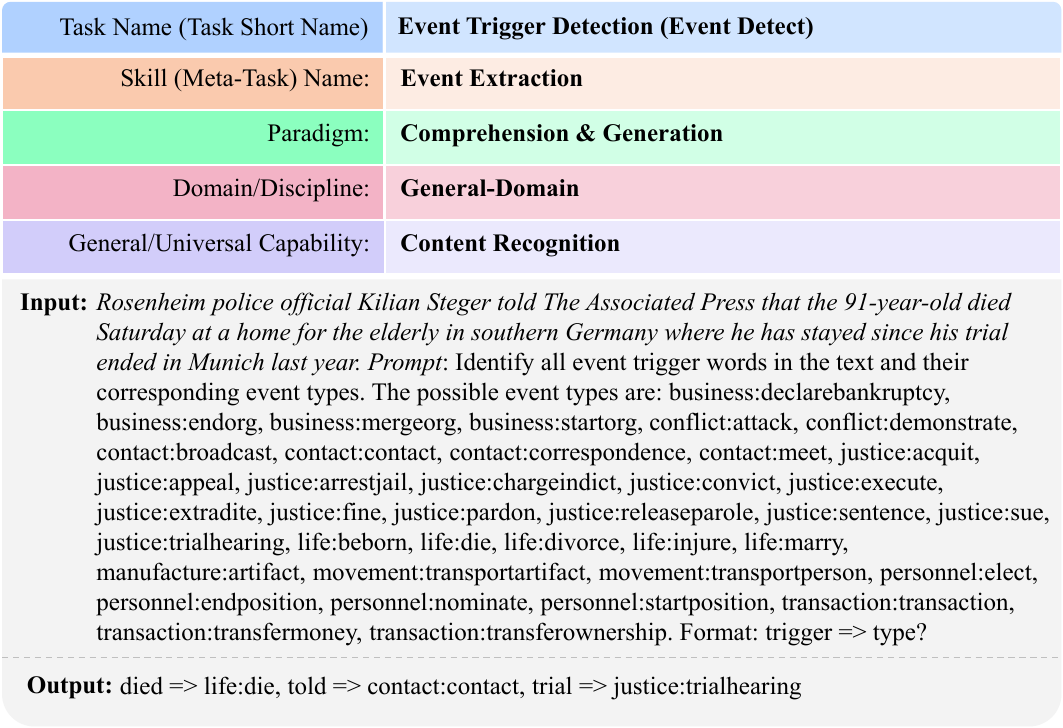}
\caption{Event Trigger Detection.}
\label{fig:L-20}
\vspace{-2mm}
\end{figure*}


\begin{figure*}[!h]
\centering
\includegraphics[width=0.9\linewidth]{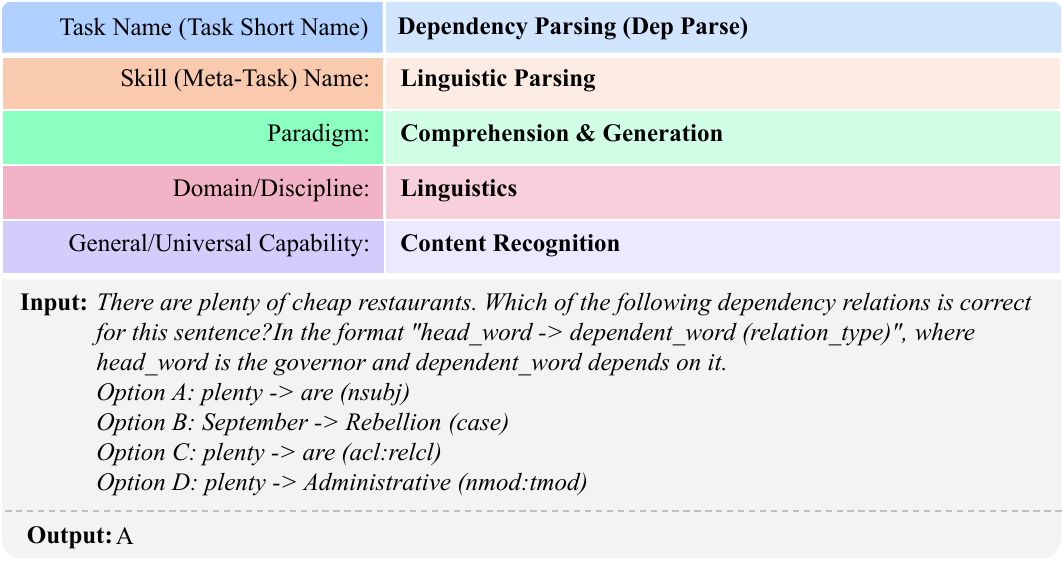}
\caption{Dependency Parsing.}
\label{fig:L-21}
\vspace{-2mm}
\end{figure*}


\begin{figure*}[!h]
\centering
\includegraphics[width=0.9\linewidth]{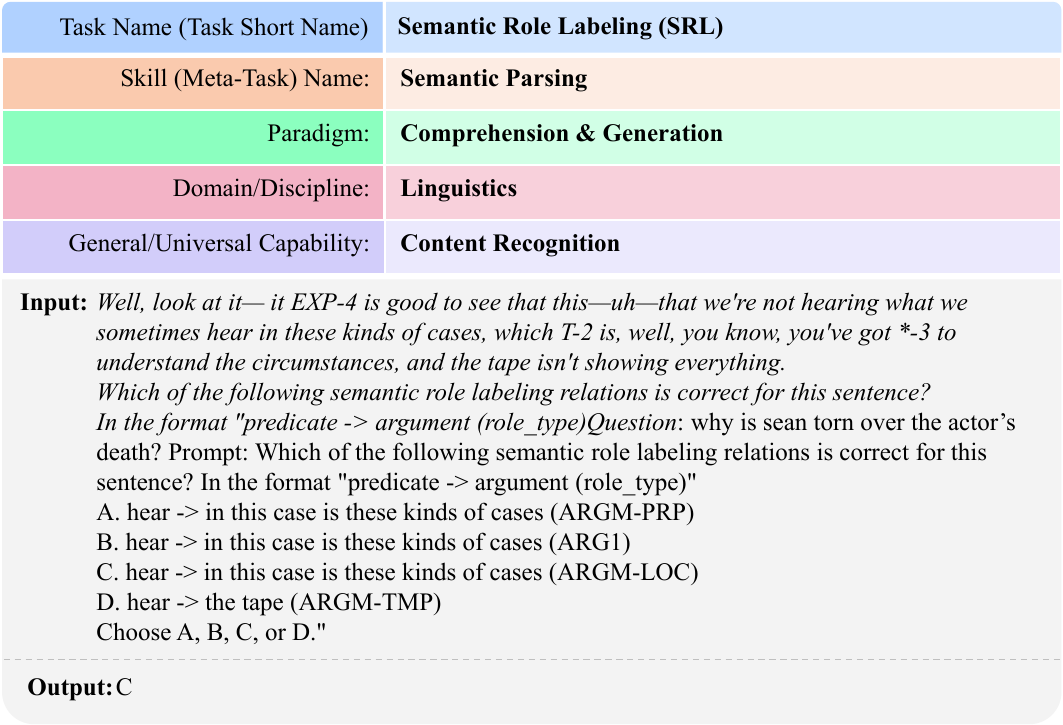}
\caption{Semantic Role Labeling.}
\label{fig:L-22}
\vspace{-2mm}
\end{figure*}


\clearpage


\section{Extension on Experimental Results}
\label{Extension on Experimental Results}

In the main text, we presented a subset of the evaluation performance of MLLMs at the skill level due to space constraints. Here, we provide the complete results on all specific tasks. The results are organized based on modality and task paradigm distinctions.

It is important to note that we conducted inference using different MLLMs' open-source codebases or APIs. However, the choice of prompts for different tasks can significantly impact the model's performance on a given task. For instance, some models require highly detailed and specific in-context instructions in the input prompt to achieve their best performance.
To ensure fairness, our team applied a uniform prompt across all MLLMs for a given task, without any model-specific prompt tuning. If the model developers are unsatisfied with the presented results, we welcome them to adjust the prompts to obtain more representative and improved scores.

\subsection{Results of Image-related Tasks}

\paragraph{Image Comprehension Results.}
The complete results of all models on image comprehension are presented in Table~\ref{tab:overall-results-image-comp-I-C-1-2-3-4-5} to Table~\ref{tab:overall-results-image-comp-I-C-38-39-40}.

\begin{table*}[t!]
\centering
  \fontsize{8.5}{10.5}\selectfont 
  \setlength{\tabcolsep}{1.mm}
\caption{
Results on \textcolor{greenCom}{Image Comprehension Group}, from \#I-C-1 to \#I-C-5.
Scores over SoTA specialists are bolded. 
}
\vspace{-2mm}
\label{tab:overall-results-image-comp-I-C-1-2-3-4-5}
\begin{tabular}{p{4.2cm} ccc c cc c cccccc}
\toprule
\rowcolor{bg-tb-heavey-vision} & \multicolumn{3}{c}{\bf \#I-C-1} & \multicolumn{1}{c}{\bf \#I-C-2} & \multicolumn{2}{c}{\bf \#I-C-3} & \multicolumn{1}{c}{\bf \#I-C-4} & \multicolumn{6}{c}{\bf \#I-C-5} \\ 

\rowcolor{bg-tb-heavey-vision}  & \multicolumn{3}{c}{\tiny (Behavior Recog)} &  \multicolumn{1}{c}{\tiny(Code Gen)} & \multicolumn{2}{c}{\tiny(Crack Det)}  &  \multicolumn{1}{c}{\tiny(Disease Det)} &   \multicolumn{6}{c}{\tiny(Disease Recog)}\\

\cmidrule(r){2-4}  \cmidrule(r){5-5}  \cmidrule(r){6-7}  \cmidrule(r){8-8}  \cmidrule(r){9-14} 

\rowcolor{bg-tb-heavey-vision} \multicolumn{1}{c}{\multirow{-3}{*}{\bf  Model}}  &  \#1 $\uparrow$ & \#2 $\uparrow$ & \#3 $\uparrow$ & \#1 $\uparrow$ & \#1 $\uparrow$ & \#2 $\uparrow$ & \#1 $\uparrow$ & \#1 $\uparrow$ & \#2 $\uparrow$ & \#3 $\uparrow$ & \#4 $\uparrow$ & \#5 $\uparrow$ & \#6 $\uparrow$ \\

\midrule\midrule

\multirow{1}{*}{\bf SoTA Specialist} & 21.70 & 82.30 & 49.80 & 53.32 & 63.08 & 21.00 & 22.30 & 40.52 & 40.50 & 35.90 & 38.42 & 16.40 & 62.40 \\

\midrule

GPT4V & \textbf{57.00} & \textbf{87.97} & \textbf{63.30} & \textbf{58.64} & \textbf{79.08} & 0.00 & 0.00 & \textbf{56.60} & \textbf{99.75} & \textbf{90.92} & 38.01 & \textbf{49.40} & 62.40 \\
\rowcolor{bg-tb-light-vision}
GPT4o & \textbf{58.70} & \textbf{90.10} & \textbf{72.80} & \textbf{63.42} & \textbf{86.46} & 0.00 & 0.00 & \textbf{69.25} & \textbf{98.64} & \textbf{94.97} & \textbf{50.89} & \textbf{53.00} & \textbf{62.60} \\
GPT4o-mini & \textbf{25.70} & \textbf{88.50} & \textbf{70.20} & \textbf{61.38} & \textbf{77.85} & 0.00 & 0.00 & \textbf{75.00} & \textbf{96.93} & \textbf{50.14} & \textbf{44.55} & \textbf{40.40} & 49.60 \\
\rowcolor{bg-tb-light-vision}
GPT-4o-4096 & \textbf{24.90} & \textbf{90.00} & \textbf{89.40} & 0.00 & \textbf{89.54} & 0.00 & 0.00 & \textbf{74.43} & \textbf{99.75} & \textbf{99.88} & \textbf{63.56} & \textbf{52.00} & 62.40 \\
ChatGPT-4o-latest & \textbf{29.60} & 66.90 & \textbf{80.40} & \textbf{57.28} & \textbf{85.54} & 0.00 & 0.00 & \textbf{63.22} & \textbf{98.40} & \textbf{97.23} & 37.21 & \textbf{47.20} & 59.34 \\
\rowcolor{bg-tb-light-vision}
Claude-3.5-Sonnet & \textbf{46.67} & \textbf{87.90} & \textbf{68.60} & \textbf{60.86} & \textbf{80.30} & 0.00 & 0.00 & \textbf{66.32} & \textbf{97.72} & \textbf{78.40} & \textbf{44.37} & \textbf{47.38} & 57.23 \\
Claude-3.5-Opus & \textbf{43.16} & \textbf{88.58} & \textbf{64.40} & \textbf{57.69} & \textbf{79.90} & 0.00 & 0.00 & \textbf{63.10} & \textbf{96.64} & \textbf{76.74} & \textbf{43.48} & \textbf{45.40} & 54.76 \\
\rowcolor{bg-tb-light-vision}
Emu2-32B & \textbf{32.11} & 82.26 & 46.91 & 7.31 & \textbf{73.23} & 0.00 & 0.00 & \textbf{41.67} & 40.50 & 35.90 & \textbf{40.99} & \textbf{38.40} & 50.40 \\
DetGPT & \textbf{24.49} & 76.13 & \textbf{55.92} & 0.00 & 59.07 & 16.08 & 3.25 & 31.89 & 39.98 & 28.73 & 31.09 & \textbf{26.20} & 44.60 \\
\rowcolor{bg-tb-light-vision}
InternVL2.5-8B & \textbf{27.10} & 73.57 & \textbf{79.20} & 4.86 & 49.85 & 0.00 & 0.00 & \textbf{48.56} & \textbf{53.44} & 6.15 & 30.11 & \textbf{27.20} & \textbf{63.00} \\
InternVL2.5-4B & \textbf{32.30} & 63.77 & 39.80 & 4.97 & 34.15 & 0.00 & 0.00 & \textbf{60.06} & \textbf{46.31} & 21.65 & 36.63 & \textbf{25.20} & 62.20 \\
\rowcolor{bg-tb-light-vision}
InternVL2.5-2B & \textbf{26.80} & 76.73 & \textbf{61.80} & 4.64 & \textbf{66.22} & 0.00 & 0.00 & 37.36 & \textbf{90.79} & 23.04 & \textbf{39.41} & \textbf{31.20} & 60.40 \\
Monkey-10B-chat & \textbf{24.60} & \textbf{83.03} & 47.80 & 0.00 & 37.23 & 0.00 & 0.00 & 36.49 & \textbf{94.59} & 0.00 & 28.12 & \textbf{38.00} & 62.40 \\
\rowcolor{bg-tb-light-vision}
DeepSeek-VL-7B-Chat & 20.00 & 79.70 & \textbf{60.20} & 4.27 & \textbf{67.57} & 0.00 & 0.00 & 37.64 & \textbf{100.00} & 19.57 & \textbf{43.56} & \textbf{28.20} & 62.40 \\
Qwen2-VL-7B & 20.00 & \textbf{85.13} & \textbf{89.40} & 3.61 & 58.78 & 0.00 & 0.00 & \textbf{72.70} & \textbf{48.86} & \textbf{73.72} & \textbf{42.38} & \textbf{38.40} & 56.20 \\
\rowcolor{bg-tb-light-vision}
Qwen-VL-Chat & 20.00 & 72.66 & \textbf{56.39} & 3.40 & 44.30 & 0.00 & 0.00 & 27.01 & \textbf{48.85} & \textbf{40.05} & 26.93 & \textbf{20.40} & \textbf{63.20} \\
MoE-LLAVA-Phi2-2.7B-4e-384 & 20.20 & 67.67 & \textbf{63.55} & 1.90 & \textbf{64.62} & 0.00 & 0.00 & \textbf{52.59} & \textbf{48.86} & 34.24 & 36.83 & \textbf{20.00} & \textbf{62.60} \\
\rowcolor{bg-tb-light-vision}
mPLUG-Owl2-LLaMA2-7b & 21.09 & 75.76 & \textbf{60.74} & 0.00 & 52.00 & 0.00 & 0.00 & \textbf{47.87} & \textbf{47.60} & \textbf{40.00} & 18.02 & \textbf{21.20} & 45.60 \\
Phi-3.5-Vision-Instruct & 20.40 & \textbf{83.97} & \textbf{61.60} & 3.44 & \textbf{68.31} & 0.00 & 0.00 & \textbf{50.86} & \textbf{59.54} & \textbf{54.34} & 25.74 & \textbf{26.80} & 38.40 \\
\rowcolor{bg-tb-light-vision}
Cambrian-1-8B & 11.10 & 75.73 & 48.29 & 0.00 & 55.08 & 0.00 & 0.00 & 38.51 & 0.42 & \textbf{43.00} & 33.66 & \textbf{19.60} & 5.60 \\
MiniGPT4-LLaMA2-7B & \textbf{28.80} & 58.70 & 27.52 & 0.40 & \textbf{65.23} & 0.00 & 0.00 & \textbf{51.44} & \textbf{82.65} & \textbf{77.18} & \textbf{97.32} & 10.74 & 55.03 \\
\rowcolor{bg-tb-light-vision}
InternVL-Chat-V1-5 & \textbf{45.80} & \textbf{86.90} & \textbf{81.40} & 0.00 & \textbf{79.60} & 0.00 & 0.00 & \textbf{64.30} & \textbf{83.60} & \textbf{56.00} & 34.60 & \textbf{32.60} & 49.60 \\
Mini-InternVL-Chat-4B-V1-5 & \textbf{34.40} & 81.10 & \textbf{53.40} & 0.00 & \textbf{72.60} & 0.00 & 0.00 & \textbf{64.60} & \textbf{65.70} & \textbf{58.20} & 27.90 & \textbf{27.00} & 56.00 \\
\rowcolor{bg-tb-light-vision}
InternLM-XComposer2-VL-1.8B & 20.60 & 79.90 & 43.40 & 3.40 & \textbf{75.00} & 0.00 & 0.00 & \textbf{42.50} & \textbf{78.90} & \textbf{67.50} & 21.30 & \textbf{21.60} & 33.40 \\
GPT4RoI & 21.70 & 57.20 & 35.80 & 8.10 & \textbf{69.80} & 0.00 & 0.00 & \textbf{51.10} & 24.00 & 14.30 & 3.70 & \textbf{16.70} & 51.60 \\
\rowcolor{bg-tb-light-vision}
GLaMM & 0.00 & 0.00 & 0.00 & 0.00 & 0.00 & 0.00 & 0.00 & 0.00 & 0.00 & 0.00 & 0.00 & 0.00 & 0.00 \\
LLaVA-NeXT-13B & \textbf{42.60} & 80.73 & \textbf{66.80} & 8.67 & \textbf{73.85} & 0.00 & 0.00 & \textbf{52.87} & \textbf{51.97} & \textbf{47.21} & \textbf{42.18} & \textbf{35.60} & 44.80 \\
\rowcolor{bg-tb-light-vision}
LLaVA-NeXT-34B & \textbf{52.60} & \textbf{83.90} & \textbf{70.40} & 10.32 & \textbf{76.00} & 0.00 & 0.00 & \textbf{60.35} & \textbf{56.39} & \textbf{43.44} & \textbf{39.80} & \textbf{37.40} & 55.20 \\
Pixtral 12B & \textbf{46.40} & 77.43 & \textbf{65.20} & 3.25 & \textbf{65.85} & 0.00 & 0.00 & \textbf{54.02} & \textbf{46.07} & \textbf{52.93} & 34.06 & \textbf{31.80} & 49.20 \\
\rowcolor{bg-tb-light-vision}
SEED-LLaMA-13B & \textbf{28.30} & 57.93 & \textbf{53.80} & 0.00 & \textbf{63.69} & 0.00 & 0.00 & 34.48 & \textbf{46.68} & \textbf{55.87} & 30.69 & \textbf{28.20} & 47.60 \\
BLIP2 & \textbf{33.90} & 71.47 & 44.60 & 0.00 & 51.20 & 0.00 & 0.00 & \textbf{46.26} & \textbf{40.91} & 19.13 & 30.10 & \textbf{25.00} & 42.80 \\
\rowcolor{bg-tb-light-vision}
MiniMonkey & 4.90 & 34.77 & 49.40 & 0.00 & 12.62 & 0.00 & 0.00 & 0.00 & 0.00 & 0.00 & 0.20 & 4.08 & \textbf{69.20} \\
DeepSeek-VL-7B & \textbf{23.30} & 79.53 & \textbf{57.80} & 0.00 & \textbf{67.69} & 0.00 & 0.00 & \textbf{42.66} & \textbf{95.45} & \textbf{37.50} & 32.28 & \textbf{26.20} & \textbf{64.60} \\
\rowcolor{bg-tb-light-vision}
LISA & 0.00 & 0.00 & 0.00 & 0.00 & 0.00 & 0.00 & 0.00 & 0.00 & 0.00 & 0.00 & 0.00 & 0.00 & 0.00 \\
CogVLM-Chat & \textbf{46.35} & \textbf{85.57} & \textbf{64.40} & 12.32 & \textbf{84.62} & 0.00 & 0.00 & \textbf{54.31} & \textbf{58.48} & \textbf{52.51} & \textbf{42.38} & \textbf{36.80} & 54.20 \\
\rowcolor{bg-tb-light-vision}
ShareGPT4V-7B & \textbf{40.76} & 78.20 & \textbf{54.20} & 8.45 & \textbf{72.92} & 0.00 & 0.00 & 38.79 & \textbf{49.51} & \textbf{48.32} & 30.50 & \textbf{31.40} & 48.40 \\
ShareGPT4V-13B & \textbf{44.19} & 80.03 & \textbf{57.60} & 10.37 & \textbf{75.69} & 0.00 & 0.00 & \textbf{49.43} & \textbf{53.69} & \textbf{51.96} & 35.45 & \textbf{35.60} & 52.60 \\
\rowcolor{bg-tb-light-vision}
BLIP-3 (XGen-MM) & \textbf{43.28} & \textbf{82.63} & \textbf{61.80} & 8.54 & \textbf{82.15} & 0.00 & 0.00 & \textbf{52.87} & \textbf{59.71} & \textbf{55.45} & 37.43 & \textbf{34.20} & 50.60 \\
AnyGPT & 12.20 & 45.70 & 47.60 & 0.00 & 53.54 & 0.00 & 0.00 & 17.15 & \textbf{43.98} & \textbf{47.17} & 20.59 & 11.20 & 18.00 \\
\rowcolor{bg-tb-light-vision}
MiniCPM3-4B & \textbf{47.30} & 79.13 & \textbf{68.80} & 4.68 & \textbf{74.77} & 0.00 & 0.00 & \textbf{57.18} & \textbf{56.76} & \textbf{49.16} & \textbf{39.80} & \textbf{36.80} & 49.20 \\
LaVIT-V2 (7B) & \textbf{36.90} & 55.17 & \textbf{55.40} & 0.00 & 62.15 & 0.00 & 0.00 & 34.48 & \textbf{49.88} & \textbf{52.23} & 32.87 & \textbf{27.40} & 48.60 \\
\rowcolor{bg-tb-light-vision}
GLM-VL-Chat & \textbf{51.10} & 72.20 & \textbf{70.20} & 17.92 & \textbf{81.85} & 0.00 & 0.00 & \textbf{63.79} & \textbf{53.44} & \textbf{56.28} & \textbf{44.16} & \textbf{39.80} & 54.80 \\
Gemini-1.5-Pro & \textbf{47.60} & \textbf{91.20} & \textbf{78.20} & 23.41 & \textbf{78.77} & 0.00 & 0.00 & \textbf{69.54} & \textbf{98.90} & \textbf{54.89} & \textbf{44.55} & \textbf{46.20} & 60.20 \\
\rowcolor{bg-tb-light-vision}
Gemini-1.5-Flash & \textbf{42.10} & \textbf{87.30} & \textbf{71.60} & 25.79 & \textbf{75.69} & 0.00 & 0.00 & \textbf{67.24} & \textbf{98.03} & \textbf{52.65} & \textbf{40.20} & \textbf{42.20} & 56.40 \\
OMG-LLaVA-InternLM20B & 3.10 & 4.20 & 1.40 & 4.03 & 3.70 & 0.00 & 0.00 & 6.90 & 7.25 & 3.77 & 1.58 & 2.60 & 3.40 \\
\rowcolor{bg-tb-light-vision}
Idefics3-8B-Llama3 & \textbf{43.10} & 79.30 & \textbf{62.80} & 11.50 & \textbf{72.00} & 0.00 & 0.00 & \textbf{58.62} & \textbf{69.41} & \textbf{50.98} & \textbf{42.19} & \textbf{41.80} & 52.20 \\
NExT-GPT-V1.5 & \textbf{34.89} & 66.38 & 39.87 & 3.70 & \textbf{65.34} & 0.00 & 0.00 & 36.45 & \textbf{43.79} & \textbf{53.47} & 21.56 & 15.40 & 32.50 \\
\rowcolor{bg-tb-light-vision}
Vitron-V1 & \textbf{39.78} & 60.75 & 42.39 & 3.90 & \textbf{66.75} & \textbf{36.40} & 2.30 & 29.38 & \textbf{50.32} & \textbf{53.46} & 32.42 & 13.68 & 34.70 \\
Otter & 0.00 & 7.70 & 31.26 & 0.00 & 0.00 & 0.00 & 0.00 & 18.15 & 0.00 & 0.00 & 0.00 & 1.00 & 0.20 \\
\rowcolor{bg-tb-light-vision}
Show-o & 20.00 & 0.18 & 4.21 & 0.00 & 0.00 & 0.00 & 0.00 & 31.89 & \textbf{51.14} & \textbf{54.33} & 20.19 & 2.21 & 37.40 \\
NExT-Chat & 19.92 & 45.76 & \textbf{63.54} & 0.00 & \textbf{68.12} & 0.00 & 0.00 & 26.82 & \textbf{42.61} & \textbf{47.31} & 11.74 & \textbf{22.41} & 16.64 \\
\rowcolor{bg-tb-light-vision}
Yi-vision-v2 & \textbf{24.60} & 79.40 & \textbf{82.96} & 2.13 & \textbf{76.31} & 0.00 & 0.00 & \textbf{47.13} & \textbf{52.95} & 15.78 & \textbf{40.79} & \textbf{30.20} & 60.00 \\
Qwen2-VL-72B & \textbf{24.13} & \textbf{84.73} & \textbf{92.08} & 5.74 & \textbf{71.27} & 0.00 & 0.00 & \textbf{80.79} & \textbf{46.72} & \textbf{53.24} & \textbf{55.30} & \textbf{41.00} & 62.40 \\

\bottomrule
\end{tabular}%
\end{table*}
\clearpage

\begin{table*}[t!]
\centering
  \fontsize{8.5}{10.5}\selectfont 
  \setlength{\tabcolsep}{1.8mm}
\caption{
Results on \textcolor{greenCom}{Image Comprehension Group},\#I-C-6, part A.
}
\vspace{-2mm}
\label{tab:overall-results-image-comp-I-C-6-a}
\begin{tabular}{p{4.2cm} ccccccccccc}
\toprule
\rowcolor{bg-tb-heavey-vision} & \multicolumn{11}{c}{\bf \#I-C-6}  \\ 

\rowcolor{bg-tb-heavey-vision}  & \multicolumn{11}{c}{\tiny (Doc VQA)}  \\

\cmidrule(r){2-12}  

\rowcolor{bg-tb-heavey-vision} \multicolumn{1}{c}{\multirow{-3}{*}{\bf  Model}}  &  \#1 $\uparrow$ & \#2 $\uparrow$ & \#3 $\uparrow$ & \#4 $\uparrow$ & \#5 $\uparrow$ & \#6 $\uparrow$ & \#7 $\uparrow$ & \#8 $\uparrow$ & \#9 $\uparrow$ & \#10 $\uparrow$ & \#11 $\uparrow$ \\

\midrule\midrule

\multirow{1}{*}{\bf SoTA Specialist} & 32.72 & 10.64 & 11.76 & 19.40 & 17.80 & 26.71 & 20.92 & 30.94 & 16.11 & 23.91 & 13.79 \\

\midrule

GPT4V & \textbf{45.23} & \textbf{40.78} & \textbf{35.64} & \textbf{38.80} & \textbf{46.46} & \textbf{35.61} & \textbf{28.61} & 27.61 & \textbf{30.61} & \textbf{31.62} & \textbf{28.61} \\
\rowcolor{bg-tb-light-vision}
GPT4o & \textbf{48.16} & \textbf{48.58} & \textbf{41.58} & \textbf{42.63} & \textbf{50.51} & \textbf{39.04} & \textbf{29.04} & 30.04 & \textbf{31.04} & \textbf{32.04} & \textbf{31.23} \\
GPT4o-mini & \textbf{41.18} & \textbf{38.30} & \textbf{40.59} & \textbf{38.80} & \textbf{45.45} & \textbf{30.83} & \textbf{30.83} & \textbf{31.83} & \textbf{32.83} & \textbf{33.83} & \textbf{30.23} \\
\rowcolor{bg-tb-light-vision}
GPT-4o-4096 & \textbf{49.63} & \textbf{47.52} & \textbf{46.53} & \textbf{46.45} & \textbf{51.18} & \textbf{45.21} & \textbf{43.51} & \textbf{46.41} & \textbf{48.59} & \textbf{42.81} & \textbf{46.29} \\
\bottomrule
\end{tabular}%
\end{table*}
\clearpage

\begin{table*}[t!]
\centering
  \fontsize{8.5}{10.5}\selectfont 
  \setlength{\tabcolsep}{2mm}
\caption{
Results on \textcolor{greenCom}{Image Comprehension Group},\#I-C-6, part B.
}
\vspace{-2mm}
\label{tab:overall-results-image-comp-I-C-6-b}
\begin{tabular}{p{4.2cm} cccccccccc}
\toprule
\rowcolor{bg-tb-heavey-vision} & \multicolumn{10}{c}{\bf \#I-C-6}  \\ 

\rowcolor{bg-tb-heavey-vision}  & \multicolumn{10}{c}{\tiny (Doc VQA)}  \\

\cmidrule(r){2-11}  

\rowcolor{bg-tb-heavey-vision} \multicolumn{1}{c}{\multirow{-3}{*}{\bf  Model}}  &  \#12 $\uparrow$ & \#13 $\uparrow$ & \#14 $\uparrow$ & \#15 $\uparrow$ & \#16 $\uparrow$ & \#17 $\uparrow$ & \#18 $\uparrow$ & \#19 $\uparrow$ & \#20 $\uparrow$ & \#21 $\uparrow$ \\

\midrule\midrule

\multirow{1}{*}{\bf SoTA Specialist} & 11.76 & 21.51 & 21.51 & 24.06 & 14.02 & 11.64 & 23.26 & 16.67 & 19.51 & 82.20 \\

\midrule

GPT4V & \textbf{32.35} & \textbf{53.70} & \textbf{35.70} & \textbf{34.10} & \textbf{25.90} & \textbf{14.30} & \textbf{51.50} & \textbf{41.20} & \textbf{36.20} & 43.21 \\
\rowcolor{bg-tb-light-vision}
GPT4o & \textbf{33.82} & \textbf{50.90} & \textbf{44.00} & \textbf{35.70} & \textbf{26.10} & \textbf{43.80} & \textbf{48.70} & \textbf{42.20} & \textbf{34.30} & 49.16 \\
GPT4o-mini & \textbf{29.42} & \textbf{40.40} & \textbf{42.00} & \textbf{30.90} & \textbf{33.30} & \textbf{25.00} & \textbf{39.60} & \textbf{40.20} & \textbf{31.60} & 40.08 \\
\rowcolor{bg-tb-light-vision}
GPT-4o-4096 & \textbf{38.24} & \textbf{48.11} & \textbf{44.37} & \textbf{45.94} & \textbf{42.58} & \textbf{44.60} & \textbf{53.49} & \textbf{46.67} & \textbf{47.76} & \textbf{95.90} \\
\bottomrule
\end{tabular}%
\end{table*}
\clearpage

\begin{table*}[t!]
\centering
  \fontsize{8.5}{10.5}\selectfont 
  \setlength{\tabcolsep}{1.mm}
\caption{
Results on \textcolor{greenCom}{Image Comprehension Group}, from \#I-C-7 to \#I-C-10.
}
\vspace{-2mm}
\label{tab:overall-results-image-comp-I-C-7-8-9-10}
\begin{tabular}{p{4.2cm} cccc c cc cccccc}
\toprule
\rowcolor{bg-tb-heavey-vision} & \multicolumn{4}{c}{\bf \#I-C-7} & \multicolumn{1}{c}{\bf \#I-C-8} & \multicolumn{2}{c}{\bf \#I-C-9} & \multicolumn{6}{c}{\bf \#I-C-10} \\ 

\rowcolor{bg-tb-heavey-vision}  & \multicolumn{4}{c}{\tiny (Emotion Det)} &  \multicolumn{1}{c}{\tiny (Graph Cls)} & \multicolumn{2}{c}{\tiny (Hallucination Det)}  &   \multicolumn{6}{c}{\tiny (Img Cap)}\\

\cmidrule(r){2-5}  \cmidrule(r){6-6}  \cmidrule(r){7-8}  \cmidrule(r){9-14} 

\rowcolor{bg-tb-heavey-vision} \multicolumn{1}{c}{\multirow{-3}{*}{\bf  Model}}  &  \#1 $\uparrow$ & \#2 $\uparrow$ & \#3 $\uparrow$ & \#4 $\uparrow$ & \#1 $\uparrow$  & \#1 $\uparrow$ & \#2 $\uparrow$ & \#1 $\uparrow$ & \#2 $\uparrow$ & \#3 $\uparrow$ & \#4 $\uparrow$ & \#5 $\uparrow$ & \#6 $\uparrow$ \\

\midrule\midrule

\multirow{1}{*}{\bf SoTA Specialist} & 31.12 & 59.44 & 41.20 & 52.30 & 15.67 & 41.60 & 60.80 & 17.84 & 62.99 & 23.05 & 18.40 & 17.65 & 28.10 \\

\midrule

GPT4V & \textbf{51.06} & \textbf{61.45} & \textbf{53.83} & \textbf{80.60} & 0.00 & 12.80 & 21.00 & 2.63 & 42.22 & 15.80 & \textbf{18.70} & \textbf{27.21} & 18.69 \\
\rowcolor{bg-tb-light-vision}
GPT4o & \textbf{63.65} & \textbf{63.86} & \textbf{61.51} & \textbf{86.30} & 0.00 & \textbf{69.60} & \textbf{66.00} & 1.30 & 48.10 & 16.51 & \textbf{19.81} & \textbf{30.42} & 23.30 \\
GPT4o-mini & \textbf{61.38} & 56.62 & \textbf{64.57} & \textbf{84.30} & 0.00 & 40.80 & \textbf{78.20} & 4.00 & 43.10 & 18.64 & 15.05 & \textbf{25.62} & 18.60 \\
\rowcolor{bg-tb-light-vision}
GPT-4o-4096 & \textbf{65.63} & \textbf{64.66} & \textbf{66.39} & \textbf{88.70} & 0.00 & \textbf{68.20} & \textbf{70.80} & 1.00 & 45.45 & 17.36 & \textbf{20.68} & \textbf{28.86} & 15.72 \\
\bottomrule
\end{tabular}%
\end{table*}
\clearpage

\begin{table*}[t!]
\centering
  \fontsize{8.5}{10.5}\selectfont 
  \setlength{\tabcolsep}{2.2mm}
\caption{
Results on \textcolor{greenCom}{Image Comprehension Group}, from \#I-C-11 to \#I-C-14.
}
\vspace{-2mm}
\label{tab:overall-results-image-comp-I-C-11-12-13-14}
\begin{tabular}{p{4.2cm} c cc cccc cc}
\toprule
\rowcolor{bg-tb-heavey-vision} & \multicolumn{1}{c}{\bf \#I-C-11} & \multicolumn{2}{c}{\bf \#I-C-12} & \multicolumn{4}{c}{\bf \#I-C-13} & \multicolumn{2}{c}{\bf \#I-C-14} \\ 

\rowcolor{bg-tb-heavey-vision}  & \multicolumn{1}{c}{\tiny (Img Depth Est)} &  \multicolumn{2}{c}{\tiny (Img Inst Seg)} & \multicolumn{4}{c}{\tiny (Img OCR)}  &  \multicolumn{2}{c}{\tiny (Img Recog)} \\

\cmidrule(r){2-2}  \cmidrule(r){3-4}  \cmidrule(r){5-8} \cmidrule(r){9-10} 

\rowcolor{bg-tb-heavey-vision} \multicolumn{1}{c}{\multirow{-3}{*}{\bf  Model}} & \#1 $\downarrow$  & \#1 $\uparrow$ & \#2 $\uparrow$ &  \#1 $\uparrow$ & \#2 $\uparrow$ & \#3 $\uparrow$ & \#4 $\uparrow$ & \#1 $\uparrow$ & \#2 $\uparrow$ \\

\midrule\midrule

\multirow{1}{*}{\bf SoTA Specialist} & 36.40 & 66.50 & 63.80 & 45.03 & 95.14 & 10.23 & 24.70 & 29.80 & 88.00 \\

\midrule

GPT4V & $\infty$ & 0.00 & 0.00 & \textbf{89.71} & 35.68 & \textbf{60.45} & 18.30 & \textbf{32.50} & \textbf{94.53} \\
\rowcolor{bg-tb-light-vision}
GPT4o & $\infty$ & 0.00 & 0.00 & \textbf{94.07} & \textbf{97.36} & \textbf{74.01} & 19.46 & 27.83 & \textbf{95.24} \\
GPT4o-mini & $\infty$ & 0.00 & 0.00 & \textbf{85.10} & \textbf{96.56} & \textbf{73.12} & 17.42 & 23.83 & \textbf{91.32} \\
\rowcolor{bg-tb-light-vision}
GPT-4o-4096 & $\infty$ & 0.00 & 0.00 & \textbf{94.22} & \textbf{96.67} & \textbf{74.45} & 0.00 & \textbf{34.00} & 32.20 \\
\bottomrule
\end{tabular}%
\end{table*}
\clearpage

\begin{table*}[t!]
\centering
  \fontsize{8.5}{10.5}\selectfont 
  \setlength{\tabcolsep}{2.4mm}
\caption{
Results on \textcolor{greenCom}{Image Comprehension Group}, from \#I-C-15 to \#I-C-16.
}
\vspace{-2mm}
\label{tab:overall-results-image-comp-I-C-15-16}
\begin{tabular}{p{4.2cm} cccccccc cc}
\toprule
\rowcolor{bg-tb-heavey-vision} & \multicolumn{8}{c}{\bf \#I-C-15} & \multicolumn{2}{c}{\bf \#I-C-16}  \\ 

\rowcolor{bg-tb-heavey-vision}  & \multicolumn{8}{c}{\tiny (Img Sem Seg)} &  \multicolumn{2}{c}{\tiny (Img Vis Grnd)}  \\

\cmidrule(r){2-9}  \cmidrule(r){10-11} 

\rowcolor{bg-tb-heavey-vision} \multicolumn{1}{c}{\multirow{-3}{*}{\bf  Model}}  &  \#1 $\uparrow$ & \#2 $\uparrow$ & \#3 $\uparrow$ & \#4 $\uparrow$ & \#5 $\uparrow$ & \#6 $\uparrow$ & \#7 $\uparrow$  & \#8 $\uparrow$ &  \#1 $\uparrow$ & \#2 $\uparrow$\\

\midrule\midrule

\multirow{1}{*}{\bf SoTA Specialist} & 19.80 & 50.40 & 88.00 & 98.70 & 83.65 & 59.36 & 47.74 & 62.15 & 84.76 & 90.91 \\

\midrule

GPT4V & 0.00 & 0.00 & 0.00 & 0.00 & 0.00 & 0.00 & 0.00 & 0.00 & 67.69 & 74.11 \\
\rowcolor{bg-tb-light-vision}
GPT4o & 0.00 & 0.00 & 0.00 & 0.00 & 0.00 & 0.00 & 0.00 & 0.00 & 76.20 & 82.56 \\
GPT4o-mini & 0.00 & 0.00 & 0.00 & 0.00 & 0.00 & 0.00 & 0.00 & 0.00 & 57.18 & 59.44 \\
\rowcolor{bg-tb-light-vision}
GPT-4o-4096 & 0.00 & 0.00 & 0.00 & 0.00 & 0.00 & 0.00 & 0.00 & 0.00 & 80.60 & 83.33 \\
\bottomrule
\end{tabular}%
\end{table*}
\clearpage

\begin{table*}[t!]
\centering
  \fontsize{8.5}{10.5}\selectfont 
  \setlength{\tabcolsep}{2.mm}
\caption{
Results on \textcolor{greenCom}{Image Comprehension Group},\#I-C-17, part A.
}
\vspace{-2mm}
\label{tab:overall-results-image-comp-I-C-17-a}
\begin{tabular}{p{4.2cm} cccccccccc}
\toprule
\rowcolor{bg-tb-heavey-vision} & \multicolumn{10}{c}{\bf \#I-C-17}  \\ 

\rowcolor{bg-tb-heavey-vision}  & \multicolumn{10}{c}{\tiny (Img VQA)}  \\

\cmidrule(r){2-11}  

\rowcolor{bg-tb-heavey-vision} \multicolumn{1}{c}{\multirow{-3}{*}{\bf  Model}}  &  \#1 $\uparrow$ & \#2 $\uparrow$ & \#3 $\uparrow$ & \#4 $\uparrow$ & \#5 $\uparrow$ & \#6 $\uparrow$ & \#7 $\uparrow$ & \#8 $\uparrow$ & \#9 $\uparrow$ & \#10 $\uparrow$ \\

\midrule\midrule

\multirow{1}{*}{\bf SoTA Specialist} & 14.36 & 36.40 & 17.32 & 43.12 & 89.38 & 17.50 & 34.30 & 80.75 & 92.72 & 78.30 \\

\midrule

GPT4V & \textbf{16.20} & 26.54 & \textbf{52.84} & 41.04 & 80.38 & \textbf{45.74} & \textbf{37.00} & 57.33 & 87.54 & 43.33 \\
\rowcolor{bg-tb-light-vision}
GPT4o & \textbf{16.00} & 23.82 & \textbf{50.31} & 31.41 & 85.75 & \textbf{69.03} & \textbf{37.00} & 53.00 & \textbf{96.10} & 51.00 \\
GPT4o-mini & 11.45 & 17.45 & \textbf{47.24} & 29.09 & 82.01 & \textbf{44.03} & 34.00 & 33.33 & \textbf{93.10} & 37.67 \\
\rowcolor{bg-tb-light-vision}
GPT-4o-4096 & \textbf{24.40} & 22.73 & \textbf{53.78} & \textbf{43.16} & 89.02 & \textbf{74.15} & \textbf{40.00} & 52.00 & \textbf{96.63} & 48.67 \\
\bottomrule
\end{tabular}%
\end{table*}
\clearpage

\begin{table*}[t!]
\centering
  \fontsize{8.5}{10.5}\selectfont 
  \setlength{\tabcolsep}{2.mm}
\caption{
Results on \textcolor{greenCom}{Image Comprehension Group},\#I-C-17, part B.
}
\vspace{-2mm}
\label{tab:overall-results-image-comp-I-C-17-b}
\begin{tabular}{p{4.2cm} cccccccccc}
\toprule
\rowcolor{bg-tb-heavey-vision} & \multicolumn{10}{c}{\bf \#I-C-17}  \\ 

\rowcolor{bg-tb-heavey-vision}  & \multicolumn{10}{c}{\tiny (Img VQA)}  \\

\cmidrule(r){2-11}  

\rowcolor{bg-tb-heavey-vision} \multicolumn{1}{c}{\multirow{-3}{*}{\bf  Model}}  &  \#11 $\uparrow$ & \#12 $\uparrow$ & \#13 $\uparrow$ & \#14 $\uparrow$ & \#15 $\uparrow$ & \#16 $\uparrow$ & \#17 $\uparrow$ & \#18 $\uparrow$ & \#19 $\uparrow$ & \#20 $\uparrow$ \\

\midrule\midrule

\multirow{1}{*}{\bf SoTA Specialist} & 49.80 & 65.40 & 66.50 & 74.70 & 73.58 & 24.40 & 71.60 & 70.70 & 51.60 & 71.40 \\

\midrule

GPT4V & 0.00 & 50.00 & \textbf{75.70} & \textbf{81.82} & \textbf{76.39} & \textbf{34.25} & 58.70 & 66.00 & 47.33 & 70.40 \\
\rowcolor{bg-tb-light-vision}
GPT4o & 0.00 & \textbf{68.10} & \textbf{79.44} & 72.72 & \textbf{78.17} & \textbf{45.00} & 66.30 & 68.33 & 12.70 & \textbf{72.20} \\
GPT4o-mini & 0.00 & 56.90 & \textbf{71.96} & \textbf{77.27} & 69.27 & \textbf{28.00} & 56.52 & 69.67 & 12.60 & 63.20 \\
\rowcolor{bg-tb-light-vision}
GPT-4o-4096 & 0.00 & 64.66 & \textbf{83.18} & \textbf{81.82} & \textbf{79.29} & \textbf{45.25} & 54.35 & 68.33 & 50.33 & \textbf{75.60} \\
\bottomrule
\end{tabular}%
\end{table*}
\clearpage

\begin{table*}[t!]
\centering
  \fontsize{8.5}{10.5}\selectfont 
  \setlength{\tabcolsep}{2mm}
\caption{
Results on \textcolor{greenCom}{Image Comprehension Group},\#I-C-17, part C.
}
\vspace{-2mm}
\label{tab:overall-results-image-comp-I-C-17-c}
\begin{tabular}{p{4.2cm} cccccccccc}
\toprule
\rowcolor{bg-tb-heavey-vision} & \multicolumn{10}{c}{\bf \#I-C-17}  \\ 

\rowcolor{bg-tb-heavey-vision}  & \multicolumn{10}{c}{\tiny (Img VQA)}  \\

\cmidrule(r){2-11}  

\rowcolor{bg-tb-heavey-vision} \multicolumn{1}{c}{\multirow{-3}{*}{\bf  Model}}  &  \#21 $\uparrow$ & \#22 $\uparrow$ & \#23 $\uparrow$ & \#24 $\uparrow$ & \#25 $\uparrow$ & \#26 $\uparrow$ & \#27 $\uparrow$ & \#28 $\uparrow$ & \#29 $\uparrow$ & \#30 $\uparrow$ \\

\midrule\midrule

\multirow{1}{*}{\bf SoTA Specialist} & 70.60 & 59.30 & 61.90 & 61.40 & 62.80 & 62.10 & 70.20 & 83.80 & 82.40 & 68.60 \\

\midrule

GPT4V & 68.60 & 22.81 & 20.40 & 24.10 & 22.64 & 22.02 & 18.01 & \textbf{93.00} & \textbf{91.40} & \textbf{71.20} \\
\rowcolor{bg-tb-light-vision}
GPT4o & \textbf{74.00} & 37.30 & 40.30 & 43.45 & 35.47 & 39.87 & 29.60 & \textbf{93.80} & \textbf{94.30} & 65.00 \\
GPT4o-mini & 65.60 & 26.34 & 23.70 & 29.07 & 26.52 & 25.01 & 25.20 & \textbf{90.50} & \textbf{89.50} & 68.50 \\
\rowcolor{bg-tb-light-vision}
GPT-4o-4096 & \textbf{77.00} & 27.95 & 49.51 & 52.05 & 43.15 & 48.03 & 43.31 & \textbf{94.40} & \textbf{95.60} & \textbf{69.60} \\
\bottomrule
\end{tabular}%
\end{table*}
\clearpage

\begin{table*}[t!]
\centering
  \fontsize{8.5}{10.5}\selectfont 
  \setlength{\tabcolsep}{2.mm}
\caption{
Results on \textcolor{greenCom}{Image Comprehension Group},\#I-C-17, part D.
}
\vspace{-2mm}
\label{tab:overall-results-image-comp-I-C-17-d}
\begin{tabular}{p{4.2cm} cccccccccc}
\toprule
\rowcolor{bg-tb-heavey-vision} & \multicolumn{10}{c}{\bf \#I-C-17}  \\ 

\rowcolor{bg-tb-heavey-vision}  & \multicolumn{10}{c}{\tiny (Img VQA)}  \\

\cmidrule(r){2-11}  

\rowcolor{bg-tb-heavey-vision} \multicolumn{1}{c}{\multirow{-3}{*}{\bf  Model}}  &  \#31 $\uparrow$ & \#32 $\uparrow$ & \#33 $\uparrow$ & \#34 $\uparrow$ & \#35 $\uparrow$ & \#36 $\uparrow$ & \#37 $\uparrow$ & \#38 $\uparrow$ & \#39 $\uparrow$ & \#40 $\uparrow$ \\

\midrule\midrule

\multirow{1}{*}{\bf SoTA Specialist} & 75.00 & 67.20 & 68.40 & 66.80 & 48.80 & 56.60 & 75.80 & 31.00 & 20.71 & 29.20 \\

\midrule

GPT4V & \textbf{77.10} & \textbf{70.40} & \textbf{73.90} & \textbf{82.80} & \textbf{89.40} & \textbf{98.20} & 14.80 & 19.60 & 16.83 & 18.30 \\
\rowcolor{bg-tb-light-vision}
GPT4o & 74.20 & \textbf{71.80} & \textbf{72.10} & \textbf{83.70} & \textbf{92.40} & \textbf{91.80} & 27.20 & 30.20 & 18.04 & 19.20 \\
GPT4o-mini & 70.50 & 35.80 & 39.00 & \textbf{80.60} & \textbf{77.60} & \textbf{63.40} & 10.60 & 7.80 & 15.23 & 18.70 \\
\rowcolor{bg-tb-light-vision}
GPT-4o-4096 & 75.00 & \textbf{76.60} & \textbf{77.40} & \textbf{83.00} & \textbf{91.00} & \textbf{87.00} & 25.80 & \textbf{47.51} & 15.83 & 0.00 \\
\bottomrule
\end{tabular}%
\end{table*}
\clearpage

\begin{table*}[t!]
\centering
  \fontsize{8.5}{10.5}\selectfont 
  \setlength{\tabcolsep}{2.mm}
\caption{
Results on \textcolor{greenCom}{Image Comprehension Group},\#I-C-18, part A.
}
\vspace{-2mm}
\label{tab:overall-results-image-comp-I-C-18-a}
\begin{tabular}{p{4.2cm} cccccccccc}
\toprule
\rowcolor{bg-tb-heavey-vision} & \multicolumn{10}{c}{\bf \#I-C-18}  \\ 

\rowcolor{bg-tb-heavey-vision}  & \multicolumn{10}{c}{\tiny (Ind-Anomaly Det)}  \\

\cmidrule(r){2-11}  

\rowcolor{bg-tb-heavey-vision} \multicolumn{1}{c}{\multirow{-3}{*}{\bf  Model}}  &  \#1 $\uparrow$ & \#2 $\uparrow$ & \#3 $\uparrow$ & \#4 $\uparrow$ & \#5 $\uparrow$ & \#6 $\uparrow$ & \#7 $\uparrow$ & \#8 $\uparrow$ & \#9 $\uparrow$ & \#10 $\uparrow$ \\

\midrule\midrule

\multirow{1}{*}{\bf SoTA Specialist} & 90.30 & 60.40 & 56.30 & 72.80 & 61.50 & 88.40 & 75.60 & 93.50 & 83.10 & 58.70 \\

\midrule

GPT4V & 0.00 & 0.00 & 0.00 & 0.00 & 0.00 & 0.00 & 0.00 & 0.00 & 0.00 & 0.00 \\
\rowcolor{bg-tb-light-vision}
GPT4o & 0.00 & 0.00 & 0.00 & 0.00 & 0.00 & 0.00 & 0.00 & 0.00 & 0.00 & 0.00 \\
GPT4o-mini & 0.00 & 0.00 & 0.00 & 0.00 & 0.00 & 0.00 & 0.00 & 0.00 & 0.00 & 0.00 \\
\rowcolor{bg-tb-light-vision}
GPT-4o-4096 & 0.00 & 0.00 & 0.00 & 0.00 & 0.00 & 0.00 & 0.00 & 0.00 & 0.00 & 0.00 \\
\bottomrule
\end{tabular}%
\end{table*}
\clearpage

\begin{table*}[t!]
\centering
  \fontsize{8.5}{10.5}\selectfont 
  \setlength{\tabcolsep}{2.mm}
\caption{
Results on \textcolor{greenCom}{Image Comprehension Group},\#I-C-18, part B.
}
\vspace{-2mm}
\label{tab:overall-results-image-comp-I-C-18-b}
\begin{tabular}{p{4.2cm} ccccccccc}
\toprule
\rowcolor{bg-tb-heavey-vision} & \multicolumn{9}{c}{\bf \#I-C-18}  \\ 

\rowcolor{bg-tb-heavey-vision}  & \multicolumn{9}{c}{\tiny (Ind-Anomaly Det)}  \\

\cmidrule(r){2-10}  

\rowcolor{bg-tb-heavey-vision} \multicolumn{1}{c}{\multirow{-3}{*}{\bf  Model}}  &  \#11 $\uparrow$ & \#12 $\uparrow$ & \#13 $\uparrow$ & \#14 $\uparrow$ & \#15 $\uparrow$ & \#16 $\uparrow$ & \#17 $\uparrow$ & \#18 $\uparrow$ & \#19 $\uparrow$ \\

\midrule\midrule

\multirow{1}{*}{\bf SoTA Specialist} & 90.00 & 75.20 & 64.80 & 81.90 & 85.20 & 62.10 & 76.20 & 23.10 & 73.60 \\

\midrule

GPT4V & 0.00 & 0.00 & 0.00 & 0.00 & 0.00 & 0.00 & 0.00 & 0.00 & 0.00 \\
\rowcolor{bg-tb-light-vision}
GPT4o & 0.00 & 0.00 & 0.00 & 0.00 & 0.00 & 0.00 & 0.00 & 0.00 & 0.00 \\
GPT4o-mini & 0.00 & 0.00 & 0.00 & 0.00 & 0.00 & 0.00 & 0.00 & 0.00 & 0.00 \\
\rowcolor{bg-tb-light-vision}
GPT-4o-4096 & 0.00 & 0.00 & 0.00 & 0.00 & 0.00 & 0.00 & 0.00 & 0.00 & 0.00 \\
\bottomrule
\end{tabular}%
\end{table*}
\clearpage

\begin{table*}[t!]
\centering
  \fontsize{8.5}{10.5}\selectfont 
  \setlength{\tabcolsep}{2.6mm}
\caption{
Results on \textcolor{greenCom}{Image Comprehension Group}, from \#I-C-19 to \#I-C-20.
}
\vspace{-2mm}
\label{tab:overall-results-image-comp-I-C-19-20}
\begin{tabular}{p{4.2cm} cccccc c}
\toprule
\rowcolor{bg-tb-heavey-vision} & \multicolumn{6}{c}{\bf \#I-C-19} & \multicolumn{1}{c}{\bf \#I-C-20} \\ 

\rowcolor{bg-tb-heavey-vision}  & \multicolumn{6}{c}{\tiny (Med Seg)}  &  \multicolumn{1}{c}{\tiny (Keypoint Det)} \\

\cmidrule(r){2-7}   \cmidrule(r){8-8} 

\rowcolor{bg-tb-heavey-vision} \multicolumn{1}{c}{\multirow{-3}{*}{\bf  Model}}  & \#1 $\uparrow$ & \#2 $\uparrow$ & \#3 $\uparrow$ & \#4 $\uparrow$ & \#5 $\uparrow$ & \#6 $\uparrow$ &  \#1 $\uparrow$ \\

\midrule\midrule

\multirow{1}{*}{\bf SoTA Specialist} & 50.10 & 16.30 & 25.85 & 23.55 & 53.40 & 37.88 & 95.70 \\

\midrule

GPT4V & 0.00 & 0.00 & 0.00 & 0.00 & 0.00 & 0.00 & 0.00 \\
\rowcolor{bg-tb-light-vision}
GPT4o & 0.00 & 0.00 & 0.00 & 0.00 & 0.00 & 0.00 & 0.00 \\
GPT4o-mini & 0.00 & 0.00 & 0.00 & 0.00 & 0.00 & 0.00 & 0.00 \\
\rowcolor{bg-tb-light-vision}
GPT-4o-4096 & 0.00 & 0.00 & 0.00 & 0.00 & 0.00 & 0.00 & 0.00 \\
\bottomrule
\end{tabular}%
\end{table*}
\clearpage

\begin{table*}[t!]
\centering
  \fontsize{8.5}{10.5}\selectfont 
  \setlength{\tabcolsep}{1.8mm}
\caption{
Results on \textcolor{greenCom}{Image Comprehension Group}, from \#I-C-21 to \#I-C-22.
}
\vspace{-2mm}
\label{tab:overall-results-image-comp-I-C-21-22}
\begin{tabular}{p{4.2cm} cccccc cccccc}
\toprule
\rowcolor{bg-tb-heavey-vision} & \multicolumn{6}{c}{\bf \#I-C-21} & \multicolumn{6}{c}{\bf \#I-C-22} \\ 

\rowcolor{bg-tb-heavey-vision}  & \multicolumn{6}{c}{\tiny (Multi-img VQA)} &  \multicolumn{6}{c}{\tiny (MM Dialog)} \\

\cmidrule(r){2-7}  \cmidrule(r){8-13} 

\rowcolor{bg-tb-heavey-vision} \multicolumn{1}{c}{\multirow{-3}{*}{\bf  Model}}  &  \#1 $\uparrow$ & \#2 $\uparrow$ & \#3 $\uparrow$ & \#4 $\uparrow$ & \#5 $\uparrow$ & \#6 $\uparrow$ & \#1 $\uparrow$ & \#2 $\uparrow$ & \#3 $\uparrow$ & \#4 $\uparrow$ & \#5 $\uparrow$ & \#6 $\uparrow$ \\

\midrule\midrule

\multirow{1}{*}{\bf SoTA Specialist} & 88.20 & 87.60 & 75.00 & 49.60 & 59.80 & 59.80 & 51.90 & 51.60 & 53.40 & 53.10 & 51.60 & 40.80 \\

\midrule

GPT4V & 46.20 & 47.20 & \textbf{82.40} & \textbf{94.00} & \textbf{84.80} & \textbf{76.80} & 23.45 & \textbf{55.06} & 28.00 & 52.38 & 37.05 & 26.80 \\
\rowcolor{bg-tb-light-vision}
GPT4o & 61.60 & 59.80 & \textbf{95.60} & \textbf{98.20} & \textbf{85.00} & \textbf{87.60} & 21.84 & \textbf{62.09} & 28.13 & \textbf{53.13} & 38.09 & 34.40 \\
GPT4o-mini & 26.80 & 28.40 & \textbf{96.60} & \textbf{97.80} & \textbf{80.80} & \textbf{66.20} & 22.79 & 47.74 & 24.94 & 42.82 & 32.13 & 11.60 \\
\rowcolor{bg-tb-light-vision}
GPT-4o-4096 & 51.40 & 50.00 & \textbf{96.00} & \textbf{98.80} & \textbf{91.00} & \textbf{78.20} & 20.88 & \textbf{59.21} & 30.22 & 50.20 & 40.12 & 2.40 \\
\bottomrule
\end{tabular}%
\end{table*}
\clearpage

\begin{table*}[t!]
\centering
  \fontsize{8.5}{10.5}\selectfont 
  \setlength{\tabcolsep}{1.2mm}
\caption{
Results on \textcolor{greenCom}{Image Comprehension Group}, \#I-C-23.
}
\vspace{-2mm}
\label{tab:overall-results-image-comp-I-C-23}
\begin{tabular}{p{4.2cm} cccccccccccccc}
\toprule
\rowcolor{bg-tb-heavey-vision} & \multicolumn{14}{c}{\bf \#I-C-23} \\ 

\rowcolor{bg-tb-heavey-vision}  & \multicolumn{14}{c}{\tiny (MM Reason)} \\

\cmidrule(r){2-15}   

\rowcolor{bg-tb-heavey-vision} \multicolumn{1}{c}{\multirow{-3}{*}{\bf  Model}}  & \#1 $\uparrow$ & \#2 $\uparrow$ & \#3 $\uparrow$ & \#4 $\uparrow$ & \#5 $\uparrow$ & \#6 $\uparrow$ & \#7 $\uparrow$ & \#8 $\uparrow$ & \#9 $\uparrow$ & \#10 $\uparrow$ & \#11 $\uparrow$  & \#12 $\uparrow$  & \#13 $\uparrow$  & \#14 $\uparrow$\\

\midrule\midrule

\multirow{1}{*}{\bf SoTA Specialist} & 68.80 & 69.80 & 77.20 & 50.00 & 49.80 & 64.60 & 80.20 & 70.60 & 80.60 & 47.20 & 66.20 & 67.60 & 74.20 & 56.80 \\

\midrule

GPT4V & 0.00 & 0.00 & 72.40 & 22.80 & 17.80 & \textbf{91.00} & \textbf{91.20} & \textbf{79.00} & \textbf{81.80} & 23.80 & 56.00 & 56.60 & \textbf{80.90} & 30.90 \\
\rowcolor{bg-tb-light-vision}
GPT4o & 0.00 & 0.00 & 62.20 & 10.40 & 8.80 & \textbf{97.40} & \textbf{97.40} & \textbf{75.80} & 77.40 & 37.00 & 50.40 & 51.40 & \textbf{81.40} & 31.20 \\
GPT4o-mini & 0.00 & 0.00 & 76.40 & 18.80 & 19.20 & \textbf{92.40} & \textbf{92.60} & \textbf{71.60} & 66.80 & 17.80 & 56.40 & 53.20 & 73.20 & 34.10 \\
\rowcolor{bg-tb-light-vision}
GPT-4o-4096 & 0.00 & 40.60 & 75.00 & 7.80 & 6.60 & \textbf{96.00} & \textbf{96.60} & \textbf{77.80} & 79.20 & 31.60 & 51.00 & 52.00 & \textbf{82.20} & 39.00 \\
\bottomrule
\end{tabular}%
\end{table*}
\clearpage

\begin{table*}[t!]
\centering
  \fontsize{8.5}{10.5}\selectfont 
  \setlength{\tabcolsep}{1.8mm}
\caption{
Results on \textcolor{greenCom}{Image Comprehension Group}, from \#I-C-24 to \#I-C-25.
}
\vspace{-2mm}
\label{tab:overall-results-image-comp-I-C-24-25}
\begin{tabular}{p{4.2cm} ccccccccc c }
\toprule
\rowcolor{bg-tb-heavey-vision} & \multicolumn{9}{c}{\bf \#I-C-24} & \multicolumn{1}{c}{\bf \#I-C-25} \\ 

\rowcolor{bg-tb-heavey-vision}  & \multicolumn{9}{c}{\tiny (Object Count)} &  \multicolumn{1}{c}{\tiny (NMT)} \\

\cmidrule(r){2-10}  \cmidrule(r){11-11} 

\rowcolor{bg-tb-heavey-vision} \multicolumn{1}{c}{\multirow{-3}{*}{\bf  Model}}  &  \#1 $\downarrow$ & \#2 $\downarrow$ & \#3 $\downarrow$ & \#4 $\downarrow$ & \#5 $\downarrow$ & \#6 $\downarrow$ & \#7 $\downarrow$ & \#8 $\downarrow$ & \#9 $\downarrow$ & \#1 $\uparrow$ \\

\midrule\midrule

\multirow{1}{*}{\bf SoTA Specialist} & 5.95 & 9.00 & 26.84 & 8.13 & 16.41 & 17.91 & 28.16 & 10.00 & 27.00 & 78.00 \\

\midrule

GPT4V & 8.41 & \textbf{3.20} & 31.29 & 9.79 & 26.91 & \textbf{10.72} & \textbf{10.86} & 32.21 & \textbf{11.14} & 72.20 \\
\rowcolor{bg-tb-light-vision}
GPT4o & 9.32 & \textbf{2.30} & 31.17 & 11.85 & 26.65 & \textbf{10.76} & \textbf{5.29} & 23.01 & \textbf{15.70} & \textbf{93.00} \\
GPT4o-mini & 10.06 & \textbf{2.80} & 30.98 & 13.62 & 43.87 & \textbf{12.87} & \textbf{10.56} & 26.30 & \textbf{14.66} & \textbf{83.00} \\
\rowcolor{bg-tb-light-vision}
GPT-4o-4096 & 7.47 & 23.66 & 31.11 & 14.65 & 21.75 & \textbf{9.93} & \textbf{4.69} & 16.26 & \textbf{16.00} & \textbf{88.40} \\
\bottomrule
\end{tabular}%
\end{table*}
\clearpage

\begin{table*}[t!]
\centering
  \fontsize{8.5}{10.5}\selectfont 
  \setlength{\tabcolsep}{1.4mm}
\caption{
Results on \textcolor{greenCom}{Image Comprehension Group},\#I-C-26, part A.
}
\vspace{-2mm}
\label{tab:overall-results-image-comp-I-C-26-a}
\begin{tabular}{p{4.2cm} ccccccccccccc}
\toprule
\rowcolor{bg-tb-heavey-vision} & \multicolumn{13}{c}{\bf \#I-C-26}  \\ 

\rowcolor{bg-tb-heavey-vision}  & \multicolumn{13}{c}{\tiny (Object Det)}  \\

\cmidrule(r){2-14}  

\rowcolor{bg-tb-heavey-vision} \multicolumn{1}{c}{\multirow{-3}{*}{\bf  Model}}  &  \#1 $\uparrow$ & \#2 $\uparrow$ & \#3 $\uparrow$ & \#4 $\uparrow$ & \#5 $\uparrow$ & \#6 $\uparrow$ & \#7 $\uparrow$ & \#8 $\uparrow$ & \#9 $\uparrow$ & \#10 $\uparrow$ & \#11 $\uparrow$ & \#12 $\uparrow$  & \#13 $\uparrow$\\

\midrule\midrule

\multirow{1}{*}{\bf SoTA Specialist} & 87.20 & 87.80 & 32.60 & 41.10 & 73.80 & 29.00 & 34.00 & 62.40 & 93.20 & 26.20 & 37.60 & 18.10 & 61.60 \\

\midrule

GPT4V & 0.00 & 0.00 & 0.00 & 0.00 & 0.00 & 0.00 & 0.00 & 0.00 & 0.00 & 0.00 & 0.00 & 0.00 & 0.00 \\
\rowcolor{bg-tb-light-vision}
GPT4o & 0.00 & 0.00 & 0.00 & 0.00 & 0.00 & 0.00 & 0.00 & 0.00 & 0.00 & 0.00 & 0.00 & 0.00 & 0.00 \\
GPT4o-mini & 0.00 & 0.00 & 0.00 & 0.00 & 0.00 & 0.00 & 0.00 & 0.00 & 0.00 & 0.00 & 0.00 & 0.00 & 0.00 \\
\rowcolor{bg-tb-light-vision}
GPT-4o-4096 & 0.00 & 0.00 & 0.00 & 0.00 & 0.00 & 0.00 & 0.00 & 0.00 & 0.00 & 0.00 & 0.00 & 0.00 & 0.00 \\
\bottomrule
\end{tabular}%
\end{table*}
\clearpage

\begin{table*}[t!]
\centering
  \fontsize{8.5}{10.5}\selectfont 
  \setlength{\tabcolsep}{1.1mm}
\caption{
Results on \textcolor{greenCom}{Image Comprehension Group},\#I-C-26, part B.
}
\vspace{-2mm}
\label{tab:overall-results-image-comp-I-C-26-b}
\begin{tabular}{p{4.2cm} ccccccccccccc}
\toprule
\rowcolor{bg-tb-heavey-vision} & \multicolumn{13}{c}{\bf \#I-C-26}  \\ 

\rowcolor{bg-tb-heavey-vision}  & \multicolumn{13}{c}{\tiny (Object Det)}  \\

\cmidrule(r){2-14}  

\rowcolor{bg-tb-heavey-vision} \multicolumn{1}{c}{\multirow{-3}{*}{\bf  Model}}  &  \#14 $\uparrow$ & \#15 $\uparrow$ & \#16 $\uparrow$ & \#17 $\uparrow$ & \#18 $\uparrow$ & \#19 $\uparrow$ & \#20 $\uparrow$ & \#21 $\uparrow$ & \#22 $\uparrow$  & \#23 $\uparrow$ &  \#24 $\uparrow$ & \#25 $\uparrow$ & \#26 $\uparrow$\\

\midrule\midrule

\multirow{1}{*}{\bf SoTA Specialist} & 27.20 & 68.30 & 81.10 & 86.40 & 18.20 & 17.20 & 31.60 & 78.90 & 28.30 & 48.10 & 3.50 & 39.40 & 22.40 \\

\midrule

GPT4V & 0.00 & 0.00 & 0.00 & 0.00 & 0.00 & 0.00 & 0.00 & 0.00 & 0.00 & 0.00 & 0.00 & 0.00 & 0.00 \\
\rowcolor{bg-tb-light-vision}
GPT4o & 0.00 & 0.00 & 0.00 & 0.00 & 0.00 & 0.00 & 0.00 & 0.00 & 0.00 & 0.00 & 0.00 & 0.00 & 0.00 \\
GPT4o-mini & 0.00 & 0.00 & 0.00 & 0.00 & 0.00 & 0.00 & 0.00 & 0.00 & 0.00 & 0.00 & 0.00 & 0.00 & 0.00 \\
\rowcolor{bg-tb-light-vision}
GPT-4o-4096 & 0.00 & 0.00 & 0.00 & 0.00 & 0.00 & 0.00 & 0.00 & 0.00 & 0.00 & 0.00 & 0.00 & 0.00 & 0.00 \\
\bottomrule
\end{tabular}%
\end{table*}
\clearpage

\begin{table*}[t!]
\centering
  \fontsize{8.5}{10.5}\selectfont 
  \setlength{\tabcolsep}{1.5mm}
\caption{
Results on \textcolor{greenCom}{Image Comprehension Group},\#I-C-26, part C.
}
\vspace{-2mm}
\label{tab:overall-results-image-comp-I-C-26-c}
\begin{tabular}{p{4.2cm} ccccccccccc}
\toprule
\rowcolor{bg-tb-heavey-vision} & \multicolumn{11}{c}{\bf \#I-C-26}  \\ 

\rowcolor{bg-tb-heavey-vision}  & \multicolumn{11}{c}{\tiny (Object Det)}  \\

\cmidrule(r){2-12}  

\rowcolor{bg-tb-heavey-vision} \multicolumn{1}{c}{\multirow{-3}{*}{\bf  Model}}  & \#27 $\uparrow$ & \#28 $\uparrow$ & \#29 $\uparrow$ & \#30 $\uparrow$ & \#31 $\uparrow$  &  \#32 $\uparrow$ & \#33 $\uparrow$ & \#34 $\uparrow$ & \#35 $\uparrow$ & \#36 $\uparrow$ & \#37 $\uparrow$ \\

\midrule\midrule

\multirow{1}{*}{\bf SoTA Specialist} & 67.60 & 63.40 & 10.90 & 48.60 & 38.00 & 97.10 & 61.10 & 97.50 & 25.90 & 58.20 & 64.20 \\

\midrule

GPT4V & 0.00 & 0.00 & 0.00 & 0.00 & 0.00 & 0.00 & 0.00 & 0.00 & 0.00 & 0.00 & 0.00 \\
\rowcolor{bg-tb-light-vision}
GPT4o & 0.00 & 0.00 & 0.00 & 0.00 & 0.00 & 0.00 & 0.00 & 0.00 & 0.00 & 0.00 & 0.00 \\
GPT4o-mini & 0.00 & 0.00 & 0.00 & 0.00 & 0.00 & 0.00 & 0.00 & 0.00 & 0.00 & 0.00 & 0.00 \\
\rowcolor{bg-tb-light-vision}
GPT-4o-4096 & 0.00 & 0.00 & 0.00 & 0.00 & 0.00 & 0.00 & 0.00 & 0.00 & 0.00 & 0.00 & 0.00 \\
\bottomrule
\end{tabular}%
\end{table*}
\clearpage

\begin{table*}[t!]
\centering
  \fontsize{8.5}{10.5}\selectfont 
  \setlength{\tabcolsep}{6.mm}
\caption{
Results on \textcolor{greenCom}{Image Comprehension Group}, \#I-C-27.
}
\vspace{-2mm}
\label{tab:overall-results-image-comp-I-C-27}
\begin{tabular}{p{4.2cm} cccc}
\toprule
\rowcolor{bg-tb-heavey-vision} & \multicolumn{4}{c}{\bf \#I-C-27}  \\ 

\rowcolor{bg-tb-heavey-vision}  & \multicolumn{4}{c}{\tiny (Object Mat)} \\

\cmidrule(r){2-5}   

\rowcolor{bg-tb-heavey-vision} \multicolumn{1}{c}{\multirow{-3}{*}{\bf  Model}}  &  \#1 $\downarrow$ & \#2 $\downarrow$ & \#3 $\uparrow$ & \#4 $\downarrow$ \\

\midrule\midrule

\multirow{1}{*}{\bf SoTA Specialist} & 18.10 & 16.60 & 16.00 & 28.90 \\

\midrule

GPT4V & $\infty$ & $\infty$ & 0.00 & $\infty$ \\
\rowcolor{bg-tb-light-vision}
GPT4o & $\infty$ & $\infty$ & 0.00 & $\infty$ \\
GPT4o-mini & $\infty$ & $\infty$ & 0.00 & $\infty$ \\
\rowcolor{bg-tb-light-vision}
GPT-4o-4096 & $\infty$ & $\infty$ & 0.00 & $\infty$ \\
\bottomrule
\end{tabular}%
\end{table*}
\clearpage

\begin{table*}[t!]
\centering
  \fontsize{8.5}{10.5}\selectfont 
  \setlength{\tabcolsep}{1.7mm}
\caption{
Results on \textcolor{greenCom}{Image Comprehension Group},\#I-C-28, part A.
}
\vspace{-2mm}
\label{tab:overall-results-image-comp-I-C-28-a}
\begin{tabular}{p{4.2cm} cccccccccccc}
\toprule
\rowcolor{bg-tb-heavey-vision} & \multicolumn{12}{c}{\bf \#I-C-28}  \\ 

\rowcolor{bg-tb-heavey-vision}  & \multicolumn{12}{c}{\tiny (Object Recog)}  \\

\cmidrule(r){2-13}  

\rowcolor{bg-tb-heavey-vision} \multicolumn{1}{c}{\multirow{-3}{*}{\bf  Model}}  &  \#1 $\uparrow$ & \#2 $\uparrow$ & \#3 $\uparrow$ & \#4 $\uparrow$ & \#5 $\uparrow$ & \#6 $\uparrow$ & \#7 $\uparrow$ & \#8 $\uparrow$ & \#9 $\uparrow$ & \#10 $\uparrow$ & \#11 $\uparrow$  & \#12 $\uparrow$ \\

\midrule\midrule

\multirow{1}{*}{\bf SoTA Specialist} & 62.16 & 94.85 & 52.58 & 18.45 & 57.67 & 75.21 & 54.85 & 75.21 & 82.22 & 45.89 & 88.73 & 62.38 \\

\midrule

GPT4V & \textbf{98.24} & 93.22 & \textbf{63.03} & \textbf{48.93} & \textbf{94.50} & \textbf{96.04} & \textbf{84.47} & \textbf{97.57} & \textbf{92.13} & \textbf{72.84} & \textbf{96.41} & \textbf{98.91} \\
\rowcolor{bg-tb-light-vision}
GPT4o & \textbf{98.43} & \textbf{95.53} & \textbf{82.87} & \textbf{66.95} & \textbf{96.10} & \textbf{97.30} & \textbf{95.63} & \textbf{97.14} & \textbf{93.13} & \textbf{77.05} & \textbf{97.50} & \textbf{99.87} \\
GPT4o-mini & \textbf{99.43} & 94.30 & \textbf{78.33} & \textbf{49.36} & \textbf{95.83} & \textbf{94.79} & \textbf{94.41} & \textbf{97.54} & \textbf{92.93} & \textbf{73.26} & \textbf{94.69} & \textbf{91.71} \\
\rowcolor{bg-tb-light-vision}
GPT-4o-4096 & \textbf{98.82} & \textbf{95.39} & \textbf{83.79} & \textbf{65.24} & \textbf{99.00} & \textbf{99.77} & \textbf{94.42} & \textbf{99.79} & \textbf{93.13} & \textbf{80.84} & \textbf{96.72} & \textbf{99.88} \\
\bottomrule
\end{tabular}%
\end{table*}
\clearpage

\begin{table*}[t!]
\centering
  \fontsize{8.5}{10.5}\selectfont 
  \setlength{\tabcolsep}{1.6mm}
\caption{
Results on \textcolor{greenCom}{Image Comprehension Group},\#I-C-28, part B.
}
\vspace{-2mm}
\label{tab:overall-results-image-comp-I-C-28-b}
\begin{tabular}{p{4.2cm} cccccccccc}
\toprule
\rowcolor{bg-tb-heavey-vision} & \multicolumn{10}{c}{\bf \#I-C-28}  \\ 

\rowcolor{bg-tb-heavey-vision}  & \multicolumn{10}{c}{\tiny (Object Recog)}  \\

\cmidrule(r){2-11}  

\rowcolor{bg-tb-heavey-vision} \multicolumn{1}{c}{\multirow{-3}{*}{\bf  Model}}  & \#13 $\uparrow$ & \#14 $\uparrow$ & \#15 $\uparrow$ & \#16 $\uparrow$ & \#17 $\uparrow$ & \#18 $\uparrow$ & \#19 $\uparrow$ & \#20 $\uparrow$ & \#21 $\uparrow$ & \#22 $\uparrow$ \\

\midrule\midrule

\multirow{1}{*}{\bf SoTA Specialist} & 15.73 & 51.30 & 64.00 & 17.67 & 17.30 & 27.00 & 16.70 & 20.67 & 17.80 & 80.00 \\

\midrule

GPT4V & \textbf{19.96} & \textbf{95.33} & \textbf{96.33} & \textbf{88.67} & \textbf{97.33} & \textbf{74.33} & \textbf{88.33} & \textbf{64.00} & \textbf{31.21} & 0.00 \\
\rowcolor{bg-tb-light-vision}
GPT4o & \textbf{39.69} & \textbf{97.33} & \textbf{97.33} & \textbf{93.33} & \textbf{98.67} & \textbf{79.67} & \textbf{90.67} & \textbf{92.33} & \textbf{53.50} & 0.00 \\
GPT4o-mini & \textbf{17.54} & \textbf{94.00} & \textbf{96.67} & \textbf{90.00} & \textbf{98.00} & \textbf{79.67} & \textbf{90.33} & \textbf{89.67} & 10.19 & 0.00 \\
\rowcolor{bg-tb-light-vision}
GPT-4o-4096 & \textbf{34.21} & \textbf{98.00} & \textbf{97.33} & \textbf{92.00} & \textbf{97.33} & \textbf{77.00} & \textbf{88.67} & \textbf{91.67} & \textbf{54.14} & 0.00 \\
\bottomrule
\end{tabular}%
\end{table*}
\clearpage

\begin{table*}[t!]
\centering
  \fontsize{8.5}{10.5}\selectfont 
  \setlength{\tabcolsep}{1.8mm}
\caption{
Results on \textcolor{greenCom}{Image Comprehension Group},\#I-C-28, part C.
}
\vspace{-2mm}
\label{tab:overall-results-image-comp-I-C-28-c}
\begin{tabular}{p{4.2cm} cccccccccc}
\toprule
\rowcolor{bg-tb-heavey-vision} & \multicolumn{10}{c}{\bf \#I-C-28}  \\ 

\rowcolor{bg-tb-heavey-vision}  & \multicolumn{10}{c}{\tiny (Object Recog)}  \\

\cmidrule(r){2-11}  

\rowcolor{bg-tb-heavey-vision} \multicolumn{1}{c}{\multirow{-3}{*}{\bf  Model}}  & \#23 $\uparrow$ & \#24 $\uparrow$ & \#25 $\uparrow$ & \#26 $\uparrow$ & \#27 $\uparrow$ & \#28 $\uparrow$ & \#29 $\uparrow$ & \#30 $\uparrow$ & \#31 $\uparrow$  & \#32 $\uparrow$ \\

\midrule\midrule

\multirow{1}{*}{\bf SoTA Specialist} & 86.40 & 46.80 & 77.90 & 97.33 & 22.46 & 56.50 & 23.50 & 27.40 & 85.31 & 91.60 \\

\midrule

GPT4V & 82.17 & \textbf{49.00} & \textbf{95.98} & \textbf{98.30} & 15.50 & \textbf{57.90} & \textbf{43.60} & \textbf{43.60} & \textbf{89.30} & 53.30 \\
\rowcolor{bg-tb-light-vision}
GPT4o & 24.90 & \textbf{59.00} & \textbf{97.39} & 73.00 & \textbf{36.67} & \textbf{60.10} & \textbf{54.20} & \textbf{63.20} & \textbf{92.40} & 80.00 \\
GPT4o-mini & 66.60 & 44.00 & \textbf{96.98} & \textbf{98.60} & \textbf{34.55} & \textbf{58.20} & \textbf{50.30} & \textbf{67.20} & \textbf{90.30} & 79.10 \\
\rowcolor{bg-tb-light-vision}
GPT-4o-4096 & 48.58 & \textbf{57.80} & \textbf{97.59} & \textbf{98.67} & \textbf{41.65} & \textbf{73.30} & \textbf{38.99} & \textbf{76.39} & 25.20 & 91.00 \\
\bottomrule
\end{tabular}%
\end{table*}
\clearpage

\begin{table*}[t!]
\centering
  \fontsize{8.5}{10.5}\selectfont 
  \setlength{\tabcolsep}{2.2mm}
\caption{
Results on \textcolor{greenCom}{Image Comprehension Group}, from \#I-C-29 to \#I-C-33.
}
\vspace{-2mm}
\label{tab:overall-results-image-comp-I-C-29-30-31-32-33}
\begin{tabular}{p{4.2cm} c c ccc c c}
\toprule
\rowcolor{bg-tb-heavey-vision} & \multicolumn{1}{c}{\bf \#I-C-29} & \multicolumn{1}{c}{\bf \#I-C-30} & \multicolumn{3}{c}{\bf \#I-C-31} & \multicolumn{1}{c}{\bf \#I-C-32} & \multicolumn{1}{c}{\bf \#I-C-33} \\ 

\rowcolor{bg-tb-heavey-vision}  & \multicolumn{1}{c}{\tiny (Panoptic Seg)} &  \multicolumn{1}{c}{\tiny (Pose Recog)} & \multicolumn{3}{c}{\tiny (Rel Reason)}  &  \multicolumn{1}{c}{\tiny (RGBD-Sem Seg)} &   \multicolumn{1}{c}{\tiny (Ripe Recog)}\\

\cmidrule(r){2-2}  \cmidrule(r){3-3}  \cmidrule(r){4-6}  \cmidrule(r){7-7} \cmidrule(r){8-8}   

\rowcolor{bg-tb-heavey-vision} \multicolumn{1}{c}{\multirow{-3}{*}{\bf  Model}}  & \#1 $\uparrow$ & \#1 $\uparrow$ &  \#1 $\uparrow$ & \#2 $\uparrow$ & \#3 $\uparrow$ & \#1 $\uparrow$ & \#1 $\uparrow$ \\

\midrule\midrule

\multirow{1}{*}{\bf SoTA Specialist} & 64.10 & 35.90 & 63.80 & 30.40 & 25.20 & 57.20 & 54.60 \\

\midrule

GPT4V & 0.00 & \textbf{97.98} & \textbf{67.50} & \textbf{31.37} & 21.27 & 0.00 & \textbf{90.40} \\
\rowcolor{bg-tb-light-vision}
GPT4o & 0.00 & \textbf{98.79} & \textbf{72.30} & \textbf{38.24} & 22.35 & 0.00 & \textbf{90.40} \\
GPT4o-mini & 0.00 & \textbf{98.49} & \textbf{64.21} & 30.31 & 20.14 & 0.00 & \textbf{96.40} \\
\rowcolor{bg-tb-light-vision}
GPT-4o-4096 & 0.00 & \textbf{99.60} & 29.20 & 0.00 & 0.00 & 0.00 & \textbf{91.80} \\
\bottomrule
\end{tabular}%
\end{table*}
\clearpage

\begin{table*}[t!]
\centering
  \fontsize{8.5}{10.5}\selectfont 
  \setlength{\tabcolsep}{1.5mm}
\caption{
Results on \textcolor{greenCom}{Image Comprehension Group}, from \#I-C-34 to \#I-C-37.
}
\vspace{-2mm}
\label{tab:overall-results-image-comp-I-C-34-35-36-37}
\begin{tabular}{p{4.2cm} ccccccc c cc c }
\toprule
\rowcolor{bg-tb-heavey-vision} & \multicolumn{7}{c}{\bf \#I-C-34} & \multicolumn{1}{c}{\bf \#I-C-35} & \multicolumn{2}{c}{\bf \#I-C-36} & \multicolumn{1}{c}{\bf \#I-C-37} \\ 

\rowcolor{bg-tb-heavey-vision}  & \multicolumn{7}{c}{\tiny (Safe Det)} &  \multicolumn{1}{c}{\tiny (SG Gen)} & \multicolumn{2}{c}{\tiny (Scene Recog)}  &  \multicolumn{1}{c}{\tiny (Sign Recog)} \\

\cmidrule(r){2-8}  \cmidrule(r){9-9} \cmidrule(r){10-11}  \cmidrule(r){12-12}

\rowcolor{bg-tb-heavey-vision} \multicolumn{1}{c}{\multirow{-3}{*}{\bf  Model}}  & \#1 $\uparrow$ & \#2 $\uparrow$ & \#3 $\uparrow$ & \#4 $\uparrow$ & \#5 $\uparrow$ & \#6 $\uparrow$ & \#7 $\uparrow$  & \#1 $\uparrow$ &  \#1 $\uparrow$ & \#2 $\uparrow$ & \#1 $\uparrow$ \\

\midrule\midrule

\multirow{1}{*}{\bf SoTA Specialist} & 97.40 & 50.40 & 47.90 & 48.40 & 48.40 & 76.30 & 74.10 & 29.60 & 83.00 & 91.20 & 98.00 \\

\midrule

GPT4V & 0.00 & 0.00 & 0.00 & 0.00 & 0.00 & 0.00 & 0.00 & \textbf{31.64} & \textbf{88.20} & 90.00 & 22.22 \\
\rowcolor{bg-tb-light-vision}
GPT4o & 0.00 & 0.00 & 0.00 & 0.00 & 0.00 & 0.00 & 0.00 & \textbf{33.47} & \textbf{91.40} & 91.00 & 35.56 \\
GPT4o-mini & 0.00 & 0.00 & 0.00 & 0.00 & 0.00 & 0.00 & 0.00 & \textbf{30.67} & \textbf{92.00} & 90.80 & 25.00 \\
\rowcolor{bg-tb-light-vision}
GPT-4o-4096 & 0.00 & 0.00 & 0.00 & 0.00 & 0.00 & 0.00 & 0.00 & 0.00 & \textbf{93.20} & 91.00 & 35.93 \\
\bottomrule
\end{tabular}%
\end{table*}
\clearpage

\begin{table*}[t!]
\centering
  \fontsize{8.5}{10.5}\selectfont 
  \setlength{\tabcolsep}{1.5mm}
\caption{
Results on \textcolor{greenCom}{Image Comprehension Group}, from \#I-C-38 to \#I-C-40.
}
\vspace{-2mm}
\label{tab:overall-results-image-comp-I-C-38-39-40}
\begin{tabular}{p{4.2cm} cccccc ccccc c}
\toprule
\rowcolor{bg-tb-heavey-vision} & \multicolumn{6}{c}{\bf \#I-C-38} & \multicolumn{5}{c}{\bf \#I-C-39} & \multicolumn{1}{c}{\bf \#I-C-40}  \\ 

\rowcolor{bg-tb-heavey-vision}  & \multicolumn{6}{c}{\tiny (Vis Rel Inf)} &  \multicolumn{5}{c}{\tiny (Vis-Story Gen)} & \multicolumn{1}{c}{\tiny (Weather Recog)} \\

\cmidrule(r){2-7}  \cmidrule(r){8-12}  \cmidrule(r){13-13} 

\rowcolor{bg-tb-heavey-vision} \multicolumn{1}{c}{\multirow{-3}{*}{\bf  Model}}  & \#1 $\uparrow$ & \#2 $\uparrow$ & \#3 $\uparrow$ & \#4 $\uparrow$ & \#5 $\uparrow$ & \#6 $\uparrow$ & \#1 $\uparrow$ & \#2 $\uparrow$ & \#3 $\uparrow$ & \#4 $\uparrow$ & \#5 $\uparrow$ & \#1 $\uparrow$ \\

\midrule\midrule

\multirow{1}{*}{\bf SoTA Specialist} & 30.50 & 35.50 & 35.80 & 38.40 & 38.40 & 59.02 & 31.90 & 44.60 & 40.20 & 28.90 & 36.50 & 82.02 \\

\midrule

GPT4V & 5.50 & 18.35 & 16.17 & 16.97 & 16.95 & \textbf{61.30} & 20.98 & 19.34 & 17.76 & 16.50 & 15.81 & \textbf{84.84} \\
\rowcolor{bg-tb-light-vision}
GPT4o & 7.40 & 21.11 & 14.80 & 16.40 & 16.46 & \textbf{72.60} & 22.16 & 23.56 & 22.65 & 18.90 & 18.32 & \textbf{87.88} \\
GPT4o-mini & 9.40 & 19.27 & 13.82 & 16.71 & 16.37 & \textbf{60.40} & 20.35 & 19.37 & 18.05 & 14.16 & 16.27 & \textbf{84.65} \\
\rowcolor{bg-tb-light-vision}
GPT-4o-4096 & 6.60 & 22.17 & 14.42 & 16.84 & 27.33 & 18.60 & 23.81 & 25.81 & 23.42 & 19.46 & 19.11 & \textbf{85.66} \\
\bottomrule
\end{tabular}%
\end{table*}
\clearpage

\clearpage
\paragraph{Image Generation Results.}
The complete results of all models on image generation are presented in Table~\ref{tab:overall-results-image-gen-I-G-1-2-3} to Table~\ref{tab:overall-results-image-gen-I-G-15}.

\begin{table*}[!h]
\centering
  \fontsize{8.5}{15}\selectfont 
  \setlength{\tabcolsep}{1.8mm}
\caption{
Results on \textcolor{blueGen}{Image Generation} Group, from \#I-G-1 to \#I-G-3.
}
\vspace{-2mm}
\label{tab:overall-results-image-gen-I-G-1-2-3}
\begin{tabular}{p{4.2cm} c c ccccccc}
\toprule
\rowcolor{bg-tb-heavey-vision} & \multicolumn{1}{c}{\bf \#I-G-1} & \multicolumn{1}{c}{\bf \#I-G-2} & \multicolumn{7}{c}{\bf \#I-G-3}  \\ 

\rowcolor{bg-tb-heavey-vision}  & \multicolumn{1}{c}{\tiny (Edge2Img Gen)} &  \multicolumn{1}{c}{\tiny (EEG2Img Gen)} & \multicolumn{7}{c}{\tiny (Img Denose)}\\

\cmidrule(r){2-2}  \cmidrule(r){3-3}  \cmidrule(r){4-10} 

\rowcolor{bg-tb-heavey-vision} \multicolumn{1}{c}{\multirow{-3}{*}{\bf  Model}}  &  \#1$\downarrow$ & \#1$\uparrow$ & \#1$\uparrow$ & \#2$\uparrow$ & \#3$\uparrow$ & \#4$\uparrow$ & \#5$\uparrow$ & \#6$\downarrow$ & \#7$\uparrow$ \\

\midrule\midrule

\multirow{1}{*}{\bf SoTA Specialist} & 18.70 & 45.40 & 34.56 & 34.55 & 31.45 & 33.92 & 34.60 & 42.05 & 25.28 \\

\midrule

Emu2-32B & 93.52 & 0.00 & 15.77 & 19.91 & 20.31 & 20.57 & 23.46 & 143.93 & 0.00 \\
\rowcolor{bg-tb-light-vision} SEED-LLaMA-13B & 127.10 & 0.00 & 15.29 & 16.85 & 18.45 & 18.72 & 19.71 & 170.65 & 0.00 \\
AnyGPT & 158.21 & 0.00 & 17.69 & 15.93 & 19.46 & 20.04 & 20.45 & 189.73 & 0.00 \\
\rowcolor{bg-tb-light-vision} LaVIT-V2 (7B) & 79.79 & 0.00 & 18.76 & 20.87 & 21.59 & 23.20 & 24.02 & 111.03 & 0.00 \\
Next-GPT-V1.5 & 49.71 & 0.00 & 10.45 & 11.36 & 13.62 & 6.38 & 0.00 & 0.00 & 0.00 \\
\rowcolor{bg-tb-light-vision} Vitron-V1 & 19.78 & 0.00 & 14.59 & 12.68 & 16.82 & 7.56 & 19.55 & 76.86 & 0.00 \\
Show-o & $\infty$ & 0.00 & 0.00 & 0.00 & 0.00 & 0.00 & 0.00 & 0.00 & 0.00 \\

\bottomrule
\end{tabular}%
\end{table*}

\begin{table*}[!h]
\centering
  \fontsize{8.5}{15}\selectfont 
  \setlength{\tabcolsep}{1.mm}
\caption{
Results on \textcolor{blueGen}{Image Generation} Group, from \#I-G-4 to \#I-G-8.
}
\vspace{-2mm}
\label{tab:overall-results-image-gen-I-G-4-5-6-7-8}
%
%
\end{table*}

\clearpage

\begin{table*}[!h]
\centering
  \fontsize{8.5}{15}\selectfont 
  \setlength{\tabcolsep}{0.7mm}
\caption{
Results on \textcolor{blueGen}{Image Generation} Group, from \#I-G-9 to \#I-G-14.
}
\vspace{-2mm}
\label{tab:overall-results-image-gen-I-G-9-10-11-12-13-14}
%
%
\end{table*}

\begin{table*}[!h]
\centering
  \fontsize{8.5}{15}\selectfont 
  \setlength{\tabcolsep}{1.8mm}
\caption{
Results on \textcolor{blueGen}{Image Generation} Group, from \#I-G-15.
}
\vspace{-2mm}
\label{tab:overall-results-image-gen-I-G-15}
%
%
\end{table*}

\clearpage
\subsection{Results of Video-related Tasks}

\paragraph{Video Comprehension Results.}
The complete results of all models on video comprehension are presented from Table~\ref{tab:overall-results-video-com-V-C-1-a} to Table~\ref{tab:overall-results-video-com-V-C-19-20}.

\begin{table*}[!h]
\centering
  \fontsize{8.5}{11}\selectfont 
  \setlength{\tabcolsep}{2.5mm}
\caption{
Results on \textcolor{greenCom}{Video Comprehension} Group, from \#V-C-1 (a).
}
\vspace{-2mm}
\label{tab:overall-results-video-com-V-C-1-a}
%
\end{table*}

\begin{table*}[!h]
\centering
  \fontsize{8.5}{11}\selectfont 
  \setlength{\tabcolsep}{2.8mm}
\caption{
Results on \textcolor{greenCom}{Video Comprehension} Group, from \#V-C-1 (b).
}
\vspace{-2mm}
\label{tab:overall-results-video-com-V-C-1-b}
%
%
\end{table*}

\begin{table*}[!h]
\centering
  \fontsize{8.5}{11}\selectfont 
  \setlength{\tabcolsep}{2.8mm}
\caption{
Results on \textcolor{greenCom}{Video Comprehension} Group, from \#V-C-2 (a).
}
\vspace{-2mm}
\label{tab:overall-results-video-com-V-C-2-a}
%
%
\end{table*}

\begin{table*}[!h]
\centering
  \fontsize{8.5}{11}\selectfont 
  \setlength{\tabcolsep}{2.8mm}
\caption{
Results on \textcolor{greenCom}{Video Comprehension} Group, from \#V-C-2 (b).
}
\vspace{-2mm}
\label{tab:overall-results-video-com-V-C-2-b}
%
%
\end{table*}

\begin{table*}[!h]
\centering
  \fontsize{8.5}{11}\selectfont 
  \setlength{\tabcolsep}{3mm}
\caption{
Results on \textcolor{greenCom}{Video Comprehension} Group, from \#V-C-3 (a).
}
\vspace{-2mm}
\label{tab:overall-results-video-com-V-C-3-a}
%
%
\end{table*}

\begin{table*}[!h]
\centering
  \fontsize{8.5}{11}\selectfont 
  \setlength{\tabcolsep}{3mm}
\caption{
Results on \textcolor{greenCom}{Video Comprehension} Group, from \#V-C-3 (b).
}
\vspace{-2mm}
\label{tab:overall-results-video-com-V-C-3-b}
%
%
\end{table*}

\begin{table*}[!h]
\centering
  \fontsize{8.5}{11}\selectfont 
  \setlength{\tabcolsep}{1.2mm}
\caption{
Results on \textcolor{greenCom}{Video Comprehension} Group, from \#V-C-4 to \#V-C-5.
}
\vspace{-2mm}
\label{tab:overall-results-video-com-V-C-4-5}
%
%
\end{table*}

\begin{table*}[!h]
\centering
  \fontsize{8.5}{11}\selectfont 
  \setlength{\tabcolsep}{1.5mm}
\caption{
Results on \textcolor{greenCom}{Video Comprehension} Group, from \#V-C-6 to \#V-C-9.
}
\vspace{-2mm}
\label{tab:overall-results-video-com-V-C-6-7-8-9}
%
%
\end{table*}

\begin{table*}[!h]
\centering
  \fontsize{8.5}{11}\selectfont 
  \setlength{\tabcolsep}{1.2mm}
\caption{
Results on \textcolor{greenCom}{Video Comprehension} Group, from \#V-C-10 to \#V-C-13.
}
\vspace{-2mm}
\label{tab:overall-results-video-com-V-C-11-12-13-14}
%
%
\end{table*}

\begin{table*}[!h]
\centering
  \fontsize{8.5}{11}\selectfont 
  \setlength{\tabcolsep}{3.3mm}
\caption{
Results on \textcolor{greenCom}{Video Comprehension} Group, from \#V-C-14.
}
\vspace{-2mm}
\label{tab:overall-results-video-com-V-C-14}
%
%
\end{table*}

\begin{table*}[!h]
\centering
  \fontsize{8.5}{11}\selectfont 
  \setlength{\tabcolsep}{1.7mm}
\caption{
Results on \textcolor{greenCom}{Video Comprehension} Group, from \#V-C-15 to \#V-C-16.
}
\vspace{-2mm}
\label{tab:overall-results-video-com-V-C-15-16}
%
%
\end{table*}

\begin{table*}[!h]
\centering
  \fontsize{8.5}{11}\selectfont 
  \setlength{\tabcolsep}{2.6mm}
\caption{
Results on \textcolor{greenCom}{Video Comprehension} Group, from \#V-C-17 to \#V-C-18.
}
\vspace{-2mm}
\label{tab:overall-results-video-com-V-C-17-18}
%
%
\end{table*}

\clearpage
\paragraph{Video Generation Results.}
The complete results of all models on video generation are presented from Table~\ref{tab:overall-results-video-gen-V-G-1} to Table~\ref{tab:overall-results-video-gen-V-G-5-6}.

\begin{table*}[!h]
\centering
  \fontsize{8.5}{11}\selectfont 
  \setlength{\tabcolsep}{1.2mm}
\caption{
Results on \textcolor{blueGen}{Video Generation} Group, from \#V-G-1.
}
\vspace{-2mm}
\label{tab:overall-results-video-gen-V-G-1}
%
%
\end{table*}

\begin{table*}[!h]
\centering
  \fontsize{8.5}{11}\selectfont 
  \setlength{\tabcolsep}{3.2mm}
\caption{
Results on \textcolor{blueGen}{Video Generation} Group, from \#V-G-2 and \#V-G-3.
}
\vspace{-2mm}
\label{tab:overall-results-video-gen-V-G-2-3}
%
%
\end{table*}

\begin{table*}[!h]
\centering
  \fontsize{8.5}{11}\selectfont 
  \setlength{\tabcolsep}{1.8mm}
\caption{
Results on \textcolor{blueGen}{Video Generation} Group, from \#V-G-4.
}
\vspace{-2mm}
\label{tab:overall-results-video-gen-V-G-4}
%
%
\end{table*}

\begin{table*}[!h]
\centering
  \fontsize{8.5}{11}\selectfont 
  \setlength{\tabcolsep}{1.2mm}
\caption{
Results on \textcolor{blueGen}{Video Generation} Group, from \#V-G-5 and \#V-G-6.
}
\vspace{-2mm}
\label{tab:overall-results-video-gen-V-G-5-6}
%
%
\end{table*}

\clearpage

\subsection{Results of Audio-related Tasks}

\paragraph{Audio Comprehension Results.}
The complete results of all models on audio comprehension are presented in Table~\ref{tab:overall-results-audio-comp-A-C-1-2-3-4} and Table~\ref{tab:overall-results-audio-comp-A-C-5-6-7-8-9}.

\begin{table*}[!h]
\centering
  \fontsize{8.5}{12}\selectfont 
  \setlength{\tabcolsep}{1.3mm}
\caption{
Results on \textcolor{greenCom}{Audio Comprehension} Group, from \#A-C-1 to \#A-C-4.
}
\vspace{-2mm}
\label{tab:overall-results-audio-comp-A-C-1-2-3-4}
\begin{tabular}{p{4.2cm} cccccc c cc c cc}
\toprule
\rowcolor{bg-tb-heavey-audio} & \multicolumn{4}{c}{\bf \#A-C-1} & \multicolumn{3}{c}{\bf \#A-C-2} & \multicolumn{1}{c}{\bf \#A-C-3} &  \multicolumn{4}{c}{\bf \#A-C-4}\\ 

\rowcolor{bg-tb-heavey-audio}  & \multicolumn{4}{c}{\tiny (Acnt Analy)} &  \multicolumn{3}{c}{\tiny (Ctnt Analy)} & \multicolumn{1}{c}{\tiny (SpeechEmo Analy)} &  \multicolumn{4}{c}{\tiny (Music Analy)} \\

\cmidrule(r){2-5}  \cmidrule(r){6-8}  \cmidrule(r){9-9}   \cmidrule(r){10-13}

\rowcolor{bg-tb-heavey-audio} \multicolumn{1}{c}{\multirow{-3}{*}{\bf  Model}}  &  \#1$\uparrow$ & \#2$\uparrow$ & \#3$\uparrow$ & \#4$\uparrow$ & \#1$\uparrow$ & \#2$\uparrow$ & \#3$\uparrow$ & \#1$\uparrow$ & \#1$\uparrow$ & \#2$\uparrow$ & \#3$\uparrow$ & \#4$\uparrow$ \\

\midrule\midrule

\multirow{1}{*}{\bf SoTA Specialist} & 75.20&	85.60&	90.33&	97.95&	98.76&	95.29&	43.20&	70.62&	83.50	&69.3&	74.00&	89.20 \\

\midrule

Qwen-Audio-Chat & 56.70 & 75.00 & 55.40 & 40.60 & 93.00 & 76.80 & 36.50 & \textbf{76.80} & 61.20 & 55.20 & 33.00 & 1.40 \\ 
\rowcolor{bg-tb-light-audio} Qwen2-Audio-Instruct & 45.80 & \textbf{99.20} & 53.20 & 92.40 & 94.40 & 82.80 & \textbf{47.20} & 61.40 & 75.20 & 47.80 & 19.40 & 4.80 \\ 
 GAMA & 23.40 & \textbf{86.50} & 32.40 & 85.70 & 80.90 & 77.50 & 34.20 & 68.00 & 63.50 & 63.90 & \textbf{85.40} & 0.00 \\ 
\rowcolor{bg-tb-light-audio}  Pengi & 21.40 & 78.90 & 35.00 & 76.20 & 78.00 & 65.70 & 36.50 & 56.70 & 61.90 & 39.20 & 46.00 & 0.00 \\ 
 SALMONN-7B & 24.70 & 80.50 & 85.00 & 65.10 & 65.40 & 58.30 & 24.10 & 45.56 & 60.50 & 16.20 & 33.70 & 0.00 \\ 
\rowcolor{bg-tb-light-audio} SALMONN-13B & 26.04 & 84.20 & \textbf{93.50} & 67.80 & 68.70 & 68.90 & 31.40 & 67.80 & 63.80 & 19.40 & 34.60 & 0.00 \\ 
 WavLLM & 37.45 & 70.15 & 90.00 & 60.20 & 35.60 & 63.10 & 24.50 & \textbf{71.20} & 56.80 & 49.80 & 12.50 & 1.20 \\ 
\rowcolor{bg-tb-light-audio}  NExT-GPT-V1.5 & 12.50 & 55.40 & 40.50 & 64.50 & 30.50 & 36.80 & 20.10 & 65.80 & 56.80 & 43.70 & 5.50 & 0.80 \\ 
 PandaGPT (13B) & 6.50 & 58.30 & 57.30 & 45.10 & 24.50 & 26.40 & 9.80 & 45.20 & 32.40 & 45.60 & 5.40 & 0.50 \\ 
\rowcolor{bg-tb-light-audio}  ImageBind-LLM & 10.50 & 65.00 & 25.40 & 50.10 & 40.70 & 38.50 & 18.50 & 56.80 & 42.30 & 25.70 & 10.40 & 0.00 \\ 
 ModaVerse-7b-v0 & 10.50 & 50.40 & 35.20 & 40.30 & 18.70 & 20.10 & 10.30 & 32.80 & 24.00 & 32.60 & 4.20 & 0.00 \\ 
\rowcolor{bg-tb-light-audio}  Any-GPT & 34.60 & 62.10 & 15.00 & 66.30 & 37.40 & 46.10 & 12.90 & 63.40 & 68.40 & \textbf{78.90} & 45.00 & 0.00 \\ 
 Unified-io-2-XXL & 22.50 & 45.90 & 13.50 & 38.70 & 34.50 & 32.50 & 15.80 & 56.10 & 56.70 & 36.80 & 20.10 & 0.70 \\

\bottomrule
\end{tabular}%
\end{table*}

\begin{table*}[!h]
\centering
  \fontsize{8.5}{12}\selectfont 
  \setlength{\tabcolsep}{1.6mm}
\caption{
Results on \textcolor{greenCom}{Audio Comprehension} Group, from \#A-C-5 to \#A-C-9.
}
\vspace{-2mm}
\label{tab:overall-results-audio-comp-A-C-5-6-7-8-9}
\begin{tabular}{p{4.2cm} cccccc c cc c cc}
\toprule
\rowcolor{bg-tb-heavey-audio} & \multicolumn{3}{c}{\bf \#A-C-5} & \multicolumn{2}{c}{\bf \#A-C-6} & \multicolumn{2}{c}{\bf \#A-C-7} &  \multicolumn{2}{c}{\bf \#A-C-8} & \multicolumn{3}{c}{\bf \#A-C-9} \\ 

\rowcolor{bg-tb-heavey-audio}  & \multicolumn{3}{c}{\tiny (Aud-Tech Analy)} &  \multicolumn{2}{c}{\tiny (Aud Analy)} & \multicolumn{2}{c}{\tiny (Aud QA)} &  \multicolumn{2}{c}{\tiny (Animal-Sound Det)
} &  \multicolumn{3}{c}{\tiny (Envir-Sound Det)} \\

\cmidrule(r){2-4}  \cmidrule(r){5-6}  \cmidrule(r){7-8}   \cmidrule(r){9-10}  \cmidrule(r){11-13}

\rowcolor{bg-tb-heavey-audio} \multicolumn{1}{c}{\multirow{-3}{*}{\bf  Model}}  &  \#1$\uparrow$ & \#2$\uparrow$ & \#3$\uparrow$ & \#1$\uparrow$ & \#2$\uparrow$ & \#1$\uparrow$ & \#2$\uparrow$ & \#1$\uparrow$ & \#2$\uparrow$ & \#1$\uparrow$ & \#2$\uparrow$ & \#3$\uparrow$ \\

\midrule\midrule

\multirow{1}{*}{\bf SoTA Specialist} & 69.40 & 80.30 & 65.90 & 70.50 & 55.30 & 38.90 & 78.50 & 77.60 & 78.20 & 76.40 & 86.70 & 71.10 \\

\midrule

Qwen-Audio-Chat & 63.40 & 58.40 & 21.34 & 19.32 & 20.25 & 31.27 & \textbf{81.60} & \textbf{86.30} & \textbf{84.00} & \bf 86.11 & \bf 92.60 & 56.80 \\ 
\rowcolor{bg-tb-light-audio} Qwen2-Audio-Instruct & 65.70 & 61.40 & 10.36 & 16.82 & 10.08 & 34.56 & \textbf{88.80} & \textbf{87.50} & 70.40 & \bf 71.58 & \bf 86.80 & 45.60 \\ 
GAMA & 23.50 & 25.40 & 6.40 & 25.40 & 28.50 & 23.60 & 74.10 & \textbf{89.60} & \textbf{81.50} & {72.30} & 80.50 & 32.60 \\ 
\rowcolor{bg-tb-light-audio} Pengi & 38.40 & 15.70 & 5.20 & 16.80 & 22.30 & 15.40 & 70.50 & \textbf{83.60} & 71.20 & 68.40 & 78.40 & 36.70 \\ 
SALMONN-7B & 31.70 & 17.90 & 5.10 & 15.40 & 15.60 & 17.80 & 60.40 & {75.80} & 65.00 & 57.80 & 66.40 & 24.60 \\ 
\rowcolor{bg-tb-light-audio} SALMONN-13B & 32.30 & 35.40 & 6.30 & 21.00 & 17.72 & 19.00 & 68.90 & \textbf{79.60} & 73.50 & 60.60 & 75.90 & 33.50 \\ 
WavLLM & 53.60 & 31.40 & 8.90 & 29.70 & 23.40 & 13.50 & 78.00 & \textbf{86.40} & 36.40 & 74.30 & 77.80 & 41.60 \\ 
\rowcolor{bg-tb-light-audio} NExT-GPT-V1.5 & 24.60 & 16.50 & 2.30 & 16.80 & 34.50 & 25.60 & 70.30 & 76.90 & 63.50 & 69.50 & {80.90} & 57.90 \\ 
PandaGPT-13B & 21.50 & 3.60 & 0.30 & 10.50 & 30.50 & 15.30 & 69.20 & 52.90 & 56.70 & 61.50 & 80.10 & 55.90 \\ 
\rowcolor{bg-tb-light-audio} ImageBind-LLM & 26.70 & 14.30 & 2.00 & 14.30 & 22.50 & 23.40 & 67.30 & 63.40 & 59.20 & 36.50 & 72.60 & 57.10 \\ 
ModaVerse-7b-v0 & 14.20 & 4.10 & 1.50 & 2.50 & 15.30 & 13.80 & 56.30 & 46.80 & 51.60 & 58.70 & 75.70 & 46.00 \\ 
\rowcolor{bg-tb-light-audio} Any-GPT & 28.40 & 10.80 & 9.60 & 36.10 & 36.70 & 28.90 & 76.40 & 62.10 & 73.80 & 45.20 & 52.30 & 36.40 \\ 
Unified-io-2-XXL & 33.10  & 5.40 & 7.90 & 41.30 & 35.40 & 12.30 & 65.10 & 57.30 & 69.70 & 56.80 & 79.40 & 45.70 \\

\bottomrule
\end{tabular}%
\end{table*}

\clearpage
\paragraph{Audio Generation Results.}
The complete results of all models on audio generation are presented in Table~\ref{tab:overall-results-audio-gen-A-G-1-2-3-4-5-6-7} and Table~\ref{tab:overall-results-audio-gen-A-G-8-9-10-11}.

\begin{table*}[!h]
\centering
  \fontsize{8.5}{13}\selectfont 
  \setlength{\tabcolsep}{1.2mm}
\caption{
Results on \textcolor{blueGen}{Audio Generation} Group, from \#A-G-1 to \#A-G-7.
}
\vspace{-2mm}
\label{tab:overall-results-audio-gen-A-G-1-2-3-4-5-6-7}
\begin{tabular}{p{4.2cm} cccccc c cc c}
\toprule
\rowcolor{bg-tb-heavey-audio} & \multicolumn{1}{c}{\bf \#A-G-1} & \multicolumn{1}{c}{\bf \#A-G-2} & \multicolumn{2}{c}{\bf \#A-G-3} &  \multicolumn{2}{c}{\bf \#A-G-4} & \multicolumn{2}{c}{\bf \#A-G-5} & \multicolumn{1}{c}{\bf \#A-G-6} & \multicolumn{1}{c}{\bf \#A-G-7}\\ 

\rowcolor{bg-tb-heavey-audio}  & \multicolumn{1}{c}{\tiny (Audio Edit)} &  \multicolumn{1}{c}{\tiny (Dialog Gen)} & \multicolumn{2}{c}{\tiny (EmoSpeech Gen)} &  \multicolumn{2}{c}{\tiny (TTS)} & \multicolumn{2}{c}{\tiny (Txt2Aud)} & \multicolumn{1}{c}{\tiny (Img2Aud Gen)} &  \multicolumn{1}{c}{\tiny (V2A)}  \\

\cmidrule(r){2-2}  \cmidrule(r){3-3}  \cmidrule(r){4-5}   \cmidrule(r){6-7}  \cmidrule(r){8-9} \cmidrule(r){10-10} \cmidrule(r){11-11}

\rowcolor{bg-tb-heavey-audio} \multicolumn{1}{c}{\multirow{-3}{*}{\bf  Model}}  &  \#1$\uparrow$ & \#1$\uparrow$ & \#1$\downarrow$ & \#2$\uparrow$ & \#1$\downarrow$ & \#2$\uparrow$ & \#1$\uparrow$ & \#2$\uparrow$ & \#1$\uparrow$ & \#1$\downarrow$ \\

\midrule\midrule

\multirow{1}{*}{\bf SoTA Specialist} & 31.50 & 3.82 & 4.12 & 3.15 & 5.60 & 3.76 & 47.04 & 36.03 & 51.40 & 11.52 \\

\midrule

LLaMA-Omni & 0.00 & 3.01 & 5.61 & 2.36 & 20.15 & 0.00 & 0.00 & 0.00 & 0.00 & $\infty$ \\ 
\rowcolor{bg-tb-light-audio} Unified-io 2 & 18.36 & 2.03 & 7.86 & 2.35 & {78.50} & 2.54 & 21.78 & 11.03 & 24.31 & 16.97 \\ 
Any-GPT & 23.50 & 3.24 & 6.98 & 2.15 & 65.80 & 1.35 & 16.52 & 10.24 & 14.05 & {27.49} \\ 
\rowcolor{bg-tb-light-audio} Next-GPT-V1.5 & 13.60 & 1.15 & 6.78 & 1.35 & {100.00} & 1.02 & \textbf{53.68} & 15.34 & 1.35 & 12.36 \\ 
AudioGPT & 0.50 & 1.32 & 5.32 & \textbf{3.89} & 45.20 & 0.00 & 48.63 & 10.32 & 0.00 & $\infty$ \\ 
\rowcolor{bg-tb-light-audio} SpeechGPT & 0.10 & 2.79 & 5.74 & 3.14 & 63.70 & 0.00 & 0.00 & 0.00 & 0.00 & $\infty$ \\ 
ModaVerse & 12.30 & 1.15 & {7.52} & 1.05 & {100.00} & 1.00 & \textbf{50.33} & 7.65 & 1.05 & {16.45} \\

\bottomrule
\end{tabular}%
\end{table*}

\begin{table*}[!h]
\centering
  \fontsize{8.5}{13}\selectfont 
  \setlength{\tabcolsep}{1.3mm}
\caption{
Results on \textcolor{blueGen}{Audio Generation} Group, from \#A-G-8 to \#A-G-11.
}
\vspace{-2mm}
\label{tab:overall-results-audio-gen-A-G-8-9-10-11}
\begin{tabular}{p{4.2cm} c ccc cccc cc }
\toprule
\rowcolor{bg-tb-heavey-audio} & \multicolumn{1}{c}{\bf \#A-G-8} & \multicolumn{3}{c}{\bf \#A-G-9} & \multicolumn{4}{c}{\bf \#A-G-10} &  \multicolumn{2}{c}{\bf \#A-G-11} \\ 

\rowcolor{bg-tb-heavey-audio}  & \multicolumn{1}{c}{\tiny (Style Trans)} &  \multicolumn{3}{c}{\tiny (Speech Trans)} & \multicolumn{4}{c}{\tiny (Music Gen)} &  \multicolumn{2}{c}{\tiny (Music Trans)} \\

\cmidrule(r){2-2}  \cmidrule(r){3-5}  \cmidrule(r){6-9}  \cmidrule(r){10-11} 

\rowcolor{bg-tb-heavey-audio} \multicolumn{1}{c}{\multirow{-3}{*}{\bf  Model}}  &  \#1$\downarrow$ & \#1$\downarrow$ & \#2$\downarrow$ & \#3$\downarrow$ & \#1$\uparrow$ & \#2$\uparrow$ & \#3$\uparrow$ & \#4$\downarrow$ & \#1$\uparrow$ & \#2$\uparrow$ \\

\midrule\midrule

\multirow{1}{*}{\bf SoTA Specialist} & 6.80 & 7.10 & 7.70 & 10.20 & 25.80 & 59.01 & 2.85 & 3.87 & 28.16 & 12.50 \\

\midrule

LLaMA-Omni & 45.30 & 89.36 & 93.56 & 100.00 & 0.00 & 0.00 & 0.00 & $\infty$ & 0.00 & 0.00 \\ 
\rowcolor{bg-tb-light-audio} Unified-io 2 & 86.23 & 93.21 & {100.00} & 90.36 & 0.00 & 0.00 & 0.00 & $\infty$ & {3.15} & {1.32} \\ 
Any-GPT & 45.36 & 56.89 & 95.45 & 99.34 & 0.00 & 0.00 & 0.00 & $\infty$ & {1.28} & 3.65 \\ 
\rowcolor{bg-tb-light-audio} Next-GPT & 96.70 & 99.30 & 98.40 & {100.00} & 0.00 & 0.00 & 0.00 & $\infty$ & {8.76} & {6.78} \\ 
AudioGPT & 46.30 & {45.68} & 94.25 & {100.00} & 0.00 & 0.00 & 0.00 & $\infty$ & 0.00 & 0.00 \\ 
\rowcolor{bg-tb-light-audio} SpeechGPT & {30.24} & 57.96 & {98.67} & {100.00} & 0.00 & 0.00 & 0.00 & $\infty$ & 0.00 & 0.00 \\ 
ModaVerse & {100.00} & {100.00} & {100.00} & {100.00} & 0.00 & 0.00 & 0.00 & $\infty$ & {3.59} & {4.75} \\

\bottomrule
\end{tabular}%
\end{table*}

\clearpage

\subsection{Results of 3D-related Tasks}

\paragraph{3D Comprehension Results.}
The complete results of all models on 3D comprehension are presented in Table~\ref{tab:overall-results-3d-comp-D-C-1-2-3-4} to Table~\ref{tab:overall-results-3d-comp-D-C-11-12-13}.

\begin{table*}[!h]
\centering
  \fontsize{8.5}{13}\selectfont 
  \setlength{\tabcolsep}{1.3mm}
\caption{
Results on \textcolor{greenCom}{3D Comprehension} Group, from \#D-C-1 to \#D-C-4.
}
\vspace{-2mm}
\label{tab:overall-results-3d-comp-D-C-1-2-3-4}
\begin{tabular}{p{4.2cm} ccccc cc cc c}
\toprule
\rowcolor{bg-tb-heavey-3D} & \multicolumn{5}{c}{\bf \#D-C-1} & \multicolumn{2}{c}{\bf \#D-C-2} & \multicolumn{2}{c}{\bf \#D-C-3} & \multicolumn{1}{c}{\bf \#D-C-4}  \\ 

\rowcolor{bg-tb-heavey-3D}  & \multicolumn{5}{c}{\tiny (3D-Human Cls)} &  \multicolumn{2}{c}{\tiny (3D-Struct Cls)} & \multicolumn{2}{c}{\tiny (Tech Cls)} &  \multicolumn{1}{c}{\tiny (Indoor-Scene Seg)} \\

\cmidrule(r){2-6}  \cmidrule(r){7-8}  \cmidrule(r){9-10}   \cmidrule(r){11-11} 

\rowcolor{bg-tb-heavey-3D} \multicolumn{1}{c}{\multirow{-3}{*}{\bf  Model}}  &  \#1$\uparrow$ & \#2$\uparrow$ & \#3$\uparrow$ & \#4$\uparrow$ & \#5$\uparrow$ & \#1$\uparrow$ & \#2$\uparrow$ & \#1$\uparrow$ & \#2$\uparrow$ & \#1$\uparrow$ \\

\midrule\midrule

\multirow{1}{*}{\bf SoTA Specialist} & 91.18 & 96.25 & 94.29 & 99.50 & 100.00 & 99.20 & 97.50 & 95.56 & 100.00 & 78.50 \\

\midrule

3D-VisTA & 0.00 & 0.00 & 0.00 & 0.00 & 0.00 & 0.00 & 0.00 & 0.00 & 0.00 & 0.00 \\ 
\rowcolor{bg-tb-light-3D} PointLLM-7B & 6.36 & 59.28 & 45.55 & 48.12 & 71.50 & 0.00 & 15.00 & 65.71 & 80.00 & 0.00 \\ 
PointLLM-13B & 5.90 & 56.42 & 52.77 & 49.87 & 79.00 & 5.00 & 15.00 & 69.28 & 87.00 & 0.00 \\ 
\rowcolor{bg-tb-light-3D} 3D-LLM & 0.00 & 0.00 & 0.00 & 0.00 & 0.00 & 0.00 & 0.00 & 0.00 & 0.00 & 0.00 \\ 
AvatarGPT & 0.00 & 0.00 & 0.00 & 0.00 & 0.00 & 0.00 & 0.00 & 0.00 & 0.00 & 0.00 \\

\bottomrule
\end{tabular}%
\end{table*}

\begin{table*}[!h]
\centering
  \fontsize{8.5}{13}\selectfont 
  \setlength{\tabcolsep}{1.mm}
\caption{
Results on \textcolor{greenCom}{3D Comprehension} Group, from \#D-C-5 to \#D-C-9.
}
\vspace{-2mm}
\label{tab:overall-results-3d-comp-D-C-5-6-7-8-9}
\begin{tabular}{p{4.2cm} cccc cccc c c c}
\toprule
\rowcolor{bg-tb-heavey-3D} & \multicolumn{1}{c}{\bf \#D-C-5} & \multicolumn{1}{c}{\bf \#D-C-6} & \multicolumn{1}{c}{\bf \#D-C-7} &  \multicolumn{6}{c}{\bf \#D-C-8} & \multicolumn{1}{c}{\bf \#D-C-9}\\ 

\rowcolor{bg-tb-heavey-3D}  & \multicolumn{1}{c}{\tiny (Outdoor-Scene Seg)} &  \multicolumn{1}{c}{\tiny  (Indoor-Inst Seg)} & \multicolumn{1}{c}{\tiny (Pose Est)} &  \multicolumn{6}{c}{\tiny (Part Seg)
} & \multicolumn{1}{c}{\tiny (3D Track)} \\

\cmidrule(r){2-2}  \cmidrule(r){3-3}  \cmidrule(r){4-4}   \cmidrule(r){5-10} \cmidrule(r){11-11}

\rowcolor{bg-tb-heavey-3D} \multicolumn{1}{c}{\multirow{-3}{*}{\bf  Model}}  &  \#1$\uparrow$ & \#1$\uparrow$ & \#1$\downarrow$ & \#1$\uparrow$ & \#2$\uparrow$ & \#3$\uparrow$ & \#4$\uparrow$ & \#5$\uparrow$ & \#6$\uparrow$ & \#1$\uparrow$ \\

\midrule\midrule

\multirow{1}{*}{\bf SoTA Specialist} &  70.02 & 81.20 & 55.00 & 87.31 & 93.40 & 88.62 & 89.38 & 81.44 & 89.52 & 75.20 \\

\midrule

3D-VisTA & 0.00 & 0.00 & $\infty$ & 0.00 & 0.00 & 0.00 & 0.00 & 0.00 & 0.00 & 0.00 \\ 
\rowcolor{bg-tb-light-3D} PointLLM-7B & 0.00 & 0.00 & $\infty$ & 0.00 & 0.00 & 0.00 & 0.00 & 0.00 & 0.00 & 0.00 \\ 
PointLLM-13B & 0.00 & 0.00 & $\infty$ & 0.00 & 0.00 & 0.00 & 0.00 & 0.00 & 0.00 & 0.00 \\ 
\rowcolor{bg-tb-light-3D} 3D-LLM & 0.00 & 0.00 & $\infty$ & 0.00 & 0.00 & 0.00 & 0.00 & 0.00 & 0.00 & 0.00 \\ 

AvatarGPT & 0.00 & 0.00 & $\infty$ & 0.00 & 0.00 & 0.00 & 0.00 & 0.00 & 0.00 & 0.00 \\

\bottomrule
\end{tabular}%
\end{table*}

\begin{table*}[!h]
\centering
  \fontsize{8.5}{13}\selectfont 
  \setlength{\tabcolsep}{1.1mm}
\caption{
Results on \textcolor{greenCom}{3D Comprehension} Group, from \#D-C-10 to \#D-C-13.
}
\vspace{-2mm}
\label{tab:overall-results-3d-comp-D-C-11-12-13}
\begin{tabular}{p{4.2cm} c c ccccccc c}
\toprule
\rowcolor{bg-tb-heavey-3D} & \multicolumn{1}{c}{\bf \#D-C-10} & \multicolumn{1}{c}{\bf \#D-C-11} & \multicolumn{7}{c}{\bf \#D-C-12} & \multicolumn{1}{c}{\bf \#D-C-13}\\ 

\rowcolor{bg-tb-heavey-3D} & \multicolumn{1}{c}{\tiny (3D-Geo Analy)} & \multicolumn{1}{c}{\tiny (3D Det)} &  \multicolumn{7}{c}{\tiny (3D QA)} & \multicolumn{1}{c}{\tiny (3D-Motion Analy)} \\

\cmidrule(r){2-2}  \cmidrule(r){3-3}  \cmidrule(r){4-10}   \cmidrule(r){11-11}

\rowcolor{bg-tb-heavey-3D} \multicolumn{1}{c}{\multirow{-3}{*}{\bf  Model}}  &  \#1$\downarrow$ & \#1$\uparrow$ & \#1$\uparrow$ & \#2$\uparrow$ & \#1$\uparrow$ & \#2$\uparrow$ & \#1$\uparrow$ & \#2$\uparrow$ & \#1$\uparrow$ & \#1$\uparrow$ \\

\midrule\midrule

\multirow{1}{*}{\bf SoTA Specialist} & 9.96 & 68.52 & 12.40 & 35.60 & 67.20 & 48.50 & 71.40 & 49.10 & 45.80 & 22.30 \\

\midrule

3D-VisTA & $\infty$ & 0.00 & \textbf{16.00} & 34.80 & 63.30 & 45.40 & 69.80 & 47.20 & \textbf{48.10} & 0.00 \\ 
\rowcolor{bg-tb-light-3D} PointLLM-7B & $\infty$ & 0.00 & 0.00 & 0.00 & 0.00 & 0.00 & 0.00 & 0.00 & 0.00 & 0.00 \\ 
PointLLM-13B & $\infty$ & 0.00 & 0.00 & 0.00 & 0.00 & 0.00 & 0.00 & 0.00 & 0.00 & 0.00 \\ 
\rowcolor{bg-tb-light-3D} 3D-LLM & $\infty$ & 0.00 & {12.00} & \textbf{36.50} & 65.60 & 47.20 & 68.80 & 48.00 & 46.30 & 0.00 \\ 

AvatarGPT & $\infty$ & 0.00 &  0.00 & 0.00 & 0.00 & 0.00 & 0.00 & 0.00 & 0.00 & 12.70\\

\bottomrule
\end{tabular}%
\end{table*}

\clearpage
\paragraph{3D Generation Results.}
The complete results of all models on 3D generation are presented in Table~\ref{tab:overall-results-3D-gen-A-G-1-2-3-4} and Table~\ref{tab:overall-results-3D-gen-D-G-5-6-7-8-9}.

\begin{table*}[!h]
\centering
  \fontsize{8.5}{14}\selectfont 
  \setlength{\tabcolsep}{1.2mm}
\caption{
Results on \textcolor{blueGen}{3D Generation} Group, from \#D-G-1 to \#D-G-4.
}
\vspace{-2mm}
\label{tab:overall-results-3D-gen-A-G-1-2-3-4}
\begin{tabular}{p{4.2cm} c cc cccc cccc}
\toprule
\rowcolor{bg-tb-heavey-3D} & \multicolumn{1}{c}{\bf \#D-G-1} & \multicolumn{2}{c}{\bf \#D-G-2} & \multicolumn{4}{c}{\bf \#D-G-3} & \multicolumn{4}{c}{\bf \#D-G-4}  \\ 

\rowcolor{bg-tb-heavey-3D}  & \multicolumn{1}{c}{\tiny (PC Complt)} &  \multicolumn{2}{c}{\tiny (PC2M Recon)} & \multicolumn{4}{c}{\tiny (Txt2PC Gen)} &  \multicolumn{4}{c}{\tiny (Txt2M Gen)} \\

\cmidrule(r){2-2}  \cmidrule(r){3-4}  \cmidrule(r){5-8}   \cmidrule(r){9-12} 

\rowcolor{bg-tb-heavey-3D} \multicolumn{1}{c}{\multirow{-3}{*}{\bf  Model}}  &  \#1$\downarrow$ & \#1$\downarrow$ & \#2$\downarrow$ & \#1$\uparrow$ & \#2$\uparrow$ & \#3$\uparrow$ & \#4$\uparrow$ & \#1$\uparrow$ & \#2$\uparrow$ & \#3$\uparrow$ & \#4$\uparrow$\\

\midrule\midrule

\multirow{1}{*}{\bf SoTA Specialist} & 0.22 & 9.32E-05 & 4.93E-05 & 24.94 & 25.10 & 24.07 & 23.56 & 26.42 & 25.22 & 25.93 & 25.18 \\

\midrule

MotionGPT-1 & $\infty$ & $\infty$ & $\infty$ & 0.00 & 0.00 & 0.00 & 0.00 & 0.00 & 0.00 & 0.00 & 0.00 \\ 
\rowcolor{bg-tb-light-3D} MotionGPT-2 & $\infty$ & $\infty$ & $\infty$ & 0.00 & 0.00 & 0.00 & 0.00 & 0.00 & 0.00 & 0.00 & 0.00 \\ 
LLaMA-Mesh & $\infty$ & $\infty$ & $\infty$ & 0.00 & 0.00 & 0.00 & 0.00 & 20.06 & 14.43 & 18.06 & 17.65 \\

\bottomrule
\end{tabular}%
\end{table*}

\begin{table*}[!h]
\centering
  \fontsize{8.5}{14}\selectfont 
  \setlength{\tabcolsep}{1mm}
\caption{
Results on \textcolor{blueGen}{3D Generation} Group, from \#D-G-5 to \#D-G-9.
}
\vspace{-2mm}
\label{tab:overall-results-3D-gen-D-G-5-6-7-8-9}
\begin{tabular}{p{4cm} cccc cccc c c c}
\toprule
\rowcolor{bg-tb-heavey-3D} & \multicolumn{4}{c}{\bf \#D-G-5} & \multicolumn{4}{c}{\bf \#D-G-6} & \multicolumn{1}{c}{\bf \#D-G-7} &  \multicolumn{1}{c}{\bf \#D-G-8} & \multicolumn{1}{c}{\bf \#D-G-9} \\ 

\rowcolor{bg-tb-heavey-3D}  & \multicolumn{4}{c}{\tiny (Img2PC Gen)} &  \multicolumn{4}{c}{\tiny (Img2M Gen)} & \multicolumn{1}{c}{\tiny (RGBD2PC Recon)} &  \multicolumn{1}{c}{\tiny (RGBD2Mesh Recon)} & \multicolumn{1}{c}{\tiny (Txt2Motion Gen)}  \\

\cmidrule(r){2-5}  \cmidrule(r){6-9}  \cmidrule(r){10-10} \cmidrule(r){11-11} \cmidrule(r){12-12}

\rowcolor{bg-tb-heavey-3D} \multicolumn{1}{c}{\multirow{-3}{*}{\bf  Model}}  &  \#1$\uparrow$ & \#2$\uparrow$ & \#3$\uparrow$ & \#4$\uparrow$ & \#1$\uparrow$ & \#2$\uparrow$ & \#3$\uparrow$ & \#4$\uparrow$ & \#1$\downarrow$ & \#1$\downarrow$ & \#1$\downarrow$ \\

\midrule\midrule

\multirow{1}{*}{\bf SoTA Specialist} & 77.06 & 78.27 & 78.87 & 78.04 & 83.30 & 82.88 & 84.43 & 83.96 & 6540.02 & 6540.02 & 0.23 \\

\midrule

MotionGPT-1 & 0.00 & 0.00 & 0.00 & 0.00 & 0.00 & 0.00 & 0.00 & 0.00 & $\infty$ & $\infty$ & 0.51 \\ 
\rowcolor{bg-tb-light-3D} MotionGPT-2 & 0.00 & 0.00 & 0.00 & 0.00 & 0.00 & 0.00 & 0.00 & 0.00 & $\infty$ & $\infty$ & 0.60 \\ 
LLaMA-Mesh & 0.00 & 0.00 & 0.00 & 0.00 & 0.00 & 0.00 & 0.00 & 0.00 & $\infty$ & $\infty$ & $\infty$ \\

\bottomrule
\end{tabular}%
\end{table*}

\clearpage
\subsection{Results of NLP Tasks}

All the results of all generalists on NLP tasks are shown in Table~\ref{tab:overall-results-nlp-L-1}, Table~\ref{tab:overall-results-nlp-L-2}, Table~\ref{tab:overall-results-nlp-L-3-4}, Table~\ref{tab:overall-results-nlp-L-5-6-7}, Table~\ref{tab:overall-results-nlp-L-8-9-10-11}, Table~\ref{tab:overall-results-nlp-L-12-13-14-15-16}, Table~\ref{tab:overall-results-nlp-L-17-a}, Table~\ref{tab:overall-results-nlp-L-17-b}, Table~\ref{tab:overall-results-nlp-L-18}, Table~\ref{tab:overall-results-nlp-L-19}, and Table~\ref{tab:overall-results-nlp-L-20-21-22}.

{\centering
  \fontsize{8.5}{10.5}\selectfont 
  \setlength{\tabcolsep}{2.mm}

\captionof{table}{
Results on \textcolor{greenCom}{NLP Group}, \#L-1.
}
\vspace{-3mm}

\label{tab:overall-results-nlp-L-1}


}

\clearpage

{\centering
  \fontsize{8.5}{10.5}\selectfont 
  \setlength{\tabcolsep}{2.mm}
\captionof{table}{
Results on \textcolor{greenCom}{NLP Group}, \#L-2.
}
\vspace{-3mm}
\label{tab:overall-results-nlp-L-2}
%
}
\clearpage

{\centering
  \fontsize{8.5}{10.5}\selectfont 
  \setlength{\tabcolsep}{1.2mm}
\captionof{table}{
Results on \textcolor{greenCom}{NLP Group}, from \#L-3 to \#L-4.
}
\vspace{-2mm}
\label{tab:overall-results-nlp-L-3-4}
%
}
\clearpage

{\centering
  \fontsize{8.5}{10.5}\selectfont 
  \setlength{\tabcolsep}{2.mm}
\captionof{table}{
Results on \textcolor{greenCom}{NLP Group}, from \#L-5 to \#L-7.
}
\vspace{-2mm}
\label{tab:overall-results-nlp-L-5-6-7}
%
}
\clearpage

{\centering
  \fontsize{8.5}{10.5}\selectfont 
  \setlength{\tabcolsep}{1.3mm}
\captionof{table}{
Results on \textcolor{greenCom}{NLP Group}, from \#L-8 to \#L-11.
}
\vspace{-2mm}
\label{tab:overall-results-nlp-L-8-9-10-11}
%
}
\clearpage

{\centering
  \fontsize{8.5}{10.5}\selectfont 
  \setlength{\tabcolsep}{1.1mm}
\captionof{table}{
Results on \textcolor{greenCom}{NLP Group}, from \#L-12 to \#L-16.
}
\vspace{-2mm}
\label{tab:overall-results-nlp-L-12-13-14-15-16}
%
}
\clearpage

{\centering
  \fontsize{8.5}{10.5}\selectfont 
  \setlength{\tabcolsep}{2.4mm}
\captionof{table}{
Results on \textcolor{greenCom}{NLP Group}, \#L-17, part A.
}
\vspace{-2mm}
\label{tab:overall-results-nlp-L-17-a}
%
}
\clearpage

{\centering
  \fontsize{8.5}{10.5}\selectfont 
  \setlength{\tabcolsep}{2.4mm}
\captionof{table}{
Results on \textcolor{greenCom}{NLP Group}, \#L-17, part B.
}
\vspace{-2mm}
\label{tab:overall-results-nlp-L-17-b}
%
}
\clearpage

{\centering
  \fontsize{8.5}{10.5}\selectfont 
  \setlength{\tabcolsep}{1.6mm}
\captionof{table}{
Results on \textcolor{greenCom}{NLP Group}, \#L-18.
}
\vspace{-2mm}
\label{tab:overall-results-nlp-L-18}
%
}
\clearpage

{\centering
  \fontsize{8.5}{10.5}\selectfont 
  \setlength{\tabcolsep}{2.6mm}
\captionof{table}{
Results on \textcolor{greenCom}{NLP Group}, \#L-19.
}
\vspace{-2mm}
\label{tab:overall-results-nlp-L-19}
%
}
\clearpage

{\centering
  \fontsize{8.5}{10.5}\selectfont 
  \setlength{\tabcolsep}{2.6mm}
\captionof{table}{
Results on \textcolor{greenCom}{NLP Group}, from \#L-20 to \#L-22.
}
\vspace{-2mm}
\label{tab:overall-results-nlp-L-20-21-22}
%
}
\clearpage

\section{Statement}

\subsection{Ethical Statement}

This work adheres to a rigorous ethical framework to ensure the responsible development, evaluation, and deployment of multimodal generalists. 
Below, we elaborate on the key ethical considerations.
These ethical measures ensure that \texttt{General-Bench} serves as a responsible and inclusive benchmark, contributing to the sustainable and equitable development of multimodal AI systems.

\paragraph{Privacy and Data Protection.} 
The benchmark and evaluation process ensure strict compliance with privacy regulations. All tasks and datasets used in \texttt{General-Bench} are carefully curated to exclude personally identifiable information (PII). 
To safeguard privacy, any data derived from public sources is anonymized, and sensitive content is filtered out. 
Our procedures align with relevant data protection standards, such as GDPR and CCPA, emphasizing our commitment to ethical research practices.

\paragraph{Data Collection.} 
The dataset for \texttt{General-Bench} is built using publicly available resources or through collaborations with contributors who explicitly consented to their data being included. 
Data collection protocols are designed to prioritize ethical sourcing, ensuring that contributors understand their rights, including the ability to withdraw their data at any time. This ensures transparency and fairness throughout the dataset construction process.

\paragraph{Annotator Compensation.} 
Human annotators play a crucial role in ensuring the high quality of the \texttt{General-Bench} dataset. 
We engage well-trained annotators, including postgraduate students and crowdsourcing professionals, and provide fair compensation for their work. 
Annotators are either volunteered to contribute, or paid based on the estimated time required to complete specific tasks.
All are signed to give their best efforts in data annotation and model implementation to ensure the work quality.

\paragraph{Bias and Fairness.} 
Recognizing the potential biases in AI systems, we take active measures to analyze and mitigate biases related to gender, ethnicity, language, and other sociocultural factors within the dataset and evaluation tasks. 
Diverse and representative data collection practices are employed across multiple modalities and languages. 
While we acknowledge that complete eradication of bias is challenging, we strive to identify and address biases as the benchmark evolves.

\paragraph{Intellectual Property Protection.} 
All datasets and tasks included in \texttt{General-Bench} respect intellectual property rights. Data collected from external sources is fully repurposed and modified, and is used under proper licensing agreements, ensuring compliance with intellectual property laws. 
Open-sourced models are strictly used according to their licenses.
Models evaluated via APIs are handled according to their respective terms of use, and no proprietary content is redistributed without permission.

\paragraph{Misuse Potential.} 
We are aware of the potential risks associated with misuse of multimodal intelligence technologies, such as applications in surveillance or the manipulation of public opinion. To mitigate such risks, we have developed guidelines to encourage ethical use. 
These guidelines emphasize the importance of transparency, accountability, and consent in any application or further development of the technologies evaluated in this work.

\paragraph{Accessibility and Inclusivity.} 
In alignment with our commitment to fostering inclusivity in the AI research community, all code, tasks, and datasets related to \texttt{General-Bench} are openly available. 
This ensures that researchers from diverse backgrounds and varying resource levels can equally contribute to, and benefit from, advancements in multimodal generalist research.

\paragraph{Evaluations and Performance.} 
We clarify that the performances of all models reported in this paper—including both specialists and generalists—are influenced by the specific testing environment. This includes factors such as the size and content of the dataset, as well as the parameters used in the reproduced code.  
As we continuously update the dataset, the evaluation results presented in this paper may differ from those obtained in future versions.  
We emphasize that such differences are considered reasonable and expected deviations, and should not raise any concerns.  
Our leaderboard is open and under active development, and we warmly welcome participation from external practitioners.

\clearpage

\subsection{Author Contribution}

All authors contributed to this project in various capacities, including idea conceptualization, data annotation, model implementation, paper writing, and project supervision. 
To provide a transparent overview, Table~\ref{tab:author-contribution} summarizes the contributions and responsibilities of all co-authors.
Given the extensive scope and workload of this project, we enlisted the help of a large group of contributors. 
Among them, some individuals made contributions but did not qualify for co-authorship due to partial involvement or insufficient or voluntary contributions. 
Nevertheless, we acknowledge their efforts and list them in Table~\ref{tab:unauthor-contribution} to express our gratitude for their support.

{
\fontsize{8.5}{11}\selectfont 
\setlength{\tabcolsep}{1.mm}
\captionof{table}{
Detailed author list and contribution statement.
}
\vspace{-2mm}
\label{tab:author-contribution}
\begin{longtable}{p{0.2cm}p{2cm}p{1.5cm}p{5.5cm}p{3cm}p{3.5cm}}
\hline
\textbf{\#} & \textbf{Name} & \textbf{Group} & \multicolumn{1}{c}{\textbf{Role}} & \textbf{Responsible Datasets} & \textbf{Responsible Models} \\
\hline
\endfirsthead
\hline
\textbf{\#} & \textbf{Name} & \textbf{Group} & \multicolumn{1}{c}{\textbf{Role}} & \textbf{Responsible Datasets} & \textbf{Responsible Models} \\
\hline
\endhead
\hline
\endfoot
\hline
\endlastfoot
1 & Hao Fei & All & 
1) Project general leader: design and implement the idea, including General-Level evaluation and General-Bench planning. \newline
2) Worker for the audio group. \newline
3) Designed data collection methods and selected models for verification. \newline
4) Responsible for all paper writing, illustrations, and polishing. \newline
5) Managed online deployment of data and automated evaluation systems. \newline
6) Maintained the project website. 
7) Provided computing resources.
& All audio-generation datasets & All specialists in audio-generation tasks \newline
Audio MLLMs: WavLLM, ImageBind-LLM, Unified-io-2-XXL, ModaVerse-7b-v0, AudioGPT-GPT4, SpeechGPT-7B-com, LLaMA-Omni\\
\hline
2 & Yuan Zhou & Working for Image group & 
1) Project co-leader:   give the formal text and formula definitions for the 5 levels of the General-Level evaluation framework, along with the corresponding formula derivation. \newline
2) Led the image group, managing tasks and execution. \newline
3) Constructed and polished over 150 datasets; implemented around 30 SoTA specialists and 2 MLLMs. \newline
4) Verified task and data management, and deployed systems. \newline
5) Developed evaluation scripts and automated testing systems. 
& All image-related datasets & Image-oriented MLLMs: GPT4-o, GPT4-o-mini, GPT4-V \\
\hline
3 & Juncheng Li & Working for Image group & 
Project co-leader for image group: Led the image group for supervised dataset collection and image-based MLLMs evaluation.
& All image-related datasets & Specialists and MLLMs supporting image-related skills \\
\hline
4 & Xiangtai Li & Working for Video group & 
Project co-leader for video group: Led the video group for supervised datasets collection and video-based MLLMs evaluation. 
& All video-related datasets & Specialists and MLLMs supporting video-related skills \\
\hline
5 & Qingshan Xu & Working for 3D group & 
Project co-leader for 3D group: Led the 3D group for supervised datasets collection and 3D-based MLLMs evaluation. 
& All 3D-related datasets & Specialists and MLLMs supporting 3D-related skills \\
\hline
6 & Bobo Li & Working for Language group & 
Project co-leader for Language group: Led the Language group for supervised datasets collection and Language-based MLLMs evaluation. 
& Language-related datasets: L-2, 14, 15, 16, 17, 18, 19, 20, 21, 22 & Specialists supporting L-2, 14, 15, 16, 17, 18, 19, 20, 21, 22 skills, and MLLMs including Qwen2.5-7B-Instruct, Baichuan2-7B-Base, Vicuna-7b-V1.5, Falcon3-7B-Instruct, Ministral-8B-Instruct-2410 \\
\hline
7 & Shengqiong Wu & Working for Audio group & 
Project co-leader for the audio group: Led the audio comprehension group for the collection of supervised datasets and the evaluation of audio-based MLLMs. 
& Audio-related datasets: A-C-1, 2, 3, 4, 5, 6, 7, 8, 9 & Specialists supporting A-C-1, 2, 3, 4, 5, 6, 7, 8, 9 skills, and MLLMs including  Qwen-Audio-Chat, Qwen2-Audio-Instruct, Vitron-V1, GAMA, Pengi, WavLLM, SALMONN-7B	SALMONN-13B	SpeechGPT-7B-com, AudioGPT-GPT4, AnyGPT, PandaGPT-13B, ImageBind-LLM, ModaVerse-7b-v0, Unified-io-2-XXL, NExT-GPT-V1.5\\
\hline
8 & Yaoting Wang & Working for Image group & Implement image-related MLLMs for evaluation & \multicolumn{1}{c}{/} & MLLMs including Qwen2-VL-7B, Qwen-VL-Chat, MoE-LLAVA-Phi2-2.7B-4e-384, mPLUG-Owl2-LLaMA2-7b, Phi-3.5-Vision-Instruct, Cambrian-1-8B, MiniGPT4-LLaMA2-7B \\
\hline
9 & Junbao Zhou & Working for 3D group & Collect 3D-related datasets and Implement 3D-related Specialists and MLLMs for evaluation  & 3D-related datasets: D-C-1, 2, 3, 4, 5, 6, 7, 8, 9, 10, 11, 12, 13 &  Specialists supporting  D-C-1, 2, 3, 4, 5, 6, 7, 8, 9, 10, 11, 12, 13 skills, and MLLMs including 3D-LLM-2.1B, PointLLM-7B, PointLLM-13B, 3D-VisTA \\
\hline
10 & Jiahao Meng &  Working for Video group  & Collect video-related datasets and Implement video-related Specialists and MLLMs for evaluation  &  Video-related datasets: V-C-2, 3, 4, 20  & Specialists supporting V-C-2, 3, 4, 20 skills, and MLLMs including Long-LLaVA-9B, DeepSeek-VL-2-small, DeepSeek-VL-2, LLaVA-One-Vision-7B, LLaVA-One-Vision-72B\\
\hline
11 & Qingyu Shi & Working for Video group & Collect video-related datasets and Implement video-related Specialists and MLLMs for evaluation & Video-related datasets: V-G-1, 2, 3, 4 & Specialists supporting V-G-1, 2, 3, 4 tasks, and MLLMs including VidAgent \\
\hline
12 & Zhiyuan Zhou & Working for Image group & Collect image-related datasets and Implement image-related Specialists and MLLMs for evaluation & Image-related datasets: I-C-5, 7, 14, 15, 26, 28, 30, 34 & Specialists supporting I-C-5, 7, 14, 15, 26, 28, 30, 34 skills, and MLLMs including InternVL2\_5-2B, InternVL2\_5-4B, InternVL2\_5-8B, Monkey-10B-chat, DeepSeek-VL-7B-Chat \\
\hline
13 & Liangtao Shi & Working for Image group & Collect image-related datasets and Implement image-related Specialists and MLLMs for evaluation & Image-related datasets: I-C-5, 13, 19, 26, 27, 28 & Specialists supporting I-C-5, 13, 19, 26, 27, 28 skills, and MLLMs including InternVL-Chat-V1-5, Mini‑InternVL‑Chat‑4B‑V1‑5, InternLM-XComposer2-VL-1.8B, GPT4RoI-7B, GLaMM\\
\hline
14 & Minghe Gao & Working for Image group  & Collect image-related datasets and Implement image-related Specialists and MLLMs for evaluation & Image-related datasets: I-C-8, 9, 17, 21, 25, 26, 28, 31, 35 &   Specialists supporting I-C-8, 9, 17, 21, 25, 26, 28, 31, 35 skills, and MLLMs including BLIP2, miniMonkey, DeepSeek-VL-7B-Base, LISA\\
\hline
15 & Daoan Zhang &  Working for 3D group & Collect 3D-related datasets and Implement 3D-related Specialists and MLLMs for evaluation  & 3D-related datasets: D-G-1, 2, 3, 4, 5, 6, 7, 8, 9 & Specialists supporting D-G-1, 2, 3, 4, 5, 6, 7, 8, 9 skills, and MLLMs including MotionGPT-T5, MotionGPT-LLaMA, AvatarGPT, LLaMA-mesh\\
\hline
16 & Zhiqi Ge & Working for Image group & Collect image-related datasets and Implement image-related Specialists and MLLMs for evaluation &  Image-related datasets: I-C-4, 26, 34 & Specialists supporting I-C-4, 26, 34 skills, and MLLMs including Claude-3.5-Sonnet, Claude-3.5-Opus, Emu2-37B, DetGPT\\
\hline
17 & Weiming Wu & Working for Image group &  Implement image-related MLLMs for evaluation  & \multicolumn{1}{c}{/} & MLLMs including Otter, Show-o, NExT-Chat, Yi-vision-v2, Qwen2-VL-72B \\
\hline
18 & Siliang Tang & Working for Image group  & Collect image-related datasets and Implement image-related Specialists and MLLMs for evaluation & Image-related datasets: I-C-26, 34 & Specialists supporting I-C-26, 34 skills, and MLLMs including Claude-3.5-Sonnet, Claude-3.5-Opus, Emu2-37B, DetGPT \\
\hline
19 & Kaihang Pan &  Working for Image group & Collect image-related datasets and Implement image-related Specialists and MLLMs for evaluation & Image-related datasets: I-C-17, 22, 23, 38, 39 & Specialists supporting I-C-17, 22, 23, 38, 39 skills, and MLLMs including Pixtral-12B, SEED-LLaMA-13B, LLaVA-NeXT-13B, LLaVA-NeXT-34B\\
\hline
20 & Yaobo Ye &  Working for Image group &  Collect image-related datasets and Implement image-related Specialists and MLLMs for evaluation & Image-related datasets: I-C-3, 7, 28, I-G-1, 3, 4, 5, 6, 7, 8, 12, 14, 15 & Specialists supporting I-C-3, 7, 28, I-G-1, 3, 4, 5, 6, 7, 8, 12, 14, 15 skills, and MLLMs including BLIP-3 (XGen-MM), CogVLM-Chat, ShareGPT4V-7B, ShareGPT4V-13B \\
\hline
21 & Haobo Yuan & Working for Video group & Collect video-related datasets and Implement video-related Specialists and MLLMs for evaluation & Video-related datasets: V-C-5, 6, 7, 8, 9, 10, 11, 12, 13 & Specialists supporting V-C-5, 6, 7, 8, 9, 10, 11, 12, 13 skills, and MLLMs including InternVL-2-8B, InternVL-2.5-8B, InternVL-2-26B, InternVL-2.5-26B \\
\hline
22 & Tao Zhang &  Working for Video group &  Collect video-related datasets and Implement video-related Specialists and MLLMs for evaluation & Video-related datasets: V-C-14, 15, 16, 17, 18, 19 & Specialists supporting V-C-14, 15, 16, 17, 18, 19 skills, and MLLMs including CoLVA-2B, CoLVA-4B, Sa2VA-8B, Sa2VA-26B\\
\hline
23 & Tianjie Ju & Working for Language group & Collect language-related datasets and Implement language-related Specialists and MLLMs for evaluation  & Language-related datasets: L-1, 2, 3, 4, 5, 6, 7, 8, 9, 10, 11, 12, 13 & Specialists supporting L-1, 2, 3, 4, 5, 6, 7, 8, 9, 10, 11, 12, 13 skills, and MLLMs including Meta-Llama-3.1-8B-Instruct, Gemma-2-9b-it, GPT-J	ChatGLM-6B, InternLM2-Chat-7B, Yi-lightning \\
\hline
24 & Zixiang Meng & Working for Image and Video group & Collect image, video-related datasets and Implement image, video-related Specialists and MLLMs for evaluation   & Image-related datasets: I-C-6, 28, 38, and Video-related datasets: V-C-20 & Specialists supporting I-C-6, 28, 38 and V-C-20 skills, and MLLMs including Gemini-1.5-Pro, Gemini-1.5-Flash, OMG-LLaVA-InternLM20B, Idefics3-8B-Llama3\\
\hline
25 & Shilin Xu &  Working for Video group &  Collect video-related datasets and Implement video-related Specialists and MLLMs for evaluation & Video-related datasets: V-C-1, 2, 4 & Specialists supporting V-C-1, 2, 4 skills, and MLLMs including InternVL-2.5-8B, InternVL-2.5-26B, Qwen2-VL-7B, Qwen2-VL-72B \\
\hline
26 & Liyu Jia & Work for image group &  Collect image-related datasets and Implement image-related Specialists and MLLMs for evaluation  & image-related dataset collection: I-C-6, 17, and & MLLMs including GPT4-o-4096. \\ 
\hline
27 & Wentao Hu & Work for image group &  Collect image-related datasets and Implement image-related Specialists and MLLMs for evaluation   & image-related dataset collection: I-C-17, 28 & MLLMs including gpt-3.5-turbo, chatgpt4-o-latest. \\
\hline
28 & Meng Luo & Working for Video group &  Collect video-related datasets and Implement video-related Specialists and MLLMs for evaluation & Video-related datasets: V-C-2, V-G-1, 5, 6 & Specialists supporting V-C-2, V-G-1, 5, 6 skills, and MLLMs including LM4LV  \\
\hline

\hline
29 &  Jiebo Luo & Discussion \& Advisory & 
Discussed the high-level directions and goals of the project. Provided important and insightful feedback for the overall system design. 
& \multicolumn{1}{c}{/} & \multicolumn{1}{c}{/}
\\
\hline
30 & Tat-Seng Chua & Discussion \& Advisory & 
Discussed the high-level directions and goals of the project. Provided important and insightful feedback for the overall system design. 
& \multicolumn{1}{c}{/} & \multicolumn{1}{c}{/}
\\
\hline
31 & Hanwang Zhang & Project \newline Supervision & 
1) Project co-supervisor, conceptualized the idea of General-Level, and the entire process. \newline
2) Provided computing resources. 
& \multicolumn{1}{c}{/} & \multicolumn{1}{c}{/}
\\
\hline
\hline
32 & Shuicheng Yan & Project \newline Supervision & 
1) Project co-supervisor, co-conceptualized the idea of General-Level, and supervised the entire process. \newline
2) Provided computing resources. 
& \multicolumn{1}{c}{/} & \multicolumn{1}{c}{/}
\\

\hline
\end{longtable}
}
\clearpage

{
\begin{table}[!t]
\centering
\fontsize{8.5}{18}\selectfont 
\setlength{\tabcolsep}{1.5mm}
\caption{
List of some contributors without authorship.
}
\vspace{-2mm}
\label{tab:unauthor-contribution}
\begin{tabular}{p{1cm}p{3cm}p{11cm}}
\hline
\textbf{\#} & \textbf{Name} & \multicolumn{1}{c}{\textbf{Contribution}} \\
\hline

1 & Zhengzhe Liu & 
Contributed to image group: assisted in image-oriented dataset preparation and model testing. \\
\hline
2 & Zhongze Luo & Evaluating Yi-vision-v2 on 77 dataset; evaluating Show-o on 65 datasets. \\ \hline

3 & Chunhan Li & Evaluating Show-o on 136 image comprehension datasets and partial 67 datasets. \\ \hline

4 & Qirui Huang & Evaluating Yi-lightning on 117 NLP datasets and evaluating Show-o on 44 image generation. \\ \hline

5 & Jiaxin Zhu & Assist in evaluating certain MLLMs on certain datasets. \\ \hline

6 & Ming Lei & Evaluating Otter on certain datasets. \\ \hline

7 & Zhangyu Wang & Evaluating Otter on certain datasets. \\ 
\hline
8 & Lin Liu & Contributed to the preparation of image-related task data during phases 1 and 2. \\ \hline
9 & Chengjie Zhou & Contributed to the preparation of NLP task data during phase 1. \\ \hline
10 & Yucheng Han & Contributed to the preparation of image-related task data during phase 1. \\ \hline
11 & Peng Zhou & Contributed to the preparation of image-related task data during phase 1. \\ \hline
12 & Luanyuan Dai & Contributed to the preparation of image-related task data during phase 2. \\ \hline
13 & Yuxuan Liu & Contributed to the preparation of image-related task data during phase 2. \\ \hline
14 & Xun Jiang & Contributed to the preparation of image-related task data during phase 2. \\ \hline
15 & Peisuo Li & Contributed to the preparation of image-related task data during phase 2. \\ \hline
16 & Xu Zhang & Contributed to the preparation of image-related task data during phase 2. \\ \hline
17 & Wenjie Zhuo & Contributed to the preparation of image-related task data during phase 2. \\ \hline
18 & Lianyuan Fan & Contributed to 3D generation-related data collection during phase 2 (incomplete). \\ \hline
\end{tabular}

\end{table}
}

\end{document}